%% file: thesis.tex
\documentclass[twosided]{uocthesis}

\usepackage{subfiles}
\usepackage{pdflscape}
\usepackage{etoolbox}
\usepackage{graphicx}
\usepackage{tipa}
\usepackage{eurosym}
\usepackage{tabto}
\renewcommand{\sectionmark}[1]{%
  \markright{Section \thesection}%
}

\usepackage{enumitem}
\usepackage{siunitx}
\usepackage{booktabs, makecell, multirow}

\usepackage{indentfirst}
\usepackage{tabularx}
\usepackage[flushleft]{threeparttable}
\usepackage[center,tableposition=top,skip=12pt]{caption}
\usepackage{ragged2e}

\newcommand{\tablenoteparagraph}[1]{%
  \vspace{12pt}%
  \footnotesize%
  \doublespacing%
  \noindent%
  \begin{minipage}{\linewidth}%
    \setlength{\parindent}{0pt}%
    \setlength{\parskip}{0pt}%
    \noindent%
    \justifying
    #1%
  \end{minipage}%
}

\usepackage{siunitx}
\usepackage{array}
\newcolumntype{Y}{>{\Centering\arraybackslash}X}

\usepackage{subcaption}
\usepackage{soul}
\usepackage{lineno}

\usepackage{float}
\newfloat{floatquote}{tbp}{loq}
\floatname{floatquote}{Excerpt}

\newfloat{floattables}{tbp}{loq}
\floatname{floattables}{Tables}

\usepackage{mdframed}
\mdfdefinestyle{MyQuoteFrame}{%
linecolor=black,
rightline=false, topline=false, bottomline=false,
outerlinewidth=0.1pt,
innerleftmargin=3em,
backgroundcolor=white}

\usepackage{natbib}

\usepackage{xcolor}
\usepackage{hyperref}
\hypersetup{
  colorlinks=true,
  citecolor=violet,
  linkcolor=violet,
  urlcolor=violet}

\usepackage{tikz}
\usetikzlibrary{arrows.meta,chains,positioning,shapes.geometric}
\usepackage{tipa}

\usepackage{CJKutf8}
\usepackage[T1]{fontenc}
\usepackage[utf8]{inputenc}

\usepackage{amsmath}
\newcommand{\eqname}[1]{\tag*{#1}}% Tag equation with name

\usepackage{silence}
\usepackage[nogroupskip,acronym,toc]{glossaries}

\newglossary[nlg]{dialect}{nlo}{nli}{New Zealand English}
\newglossary[nlg]{maori}{nlo}{nli}{te reo Māori}
\newglossary[nlg]{name}{nlo}{nli}{Place Names and Proper Nouns}
\newglossary[nlg]{analysis}{nlo}{nli}{Miscellaneous Terms}
\newglossary[nlg]{tech}{nlo}{nli}{Technical Abbreviations}
\newglossary[nlg]{misc}{nlo}{nli}{Miscellaneous Abbreviations}

\makenoidxglossaries

\loadglsentries{foo.tex}

    \Title{Language, Place, and Social Media:\\ Geographic Dialect Alignment in New Zealand}
    \Author{Sidney Gig-Jan Wong}
    \Year{2026}

    \Supervisor{Dr Benjamin Adams}
        \Department{Computer Science and Software Engineering, University of Canterbury}
    \Supervisor{Dr Jonathan Dunn}
        \Department{Department of Linguistics, University of Illinois Urbana-Champaign}
    \Supervisor{Dr Kong Meng Liew}
        \Department{School of Psychology, Speech and Hearing, University of Canterbury}
    \Supervisor{Professor Jen Hay}
        \Department{Department of Linguistics, University of Canterbury}

\begin{document}
\begin{CJK}{UTF8}{bsmi}

    \prelimpages
    \titlepage

    \quoteslip

    \abstract{

    Variation and change are fundamental properties of human language, and the emergent patterns observed within these dynamic systems are systematic and meaningful. While the focus of sociolinguists has traditionally rested on spoken language, the advent of computer-mediated communication (CMC) and social media has demonstrated that written language likewise reveals structured patterns of variation and change. As social networks, social media platforms function as a `natural laboratory' to investigate the interplay of social identity and language use.

    While social media dialectology is not a new area of study, there are compelling practical reasons to explore variation and change within the digital sphere. With recent advancements in Natural Language Processing (NLP), there is an increased awareness of how well georeferenced social media language data represents an underlying population. This need is particularly salient for low-resource similar languages, varieties, and dialects where data availability is often limited.
    
    As an intra-disciplinary problem within linguistics, I reconceptualise this issue within computational sociolinguistics, aiming to understand the extent to which place-based social media networks align with the linguistic context of underlying geographic dialect communities. I refer to this phenomenon as geographic dialect alignment. Focusing on the sociolinguistic context of New Zealand and communities on Reddit, my primary research question is: to what extent can we observe geographic dialect alignment in place-informed social media communities? More specifically, do digital communities reflect patterns of language variation and change similar to those observed within and across geographically defined dialect communities?

    Of particular interest to my research is the role of the social construction of space - conceptualised as place - in shaping language variation and change, as well as the perceptions of users on Reddit. Both aspects remain under-explored in social media dialectology. To address this gap, I explore the following secondary research questions: 1) do users in place-based communities associate language-use with a place identity? 2) is there a relationship between geographic dialect communities and place-based communities? 3) do place-related communities form a contiguous speech community?

    In the first phase, I selected a sample of post submissions from r/newzealand - the primary place-based community associated with New Zealand - and identified two selfposts specifically focused on New Zealand English and local language use. For the qualitative analysis, I employed discourse analysis to determine the situated meanings within these posts. The objective of this analysis was to reconnect the producers of language to their discourse, humanizing the data before quantification. Subsequently, I applied thematic analysis to the associated comment threads to curate a user-informed inventory consisting of 51 lexical, 3 morphosyntactic, and 13 semantic features.

    In the second phase, I analysed the distribution of these user-informed lexical and morphosyntactic features across six country-level place-based communities on Reddit to evaluate the accuracy of user intuition. The findings indicated that while user intuitions were largely incongruent with the data, the distribution of these features remained systematic and meaningful across country-level communities. Moreover, non-linguistic user behaviour - specifically temporal engagement patterns - emerged as a significant indicator for identifying non-local users, whose presence often correlated with an increase in innovative variants.

    In the third phase, I explored alternative computational approaches for detecting language variation across place-based communities on Reddit. Consistent with existing literature, traditional text classification methods proved ineffective for identifying latent linguistic variation at both the country and city levels. However, advanced language modelling techniques - specifically Word2Vec embeddings - facilitated the detection of variation across the user-informed semantic variables. By comparing word vector representations trained on discrete place-based communities using cosine similarity, I was able to quantify the degree of semantic shift and geographic alignment across the digital landscape.
    
    In the fourth and final phase, I expanded the corpus to include a broader network of New Zealand-related communities. By identifying user informed recommendations from r/NZMetaHub, I incorporated an additional 32 subreddits from the Pushshift Repository. Utilising Computational Construction Grammar, I confirmed that these communities maintain a high degree of grammatical similarity. I then examined diachronic semantic shift within the 13 user-informed semantic variables. Although only three variables exhibited the expected shifts, the results for `chippy' (transitioning from a `potato chip' to the nickname of a former Prime Minister) and `snapper' (shifting from a 'transport card' back to the `fish species') suggest that the diachronic embedding models successfully captured semantic changes unique to the New Zealand sociolinguistic context.

    Based on my analysis of user-informed lexical, morphosyntactic, and semantic variables, the findings suggest that geographic dialect alignment is observable within place-informed social media communities for New Zealand-related subreddits. Regarding the secondary research questions, I found that users in place-based communities generally associated specific language use with a distinct place-identity, and that these digital communities tended to form a contiguous speech community. However, the relationship between established geographic dialect communities and their digital counterparts was not straightforward when assessed through user-informed variables, indicating a complex layering of traditional regionalisms and emergent digital norms.

    Some of the limitations of this study stemmed from the reliance on user-informed variables, which inherently shaped the direction and scope of the analysis. Additional constraints included data sparsity within specific regional subsets and potential model bias introduced during the analytical pipeline. A significant theoretical limitation was the restricted engagement with traditional sociolinguistic frameworks; this reflects a broader historical emphasis on spoken language within the field, which complicates the direct application of established theories to computer-mediated data. To mitigate this, I prioritised methodological rigour and the development of a robust computational pipeline, with the objective of bridging this theoretical gap in future research.

    The thesis makes several distinct contributions to the field by introducing place-informed social media dialectology and implementing advanced language modelling techniques. By integrating user perceptions to evaluate the degree of language variation and change, this work addresses critical gaps in the existing literature regarding digital vernaculars. Furthermore, this research produced a comprehensive corpus of New Zealand-related Reddit communities - comprising 4.26 billion unprocessed words - providing a substantial and valuable resource for future sociolinguistic and computational inquiry.

    In terms of future research, there is significant potential to utilise state-of-the-art transformer-based large language models (LLMs) to examine semantic shift through contextual embeddings, though such approaches remain computationally resource-intensive. There is also a critical opportunity to develop a dedicated NLP benchmark for New Zealand English to improve model performance on local varieties. Further directions include extending this methodology to additional platforms - such as Twitter/X - and expanding into multimodal analysis (integrating spoken data). Finally, perceptual dialectology on social media represents a promising avenue for understanding metalinguistic awareness, particularly as researcher access to platform data continues to evolve.
    
    }

    \public{

    This thesis investigates geographic dialect alignment in place-informed social media communities, focussing on New Zealand-related Reddit communities. By integrating qualitative analyses of user perceptions with computational methods, the study examines how language use reflects place identity and patterns of language variation and change based on user-informed lexical, morphosyntactic, and semantic variables. The findings show that users generally associate language with place, and place-related communities form a contiguous speech community, though alignment between geographic dialect communities and place-related communities remains complex. Advanced language modelling, including static and diachronic Word2Vec language embeddings, revealed semantic variation across place-based communities and meaningful semantic shifts within New Zealand English. The research involved the creation of a corpus containing 4.26 billion unprocessed words, which offers a valuable resource for future study. Overall, the results highlight the potential of social media as a natural laboratory for sociolinguistic inquiry.
    
    }

    \tableofcontents
    \listoffigures
    \listoftables

    \acknowledgments{

    If I had to properly acknowledge everyone who supported me through this academic journey, it would probably be as long as this thesis. First and foremost, I want to thank Toi Hangarau Geospatial Research Institute (GRI) and the University of Canterbury | Te Whare Wānanga o Waitaha (UC) for giving me this opportunity to pursue this PhD in Linguistics. I also extend my gratitude to everyone at the Department of Linguistics and the New Zealand Institute of Language, Brain and Behaviour who have supported me at every stage of my academic career.

    I also want to thank Fulbright New Zealand Te Tūāpapa Mātauranga o Aotearoa me Amerika for providing me the opportunity to learn and grow as a researcher at the University of Illinois Urbana-Champaign (UIUC). I would like to thank the Department of Linguistics at UIUC for hosting me during that time.
    
    I am deeply grateful to Associate Professor Benjamin Adams for his primary supervision and for shaping me into an independent thinker. I am grateful to Associate Professor Jonathan Dunn for introducing me to the world of computational sociolinguistics. Thank you both for supporting me throughout this journey. I also want to thank Professor Jennifer Hay for your unwavering support and to Dr. Kong Meng Liew for showing me different ways to explore data.

    I would like to thank the examiners, Associate Professor Hunter Hatfield (University of Otago | Ōtākou Whakaihu Waka) and Professor Benedikt Szmrecsanyi (KU Leuven), for offering their time and expertise in examining my thesis. I would also like to thank the examination chair, Associate Professor Lynn Clark (UC), for coordinating the examination of my thesis.

    As cross-disciplinary researcher, I had the pleasure of being part of not one, but two, PhD cohorts. In the GRI, I want to thank my officemates Dr. Martin Nguyen, Andrea Pozo Estivariz , Clevon Ash, Katherine Booker, David Pedley, and Sunil Tamang. In linguistics, my thanks go to Dr. Gia Hurring, Matthew Durward, and Peiman Pishyar-Dehkordi. To Luke Parkinson, thank you for your continued support. I count your friendship as one of the very best outcomes of this PhD journey.
    
    My time at the University of Illinois at Urbana-Champaign was equally special. I want to thank Amy Atiles and Ryan Corrigan for hosting me in the printing room and to Tai Armstrong, Jonathan Soelle, and Martine Gallardo for letting me win Magic: The Gathering - once in a while.
    
    I want to thank Dr. Mackay Price, Dr. Patrick Hancock, and Dr. Amit Barde for being my proofreaders and validators. And to Dr. Taylor Winter, who knows exactly which buttons for me to do what I need to do. Special thanks to Marilyn and Sin\'{e}ad for taking care of me during this time, and to Adam, Ian, Ross, Toby, and Felix, who have had to deal with me over the years.

    I am forever grateful to my mum and dad for their love and guidance. And to Dr. Barry Wong, my older brother and the first Dr. Wong, who has been there for me since I can remember. We did it! 
    
    Finally, to Jake for standing beside me through it all. This thesis is as much a testament to your patience and love as it is to my research. I dedicate this thesis to you.
    
    }

    \notes{
    
    \paragraph*{New Zealand}
    
    In this thesis, I use \textit{New Zealand} in reference to the geopolitical entity that occupies the whenua of tāngata whenua. This whenua is known by many names, such as Aotearoa and Te Waipounamu. I follow the guidance of Ranginui Walker (Whakatōhea), who suggested that \textit{New Zealand} might be useful as a shorthand device for the settler-colonial entity \citep{ballantyne_place_2011}. However, it does not simply equate to both the land and the people associated with it. Toitū te Tiriti.
    
    \paragraph*{New Zealand English and te reo Māori}
    
    \acrfull{NZE} and borrowings from te reo Māori are used frequently throughout this thesis. Where appropriate, definitions of unfamiliar terms from \acrshort{NZE} and te reo Māori are provided in a separate glossary (including a list of proper nouns and place names) located after the References. These definitions can be accessed via embedded hyperlinks throughout the text.
    
    \paragraph*{Ethical Considerations}
    
    All ethical considerations in this thesis were made with reference to the guidelines for social media research outlined by the Human Research Ethics Committee (HREC). Data reproduced within this thesis does not include any personally identifiable information. As no identifiable or depersonalised information has been included in the publication of this thesis, a formal application with the HREC was not required, in accordance with the committee's protocols for non-human participant research involving public datasets.

    \paragraph*{AI Use Declaration}

    I acknowledge the use of Google Gemini (\href{https://gemini.google.com/}{https://gemini.google.com/}) to proofread and refine the post-submission draft of this thesis for clarity, grammar, and consistency with \acrshort{NZE} orthography. The resulting output was subsequently modified to better represent my own style and authorial tone.
    
    }

    \dedication{for Jake}

    \vita{

    \sloppy

    \subsubsection*{Education}

    \begin{itemize}[nolistsep]
        \item \textbf{Doctor of Philosophy} (2026), Linguistics - Geospatial Research Institute, University of Canterbury. Supervisors: Associate Professor Benjamin Adams, Associate Professor Jonathan Dunn, Professor Jennifer Hay, and Dr. Kong Meng Liew
        \item \textbf{Master of Applied Data Science} (with Distinction), University of Canterbury
        \item \textbf{Master of Linguistics} (with Distinction), University of Canterbury
        \item \textbf{Bachelor of Science} (Linguistics), University of Canterbury
    \end{itemize}

    \subsubsection*{Research Output}
    
    \paragraph*{Peer-Reviewed Publications}
    
    \begin{itemize}[nolistsep]
        \item Wong, S., Adams, B., and Dunn, J. (in press). Detecting Linguistic Diversity on Social Media. \textit{Cartography and GIScience in Australasia and Oceania: Including twenty years of GeoCart} (Ed: A. Moore.).
        \item Dunn, J., \& Wong, S. (2025). Language Contact and Population Contact as Sources of Dialect Similarity. \textit{Languages}, 10(8), 188. https://doi.org/10.3390/languages10080188
        \item Dunn, J., \& Wong, S. (2022). Stability of Syntactic Dialect Classification Over Space and Time. \textit{Proceedings in the 29th International Conference on Computational Linguistics}, October 12-17 2022, Gyeongju, Republic of Korea. https://aclanthology.org/2022.coling-1.3
        \item Wong, S; Dunn, J, \& Adams, B. (2022). Comparing Measures of Linguistic Diversity Across Social Media Language Data and Census Data at Subnational Geographic Areas. \textit{Proceedings in New Zealand Geospatial Research Conference}, August 29-30 2022, Massey University, Wellington, New Zealand. https://doi.org/10.6084/m9.figshare.20686324.v1
    \end{itemize}

    \paragraph*{Conference Presentations}

    \begin{itemize}[nolistsep]
        \item Wong, S. (2025, August). Constructing a Sense of Place in Social Media. Abstract presented at the Thirteenth International Conference on Geographic Information Science (GIScience 2025), Aug 26-29, Christchurch, New Zealand. https://doi.org/10.5281/zenodo.16777132
        \item Wong, S. G.-J. (2024, November). The Interaction of Space, Place, and Linguistic Variation in Social Media. Poster presented at the 52 New Ways of Analyzing Variation (NWAV), Nov 7-9, Miami Beach. https://doi.org/10.6084/m9.figshare.27642444.v2
        \item Wong, S. G.-J. (2024, October). End of a Golden Era: The Uncertain Future of Social Media Research. Poster presented at the Academic Data Science Alliance (ADSA) Annual Meeting, Oct 29-31, University of Michigan Ann Arbor. https://doi.org/10.6084/m9.figshare.27365991.v1
        \item Wong, S. G.-J. (2024, June). Geographic Dialect Bias in Multiclass Classification Models. Paper presented at the Sociolinguistics Symposium 25 (SS25), Jun 24-27, Curtin University Perth. https://doi.org/10.6084/m9.figshare.26087992.v1
        \item Wong, S. G.-J. (2023, November). Topic stability of New Zealand English on social media. Paper presented at the Arts, Law, Psychology and Social Science Postgraduate Conference, University of Waikato.
        \item Wong, S. G.-J. (2023, February). Developing a global model of emerging non-geographic digital dialects. Paper presented at the Sprachwissenschaftliche Tagung für Promotionsstudierende, Feb 24-26, Augsburg/Budapest/Vienna.
        \item Wong, S. G.-J. (2022, December). Navigating Aotearoa New Zealand’s Multilingual Social Media Landscape. Paper presented at the Linguistic Society of New Zealand Conference, Dec 8-9, University of Otago. https://doi.org/10.6084/m9.figshare.24053823.v2
    \end{itemize}

    \paragraph*{Invited Talks and Seminars}

    \begin{itemize}[nolistsep]
        \item Wong, S. (2025, July). Locating a Sense of Place on Social Media: A focus on Aotearoa New Zealand. Presentation for the Deputy Vice Chancellor R\&I Visit to the Geospatial Research Institute, Jul 23, University of Canterbury. https://doi.org/10.6084/m9.figshare.29857097.v1
        \item Wong, S. (2025, April). Social Media Sociolinguistics: Theory and Application. Invited speaker at League of Linguists Undergraduate Linguistics Student Organisation, April 3, University of Illinois Urbana-Champaign. https://doi.org/10.6084/m9.figshare.28727204.v1
        \item Wong, S. (2025, March). A Sense of Place in Digital Spaces: The Role of Geography in Social Media Dialectology. Invited speaker at the Virginia Tech Language Sciences (VTLx) Speaker Series, Mar 19, Virginia Polytechnic Institute and State University. https://doi.org/10.6084/m9.figshare.28639592.v1
        \item Wong, S. (2025, February). Beyond the Variable: Application of NLP in Sociolinguistics Research. Invited speaker at the Computational Psycholinguistics of Listening and Speaking (CPLS) Lab Research Meeting, Feb 10, University of California Santa Barbara. https://doi.org/10.6084/m9.figshare.28387244.v1
        \item Wong, S. G.-J. (2022, November). Using social media language data to understand linguistic diversity. Invited talk for the Association of Local Government Information Managers, Nov 23, Geospatial Research Institute.
        \item Wong, S. G.-J. (2022, November). \#linguistics: exploring variation and change using social media language data. Seminar presented at the Linguistics Seminar, Nov 2, University of Waikato.
        \item Wong, S. G.-J. (2022, June). Geospatial Linguistics: Connecting Language to the Land. Seminar presented at New Zealand Institute of Language, Brain and Behaviour Seminar, Jun 30, University of Canterbury.
    \end{itemize}

    \subsubsection*{Teaching}

    \begin{itemize}[nolistsep]
        \item Tutor - Linguistics, 172.232 Language and Society in New Zealand. Te Kunenga ki Pūrehuroa | Massey University
        \item Teaching Fellow (Course Convener), LINGS303-23B Sociolinguistics. Te Whare Wānanga o Waikato | University of Waikato
        \item Teaching Fellow (Lecturer), LINGS301-22B Research Apprenticeship. Te Whare Wānanga o Waikato | University of Waikato
    \end{itemize}

    \subsubsection*{Scholarships, Grants, and Fellowships}

    \begin{itemize}[nolistsep]
        \item Postdoctoral Fellowship in Complex Systems, Te Pūnaha Matatini $|$ New Zealand Centre of Research Excellence for Complex Systems: postdoctoral fellow in the Modelling for Impact Hub to model an entire country as a complex system.
        \item Fulbright Science and Innovation Graduate Award, Fulbright New Zealand $|$ Te Tūāpapa Mātauranga o Aotearoa me Amerika: to conduct research on automatic hate speech detection on social media at the University of Illinois Urbana-Champaign
        \item Ngā Kōrero Tuku Iho, New Zealand Oral History Grant, Manatū Taonga $|$ Ministry for Culture and Heritage: grant received in conjunction with Chelsea Wong She and Eda Tang to record oral histories in the project \textit{Cantonese Heritage and Culture in Aotearoa New Zealand}
        \item Research Grant, Pegasus Health: for research expenses
        \item Travel Grant, European Cooperation in Science and Technology: to attend the second COST funded Action CA19103 LGBTI+ Social and Economic (in)equalities Summer School at Venice International University
        \item Travel Grant, Department of Language Science, University of California Irvine:  to attend the Summer School on Computational Cognitive Modelling for Language at the University of California Irvine
        \item Geospatial Research Institute Toi Hangarau PhD Scholarship, Toi Hangarau | Geospatial Research Institute for tuition fees and research expenses
        \item University of Canterbury Doctoral Scholarship, Te Whare Wānanga o Waitaha | University of Canterbury
    \end{itemize}
     
    }

\textpages
\mainmatter
    
% -----------------------------
% Chapter 1: Introduction
% -----------------------------
    
\chapter{Introduction}
\markboth{Introduction}{}
\label{chap:introduction}

\section{Chapter Outline}

    The purpose of this chapter is to contextualise the research and establish the framework for the study. Following a brief introduction in Section \ref{intro:introduction}, Section \ref{intro:research_context} introduces the overarching thesis aim and specific objectives. I then define the primary and secondary research questions in Section \ref{intro:research_questions}. The subsequent sections detail the four research phases (Section \ref{intro:research_phase}) and provide a comprehensive outline of the thesis structure (Section \ref{intro:outline}).

\section{Introduction}
\label{intro:introduction}

    The rapid expansion of social media since the 1990s \citep{edosomwan_history_2011} has resulted in an unprecedented explosion of linguistic data. On Reddit alone, users contribute over 10 million comments and submissions each month \citep{amaya_new_2021}, providing a vast digital record of contemporary human interaction. This volume of information has not only enabled the development of novel language modelling techniques \citep{radford_language_2019} but has also provided a unique laboratory for investigating patterns of linguistic variation and change. However, as these digital corpora grow, so too does the tension between digital space and physical place. We must critically evaluate how closely digital language use aligns with the physical geographic communities it purportedly represents.

    While social media dialectology is now a well-established field \citep{calude_linguistics_2023}, there are compelling theoretical and practical reasons to look beyond mere data volume. Advancements in \acrfull{NLP} \citep{joshi_natural_2025} have heightened concerns regarding the accuracy with which georeferenced data reflects underlying populations \citep{johnson_geography_2016}. These considerations are particularly vital for non-dominant varieties and dialects, such as \acrshort{NZE}, where traditional sociolinguistic data is often scarce \citep{zampieri_natural_2020}.

    This research interrogates the treatment of place and the role of user perception in language variation and change. Both areas represent significant gaps in the current literature \citep{nguyen_dialect_2021}. Rather than treating social media as a mere coordinate of absolute spatial data, I argue that digital communities are social constructions that perform identity through language. To address this, I reconceptualise the relationship between digital production and physical identity as a question of \textit{geographic dialect alignment}: the degree to which a digital language sample accurately reflects the linguistic and social realities of its associated population.

\section{Aims and Objectives}
\label{intro:research_context}

    Before outlining the primary contributions of this thesis, I first define the sites of interest. Geographically, my focus is on the sociolinguistic context of New Zealand; digitally, the study centres on Reddit as the primary source of social media language data. The aim of this research is to examine patterns of geographic dialect alignment between New Zealand - as a distinct geographic dialect community - and the language-use within place-informed communities on Reddit. This thesis presents the first study to integrate methods from \acrfull{NLP} and perceptual dialectology to investigate language variation and change in \acrshort{NZE}. More specifically, it addresses the research gap regarding the role of user perception in language variation and change on social media platforms \citep{nguyen_dialect_2021}. I argue that user perceptions complement data-driven approaches in \acrshort{NLP} and computational sociolinguistics \citep{nguyen_computational_2016}. A secondary, more practical contribution is the development of an integrated multi-method pipeline combining sociolinguistics and computational linguistics to explore data quality and availability in non-dominant varieties such as \acrshort{NZE}. Finally, this research contributes to the broader literature on \acrshort{NZE}, which has historically focussed on spoken language.

\subsubsection{Place-informed Social Media Dialectology}

    Studies have demonstrated geographic dialect alignment between physical communities and X, formerly known as Twitter\footnote{Throughout the thesis, I have stylised the name of this platform as Twitter\textsuperscript{X} to account for this name change.} (\citealp{grieve_mapping_2019}; \citealp{dunn_global_2019}), there is little evidence of similar alignment in place-based social media platforms like Reddit, with the notable exception of \citet{hamre_geographic_2024}. This research extends the existing body of spatially informed literature beyond Twitter\textsuperscript{X} to investigate how these geolinguistic relationships are preserved across geographic and digital contexts. By shifting the focus from absolute space to socially constructed space (or place), I engage with linguistic literature regarding enregisterment \citep{johnstone_pittsburghese_2009}, place attachment (\citealp{reed_place_2020}; \citealp{carmichael_locating_2023}), and place identity \citep{ilbury_tale_2022} as they relate to language-use on social media.

\subsubsection{Perceptual Dialectology and NLP}

    Interactional context is often missing from corpus-assisted research \citep{seaver_nice_2015}. Similarly, the role of user perception in social media dialectology remains under-explored, with only a handful of existing studies. This gap is partly due to the fact user perceptions are frequently undervalued in empirical research, as the views of lay-people do not always mirror actual usage \citep{preston_language_2010}. A key point of difference in this research is the integration of both interactional context and user perception to evaluate language variation and change across place-based Reddit communities. By doing so, I maintain the connection between the producer of language and their discourse \citep{matheson_discourse_2023}, allowing me to establish a link between user attitudes, ideologies, and language-use. To achieve this, I adopt a passive data collection approach \citep{rocha-silva_passive_2024}. Unlike active collection methods (such as distributing surveys), passive collection involves analysing social media posts within their natural context with minimal intervention. I implement this through thematic analysis to derive a set of user-informed variables directly from the discourse \citep{braun_using_2006}.

\subsubsection{Sociolinguistics in New Zealand}

    While \acrshort{NZE} is one of the more extensively documented varieties of English \citep{gordon_new_2004}, research has historically focussed on spoken language. Consequently, linguistic studies exploring \acrshort{NZE} within social media contexts remain scarce and are largely restricted to Twitter\textsuperscript{X} (\citealp{trye_maori_2019}; \citealp{trye_hybrid_2020}; and \citealp{calude_arehashtagswords_2024}). This thesis addresses this gap by investigating the social and linguistic landscape of New Zealand Reddit. Specifically, my findings examine how written \acrshort{NZE} manifests within this digital context and how it aligns with established linguistic patterns.

\subsection{Broader Applications}

    Geographic dialect alignment can also be conceptualised through the components of an ideal language sample intended to represent an underlying population \citep{dourish_re-space-ing_2006}. This is a long-standing inquiry in linguistics; early dialectologists relied on the speech of \acrfullpl{NORM} speakers, who were viewed as ideal representatives of a speech community due to their perceived resistance to language change \citep{chambers_dialectology_1998}. However, the `ideal' speaker may not provide the ideal sample for developing language technologies if their speech does not reflect contemporary language-in-use \citep{cabitza_toward_2023}. Indeed, users of these technologies may prefer context-specific dialect input that reflects the diverse ways variation manifests in the real-world \citep{blaschke_what_2024}. While evidence suggests broad geographic dialect alignment between spatial language data as a source of ground-truth linguistic behaviour of populations (\citealp{dunn_measuring_2020}; \citealp{dunn_language_2025}), there is limited evidence to suggest that similar alignment exists within implicit or place-based (placial) language data \citep{nguyen_dialect_2021}.

\section{Research Questions}
\label{intro:research_questions}

    Focussing on geographic dialect alignment between New Zealand and Reddit, my primary research question asks (to what extent): 
    
    \begin{quote}
        Can we observe geographic dialect alignment in place-informed social media communities?
    \end{quote}

    Specifically, I aim to determine whether the language used within New Zealand-related Reddit communities aligns with the linguistic patterns speakers expect to encounter in New Zealand. I predict that geographic dialect alignment will be detectable in these communities and that this alignment will be uniquely characteristic of \acrshort{NZE}. Consequently, the null hypotheses are twofold: first, that no relationship exists between language use and place-based community membership; and second, that such language use bears no specific connection to New Zealand. The premise for the New Zealand context is straightforward: as \acrshort{NZE} is the dominant variety in the region, Reddit communities associated with New Zealand should exhibit a higher frequency of its distinct linguistic features. Conversely, features not associated with \acrshort{NZE} should appear significantly less frequently within these digital spaces.

\subsection{Secondary Research Questions}

    Under the primary research question rest three complementary secondary questions to guide my research. These are:

    \vspace{12pt}

    \begin{itemize}[nolistsep]
        \item[SQ1] Do users in place-based communities associate language-use with a place identity?
        \item[SQ2] is there a relationship between geographic dialect communities and place-based communities?
        \item[SQ3] Do place-related communities form a contiguous speech community?
    \end{itemize}

\subsubsection{Secondary Research Question 1}

    The purpose of \acrfull{SQ1} is to determine whether users within place-related communities associate their place identity with specific linguistic forms. While previous studies often treated social media as a mere digital extension of the \textit{real-world} (\citealp{eisenstein_diffusion_2014}; \citealp{grieve_mapping_2019}), I conceptualise place-based communities on Reddit as `places' in their own right. \acrfull{SQ1} aims to establish a link between perception and production by examining user attitudes and ideologies towards \acrshort{NZE} within the specific context of \texttt{r/newzealand}. Furthermore, I investigate whether users are motivated to employ language in a manner that is - or is not - congruent with their perceived place identity. The null hypothesis for this question is that users within these communities do not associate specific linguistic practices with their place identity.

\subsubsection{Secondary Research Question 2}

    Whereas \acrshort{SQ1} focussed on language perception, \acrfull{SQ2} examines linguistic production within place-based communities. From a linguistic perspective, variation is both predictable and meaningful, often serving to index specific social identities \citep{weinreich_empirical_1968}. If geographic dialect alignment is present in place-based communities on Reddit, it should be reflected in the manifestation of linguistic variation. To address \acrshort{SQ2}, I analyse both discrete, user-informed sociolinguistic variables and  high-dimensional representations of language, such as language embedding models \citep{mikolov_efficient_2013}. The null hypothesis for this question is that there is no significant association between linguistic practice and place-based community membership.

\subsubsection{Secondary Research Question 3}

    \acrfull{SQ3} assumes that place-related communities on Reddit exhibit coherence, forming a contiguous speech community rather than a random collection of users and groups lacking shared norms. While place-based communities (such as \texttt{r/newzealand}) are implicitly georeferenced by their name, this is not necessarily true for place-related communities (such as \texttt{r/WinstonJerk}). This approach is grounded in the inclusive definition of a speech community proposed by \citet[p.362]{gumperz_introduction_1996} who argues ``speech communities, broadly conceived, can be regarded as collectivities of social networks''. The null hypothesis for this question is that place-related communities do not form a contiguous speech community.

\section{Research Phases}
\label{intro:research_phase}

    The thesis is organised into four research phases that broadly correspond to the three secondary research questions. The findings from each phase collectively address the primary research question: to what extent is geographic dialect alignment observable in place-informed social media communities? In brief, the analytical pipeline involves situating language use within its social context, establishing a list of evaluative features, and applying language modelling techniques. A synopsis of each phase is provided in the following sections.

\paragraph{Phase 1: User Intuition and Place Identity}

    The purpose of Phase 1 is to understand language-use within its social context. I adopt an inductive approach to social dialectology by establishing the relationship between users within \texttt{r/newzealand} - the primary place-based community associated with New Zealand - and their language-use. Specifically, I employ discourse analysis \citep{gee_introduction_2005} and thematic analysis \citep{braun_using_2006} to develop a list of user-informed features associated with \acrshort{NZE}. A key benefit of this approach is that user perception plays a primary role defining the parameters of the variety. Furthermore, by applying discourse analysis, I develop insights into the way users interact within place-based communities to better understand their perceptions of language-use. In the absence of established benchmarks for evaluating \acrshort{NZE}, a secondary purpose of this phase is to derive a robust set of features characteristics of the variety.
    
\paragraph{Phase 2: User-Informed Sociolinguistic Variables}

    This phase builds on the initial findings by examining the distribution of the identified user-informed sociolinguistic variables. I analyse these distributions across several country-level communities, including \texttt{r/canada}, \texttt{r/usa}, \texttt{r/ireland}, \texttt{r/unitedkingdom}, and \texttt{r/australia}. Given that the majority of research into linguistic variation on social media adopts a variable-based approach, the findings from this phase address the existing research gap between dialectology on Twitter and Reddit (notably extending the work of \citealp{hamre_geographic_2024}).

\paragraph{Phase 3: Dialect Modelling and Language Embeddings}

    Having developed an understanding of language-use within place-based communities and explored avenues for evaluating geographic dialect alignment in \acrshort{NZE}, I then introduce \acrshort{NLP} dialect modelling techniques - such as text classification and language embedding models - to study linguistic variation in a high-dimensional space \citep{dunn_stability_2022}. I extend this analysis beyond \texttt{r/newzealand} and inner-circle country-level communities (as defined by \citealp{kachru_standards_1985}) to include outer-circle contexts (such as \texttt{r/Kenya}, \texttt{r/southafrica}, \texttt{r/india}, \texttt{r/pakistan}, \texttt{r/malaysia}, and \texttt{r/Philippines}) as well as New Zealand city-level communities (including \texttt{r/auckland}, \texttt{r/Tauranga}, \texttt{r/thetron}, \texttt{r/Wellington}, \texttt{r/chch}, and \texttt{r/dunedin}).

\paragraph{Phase 4: Social Networks and Diachronic Embeddings}

    The final phase addresses two critical issues in social media dialect research: data availability and language change \citep{nguyen_dialect_2021}. I first examine \texttt{r/NZMetaHub} to curate a comprehensive list of New Zealand-related Reddit communities. I then interrogate the suitability of this expanded corpus using \acrshort{C2xG} to determine grammatical similarity across these spaces \citep{dunn_computational_2017}. The primary purpose of this analysis is to define New Zealand-related communities a contiguous speech community - or more simply, \textit{a network of networks} \citep{hymes_scope_2020}. To do so, I examine the grammatical relationships between communities in conjunction with user-derived behavioural measures. Building on the work of  \citet{dunn_mapping_2019}, \citet{dunn_stability_2022}, and \citet{dunn_language_2025}, I then leverage this established New Zealand Reddit data set to train diachronic embedding models \citep{kutuzov_diachronic_2018}, allowing me to evaluate semantic shift within the previously identified user-informed variables.

\section{System Requirements}
\label{intro:system_requirements}

    All analyses were conducted using Python 3 \citep{van_rossum_interactively_1991, python_software_foundation_python_2025} within the Google Colab environment \citep{google_google_2025}, a hosted Jupyter Notebook service \citep{perez_ipython_2007} providing access to high-performance computing resources, including GPUs and TPUs. I employed the \texttt{pandas} \citep{mckinney_data_2010, the_pandas_development_team_pandas-devpandas_2025} and \texttt{NumPy} \citep{harris_array_2020} libraries for data manipulation and analysis. Data visualisations were generated using \texttt{Matplotlib} \citep{hunter_matplotlib_2007} and the statistical plotting library \texttt{seaborn} \citep{waskom_seaborn_2021}.

\section{Data Availability}
\label{intro:data_availability}

    The code for this research is available via the Open Science Foundation repository: \href{https://osf.io/bw7eh/}{https://osf.io/bw7eh/}. Additionally, the datasets and models developed for this thesis are hosted across several HuggingFace repositories:

    \begin{itemize}[nolistsep]
        \item The Reddit Corpus of Global Language Use (RCGLU), utilised in Chapter \ref{chap:user_variables} User Intuitions and Place Identity and Chapter \ref{chap:dialect_classification} Dialect Modelling and Language Embeddings: \href{https://doi.org/10.57967/hf/8197}{https://doi.org/10.57967/hf/8197}.
        \item The New Zealand Reddit (NZR) Corpus, utilised in Chapter \ref{chap:dialect_classification} Dialect Modelling and Language Embeddings and Chapter \ref{chap:construction_grammar} Social Networks and Diachronic Embeddings: \href{https://doi.org/10.57967/hf/8196}{https://doi.org/10.57967/hf/8196}.
        \item The Reddit Word2Vec models trained for Chapter \ref{chap:dialect_classification} Dialect Modelling and Language Embeddings: \href{https://doi.org/10.57967/hf/8198}{https://doi.org/10.57967/hf/8198}.
        \item The diachronic Reddit Word2Vec models trained for Chapter \ref{chap:construction_grammar} Social Networks and Diachronic Embeddings: \href{https://doi.org/10.57967/hf/8200}{https://doi.org/10.57967/hf/8200}.
        \item The New Zealand Reddit Computational Construction Grammar model trained for Chapter \ref{chap:construction_grammar} Social Networks and Diachronic Embeddings: \href{https://doi.org/10.57967/hf/8199}{https://doi.org/10.57967/hf/8199}.
    \end{itemize}

\section{Outline}
\label{intro:outline}

    The thesis is organised into three main parts: a review of the literature and corpus dimensions (Chapters \ref{chap:literature} and \ref{chap:corpus_dimensions}), the core studies (Chapters \ref{chap:user_intuitions}, \ref{chap:user_variables}, \ref{chap:dialect_classification}, and \ref{chap:construction_grammar}), and the conclusion (Chapter \ref{chap:conclusion}). Each study chapter corresponds to a specific research phase and includes its own motivation, methodology, results, and discussion. While these chapters are self-contained, they are designed to build upon one another sequentially. A synopsis of each is provided below.
    
    In \textbf{Chapter \ref{chap:literature} Literature Review}, I review relevant theoretical and methodological approaches to analysing language variation and change in \acrfull{CMC}. I discuss geographic and linguistic perspectives on \textit{place} and how these contribute to the conceptual framework of geographic dialect alignment. Finally, I contextualise this research by introducing the sociolinguistic context of New Zealand and the distinct features of \acrshort{NZE}.
    
    In \textbf{Chapter \ref{chap:corpus_dimensions} Corpus Dimensions}, I describe the linguistic landscape of Reddit. While this chapter does not directly address the research questions, it is integral to understand Reddit as a specific register of language. I employ the Situational Characteristics of Registers and Genres framework to guide this analysis.
    
    In \textbf{Chapter \ref{chap:user_intuitions} User Intuitions and Place Identity}, I consider how users in the place-based community \texttt{r/newzealand} index their place identity through language use. This chapter addresses \acrshort{SQ1} by examining whether users associate specific linguistic features with a New Zealand identity.
    
    In \textbf{Chapter \ref{chap:user_variables} User-Informed Sociolinguistic Variables}, I examine the distribution of these user-informed features across various country-level communities. This chapter addresses \acrshort{SQ2} by testing the relationship between physical geographic dialect communities and digital place-based communities.
    
    In \textbf{Chapter \ref{chap:dialect_classification} Dialect Modelling and Language Embeddings}, I use text classification models to illustrate geographic sampling bias and the impact of sample size in high-dimensional language modelling. Building on these insights, I develop embedding models to evaluate user-informed semantic variables, further addressing \acrshort{SQ2}.
    
    In \textbf{Chapter \ref{chap:construction_grammar} Social Networks and Diachronic Embeddings}, I extend the analysis beyond primary place-based communities to consider the New Zealand Reddit ecosystem as a speech community. I use \acrshort{C2xG} to measure the grammatical distance between New Zealand-related communities at both the user and community levels. Finally, I train an embedding model to evaluate diachronic change, addressing both \acrshort{SQ2} and \acrshort{SQ3} regarding the contiguity of place-related speech communities.
    
    In the final chapter, \textbf{Chapter \ref{chap:conclusion} Conclusion}, I synthesise the key findings from the four research phases. I discuss how these results expand upon existing literature and support a collective understanding of geographic dialect alignment within the New Zealand context. I also address the limitations of the research and propose future directions for exploring geolinguistic variation in digital spaces.

    Following the bibliography, the Appendix includes an overview of \acrshort{NZE} features (Appendix \ref{app:nze}), miscellaneous figures and tables (Appendix \ref{app:misc}), and statistical models (Appendix \ref{app:models}). This is followed by glossaries for place names and other proper nouns, \acrshort{NZE}, te reo Māori, and miscellaneous definitions.

% -----------------------------
% Chapter 2: Literature Review
% -----------------------------

\chapter{Literature Review}
\markboth{Literature Review}{}
\label{chap:literature}

\section{Chapter Outline}
\label{review:outline}

    I begin the literature review by introducing fundamental concepts within dialectology and sociolinguistics (Section \ref{review:lvc}). Following this, I describe relevant \acrshort{NLP} approaches (Section \ref{review:NLP_considerations}) and provide a review of a primary social media platform: Twitter\textsuperscript{X} (Section \ref{review:x_twitter}). I then offer relevant theoretical perspectives from human geography and linguistics regarding language use and place identity (Section \ref{review:theoretical_perspectives}). Finally, I introduce the sociolinguistic context of New Zealand, where I describe the specific features associated with \acrshort{NZE} (Section \ref{review:nze}).

\section{Introduction}
\label{review:introduction}

    Social media sociolinguistics and dialectology is by no means a new field \citep{calude_linguistics_2023}. An early example of research within a digital platform involved a dialectological study of an \acrshort{IRC} channel \citep{androutsopoulos_exploring_2004}, where researchers investigated \texttt{\#mannheim} to identify features specific to the Mannheim variety of German. Recently, the greatest advancements in computational sociolinguistics have occurred within \acrshort{NLP}, particularly in distinguishing closely related varieties, dialects, and languages \citep{zampieri_natural_2020}. These approaches broadly fall into two categories: \acrfull{NLU} and \acrfull{NLG}. Of primary interest to social media dialectology are \acrshort{NLU} tasks, such as dialect identification and sentiment classification \citep{joshi_natural_2025}. In contrast, generation tasks involve processing text sequences to produce new content, such as summarising long-form documents.

    Computer Mediated Communication (\acrshort{CMC}) is defined by the requirement of a machine to transmit linguistic signals \citep{biber_register_2009}. While early \acrshort{CMC} was heavily constrained by the space and cost limitations of protocols like SMS, modern social media has evolved into a diverse global ecosystem. A significant development in this landscape was the emergence of Locative Social Media, or \acrfullpl{LBSN}, such as Foursquare, which merged social networking with geographic documentation \citep{evans_locative_2015}. A defining feature of these platforms is the capture of geographic information, categorised as either explicit or implicit \citep{hu_geo-text_2018}. Explicit information refers to precise spatial coordinates, whereas implicit information includes physical addresses, location metadata, or geo-semantic information (such as toponyms) within text and user profiles \citep{hu_geo-text_2018, marti_social_2019}.
    
    The introduction of \acrshortpl{LBSN} has enabled researchers to examine linguistic variation based on these locations, a shift further facilitated by the launch of official \acrshortpl{API} \citep{junger_brief_2021}. For providers, managing an \acrshort{API} allows for controlled data access \citep{bucher_objects_2013}, while for researchers, it provides a systematic method for collecting the vast corpora necessary for modern \acrshort{NLP} and computational dialectology. This technological shift has transformed social media into a primary site for geolinguistic enquiry. However, the move from explicit spatial coordinates to implicit, community-driven data requires a parallel shift in theoretical focus. While \acrshortpl{API} provide the means to retrieve vast quantities of data, the interpretation of this data necessitates a more in-depth understanding of how geographic space is perceived and enacted by users. 
    
    In the following sections, I transition from linguistic frameworks to technical mechanisms required to analyse variation within these digital environments, focusing specifically on the distinction between absolute space and the socially constructed concept of place. Within the context of this thesis, these intra-disciplinary perspectives are essential for identifying how New Zealand-specific linguistic features - from unique lexical items to distinct morphosyntactic constructions - are enregistered and maintained within the digital \textit{places} of the \acrshort{NZE} ecosystem.

\section{Language Variation and Change}
\label{review:lvc}

    Variation is inherent to language \citep{weinreich_empirical_1968}. If language and its speakers are perceived as a complex adaptive social system (\citealp{steels_language_2000}; \citealp{ellis_language_2009}; \citealp{the_five_graces_group_language_2009}), then variation is the catalyst of change. These variations are typically categorised into four forms: diaphasic (style or register), diastratic (social groups), diachronic (time), and diatopic (space) \citep{zampieri_natural_2020}. These dimensions are deeply interrelated and collectively contribute to our understanding of \textit{orderly heterogeneity} \citep{weinreich_empirical_1968} - the principle that linguistic variation is not random but is instead a structured system governed by social and contextual factors under the framework of variationist sociolinguistics. In this section, I review the relevant literature from traditional dialectology and sociolinguistics that informs this multidimensional approach to linguistic variation.

\subsection{Dialectology}
\label{review:dialectology}

    Dialectology is the scientific study of language variation across geographic regions (dialect geography) and social groups (sociolinguistics) \citep{trudgill_dialect_1983}. As a discipline, it examines the development, diffusion, and evaluation of linguistic varieties. While the boundaries between dialect geography and sociolinguistics are permeable, it is important to note that \textit{sociolinguistics} carries a different connotation within British and North American academic traditions (\citealp{trudgill_social_1971}; \citealp{labov_social_2006}). To document these phenomena, dialectologists rely on mapping and the identification of isoglosses as their primary toolkit, a methodology traditionally applied to the speech of \acrshort{NORM} speakers to produce comprehensive linguistic atlases \citep{chambers_dialectology_1998}.
    
    While a dialect can be broadly defined as a social or regional variety of a language \citep{trudgill_dialect_1983}, the term continues to carry negative connotations for non-linguists. To address these misconceptions, \citet{wolfram_american_2016} outlined what a dialect is not:

    \vspace{12pt}
    
    \begin{enumerate}[nolistsep]
        \item A dialect is something \textit{someone} else speaks.
        \item Dialects always have highly noticeable features that set them apart.
        \item Only varieties of a language spoken by socially disfavoured groups are dialects.
        \item Dialects result from unsuccessful attempts to speak the ``correct'' form of a language.
        \item Dialects have no linguistic patterning in their own right; they are deviations from standard speech.
        \item Dialects inherently carry negative social connotations.
    \end{enumerate}

    \vspace{12pt}

\subsubsection{Models of Language Change}

    Dialectologists observed patterns of language change by visualising the spatial distribution of linguistic features. The two prevailing models include wave theory \citep{schmidt_verwandtschaftsverhaltnisse_1872} and the gravity model \citep{trudgill_social_1971}. The wave theory of change proposes that innovative linguistic features emerge from a central point and are gradually incorporated into nearby communities \citep{schmidt_verwandtschaftsverhaltnisse_1872}. This model emerged as an alternative to the tree model, which viewed the evolution of language through a `genetic' relationship \citep{schleicher_ersten_1853}. Support for the wave model arose from observing the distribution of areal features in neighbouring Germanic languages - patterns that the linear descent of the tree model could not adequately explain \citep{chambers_dialectology_1998}.
    
    A key assumption of the wave model is that linguistic innovations spread continuously - a phenomenon known as the neighbourhood effect. However, advancements in dialectology have revealed that the spread of innovative forms is often, in fact, discontinuous \citep{chambers_dialectology_1998}. The gravity model, also known as the geographic diffusion model, proposes that innovative linguistic features `jump' between major urban centres in a spatially discontinuous manner \citep{trudgill_social_1971}. Simply put, the model predicts that innovative linguistic forms are more likely to emerge in larger population centres. It quantifies the level of linguistic influence a location exerts on its surrounding areas based on priors such as interaction rates, population size, and physical distance \citep{trudgill_social_1971}. 

\subsubsection{Perceptual Dialectology}

    Perceptual dialectology (or folk linguistics) examines how non-linguists perceive, map, and evaluate language variation \citep{niedzielski_folk_2000}. Early methodological approaches included the Little Arrow method, developed in the 1950s, where participants indicated perceived linguistic similarities or differences between geographic regions in the Netherlands \citep{rensink_informant_2008}. During the same period, \citet{grootaers_new_1962} established the concept of \textit{difference boundaries} to map subjective linguistic perceptions in Japan. \citet{preston_five_1986} formalised these inquiries by introducing a five-point method - comprising draw-a-map tasks, degree of difference ratings, pleasantness/correctness scales, voice identification, and qualitative interviews to produce detailed maps of perceptual dialect areas. While these tasks (such as draw-a-map) originated in cultural geography, they have become central to understanding the sociolinguistic landscape. Although the perceptions of laypeople do not always mirror actual usage \citep{preston_language_2010}, perceptual dialectology provides critical insights into the link between language and social identity. Ultimately, speaker perception plays a vital role in both language maintenance and linguistic change \citep{niedzielski_folk_2000}.

\subsubsection{Dialectometry}

    The introduction of quantitative and computational methods to dialectology led to the development of dialectometry \citep{chambers_dialectology_1998}. Foundational work in this field applied statistical techniques to analyse large-scale linguistic corpora, moving away from purely qualitative descriptions (\citealp{seguy_relation_1971}; \citealp{goebl_dialectometry_1993}). A significant advancement was the introduction of the Levenshtein distance (or edit distance) by \citet{nerbonne_phonetic_1996} as a method for objectively measuring phonetic differences between varieties. This string-based metric quantifies the number of insertions, deletions, and substitutions required to transform one sequence into another. Another prominent measure in the field is \acrfull{GIW}, a frequency-weighted similarity metric used to assess the importance of specific linguistic features across regions \citep{lafkioui_dialectometry_2008}.
    
    The key paradigmatic shift from traditional dialectology to dialectometry was the move toward corpus-informed approaches, enabling the aggregation and quantitative analysis of linguistic features at scale. This evolution gave rise to specialised fields such as corpus-based dialectometry \citep{szmrecsanyi_corpus-based_2011} and sociolectometry \citep{ruette_semantic_2014}. Further contributions to the field include the validation of lexical distance measures \citep{goebl_recent_2006} and the formulation of the Fundamental Dialectological Postulate, which asserts that ``geographically proximate varieties tend to be more similar than distant ones'' \citep[p.154]{nerbonne_toward_2007}. Additionally, researchers have investigated the applicability of these metrics within smaller speech communities, testing the limits of quantitative regional analysis \citep{nerbonne_geographic_2007}.
    
\subsection{Sociolinguistics}
\label{review:sociolinguistics}

    Sociolinguistics - under the Labovian tradition - and linguistic anthropology emerged from the American anthropological tradition known as the ``ethnography of speaking'' \citep{hymes_ethnography_1962}. The initial scope of the field, as defined by \citet{hymes_scope_1972}, was to provide an integrated approach to linguistic description that met the scientific and practical needs of inquiry while countering the prevailing `intuitionist' approach to linguistic theory. This perspective is best exemplified by the generative grammar paradigm led by Noam Chomsky, which focuses on language \textit{competence} - the abstract, underlying knowledge of grammar \citep{chomsky_aspects_1965}. In contrast, sociolinguists adopt an inductive, or `frequentist', approach to study language \textit{performance}, focusing on the actual use of language within its various social contexts \citep{hymes_scope_1997}.

\subsubsection{Variationist Perspectives}
\label{review:variationist_sociolinguistics}

    Sociolinguistics is generally divided into two distinct yet complementary approaches: variationist sociolinguistics, which quantifies the relationship between social factors and linguistic variation \citep{labov_sociolinguistic_1972}, and interactional sociolinguistics, which focuses on the social context of specific interactions \citep{gumperz_discourse_1982}. Focusing on language use at the societal level, variationist sociolinguistics operates on the core assumption that non-linguistic factors influence linguistic variation and change \citep{weinreich_empirical_1968}. The approach to this relationship has evolved significantly over time; \citet{eckert_three_2012} identified three paradigmatic shifts within the field, categorised as the first wave, the ethnographic second wave, and the stylistic third wave. 
    
    Beginning in the late 1960s and 1970s, the first wave of sociolinguistic enquiry was marked by the introduction of quantitative empiricism to linguistics \citep{eckert_three_2012}. William Labov introduced the foundational methods of variationist sociolinguistics through his seminal field research on Martha's Vineyard and in New York City \citep{labov_sociolinguistic_1972}. These studies established a quantifiable relationship between a speaker's social identity and their use of the vernacular, defined as the ordinary, informal form of spoken language situated in its natural social context \citep{labov_study_1968}. 
    
    The primary object of study was the sociolinguistic variable, defined as a ``linguistic variable that shifts from one context to another'' \citep[p.112]{labov_principles_1972}. Structurally, the variable represents a single unit of language with two or more ways of saying the same thing, known as variants. These variants are considered to be semantically equivalent - meaning they carry the same denotative message - but differ in their social or stylistic significance. The seminal study on Martha's Vineyard revealed that the choice between these linguistic variants is not arbitrary; rather, it is systematically influenced by the social order and norms within a speech community \citep{labov_sociolinguistic_1972}. By quantifying these distributions, the first wave established that linguistic variation is the primary mechanism through which social meaning is encoded and through which language change is initiated.
    
\subsubsection{Ethnographic Perspectives}
\label{review:social_networks}

    In contrast to the first wave, the second paradigm - spanning the 1980s to the 1990s - placed a heavy emphasis on social agency and how speakers express local or class identity through vernacular language \citep{eckert_three_2012}. This paradigm was inspired by the work of John J. Gumperz on interactional sociolinguistics, an approach that relies on discourse analysis to understand how speakers use language to create meaning through social interaction within a network \citep{gumperz_discourse_1982}. This period saw sociolinguists re-engage with linguistic anthropology, as the two disciplines had drifted apart since their inception \citep{duranti_language_2003}. A key contribution during this time was Lesley Milroy’s seminal study on Belfast English, which established a clear association between a speaker’s degree of community integration and their linguistic choices.
    
    By recruiting participants through the friend-of-a-friend approach, \citet{milroy_language_1987} observed how speakers in different social contexts used variants of eight variable vernacular features. A Network Strength Score was assigned to each speaker to quantify their level of community integration. In one neighbourhood, a high Network Strength Score was directly linked to a greater use of vernacular variants. Similarly, vernacular forms were more prominent in the speech of men, who belonged to denser networks, than in the speech of women. \citet[p.179]{milroy_language_1987} concluded that ``the closer an individual's network ties are with [their] local community, the closer [their] language approximated to localised vernacular norms''. These results suggest that social networks function as a critical norm-enforcement mechanism, conditioning linguistic behaviour based on community integration.
    
\subsubsection{Stylistic Perspectives}

    Building on these foundations, the third wave of sociolinguistics shifted the focus from broad social categories towards the active construction of persona and stance. The New York Department Store study, originally published in \citet{labov_social_2006}, was part of a broader investigation into New York English \citep{labov_sociolinguistic_1972}. While the Martha's Vineyard study focused on conservative and innovative variants, the Department Store study aimed to identify patterns of overt and covert prestige mediated by \textit{style}. In New York City, the rhotic variant was associated with overt prestige, aligning with the standard variety and formal contexts. Whereas the first-wave and ethnographic approaches interpret the relationship between linguistic variation and social attributes in a constitutive manner, the third sociolinguistic paradigm views variation as a deliberate stylistic practice by speakers motivated by their social identities \citep{eckert_three_2012}.
    
    This deliberate stylistic practice is best exemplified in \citet{kiesling_mens_1998}, who found that the use of apical -\textit{ing} (commonly known as \textit{g}-dropping) among fraternity members was associated with a working-class identity. Furthermore, this variant indexed specific stances - such as being hard-working, rebellious, casual, or confrontational - depending on the interactional context. \citet{silverstein_indexical_2003} formalised this meta-pragmatic process as the indexical order, drawing on the class stratification data from the New York study \citep{labov_social_2006}. The indexical order describes the process by which a group identity becomes salient, allowing linguistic features associated with that identity to index group membership.

\subsection{Summary}

    While sociolinguistics has historically prioritised the spoken vernacular \citep{labov_study_1968}, scholars have increasingly noted that the language used within social media platforms blurs the traditional boundaries between speech and writing \citep{cutler_introduction_2022}. Often described as \textit{written orality}, this medium represents a ``complex union of writing in digital space: written speech, written vernacular, silent orality, written orality, digital orality, and/or literate orality'' \citep[p.19]{cutler_introduction_2022}. This hybridity arises because, like all other modes of language, the act of digital writing requires engagement with diverse linguistic resources - semantic, pragmatic, and metapragmatic - alongside broader sociocultural frameworks \citep{blommaert_writing_2013}. Consequently, the rapid evolution of social media as a dominant form of \acrshort{CMC} has introduced significant opportunities and challenges for traditional methods of analysing linguistic variation and change \citep{nguyen_dialect_2021}.

\section{Natural Language Processing for Social Media}
\label{review:NLP_considerations}

    \acrfull{NLP} is primarily concerned with the processing, analysis, and generation of human language, encompassing both written text and spoken speech. Core tasks within the field include language classification, generation and transformation, and information retrieval. The methodologies employed to achieve these tasks directly complement corpus-driven approaches in linguistics. In this context, a corpus is a large, structured collection of digitised text designed for systematic linguistic analysis. Early foundational datasets, such as the Brown Corpus \citep{francis_brown_1979}, were considered substantial for containing one million word tokens.
    
    However, technological advancements have exponentially increased our capacity for the collection, storage, and management of data. For instance, the \acrfull{CGLU} contains 423 billion words from the web and 20 billion words from social media \citep{dunn_mapping_2020}. The following section provides a review of computational models of language (Section \ref{review:computational_models}), with a particular focus on language and dialect modelling. In Section \ref{review:x_twitter}, I discuss how these approaches have been applied to social media data. Although social media encompasses various multi-modal forms of \acrshort{CMC}, this review focuses specifically on written language.
    
\subsection{Computational Models of Language}
\label{review:computational_models}

    When working with written text, language can be decomposed into orthography (encompassing both autonomous and sociocultural models), spelling, and writing systems \citep{sebba_spelling_2008}. In alphabetic systems like English, words and characters - including whitespace - serve as the fundamental units of analysis. Computational models provide the necessary framework for linguists to manipulate and analyse these units within large-scale corpora. In this section, I describe the fundamental \acrshort{NLP} techniques for processing language data and the specific task of dialect modelling. Furthermore, I introduce \acrshort{C2xG}, a usage-based computational approach to \acrshort{CxG} \citep{dunn_computational_2024}.

\subsubsection{Language Modelling Approaches}
\label{review:language_modelling}

    Early computational models were constrained by both limited data availability and computational inefficiency \citep{ghaseminejad_raeini_evolution_2025}. While originally focused on machine translation, the field of \acrshort{NLP} has undergone significant paradigmatic shifts, evolving from rule-based logico-grammatical systems toward modern artificial intelligence frameworks \citep{jones_natural_1994}. Fundamental approaches to language modelling have transitioned through statistical, neural, and transformer-based techniques \citep{ghaseminejad_raeini_evolution_2025}. These methodologies represent a steady progression from context-poor statistical models toward context-sensitive architectures specifically designed to resolve challenges such as lexical ambiguity; this is particularly evident in the transition from statistical models and static embeddings to dynamic models that can distinguish between homonymy and polysemy \citep{krovetz_homonymy_1997}. An overview of these prevailing language modelling approaches is provided in the sections below.

\paragraph{Statistical Language Models}

    \begin{table}
        \scriptsize
        \onehalfspacing
        \centering \caption{\label{tab:lexical_semantic} Lexical Semantic Relationships}
        \renewcommand{\arraystretch}{1.4}
            \begin{tabularx}{\linewidth}{l*{1}{>{\arraybackslash}X}}
            \toprule
            \textbf{Relationships} & \textbf{Definition} \\ 
            \midrule
            Antonymy & words with opposing meanings \\
            Homonymy & words which share an identical form but maintain distinct unrelated meanings \\
            Hyponymy & a word that is a subcategory of a more general class of words \\
            Hypernymy & a word that encompasses the meanings of a subcategory of words  \\
            Meronymy & a word that denotes part of a larger whole \\
            Metonymy & a word or figure of speech that replaces the name of another word \\
            Polysemy & a word with multiple related meanings \\
            Synonymy & words which share an identical or very similar meaning  \\
            \bottomrule
            \end{tabularx}
    \end{table}

    Prior to the popularisation of of deep learning, statistical language models served as the primary tools for computational linguistics \citep{hearst_review_1994}. Fundamental models techniques Markov chains (\citealp{markov_example_2006}; \citealp{shannon_mathematical_1948}) and Hidden Markov Models \citep{baum_statistical_1966}. An $n$-gram model is a probabilistic tool that predicts the next item in a sequence - based on the preceding $n-1$ items \citep{dunn_natural_2022}. Standard variations include unigram ($n=1$), bigram ($n=2$), or trigram ($n=3$) models. For instance, a word-level trigram model references the previous two words to predict the most likely subsequent word. These statistical language models were capable of performing the majority of \acrshort{NLP} tasks including machine translation, speech recognition, and information retrieval \citep{jurafsky_speech_2026}.
    
    While $n$-gram models are simple and efficient, they often treat words as an unordered set rather than a cohesive sequence \citep{krovetz_homonymy_1997}. Consequently, complex grammatical relationships - such as morphosyntactic or semantic dependencies - are often lost. This is particularly problematic for the 1-gram model, also known as the bag-of-words model \citep{harris_distributional_1954}, which fails to account for lexical semantic relationships like homonymy and polysemy. Similarly, multi-word entities, such as place names, may be inadvertently separated. Some of these issues can be mitigated during pre-processing by merging high-frequency $n$-grams using techniques such as Average Mutual Information \citep{church_word_1990}. Although Term Frequency-Inverse Document Frequency was introduced to refine this by weighting importance across a corpus \citep{sparck_jones_statistical_1972}, it still struggles to capture deeper contextual meaning or the nuances of word order.

\paragraph{Neural Language Models}
    
    Neural language models, such as \acrfullpl{RNN} and Word2Vec \citep{mikolov_efficient_2013}, utilise neural networks to predict word sequences. Word2Vec is a shallow, two-layer neural network that converts text into numerical vectors \citep{mikolov_efficient_2013}. These vectors, also known as word embeddings, typically rely on one of two architectures: \acrfull{CBOW} or \acrfull{SGNS}. The \acrshort{CBOW} model predicts a target word based on its neighbouring words, whereas the \acrshort{SGNS} model predicts the surrounding context based on a specific target word. Consequently, \acrshort{CBOW} models are often faster to train on large datasets, whereas \acrshort{SGNS} is more computationally expensive but often superior for capturing rare words. \acrfull{GloVe} is an alternative embedding approach; unlike the predictive nature of Word2Vec, it is a count-based model that trains on a global word-word co-occurrence matrix \citep{pennington_glove_2014}.
    
    Word embedding models address several limitations of statistical language models by capturing the meaning of a word based on its surrounding context \citep{mikolov_efficient_2013}. These lexical semantic relationships are examined using vector semantics based on the distributional hypothesis, which posits that semantic similarity corresponds to geometric similarity in a high-dimensional space \citep{harris_distributional_1954}. Consequently, semantic relationships between words can be quantified using distance metrics such as cosine similarity. For example, the relational distance between countries (e.g., \textit{China} and \textit{Russia}) and their respective capital cities (\textit{Beijing} and \textit{Moscow}) remains stable within the vector space. The introduction of these word embedding architectures has enabled researchers to systematically compare vector representations trained on data from different geographic dialects \citep{bamman_distributed_2014}.

\paragraph{Transformer-Based Models}

    The introduction of transformer-based models - the architecture underpinning modern \acrfullpl{LLM} - marked a shift from static vectors, such as Word2Vec and \acrshort{GloVe}, to context-sensitive embeddings \citep{vaswani_attention_2017}. Because Word2Vec assigns a single, fixed vector to each word regardless of its usage, these models often struggle to distinguish between homonyms or represent the nuanced variations of polysemous words \citep{hedderich_using_2019}. The core innovations of the transformer architecture are the self-attention mechanism and the encoder-decoder structure. A prominent example is the encoder-only \acrfull{BERT} model \citep{devlin_bert_2019}. Unlike static word embedding models, \acrshort{BERT} is `context-aware', as it analyses the linguistic environment both before and after a masked token to generate a dynamic representation.
    
    A significant challenge of these \acrshortpl{LLM} is the massive volume of data required for training from scratch. Consequently, researchers typically fine-tune pre-trained models on specialised datasets to optimise them for specific \acrshort{NLP} tasks. Popular open-source examples include \texttt{RoBERTa} \citep{zhuang_robustly_2021}, \texttt{XLM-RoBERTa} \citep{conneau_unsupervised_2020}, and \texttt{GPT-2} \citep{radford_language_2019}. The weights of these models can be adjusted through domain adaptation to improve performance in niche contexts \citep{ganin_unsupervised_2015}; however, significant ethical concerns remain. Because \acrshortpl{LLM} can inadvertently learn and propagate harmful stereotypes from their training data \citep{steed_upstream_2022}, they often exhibit issues such as geographic bias \citep{manvi_large_2024}. This is particularly detrimental to the representation of low-resource languages and dialectal varieties \citep{dunn_geographically-balanced_2020}.

\subsubsection{Dialect Modelling}
\label{review:dialect_modelling}

    One research task within \acrshort{NLU} is dialect modelling. Within the field of \acrshort{NLP}, this often involves tasks such as dialect identification and sentiment classification \citep{joshi_natural_2025}. Machine translation also benefits from dialect modelling, which facilitates translating between closely related languages and varieties \citep{zampieri_natural_2020}. Early approaches to this area of inquiry typically employed distance-based metrics to measure differences between two texts \citep{joshi_natural_2025}. Dialect modelling is now largely conceived as a classification problem, typically using word- or character-level $n$-grams to distinguish between varieties \citep{taha_comprehensive_2024}. However, issues like homonymy and polysemy not only effect information retrieval within languages \citep{krovetz_homonymy_1997}, but also across similar languages, varieties, and dialects \citep{zampieri_natural_2020}.

\paragraph{Distance-Based Metrics}

    Similar to the Levenshtein distance (or edit distance) in dialectometry \citep{nerbonne_phonetic_1996}, distance measures provide a standardised, interpretable framework for quantifying relationships between two targets \citep{mikolov_efficient_2013}. For example, Levenshtein distance measures the sequence difference between two strings. Common distance measures include \citep{kgosietsile_cosine_2025}:
    
    \begin{itemize}
        \item \textbf{Cosine Similarity}: Used primarily in high-dimensional text analysis to determine the orientation between two entities. Unlike Euclidean distance, this approach is invariant to vector magnitudes, focusing entirely on the angle between vectors rather than their size or frequency.
        \item \textbf{Euclidean Distance}: Serves as a geometric measure of the straight-line distance between two points in a multidimensional space. By calculating the difference in magnitude across various dimensions, this metric quantifies the literal space between coordinates.
    \end{itemize}
    
    Other common measures include Hamming distance, Jaccard similarity, Earth Mover's distance, and Mahalanobis distance \citep{kgosietsile_cosine_2025}. Additionally, Burrows' Delta \citep{burrows_delta_2002} - based on Manhattan distance - is frequently used in stylometry for authorship attribution to calculate the distance between two texts.
    
    One of the first studies to demonstrate diachronic semantic change was \citet{kim_temporal_2014}, who used static word embeddings on the Google Books Corpus. This approach was later formalised by \citet{hamilton_diachronic_2016}, revealing that changes in these vector representations follow predictable linguistic laws. However, much like the $n$-gram models used in shallow text classifiers, the bag-of-words nature of static embedding models can still obscure the syntactic relationship between words, thereby eliminating latent linguistic variation that would otherwise be observed in the grammar.
    
    While transformer-based models appear more reliable for \acrshort{NLU} and \acrshort{NLG} tasks \citep{devlin_bert_2019}, traditional distance measures like cosine similarity can become less effective in extremely high-dimensional spaces if they ignore crucial vector magnitude data \citep{kgosietsile_cosine_2025}. Furthermore, the lack of ground-truth metadata for the corpora used to train many open-source models raises questions about whether these representations truly track diachronic change or merely reflect shifts in the underlying data sources \citep{koplenig_impact_2017}.
    
\paragraph{Text Classification}

    While text classification remains the primary approach for distinguishing geographic dialects, the overarching goal of dialect modelling is to examine linguistic variation within a high-dimensional space \citep{dunn_global_2019}. This contrasts with earlier methods in dialectology and sociolinguistics that relied on a limited set of discrete features. Text classification models follow a well-established pipeline: (a) feature extraction, (b) dimensionality reduction, (c) model training, and (d) evaluation \citep{kowsari_text_2019}. Within this framework, certain architectures - such as \acrfull{SVM} \citep{joachims_text_1998} - and features - such as word or character 
    $n$-grams \citep{taha_comprehensive_2024} - have become industry standards.
    
    As a tool for language understanding, text classification differentiates text samples based on high-frequency token features \citep{joshi_natural_2025}. However, social media data is typically characterised by a high density of named entities, such as toponyms and local abbreviations \citep{eisenstein_latent_2010}. While researchers can iteratively prune these entities to minimise their influence on the model, such interventions raise critical questions regarding the extent to which georeferenced data can truly represent underlying populations \citep{johnson_geography_2016}.
    
    Furthermore, inherent confounds such as sample length \citep{figueroa_text_2013} suggest that text classification is an imperfect proxy for exploring linguistic variation in high-dimensional spaces. Ultimately, relying solely on classification may obscure the nuances of geographic dialects, making it a limited approach for capturing the full complexity of geolinguistic variation \citep{nguyen_dialect_2021}.

\subsubsection{Computational Construction Grammar}

    So far, I have described statistical, neural, and transformer-based computational models of language. The following section describes a usage-based approach to modelling language, which posits that usage events (or linguistic experience) are crucial to the ongoing structural development of linguistic systems \citep{diaz-campos_handbook_2023}. One such approach is \acrshort{C2xG} \citep{dunn_computational_2017}, which accounts for morphosyntax - the intersection of the lexicon and grammar. \acrfull{CxG} itself is a usage-based framework for syntax \citep{goldberg_constructions_2006} that focuses on how specific forms become entrenched. For example, in \acrshort{NZE}, \textit{in the weekend} is preferred over the American English \textit{on the weekend}, or the older British English variety \textit{at the weekend} \citep{trudgill_international_2017}.
    
    Essentially, \acrshort{CxG} posits that the lexicon and syntax are inseparable and interdependent components, forming a single lexico-grammar. This contrasts with traditional models that treat the lexicon and grammar as distinct modules. Instead, a grammar is composed of a set of constructions (a \textit{constructicon}), which act as constraint-based representations \citep{dunn_computational_2017}. These slot-constraints vary in their levels of abstractness; slot-fillers include lexical or item-specific constraints, syntactic or form-based constraints, and semantic or meaning-based constraints. Examples of syntactic (\ref{c2xg:1}) and semantic (\ref{c2xg:2}) constraints are presented below:

    \begin{center}
    \singlespacing
    \begin{align}
        \textnormal{[\textsc{syn}:\textsc{v} -- \textsc{syn}:\textsc{prp}]} \label{c2xg:1} \\
        \notag \\
        \textnormal{\texttt{breaks down}} \tag{i} \\
        \textnormal{\texttt{breaks into}} \tag{ii} \\
        \textnormal{\texttt{breaks with}} \tag{iii} \\
        \textnormal{\texttt{brushes against}} \tag{iv} \\
        \textnormal{\texttt{cooks through}} \tag{v} \\
        \textnormal{...} \notag \\
        \textnormal{\texttt{wipes out}} \tag{\textit{vi}} \\
        \notag \\
        \eqname{Source: Reddit}
    \end{align}
    \end{center}

    \begin{center}
    \singlespacing
    \begin{align}
        \textnormal{[\textsc{syn}:\textsc{mod} > \textsc{syn}:\textsc{aux} -- -- \textsc{syn}:\textsc{v}]} \label{c2xg:2} \\
        \notag \\
        \textnormal{\texttt{can be done}} \tag{i} \\
        \textnormal{\texttt{can be found}} \tag{ii} \\
        \textnormal{\texttt{can be ignored}} \tag{iii} \\
        \textnormal{\texttt{can be replaced}} \tag{iv} \\
        \textnormal{\texttt{can't be done}} \tag{v} \\
        \textnormal{...} \notag \\
        \textnormal{\texttt{could be done}} \tag{\textit{vi}} \\
        \notag \\
        \eqname{Source: Reddit}
    \end{align}
    \end{center}
    \vspace{12pt}

    The syntactic example (\ref{c2xg:1}) represents a phrasal verb pattern; meanwhile, the semantic example (\ref{c2xg:2}) represents a passive modal pattern. As a set of discrete features, these constructions can be used to distinguish and evaluate corpora \citep{dunn_stability_2022}. These constructions offer a more cognitively and socially realistic model of language in contrast to the bag-of-words approach \citep{morin_dialect_2020}.
    
\subsection{Twitter: the Digital Town Square}
\label{review:x_twitter}

    One of the most influential social media platforms within research contexts is X, formerly known as Twitter \citep{mir_rise_2023}. A bibliometric analysis revealed that 6,193 publications have utilised Twitter\textsuperscript{X} as a primary data source \citep{mir_rise_2023}. Research involving the platform generally falls into four main categories: social network analysis, sentiment analysis, spam and bot detection, and hate speech detection \citep{antonakaki_survey_2021}. Before its acquisition by Elon Musk in 2022, researchers could retrieve up to 10 million posts (tweets) per month for free, with access to both real-time and historical data \citep{antonakaki_survey_2021}. This included both explicit and implicit geographic information accessible through the \acrshort{API} \citep{marti_social_2019}.
    
    Because Twitter\textsuperscript{X} has historically been the dominant source of social media language data, it has naturally become the primary focus of social media dialectology. Founded on 21 March 2006, Twitter\textsuperscript{X} is a social networking service where users produce character-constrained messages for a virtual audience. At its peak, the platform hosted 368.4 million users. While it encourages interaction, there is no inherent expectation for users to engage actively with one another. A distinguishing feature of Twitter\textsuperscript{X} is the ability to broadcast views, ideas, and news in real-time; notably, journalists often report breaking news on the platform before official publication. This broadcasting function contrasts with platforms like Facebook - and its predecessors Bebo and MySpace - which are typically restricted to closer familial and social networks.

\subsubsection{Geographic Variation}

    Historically, research using Twitter\textsuperscript{X} has focused on linguistic variation as a source of georeferenced social media language data, with numerous attempts to apply variationist and sociolinguistic theory to these corpora. One of the first studies to integrate georeferenced social media data with dialectology was conducted by \citet{eisenstein_diffusion_2014}. The researchers modelled linguistic diffusion across the contiguous United States using a corpus of 107 million posts (tweets) from 2.7 million users. By mapping the frequency distributions of 2,603 lexical items across the 200 largest \acrfullpl{MSA}, they observed a non-uniform spread of innovative items concentrated in densely populated urban centres - a pattern consistent with the Gravity Model \citep{trudgill_social_1971}. Subsequent studies have confirmed that similar models of linguistic diffusion remain observable on the platform \citep{ilbury_using_2024}.
    
    Machine learning techniques have also been employed to identify distinct dialect regions. By clustering a distance matrix using \acrfull{PCA}, \citet{huang_understanding_2016} identified between three and eight dialect areas across the United States. This was achieved by calculating the frequency distributions of 59 alternating lexical features (e.g., \textit{bag} versus \textit{sack}) for each county. Notably, both \citet{jones_toward_2015} and \citet{huang_understanding_2016} found strong alignment between Twitter\textsuperscript{X} data and the dialect areas previously established by \citet{labov_dialects_2008}.

    Further studies have tested the suitability of Twitter\textsuperscript{X} as a viable alternative to traditional dialectological sources. \citet{grieve_mapping_2019} compared the frequency of 36 lexical variants (such as \textit{mate}, \textit{pal}, and \textit{buddy}) within the United Kingdom and found broad alignment between social media and traditional data. Regarding language change, \citet{grieve_mapping_2018} traced the emergence of innovative lexical features in the United States over 399 days. While they did not identify clear regional patterns, they successfully documented real-time lexical emergence - a process recently expanded upon by \citet{wurschinger_social_2021}, who combined diffusion and emergence models.

    Other approaches involve the visual comparison of Twitter\textsuperscript{X} distributions with existing dialect atlases. For \acrfull{AAVE}, \citet{jones_toward_2015} utilised non-standard orthography as a diagnostic feature. While research has primarily focused on lexis, some studies have examined phonology. Similar to \citet{jones_toward_2015}, \citet{dijkstra_using_2021} used non-standard orthography to identify a change in progress in Frisian relative pronouns undergoing word-final -\textit{t} deletion (e.g., \textit{dy't}$\sim$\textit{dy}). Their results indicated that younger users were more likely to adopt the innovative \textit{t}-less form, with social attributes deduced from user profile analysis \citep{dijkstra_using_2021}.

\subsubsection{Language Modelling}

    To determine the suitability of shallow classifiers for dialect modelling, \citet{dunn_stability_2022} georeferenced language data from Twitter\textsuperscript{X} for four national varieties of English: Australia, Canada, Ireland, and New Zealand. The results indicated that the top predictors were primarily toponyms (e.g., \textit{canada}), named entities (e.g., \textit{kpa}), or public figures (e.g., \textit{trudeau}) associated with those regions. These findings are consistent with early research into social media dialectology, which suggests that language use on these platforms is typically characterised by a high volume of named entities and local abbreviations \citep{eisenstein_latent_2010}. Crucially, neither of these features is associated with latent dialect variation.
    
    As with much of the existing literature, work in this area has largely focused on Twitter\textsuperscript{X} and country-level varieties (\citealp{dunn_modeling_2019}; \citealp{dunn_language_2025}). By analysing online language from CommonCrawl and Twitter\textsuperscript{X} across seven language conditions - including Arabic, English, French, German, Portuguese, Russian, and Spanish - \citet{dunn_global_2019} tested the suitability of \acrshort{C2xG} for text classification within four Inner Circle varieties of English. The results demonstrated that \acrshort{C2xG} features were more reliable than function words or lexical $n$-grams for dialect classification tasks.

\subsubsection{Limitations}

    The availability of explicit geographic information has enabled researchers to explore linguistic variation at scale by employing methodologies from \acrshort{NLP} \citep{marti_social_2019}. For linguists, the existence of variation is a natural feature of language rather than an inherent flaw; however, the over- or under-representation of specific varieties becomes a critical issue in the development of language technologies \citep{joshi_natural_2025}. Many of these systems ingest vast quantities of social media data, often without accounting for demographic or geographic balance. One unintended consequence of this practice is the introduction of algorithmic bias, which can skew the performance of these systems across different communities \citep{hovy_social_2016}.

\paragraph{Social Bias}

    While georeferenced data from Twitter\textsuperscript{X} has become a foundational source for studying language production, it is subject to significant limitations, most notably a lack of explicit sociodemographic information which must often be inferred from secondary sources \citep{sloan_knowing_2013}. While some researchers caution against assuming these users are representative of the public, others argue that, at a macro-level, linguistic trends on the platform are inevitably shaped by broader demographic properties \citep{eisenstein_diffusion_2014}. Nevertheless, empirical evidence suggests that Twitter\textsuperscript{X} users are not representative of underlying geographic populations \citep{sloan_who_2015}.
    
    In the United States, the user base is inherently skewed toward specific demographics: typically younger, more urban, and more likely to belong to ethnic minority groups \citep{mislove_understanding_2011}. Furthermore, users who enable georeferencing often share distinct socioeconomic characteristics, such as specific educational backgrounds and professional occupations in management, science, or the arts \citep{hu_geo-text_2018}. In an attempt to integrate variables such as age, income, and race \citet{eisenstein_diffusion_2014} found that \acrshortpl{MSA} with similar racial profiles exhibited higher linguistic influence; however, this may be an artefact of the skewed sample rather than a true reflection of the underlying population.
    
    Similar challenges were observed by \citet{dijkstra_using_2021}, who noted that while the automatic detection of written Frisian was successful, retrieving social attributes like age and gender was problematic. For instance, only 82.5\% of users provided a birth year, forcing researchers to estimate the remaining 17.5\% through profile pictures or cross-referencing platforms like Facebook or LinkedIn. The lack of standardised metadata for language, age, and gender remains a primary challenge, as researcher-led identification of these social attributes raises both methodological and ethical concerns \citep{williams_towards_2017}.

\paragraph{Geographic Bias}

    In the absence of detailed metadata, it is difficult to examine the sociodemographic make-up of population of texts \citep{nguyen_computational_2016}. To account for possible biases, \citet{dunn_measuring_2020} outlined four specific forms of diversity-related bias that occur when using social media corpora for scientific inquiry. The first is production bias, or the over-representation of digital data from a single source (\citealp{jurgens_incorporating_2017}; \citealp{kulshrestha_geographic_2012}). The second is sampling bias, which occurs when a specific demographic group is disproportionately represented in the digital record \citep{dunn_mapping_2019}, a phenomenon that often extends across various social attributes \citep{mislove_understanding_2011}.
    
    The third form is non-local bias, referring to the linguistic noise created by transient or non-resident populations (\citealp{graham_where_2014}; \citealp{johnson_geography_2016}). Interestingly, the COVID-19 pandemic provided a unique opportunity to evaluate this bias while widespread travel restrictions were in place \citep{dunn_measuring_2020}. The final form is majority language bias \citep{lackaff_local_2016}, where dominant languages overshadow regional varieties. Mitigation strategies include the deliberate inclusion of diverse georeferenced data during model development \citep{jurgens_incorporating_2017} or the use of population-based sampling during corpus production, which has proven effective in correcting both geographic and demographic imbalances \citep{dunn_geographically-balanced_2020}.

\subsection{Summary}
    
    While the focus of this section has been on the opportunities and challenges of applying \acrshort{NLP} techniques to social media analysis, certain limitations remain difficult to address through a purely computational lens. Just as early forms of \acrshort{CMC} were constrained by hardware, modern users must navigate the recommender systems and algorithms built into contemporary platforms (\citealp{van_der_nagel_networks_2018}; \citealp{steen_you_2023}). For example, users on TikTok have modified their linguistic practices to circumvent algorithmic content moderation \citep{steen_you_2023}. This phenomenon - known as \textit{Algospeak} - involves various word-formation strategies, including analogy (e.g., \textit{leg booty community} for LGBT community), the use of numeric characters (e.g., \textit{str8}), non-alphanumeric substitutions (e.g., \textit{sh!t}), and emojis \citep{steen_you_2023}. An additional consideration for modern social media research is the increasing presence of machine-generated language, which further complicates the analysis of human linguistic variation \citep{sadiq_deepfake_2023}. Ultimately, these evolving digital constraints and artificial influences necessitate a more nuanced, \textit{placial} approach to dialectology that accounts for the interactional context and user agency within specific online communities.

\section{Language in the Construction of Place}
\label{review:theoretical_perspectives}

    While these computational and algorithmic hurdles discussed in the previous section are significant, they also highlight the urgent need for a more comprehensive framework in social media research. \citet{nguyen_dialect_2021} noted several critical gaps within computational sociolinguistics, including the need to incorporate geographic and social factors, the treatment of place, the inclusion of user perception, and the bottom-up discovery of features. Furthermore, there remains a pressing need to investigate the role of language variation and change across multiple social media platforms and modalities. Currently, the role of \textit{place} - or implicit geographic information - remains an under-explored area of social media dialectology \citep{nguyen_dialect_2021}. Much of this neglect stems from a preference for ground truth in machine learning, which often introduces a bias in ground truth itself \citep{sogaard_selection_2014}. Despite this, greater attention is required to evaluate georeferenced corpora in the absence of traditional ground truth markers \citep{williams_twitter_2017}. Given the increasing demand for high-quality georeferenced language data - particularly for low-resource languages, dialects, and varieties \citep{zampieri_natural_2020} - it is vital to examine how the concept of place functions within social media platforms.
    
\subsection{Geographic Perspectives}
\label{review:human_geography}
        
    Geography, in the form of spatiality, synchronises the relationship between people and their environment, including other people, on the Earth's surface. This can be conceived in terms of absolute space, a dimension defined by abstract geometries such as distance, direction, size, and volume. Before the Spatial Turn \citep{withers_place_2009}, human geographers often relied on phenomenological approaches, positing that the ``foundations of geographical knowledge lie in the direct experiences and consciousness we have of the world we live in'' \citep[p.4]{relph_place_1976}. In practice, this means that phenomenological geography, as a method, incorporate[s] careful looking, reflection and description'' \citep[p.41]{seamon_phenomenology_1979}. In the following sections, I provide a definition of place and space as they pertain to human geography and their relevance to linguistic variation.

\subsubsection{Sense of Place}

    \begin{figure}
      \centering
      
        \begin{tikzpicture}[
        node distance = 1mm and 1mm,
        start chain = going right,
        alg/.style = {
            draw=none,
            align=center,
            text width=45mm,
            font=\linespread{0.8}}]
        \begin{scope}[every node/.append style={on chain}]
            \node (n1) [alg] {};
            \node (n2) [alg] {\textbf{Place}};
            \node (n3) [alg] {};
        \end{scope}
            \node (t1) [below left= 2cm and -2cm of n2] [alg] {\textbf{Locale}};
            \node (t2) [below left= 2cm and -4.75cm of n2] [alg] {\textbf{Location}};
            \node (t3) [below right= 2cm and -2cm of n2] [alg] {\textbf{Sense of Place}};
            \node (u1) [below left=3.5cm and -2cm of n2] [alg] {\textbf{Setting}};
            \node (u2) [below left= 3.5cm and -4.75cm of n2] [alg] {\textbf{Connection}};
            \node (u3) [below right=3.5cm and -2cm of n2] [alg] {\textbf{Identity}};
        % arrows
        \draw[-] (n2) -- (t1);
        \draw[-] (n2) -- (t2);
        \draw[-] (n2) -- (t3);
        \draw[-] (t1) -- (u1);
        \draw[-] (t2) -- (u2);
        \draw[-] (t3) -- (u3);
    \end{tikzpicture}

      \vspace{6pt}
      \caption{Tripartite Model of Place}
      \label{fig:place}
      
      \vspace{6pt}
      \captionsetup{font=footnotesize, labelformat=empty, justification=justified, singlelinecheck=false}
      \caption*{\setstretch{2}\textbf{Description}: Tripartite model of place from \citet{agnew_place_1987} as adapted from \citet{reed_importance_2020}. The three fundamental components are linked to setting, connection, and identity.}
      
    \end{figure}

    Canadian geographer Edward Relph explained that place ``has a range of subtleties and significances as great as the range of human experiences and intention'' \citep[p.26]{relph_place_1976}. This suggests there are as many interpretations of place as there are - or have been - individuals to experience them. Rather than a singular definition, \citet{relph_place_1976} examines the \textit{essence} of place through its relationship to location, landscape, time, and community. Similarly, Yi-Fu Tuan foregrounds the relational aspect of place, noting its ordinary usage implies two meanings: ``one's position in society and spatial location'' \citep[p.408]{tuan_space_1979}. \citet[p.409]{tuan_space_1979} further elaborates that even ``spatial location derives from position in society rather than vice-versa,'' drawing on the sociological frameworks of \citet{sorokin_sociocultural_1943}.
    
    As an essentially contested concept \citep{gallie_essentially_1956}, British-American political geographer John A. Agnew refrained from offering a singular definition of \textit{place}. Instead, \citet{agnew_place_1987} proposed that the development of a \textit{meaningful place} is a structurated process comprising three fundamental aspects: \textit{location}, \textit{locale}, and \textit{sense of place}. This tripartite model is illustrated in Figure \ref{fig:place}, adapted from \citet{reed_importance_2020}. As interpreted by \citet[pp.639-640]{withers_place_2009}, \textit{location} refers to absolute location, while \textit{locale} represents the material settings for social relations; finally, \textit{sense of place} encompasses the affective attachment individuals have toward that specific environment.

\subsubsection{Digital Placemaking}
\label{review:digital_placemaking}

    Spatial metaphors such as \textit{space} and \textit{place} have long been utilised in computer science to facilitate interaction within collaborative and communicative environments \citep{graham_end_1998}. While \acrfull{GIS} often rely on the digital twin as a virtual model of physical objects, social media facilitates a more active form of digital placemaking. This is exemplified by the concept of the \textit{spatial self}, which refers to the process of online self-presentation rooted in offline physical activities \citep{schwartz_spatial_2015}. Rather than being aspatial or detached from physical reality \citep{gieryn_space_2000}, digital platforms serve as venues for location-based performance. Here, users leverage their physical experiences to communicate identity, transforming abstract digital \textit{space} into meaningful \textit{place} through the subjective social construction of their settings \citep{zimmerbauer_image_2011}. In this context, digital placemaking becomes a performative act where language and location coalesce to signal community belonging.

\subsection{Linguistic Perspectives}
\label{review:linguistic_perspectives}

    Linguists have not been operating in an entirely \textit{placeless} manner. By adopting an ethnographic approach, \citet{labov_principles_1972} engaged closely with both the production and the attitudes of speakers. In the Martha's Vineyard study, \citet{labov_sociolinguistic_1972} noted that the conservative variant signalled Up-island status, serving as a marker of local orientation and a sense of belonging; here, each successive group of islanders functioned as the reference group for new speakers. Ultimately, ``linguistic systems are exercised by speakers, in social space'' \citep[p.573]{patrick_speech_2004}, where individuals actively engage in social and communal activities.

\subsubsection{Speech Community}
\label{review:speech_community}

    Drawing from \citet{milroy_language_1987}, a speaker's community integration is determined by specific criteria: kinship ties with multiple households, a shared workplace with neighbourhood peers of the same gender, regular participation in territorial activities (such as local sports or social clubs), and voluntary association with co-workers outside formal hours. Although a community does not automatically denote a physical location \citep{agnew_place_1987}, the term speech community has historically functioned as a conceptual placeholder for place. After all, ``community can be defined and identified in terms of space, place, affiliation, practices and any combination of these terms'' \cite[p.1]{morgan_speech_2014}. For this research, I adopt the definition of a speech community from \citet{holmes_introduction_2013}:
    
    \begin{quote}
        \footnotesize
        ``[A] group of people who share the same rules of speaking. People who belong to the same speech community interpret events similarly, and know the norms for behaving appropriately in the regular communicative events of the community.'' \cite[p.377]{holmes_introduction_2013}
    \end{quote}
    
    In this concept, community - rather than speech - is the operative term, and the linguistic understanding of what constitutes a community has evolved significantly. The concept of the speech community predates modern sociolinguistics, having been popularised by Leonard Bloomfield during the development of American structural linguistics \citep{bloomfield_language_1933}. The term was likely a translation of the German \textit{Sprachgemeinschaft} (`language community'), inspired by Bloomfield's time with the Neogrammarians \citep{fishman_sociology_1971}. Crucially, Bloomfield defined a speech community as a group of people who use the same system of speech-signals'' \cite[p.29]{bloomfield_language_1933}, reflecting the ethnolinguistic nationalism prevalent in that period \cite[p.3]{morgan_speech_2014}. Since then, the concept has undergone substantial revision; I have summarised these theoretical shifts in Table \ref{tab:speech_community} with different definitions of a \textit{speech community}.

    \begin{table}
        \scriptsize
        \onehalfspacing
        \centering \caption{\label{tab:speech_community} Definitions of Speech Community}
        \renewcommand{\arraystretch}{1.4}
            \begin{tabularx}{\linewidth}{l*{1}{>{\arraybackslash}X}}
            \toprule
            \textbf{Source} & \textbf{Definition} \\
            \midrule
                \citet{bloomfield_language_1933} & ``A speech-community is a group of people who interact by means of speech'' (p.42) \\
                \citet{hymes_models_1967} & ``Tentatively, a speech community is defined as a community sharing both rules for the conduct and interpretation of acts of speech, and rules for the interpretation of at least one common linguistic code.'' (p.18) \\
                \citet{gumperz_speech_1968} & ``Any human aggregate characterized by regular and frequent interaction by means of a shared body of verbal signs and set off from similar aggregates by significant differences in language usage'' (p.381) \\
                \citet{lyons_new_1970} & ``All people who use a given language (or dialect)'' (p.326) \\
                \citet{labov_principles_1972} & ``The speech community is not defined by any marked agreement in the use of language elements, so much as by participation in a set of shared norms; these norms may be observed in overt types of evaluative behaviour, and by the uniformity of abstract patterns of variation which are invariant in respect to particular levels of usage'' (pp.120-121) \\ 
            \bottomrule
            \end{tabularx}
    \end{table} 

\subsubsection{Enregisterment}
\label{review:enregisterment}

    Serious engagement between linguistics and place is most notably attributed to Barbara Johnstone and her work on `Pittsburghese' \citep{johnstone_pittsburghese_2009}. Drawing on the framework of the indexical order, \citeauthor{johnstone_pittsburghese_2009} observed that speakers develop perceptual contrasts in social identity through the metapragmatic process of enregisterment \citep{agha_social_2003}. This social process occurs when specific linguistic features become linked to a socially recognised register or variety. A prominent example is the second-person plural pronoun yinz, which has come to signal the variety of English spoken in Pittsburgh \citep{johnstone_pittsburghese_2009}. This process is expressed through the following simplified schema:

    \vspace{12pt}
    
    \begin{enumerate}[nolistsep]
        \item ``\textit{n}-th-Order Indexical'': A feature whose use can be correlated with a sociodemographic identity (e.g., region or class) or a semantic function (e.g., number-marking).
        \item ``\textit{n}+1-th-Order Indexical'': An \textit{n}-th order indexical feature that has assigned ``an ethno-metapragmatic driven native interpretation'' \cite[p.212]{silverstein_indexical_2003}, i.e., a meaning in terms of one or more native ideologies (the idea that certain people speak more correctly than others, for example).
        \item ``(\textit{n}+1)+1-th-Order Indexical'': An indexical phenomenon at order \textit{n}+1 can come to have another, (\textit{n}+1)+1-th-order, indexical meaning when a subset of its features come to be perceived as meaningful according to another ideological schema.
    \end{enumerate}

    \vspace{12pt}

    Put simply, speakers first notice that certain linguistic variants correlate with specific social traits (\textit{n}-th-Order); they then attribute social meaning to these variants and style-shift accordingly, mediated by ideology (\textit{n}+1-th-Order). Finally, the broader community associates the sum of these linguistic variants to index a cohesive social identity ((\textit{n}+1)+1-th-Order). \citeauthor{johnstone_pittsburghese_2009}'s key finding was that variants common in Southwestern Pennsylvania were enregistered to a Pittsburgh identity, facilitating the formation of \textit{Pittsburghese} and the associated social persona of the \textit{Yinzer}.

\subsection{Sociotheoretical Perspectives}

    \begin{figure}
    \centering
    \footnotesize
    \begin{tikzpicture}[
        node distance = 1mm and 1mm,
        start chain = going right,
        blg/.style = {
            draw=white,
            align=center,
            text width=45mm,
            font=\linespread{0.8}}]
        \begin{scope}[every node/.append style={on chain}]
            \node (m1) [blg] {\textbf{Social structure}};
            \node (m2) [blg] {\textbf{Social action}};
            \node (m3) [blg] {\textbf{Social behaviour}};
        \end{scope}
            \node (v1) [below=of m1] [blg] {Society has fixed structures which constrain individual\\action};
            \node (v2) [below=of m2] [blg] {Social meaning is achieved\\via communicative\\interaction};
            \node (v3) [below=of m3] [blg] {Society construed as\\a set of observable\\individual behaviours};
            \node (w1) [below left= 2cm and -2cm of v2] [blg] {\textbf{Rational action}};
            \node (w2) [below right= 2cm and -2cm of v2] [blg] {\textbf{Praxis}};
            \node (x1) [below left=2.7cm and -2cm of v2] [blg] {Language as strategic\\motivated choices};
            \node (x2) [below right=2.7cm and -2cm of v2] [blg] {Meanings are reciprocal\\interactional achievements};
        % arrows
        \draw[->] (v2) -- (w1);
        \draw[->] (v2) -- (w2);
    \end{tikzpicture}
    \vspace{0.5cm}
    \caption{\label{fig:social_perspectives} Sociotheoretical perspectives adapted from \citet{coupland_introduction_2001}}
    \end{figure}
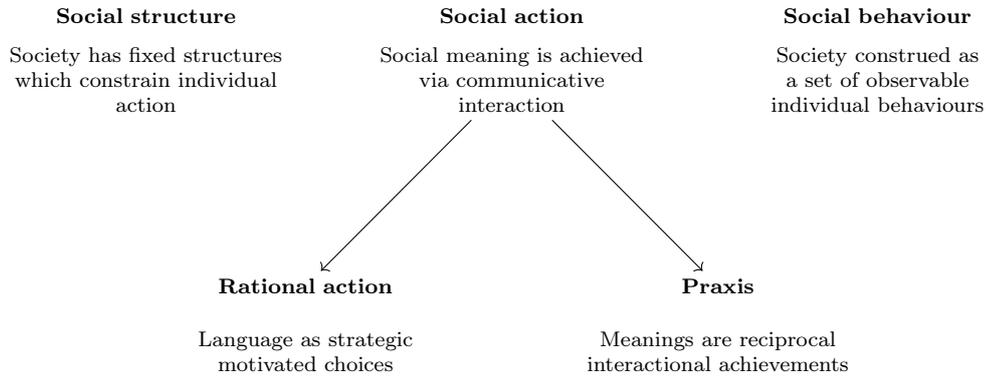

    With a greater understanding of how \textit{place} is conceived in geography and linguistics, the purpose of the following section is to unify the two perspectives through the lens of social theory. In its broadest definition, social theory refers to the analytical frameworks that are used to study and interpret social phenomena \citep{seidman_contested_2017}. Both social theory and linguistic theory share a common genealogy stemming from the structuralism movement \citep{coupland_introduction_2001}. Nineteenth and 20\textsuperscript{th} Century thinkers, such as Durkheim, theorised that societies are complex systems made up of \textit{structures}, underlying systems and patterns, and \textit{functions}, the role these structures play in maintaining social order and stability \citep{sawyer_social_2005}. 
    
    This Structuralist view of society would later develop into \textit{Structural Functionalism}, initiated by American sociologist Talcott Parsons, which have adopted the view that social action (or change) as an adaptive response to some tension in the social system \citep{parsons_structure_1949}. Other epistemological perspectives relevant to social theory are \textit{constructivism}, a cognitive model that ``assumes that people, including researchers, construct the realities in which they participate'' \cite[p.187]{charmaz_constructing_2006}, and \textit{social constructionism} \citep{berger_social_1966} that ``assumes that people create social reality(ies) through individual and collective actions'' \cite[p.189]{charmaz_constructing_2006}. In the adapted model \citep{reed_importance_2020}, locale is extended to include \textit{setting}, connection to \textit{connection}, and \textit{identity}.

\subsubsection{Place Orientation}

    Place has been reconfigured by linguists as place orientation, defined as ``the ways speakers might relate to varied places in which they have spent time'' \citep[p.19]{carmichael_language_2025}. Linguists have approached the measurement of place orientation from various perspectives; one of the earliest was the Cosmopolitan Orientation metric, a composite measure based on speaker attitudes and linguistic habits \citep{solomon_phonological_1999}. This approach is conceptually similar to the Network Strength Score used by \citet{milroy_language_1987}, where speakers are assigned a score based on their integration into local community, workplace, and leisure networks.
    
    \citet[pp.20-22]{carmichael_language_2025} compiled a comprehensive list of place orientation metrics from recent linguistic literature, which I have reproduced in Table \ref{tab:place_attachementa}. These measures are largely derived from three sources: qualitative accounts of personal experience and embodiment (\citealp{schoux_casey_postvocalic_2013}; \citealp{pabst_putting_2022}), demographic background and residential history (\citealp{chambers_region_2000}; \citealp{beaman_identity_2021}; \citealp{jeszenszky_effects_2024}), or survey responses targeting self-identification and future prospects (\citealp{solomon_phonological_1999}; \citealp{reed_sounding_2016}; \citealp{monka_place_2020}; \citealp{beaman_identity_2021}; \citealp{carmichael_locating_2023}).
        
    \begin{table}
        \scriptsize
        \onehalfspacing
        \centering \caption{\label{tab:place_attachementa} Measuring Place Orientation}
        \renewcommand{\arraystretch}{1.4}
            \begin{tabularx}{\linewidth}{ll*{1}{>{\arraybackslash}X}}
            \toprule
            \textbf{Framework} & \textbf{Reference} & \textbf{Metrics} \\ 
            \midrule
            Cosmopolitan Orientation & \citet{solomon_phonological_1999} & 
            (a) experience with and attitudes towards larger urban areas; (b) use of indigenous language in conversation settings \\
            Regionality Index & \citet{chambers_region_2000} & 
            (a) where the speaker was raised from 8 to 18; (b) where the speaker was born; (c) where the speaker lives now; (d) where the speaker's mother and father were born \\
            Local Orientation & \citet{schoux_casey_postvocalic_2013} &
            (a) primarily brought up local topics through the lens of personal experience, and took a strong New Orleans-centric perspective across topics; and, (b) specifically discussed themselves as embodying New Orleanian-ness \\
            Rootedness Score & \citet{reed_sounding_2016} & 
            (a) willingness to relocate; (b) travel habits; (c) self-identification with region; (d) familial connection; (e) areal identification ranking; (f) local integration; (g) centrality of place identity \\
            Index of Local Attachment & \citet{monka_place_2020} & 
            (a) mother's geographical background; (b) father's geographical background; (c) places of residence; (d) places of schooling; (e) location of spare-time job; (f) location of leisure activity; (g) geographical location of friends; (h) future geographical prospects \\
            Swabian Mobility Index & \citet{beaman_identity_2021} & 
            (a) residential dispersion; (b) residential distance \\
            Swabian Orientation & \citet{beaman_identity_2021} & 
            (a) Swabian allegiance; (b) Swabian language attitudes; (c) Swabian cultural competence; (d) Swabian language usage \\
            Local Affiliation Score & \citet{pabst_putting_2022} & 
            (a) description as hard-working; (b) past or present engagement in hunting; (c) past or present engagement in other outdoor activities; (d) description as down to earth, enjoying the simple things in life; (e) mention of the interviewee helping other people; (f) focus on positive aspects of local life and culture; (g) limited time living outside of Northern Maine \\
            MPOMs & \citet{carmichael_locating_2023} & 
            (a) identification as a Chalmatian; (b) stated desire to leave Chalmette before the storm; (c) residential history; (d) school location; (e) workplace location \\
            Linguistic Mobility Index & \citet{jeszenszky_effects_2024} & 
            (a) mother's regional origin; (b) father's regional origin; (c) partner's regional origin; (d) place education; (e) workplace; (f) external residence or time spent living elsewhere \\
            \bottomrule
            \end{tabularx}
    \end{table}

\subsubsection{Place Identity}
\label{review:place_identity}

    While traditional metrics of place orientation rely on physical proximity and residential history, the rise of digital interaction has necessitated a broader understanding of how these orientations are expressed within virtual environments. Early research into \acrshort{CMC} primarily sought to distinguish digital interaction from traditional linguistic forms. For instance, \citet{wick_speech_1997} argued that listservs - subscribable email lists - constituted legitimate speech communities due to their shared communicative norms. To account for these digital spaces, \citet{wick_speech_1997} proposed an adapted definition:
    
    \begin{quote}
        \small
        ``[A] speech community is an organisation of diversity, a group that shares at least one common language, that shares knowledge of the rules for the use and interpretation of speech and attitudes toward speech, and that constitutes itself in the speaking even as it creates a history to ground future speaking.'' \cite[p.14-15]{wick_speech_1997}
    \end{quote}

    Synthesising the foundational work of \citet{gumperz_types_1962}, \citet{hymes_linguistic_1968}, and \citet{labov_sociolinguistic_1972}, this definition recognises that online communities are governed by internal rules and linguistic norms \citep{morgan_speech_2014}. This shift coincided with \citeauthor{gumperz_introduction_1996}'s most inclusive definition, which reframes speech communities as ``collectivities of social networks'' \citep[p.362]{gumperz_introduction_1996}.
    
    Modern scholarship has extended these concepts to social media, where a variety of language is viewed as a population of texts defined by dialect, register, and period \citep{grieve_sociolinguistic_2025}. Processes of enregisterment have been observed on platforms such as Twitter\textsuperscript{X}, where users employ specific features to index digital personae \citep{ilbury_sassy_2020}. Crucially, \citet{ilbury_tale_2022} draws an explicit link to place identity, demonstrating that online communities occupying the same physical geography may modify their language use to negotiate their relationship with a local neighbourhood. In this context, social media becomes a site where place identity is not just reflected, but actively performed and enregistered through digital discourse.

\subsection{Summary}
\label{review:place_summary}

    Drawing from \acrfull{GISc}, \citet[p.1-2]{scheider_places_2010} offered a non-reductionist definition of places where: (a) places are located, but are not locations; (b) places are primary categories of human experience and social constructs; (c) places have stabilising functions that afford insideness; (d) places have material settings (surface layouts). While these criteria traditionally describe physical environments, they provide a powerful lens for examining digital spaces as meaningful \textit{places} rather than abstract data points. Crucially, language serves as one of the primary tools through which these social constructs and stabilising functions are enacted, allowing researchers to computationally analyse the \textit{insideness} of a community by tracing the shared linguistic norms and identities that define it.

\section{Sociolinguistic Context of New Zealand}
\label{review:nze}

    While the theoretical frameworks of language variation, change, and enregisterment can be applied across various contexts, their specific manifestation is deeply rooted in the socio-historical and linguistic conditions of individual regions. In the case of New Zealand, these digital performances occur within a unique post-colonial landscape shaped by the complex relationship between its indigenous heritage and its colonial origins.
    
    New Zealand has two official languages: te reo Māori and \acrfull{NZSL}. According to the 2018 Census, these account for 4.0\% and 0.5\% of the languages used in New Zealand, respectively \citep{stats_nz_place_2024}. English is the most widely spoken language - used by approximately 95.4\% of the population - and serves as the de facto official language of the country \citep{stats_nz_place_2024}. The dominance of English across New Zealand society is a direct result of British colonisation beginning in the 19\textsuperscript{th} century, compounded by the subsequent impacts of globalisation \citep{kachru_standards_1985}.
    
    The origins of \acrshort{NZE} can be traced to the arrival and settlement of British colonists during the 18\textsuperscript{th} and 19\textsuperscript{th} centuries \citep{gordon_new_2004}. Following the signing of Te Tiriti o Waitangi (the Treaty of Waitangi) in 1840, English was elevated to the official administrative language by the 1860s, initiating a significant language shift within te reo Māori-speaking communities. The introduction of compulsory primary education in 1877 further accelerated the process of koineization, or dialect mixing, among children whose parents spoke various regional varieties of English.

\subsection{Features of New Zealand English}

    This section provides a comprehensive overview of \acrshort{NZE} for those unfamiliar with the variety. Given the absence of phonological and prosodic cues in written English, I focus specifically on lexical and grammatical features to distinguish \acrshort{NZE} from other varieties. This approach prioritises the \textit{written orality} of the digital medium \citep{cutler_digital_2022}, where social identity is primarily enregistered through word choice and morphosyntactic variation rather than the spoken accent.

\subsubsection{Lexis}
\label{review:nze_lexis}

    Although \acrshort{NZE} shares 95\% of its lexical items with other varieties of English \citep{deverson_handling_2000}, vocabulary remains one of its most salient markers. Interest in \textit{New Zealandisms} was notably sparked by \citet{gordan_hunting_1957}, followed by \citeauthor{turner_english_1966}'s curation of over 800 entries associated with Australasian Englishes. The now-defunct New Zealand Dictionary Centre managed a database of approximately 42,000 general \acrshort{NZE} words \citep{bardsley_lexicography_2009}, while major published sources include \citet{orsman_new_1994}, which contains 4,500 headwords, and more recent editions of the Oxford New Zealand Programme, which include 12,000 regional entries.

    A defining characteristic of the \acrshort{NZE} lexicon is the extensive borrowing from te reo Māori, with an estimated six out of every 1,000 words in speech being loanwords \citep{macalister_trends_1999}. These borrowings encompass toponyms (\gls{aotearoa}, \gls{aoraki}), flora and fauna (\gls{kowhai}, \gls{kiwi}), and social or material culture (\gls{marae}, \gls{whanau_mao}). Furthermore, greetings and salutations (\gls{kia_ora}, \gls{nga_mihi}) have become common in general \acrshort{NZE} usage \citep{maclagan_englishes_2020}. While early studies identified an active Māori vocabulary of approximately 80 words (excluding place names) among \acrshort{NZE} speakers \citep{macalister_weka_2006}, recent research suggests this number is likely higher due to shifting social attitudes \citep{de_bres_attitudes_2010}. Outside these indigenous influences, \acrshort{NZE} has been significantly shaped by contact with Australian English during the 19\textsuperscript{th} century.

    As of March 2025, the Oxford English Dictionary Online includes 2,962 entries for Australasian English, 1,404 of which are coded specifically as \acrshort{NZE}. These entries result from various word-formation processes, including abbreviations, hypocoristics, compounding, and conversion. To provide a deeper overview of these features, I describe specific word-formation processes in the Appendix \ref{app:nze}: New Zealand English Features.

\subsubsection{Morphosyntactic Features}
\label{review:morphosyntax}

    In contrast to its distinctive lexicon, there appear to be few morphosyntactic constructions that are unique to \acrshort{NZE} \citep{bauer_grammatical_2007}. Early research often relied on benchmarking \acrshort{NZE} features against British and American (and occasionally Australian) standards \citep{hundt_new_1998}. However, \citet{bauer_grammatical_2007} took a different approach to profiling the variety by identifying what it lacks compared to other Englishes, specifically in terms of aspect marking and noun-verb agreement. These absent features include habitual-\textit{be} (\ref{aspect:1}a), found in Irish English and \acrshort{AAVE}; perfect-\textit{be} (\ref{aspect:1}b), seen in Old English; the after-perfect (\ref{aspect:1}c) of Irish English; the done-perfect (\ref{aspect:1}d) of \acrshort{AAVE}; and the durative a-prefix (\ref{aspect:1}e) typical of Appalachian English.

    \vspace{-12pt}
    \begin{center}
    \singlespacing
    \begin{align}
        \textnormal{New Zealand English (Ungrammatical)} \label{aspect:1} \\
        \notag \\
        \textnormal{\texttt{*They (do) be working the mornin}} \tag{a} \\ 
        \textnormal{\texttt{*They are come already}} \tag{b} \\
        \textnormal{\texttt{*They are after crashing the car}} \tag{c} \\
        \textnormal{\texttt{*They done finished this}} \tag{d} \\ 
        \textnormal{\texttt{*They are a-singing}} \tag{e} \\
        \notag \\
        \eqname{\citep{bauer_grammatical_2007}}
    \end{align}
    \end{center}

    \begin{center}
    \singlespacing
    \begin{align}
        \textnormal{New Zealand English (Grammatical)} \label{aspect:2} \\
        \notag \\
        \textnormal{\texttt{They regularly work in the morning}} \tag{a} \\
        \textnormal{\texttt{They have come already}} \tag{b} \\
        \textnormal{\texttt{They have just crashed the car}} \tag{c} \\
        \textnormal{\texttt{They have already finished this}} \tag{d} \\
        \textnormal{\texttt{They are singing}} \tag{e} \\
        \notag \\
        \eqname{\citep{bauer_grammatical_2007}}
    \end{align}
    \end{center}
    \vspace{12pt}

    By defining \acrshort{NZE} through these absences, \citet{bauer_grammatical_2007} suggests that the variety can be conceptualised as a specific subset of possible English grammatical constructions. The standard \acrshort{NZE} equivalents for these forms are provided in (\ref{aspect:2}a--e). Despite this perceived lack of uniqueness, a robust body of literature documenting the morphosyntactic character of \acrshort{NZE} has developed (\citealp{bauer_new_1987, quinn_variation_1995, hundt_new_1998, quinn_variation_2000, bauer_grammatical_2007, hay_new_2008}). Drawing on these studies, I provide a detailed description of the morphological and syntactic features associated with \acrshort{NZE} in Appendix \ref{app:nze}: New Zealand English Features.

\subsection{Languages, Dialects, Accents}
\label{review:variation_nze}

    The most comprehensive corpus-based study on morphosyntactic variation in \acrshort{NZE} to date, conducted by \citet{hundt_new_1998}, famously noted that the variety is ``virtually indistinguishable'' \citep[p.139]{hundt_new_1998} from Australian English. This perceived lack of distinguishing grammatical features has been re-asserted by \citet{hay_new_2008} and reiterated by \citet{trudgill_international_2017}, who identified only three specific grammatical features - two of which are shared with Australian English. This overlap is so pronounced that \citet{kiesling_english_2020} did not list a single unique grammatical feature in a broad overview of English across Australia and New Zealand. Yet, while \acrshort{NZE} maintains a ``reputation for being extremely homogenous'' \citep[p.169]{bauer_can_2002}, internal variation does exist. In the following sections, I describe the varieties associated with specific ethnic groups (ethnolects), geographic regions (regiolects), and other distinct social groups.

\subsubsection{Ethnolects}

    There are two well-attested varieties of English spoken in New Zealand: Māori English and \acrshort{NZE} \citep{stubbe_talking_2000}. Māori English has been described both as a distinct variety of English and as an ethnolect of \acrshort{NZE} associated with speakers of Māori descent or identity \citep{maclagan_englishes_2020}. While early interactions between Māori and European colonists in the 18\textsuperscript{th} century occurred primarily in te reo Māori \citep{maclagan_maori_2008}, Māori English subsequently followed a developmental trajectory similar to other New Zealand varieties as a product of intensive language contact \citep{degani_language_2012}. Crucially, however, Māori English was also the result of a language shift driven by the forced displacement of te reo Māori within Māori communities \citep{maclagan_maori_2008}. Additionally, \citet{maclagan_maori_2008} notes the development of Pasifika English, associated with Pasifika ethnic groups, which superficially shares several linguistic characteristics with Māori English.

\subsubsection{Regiolects}

    \citet{bauer_playground_2003} examined the language of children - or playground talk - noting significant lexical variation in primary schools across the country. Through surveys sent to 150 primary schools, they elicited language use across various settings, including greetings, farewells, and play-based expressions. This study identified three broad dialect zones - the Northern, Central, and Southern regions - which were further divided into eleven subregions. Among these, the most well-documented regional variety is the Southland dialect, distinguished primarily by its rhoticity \citep{bartlett_regional_1992}. Additionally, research has suggested the existence of a distinct variety in the Taranaki region \citep{ainsworth_regional_2004}, while more recent scholarship has proposed the emergence of Multicultural Auckland English \citep{ballard_new_2025}.

\subsubsection{Other Sociolects}

    Other forms of linguistic variation within \acrshort{NZE}, primarily in the lexicon, are often constrained to specific semantic domains. Numerous lexical items are unique to the rural sector of New Zealand \citep{bardsley_specialist_2006}, with other specialised domains including agriculture, sports, and horse racing \citep{hay_new_2008}. For example, lexical features specific to rugby and related team sports include \textit{boilover} (an unexpected sporting outcome), \textit{to seagull} (to run in the backline), and \textit{dropie}, a clipped hypocoristic form of \textit{drop goal}.

\subsection{Attitudes and Ideologies}

    To conclude this review of the New Zealand context, I examine how language attitudes and ideologies - mediated by identity - have influenced linguistic variation and change. Within the popular consciousness, \acrshort{NZE} is still frequently reduced to the concept of the New Zealand accent'' \citep{bayard_antipodean_1991, donald_its_2018, de_bres_sexiest_2021}. Public interest in the variety was famously sparked by Ian Gordon’s article, Hunting New Zealandisms,'' which advocated for the inclusion of regional lexical features in the \textit{Oxford English Dictionary} \citep{gordan_hunting_1957}. This advocacy emerged during a period of burgeoning national identity; however, early research often continued to benchmark \acrshort{NZE} against British or American standards, rarely treating it as a distinct variety of interest \citep{hundt_new_1998}. In many ways, the \textit{cultural cringe} - the internalised inferiority of local culture - permeated early research practices.
    
    Historical records, including school inspector reports and periodicals, reveal that the emerging New Zealand variety was once highly stigmatised \citep{abell_this_1990}. During the mid-20th century, the `Queen’s English' and the `standard' speech of the BBC served as the prestige models to which subjects across the Commonwealth calibrated their speech \citep{bell_inventing_1996}. Local television and drama were often perceived as amateur simply because they sounded ``local'' \citep[p.22]{bell_inventing_1996}. Associated with the image of New Zealand as merely ``an English farm in the South Pacific'', the local variety was deemed undesirable, leading to the common practice of broadcasters receiving elocution lessons to reduce their accents as late as the 1980s \citep{bell_inventing_1996}.

    These linguistic ideologies remain relevant today. As of November 2023, the Sixth National Government and its coalition partners have proposed making English a \textit{de jure} official language of New Zealand, partly in response to perceived threats to national identity. This move is arguably driven by a perception that \acrshort{NZE} is being eclipsed by dominant Inner Circle varieties, such as American English, as well as the increasing influence of Outer Circle varieties and non-native learners. Ultimately, these entrenched attitudes and modern political anxieties demonstrate that user perceptions are not merely secondary observations; they are the primary ideological drivers that determine which linguistic features are enregistered, preserved, or abandoned within the community.

\section{Chapter Summary}

    This review has traced the evolution of dialectology from traditional \textit{spatial} models to modern \textit{placial} frameworks, highlighting the role of user perception and identity in the enregisterment of linguistic varieties. While the technical capabilities of \acrshort{NLP} have enabled the analysis of social media at an unprecedented scale, a significant gap remains in our understanding of how these digital \textit{places} function as coherent speech communities, particularly within the New Zealand context.
    
    Existing research has largely treated social media as a proxy for physical location, often overlooking the ideological drivers - such as the \textit{cultural cringe} or national identity anxieties - that influence how users perform their identity online. By applying high-dimensional modelling \acrshort{C2xG} approach to the New Zealand Reddit ecosystem, this thesis addresses these gaps, moving beyond simple text classification to explore the deep grammatical and semantic structures that signal a shared \textit{sense of place}. The socio-historical trajectory of \acrshort{NZE} - moving from a stigmatised \textit{farm} variety to a core component of national identity - demonstrates that linguistic variation is not merely a matter of geographic distance, but a product of complex social attitudes and ideologies. This complexity justifies the multi-method pipeline employed in this thesis.
    
    Because \acrshort{NZE} is often described as \textit{homogenous} or \textit{indistinguishable} in its formal grammar (\citealp{bauer_grammatical_2007}; \citealp{kiesling_english_2020}), a standard text-classification approach alone would likely fail to capture the subtle nuances of the variety. By combining discourse and thematic analysis (Phase 1) with variable-based frequency distributions (Phase 2), I am able to identify \textit{user-informed} sociolinguistic variables that reflect actual community perceptions rather than top-down assumptions. Furthermore, the integration of high-dimensional embeddings (Phase 3) and \acrshort{C2xG} (Phase 4) allows for the detection of latent patterns in the lexico-grammar that traditional dialectology has historically overlooked.
    
    Ultimately, this multi-method approach ensures that the analysis of \textit{place} on Reddit is not reduced to simple coordinates, but is instead treated as a robust, enregistered performance of identity. By bridging the gap between qualitative sociolinguistic theory and quantitative \acrshort{NLP} techniques, this research provides a comprehensive account of how geographic dialect alignment manifests in the digital age.

% -----------------------------
% Chapter 4: Corpus Dimensions
% -----------------------------

\chapter{Corpus Dimensions}
\markboth{Corpus Dimensions}{}
\label{chap:corpus_dimensions}

\section{Chapter Outline}
\label{corpus:chapter_outline}

    In this chapter, I outline the corpus characteristics and dimensions of my primary dataset: Reddit. Following a brief introduction in Section \ref{corpus:intro}, I describe the platform's situational characteristics (Section \ref{corpus:reddit_characteristics}) before providing a high-level overview of the data processing procedures employed to refine the corpus for analysis (Section \ref{corpus:data}).

\section{Reddit: the Front Page of the Internet}
\label{corpus:intro}

    Reddit was founded in June 2005 by Steve Huffman and Alexis Ohanian, with Aaron Swartz joining later that year. As of late 2024, the platform hosts over 100,000 active communities and averages approximately 97.2 million daily active users. A central feature of the platform is the karma system, a reputation metric where users, or Redditors, earn points through community engagement \citep{lagorio-chafkin_we_2018}. Users are incentivised to contribute through `Post karma' and `Comment karma', which can unlock specific achievements or grant access to restricted subreddits. Unlike platforms that rely on explicit Global Positioning Systems data, Reddit provides implicit georeferenced data through language used within its various place-based communities.

    \citet{panek_understanding_2022} established a typology of Reddit communities based on their specific purpose (summarised in Table \ref{tab:reddit_typology}). Of particular interest are place-based communities - subreddits dedicated to a specific country, region, or city. While the platform-wide reddiquette provides a general code of conduct, it also contains specific expectations regarding linguistic behaviour. For instance, a long-standing norm encourages the use of ``proper grammar, capitalization, and spelling'' to maintain discourse quality. Users are further discouraged from using `time-sensitive' words like `breaking' in titles or making low-effort comments that `lack content'. Beyond these platform-wide guidelines, users must also adhere to specific rules established by community moderators, known as \glspl{mod}, who enforce local norms within each subreddit.

    \begin{table}
        \scriptsize
        \centering
            
            \caption{Typology of Reddit Communities}
            \label{tab:reddit_typology}
            
        \renewcommand{\arraystretch}{1.4}
            \begin{tabular}{lp{0.7\linewidth}}
            \hline
            \textbf{Category} & \textbf{Definition} \\
            \hline
            \\
            Spectacle & A community where submission posts are typically images or short videos with minimal context provided in the post title. A subtype of submission posts are intended to evoke a sense of moral judgement. \\
            \acrshort{AAF} & An abbreviation of \acrfull{AAF}, this is a community concerned with media texts (including books, music, movies, video games) and other spectator experiences (such as sports). \\
            The Public Sphere & A community focussed on matters of public or societal concern such as current events often specific to the socio-political context of the United States. \\
            PEOSC & An abbreviation of Personal experience, opinion sharing, and conversation, this is a community concerned with personal experiences or opinions of users. \\
            Educational & A community serving a similar function as an internet forum to exchange information, links to resources, and for learning or developing skills in a particular domain. \\
            Creativity & A community dedicated to showcasing creative works and to discuss the process of creativity. \\
            Place-based & A community dedicated to particular countries, regions, or cities. \\
            \\
            \hline
            \end{tabular}

            \tablenoteparagraph{\textbf{Table Notes}: This table outlines the typology of Reddit communities as established by \citet{panek_understanding_2022}. Columns provide the specific classification for each type of community (\texttt{Category}) alongside its corresponding functional or thematic \texttt{Definition}. All categorisations are sourced from \citet{panek_understanding_2022}.}
        
    \end{table}

\subsection{Why Reddit?}

    Over-reliance on a single data source, such as Twitter\textsuperscript{X}, is problematic and often results in findings that are unrepresentative of language use across diverse registers and contexts \citep{pechenick_characterizing_2015}. Despite its potential, linguistic engagement with Reddit remains surprisingly low. Between 2011 and 2023, only 1,285 articles and books focused on Reddit as a research site \citep{lendvai_reddit_2025} - a mere fraction of the scholarship dedicated to Twitter\textsuperscript{X} during the same period \citep{mir_rise_2023}. A bibliometric analysis further illustrates this disparity, finding that out of 727 Reddit-focused manuscripts, only eight were relevant to linguistics \citep{proferes_studying_2021}. One such study tracked the monthly relative frequency of word forms over a year to analyse lexical emergence \citep{mahler_lexical_2020}. While \citet{hamre_geographic_2024} recently conducted a comprehensive linguistic study on Reddit, the findings were limited to North American and British varieties.
    
    The reasons for this limited engagement are likely twofold. Historically, Reddit’s relatively small initial user base compared to its contemporaries may have hindered its adoption as a primary research site \citep{lendvai_reddit_2025}. Additionally, the field’s historical preference for absolute physical locations over implicit geographic information (such as community names) may have sidelined Reddit in dialectological research. However, there is an increasing recognition of the value of implicit geography, as it accounts for the socially constructed perspectives of space that are central to modern sociolinguistic theory \citep{nguyen_dialect_2021}.
    
\subsection{New Zealand Reddit}

    Reddit was first introduced to the New Zealand digital landscape with the creation of \texttt{r/newzealand} on 23 March 2008, nearly three years after the platform's global launch in June 2005 \citep{lagorio-chafkin_we_2018}. According to a 2024 survey of active internet users in New Zealand, 7\% of the population uses Reddit daily \citep{internetnz_internetnz_2025}. While this represents approximately half the daily activity of platforms such as TikTok (14\%) and Snapchat (13\%), \texttt{r/newzealand} remains the dominant place-based community for the country, ranking within the top 1\% of all subreddits globally and serving as the third most popular community in Oceania after \texttt{r/australia} and \texttt{r/sydney}. Since its inception, \texttt{r/newzealand} has cultivated a distinct national identity through its visual and social markers. The community's avatar - or `snoomoji' - features an outline of a Kiwi with red eyes and an antenna. The red eyes are a direct reference to the `Laser Kiwi' flag design from the 2015–2016 New Zealand flag referendums, while the antenna identifies the character as a variation of Snoo, the global Reddit mascot.

\section{Situational Characteristics}
\label{corpus:reddit_characteristics}

    This research conceptualises social media platforms, such as Reddit and Twitter\textsuperscript{X}, as specialised registers of language use. A register is defined as a variety of natural language that occurs within a specific situation of use \citep{grieve_sociolinguistic_2025}. Furthermore, register differences exist within these platforms themselves, driven by varying production circumstances and communicative purposes. To determine how these platforms function as self-contained registers, I employ the Situational Characteristics of Registers and Genres framework \citep{biber_register_2009}. This model consists of seven situational characteristics: participants, relations among participants, channel, production circumstances, setting, communicative purposes, and topic.

\subsection{Participants}

    As previously established, the primary participants in communities like \texttt{r/newzealand} are the users. Interaction is inherently fluid, meaning each user has the potential to function as both the addressor and the addressee. A user who initiates a submission - the \acrfull{OP} - acts as the addressor with the intent of reaching other users. When these users respond, their roles shift from addressee to addressor. The scale of this participant base is significant; as of May 2025, \texttt{r/newzealand} hosted 786,000 subscribers, representing a 22.81\% increase from the 640,000 recorded in August 2024. However, these figures do not account for the high volume of unregistered visitors or `on-lookers'. Users who consume content without actively posting or subscribing are known as lurkers. Additionally, some participants may adopt temporary `burner' accounts to maintain anonymity when sharing sensitive or personal content.

\subsection{Relations Among Participants}
    
    \texttt{r/newzealand} is maintained by a team of voluntary moderators, or \glspl{mod}, who are responsible for establishing and enforcing the community's specific norms \citep{panek_understanding_2022}. These moderators possess the authority to warn, temporarily ban, or permanently remove users and content that breach established rules. Reflecting the scale of these responsibilities, moderator teams are often sized proportionally to their community; for example, \texttt{r/science} maintains a massive team of over 1,500 moderators to manage its 34.3 million members as of May 2025. 

    In \texttt{r/newzealand}, the community is governed by a set of 11 rules that differentiate between unacceptable behaviours (such as bigotry and low-effort posts) and ideal participation. Of particular significance to the register is the first rule, which mandates that all submissions must ``directly relate to New Zealand''. This requirement ensures that content remains relevant to the local \textit{place}, even excluding New Zealand-sourced content if the subject itself is not applicable to the national context. 

    Beyond these local regulations, users are subject to reddiquette - a site-wide portmanteau of `Reddit' and `etiquette' - which outlines expected conduct for voting, commenting, and submissions. This environment fosters a unique knowledge-sharing dynamic where moderators act as facilitators rather than necessarily being subject-matter experts, and commenters often contribute more specialised domain knowledge than the \acrshortpl{OP} \citep{panek_understanding_2022}.

\subsection{Channel}

    Users on \texttt{r/newzealand} primarily interact via the website (\href{https://www.reddit.com/}{reddit.com}) or the mobile application. While both channels provide access to the same core content, they differ in appearance and layout. Despite these interface variations, the community homepage remains highly customisable, allowing moderators to localise the space by including relevant bookmarks. The following resources are currently integrated into the \texttt{r/newzealand} interface:
    
    \begin{itemize}
        \item \textbf{Mental Health Help}: This bookmark redirects users to a moderator-authored post containing a comprehensive list of New Zealand-based mental health support services and resources. Before being archived, the submission garnered 275 votes and 167 comments, reflecting its significance to the community.
        \item \textbf{Contact the Mods}: This link initiates the Reddit messaging service, allowing users to private message the moderator team. This function is primarily used to report content or users that breach community rules, with a drop-down subject line provided to categorise the specific nature of the report.
        \item \textbf{Search and Filter}: The remaining bookmarks, \textit{Search by Post Flair} and \textit{Hide Posts Flared}, enable users to filter the subreddit's feed based on specific categories. These \glspl{flair} serve as a primary navigational tool, which I discuss further under the situational characteristic of \textbf{Topic}.
    \end{itemize}

\subsection{Production Circumstances}

    Production circumstances on Reddit are categorised into two primary phenomena: post submissions and comments. While any registered user may post on \texttt{r/newzealand}, specific requirements must often be met to engage with politically sensitive content. Users may respond to submissions by up-voting, down-voting, commenting, or awarding; these actions generate engagement insights, though total view counts remain visible only to the \acrshort{OP} and \glspl{mod}. Submissions are further divided into several distinct types:
    
    \begin{itemize}
        \item \textbf{Post}: The standard submission consists of a title (limited to 300 characters), a \acrshort{URL}, and optional body text (up to 10,000 characters). A preview of the linked content typically appears alongside the text. Users are encouraged to apply a \gls{flair} to group submissions by topic or to provide content warnings, placing the onus of self-censorship on the subscriber.
        \item \textbf{Selfpost}: A selfpost is a submission without external links, comprising a title, body text, and relevant \glspl{flair} or tags (e.g., \textit{Spoiler}). The body text, or `selftext', is subject to a 10,000-character limit.
        \item \textbf{Image and Video Posts}: These follow the standard submission structure but feature an uploaded media file in place of body text. While various formats are supported (e.g., .png, .webp, .gif), many users prefer to link to third-party hosting sites like Imgur, which was originally developed to meet Reddit’s specific media needs \citep{lagorio-chafkin_we_2018}.
        \item \textbf{Poll}: Poll posts allow users to create time-limited surveys with up to six options. Guidelines suggest providing clear, concise options and a defined voting duration of between one and seven days.
        \item \textbf{\acrfull{AMA}}: the \acrshort{AMA} is a highly interactive format where guests must often verify their identity - typically via a `selfie' with a handwritten note - before engaging with the community in real-time.
    \end{itemize}
    
    The second production phenomenon consists of comments posted by registered users. Comments may contain text, images, or \acrfullpl{GIF} and are subject to the same interactive actions and moderation as top-level posts. These interactions feed into sorting algorithms that categorise comments as Best, Top, New, or Controversial, thereby increasing the relevance and visibility of high-quality or timely discourse within a thread.

\subsection{Setting}

    Neither the time nor the place of communication is shared by users, which characterises the setting as primarily asynchronous. While submissions and comments are typically displayed in chronological order (facilitating near real-time interaction), they are indexed based on \acrfull{UTC}. Furthermore, the visibility of these posts is often subject to Reddit's internal recommender systems and algorithmic sorting. Moderators may also choose to delay or remove content that fails to meet community requirements. Finally, the ability for users to revise, edit, or delete their existing submissions and comments adds a layer of post-production malleability to the register that distinguishes it from more ephemeral forms of \acrshort{CMC}.
    
\subsection{Communicative Purposes}  

    The communicative purpose on Reddit is largely dependent on the intended audience of each individual community. For example, \texttt{r/reddit} serves as an official broadcast channel for platform administrators, whereas \texttt{r/newzealand} functions as a place-based community dedicated to topics concerning New Zealand \citep{panek_understanding_2022}. This purpose is explicitly stated in the subreddit’s headline - ``Aotearoa | New Zealand: Tomorrow's Headlines Today!'' - and its tagline - ``r/NewZealand, this is New Zealand Today''. While \texttt{r/newzealand} serves as the primary national forum, it is part of a broader ecosystem of New Zealand-centric subreddits. Numerous special interest communities are promoted within \texttt{r/newzealand} to cater to specific regional or topical needs; a summary of these promoted communities is presented in Table \ref{tab:promoted_communities}.

    \begin{table}
        \scriptsize
        \centering
            
            \caption{Promoted New Zealand-Related Communities on Reddit}
            \label{tab:promoted_communities}
            
            \renewcommand{\arraystretch}{1.4}
            \begin{tabularx}{\linewidth}{l*{3}{>{\centering\arraybackslash}X}}
                \toprule
                \textbf{Community} & \textbf{Category} & \textbf{Created Date} & \textbf{Members} \\
                \midrule
                \addlinespace[1em]
                \texttt{r/auckland} & Place-based & Nov 22, 2009 & 228,924 \\
                \texttt{r/thetron} & Place-based & Feb 17, 2011 & 43,053 \\
                \texttt{r/hawkesbay} & Place-based & - & 3,601 \\
                \texttt{r/palmy} & Place-based & - & 3,601 \\
                \texttt{r/Wellington} & Place-based & Sep 2, 2010 & 124,348 \\
                \texttt{r/Nelsonnz} & Place-based & - & 2,422 \\
                \texttt{r/chch} & Place-based & Feb 14, 2011 & 66,073 \\
                \texttt{r/queenstown} & Place-based & - & 23,844 \\
                \texttt{r/dunedin} & Place-based & Sep 6, 2010 & 36,069 \\
                \texttt{r/AveragePicsOfNZ} & Spectacle & Jul 8, 2018 & 83,736 \\
                \texttt{r/diynz} & Educational & Feb 6, 2019 & 50,140 \\
                \texttt{r/LegalAdviceNZ} & Educational & Dec 2, 2018 & 37,152 \\
                \texttt{r/MapsWithoutNZ} & Spectacle & Sep 16, 2015 & 126,440 \\
                \texttt{r/PersonalFinanceNZ} & Educational & Jul 6, 2015 & 126,158 \\
                \addlinespace[1em]
                \bottomrule
            \end{tabularx}
    
            \tablenoteparagraph{\textbf{Table Notes}: This table categorises special interest communities associated with New Zealand, detailing the community name (\texttt{Community}), its creation date (\texttt{Created Date}), and subscriber count as of June 2025 (\texttt{Members}). Each community is classified by type (\texttt{Category}) according to the typology of Reddit communities established by \citet{panek_understanding_2022}. The data indicates that the majority of these special interest groups are additional place-based communities; all data is sourced from Reddit.}
            
    \end{table}
    
\subsection{Topic}
    
    \glspl{flair} are established by moderators, and \acrshortpl{OP} are encouraged to categorise each submission accordingly. Users can subsequently filter or block content based on these designations. In the case of \texttt{r/newzealand}, there are 22 distinct \glspl{flair} (including \textit{Politics}, \textit{News}, \textit{Discussion}, and \textit{Advice}) alongside three site-wide tags: \acrshort{NSFW} (mature content), \textit{Spoilers}, and \textit{Brand Affiliate}. One notable category is the Shitpost, which refers to provocative submissions designed to elicit a reaction; however, these are often misinterpreted at face value by unaware users. While encouraged, the use of \glspl{flair} remains an imperfect solution for topic categorisation, as submissions frequently overlap multiple categories and \glspl{flair} are subject to subjective interpretation or removal.
    
    Several \glspl{flair} are specifically rooted in the New Zealand cultural context, such as \gls{maoritanga} and \gls{kiwiana}. Particularly relevant to the construction of place identity is the \textit{Meta} \gls{flair}, used for discussions regarding \texttt{r/newzealand} itself \citep{singer_evolution_2014}. The presence of this \gls{flair} suggests that users are acutely aware that while the subreddit is a place-based community, it functions as a distinct social construct rather than a literal extension of the physical country. These self-referential discourses are part of a broader platform-wide phenomenon known as Metareddit, or the ``Theory of Reddit'' \citep{hagen_no_2023}. In Chapter \ref{chap:construction_grammar}, I utilise the \texttt{r/NZMetaHub} to further investigate the structure of this wider New Zealand Reddit ecosystem.
    
\section{Sources of Data}
\label{corpus:data}

    My primary source of Reddit data is derived from the Pushshift Dumps, an ongoing data collection effort targeting the top 40,000 communities on the platform via the Pushshift \acrshort{API} \citep{baumgartner_pushshift_2020}. This data is hosted in the Academic Torrent repository (the \texttt{Watchful1} database), which became a vital resource following the 2023 revision of Reddit’s Developer Platform and Data \acrshort{API} access rules. My baseline corpus is structured into three tiers: (1) 12 country-level communities used for the cross-national analyses in Chapters \ref{chap:user_variables} and \ref{chap:dialect_classification}; (2) six city-level communities utilised in Chapters \ref{chap:dialect_classification} and \ref{chap:construction_grammar}; and (3) a network of 33 New Zealand-related and 14 peripheral communities specifically examined in Chapter \ref{chap:construction_grammar}.

\section{Data Processing}

    I applied several general data processing steps to the Reddit data, including the removal of all observations that were deleted by users or removed by \glspl{mod}. I also removed all duplicate entries from the final dataframe. To align the data with the Situational Characteristics framework \citep{biber_register_2009}, I categorised the linguistic data into four distinct text-types based on their production circumstances: post submission titles (\gls{rpost}), selfpost titles (\gls{rstitle}), selfpost body texts (\gls{rstext}), and comments (\gls{rcomm}). These categories were determined by filtering for titles associated with external media \acrfullpl{URL} (\gls{rpost}) versus those associated with internal `selftext' bodies (\gls{rstitle} and \gls{rstext}). Comments (\gls{rcomm}) were processed as a separate dataset.
    
    Following these initial processing steps, the final dataset for the 12 country-level communities comprised 98,116,389 observations. This followed the removal of 9.2\% ($n=10,046,750)$  of deleted and duplicated entries from an initial total of 108,163,139. For the New Zealand-related communities, deleted and duplicated observations accounted for 6.3\% ($n=1,551,075)$ of the raw data, resulting in a final corpus of 23,208,051 observations across 33 subreddits. In both datasets, comments represented the most numerous text-type.

\section{Chapter Summary}

    This chapter has established the corpus dimensions and technical framework of the research by situating Reddit as a primary site for geolinguistic enquiry. By applying the Situational Characteristics framework, I have defined the Reddit ecosystem not merely as a data source, but as a collection of specialised registers governed by unique production circumstances and community norms.

% -----------------------------
% Chapter 4: User Intuitions and Place Identity
% -----------------------------

\chapter{User Intuitions and Place Identity}
\markboth{User Intuitions and Place Identity}{}
\label{chap:user_intuitions}

\section{Chapter Outline}
\label{user:chapter_outline}

    This chapter consists of two qualitative analyses to determine language use in context: discourse analysis and thematic analysis. I provide a background and my motivation in Section \ref{user:introduction} and my corpus sampling strategy in Section \ref{user:sampling_strategy}. I then apply a discourse analysis framework to two language-related selfposts in Section \ref{user:discourse_analysis}, followed by an interim summary in Section \ref{user:interim_summary}. In Section \ref{user:content_analysis}, I use thematic analysis to determine a list of user-informed features from the comments associated with the two language-related selfposts. I conclude the chapter with a discussion synthesising the findings from my two qualitative approaches in Section \ref{user:discussion} and describe the key findings in Section \ref{user:conclusion}. The findings from this chapter contribute to \acrshort{SQ1}: Do users in place-based communities associate language use with a place identity?

\section{Background and Motivation}
\label{user:introduction}

    Of interest to me were the place-based communities on Reddit which are communities dedicated to particular countries, regions, or cities \citep{panek_understanding_2022}. As a form of georeferenced social media language data, these place-based communities offer implicit geographic information about the users \citep{marti_social_2019}. One way to conceptualise place-based communities on Reddit is that they are communities of practice \citep{leuckert_towards_2020}. A \textit{community of practice} is defined as a group of people who share a common interest who develop shared knowledge and practices through regular mutual engagement \citep{eckert_communities_2005}. In the case of \texttt{r/newzealand}, users abide by a set of community rules (such as ``Submissions must directly relate to New Zealand''). Since Reddit makes up only a small proportion of users in New Zealand's social media landscape \citep{internetnz_internetnz_2025}, I need to first determine if there are pre-existing perceptions and ideologies towards the role of language use in a place-based community associated with New Zealand.
    
    Ultimately, I am interested in how language is used to construct and shape users' perception of \texttt{r/newzealand} as a place-based community and how language is used as a place-making activity \citep{johnstone_place_2004}. One methodological approach is discourse analysis \citep{gee_introduction_2005}. Computational discourse analysis often involves techniques such as sentiment analysis \citep{ferrer_discovering_2021} or topic modelling \citep{stine_comparative_2020} based on the representative context of the language samples \citep{dourish_what_2004}. Some criticisms of corpus-assisted discourse analysis are that it risks disconnecting the producers of language from their discourses \citep{matheson_discourse_2023}. Conscious that the successive phases of my research are dominated by corpus-based approaches that obfuscate the social resources of users, I initiate my analysis by examining the interactional context of my place-based community \citep{seaver_nice_2015}.
    
    Except for a handful of unpublished theses (\citealp{merrit_analysis_2012}; \citealp{birznieks_perpetuation_2020}; \citealp{desmarais_men_2020}; \citealp{zoric_constructing_2024}), Reddit is seldom used as a source of discursive enquiry in linguistics. Whereas corpus-assisted approaches offer some privacy by aggregating user-generated language, methods and approaches to discourse analysis do raise complex ethical concerns about consent and anonymity (\citealp{adams_scraping_2022}; \citealp{gliniecka_ethics_2023}; \citealp{rocha-silva_passive_2024}). This does not mean Reddit is off-limits, as some researchers have adapted existing methods to address these concerns. For example, Reddit has become a primary source in medical internet research with a focus on sensitive topics such as mental health and substance use \citep{lendvai_reddit_2025}.
    
    Qualitative approaches such as thematic analysis \citep{braun_using_2006} allow me to examine and synthesise sensitive data while maintaining the privacy of users \citep{ison_i_2025}. An example of this was \citet{tarnarutckaia_myth_2020}, who analysed user comments in \texttt{r/dragonage} to examine user attitudes towards the speech of French and French-accented non-player characters. The findings suggest Reddit is a rich source of perceptual language data observed within its social context.

    One assumption made when using georeferenced social media data is that this form of language is not only representative of users \citep{sloan_who_2015}, but also that users are motivated to use language that reflects their geographic dialect \citep{staehr_dialect_2019}. Users may modulate their language use as a result of their own attitudes and ideologies. In the case of \acrshort{NZE}, it is possible users may deliberately modify their New Zealand accent or lexical choices in digital settings as a result of pre-existing attitudes such as cultural cringe (\citealp{bayard_antipodean_1991}; \citealp{donald_its_2018}; \citealp{de_bres_sexiest_2021}). Therefore, the goals of the current chapter are: a) to examine the language use of place-based communities in its social context; and b) to identify linguistic features users associate with \acrshort{NZE}.

\section{Sampling Strategy}
\label{user:sampling_strategy}

    \begin{table}
        \scriptsize
        \renewcommand{\arraystretch}{1.4}
        \centering
        \caption{\label{tab:sampling_strategy} Language-related submissions on r/newzealand as of May 2025.}
        \begin{tabular}{p{0.2cm}p{9cm}lcc}
            \hline 
             & Submission & Created & Score & Comment \\
            \hline 
            \\
            \textbf{English} \\
            (a) & \href{https://www.reddit.com/r/newzealand/comments/x13piy/}{Biggest NZ urban legends: `English is not an official language'} & 3y ago & 83 & 156 \\
            (b) & \href{https://www.reddit.com/r/newzealand/comments/1cvyl7c/}{Teaching English Overseas} & 1y ago & 3 & 4 \\
            (c) & \href{https://www.reddit.com/r/newzealand/comments/cszdd3/}{How much longer will New Zealand English, as a distinct dialect, last for?} & 6y ago & 21 & 97 \\
            (d) & \href{https://www.reddit.com/r/newzealand/comments/155d1f3/}{What changes have you noticed in NZ English in your lifetime?} & 2y ago & 113 & 703 \\
            \\
            \textbf{language} \\
            (e) & \href{https://www.reddit.com/r/newzealand/comments/smdk6k/}{When I talk to my kids I worry that our NZ identify is being weakened by exposure to so much US / UK / Aus content, how do protect our language?} & 3y ago & 256 & 504 \\
            (f) & \href{https://www.reddit.com/r/newzealand/comments/1bcj86o/}{Maori language on tv needs to be subtitled} & 1y ago & 392 & 108 \\
            (g) & \href{https://www.reddit.com/r/newzealand/comments/o6n4s9/}{Why do you think old people feel so threatened by the use of Maori language being used in the media and with place names etc? The amount of people against it is a lot more than I thought. Struggling to see their side of reasoning..} & 4y ago & 235 & 426 \\
            (h) & \href{https://www.reddit.com/r/newzealand/comments/16vq76h/}{Wanting to learn sign language. \acrshort{NZSL} or ASL?} & 2y ago & 3 & 4 \\
            \\
            \textbf{accent} \\
            (i) & \href{https://www.reddit.com/r/newzealand/comments/1aufjbf/}{I just realized how much I lost my accent :(} & 1y ago & 444 & 173 \\
            (j) & \href{https://www.reddit.com/r/newzealand/comments/g9xksm/}{The New Zealand accent broken down. I mean, is he wrong? [Grinning Face with Sweat emoji]} & 5y ago & 778 & 127 \\
            (k) & \href{https://www.reddit.com/r/newzealand/comments/xyg1ee/}{A familiar accent in Ukraine} & 3y ago & 570 & 107 \\
            (l) & \href{https://www.reddit.com/r/newzealand/comments/1ctigv5/}{Will having an American accent be professionally detrimental if I want to work in hospitality, as a tour guide, or other public facing role?} & 1y ago & 84 & 230 \\
            \\
            \textbf{sayings} \\
            (m) & \href{https://www.reddit.com/r/newzealand/comments/1102c68/}{What are New Zealand's corniest sayings?} & 2y ago & 118 & 595 \\
            (n) & \href{https://www.reddit.com/r/newzealand/comments/1102c68/}{Whats your favourite kiwi lingo/sayings?} & 12y ago & 18 & 63 \\
            (o) & \href{https://www.reddit.com/r/newzealand/comments/lkr0ng/}{Common 90s playground phrase (trigger warning)} & 4y ago & 19 & 71 \\
            (p) & \href{https://www.reddit.com/r/newzealand/comments/1gob4xo/}{How to decline saying a Karakia at work} & 7mo ago & 650 & 849 \\
            \\
            \textbf{slang} \\
            (q) & \href{https://www.reddit.com/r/newzealand/comments/zpmseu/}{Teach this Canuck some kiwi slang} & 2y ago & 0 & 42 \\
            (r) & \href{https://www.reddit.com/r/newzealand/comments/1itt2xo/}{Some Kiwi slangs} & 3mo ago & 0 & 30 \\
            (s) & \href{https://www.reddit.com/r/newzealand/comments/wat7ui/}{New Zealand Slang} & 3y ago & 61 & 68 \\
            (t) & \href{https://www.reddit.com/r/newzealand/comments/16nhrfl/}{What's everyone's thoughts on New Zealand culture being affected by American culture, e.g. various trends from youth to young adults, behaviour, cringe slang and vernacular, effects on mentality and personality from American media, movies, television} & 2y ago & 40 & 147 \\
            \\
            \hline
        \end{tabular}
    \end{table}

    My primary goal in this chapter is to establish a link between language use and place identity in \texttt{r/newzealand}. My main selection criteria are selfposts with high user engagement as judged by the number of comments. The purpose of these comments is to curate a list of user-informed features, which I will later use to evaluate my dialect models based on users' perceptions. I adopt passive data collection to analyse language use in its social context while respecting the privacy of users and avoiding the reveal of otherwise sensitive information \citep{lagorio-chafkin_we_2018}. The list of submission posts is presented in Table \ref{tab:sampling_strategy}.
        
    I filtered the post submissions in \texttt{r/newzealand} using Reddit's internal recommendation algorithm to identify relevant post submissions with the following keywords: \textit{English}, \textit{language}, \textit{dialect}, \textit{accent}, \textit{sayings}, and \textit{slang}. Over half of the post titles were structured as a question directed at \texttt{r/newzealand}. The top two most relevant submissions were (a) and (b) under \textit{English}, and the fourth most relevant submission was (g) under \textit{language}. I excluded the results for \textit{dialect} as there was significant overlap with the other keywords. Based on the scores alone, the post submission that received the most engagement was a submission which received 778 upvotes under \textit{accent} (j). This post was uploaded in April 2020 and featured a video reel from TikTok - a short-form online video-sharing platform - describing the vowel space of \acrshort{NZE}. The second most relevant submission, with a score of 650, was a selfpost from November 2024 under \textit{sayings} (p). This selfpost was tagged with a \textit{Restricted} \gls{flair} due to the political nature of the discussion.

    Based on this sample, language use was a topic of interest in \texttt{r/newzealand}. Users in \texttt{r/newzealand} facilitated discussions in terms of language variation (l) and change (d) in \acrshort{NZE}, as well as the status of te reo Māori (g). The community was also being used as a forum for users who are unfamiliar with \acrshort{NZE}, as in the case of one user who affiliated with Canadian `Canuck' English (q). One post submission (t) raised the concern of American culture portrayed through the media as a threat to New Zealand culture, including \textit{cringe slang and vernacular}. This reflects wider sociolinguistic anxieties regarding linguistic homogenisation and the perceived loss of local identity. As I am primarily interested in specific examples of language use, I conduct my discourse analysis on two open-ended discussion post submissions. The first (m) (\texttt{155d1f3}) was posted in December 2023. My second post submission (d) (\texttt{1102c68}) was posted in July 2023 and tagged with the \textit{Discussion} post \gls{flair}. I chose this selfpost because the discussion was centred on language variation and change.

\section{Discourse Analysis}
\label{user:discourse_analysis}

    Discourse analysis is the study of language-in-use \citep{gee_introduction_2005}. My application of this method is driven by the need to determine the interactional context of \texttt{r/newzealand} \citep{seaver_nice_2015}. In order to establish a link between language use and place identity on Reddit, I integrate analytical tools from discourse analysis to support my understanding of language use in \texttt{r/newzealand}. By focusing on how users negotiate meaning and identity through their interactions, I can identify the specific ways in which linguistic choices serve as markers of community membership. This approach allows for a more nuanced understanding of how the digital space of \texttt{r/newzealand} is constructed as a distinct social and geographic place through discursive practice.

\subsection{Methodology}

    I apply the Seven Building Tasks \citep{gee_introduction_2005} to my two submission posts of interest. In the case of longer post submissions, I use the \textit{Narrative Analysis Framework} \citep{labov_narrative_1967}. The Seven Building Tasks allow me to examine how users construct reality through language, specifically looking at how they build significance, activities, identities, relationships, politics, connections, and sign systems within the subreddit. By pairing this with Labov's framework for longer submissions, I can better track the structural development of user stories about language use, ensuring that the `point' or evaluation of their linguistic experiences is clearly identified within the social context of the community.
    
\subsubsection{Seven Building Tasks}
    
    \citet{gee_introduction_2005} proposed a discourse analysis framework to determine how people use language to construct reality in a given communicative situation. These tasks provide the components of an ``ideal discourse analysis'' to interrogate any given language-in-use \citep[p.10]{gee_introduction_2005}. The building tasks include: building significance; building activities; building identities; building relationships; building politics; building connections; and building sign systems and knowledge. Each task serves as a lens to examine how speakers or writers use linguistic cues to signal what is important, what social activity is taking place, and what kinds of identities or relationships are being enacted within the discourse.
    
    The focus of the first two building tasks is on language-in-use: building significance and building activities. Significance, also known as the situated meaning, refers to how and what different things mean in any given situation \citep{gee_situated_2014}. As an example, the situated meaning of rules within the context of Reddit is unique to each community. In the case of \texttt{r/newzealand}, users and \glspl{mod} (itself a situated meaning) have a common understanding of what is relevant or irrelevant to New Zealand, according to the first rule. 
    
    Building activities, also known as practices, refer to the activity or set of activities in any situation. In the case of Reddit, this would relate to the production circumstances which I outlined in Chapter \ref{chap:corpus_dimensions}: Corpus Dimensions. By establishing these significance and activity tasks, users are able to frame their linguistic contributions within a shared understanding of what the subreddit represents. This framing ensures that discussions remain socially and geographically grounded, as the activity of posting on \texttt{r/newzealand} is inherently tied to the significance of the New Zealand identity.
    
    The next three building tasks focus on the social context of the situation. Building identities presupposes that the people enacted within a communicative situation are recognised as consequential. Inversely, some identities are made irrelevant within a situation. Similarly, the focus of building relationships is on how the enacted relationships are recognised as consequential. Both building tasks correspond to Participants and Relations Among Participants in the situational characteristics of registers and genres framework \citep{biber_register_2009}. Lastly, building politics refers to the distribution of social goods. The primary social good on Reddit is engagement, which involves upvotes (including downvotes) and comments. Users may withhold social goods through disengagement or other communicative strategies, such as screenshotting \citep{van_der_nagel_networks_2018}. Within the context of a place-based community, these `politics' often revolve around the validation of local knowledge, where users who demonstrate authentic \textit{Kiwiness} are rewarded with higher engagement, while those perceived as outsiders may be penalised through downvoting or social exclusion.
    
    The penultimate building task, building connections, presupposes that in any situation, connections or disconnections are made within and between a situation, as well as across situations (intertextuality). In the case of post submissions, I would consider how the situation connects or disconnects throughout the comment thread, but also intertextual references such as memes, \acrshortpl{GIF}, or external \acrshortpl{URL}. The final component, building significance for sign systems and knowledge, presupposes that for any communicative situation, there are sign systems and various ways of knowing in which I operate and orient towards, and which I value or dis-value in certain ways. Of interest to me were the inherent ideologies towards language and identity across a post submission and its comment thread. As an example, the use of te reo Māori may or may not indicate a user's stance towards this particular sign system or way of knowing. This final task is crucial for understanding how certain linguistic varieties are prioritised over others, effectively creating a hierarchy of authentic knowledge within the \texttt{r/newzealand} community.
    
\subsubsection{Narrative Structure}

    The purpose of the \textit{Narrative Analysis Framework} \citep{labov_narrative_1967} is to outline the macrostructure of a narrative by defining the constituent narrative elements: orientation, complication, evaluation, resolution, and coda. \textit{Orientation}, also known as the \textit{setting}, orients the listener to the person, place, time, and behavioural situation of the narrative. \textit{Complication}, further broken down into \textit{catalyst} and \textit{crisis}, includes the series of events which contribute to the narrative. The purpose of the \textit{evaluation} in a narrative is to respond to some stimulus or to emphasise the strange and unusual character of a situation by providing a break between complicating action and the result. \textit{Resolution} is the result of the narrative and is often fused with the evaluation. Lastly, the \textit{coda}, a vestigial element, serves as a functional device for returning the perspective to the present moment. By identifying these elements within user posts, I can determine how personal anecdotes are used to legitimise claims about language change. For instance, a user might use an orientation to establish their `local' credentials before moving into a complication that highlights a perceived `incorrect' usage of a \acrshort{NZE} term. In this context, the evaluation often serves as the site where linguistic ideologies - such as the aforementioned cultural cringe - are most explicitly expressed.

\subsubsection{Idealised Lines}

    The smallest unit in my analysis is the idea unit organised into idealised lines \citep{gee_linguistic_1991}. An idea unit, based on pitch glide in spoken language, contains a single new piece of information; an idealised line, also known as a tone group \citep{halliday_introduction_2004}, only contains one central idea or focus. Within each narrative element are stanzas. These include one or more idealised lines which share a unitary topic or perspective. In the case of submission posts where there is a narrative macrostructure, my largest unit of analysis is not the narrative elements, but sub-narratives which I refer to as sub-stories. These sub-stories are interrelated narratives which contribute to the overall situation. This hierarchical approach allows for a granular examination of how individual pieces of information are layered to build complex social identities and arguments within the digital environment.

\subsubsection{Evaluation}

    As a tool of inquiry, discourse analysis provides the data to support a given qualitative hypothesis \citep{gee_introduction_2005}. Based on \acrshort{SQ1}, my aim is to determine whether users in place-based communities associate language use with place identity. Firstly, I predict users do associate their place identity with their language use. I also predict users are intuitively attuned to these linguistic features which they associate with \acrshort{NZE}. This hypothesis aligns with the concept of linguistic awareness, where speakers of a minoritised or distinct variety often exhibit a heightened consciousness of the features that differentiate their speech from a perceived global or colonial standard. By applying these discourse tools, I can move beyond simple observation and systematically map how users `claim' certain lexical or phonological variables as belonging to the New Zealand landscape. This process of association is a key component of place-making, where the digital subreddit is transformed from a generic web space into a culturally specific \textit{place} through the deliberate use of local linguistic markers.

\subsection{Selfpost 1}
\label{user:submission_post1}

    \begin{figure}
      \centering
      
            \includegraphics[width=0.8\textwidth]{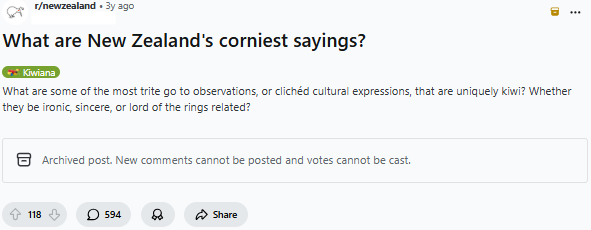}
            
      \vspace{6pt}
      \caption{Screenshot of Selfpost 155d1f3}
      \label{fig:selfpost_155d1f3}
      
    \end{figure}

    The first selfpost (see Figure \ref{fig:selfpost_155d1f3}), Selfpost 1 (\texttt{155d1f3}), was tagged with the \textit{\gls{kiwiana}} post \gls{flair}. This \gls{flair} describes the popular culture of New Zealand, including material culture (such as products, images, people, and food) and symbolic culture (including sayings and ways of life), and anything else that reflects the New Zealand lifestyle \citep{pickles_kiwi_2002}. I now apply the Seven Building Tasks \citep{gee_introduction_2005} to Selfpost 1 (\texttt{155d1f3}). As the selfpost was relatively short, containing no more than 30 words, it was not necessary to determine its narrative structure.

\paragraph{Building Significance}

    I begin with the selfpost title. The phrase \textit{New Zealand's corniest sayings} appeared to be important in this situation. This is because \textit{New Zealand} referred not only to the geopolitical entity, but also to the places and communities associated with the country. Similarly, \textit{sayings} had an extended meaning, referring to all linguistic features associated with New Zealand. As for the selfpost itself, \textit{observations} and \textit{cultural expressions} referred to the perceptions of users in \texttt{r/newzealand}. These terms were used interchangeably by the \acrshort{OP} (in addition to \textit{sayings}) in reference to language use. One linguistic feature that seemed important was the descriptive modifiers used across the selfpost. These included the superlative \textit{corniest}, the adjectives \textit{trite}, \textit{clichéd}, \textit{ironic}, and \textit{sincere}, and the adverb \textit{uniquely}. The mixed sentiments of these descriptive modifiers suggest the \acrshort{OP} used them in a self-deprecating manner often associated with New Zealand humour. One descriptive modifier, \textit{Kiwi}, also appeared to be important in this situation, which I discuss in the building task of Building Identities.
    
\paragraph{Building Activities}

    As a selfpost, the primary activity should be to post a submission that relates directly to New Zealand. This situation did not appear to be the primary activity. Both the selfpost title and body text contained the wh-question construction (\textit{what}). The \acrshort{OP} used the selfpost to initiate a discussion with other users so they could contribute \textit{sayings} and \textit{cultural observations} which users perceived as \textit{corny}, \textit{trite}, \textit{clichéd}, or \textit{ironic}, as well as being \textit{sincere}, \textit{unique}, and most importantly \textit{Kiwi}. Instead, the secondary activity was elevated to the main activity in this situation.

\paragraph{Building Identities}

    The identity of the \acrshort{OP} was made relevant in this situation through the creation of the image post. A secondary identity made relevant in this situation was the group of users who engaged with this selfpost in the comment thread. One identity made irrelevant in this situation was the \glspl{mod} of \texttt{r/newzealand}. As the selfpost was deemed relevant to the place-based community and did not breach the community rules (``Submissions must directly relate to New Zealand''), moderation was not necessary. The descriptive modifier \textit{Kiwi} was used as the endonym for people associated with New Zealand. This linked the identities within the selfpost to the place-based community of \texttt{r/newzealand}. The \acrshort{OP} was not referring to the other definitions of \textit{kiwi}, such as the species of flightless bird from which the endonym was derived or the fruit. Curiously, the \acrshort{OP} used \textit{Kiwi} in lieu of the alternative endonym and variant \textit{New Zealander}.

\paragraph{Building Relationships}

    The primary social relationship in this situation was between the \acrshort{OP} and the users. Presumably, the relationship between users was transformed when both the \acrshort{OP} and users developed a mutual understanding of \textit{sayings} and \textit{cultural observations} which were \textit{uniquely Kiwi}. Due to the short length of the selfpost, it was difficult to ascertain any further relationships without referring to the comment thread.

\paragraph{Building Politics}

    User engagement was the primary social good in this situation. The selfpost received a score of 118 and 595 comments. Based on Reddit's internal metrics, the selfpost had an upvote ratio of 0.89, which suggests the post was high-quality, relevant, and engaging for other users in the community.

\paragraph{Building Connections}

    The \acrshort{OP} alluded to \textit{The Lord of the Rings} as a corny intertextual reference associated with New Zealand. The epic high fantasy novel \textit{The Lord of the Rings} by South African-born British author J.R.R. Tolkien (1892–1973) and the film trilogy directed by New Zealander Peter Jackson have had a significant impact on New Zealand's external image.
    
\paragraph{Building Significance for Sign Systems and Knowledge}

    In terms of shared systems and knowledge, there was a mutual understanding between the \acrshort{OP} and the users regarding which \textit{sayings} and \textit{cultural expressions} were \textit{corny}, \textit{trite}, \textit{clichéd}, or \textit{ironic}, as well as being \textit{sincere} and \textit{unique} to New Zealand. The perceptions of users were judged to be both true and relevant in this situation. One situated meaning that was expected in the situation, but made irrelevant, was the concept of cultural cringe \citep{phillips_cultural_1950}.
    
\subsubsection{Summary}

    Since the selfpost was relatively short, I was unable to delve deeper into the discourses within this situation. Of interest to me were the 595 comments from users who provided examples of features associated with \acrshort{NZE}. I describe these further in the thematic analysis later in this chapter.

\subsection{Selfpost 2}
\label{user:submission_post2}

    \begin{figure}
      \centering
      
            \includegraphics[width=\textwidth]{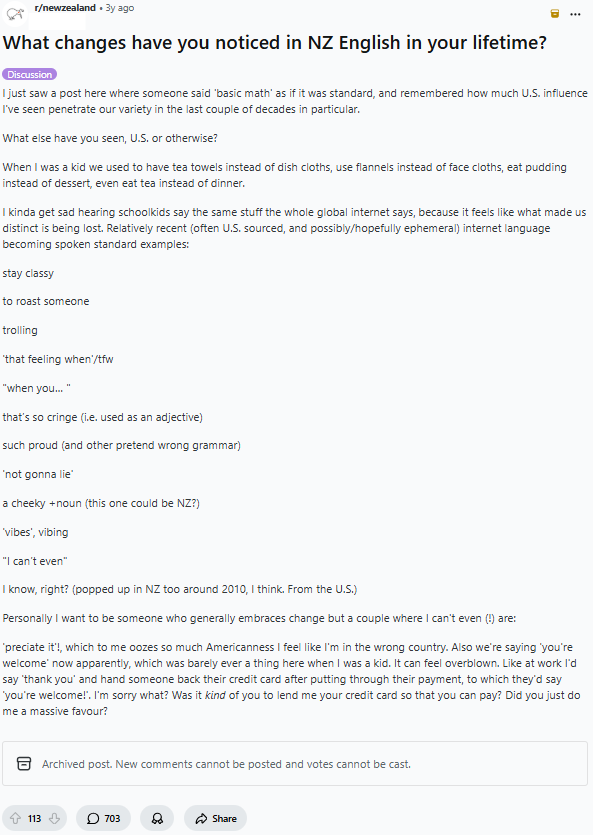}
            
      \vspace{6pt}
      \caption{Screenshot of Selfpost 2 (1102c68)}
      \label{fig:selfpost_1102c68}
      
    \end{figure}

    \begin{table*}[t!]
        \centering
        \caption{Selfpost 2 (\texttt{1102c68}): Macro-Structure (Part 1)}
            \begin{subtable}[t]{0.50\textwidth}
                \scriptsize
                \renewcommand{\arraystretch}{1.4}
                \centering
                \vspace{0.25cm}
                \begin{tabular}{lp{0.8\linewidth}}
                    \hline 
                    \textbf{Line} &  \\
                    \hline 
                    \\
                     & \textbf{Start of Sub-Story 1} \\
                     \\
                    \textbf{I} & \textbf{Orientation} \\
                    \\
                     & \textit{Stanza 1} \\
                    1 & I just saw a post here \\
                    \\
                    \textbf{II} & \textbf{Catalyst} \\
                    \\
                    & \textit{Stanza 2} \\
                    2 & where someone said `basic math' \\
                    \\
                    \textbf{III} & \textbf{Crisis} \\
                    \\
                    & \textit{Stanza 3} \\
                    3 & as if it was standard \\
                    \\
                    \textbf{IV} & \textbf{Evaluation} \\
                    \\
                    & \textit{Stanza 4} \\
                    4 & and remembered how much U.S. influence \\
                    5 & I've seen penetrate our variety \\
                    6 & in the last couple of decades \\
                    7 & in particular \\
                    \\
                    \textbf{V} & \textbf{Resolution} \\
                    \\
                    & \textit{Stanza 5} \\
                    8 & Ø \\
                    \\
                    \textbf{VI} & \textbf{Coda} \\
                    \\
                    & \textit{Stanza 6} \\
                    9 & What else have you seen \\
                    10 & U.S. or otherwise \\
                    \\
                    \\
                     & \textbf{Start of Sub-Story 2} \\
                    \\
                    \textbf{I} & \textbf{Orientation} \\
                    \\
                    & \textit{Stanza 7} \\
                    11 & When I was a kid I used to have tea towels instead of dish cloths \\
                    12 & use flannels instead of face cloths \\
                    13 & eat pudding instead of dessert \\
                    \\
                    \hline
                \end{tabular}
            \end{subtable}%
            ~ 
            \begin{subtable}[t]{0.50\textwidth}
                \scriptsize
                \renewcommand{\arraystretch}{1.4}
                \centering
                \vspace{0.25cm}
                \begin{tabular}{lp{0.8\linewidth}}
                    \hline 
                    \textbf{Line} & \\
                    \hline 
                    \\
                    14 & even eat tea instead of dinner \\
                    \\
                    \textbf{II} & \textbf{Catalyst} \\
                    \\
                    & \textit{Stanza 8} \\
                    15 & I kinda get sad \\
                    16 & hearing schoolkids say the same stuff \\
                    17 & the whole global internet says \\
                    \\
                    \textbf{III} & \textbf{Crisis} \\
                    \\
                    & \textit{Stanza 9} \\
                    18 & because it feels like what made us distinct is being lost \\
                    \\
                    \textbf{IV} & \textbf{Evaluation} \\
                    \\
                    & \textit{Stanza 10} \\
                    19 & Relatively recent \\
                    20 & often U.S. sourced \\
                    21 & and possibly \\
                    22 & hopefully \\
                    23 & ephemeral \\
                    24 & internet language becoming spoken standard \\
                    \\
                    \textbf{V} & \textbf{Resolution} \\
                    \\
                    & \textit{Stanza 11} \\
                    25 & Ø \\
                    \\
                    \textbf{VI} & \textbf{Coda} \\
                    \\
                    & \textit{Stanza 12} \\
                    26 & examples \\
                    \\
                    & \textit{Stanza 13} \\
                    27 & stay classy \\
                    \\
                    & \textit{Stanza 14} \\
                    28 & to roast someone \\
                    \\
                    & \textit{Stanza 15} \\
                    29 & trolling \\
                    \\
                    & \textit{Stanza 16} \\
                    30 & `that feeling when' \\
                    \\
                    \hline
                \end{tabular}
            \end{subtable}
    \end{table*}

    \begin{table*}[t!]
        \centering
        \caption{Selfpost 2 (\texttt{1102c68}): Macro-Structure (Part 2)}
            \begin{subtable}[t]{0.50\textwidth}
                \scriptsize
                \renewcommand{\arraystretch}{1.4}
                \centering
                \vspace{0.25cm}
                \begin{tabular}{lp{0.8\linewidth}}
                    \hline 
                    \textbf{Line} &  \\
                    \hline 
                    \\
                    31 & tfw \\
                    \\
                    & \textit{Stanza 17} \\
                    32 & `when you ...' \\
                    \\
                    & \textit{Stanza 18} \\
                    33 & that's so cringe \\
                    34 & i.e. used as an adjective \\
                    \\
                    & \textit{Stanza 19} \\
                    35 & such proud \\
                    36 & and other pretend wrong grammar \\
                    \\
                    & \textit{Stanza 20} \\
                    37 & `not gonna lie' \\
                    \\
                    & \textit{Stanza 21} \\
                    38 & a cheeky + noun \\
                    39 & this one could be NZ \\
                    \\
                    & \textit{Stanza 22} \\
                    40 & `vibes' \\
                    41 & vibing \\
                    \\
                    & \textit{Stanza 23} \\
                    42 & `I can't even' \\
                    \\
                    & \textit{Stanza 24} \\
                    43 & I know, right? \\
                    44 & popped up in NZ too \\
                    45 & around 2010 \\
                    46 & I think \\
                    47 & From the U.S. \\
                    \\
                    \\
                    & \textbf{Start of Sub-Story 3} \\
                    \\
                    \textbf{I} & \textbf{Orientation} \\
                    \\
                    & \textit{Stanza 25} \\
                    48 & Personally \\
                    \\
                    \textbf{II} & \textbf{Catalyst} \\
                    \\
                    & \textit{Stanza 26} \\
                    \\
                    \hline
                \end{tabular}
            \end{subtable}%
            ~ 
            \begin{subtable}[t]{0.50\textwidth}
                \scriptsize
                \renewcommand{\arraystretch}{1.4}
                \centering
                \vspace{0.25cm}
                \begin{tabular}{lp{0.8\linewidth}}
                    \hline 
                    \textbf{Line} & \\
                    \hline 
                    \\
                    49 & I want to be someone who generally embraces \\
                    & change \\
                    50 & but a couple \\
                    51 & where I can't even \\
                    52 & are `preciate it' \\
                    53 & which to me oozes \\
                    54 & so much Americanness \\
                    \\
                    \textbf{III} & \textbf{Crisis} \\
                    \\
                    & \textit{Stanza 27} \\
                    55 & I feel like I'm in the wrong country \\
                    \\
                    \textbf{IV} & \textbf{Evaluation} \\
                    \\
                    & \textit{Stanza 28} \\
                    56 & Ø \\
                    \\
                    \textbf{V} & \textbf{Resolution} \\
                    \\
                    & \textit{Stanza 29} \\
                    57 & Ø \\
                    \\
                    \textbf{VI} & \textbf{Coda} \\
                    \\
                    \\
                    & \textbf{Start of Sub-sub-story 1} \\
                    \\
                    \textbf{I} & \textbf{Orientation} \\
                    \\
                    & \textit{Stanza 30} \\
                    58 & Ø \\
                    \\
                    \textbf{II} & \textbf{Catalyst} \\
                    \\
                    & \textit{Stanza 31} \\
                    59 & Also \\
                    60 & we're saying `you're welcome' now \\
                    61 & apparently \\
                    \\
                    \textbf{III} & \textbf{Crisis} \\
                    \\
                    & \textit{Stanza 32} \\
                    62 & which was barely ever a thing here \\
                    63 & when I was a kid \\
                    \\
                    \hline
                \end{tabular}
            \end{subtable}
    \end{table*}

    \begin{table*}[t!]
        \centering
        \caption{Selfpost 2 (\texttt{1102c68}): Macro-Structure (Part 3)}
            \begin{subtable}[t]{0.50\textwidth}
                \scriptsize
                \renewcommand{\arraystretch}{1.4}
                \centering
                \vspace{0.25cm}
                \begin{tabular}{lp{0.8\linewidth}}
                    \hline 
                    \textbf{Line} &  \\
                    \hline 
                    \\
                    \textbf{IV} & \textbf{Evaluation} \\
                    \\
                    & \textit{Stanza 34} \\
                    65 & Ø \\
                    \\
                    \textbf{VI} & \textbf{Coda} \\
                    \\
                    \\
                    & \textbf{Start of Sub-sub-story Proper} \\
                    \\
                    \textbf{I} & \textbf{Orientation} \\
                    \\
                    & \textit{Stanza 35} \\
                    66 & Like at work \\
                    \\
                    \textbf{II} & \textbf{Catalyst} \\
                    \\
                    & \textit{Stanza 36} \\
                    67 & I'd say `thank you' \\
                    68 & and hand someone back their credit card \\
                    69 & after putting through their payment \\
                    \\
                    \\
                    \hline
                \end{tabular}
            \end{subtable}%
            ~ 
            \begin{subtable}[t]{0.50\textwidth}
                \scriptsize
                \renewcommand{\arraystretch}{1.4}
                \centering
                \vspace{0.25cm}
                \begin{tabular}{lp{0.8\linewidth}}
                    \hline 
                    \textbf{Line} & \\
                    \hline 
                    \\
                    \textbf{III} & \textbf{Crisis} \\
                    \\
                    & \textit{Stanza 37} \\
                    70 & to which they'd say `you're welcome!' \\
                    \\
                    \textbf{IV} & \textbf{Evaluation} \\
                    \\
                    & \textit{Stanza 38} \\
                    71 & I'm sorry what? \\
                    72 & Was it kind of you to lend me your credit card \\
                    73 & so that you can pay? \\
                    74 & Did you just do me a massive favour? \\
                    \\
                    \textbf{V} & \textbf{Resolution} \\
                    \\
                    & \textit{Stanza 39} \\
                    75 & Ø \\
                    \\
                    \textbf{VI} & \textbf{Coda} \\
                    \\
                    & \textit{Stanza 40} \\
                    76 & Ø \\
                    \\
                    \hline
                \end{tabular}
            \end{subtable}
    \end{table*}

    I now apply the Seven Building Tasks \citep{gee_introduction_2005} to Selfpost 2 (see \ref{fig:selfpost_1102c68}). As the selfpost was presented as an extended narrative, I have applied the Narrative Analysis Framework \citep{labov_narrative_1967} to determine the macrostructure of the selfpost by organising the submission into its constituent narrative elements: an orientation, a catalyst, a crisis, an evaluation, a resolution, and the coda. I identified three sub-stories, 40 stanzas, and 76 idealised lines. While each of the sub-stories corresponded with the narrative elements (the orientation, catalyst, and crisis in sub-story 1; evaluation and resolution in sub-story 2; and the coda in sub-story 3), I analysed each sub-story as a standalone situation contributing to the extended narrative of the selfpost.
    
\paragraph{Building Significance}

    The first sub-story was the shortest of the three sub-stories. The narrative elements of the first sub-story included an orientation, a catalyst, a crisis, an evaluation, and a coda. The sub-story did not conclude with a resolution. The \acrshort{OP} identified themselves as the primary actor, \texttt{r/newzealand} as the location, and their reaction to another selfpost as the behavioural motivator. The \acrshort{OP} described the complicating situation over the preceding idealised lines. Here, the \acrshort{OP} identified the presence of \textit{basic math} as the main catalyst for their selfpost. The \acrshort{OP} associated this feature with the United States and perceived the presence of this feature - and, by extension, American English - as a threat to their variety of English. The \acrshort{OP} then concluded the sub-story in the coda by addressing the other users directly with a question, intending to prompt them to engage in this extended narrative.

    The second sub-story comprised an orientation, a catalyst, a crisis, an evaluation, and a coda. Similarly, sub-story 2 did not conclude with a resolution. The purpose of the second sub-story was to establish a sense of contrast between the \acrshort{OP}'s perceived language use in the present and the past. Once again, the \acrshort{OP} oriented the users by identifying themselves as the primary actor within the sub-story and the time as situated in the past (\textit{When I was a kid}). By using the \textit{instead of} construction, the \acrshort{OP} established a temporal contrast between one set of features associated with the past and another set of features associated with present language use. The \acrshort{OP} concluded the narrative element by identifying the language use of \textit{schoolkids} as the main contributor to a lost sense of collective identity in New Zealand.

    During the evaluation, the \acrshort{OP} once again attributed the crisis to being \textit{US sourced} and caused by \textit{internet language}. The coda further reinforced the complicating situation observed in sub-story 2 by including examples of lexico-semantic features (and morphosyntactic constructions) which were presumably \textit{US sourced}, stemming from \textit{internet language}, and being used by \textit{schoolkids}. By listing these constructions, the \acrshort{OP} insinuated that they were not associated with \textit{our variety}, which was under threat from \textit{US influence}. The \acrshort{OP} briefly questioned the validity of their claims (\textit{this one could be NZ}) by showing uncertainty in their judgements (\textit{I think from the US}).

    The third sub-story was the most structurally complex, with its own orientation, catalyst, and crisis. The coda marked the beginning of a sub-sub-story. In the orientation, the \acrshort{OP} no longer provided a place, time, or behavioural motivator. This was because the \acrshort{OP} was the only identity involved in the situation. The \acrshort{OP} identified the phrase \textit{preciate it} as the catalyst of this situation, which made them feel displaced (\textit{I feel like I'm in the wrong country}). This also constituted the crisis. \textit{Americanness} was once again the cause of this crisis, which linked this sub-sub-story back to the first and second sub-stories, with the \acrshort{OP}'s inability to accept change contributing to their negative affective stance.

    As the selfpost progressed, the extended narrative became more unstructured. Within the coda of sub-story 3, I observed the beginning of sub-sub-story 1. Sub-sub-story 1 did not provide an orientation at all. The \acrshort{OP} introduced a new complicating situation with the phrase \textit{you're welcome} which the \acrshort{OP} argued, like \textit{preciate it}, was not used in New Zealand. In this situation, the \acrshort{OP} used \textit{here} to refer to New Zealand, the physical socio-political entity. This contrasted with how the \acrshort{OP} previously used \textit{here} to describe \texttt{r/newzealand} during the orientation of sub-story 1. The \acrshort{OP} once again used \textit{when I was a kid}, last observed in sub-story 2, to re-situate the users to a time in New Zealand before \textit{US influence} or \textit{internet language}.
    
\paragraph{Building Activities}

    Similar to selfpost 1, the main activity for this selfpost was not the primary activity, which was to post a submission that directly related to New Zealand. Instead, the \acrshort{OP} used the selfpost as a platform to initiate a discussion with the users of \texttt{r/newzealand}, as indicated by the \textit{Discussion} post \gls{flair}. Not only did the \acrshort{OP} want the users to engage with the selfpost through upvoting, but the \acrshort{OP} also encouraged comments through their open-ended question in the coda of sub-story 1: \textit{What else have you seen US or otherwise?}. By including their personal experiences and opinions in this complex narrative, the \acrshort{OP} wanted the users primarily to agree with them (or, alternatively, to disagree in the comments).

\paragraph{Building Identities}

    \begin{figure}
    \footnotesize
    \begin{align} 
        \textnormal{\textbf{I} just saw a post here (Line 1)} \tag{\texttt{a}} \\ 
        \textnormal{[How much U.S. influece] \textbf{I}'ve seen penetrate our variety (Line 5)} \tag{\texttt{b}} \\
        \textnormal{When \textbf{I} was a kid [we used to have tea towels instead of dish cloths] (Line 11)} \tag{\texttt{c}} \\
        \textnormal{\textbf{I} kinda get sad [hearing schoolkids say the same stuff] (Line 15)} \tag{\texttt{d}} \\
        \textnormal{\textbf{I} think [from the U.S.] (Line 46)} \tag{\texttt{e}} \\
        \textnormal{\textbf{I} want to be someone who generally embraces change (Line 49)} \tag{\texttt{f}} \\
        \textnormal{\textbf{I} feel like I'm in the wrong country (Line 55)} \tag{\texttt{g}} \\
        \textnormal{[Which was barely a thing here] when \textbf{I} was a kid (Line 63)} \tag{\texttt{h}} \\
        \textnormal{\textbf{I}'d say `thank you' (Line 67)} \tag{\texttt{i}} \\
        \textnormal{\textbf{I}'m sorry what? (Line 71)} \tag{\texttt{j}} \\
        \label{istatements}
    \end{align}
        \vspace{-12pt}
        \label{cognitive-i}
        \caption{Selfpost 2 (\texttt{1102c68}): Cognitive \textit{I}-Statements}
    \end{figure}

    Except for quotes, the first-person singular \textit{I} was the most frequent pronoun, occurring 11 times throughout the situation. Evidently, the \acrshort{OP} was the central identity of this selfpost. In order to determine how the \acrshort{OP} mediated their situated identity throughout the selfpost, I organised the \textit{I}-statements into discrete categories based on their functions, which include expressing affect or desire, action or status, and cognitive-I statements \citep{gee_introduction_2005}. Based on this categorisation, the \acrshort{OP}'s \textit{I}-statements largely expressed affect and desire, with the bulk of the \textit{I}-statements involving an action or state. I only observed one example of the cognitive-\textit{I} category, which suggests the narrative within the selfpost leant heavily into the \acrshort{OP}'s actions as well as affect and desire.

    The users, while also a situated identity in this situation, remained a passive identity in the selfpost until the \acrshort{OP} addressed them directly. As an example, the second-person singular/plural \textit{you} was scarcely used by the \acrshort{OP} in reference to the users, only appearing in either the quotative or as an indirect object. The \acrshort{OP} involved other users in \texttt{r/newzealand} in this situation by using the inclusive plural personal pronouns (\textit{our}, \textit{we}, and \textit{us}). The \acrshort{OP} has made an assumption that other users have a mutual understanding of what was meant by \textit{our variety} and \textit{standard} language use, not only in the place-based community of \texttt{r/newzealand}, but also in New Zealand.
    
    One situated identity made irrelevant in this situation consisted of those who did not have a mutual understanding of the \textit{standard} or \textit{our variety}. In the first sub-story, the person involved in the situation (\textit{someone who said `basic math'}) was made relevant. However, they were also excluded by the OP. These relevant identities were all situated in time (the younger generation) and place (those outside of \texttt{r/newzealand} and New Zealand). More specifically, these relevant yet excluded identities were \textit{US sourced} or \textit{from the US}. As the extended narrative progressed, \textit{schoolkids} and those associated with American English and the internet were also made relevant, but excluded from the situation. These relevant yet excluded identities were not included in the \acrshort{OP}'s designation of \textit{standard}.

\paragraph{Building Relationships}

    The \acrshort{OP} utilised the inclusive plural personal pronouns (\textit{our}, \textit{we}, and \textit{us}) to construct a collective sense of identity between the \acrshort{OP} and other users in \texttt{r/newzealand}. This was best illustrated by the phrases \textit{I've seen penetrate our variety} and \textit{because it feels like what made us distinct is being lost}.

\paragraph{Building Politics}

    Over the course of the selfpost, the \acrshort{OP} attempted to build a sense of urgency with the users. A social good at risk in this situation was the \acrshort{OP}'s insider status as someone who could define speakers of \textit{our variety} and those who could uphold the \textit{standard}. The selfpost received a score of 113 and 703 comments. Using Reddit's internal recommendation system, both the best and top comment, with a score of 350, questioned the \acrshort{OP}'s definition of \textit{dish cloth} and \textit{tea towel}.

\paragraph{Building Connections}

    One situated meaning that seemed relevant to the selfpost was the word \textit{standard} (\textit{as if it was standard}; \textit{internet language becoming spoken standard}). The \acrshort{OP} listed \textit{tea towel}, \textit{flannel}, \textit{pudding}, and \textit{tea} as being associated with \acrshort{NZE} (\textit{our variety}). The \acrshort{OP} used the \textit{instead of} construction to establish a contrast between lexical features associated with \acrshort{NZE} and those that were not. By demarcating a form of language use that was indexed to a specific sociodemographic group, the \acrshort{OP} established a form of language that was considered neither \textit{our variety} nor \textit{standard}. In this situation, \textit{standard} was a discourse model relevant to the context, with \textit{standard} benchmarked to lexis associated with \acrshort{NZE}.

    In addition to the lexical features, the \acrshort{OP} also produced a list of constructions which were deemed not to be part of \textit{our variety}. The first group of lexico-semantic constructions included: \textit{basic math}, \textit{trolling}, \textit{that's so cringe}, \textit{a cheeky + noun}, and \textit{vibes}. The \acrshort{OP} described the introduction of these variants as \textit{penetrating our variety}. The last group of constructions included pragmatic features associated with American English: \textit{preciate it} and \textit{you're welcome} in response to expressions of gratitude, \textit{which was barely ever a thing here} - with \textit{here} once again being New Zealand. These constructions were mentioned in the sub-sub-story of the coda in sub-story 3. The \acrshort{OP} described the introduction of this interactional feature as something that \textit{oozes so much Americanness} and is \textit{overblown}.
    
\paragraph{Building Significance for Sign Systems and Knowledge}

    The \acrshort{OP} created a sense of contrast between \acrshort{NZE} and the \textit{global internet} by singling out the United States (\textit{US}). The \acrshort{OP} established this contrast early on in the selfpost by describing \textit{US influence} and then later on when the \acrshort{OP} once again described these differences as \textit{US sourced}. The frequency with which \textit{US} was mentioned was significant. When compared with the presence of New Zealand (\textit{NZ}), \textit{NZ} only appeared once throughout the entire selfpost.

\subsubsection{Summary}

    Across the three sub-stories, there was a linear progression of the setting from online (\textit{here} with reference to \texttt{r/newzealand}) to offline (\textit{at work}). Time played an important role in orienting the reader. Sub-story 1 occurred in the recent past, sub-story 2 began in the remote past and concluded in the recent past, and sub-story 3 concerned a situation in the present. The \acrshort{OP} mentioned how they \textit{want to be someone who generally embraces change}, but was clearly distressed by ongoing change in \acrshort{NZE} observed on \texttt{r/newzealand}. I also observed an increase in intensity as the narrative progressed across the three sub-stories, from general discomfort (\textit{as if it was standard}), to a loss of identity (\textit{because it feels like what made us distinct is lost}), to a sense of displacement (\textit{I feel like I'm in the wrong country}).

\section{Interim Summary}
\label{user:interim_summary}

    My analyses of selfpost 1 (\texttt{155d1f3}) and Selfpost 2 (\texttt{1102c68}) offered an insight into the way users engaged with other users in \texttt{r/newzealand}. Intuitively, both OPs had a preconceived idea of what is \acrshort{NZE} and what is not. This was particularly evident in Selfpost 2 (\texttt{1102c68}), where the use of inclusive-we combined with the discourse model of \textit{standard} \acrshort{NZE} established a link between language use and identity. More importantly, this suggests language use acts as a form of place-making activity in \texttt{r/newzealand} \citep{johnstone_place_2004}. I was unable to draw the same conclusions from Selfpost 1 (\texttt{155d1f3}) due to its short length.
    
    Of interest to place identity is the role of social identity threat \citep{hogg_social_2016}. The \acrshort{OP} of Selfpost 2 (\texttt{1102c68}) offered multiple examples of change in \acrshort{NZE} influenced by American English. Of the six lexical examples, I could only trace \textit{dishcloth} \citeauthor{oxford_university_press_dishcloth_2025} (OED, \citeyear{oxford_university_press_dishcloth_2025}) to American English. Conversely, \textit{math} (OED, \citeyear{oxford_university_press_math_2025}) predates the orthographic variant \textit{maths} (OED, \citeyear{oxford_university_press_maths_2024}). According to their earliest attested usage, the remaining examples - \textit{cringe}, \textit{troll}, \textit{vibes}, and \textit{cheeky} (OED, \citeyear{oxford_university_press_cringe_2023}; \citeyear{oxford_university_press_troll_2023}; \citeyear{oxford_university_press_vibes_2024}; \citeyear{oxford_university_press_cheeky_2024}) - did not originate from American English.

    Another source of social identity threat was the internet. The \acrshort{OP} was correct in their claims that these constructions were a form of \textit{internet language}. For example, \textit{stay classy} originated in a meme template \citep{lavacano201014_stay_2011}. Other constructions described by the \acrshort{OP} also originated as meme templates, such as \textit{that feeling when} \citep{hamilton_blue_2023}, \textit{not gonna lie} \citep{adam_they_2019}, and \textit{such proud} \citep{novaxp_doge_2013}. The \acrshort{OP} described the phrase \textit{such proud}, associated with the Doge meme template, as \textit{pretend wrong grammar}. Originating from the internet does not necessarily suggest these constructions are automatically part of American English.

    Lastly, the \acrshort{OP} in Selfpost 2 dedicated a significant portion of the extended narrative to the pragmatics of gratitude across geographic dialect communities. The \acrshort{OP}'s intuition about this pragmatic practice in American English was largely correct. \citet{jautz_gratitude_2008} found that speakers of American English were more likely to use \textit{you're welcome} (53.5\%) than speakers of Irish English (34.2\%) or English English (16.3\%) \citep{schneider_no_2005}. Due to the limited research on the pragmatics of gratitude in \acrshort{NZE}, I cannot confirm whether there has in fact been a diachronic change.

\section{Content Analysis}
\label{user:content_analysis} 
    
    Following my discourse analyses of the two selfposts, I now move on to my next analytical step, where I conduct a content analysis of the comment threads. As an example, I observed a small set of user-informed features (such as \textit{tea towel}, \textit{flannel}, \textit{pudding}, and \textit{tea}) in Selfpost 2 (\texttt{1102c68}) which the \acrshort{OP} associated with \acrshort{NZE}. This small list offered an insight into users' perceptions of language use in \texttt{r/newzealand}. I extend this to the comment threads for both selfposts. My primary goal is to develop a set of user-informed linguistic features associated with \acrshort{NZE} which I can later test.

\subsection{Methodology}
    
    Whereas the previous section was concerned with the production circumstances of submission posts (\gls{rstext}) and selfposts (\gls{rstext}), I now shift my focus to the comments (\gls{rcomm}). Comments vary in length from one character to the official limit of 10,000, whether the comment is a top-level post or a reply. This is the same limit as the body text of a selfpost. Naturally, comments make up the bulk of the language production on Reddit. For this reason, comments on Reddit offer immense value in social media research. I use thematic analysis to examine the comments from the two selfposts \citep{braun_using_2006}. Thematic analysis consists of six phases: 1) data familiarisation; 2) initial code generation; 3) searching for themes; 4) reviewing themes; 5) defining and naming themes; and 6) producing the report \citep{braun_using_2006}. With these six phases as a guide, I first group the comments by theme and then into specific user-informed features. Ideally, I want to establish a list of sociolinguistic variables with contrasting conservative and innovative variants.

\subsection{Cultural Cringe and National Pride}

    The \acrshort{OP} of Selfpost 1 (\texttt{155d1f3}) encouraged users to offer examples of sayings and cultural expressions that were \textit{corny}, \textit{trite}, \textit{clichéd}, \textit{ironic}, \textit{sincere}, and \textit{uniquely Kiwi}. There was a total of 595 comments associated with this selfpost. Of those 595 comments, approximately 45.7\% ($n=278$) did not explicitly provide examples of sayings or cultural expressions. Instead, the purpose of these comments was to express agreement, clarification, humour, or even confrontation.With the remaining comments, I categorised them into three broad areas: linguistic, intertextual, and social phenomena. As I am interested in specific user-informed linguistic features, I focus my thematic analysis on the linguistic phenomena only. This included phonological, lexical, morphosyntactic, and pragmatic phenomena. Where appropriate, I disaggregated comments based on the user-informed features, as some users provided more than one example within their comment.

\paragraph{Phonological Phenomena}

   Phonology received the least attention from users, with only five associated comments. Out of the five comments, two were general statements where one user described the entire \acrshort{NZE} vowel system as corny, while another user described the high rising terminal as an \textit{affliction}, which received 12 upvotes. In terms of user-informed phonological examples, one user noted metathesis in \textit{ask} (written as \textit{aks}), which received a score of eight; another user noted phonemic blending in \textit{library} ($\uparrow2$); and lastly, one user noted the \textit{brought} and \textit{bought} alternation as corny, which received a total of 95 upvotes.

\paragraph{Lexical Phenomena}
    
    I identified fifty-one comments where users described corny lexical features associated with \acrshort{NZE}. After aggregating the comments, I found 24 unique user-informed features. Only eight of these features were associated with \acrshort{NZE} or recent borrowings from Australian English. These included: \textit{\gls{arvo}}, \textit{\gls{chocka}}, \textit{\gls{kiwiana}}, \textit{\gls{savvyb}}, \textit{\gls{skux}}, \textit{\gls{spagbol}}, \textit{\gls{sweetbix}}, and \textit{\gls{tucker_fucker}}. The remaining features were not specific to \acrshort{NZE}. These included: \textit{bro}, \textit{buzzy}, \textit{coin}, \textit{corny}, \textit{grub}, \textit{gutted}, \textit{hectic}, \textit{per capita}, \textit{lovely jubbly}, \textit{mate}, \textit{missus}, \textit{poos}, \textit{ratshit}, \textit{reckon}, \textit{slammed}, and \textit{wees}.

\paragraph{Morphosyntactic Phenomena}

    I identified two morphosyntactic phenomena in Selfpost 1 (\texttt{155d1f3}). These user-informed features were the intensifier-as construction \citep{sowa_sweet_2009} and the marked second-person plural \citep{bauer_can_2002}. The first corny user-informed morphosyntactic feature, discussed in twenty comments, was the intensifier-as construction, which received a combined upvote score of 283. Some comments included both derision and praise. User-informed examples of this construction included \textit{good as gold} (or simply, \textit{good as}) ($\uparrow52$), \textit{cheap as chips} ($\uparrow44$), \textit{guttering as} ($\uparrow5$), \textit{man, that's buzzy as} ($\uparrow2$), and \textit{sweet as} ($\uparrow1$).There was also considerable interest in the marked plural pronoun (\textit{all ya'll}, \textit{you lot}, \textit{youse}). If I split the upvotes by sentiment, I note that 62 upvotes supported the negative comment (\textit{… hate both}), 65 were neutral (\textit{... is common in [New Zealand] because English doesn't have a proper [separate] plural for you, whereas [Māori] does ...}), and 24 were in support (\textit{Yous is great}).

\paragraph{Pragmatic Phenomena}

    Of the 595 comments, there were 162 comments describing pragmatic features of \acrshort{NZE} as corny. This was the largest category of features in Selfpost 1 (\texttt{155d1f3}). I categorised these features into one-word interjections, idiomatic phrases, and cross-linguistic phrases between \acrshort{NZE} and te reo Māori. Some cringy interjections included: \textit{awesome} ($\uparrow17$), \textit{boomfa} ($\uparrow2$), \textit{choice} ($\uparrow15$), \textit{chur} ($\uparrow20$), and \textit{mean} ($\uparrow5$). These interjections are often associated with positive responses. In terms of those associated with negative responses, these included \textit{crikey} ($\uparrow2$), \textit{naw} ($\uparrow5$), and \textit{stink} ($\uparrow12$). Users also noted the use of discourse markers \textit{exactly} ($\uparrow2$), \textit{obviously} ($\uparrow7$), \textit{okay} ($\uparrow2$), and \textit{yeah} ($\uparrow2$), as well as the hesitation marker \textit{so} ($\uparrow1$), as corny.

    Approximately half ($n=84$) of these user-informed pragmatic features were idiomatic or multi-word phrases. I have summarised these multi-word phrases in Table \ref{tab:idiomatic_phrases}. Of the 70 multi-word phrases, nine users noted \textit{you can't beat Wellington on a good day} (and its variants) as corny. This user-informed pragmatic feature received a combined upvote score of 592. Nine users also commented that \textit{at the end of the day} was corny. The last category of user-informed pragmatic features included code-mixing and borrowings between \acrshort{NZE} and te reo Māori. Some user-informed examples of code-mixing and borrowings include \gls{mahi} (\textit{did the mahi}, \textit{do the mahi}), \gls{tahi} (\textit{that's the tahi}), and \gls{ra} (\textit{at the end of te rā}). Another example was \gls{patu}, as in \textit{patu as}, \textit{patu feet}, and \textit{way patu}. Some extended examples include \gls{whakatauki} such as \gls{he_aha} and \gls{he_waka}. In general, some users mentioned the tokenistic use of te reo Māori as corny ($n=3$).

    \begin{table}
        \scriptsize
        \centering
            
            \caption{User-informed Cringy Multi-Word Phrases}
            \label{tab:cringe_multiword}
            
            \renewcommand{\arraystretch}{1.4}
            \begin{tabularx}{\textwidth}{cccc}
            \toprule
            Score & Multi-Word Phrase & Score & Multi-Word Phrase \\
            \midrule
                \addlinespace[1em]
                    592 & you can't beat wellington on a good day & 5 & get with the programme \\
                    548 & at the end of the day & 5 & hell to the yeah \\
                    150 & look, son & 5 & well he made a dog's breakfast of that \\
                    124 & she'll be right & 4 & today's definitely my day \\
                    98 & you know & 3 & actually not bad \\
                    60 & yeah nah & 3 & anyone who calls you a fool has seen a few \\
                    52 & you're golden & 3 & box of fluffy ducks \\
                    49 & that's primo & 3 & if it ain't broke don't fix it \\
                    47 & you can't handle the jandal & 3 & it's a different kind of heat \\
                    19 & thanking you & 3 & nah mate that's munted \\
                    18 & get a feed & 3 & take as much as you want \\
                    17 & at the coalface & 3 & well pucker innit \\
                    17 & in the trenches & 3 & whinging poms \\
                    11 & it's all about the vibes bro & 2 & aw yeah \\
                    10 & rattle your daggs & 2 & go for gold \\
                    9 & all right & 2 & it is what it is \\
                    9 & this isn't africa & 2 & it's our year \\
                    9 & you won't know yourself & 2 & listen up peeps \\
                    8 & come here ya fackin egg or I'll dong ya & 2 & not a dickie bird \\
                    8 & done like a dogs dinner & 2 & not here to fuck spiders \\
                    8 & fill your boots & 2 & silly games silly prizes \\
                    8 & if you go with the flow & 2 & that's fish n chips paper \\
                    8 & it'll be fine you won't get pregnant & 2 & that's not going anywhere \\
                    8 & it's fine once you're in & 2 & that's why \\
                    8 & thank you caller & 2 & too easy \\
                    8 & way shape or form & 2 & useless as tits on a bull \\
                    8 & you're a gentleman and a scholar & 2 & we're not here to fuck spiders \\
                    7 & if it can't be done by one guy in a garage & 2 & yeh that's the one eh \\
                    7 & sydney for beginners & 1 & doing laps of the main \\
                    7 & the thing is that & 1 & don't be sad ow \\
                    6 & alright I guess & 1 & I might come around \\
                    6 & yeah bol & 1 & that's ideal \\
                    5 & are you there & 1 & what did you actually say \\
                    5 & back in the uk & 1 & wrap your laughing gear around this/that \\
                    5 & four seasons in one day & & \\
                \addlinespace[1em]
            \bottomrule
            \end{tabularx}

        \tablenoteparagraph{\textbf{Table Note}: Columns provide the combined score (\texttt{Score}) and the multi-word phrase (\texttt{Multi-Word Phrase}). Data source: Reddit/r/newzealand. List of user-informed multi-word phrases from the comments of Self-post 1 (\texttt{155d1f3}) sorted in descending order of the score. The multi-word phrase with the highest score was \textit{you can't beat wellington on a good day} with a combined score of 592.}
        
    \end{table}

\subsubsection{Summary}

    The list of corny user-informed features does not necessarily reflect a negative affective stance from users. For example, when one user commented on how they hated the interjection \textit{sweet as}, they received two downvotes. Nor does \textit{corny} automatically suggest negative affect. Instead, I propose these features offer evidence of enregisterment (\citealp{agha_social_2003}; \citealp{johnstone_pittsburghese_2009}). More specifically, these corny user-informed features index a New Zealand identity for users on \texttt{r/newzealand}. The presence of these features supports my proposal that language, in its most salient sense, is utilised by users as a place-making activity \citep{johnstone_place_2004}.
    
    One insight from the thematic analysis was the high volume of multi-word phrases users associated with \acrshort{NZE}. I also note the absence of named entities in the user-informed features. The user-informed features from Selfpost 1 (\texttt{155d1f3}) offer limited practical utility within the context of my research. As discrete user-informed features, I may be able to compare the distribution of the lexical, morphosyntactic, and pragmatic phenomena across place-based communities. However, the lack of user-informed variants means I am limited in my analysis.
    
    Even though an in-depth thematic analysis of the social and intertextual phenomena is beyond the scope of my research, I want to briefly mention the intertextual references, as they offered a useful indicator of users' social demographics. These intertextual references alluded to other written, oral, or visual texts (such as television), suggesting that users share a similar repertoire of pop-cultural references. I present a list of these references in Table \ref{tab:intertextual}. The majority of the intertextual references originate from between the 1990s and 2010s. These intertextual references might have prompted some users to describe New Zealand's advertising industry as corny ($n=21$).
    
    Similarly, the social phenomena also offer insights into the ideologies and attitudes of users towards New Zealand culture. Approximately 8.4\% ($n=50$) commented on the culture of New Zealand. Some users ($n=14$) noted that New Zealanders' general \textit{non-confrontational} and \textit{laid-back} approach to social issues leans towards passive-aggression and apathy, as characterised by politicians in New Zealand ($n=13$). Some users ($n=8$) found the \textit{Kiwi ingenuity} trope to be corny, including outward expressions of national pride ($n=7$), which some described as the \textit{Number 8 wire} mentality. Many of these comments have a direct link to cultural cringe \citep{pickles_transnational_2011}.

    \begin{table}
        \scriptsize
        \centering
            
            \caption{User-informed Intertextual References}
            \label{tab:intertextual}
            
            \renewcommand{\arraystretch}{1.4}
            \begin{tabularx}{\textwidth}{p{0.3\textwidth}p{0.3\textwidth}*{3}{>{\centering\arraybackslash}X}}
            \toprule
            Intertextual Reference & Source/Author & Year & Example \\
            \midrule
                \addlinespace[2em]
                \multicolumn{4}{c}{\textbf{Advertising}} \\
                \addlinespace[1em]
                100\% Pure New Zealand & Tourism New Zealand & 1999- & \href{https://www.youtube.com/watch?v=Bu-1_AFWcqQ}{Link} \\
                Briscoes' Lady & Briscoes New Zealand & 1989- & \href{https://www.youtube.com/watch?v=peCDFf41S24}{Link} \\
                `Clean, Green' New Zealand & Tourism New Zealand & 1999- & - \\
                C'mon, Guys, Get Firewise & Fire and Emergency New Zealand & 2005 & \href{https://www.youtube.com/watch?v=V3ShUDNT9NA}{Link} \\
                Mate, Mate, Mate, Dave & New Zealand Transport Agency & 2007 & \href{https://www.youtube.com/watch?v=l0c7RmsrxsY}{Link} \\
                World Famous in New Zealand & L\&P & 1994 & \href{https://www.youtube.com/watch?v=qZcUbtNIwI0}{Link} \\
                On the floor & Sky Television & 2004 & \href{https://www.youtube.com/watch?v=NyRWnUpdTbg}{Link} \\
                Two Shots for Summer & New Zealand Government & 2021 & - \\
                Only CC's ees Taste Like These & Bluebirds & 1988 & \href{https://www.youtube.com/watch?v=vr2mavhGI0c}{Link} \\
                Togs, Togs, Undies & Tip Top & 2006 & \href{https://www.youtube.com/watch?v=h-Lx2ihpGbc}{Link} \\
                \addlinespace[2em]
                \multicolumn{4}{c}{\textbf{Television and Movies}} \\
                \addlinespace[1em]
                Mount Doom & \textit{Lord of the Rings} Trilogy & 2001-2003 & - \\
                `Cook the man some fucking eggs' & \textit{Once Were Warriors} & 1994 & \href{https://www.youtube.com/watch?v=H0rpeR-uYdc}{Link} \\
                `Always Blow on the Pie' & \textit{Police Ten 7} & 2009 & \href{https://www.youtube.com/watch?v=aEAHLFvD3v4}{Link} \\
                John Campbell & - & - \\
                `O for Awesome' & \textit{Wheel of Fortune} & 1992 & \href{https://www.youtube.com/watch?v=_b81P3aPv7Y}{Link} \\
                `This is the fucking news' & \textit{3News} & 2015 & \href{https://www.youtube.com/watch?v=7xv_pfIK60k}{Link} \\
                `You're not In Guatemala now' & \textit{Shortland Street} & 1992 & \href{https://www.youtube.com/watch?v=effmUm_OGz0}{Link} \\
                \addlinespace[2em]
                \multicolumn{4}{c}{\textbf{Music}} \\
                \addlinespace[1em]
                Counting the Beat & The Swingers & 1981 & \href{https://www.youtube.com/watch?v=p72Z1D1oKbw}{Link} \\
                Welcome Home & Dave Dobbyn & 2006 & \href{https://www.youtube.com/watch?v=Wlz2uEuxyyk}{Link} \\
                Loyal & Dave Dobbyn & 1988 & \href{https://www.youtube.com/watch?v=Qwf2CWCq5tc}{Link} \\
                God Defend New Zealand & Thomas Bracken & 1870s & \href{https://www.youtube.com/watch?v=CbDf0YG2xnA}{Link} \\
                Slice of Heaven & Dave Dobbyn & 1986 & \href{https://www.youtube.com/watch?v=jM04FIgoEkI}{Link} \\
                Not Many & Scribe & 2003 & \href{https://www.youtube.com/watch?v=7hyD2yAZFwE}{Link} \\
                \addlinespace[2em]
                \multicolumn{4}{c}{\textbf{Social Media}} \\
                \addlinespace[1em]
                Beached Az & The Handsomity Institute & 2008 & \href{https://www.youtube.com/watch?v=ZdVHZwI8pcA}{Link} \\
                Nek Minit & AxstaBludsta & 2011 & \href{https://www.youtube.com/watch?v=CTZyorJVeqI{&}a}{Link} \\
                \addlinespace[1em]
                \bottomrule
            \end{tabularx}

        \tablenoteparagraph{\textbf{Table Note}: Columns provide the name of the intertextual reference (\texttt{Intertextual reference}), the source or author (\texttt{Source/Author}), the year (\texttt{Year}), and a YouTube hyperlink (\texttt{Link}). Rows group user-informed examples by media type. Data source: Reddit. The user-informed intertextual references were largely associated with New Zealand advertisements. The oldest reference identified was the national anthem, \textit{God Defend New Zealand}, while the most recent was the \textit{Two Shots for Summer} government campaign initiated in response to the COVID-19 pandemic.}
        
    \end{table}

\subsection{Language Variation and Change}

    There was a total of 703 comments associated with Selfpost 2 (\texttt{1102c68}). Of the 703 comments, 24.2\% ($n=170$) were used to express agreement, clarification, humour, or other forms of communication not directly addressing the \acrshort{OP}'s request for open discussion (\textit{What else have you seen US or otherwise?}). I analysed the remaining comments and grouped them into four categories based on linguistic feature types. These were phonological, lexical, morphosyntactic, semantic, and pragmatic phenomena. Where appropriate, I disaggregated comments based on the user-informed features, as some users provided more than one example within their comment.
    
\paragraph{Phonological Phenomena}

    Beginning with phonological phenomena, I identified 44 examples within the comment thread. This was in contrast to Selfpost 1 (\texttt{155d1f3}), where I only found five related comments. As a general observation, the majority of the user-informed phonological phenomena were associated with American English. Once again, this was likely due to priming from the \acrshort{OP} in the body text of the selfpost. I grouped these comments based on phonological processes including yod-dropping, deletion and reduction, stress, and metathesis.

\subparagraph{Yod-Dropping}

    Users noted that in certain consonant clusters involving /j/ there has been a noticeable change towards yod-dropping in contrast to yod-coalescence. As an example, \textit{duke} /dju\textipa{k}/ with yod-coalescence is realised as /d\textyogh\textipa{u}\textipa{k}/ while \textit{duke} with yod-dropping is realised as /du\textipa{k}/. Other examples noted by users included \textit{stupid} realised as /\textprimstress\textipa{stu.p}\textschwa\textipa{d}/ and \textit{tuesday} realised as /\textprimstress\textipa{tjuzdei}/ with yod-dropping in contrast to /\textprimstress\textipa{st}\textesh\textipa{u.p}\textschwa\textipa{d}/ and /\textprimstress\textipa{t}\textesh\textipa{uzdei}/ with yod-coalescence respectively. The majority of users associated yod-dropping with American English.

\subparagraph{Deletion and Reduction}

    Of the deletion and reduction processes, two users noted the deletion of word initial-\textit{h} in \textit{have} and \textit{herb}. While \textit{h}-dropping was observed in earlier varieties of \acrshort{NZE} \citep{maclagan_h-dropping_1998}, \textit{h}-dropping is not often associated with \acrshort{NZE}. Once again, users associated the realised of \textit{herb} as /\textbaro\textlengthmark\textipa{b}/ with American English. One user also noted reduction in \textit{chur brutha} /t\textesh\textbaro\textlengthmark.b\textturnr\textipa{a}\dh\textschwa/ realised as /t\textesh\textbaro\textlengthmark.b\textipa{a}\dh\textschwa/ (written as \textit{chur butha}). Lastly, one user noted that the consonant cluster reduction in \textit{library} /l\textturnscripta\textsci.b\textturnr\textschwa.\textturnr\textipa{i}\textlengthmark/$\sim$/l\textturnscripta\textsci.b\textschwa\textturnr.\textturnr\textipa{i}\textlengthmark/ as /l\textturnscripta\textsci.b\textturnr\textipa{i}\textlengthmark/ (written as \textit{libry}) was a change observed in \acrshort{NZE}. The same user questioned whether \textit{Wednesday} should be realised with three syllables (/wed.n\textschwa\textipa{z}.dei/) instead of two (/wen\textipa{z}.dei/)

\subparagraph{Stress}

    Alternating and variable realisations in some lexical features were also noted by some users ($n=12$) such as \textit{data} where /\textprimstress\textipa{d}a\textlengthmark\textfishhookr.\textschwa/ (written as \textit{darta}) was associated with \acrshort{NZE}, while /deit.\textschwa/ (written as \textit{dayta}) and /\textprimstress\textipa{d}\ae\textfishhookr.\textschwa/ (written as \textit{datta}) were associated with British and American Englishes respectively. One user attributed the introduction of /\textprimstress\textipa{d}\ae\textfishhookr.\textschwa/ to the American science fiction television series \textit{Star Trek: Next Generation} (1987-1994) as opposed to a direct import from American English. Another user noted that there was a change in the realisation of \textit{yoghurt} from /\textprimstress\textipa{j}\textturnscripta\textipa{g}\textschwa\textipa{t}/ (written as \textit{YOG-hurt}) to /\textprimstress\textipa{jo}\textupsilon\textipa{g}\textschwa\textipa{t}/ (written as \textit{YO-ghurt}). Once again, the innovative variant is associated with American English. Interestingly, users did not associate variable realisation in \textit{gif} (/d\textyogh\textsci\textipa{f}/$\sim$/\textg\textsci\textipa{f}/) or \textit{often} (/\textturnscripta\textipa{ft.}\textschwa\textipa{n}/$\sim$/\textturnscripta\textipa{f.}\textschwa\textipa{n}/) with a specific variety of English.

    Five users noted a change in lexical stress. As an example, one user noted that the stress in \textit{macrame} and \textit{pastels} from the first syllable, as in /\textprimstress\textipa{m}\ae\textipa{k}\textturnr\textschwa\textsecstress\textipa{me}\textsci/ and /\textprimstress\textipa{p}\ae\textipa{s.t}\textschwa\textipa{lz}/, to the second syllable as in /\textipa{m}\textschwa\textprimstress\textipa{k}\textturnr\textipa{a}\textlengthmark\textipa{mi}/$\sim$/\textipa{m}\textschwa\textprimstress\textipa{k}\textturnr\textipa{a}\textlengthmark\textipa{me}\textsci/ and /p\ae\textipa{s}\textprimstress\textipa{t}\textrevepsilon\textipa{lz}/ respectively. Another user noted that the stress in \textit{contribute} has moved from the second syllable /k\textschwa\textipa{n}\textprimstress\textipa{t}\textturnr\textsci\textipa{b.jut}/ to the first syllable /\textprimstress\textipa{k}\textturnscripta\textipa{n.}\textipa{t}\textturnr\textsci\textsecstress\textipa{b.jut}/. Two users noted that the unstressed form of \textit{the}, /\dh\textschwa/, has become the default form with the stressed form, /\dh\textipa{i}\textlengthmark/, restricted to specific contexts.

\subparagraph{Metathesis}

    Two users noted the variable realisations of \textit{ask} /ask/ as /aks/ (written as \textit{arks}) and \textit{spaghetti} /sp\textschwa.get.ti\textlengthmark/ as /p\textschwa\textipa{s}.get.ti\textlengthmark/ (written as \textit{basgetti}) which have undergone the process of metathesis whereby speech sounds are transposed within a word segment. 

\paragraph{Lexical Phenomena}

    The next set of linguistic features were lexical phenomena. In the majority of examples, users supplied both the conservative and innovative variants. These were in addition to the user-informed lexical variables that were already listed by the \acrshort{OP} in the body text of the selfpost (which were \textit{tea towel}, \textit{flannel}, \textit{pudding}, and \textit{tea}). I have summarised these user-informed lexical variables in Table \ref{tab:lexical_change}. As I offer more details on these user-informed lexical variables in Chapter \ref{chap:user_variables}: User-Informed Sociolinguistic Variables, I have chosen not to go into great detail here to minimise duplicating my analysis. The key insight from this list was that the innovative variants were either originally from the United States or chiefly used in North American English (see Section \ref{variable:variables}).
    
    One category of user-informed lexical features included innovative-only features where users have not provided a conservative variant as a point of comparison. These innovative-only features were predominantly borrowings from other varieties of English such as \textit{bruh} \citep{oxford_university_press_bruh_2024} and \textit{finna} \citep{oxford_university_press_finna_2023} from African-American Vernacular English, \textit{doggo} originating from internet chat groups \citep{punske_me_2019}, and other varieties (\textit{dump}, \textit{eraser}, \textit{folks}, \textit{gas station}, \textit{kiwi bird}, \textit{metric ton}, \textit{reach out}, \textit{trainers}, \textit{underwear}, \textit{yuppies}, \textit{vibes}, \textit{chai tea}, \textit{unironical}). A subset of these innovative-only features originated from Australian English (\textit{\gls{bottlo}}, \textit{\gls{compo}}, \textit{\gls{footy}}, and \textit{\gls{servo}}). Five of the six Australian English innovative-only features were hypocoristics \citep{simpson_hypocoristics_2008}.

    \begin{table}
        \scriptsize
        \centering
            
            \caption{User-informed Conservative and Innovative Lexis}
            \label{tab:lexical_change}
            
            \renewcommand{\arraystretch}{1.4}
            \begin{tabularx}{\textwidth}{*{2}{>{\centering\arraybackslash}X}|*{2}{>{\centering\arraybackslash}X}}
            \toprule
            Conservative & Innovative & Conservative & Innovative \\
            \midrule
                \addlinespace[1em]
                    addictive & addicting & milliard & billion \\
                    aeroplane & airplane & mum & mom \\
                    aluminium & aluminum & nappy, nappies & diaper, diapers \\
                    anyway & anyways & next minute & nek minnit \\
                    appreciate & preciate & obliged & obligated \\
                    back lawn, lawn & backyard, yard & petrol & gas \\
                    bias & biased & poo & poop \\
                    biscuit, biscuits & cookie, cookies & pudding & dessert \\
                    bogan, bogans & redneck, rednecks & rubbish & garbage \\
                    car bonnet, bonnet & hood & rubbish & trash \\
                    car boot, boot & trunk & salvation army & salvo \\
                    boyfriend & partner & sandshoes & sneakers \\
                    chinese gooseberries & kiwifruit & secondary school & high school \\
                    chips & fries & soccer & football \\
                    college & high school & soft drink & soda \\
                    courage & brave & specifically & pacific \\
                    crisps & chips & specific & pacifically \\
                    cunt & c-bomb & stink & stoink \\
                    curtains & drapes & sweets & candy \\
                    disrespect & diss & takeaway, takeaways & take out \\
                    dummies & pacifier & tea, tea time & dinner, dinner time \\
                    fire engine & fire truck & tea towel, tea towels & dish cloth, dish cloths \\
                    fizzy drink & soda & telly & tv \\
                    fizzy drink & soft drink & the warriors & the wahs \\
                    flat & apartment & thou & you \\
                    football & soccer & toilet & bathroom \\ 
                    footpath & sidewalk & toilet & potty \\
                    fuck & f-bomb & toilet & restroom \\
                    fuel & petrol & tramping & hiking \\
                    girlfriend & partner & truck & lorry \\
                    going out & dating & tuna & tuna fish \\
                    hole in the wall & atm & twink & white out \\
                    hot dog & corn dog & ute & truck, pickup truck \\
                    iceblocks, ice blocks & ice lolly, ice lollies & verse & versing \\
                    iceblocks, ice blocks & popsicle, popsicles & vivid & sharpie \\
                    knickers & panties & vomit & puke \\
                    lego bricks & legos & you & ya'll \\
                    lift & elevator & you guys & dude \\
                    lolly, lollies & candy & you guys & ya'll \\
                    lorry, lorries & truck & yous, youse & ya'll \\
                    maths & math & yous, youse & you all \\
                \addlinespace[1em]
            \bottomrule
            \end{tabularx}

            \tablenoteparagraph{\textbf{Table Note}: This table catalogues user-informed lexical features extracted from the comments of Selfpost 2 (\texttt{1102c68}) in \texttt{r/newzealand}, contrasting conservative variants with their innovative counterparts. The features are arranged alphabetically by the conservative variant, with orthographic variants for compounds and marked plural forms separated by commas. Data is sourced from Reddit/\texttt{r/newzealand}.}
        
    \end{table}
    
\paragraph{Morphosyntactic Phenomena}

    As with Selfpost 1 (\texttt{155d1f3}), morphosyntax received the least attention from users. Even so, there were 26 comments related to this linguistic feature. Some morphosyntactic phenomena were misidentified by users as lexical phenomena. I grouped these comments into six morphosyntactic phenomena: verb conjugations, modal-of, irregular plurals, conditionals, intensifier-as, and lexico-grammatical constructions.

\subparagraph{Verb Conjugations}
 
    One user noted a change in preference for the infinitive form of verbs in verb–noun compounds in place of the gerund. More specifically, the user provided the example that \textit{swimming team} and \textit{swim team} were now used in free variation, whereas \textit{swim pool} for \textit{swimming pool} is considered ungrammatical. Other users noted an increase in the use of the present perfect with the past simple rather than the past participle, using the example \textit{he has went} in place of \textit{he has gone}. The most controversial morphosyntactic phenomenon noted by users ($n=4$) was the simple past and past participle of \textit{bring} and \textit{buy} being used interchangeably.

\subparagraph{Modal-\textit{of}}

    Users ($n=4$) noted the increase in the use of \textit{modal-of} in the modal of opportunity construction (\textit{could have}, \textit{would have}, and \textit{should have}), where the contracted forms (\textit{could've}, \textit{would've}, \textit{should've}) have seen the re-analysis of the reduced \textit{have} as \textit{of} (as in \textit{could of}, \textit{would of}, \textit{should of}) \citep{hay_new_2008}. Other users have noted that the contracted forms have experienced further reduction to \textit{coulda}, \textit{woulda}, and \textit{shoulda}.

\subparagraph{Irregular Plurals}

    Three users noted the plural forms of \textit{fish} and \textit{sheep} regularised to \textit{fishes} and \textit{sheeps}. While the plural \textit{fishes} does refer to more than one species of fish, the regularised plural form \textit{sheeps} is non-standard.

\subparagraph{Conditionals}

    One user noted a change in conditionals, such as \textit{if you would (have)}. More specifically, the user provided the following contrasting constructions: \textit{if it would have been me} and \textit{if it was/were me}. The user illustrated this with the syntactic pairs \textit{if you would have given me the money} and \textit{if you gave me the money}.

\subparagraph{Intensifier-as}

    One user from selfpost 1 (\texttt{155d1f3}) noticed a decrease in the intensifier-as construction \citep{sowa_sweet_2009}. Some examples provided by the user included \textit{choice as}, \textit{sweet as}, and \textit{hot as}.

\subparagraph{Lexico-Grammar}

    Two users noted a change in the lexico-grammar of \acrshort{NZE}. One user noted that the pronoun in \textit{come with me} has been reduced to \textit{come with}; another user noted that the indefinite article \textit{an} was disappearing; and similarly, a third user noted that the pronoun \textit{we} was disappearing. Lastly, three users noted that \textit{by accident} was being replaced with \textit{on accident}. One user argued that \textit{on accident} was associated with American English and was surprised to hear the innovative form on \textit{Shortland Street} (1992–), a New Zealand prime-time soap opera. The final example of lexico-grammatical change was the shift from \textit{close to} to \textit{close with}.

\paragraph{Semantic Phenomena}

    I have summarised the user-informed semantic phenomena in Table \ref{tab:semantic_change}. All examples of user-informed semantic phenomena provided both a conservative (a) and an innovative (b) variant. I grouped these user-informed semantic phenomena by the processes involved in semantic change. These included broadening, narrowing, metonymy, pejoration, and polysemy. Some examples represented a combination of these processes.

\subparagraph{Broadening}

    This semantic process refers to the generalisation of a word. Users observed the semantic broadening of \textit{tramp} (see \gls{tramp}), where the conservative variant refers to a long walk (synonymous with \textit{hike}), while the innovative variant refers to a homeless person.

\subparagraph{Narrowing}

    This is the opposite of broadening and refers to the specialisation of a word. Users observed five examples of semantic narrowing in \textit{fag}, \textit{flannel}, \textit{football}, \textit{tea}, and \textit{pudding}. I have provided the conservative and innovative meanings in Table \ref{tab:semantic_change}. Of considerable interest to users was \textit{football}, where users were unsure whether the innovative variants (\textit{rugby} and \textit{soccer}) had a conservative meaning.

\subparagraph{Metonymy}

    Similar to homonymy, where a word may share a surface form but hold different meanings, this refers to the process where the meaning of a word is substituted by the meaning of a closely related word. Users observed two examples of metonymy in \textit{kiwi} and \textit{rubber(s)}.

\subparagraph{Pejoration}

    Some words may develop a negative meaning over time. Some users observed the pejoration of \textit{dick} and \textit{egg}. The conservative variant of \textit{dick} refers to the diminutive of Richard, while the innovative variant refers to an obnoxious person. One semantic phenomenon that received considerable interest from users was the pejoration of \textit{egg} (see \gls{egg}) in \acrshort{NZE}. Other examples included \textit{bum}, which was a combination of semantic pejoration and narrowing, while \textit{gay} was a combination of semantic pejoration and metaphor.

\subparagraph{Polysemy}

    Users noted two examples of polysemy which were \textit{tuna} and \textit{twink}. Related to phonological phenomena, one user noted phonological change in \textit{tuna} was now realised as /t\textbaru.na/ and not /t\textesh\textipa{u}.na/ (with yod-coalescence). Whereas other examples of yod-dropping (as in \textit{duke} and \textit{stupid}) were attributed to American English, some users noted that this change was the result of borrowing from te reo Māori \gls{tuna}.

    \begin{table}
        \scriptsize
        \centering
            
            \caption{User-informed Semantic Phenomena}
            \label{tab:semantic_change}
            
            \renewcommand{\arraystretch}{1.4}
            \begin{tabularx}{\textwidth}{*{3}{>{\arraybackslash}X}l}
            \toprule
            Variable & Conservative Variant & Innovative Variant & Semantic Process \\
            \midrule
                \addlinespace[1em]
                    tramp & a long walk & a homeless person & Broadening \\
                    kiwi & a flightless bird & a fruit & Metonymy \\
                    rubber(s) & an eraser & a condom & Metonymy \\
                    fag & a cigarette & clipping of faggot & Narrowing \\
                    flannel & a face cloth & a flannel shirt & Narrowing \\
                    football & rugby/soccer & rugby/soccer & Narrowing \\
                    tea & an evening meal & a beverage & Narrowing \\
                    pudding & dessert course & a type of dessert & Narrowing \\
                    dick & diminutive of Richard & an obnoxious person & Pejoration \\
                    gay & happy, joyful, lively & homosexual & Pejoration and Narrowing \\
                    bum & to borrow & to sodomise & Pejoration and Metaphor \\
                    tuna & a fish species & New Zealand longfin eel & Polysemy \\
                    twink & correction fluid & a young gay male & Polysemy \\
                \addlinespace[1em]
            \bottomrule
            \end{tabularx}

            \tablenoteparagraph{\textbf{Table Note}: This table organises user-informed lexical variables by the semantic processes underlying their variation, such as narrowing, broadening, or shifting. Columns identify the specific variable (\texttt{Variable}), the corresponding conservative variant (\texttt{Conservative Variant}), the innovative variant (\texttt{Innovative Variant}), and the categorised semantic process (\texttt{Semantic Process}); all data is sourced from Reddit.}
        
    \end{table}

\paragraph{Pragmatic Phenomena}

    User-informed pragmatic features included interjections (such as \textit{choice}, \textit{chur}, \textit{yeah nah}, \textit{cool bananas}, and \textit{bro}), hesitation fillers (\textit{like}, \textit{literally}, \textit{momentarily}, and \textit{surely}), and phrasemic fillers (such as \textit{intents and purposes}, \textit{this is everything}, \textit{I'm here for it}, \textit{I know right}, \textit{I mean}, \textit{its so fun}, \textit{I can't even}, \textit{I couldn't care less}, \textit{can hardly}, and \textit{whatever}). Users observed an increase in these interjections, hesitation fillers, and phrasemic fillers. One user provided a list of idiomatic phrases that they observed have decreased in usage. These differ from the idiomatic multi-word phrases in Selfpost 1 (see Table \ref{tab:idiomatic_phrases}).

    Some users have noted that there has been a change from \textit{hello} to alternates such as \textit{hey} and \textit{hi} beginning in the 1990s. One user, who used to live in New Zealand in the 1960s, noted that \textit{kia ora} was rarely used as a greeting then but is now more prevalent. Another noted that there has been a change in how people ask where someone lives, from \textit{where do you live?} to \textit{where do you stay?}. In terms of expressions of gratitude, one user noted that people no longer said \textit{ta}. The most controversial change noted by users was the increase of \textit{preciate it} and \textit{you're welcome} in response to \textit{thank you}. This was likely in light of sub-story one in Selfpost 1 (\texttt{155d1f3}) (see Section \ref{user:submission_post1}).

    \begin{table}
        \scriptsize
        \centering
            
            \caption{User-informed Idiomatic Phenomena}
            \label{tab:idiomatic_phrases}
            
            \renewcommand{\arraystretch}{1.4}
            \begin{tabularx}{\textwidth}{l}
            \toprule
            Idiomatic Phrases \\
            \midrule
                \addlinespace[1em]
                    a bird in the hand is worth two in the bush \\
                    a little bird told me \\
                    back to the drawing board \\  
                    beat around the bush \\
                    believe half of what you see, and nothing of what you hear \\
                    cost an arm and a leg \\ 
                    couldn't organise a piss up in a brewery \\
                    crook as buggery \\
                    hit the nail on the head \\
                    hold your horses \\
                    get out of the scratcher \\
                    marry in haste, repent at leisure \\
                    kill two birds with one stone \\  
                    running around like a blue arsed fly \\
                    stone the crows \\
                    you look like you dropped a pound and found a penny \\
                \addlinespace[1em]
            \bottomrule
            \end{tabularx}

            \tablenoteparagraph{\textbf{Table Note}: This table lists a selection of user-informed idiomatic phrases identified within the dataset. All featured phrases were contributed by a single user, providing a qualitative snapshot of individual idiomatic usage; the data is sourced from Reddit.}
        
    \end{table}

\section{Discussion}
\label{user:discussion}

    In the thematic analysis, users provided phonological, lexical, morphosyntactic, and pragmatic features in Selfpost 1 (\texttt{155d1f3}) that were considered corny. I observed an over-representation of multi-word phrases, idiomatic phrases, and listemes. This was followed by corny user-informed lexical features. As for Selfpost 2 (\texttt{1102c68}), users offered examples of \acrshort{NZE} including phonological, lexical, morphosyntactic, semantic, and pragmatic phenomena. The user-informed features in Selfpost 2 (\texttt{1102c68}) were characteristically a set of sociolinguistic variables where two or more linguistic features contrast but refer to the same meaning or concept.

    The findings from the thematic analysis also reinforced my earlier conclusions that, for users in \texttt{r/newzealand}, there was a link between language use and their place identity by contrasting linguistic variants that were and were not associated with \acrshort{NZE}. This further suggests language use has a role in place-making in place-based communities like \texttt{r/newzealand} that are associated with a geographic dialect community \citep{johnstone_place_2004}. In terms of enregisterment (\citealp{agha_social_2003}; \citealp{johnstone_pittsburghese_2009}), some users attempted to encapsulate the entire phonological system of \acrshort{NZE} with the phraseme \textit{fish n chips} ($\uparrow4$) as a way to distinguish \acrshort{NZE} from Australian English.
    
    I suspect a degree of priming from the \acrshort{OP}, as some users also included conservative and innovative variants to illustrate their observations. There was an over-representation of comments contrasting \acrshort{NZE} with American English. This was particularly evident from comments which have directly attributed language change in \acrshort{NZE} to the United States ($n=34$). This feeds into the profile of well-meaning self-appointed experts within these communities \citep{panek_understanding_2022}. As an example, one user incorrectly categorised blending in \textit{restaurant} and \textit{history} as yod-coalescence. Even though this does suggest users were aware of speech sound changes, they may not be aware of underlying phonological processes. More generally, the presence of innovative phonological phenomena does not necessarily suggest \acrshort{NZE} is adopting more phonological features associated with American English.

    On the other end of the spectrum were shitposters. The most controversial comment, with a score of 1, openly challenged the \acrshort{OP} in Submission 1 and their claim that \textit{math} was an acceptable variant in \textit{our variety} and should be treated as the \textit{standard}. As part of the user's justification, the user alluded to the word-formation process of clipping and justified their claim by introducing another variant, \textit{aluminum}, chiefly used in North American varieties of English in contrast to \textit{aluminium}. The user alluded to Stigler's Law of Eponymy \citep{stigler_stiglers_1980} to justify their preference for \textit{aluminum}. More specifically, the user noted that there was increased rhoticity \textit{creeping into the TikTok kids' vernacular} in reference to the short-form video-sharing social media platform.
    
    One user noted that \textit{[I] bought a plate to dinner} was ungrammatical. Another user argued that regardless of the verb's surface form, it was possible to determine the verb based on its context, using the examples \textit{I brought a new phone} and \textit{I bought a plate to the dinner} to illustrate their prediction. The user argued that the prescriptive distinction between \textit{bought} and \textit{brought} was \textit{old-fashioned}, leading to hypercorrections in the case of \textit{I brought it from Pak'nSave}. In line with the shitposter persona, some users have attributed \textit{brought}-\textit{bought} alternation to younger speakers of \acrshort{NZE} and exemplified public figures such as the 40th Prime Minister of New Zealand, Jacinda Ardern, as the progenitor of this change.

    My thick description of two submissions in \texttt{r/newzealand} has offered some insights into language use in its social context. The different text types each have their own communicative purpose. As an example, the body text of a selfpost can be a short prompt (\texttt{155d1f3}) or as long as an extended narrative (\texttt{1102c68}). Other text types, such as comments, also vary in length and communicative purpose. In the case of Selfpost 1 (\texttt{155d1f3}), just under half (45.7\%) of comments were no longer than one or two words and served as discourse markers throughout the comment thread. Comments can also be as long as the body text of a selfpost. In contrast to Twitter (X), hashtags and emojis are seldom used by users on Reddit. Additionally, the upvote score does not necessarily reflect positive affect.

\section{Conclusion and Key Findings}
\label{user:conclusion}

    In summary, I conclude that some users in \texttt{r/newzealand} do associate language use with their sense of place - more specifically, the subset of users who engaged with the posts. Users were most aware of lexical features associated with \acrshort{NZE}, but they also had an awareness of morphosyntactic (intensifier-as) and pragmatic phenomena (\textit{you can't beat Wellington on a good day}). In the case of Selfpost 2 (\texttt{1102c68}), most users were able to confidently identify conservative and innovative linguistic features associated with \acrshort{NZE} (albeit sometimes incorrectly). Some users even expressed confusion as to which features were the conservative variants, such as the meaning of \textit{football} in relation to \textit{rugby} and \textit{soccer}. In analysing the comments of the two submissions, I have established a list of user-informed features associated with \acrshort{NZE} in \texttt{r/newzealand}.

% -----------------------------
% Chapter 5: User-Informed Sociolinguistic Variables
% -----------------------------

\chapter{User-Informed Sociolinguistic Variables}
\markboth{User-Informed Sociolinguistic Variables}{}
\label{chap:user_variables}

\section{Chapter Outline}
\label{variable:chapter_outline}

    This chapter builds on the findings from the previous chapter by analysing the user-informed sociolinguistic variables. I provide the background and motivation for this analysis in Section \ref{variable:introduction}. In Section \ref{variable:methdology_sociolinguistic}, I describe my methodology and introduce my inner-circle country-level communities. I then present the results from my analysis of the user-informed lexical and morphosyntactic variables in Section \ref{variable:results}. I conclude the chapter with a discussion in Section \ref{variable:discussion}, which synthesises the findings from my two qualitative approaches, and I outline the key findings in Section \ref{variable:conclusion}. The insights from this chapter contribute to both \acrshort{SQ1} and \acrshort{SQ2}: notably, whether there is a relationship between geographic dialect communities and place-based communities.

\section{Background and Motivation }
\label{variable:introduction}

    The findings from Chapter \ref{chap:user_intuitions}: User Intuitions and Place Identity suggest that users in place-based communities associated with non-dominant varieties of English, such as \texttt{r/newzealand}, perceive a change in progress. More specifically, users have correlated the changes observed in \acrshort{NZE} with the influence of American English. However, linguistic perception rarely aligns perfectly with linguistic production \citep{preston_language_2010}. In the case of \acrshort{NZE}, speakers have identified more linguistic variation than structural evidence would typically suggest \citep{duhamel_end_2015}. There is evidence to suggest that American English has developed into the \textit{lingua franca} on Twitter\textsuperscript{X} over time, as evidenced by an increase in words associated with American English \citep{goncalves_mapping_2018}. I can document this change by evaluating the sociolinguistic variable \citep{labov_social_2006}.
    
    Since the seminal studies conducted by \citet{labov_sociolinguistic_1972}, the sociolinguistic variable has played a fundamental role in the understanding of language variation and change. A sociolinguistic variable is a linguistic feature with two or more forms (variants) that vary in production in relation to social factors such as age, class, or gender. This is because a relative increase in one variant presupposes a decrease in another. As an example, \textsc{rizz} is an age-graded sociolinguistic variable that emerged on Twitch.tv, with the conservative variant (\textit{charisma}) associated with older users and the innovative variant (\textit{rizz}) associated with younger users \citep{purba_rise_2025}. Sociolinguistic variables are not restricted to lexis but also encompass other linguistic features (such as phonology, morphosyntax, and semantics) as well as non-linguistic features.

    Naturally, British and American Englishes are often viewed in binary opposition in cases of diatopic (or geographic) variation in English \citep{hickey_legacies_2005}. This Americanisation discourse is not a new phenomenon, and it is frequently linked to globalisation \citep{phillipson_americanization_2014}. However, studies into \acrshort{NZE} have found an observed shift towards American English variants in only two of the 11 lexical variables, which I have summarised in Table \ref{tab:working_class} (\citealp{bayard_antipodean_1991}; \citealp{meyerhoff_lexical_1993}; \citealp{meyerhoff_globalisation_2003}). On the contrary, \citet{meyerhoff_globalisation_2003} argued that using British and American Englishes as benchmarks for \acrshort{NZE} has led to a reductionist view of language variation and change. I cannot simply view American English as a `category killer' \citep{klein_no_2000} when it comes to processes such as linguistic diffusion \citep{meyerhoff_globalisation_2003}.
    
    Since establishing an association between language use and place identity in the previous chapter, my next step is to evaluate the intuition of users and determine the role of these user-informed variables in \texttt{r/newzealand}. With the majority of studies focusing on Twitter\textsuperscript{X} (\citealp{jones_toward_2015}; \citealp{grieve_mapping_2018}; \citealp{grieve_mapping_2019}; \citealp{ilbury_using_2024}; \citealp{goncalves_mapping_2018}; \citealp{dijkstra_using_2021}), I address this gap in the literature by evaluating sociolinguistic variation across the place-based communities on Reddit to support the findings from \citet{hamre_geographic_2024}. By evaluating the user-informed sociolinguistic variables, I also bridge the gap between user perception and production in linguistic variation on social media \citep{nguyen_dialect_2021}. In summary, the purpose of this chapter is to evaluate user-informed variables by mapping them to actual language use across country-level place-based communities on Reddit.

    \begin{table}
        \scriptsize
        \centering
            
            \caption{Reported Use of Lexical Pairs in \acrshort{NZE}}
            \label{tab:working_class}
            
            \renewcommand{\arraystretch}{1.4}
                \begin{tabularx}{\textwidth}{*{5}{>{\centering\arraybackslash}X}}
                    \toprule
                    Conservative & Innovative & 1984-5 & 1989 & User-Informed \\
                    \midrule
                    \addlinespace[1em]
                    torch & flashlight & 89/4 & 97/2 & No \\
                    biscuit & cookie & 83/5 & 95/0 & Yes \\
                    petrol & gas & 78/4 & 88/7 & Yes \\
                    jersey & sweater & 75/8 & 86/4 & No \\
                    lift & elevator & 81/8 & 85/10 & Yes \\
                    serviette & napkin & 57/12 & 83/11 & No \\
                    bonnet (car) & hood & 77/12 & 76/24 & Yes \\
                    tin (food) & can & - & 76/24 & No \\
                    pictures & movies & 36/26 & 53/37 & No \\
                    trousers & pants & 60/7 & 46/41 & No \\
                    lorry & truck & 7/70 & 3/97 & Yes \\
                    \addlinespace[1em]
                    \bottomrule
                \end{tabularx}

            \tablenoteparagraph{\textbf{Table Note}: This table, adapted from \citet{meyerhoff_globalisation_2003}, compares lexical usage across 11 pairs of conservative and innovative variants within \acrshort{NZE}. Columns show the specific variants (\texttt{Conservative} and \texttt{Innovative}), their relative proportions as reported in earlier studies by \citet{bayard_me_1989} (\texttt{1984-5}) and \citet{meyerhoff_lexical_1993} (\texttt{1989}), and whether these pairs were identified by contemporary users in \texttt{r/newzealand}. Observations indicate a relative change in six of the 11 pairs, specifically showing a proportional decrease in the conservative variant for the pairs torch/flashlight and lift/elevator. Data is sourced from \citet{bayard_me_1989}, \citet{meyerhoff_lexical_1993}, and Reddit/Pushshift \citep{baumgartner_pushshift_2020}.}
        
    \end{table}

\section{Methodology}
\label{variable:methdology_sociolinguistic}

    My primary goal in this chapter is to determine the distribution of the user-informed sociolinguistic variables across the six country-level communities described in Section \ref{variable:data}. My secondary goal is to determine the suitability of these user-informed variables as a rudimentary benchmark for \acrshort{NZE} in downstream \acrshort{NLP} tasks such as language modelling. I apply techniques from corpus linguistics and \acrshort{NLP} to extract my features. In order to control for the production circumstances of Reddit, I restrict my analysis to comments (\gls{rcomm}).

\subsection{Data}
\label{variable:data}

    I extend my analysis beyond \texttt{r/newzealand} and incorporate other country-level communities on Reddit in order to determine the distribution of these user-informed sociolinguistic variables. My primary source of Reddit data comes from the Pushshift dumps, which are an ongoing data collection effort of the top 40,000 communities on Reddit through the Pushshift \acrshort{API} \citep{baumgartner_pushshift_2020}. I adopt the Three Circles of English model to support my sampling of candidate country-level communities associated with inner-circle varieties of English \citep{kachru_standards_1985}. Inner-circle varieties refer to countries that are considered the traditional bases of English and where English continues to be the primary language. The six inner-circle varieties of English and their associated place-based communities include: Canadian English (\texttt{r/canada}), American English (\texttt{r/usa}), Irish English (\texttt{r/ireland}), British English (\texttt{r/unitedkingdom}), Australian English (\texttt{r/australia}), and \acrshort{NZE} (\texttt{r/newzealand}).

    Of the inner-circle varieties, the largest place-based community was \texttt{r/canada} with over 4.2 million subscribers, and the smallest was \texttt{r/usa} with over 120,000 subscribers. A summary of the country-level communities is presented in Table \ref{tab:inner_circle}, including the date they were created, the number of subscribers as of June 2025, the estimated population of the corresponding country as of 2024, and the broad geographic location taken from the \textit{UNdata} profiles collated by the United Nations Statistics Division\footnote{\href{https://data.un.org/default.aspx}{https://data.un.org/default.aspx}}. I also note the status of the country-level community. The descriptive measures of the country-level corpus are presented in Table \ref{tab:country_ttr}, including the total number of observations per community with the calculated type–token ratio (\textsc{ttr}) measures broken down by text type. The purpose of the \textsc{ttr} is to offer a high-level measure of lexical diversity.

    \begin{table}
        \scriptsize
        \centering
            
            \caption{Summary of Inner-Circle Country-level Communities on Reddit}
            \label{tab:inner_circle}
            
            \renewcommand{\arraystretch}{1.4}
            \begin{tabularx}{\textwidth}{ll*{3}{>{\centering\arraybackslash}X}}
            \toprule
            \textbf{Community} & \textbf{Created} & \textbf{Members} & \textbf{Population} & \textbf{Region} \\
            \midrule
                \addlinespace[1em]
                \texttt{r/canada} & Jan 25, 2008 & 4,218,715 & 39,742,000 & Northern America \\
                \texttt{r/usa} & Jan 25, 2008 & 119,176 & 345,427,000 & Northern America \\
                \texttt{r/ireland} & Mar 3, 2008 & 1,212,527 & 5,255,000 & Northern Europe \\
                \texttt{r/unitedkingdom} & Apr 2, 2008 & 5,357,844 & 69,138,000 & Northern Europe \\
                \texttt{r/australia} & Jan 26, 2008 & 2,711,245 & 26,713,000 & Oceania \\
                \texttt{r/newzealand} & Mar 23, 2008 & 793,800 & 5,214,000 & Oceania \\
                \addlinespace[1em]
            \bottomrule
            \end{tabularx}

            \tablenoteparagraph{\textbf{Table Note}: This table summarises the six inner-circle country-level communities used as a sample of the six inner-circle varieties of English, detailing each community's name (\texttt{Community}), creation date (\textit{Created}), and subscriber count as of June 2025 (\textit{Members}). To provide demographic and geographic context, the table includes estimated population figures from the United Nations as of 2024 (\textit{Population}) and regional classifications based on United Nations designations (\textit{Region}). Excluding \texttt{r/usa}, a strong positive correlation exists between community membership and estimated population (Pearson's $R=0.979$). Data is sourced from Reddit/Pushshift \citep{baumgartner_pushshift_2020} and the United Nations.}
        
    \end{table}

\subsection{Sociolinguistic Variables}
\label{variable:variables}

    As the primary modality of Reddit data is written language, I focus on the lexical and morphosyntactic features, despite users providing phonological, lexical, morphosyntactic, semantic, and pragmatic features. This is not to say previous studies have not considered phonological variation based on non-standard orthography (\citealp{jones_toward_2015}; \citealp{dijkstra_using_2021}). My user-informed sociolinguistic variables were all derived from user comments in Chapter \ref{chap:user_intuitions}: User Intuitions and Place Identity. User-informed features with no variants were excluded. I established a list of 51 user-informed lexical variables, three user-informed morphosyntactic variables, and 13 user-informed semantic variables. Out of the 51 user-informed lexical variables, five were also listed by \citet{meyerhoff_lexical_1993}: \textsc{biscuit}, \textsc{fuel}, \textsc{lift}, \textsc{bonnet}, and \textsc{lorry}. Due to the low number of user-informed morphosyntactic variables, I analyse the distribution of two additional morphosyntactic constructions across the six country-level communities. I address semantic alternations in Chapter \ref{chap:dialect_classification}: Dialect Modelling and Embedding Models.

    \begin{table}
        \scriptsize
        \centering
            
            \caption{User-informed Sociolinguistic Variables}
            \label{tab:lex_variables1}
            
            \renewcommand{\arraystretch}{1.4}
                \begin{tabularx}{\textwidth}{lllX}
                    \toprule
                    Variable & Conservative & Innovative & Usage Notes \\
                    \midrule
                    \addlinespace[2em]
                    \multicolumn{4}{c}{\textbf{Lexical Phenomena}} \\
                    \addlinespace[1em]
                    \textsc{plane} & aeroplane & airplane & chiefly North American \\
                    \textsc{aluminium} & aluminium & aluminum & chiefly North American \\
                    \textsc{lawn} & lawn & yard & chiefly North American and dialect \\
                    \textsc{lego} & lego bricks & legos & chiefly North American \\
                    \textsc{lift} & lift & elevator & chiefly North American \\
                    \textsc{biscuit} & biscuits & cookies & chiefly North American \\
                    \textsc{bogan} & bogans & rednecks & chiefly North American \\ 
                    \textsc{bonnet} & bonnet & hood & originally and chiefly U.S. \\
                    \textsc{boot} & boot & trunk & North American \\
                    \textsc{friend} & boyfriend, girlfriend & partner & \\
                    \textsc{fruit} & chinese gooseberries & kiwifruit & chiefly New Zealand \\
                    \textsc{chip} & chips, crisps & fries & originally North American \\
                    \textsc{school} & college, high school & secondary school & \\
                    \textsc{curtain} & curtains & drapes & chiefly North American \\
                    \textsc{meal} & tea time & dinner time & \\
                    \textsc{dummy} & dummy & pacifier & North American \\
                    \textsc{fire} & fire engine & fire truck & chiefly North American \\
                    \textsc{fizzy} & fizzy drink, soft drink & soda & \\
                    \textsc{flannel} & flannel & wash-cloth & United States \\
                    \textsc{flat} & flat & apartment & chiefly North American \\
                    \textsc{ball} & football & soccer, rugby & chiefly British, chiefly Australian \\
                    \textsc{path} & footpath & sidewalk & chiefly North American \\
                    \textsc{fuel} & fuel, petrol & gas & chiefly North American (originally colloquial) \\
                    \textsc{garbage} & garbage, rubbish & trash & chiefly United States \\
                    \textsc{goingout} & going out & dating & originally United States \\
                    \textsc{hotdog} & hot dog & corn dog & North American \\
                    \textsc{iceblock} & ice block, ice lolly & popsicle & originally United States \\
                    \textsc{knicker} & knickers & panties & originally United States \\
                    \textsc{lolly} & lollies, sweets & candies & chiefly North American \\
                    \textsc{lorry} & lorry & truck & originally U.S. \\
                    \textsc{maths} & maths & math & chiefly North American \\
                    \textsc{nappy} & nappy & diaper & chiefly North American \\
                    \textsc{oblige} & obliged & obligated & chiefly North American \\
                    \textsc{poo} & poo & poop & chiefly United States \\
                    \textsc{pudding} & pudding & dessert & Unites States, now also in British Usage \\
                    \textsc{salvo} & salvation army & salvo & Australian \\
                    \addlinespace[1em]
                    \bottomrule
                \end{tabularx}

            \tablenoteparagraph{\textbf{Table Note}: This table organises user-informed variables by their respective linguistic phenomena, contrasting conservative variants with their innovative counterparts. Columns detail the specific variable name, the conservative form, the innovative form, and relevant usage notes for the innovative variant as documented in the Oxford English Dictionary. All data is sourced from Reddit/Pushshift \citep{baumgartner_pushshift_2020}.}
        
    \end{table}

    \begin{table}
        \scriptsize
        \centering
            
            \caption{User-informed Sociolinguistic Variables (\textit{Continued})}
            \label{tab:lex_variables2}
            
            \renewcommand{\arraystretch}{1.4}
                \begin{tabularx}{\textwidth}{lllX}
                    \toprule
                    Variable & Conservative & Innovative & Usage Notes \\
                    \midrule
                    \addlinespace[2em]
                    \multicolumn{4}{c}{\textbf{Lexical Phenomena} (Continued)} \\
                    \addlinespace[1em]
                    \textsc{takeaway} & takeaway & takeout & chiefly North American \\
                    \textsc{tea} & tea & dinner & \\
                    \textsc{towel} & tea towel & dishcloth & \\
                    \textsc{telly} & telly & TV & originally United States \\
                    \textsc{toilet} & toilet, bathroom, restroom & potty & originally North American \\
                    \textsc{tramp1} & tramp & hike & colloquial. Originally dialect and U.S. \\
                    \textsc{tramp2} & tramp & hike & colloquial. Originally dialect and U.S. \\
                    \textsc{tramp3} & tramping & hiking & colloquial. Originally dialect and U.S. \\
                    \textsc{tuna} & tuna & tuna fish & \\
                    \textsc{twink} & twink & whiteout & originally North American \\
                    \textsc{ute} & ute & pickup truck & originally United States \\
                    \textsc{vivid} & vivid & sharpie & originally United States \\
                    \textsc{vomit} & vomit & puke & \\
                    \textsc{nekminnit} & next minute & nek minnit & \\
                    \addlinespace[2em]
                    \multicolumn{4}{c}{\textbf{Morphosyntactic Phenomena}} \\
                    \addlinespace[1em]
                    \textsc{syn1} & swimming team & swim team & \\
                    \textsc{syn2} & has gone & has went & \\
                    \textsc{syn3} & by accident & on accident & \\
                    \addlinespace[2em]
                    \multicolumn{4}{c}{\textbf{Semantic Phenomena}} \\
                    \addlinespace[1em]
                    \textsc{bum} & to acquire & to have anal sex &  \\
                    \textsc{dick} & diminutive of Richard & an annoying person & slang (originally U.S.) \\
                    \textsc{fag} & a cigarette & a gay man & North American slang \\
                    \textsc{football} & a ball game & soccer, rugby & chiefly British, chiefly Australian \\
                    \textsc{flannel} & a piece of cloth & a garment &  \\
                    \textsc{gay} & light-hearted & homosexual & Originally U.S. slang \\
                    \textsc{kiwi} & a bird & a fruit &  \\
                    \textsc{pudding} & a sweet course & a sweet dish & North American \\
                    \textsc{rubber} & an eraser & a condom & slang; chiefly North American \\
                    \textsc{tea} & an evening meal & a beverage &  \\
                    \textsc{tramp} & to walk & a vagrant &  \\
                    \textsc{tuna} & a fish & an eel & New Zealand \\
                    \textsc{twink} & correction fluid  & a gay man & slang; originally U.S. \\
                    \addlinespace[1em]
                    \bottomrule
                \end{tabularx}

            \tablenoteparagraph{\textbf{Table Note}: This table organises user-informed variables by their respective linguistic phenomena, contrasting conservative variants with their innovative counterparts. Columns detail the specific variable name, the conservative form, the innovative form, and relevant usage notes for the innovative variant as documented in the Oxford English Dictionary. All data is sourced from Reddit/Pushshift \citep{baumgartner_pushshift_2020}.}
        
    \end{table}

\subsection{Feature Extraction}

    After applying initial data cleaning procedures to the language data, such as removing deleted posts and comments by users, I tokenise the observations and then apply parts-of-speech tags using the Textblob Averaged Perceptron Tagger from the \texttt{nltk} Python package library. The purpose of this step is to identify words based on their grammatical context, which minimises the risks associated with relying solely on the surface form of these features. I also split the features into one-word (unigram) and two-word (bigram) pairs to account for the orthographic variants of compounds.

\subsection{Evaluation}

    Once I have extracted the tokens from my country-level communities, I generate two measures: token frequency ($n$) and the proportion as a percentage. While the token frequency provides a general indication of usage, the proportion allows me to identify distributional patterns within and across country-level communities.More importantly, I want to determine whether the conservative variants, as judged by users, have in fact decreased. As part of my analysis, I group these variables based on their distributional patterns. After carrying out my exploratory data analysis, I use $k$-means clustering and \acrshort{PCA} to validate these groupings.By using \acrshort{PCA} for dimensionality reduction, I can more easily visualise the clusters of lexical and morphosyntactic variants that share similar distributional profiles across the different subreddits.

\section{Results}
\label{variable:results}

    I was able to extract the 51 user-informed lexical and three user-informed morphosyntactic variables across the six inner-circle communities. There were some noticeable data gaps for some variants. More specifically, the results for \texttt{r/usa} should be interpreted with caution due to the lower than expected token frequencies, which are a result of its relatively small subscription base.
    
\subsection{Lexical Variables}
\label{variable:lexical_variables}

    I begin with my analysis of the user-informed lexical variables. I first focus on 50 of the 51 user-informed lexical variables; I discuss the results for \textsc{nekminnit} separately as a case study, as it is a user-informed lexical variable that is both specific to \acrshort{NZE} and originated from an internet meme \citep{otago_daily_times_nek_2011}.

\subsubsection{Distributional Groupings}

    I grouped the 50 user-informed lexical variables based on their distributional patterns. As an example, for the user-informed variable \textsc{boot}, all six place-based communities had a percentage frequency greater than 50\% for the conservative variant (\textit{boot}). This suggested a community-level preference for the conservative variant (\textit{boot}) over the innovative variant (\textit{trunk}). Therefore, I have grouped this user-informed lexical variable in the conservative dominant category. Across the 50 user-informed lexical variables, I observed four different distributional patterns: conservative dominant, innovative dominant, community-specific groupings, and community-specific outliers. I describe each of the distributional patterns below.

\paragraph{Conservative Dominant}

    My first distributional pattern of interest was where the conservative variants had the greatest proportion across all country-level communities. There were twelve user-informed lexical variables associated with this distributional pattern, accounting for just under a quarter of the user-informed lexical variables. The community-level preference for the conservative variants in \texttt{r/newzealand} would suggest that the intuition of users was largely incorrect. Some variable categories were English dialect-specific (such as \textsc{curtain}). One surprising observation from the conservative dominant user-informed variables was the lack of variability across the user-informed variables. Of interest to \acrshort{NZE} was the user-informed variable \textsc{goingout}, where I observed a ratio of 63:37 (rounded to two decimal places) between the conservative and innovative variants. This may suggest a change in progress for this variable. The results for this distributional pattern are presented in Table \ref{tab:conservative_only}.

\paragraph{Innovative Dominant}

    In contrast to my first distributional pattern, my second distributional pattern was where the innovative variants had the highest proportion across all country-level communities. I described this as the innovative dominant distribution. I observed this distributional pattern in ten user-informed variables. Of interest to my analysis are the variables associated with the user-informed variable \textsc{tramp}, where I observed a community-level preference in \texttt{r/newzealand} for the innovative variant \textit{hike} (53.7\%). Once again, this may suggest a change in progress for this variable. Other variables were fairly advanced in their change in progress towards the innovative variant (>70\%). The results for this distributional pattern are presented in Table \ref{tab:innovative_only}.

\paragraph{Community-Specific Groupings}

    My third distributional pattern of interest was community-specific groupings. I clustered the user-informed lexical variables by country-level communities that shared a similar distributional pattern. These community-specific groupings suggest a degree of co-variation between country-level communities. I observed this pattern in nineteen variable categories, making it by far the largest distributional pattern. Some major community-specific groupings included the Southern Hemisphere cluster (\texttt{NZ}/\texttt{AU}), the British Isles cluster (\texttt{UK}/\texttt{IE}), and the North American cluster (\texttt{CA}/\texttt{US}). I also observed a British Isles plus New Zealand cluster (\texttt{NZ}/\texttt{UK}/\texttt{IE}) and a British Isles plus United States cluster (\texttt{UK}/\texttt{IE}/\texttt{US}). Within the Southern Hemisphere cluster, I observed a community preference in \texttt{r/newzealand} for the conservative variants in the user-informed variables \textsc{ute} (98.2\%), \textsc{lolly} (94.9\%), \textsc{bogan} (70.4\%), \textsc{fuel} (67.8\%), and \textsc{iceblock} (58.3\%). The results for this distributional pattern are presented in Tables \ref{tab:community_groups1} and \ref{tab:community_groups2}.

\paragraph{Community-Specific Outliers}

    My fourth and final distributional pattern of interest was where I observed outliers specific to one community. Except for \texttt{r/ireland}, there were variable categories associated with each of the country-level communities. I observed this pattern in eight variable categories. Of interest to \acrshort{NZE}, there was a community-level preference in \texttt{r/newzealand} for the user-informed lexical variable \textsc{lawn} (56.6\%). The results for this distributional pattern are presented in Table \ref{tab:outliers}.

    \begin{table}[p]
        \scriptsize
        \centering
            
            \caption{User-Informed Lexis (Conservative Dominant)}
            \label{tab:conservative_only}
            
            \renewcommand{\arraystretch}{1.4}
                \begin{tabularx}{\textwidth}{p{1.5cm}l
                  *{6}{>{\centering\arraybackslash}X} |
                  >{\centering\arraybackslash}p{1.5cm}}
                    \toprule
                    Variable & Variant 
                    & \textsc{nz} & \textsc{au} & \textsc{uk} 
                    & \textsc{ie} & \textsc{ca} & \textsc{us}
                    & $n$ \\
                    \midrule

                    \addlinespace[1em]

                    \multirow[c]{4}{*}{\textsc{toilet}} & \underline{toilet} & \textbf{62.2} & \textbf{74.1} & \textbf{68.3} & \textbf{61.2} & \textbf{50.7} & \textbf{54.7} & 53,795 \\
                    & \underline{bathroom} & 36.8 & 24.8 & 30.8 & 37.8 & 46.6 & 38.4 & 29,845 \\
                    & \underline{restroom} & 0.6 & 0.7 & 0.5 & 0.7 & 2.0 & 7.0 & 799 \\
                    & potty & 0.4 & 0.4 & 0.4 & 0.3 & 0.7 & - & 368 \\
                    \addlinespace[1em]

                    \multirow[c]{2}{*}{\textsc{tuna}} & \underline{tuna} & \textbf{97.2} & \textbf{98.7} & \textbf{98.2} & \textbf{98.1} & \textbf{97.3} & \textbf{100.0} & 4,989 \\
                    & tuna fish* & 2.8 & 1.3 & 1.8 & 1.9 & 2.7 & - & 106 \\
                    \addlinespace[1em]

                     \multirow[c]{2}{*}{\textsc{curtain}} & \underline{curtains} & \textbf{96.0} & \textbf{96.6} & \textbf{97.8} & \textbf{97.9} & \textbf{81.3} & \textbf{80.0} & 6,965 \\
                     & drapes & 4.0 & 3.4 & 2.2 & 2.1 & 18.7 & 20.0 & 357 \\
                    \addlinespace[1em]

                    \multirow[c]{2}{*}{\textsc{teatowel}} & \underline{tea towel}* & \textbf{93.3} & \textbf{92.7} & \textbf{76.8} & \textbf{87.0} & \textbf{66.7} & \textbf{100.0} & 903 \\
                    & dishcloth* & 6.7 & 7.3 & 23.2 & 13.0 & 33.3 & - & 114 \\
                    \addlinespace[1em]

                    \multirow[c]{2}{*}{\textsc{hotdog}} & \underline{hot dog}* & \textbf{93.2} & \textbf{97.8} & \textbf{99.2} & \textbf{96.7} & \textbf{97.9} & \textbf{90.0} & 3,552 \\
                    & corn dog* & 6.8 & 2.2 & 0.8 & 3.3 & 2.1 & 10.0 & 116 \\
                    \addlinespace[1em]

                    \multirow[c]{2}{*}{\textsc{dummy}} & \underline{dummy} & \textbf{91.7} & \textbf{96.5} & \textbf{95.9} & \textbf{96.7} & \textbf{97.5} & \textbf{92.9} & 4,709 \\
                     & pacifier & 8.3 & 3.5 & 4.1 & 3.3 & 2.5 & 7.1 & 186 \\
                    \addlinespace[1em]

                    \multirow[c]{2}{*}{\textsc{flannel}} & \underline{flannel} & \textbf{90.9} & \textbf{87.3} & \textbf{98.4} & \textbf{72.9} & \textbf{93.1} & \textbf{100.0} & 752 \\
                    & washcloth* & 9.1 & 12.7 & 1.6 & 27.1 & 6.9 & - & 84 \\
                    \addlinespace[1em]

                     \multirow[c]{2}{*}{\textsc{boot}} & \underline{boot} & \textbf{89.1} & \textbf{93.4} & \textbf{92.2} & \textbf{94.6} & \textbf{74.9} & \textbf{75.0} & 30,440 \\
                     & trunk & 10.9 & 6.6 & 7.8 & 5.4 & 25.1 & 25.0 & 4,452 \\
                    \addlinespace[1em]

                     \multirow[c]{3}{*}{\textsc{chip}} & \underline{chips} & \textbf{84.2} & \textbf{85.2} & \textbf{67.9} & \textbf{60.1} & \textbf{57.5} & \textbf{57.6} & 62,567 \\
                     & \underline{crisps} & 2.4 & 2.4 & 24.0 & 28.9 & 1.0 & 4.8 & 10,187 \\
                     & fries & 13.4 & 12.4 & 8.2 & 11.0 & 41.5 & 37.6 & 16,054 \\   
                    \addlinespace[1em]

                    \multirow[c]{2}{*}{\textsc{takeaway}} & \underline{takeaway}* & \textbf{86.9} & \textbf{90.6} & \textbf{94.8} & \textbf{93.2} & \textbf{55.2} & \textbf{87.5} & 14,019 \\
                    & takeout & 13.1 & 9.4 & 5.2 & 6.8 & 44.8 & 12.5 & 2,970 \\
                    \addlinespace[1em]
                    
                     \multirow[c]{2}{*}{\textsc{goingout}} & \underline{going out} & \textbf{62.8} & \textbf{68.2} & \textbf{72.6} & \textbf{62.8} & \textbf{67.3} & \textbf{61.5} & 35,521 \\
                     & dating & 37.2 & 31.8 & 27.4 & 37.2 & 32.7 & 38.5 & 18,071 \\
                    \addlinespace[1em]
                    
                    \bottomrule
                \end{tabularx}

            \tablenoteparagraph{\textbf{Table Note}: This table details the percentage distribution and total token frequency ($n$) for user-informed lexical variables and their variants across six inner-circle country-level communities: \texttt{r/newzealand} (\textsc{nz}), \texttt{r/australia} (\textsc{au}), \texttt{r/unitedkingdom} (\textsc{uk}), \texttt{r/ireland} (\textsc{ie}), \texttt{r/canada} (\textsc{ca}), and \texttt{r/usa} (\textsc{us}). Percentages for each community are rounded to one decimal place and sum to 100\%, with conservative variants underlined and orthographic variants for compounds combined and marked with an asterisk (*). The most frequent variant within each category-community group is identified, while a hyphen (-) represents a variant with zero frequency. All data is sourced from Reddit/Pushshift \citep{baumgartner_pushshift_2020}.}
        
    \end{table}

    \begin{table}[p]
        \scriptsize
        \centering
            
            \caption{User-Informed Lexis (Innovative Dominant)}
            \label{tab:innovative_only}
            
            \renewcommand{\arraystretch}{1.4}
                \begin{tabularx}{\textwidth}{p{1.5cm}l
                  *{6}{>{\centering\arraybackslash}X} |
                  >{\centering\arraybackslash}p{1.5cm} }
                    \toprule
                    Variable & Variant 
                    & \textsc{nz} & \textsc{au} & \textsc{uk} 
                    & \textsc{ie} & \textsc{ca} & \textsc{us}
                    & $n$ \\
                    \midrule
                    
                    \addlinespace[1em]
                    
                    \multirow[c]{2}{*}{\textsc{tramp3}} & \underline{tramping} & 46.3 & 3.6 & 5.1 & 1.3 & 0.7 & - & 1,948 \\
                    & hiking & \textbf{53.7} & \textbf{96.4} & \textbf{94.9} & \textbf{98.7} & \textbf{99.3} & \textbf{100.0} & 10,455 \\
                    \addlinespace[1em]

                    \multirow[c]{2}{*}{\textsc{plane}} & \underline{aeroplane} & 29.4 & 39.3 & 42.4 & 18.9 & 1.6 & - & 1,347 \\
                    & airplane & \textbf{70.6} & \textbf{60.7} & \textbf{57.6} & \textbf{81.1} & \textbf{98.4} & \textbf{100.0} & 7,706 \\
                    \addlinespace[1em]
                    
                    \multirow[c]{2}{*}{\textsc{bonnet}} & \underline{bonnet} & 27.8 & 32.3 & 28.0 & 37.4 & 3.4 & 4.3 & 3,286 \\
                    & hood & \textbf{72.2} & \textbf{67.7} & \textbf{72.0} & \textbf{62.6} & \textbf{96.6} & \textbf{95.7} & 11,138 \\
                    \addlinespace[1em]

                    \multirow[c]{2}{*}{\textsc{lego}} & \underline{lego bricks} & 22.9 & 32.5 & 48.6 & 29.4 & 12.5 & - & 123 \\
                    & legos & \textbf{77.1} & \textbf{67.5} & \textbf{51.4} & \textbf{70.6} & \textbf{87.5} & \textbf{100.0} & 344 \\
                    \addlinespace[1em]

                    \multirow[c]{2}{*}{\textsc{tramp1}} & \underline{tramp} (\textit{n}) & 21.4 & 4.7 & 20.8 & 13.9 & 0.9 & 3.7 & 1,661 \\
                    & hike (\textit{n}) & \textbf{78.6} & \textbf{95.3} & \textbf{79.2} & \textbf{86.1} & \textbf{99.1} & \textbf{96.3} & 18,454 \\
                    \addlinespace[1em]

                    \multirow[c]{3}{*}{\textsc{friend}} & \underline{boyfriend} & 6.8 & 8.0 & 10.2 & 13.1 & 10.9 & 19.6 & 18,424 \\
                    & \underline{girlfriend} & 11.9 & 14.7 & 31.7 & 28.9 & 25.5 & 37.0 & 40,802 \\
                    & partner & \textbf{81.3} & \textbf{77.2} & \textbf{58.1} & \textbf{58.0} & \textbf{63.6} & \textbf{43.5} & 131,444 \\
                    \addlinespace[1em]

                    \multirow[c]{2}{*}{\textsc{nappy}} & \underline{nappy} & 16.7 & 14.8 & 20.0 & 18.0 & - & - & 69 \\
                    & diaper & \textbf{83.3} & \textbf{85.2} & \textbf{80.0} & \textbf{82.0} & \textbf{100.0} & \textbf{100.0} & 1,148 \\
                    \addlinespace[1em]   

                    \multirow[c]{2}{*}{\textsc{tramp2}} & \underline{tramp} (\textit{v}) & 14.5 & 3.2 & 3.4 & 4.5 & 0.6 & - & 262 \\
                    & hike (\textit{v}) & \textbf{85.5} & \textbf{96.8} & \textbf{96.6} & \textbf{95.5} & \textbf{99.4} & \textbf{100.0} & 5,356 \\
                    \addlinespace[1em]
                    
                    \multirow[c]{2}{*}{\textsc{fruit}} & \underline{chinese gooseberries} & 2.7 & 16.2 & 25.0 & - & - & - & 77 \\
                    & kiwifruit* & \textbf{97.3} & \textbf{83.8} & \textbf{75.0} & \textbf{100.0} & \textbf{100.0} & - & 2,309 \\
                    \addlinespace[1em]   

                    \multirow[c]{2}{*}{\textsc{telly}} & \underline{telly} & 0.2 & 0.3 & 0.6 & 0.9 & - & - & 1,233 \\
                    & tv & \textbf{99.8} & \textbf{99.7} & \textbf{99.4} & \textbf{99.1} & \textbf{100.0} & \textbf{100.0} & 307,172 \\
                    \addlinespace[1em]
                    
                    \bottomrule
                \end{tabularx}

            \tablenoteparagraph{\textbf{Table Note}: This table details the percentage distribution and total token frequency ($n$) for user-informed lexical variables and their variants across six inner-circle country-level communities: \texttt{r/newzealand} (\textsc{nz}), \texttt{r/australia} (\textsc{au}), \texttt{r/unitedkingdom} (\textsc{uk}), \texttt{r/ireland} (\textsc{ie}), \texttt{r/canada} (\textsc{ca}), and \texttt{r/usa} (\textsc{us}). Percentages for each community are rounded to one decimal place and sum to 100\%, with conservative variants underlined and orthographic variants for compounds combined and marked with an asterisk (*). The most frequent variant within each category-community group is identified, while a hyphen (-) represents a variant with zero frequency. All data is sourced from Reddit/Pushshift \citep{baumgartner_pushshift_2020}.}
        
    \end{table}

    \begin{table}[p]
        \scriptsize
        \centering
            
            \caption{User-Informed Lexis (Community-Specific Groupings)}
            \label{tab:community_groups1}
            
            \renewcommand{\arraystretch}{1.4}
                \begin{tabularx}{\textwidth}{p{1.5cm}l
                  *{6}{>{\centering\arraybackslash}X} |
                  >{\centering\arraybackslash}p{1.5cm} }
                    \toprule
                    Variable & Variant 
                    & \textsc{nz} & \textsc{au} & \textsc{uk} 
                    & \textsc{ie} & \textsc{ca} & \textsc{us}
                    & $n$ \\
                    \midrule
                    \addlinespace[1em]
                    
                    \multicolumn{9}{c}{Southern Hemisphere Cluster} \\
                    \addlinespace[1em]
                    
                     \multirow[c]{2}{*}{\textsc{ute}} & \underline{ute} & \textbf{98.2} & \textbf{97.2} & 12.1 & 17.1 & 0.5 & - & 7,840 \\
                     & pickup truck & 1.8 & 2.8 & \textbf{87.9} & \textbf{82.9} & \textbf{99.5} & \textbf{100.0} & 1,817 \\
                    \addlinespace[1em]

                     \multirow[c]{3}{*}{\textsc{lolly}} & \underline{lollies} & \textbf{76.1} & \textbf{77.1} & 3.9 & 3.1 & 0.8 & - & 5,744 \\
                     & \underline{sweets} & 18.8 & 17.1 & \textbf{93.6} & \textbf{93.6} & 33.1 & \textbf{66.7} & 7,503 \\
                     & candies & 5.1 & 5.7 & 2.5 & 3.3 & \textbf{66.1} & 33.3 & 1,311 \\
                    \addlinespace[1em]

                     \multirow[c]{2}{*}{\textsc{bogan}} & \underline{bogans} & \textbf{70.4} & \textbf{87.9} & 17.5 & 12.8 & 1.2 & 6.2 & 12,977 \\
                     & rednecks* & 29.6 & 12.1 & \textbf{82.5} & \textbf{87.2} & \textbf{98.8} & \textbf{93.8} & 6,333 \\
                    \addlinespace[1em]

                    \multirow[c]{2}{*}{\textsc{vivid}} & \underline{vivid} & \textbf{57.4} & \textbf{69.7} & 47.7 & 44.4 & 25.7 & - & 783 \\
                    & sharpie & 42.6 & 30.3 & \textbf{52.3} & \textbf{55.6} & \textbf{74.3} & \textbf{100.0} & 727 \\
                    \addlinespace[1em]

                     \multirow[c]{3}{*}{\textsc{fuel}} & \underline{fuel} & \textbf{38.1} & \textbf{42.6} & 34.4 & 29.5 & 24.7 & 17.4 & 142,561 \\
                     & \underline{petrol} & 29.7 & 16.3 & 18.6 & 24.2 & 0.6 & 3.6 & 56,253 \\
                     & gas & 32.2 & 41.1 & \textbf{47.0} & \textbf{46.3} & \textbf{74.7} & \textbf{79.0} & 245,384 \\
                    \addlinespace[1em]

                     \multirow[c]{3}{*}{\textsc{iceblock}} & \underline{ice block}* & \textbf{57.0} & \textbf{55.7} & 3.8 & 9.8 & 8.1 & - & 374 \\
                     & \underline{ice lolly}* & 1.3 & 1.8 & \textbf{61.5} & \textbf{66.1} & - & - & 115 \\
                     & popsicle & 41.7 & 42.5 & 34.6 & 24.1 & \textbf{91.9} & - & 548 \\
                    \addlinespace[2em]

                    \multicolumn{9}{c}{British Isles plus New Zealand Cluster} \\
                    \addlinespace[1em]

                    \multirow[c]{2}{*}{\textsc{twink}} & \underline{twink} & \textbf{68.9} & 40.0 & \textbf{65.0} & \textbf{96.5} & 25.7 & - & 694 \\
                    & whiteout* & 31.1 & \textbf{60.0} & 35.0 & 3.5 & \textbf{74.3} & \textbf{100.0} & 287 \\
                    \addlinespace[2em]
                    
                    \multicolumn{9}{c}{British Isles Cluster} \\
                    \addlinespace[1em]

                    \multirow[c]{2}{*}{\textsc{fire}} & \underline{fire engine}* & 40.6 & 20.2 & \textbf{87.8} & \textbf{62.4} & 5.4 & - & 555 \\
                    & fire truck* & \textbf{59.4} & \textbf{79.8} & 12.2 & 37.6 & \textbf{94.6} & - & 1070 \\
                    \addlinespace[1em]
                    
                    \multirow[c]{3}{*}{\textsc{garbage}} & \underline{garbage} & \textbf{40.2} & \textbf{50.7} & 30.3 & 30.0 & \textbf{75.2} & \textbf{68.4} & 85,267 \\
                    & \underline{rubbish} & 28.8 & 22.9 & \textbf{47.4} & \textbf{42.0} & 1.1 & 2.5 & 23,589 \\
                    & trash & 31.0 & 26.5 & 22.3 & 28.1 & 23.7 & 29.1 & 37,319 \\
                    \addlinespace[1em]
                    
                    \bottomrule
                \end{tabularx}

            \tablenoteparagraph{\textbf{Table Note}: This table details the percentage distribution and total token frequency ($n$) for user-informed lexical variables and their variants across six inner-circle country-level communities: \texttt{r/newzealand} (\textsc{nz}), \texttt{r/australia} (\textsc{au}), \texttt{r/unitedkingdom} (\textsc{uk}), \texttt{r/ireland} (\textsc{ie}), \texttt{r/canada} (\textsc{ca}), and \texttt{r/usa} (\textsc{us}). Percentages for each community are rounded to one decimal place and sum to 100\%, with conservative variants underlined and orthographic variants for compounds combined and marked with an asterisk (*). The most frequent variant within each category-community group is identified, while a hyphen (-) represents a variant with zero frequency. All data is sourced from Reddit/Pushshift \citep{baumgartner_pushshift_2020}.}
        
    \end{table}

    \begin{table}[p]
        \scriptsize
        \centering
            
            \caption{User-informed Lexis (Community-Specific Groupings Continued)}
            \label{tab:community_groups2}
            
            \renewcommand{\arraystretch}{1.4}
                \begin{tabularx}{\textwidth}{p{1.5cm}l
                  *{6}{>{\centering\arraybackslash}X} |
                  >{\centering\arraybackslash}p{1.5cm} }
                    \toprule
                    Variable & Variant 
                    & \textsc{nz} & \textsc{au} & \textsc{uk} 
                    & \textsc{ie} & \textsc{ca} & \textsc{us}
                    & $n$ \\
                        \midrule
                    \addlinespace[1em]
                    \multicolumn{9}{c}{British Isles plus United States Cluster} \\
                    \addlinespace[1em]

                    \multirow[c]{3}{*}{\textsc{school}} & \underline{college} & 31.0 & 28.1 & \textbf{68.2} & \textbf{85.9} & 47.7 & \textbf{77.4} & 138,888 \\
                     & \underline{high school} & \textbf{64.4} & \textbf{69.6} & 15.7 & 2.0 & \textbf{50.8} & 22.2 & 85,386 \\
                     & secondary school & 4.6 & 2.3 & 16.2 & 12.1 & 1.5 & 0.4 & 15,664 \\
                    \addlinespace[1em]

                     \multirow[c]{2}{*}{\textsc{tea}} & \underline{tea} & 40.0 & 43.4 & \textbf{73.2} & \textbf{60.2} & 42.1 & \textbf{51.9} & 73,101 \\
                     & dinner & \textbf{60.0} & \textbf{56.6} & 26.8 & 39.8 & \textbf{57.9} & 48.1 & 63,547 \\
                    \addlinespace[1em]
                    
                     \multirow[c]{2}{*}{\textsc{pudding}} & \underline{pudding} & 37.5 & 44.3 & \textbf{81.2} & \textbf{82.5} & 48.8 & \textbf{66.7} & 7,960 \\
                     & dessert & \textbf{62.5} & \textbf{55.7} & 18.8 & 17.5 & \textbf{51.2} & 33.3 & 4,438 \\
                    \addlinespace[2em]
                    
                    \multicolumn{9}{c}{North American Cluster} \\
                    \addlinespace[1em]
                    
                     \multirow[c]{2}{*}{\textsc{path}} & \underline{footpath} & \textbf{87.5} & \textbf{84.0} & \textbf{76.0} & \textbf{88.8} & 0.5 & - & 12,335 \\
                     & sidewalk & 12.5 & 16.0 & 24.0 & 11.2 & \textbf{99.5} & \textbf{100.0} & 7,054 \\
                    \addlinespace[1em]
                    
                     \multirow[c]{2}{*}{\textsc{lift}} & \underline{lift} & \textbf{79.3} & \textbf{76.7} & \textbf{88.5} & \textbf{90.0} & 48.1 & 33.3 & 13,087 \\
                     & elevator & 20.7 & 23.3 & 11.5 & 10.0 & \textbf{51.9} & \textbf{66.7} & 4,473 \\
                    \addlinespace[1em]

                     \multirow[c]{2}{*}{\textsc{aluminium}} & \underline{aluminium} & \textbf{76.4} & \textbf{82.4} & \textbf{81.2} & \textbf{71.0} & 14.1 & 4.5 & 5,735 \\
                     & aluminum & 23.6 & 17.6 & 18.8 & 29.0 & \textbf{85.9} & \textbf{95.5} & 5,482 \\
                    \addlinespace[1em]

                     \multirow[c]{2}{*}{\textsc{poo}} & \underline{poo} & \textbf{64.2} & \textbf{63.7} & \textbf{70.5} & \textbf{63.1} & 33.9 & 16.7 & 9,288 \\
                     & poop & 35.8 & 36.3 & 29.5 & 36.9 & \textbf{66.1} & \textbf{83.3} & 6,818 \\
                    \addlinespace[1em]

                     \multirow[c]{2}{*}{\textsc{biscuit}} & \underline{biscuits} & \textbf{58.1} & \textbf{60.8} & \textbf{67.1} & \textbf{65.6} & 7.9 & 31.6 & 13,597 \\
                     & cookies & 41.9 & 39.2 & 32.9 & 34.4 & \textbf{92.1} & \textbf{68.4} & 12,546 \\
                    \addlinespace[1em]

                     \multirow[c]{2}{*}{\textsc{maths}} & \underline{maths} & \textbf{57.0} & \textbf{64.1} & \textbf{88.3} & \textbf{83.0} & 2.4 & 3.2 & 47,415 \\
                     & math & 43.0 & 35.9 & 11.7 & 17.0 & \textbf{97.6} & \textbf{96.8} & 66,015 \\
                    \addlinespace[1em]

                     \multirow[c]{2}{*}{\textsc{knicker}} & \underline{knickers} & \textbf{52.5} & \textbf{50.7} & \textbf{80.8} & \textbf{86.0} & 16.5 & 33.3 & 5,181 \\
                     & panties & 47.5 & 49.3 & 19.2 & 14.0 & \textbf{83.5} & \textbf{66.7} & 3,869 \\
                    \addlinespace[1em]

                     \multirow[c]{2}{*}{\textsc{oblige}} & \underline{obliged} & \textbf{52.5} & \textbf{54.7} & \textbf{73.2} & \textbf{73.4} & 23.1 & 26.7 & 16,691 \\
                     & obligated & 47.5 & 45.3 & 26.8 & 26.6 & \textbf{76.9} & \textbf{73.3} & 16,776 \\
                    \addlinespace[1em]
                    
                    \bottomrule
                \end{tabularx}

            \tablenoteparagraph{\textbf{Table Note}: This table details the percentage distribution and total token frequency ($n$) for user-informed lexical variables and their variants across six inner-circle country-level communities: \texttt{r/newzealand} (\textsc{nz}), \texttt{r/australia} (\textsc{au}), \texttt{r/unitedkingdom} (\textsc{uk}), \texttt{r/ireland} (\textsc{ie}), \texttt{r/canada} (\textsc{ca}), and \texttt{r/usa} (\textsc{us}). Percentages for each community are rounded to one decimal place and sum to 100\%, with conservative variants underlined and orthographic variants for compounds combined and marked with an asterisk (*). The most frequent variant within each category-community group is identified, while a hyphen (-) represents a variant with zero frequency. All data is sourced from Reddit/Pushshift \citep{baumgartner_pushshift_2020}.}
        
    \end{table}

    \begin{table}[p]
        \scriptsize
        \centering
        
            \caption{Distribution of User-informed Lexis (Community-Specific Outliers)}
            \label{tab:outliers}
            
              \renewcommand{\arraystretch}{1.4}
                \begin{tabularx}{\textwidth}{p{1.5cm}l
                  *{6}{>{\centering\arraybackslash}X} |
                  >{\centering\arraybackslash}p{1.5cm} }
                    \toprule
                    Variable & Variant 
                    & \textsc{nz} & \textsc{au} & \textsc{uk} 
                    & \textsc{ie} & \textsc{ca} & \textsc{us}
                    & $n$ \\
                    \midrule
                    \addlinespace[1em]
                    
                    \multicolumn{9}{c}{\texttt{r/newzealand}} \\
                    \addlinespace[1em]
                    
                     \multirow[c]{2}{*}{\textsc{lawn}} & \underline{lawn} & \textbf{56.6} & 37.0 & 40.4 & 37.2 & 45.8 & 49.2 & 22,601 \\
                     & yard & 43.4 & \textbf{63.0} & \textbf{59.6} & \textbf{62.8} & \textbf{54.2} & \textbf{50.8} & 28,607 \\
                    \addlinespace[2em]
                    
                    \multicolumn{9}{c}{\texttt{r/australia}} \\
                    \addlinespace[1em]
                    
                     \multirow[c]{3}{*}{\textsc{fizzy}} & \underline{fizzy drink} & 10.0 & 2.5 & 10.7 & 6.0 & 0.4 & - & 608 \\
                     & \underline{soft drink} & 20.7 & \textbf{49.0} & 27.8 & 12.6 & 9.5 & 3.8 & 3,198 \\
                     & soda & \textbf{69.3} & 48.5 & \textbf{61.4} & \textbf{81.4} & \textbf{90.1} & \textbf{96.2} & 9,775 \\
                    \addlinespace[2em]
                    
                    \multicolumn{9}{c}{\texttt{r/unitedkingdom}} \\
                    \addlinespace[1em]
                    
                     \multirow[c]{2}{*}{\textsc{lorry}} & \underline{lorry} & 0.5 & 0.4 & \textbf{51.3} & 22.3 & 0.1 & 0.8 & 5,702 \\
                     & truck & \textbf{99.5} & \textbf{99.6} & 48.7 & \textbf{77.7} & \textbf{99.9} & \textbf{99.2} & 84,179 \\
                    \addlinespace[1em]

                    \multirow[c]{2}{*}{\textsc{flat}} & \underline{flat} & 13.7 & 4.8 & \textbf{54.6} & 3.7 & 1.5 & 2.0 & 7436 \\
                     & apartment & \textbf{86.3} & \textbf{95.2} & 45.4 & \textbf{96.3} & \textbf{98.5} & \textbf{98.0} & 89,251 \\
                    \addlinespace[1em]
                    
                    \multirow[c]{2}{*}{\textsc{meal}} & \underline{tea time}* & 23.2 & 15.3 & \textbf{62.6} & 33.4 & 15.5 & 33.3 & 562 \\
                     & dinner time* & \textbf{76.8} & \textbf{84.7} & 37.4 & \textbf{66.6} & \textbf{84.5} & \textbf{66.7} & 1247 \\
                    \addlinespace[2em]
                    
                    \multicolumn{9}{c}{\texttt{r/canada}} \\
                    \addlinespace[1em]
                    
                     \multirow[c]{2}{*}{\textsc{ball}} & \underline{football} & \textbf{70.6} & \textbf{65.0} & \textbf{97.5} & \textbf{74.8} & 45.0 & \textbf{71.7} & 73,260 \\
                     & \underline{soccer} & 29.4 & 35.0 & 2.5 & 25.2 & \textbf{55.0} & 28.3 & 27,786 \\
                    \addlinespace[2em]
                    
                    \multicolumn{9}{c}{\texttt{r/usa}} \\
                    \addlinespace[1em]

                    \multirow[c]{2}{*}{\textsc{salvo}} & \underline{salvation army} & \textbf{96.8} & \textbf{72.2} & \textbf{73.6} & \textbf{69.9} & \textbf{84.2} & - & 3,232 \\
                    & salvo & 3.2 & 27.8 & 26.4 & 30.1 & 15.8 & \textbf{100.0} & 721 \\
                    \addlinespace[1em]
                    
                     \multirow[c]{2}{*}{\textsc{vomit}} & \underline{vomit} & \textbf{61.1} & \textbf{88.8} & \textbf{86.8} & \textbf{74.4} & \textbf{66.2} & 40.0 & 6,289 \\
                     & puke & 38.9 & 11.2 & 13.2 & 25.6 & 33.8 & \textbf{60.0} & 2,048 \\
                    \addlinespace[1em]
                    
                    \bottomrule
                \end{tabularx}

            \tablenoteparagraph{\textbf{Table Note}: This table summarises the percentage distribution and total token frequency ($n$) of user-informed lexical variables and their variants across six inner-circle country-level communities: \texttt{r/newzealand} (\textsc{nz}), \texttt{r/australia} (\textsc{au}), \texttt{r/unitedkingdom} (\textsc{uk}), \texttt{r/ireland} (\textsc{ie}), \texttt{r/canada} (\textsc{ca}), and \texttt{r/usa} (\textsc{us}). Percentages for each community are rounded to one decimal place and sum to 100\%, with conservative variants underlined and orthographic variants for compounds combined and marked with an asterisk (*). The most frequent variant within each category-community group is highlighted, while a hyphen (-) signifies zero instances of a variant. All data is sourced from Reddit/Pushshift \citep{baumgartner_pushshift_2020}.}
        
    \end{table}

\subsubsection{Clustering Analysis}

    My initial groupings based on the distributional patterns of the user-informed variables suggest that the judgements of users were largely incorrect; the innovative variants were only dominant in ten of the 50 variables I analysed. However, the distribution of these variables was systematic and meaningful. For example, the user-informed variables 23–28 are associated with Australian English and \acrshort{NZE}, while user-informed variables 35–42 are associated with North American English.
    
\begin{figure}[p]
      \centering
      
            \includegraphics[height=0.50\textwidth]{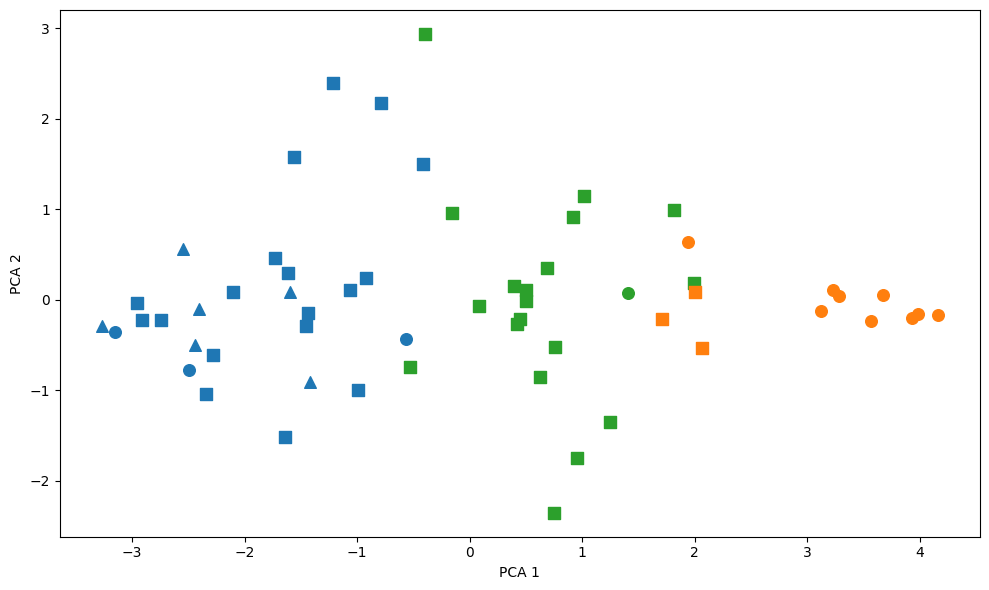}

      \vspace{6pt}
      \caption{Clustering User-informed Variables across Country-level Communities}
      \label{fig:country_pca}
      
      \vspace{6pt}
      \captionsetup{font=footnotesize, labelformat=empty, justification=justified, singlelinecheck=false}
      \caption*{\setstretch{2}\textbf{Description}: This scatterplot visualises user-informed lexical variables for conservative variants across various country-level communities, using $k$-means clustering ($k=2$) and \acrshort{PCA} for dimensionality reduction based on proportions. Markers identify distributional groupings: triangles represent conservative dominant variants, circles indicate innovative dominant variants, and squares denote community-specific groupings or outliers. The analysis reveals three clusters corresponding to these categories: conservative dominant variants (blue), non-dominant conservative variants (orange), and community-specific groupings or outliers (red). All data is sourced from Reddit/Pushshift \citep{baumgartner_pushshift_2020}.}
      
    \end{figure}

    The next phase of my analysis is to statistically evaluate these groupings. I conducted $k$-means clustering on the proportional distribution of the user-informed conservative variants across the country-level communities. I visualised the results from my clustering analysis in Figure \ref{fig:country_pca}. The clusters confirmed that the three groupings - conservative dominant, innovative dominant, and the merged community-specific groupings and outliers - were statistically significant.
    
\begin{figure}[p]
      \centering
      
            \includegraphics[height=0.60\textwidth]{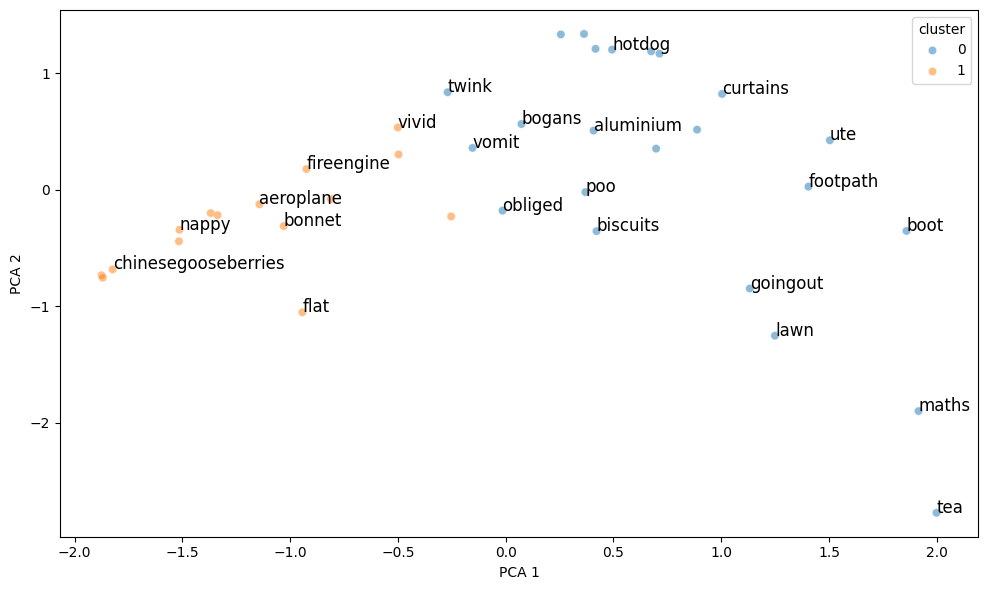}

      \vspace{6pt}
      \caption{Clustering User-informed Variables for \texttt{r/newzealand}}
      \label{fig:nzl_pca}
      
      \vspace{6pt}
      \captionsetup{font=footnotesize, labelformat=empty, justification=justified, singlelinecheck=false}
      \caption*{\setstretch{2}\textbf{Description}: This scatterplot visualises user-informed lexical variables for conservative variants within \texttt{r/newzealand}, utilizing $k$-means clustering ($k=2$) and \acrshort{PCA} for dimensionality reduction on token frequency ($n$) and proportions. The analysis reveals two distinct groupings: Cluster 0 (blue), consisting of low token frequency and non-dominant conservative variants, and Cluster 1 (orange), comprising high token frequency and dominant conservative variants. All data is sourced from Reddit/Pushshift \citep{baumgartner_pushshift_2020}.}
      
    \end{figure}

    Next, I applied the same clustering methodology to the distribution of user-informed conservative variants in \texttt{r/newzealand} using the token frequency ($n$) and proportions. The results of the clustering analysis ($k=2$) identified two clusters: one low-token-frequency and non-dominant (low proportion) cluster, and one high-token-frequency and dominant (high proportion) cluster. I visualised the results from the \texttt{r/newzealand}-only clustering analysis in Figure \ref{fig:nzl_pca}.

\subsubsection{Case Study: Nek Minute}

\begin{figure}[p]
      \centering
      
            \includegraphics[width=10cm,clip=false]{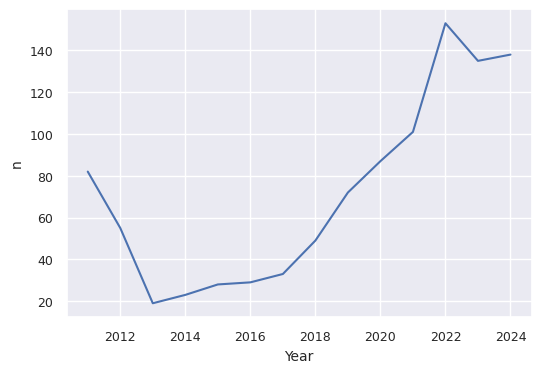}

      \vspace{6pt}
      \caption{Emergence and Growth of \textit{nek minnit}}
      \label{fig:lineplot_nek-minnit}
      
      \vspace{6pt}
      \captionsetup{font=footnotesize, labelformat=empty, justification=justified, singlelinecheck=false}
      \caption*{\setstretch{2}\textbf{Description}: This lineplot visualises the emergence and growth of the lexical item \textit{nek minnit} and its variants over time within \texttt{r/newzealand}. The feature first emerged in 2011, followed by a decline in usage in 2013, after which it has shown a consistent increase in frequency through to 2024; all data is sourced from Reddit/Pushshift \citep{baumgartner_pushshift_2020}.}
      
    \end{figure}

    One user-informed variable I excluded from my broader analysis was \textit{nek minute}. This multi-word phrase \textit{nek minute} (and its variants \textit{nek minnit} and \textit{nekminnit}) originated in a YouTube video \citep{axstabludsta_nek_2011} and can be traced to Auckland-based skateboarder Levi Hawken \citep{price_where_2016}. This variable was noted by users in the comment threads of both Selfpost 1 (\texttt{155d1f3}) and Selfpost 2 (\texttt{155d1f3}). It represents a rare instance of a dialect feature that is both distinct to \acrshort{NZE} and originated from the internet. With reference to Table \ref{tab:nek_minnit}, it is clear that this feature is specific to \texttt{r/newzealand} based on the distribution of \textit{nek minute} across the country-level communities in \gls{rcomm}. I only observed the presence of this variable in \texttt{r/newzealand} and \texttt{r/australia} (with one outlier in \texttt{r/ireland}). The innovative variant was the dominant form in \texttt{r/newzealand}. By visualising the token frequency of \textit{nek minnit} and its variants, I found that this feature was undergoing a process of lexicalisation, as shown in Figure \ref{fig:lineplot_nek-minnit}.

    \begin{table}[p]
        \scriptsize
        \centering
        
            \caption{Distribution of \textit{nek minnit} Country-Level Communities}
            \label{tab:nek_minnit}
            
            \renewcommand{\arraystretch}{1.4}
                \begin{tabularx}{\textwidth}{cl*{6}{>{\centering\arraybackslash}X} |
                  >{\centering\arraybackslash}p{2cm} }
                    \toprule
                    Variable & Variant 
                    & \textsc{nz} & \textsc{au} & \textsc{uk} 
                    & \textsc{ie} & \textsc{ca} & \textsc{us}
                    & $n$ \\
                    \midrule
                    \addlinespace[1em]
                    \multirow[c]{3}{*}{\textsc{nekminnit}} & next minute & 23.7 & \textbf{53.5} & \textbf{100.0} & \textbf{98.8} & \textbf{100.0} & - & 816 \\
                    & nek minnit & \textbf{58.2} & 35.0 & - & - & - & - & 794 \\
                    & nek minute & 18.2 & 11.5 & - & 1.2 & - & - & 253 \\
                    \addlinespace[1em]
                    \bottomrule
                \end{tabularx}\vspace{0.5cm}
                
            \tablenoteparagraph{\textbf{Table Note}: This table details the percentage distribution and total token frequency ($n$) for user-informed variables and their variants across six inner-circle country-level communities: \texttt{r/newzealand} (\textsc{nz}), \texttt{r/australia} (\textsc{au}), \texttt{r/unitedkingdom} (\textsc{uk}), \texttt{r/ireland} (\textsc{ie}), \texttt{r/canada} (\textsc{ca}), and \texttt{r/usa} (\textsc{us}). Percentages per community are rounded to one decimal place and sum to 100\%, with zero frequencies indicated by a hyphen (-). Conservative variants are underlined, while combined orthographic variants for compounds are marked with an asterisk (*). The most frequent variant within each category-community group is identified. All data is sourced from Reddit/Pushshift \citep{baumgartner_pushshift_2020}.}
        
    \end{table}

\subsection{Morphosyntactic Variables}

    In contrast to the user-informed lexical variables, I identified only six user-informed morphosyntactic constructions (including verb conjugations, modal-\textit{of}, irregular plurals, conditionals, intensifier-\textit{as}, and lexico-grammatical constructions). Of these six constructions, only three included user-informed variants, which I have presented in Table \ref{tab:morphosyntactic_alternations}. Of the three user-informed morphosyntactic variables, I observed variability across the country-level communities in only one category where the user-informed conservative variant was dominant. Surprisingly, this variability was found in \textit{swimming team} (50), where the innovative variant (\textit{swim team}) was the dominant form for \texttt{r/newzealand}, which was grouped with \texttt{r/ireland} and \texttt{r/canada}. Overall, the distributional patterns of the user-informed morphosyntactic variables did not appear to be meaningful. To supplement my analysis, I have provided a case study on irregular past participle forms and the modal-\textit{of} construction, detailing their distribution across the country-level communities.

    \begin{table}
        \scriptsize
        \centering
            
            \caption{Distribution of User-informed Morphosyntactic Constructions}
            \label{tab:morphosyntactic_alternations}
            
            \renewcommand{\arraystretch}{1.4}
                \begin{tabularx}{\textwidth}{p{1.5cm}l
                *{6}{>{\centering\arraybackslash}X} |
                  >{\centering\arraybackslash}p{1.5cm} }
                    \toprule
                    Variable & Variant 
                    & \textsc{nz} & \textsc{au} & \textsc{uk} 
                    & \textsc{ie} & \textsc{ca} & \textsc{us}
                    & $n$ \\
                    \midrule
                    \addlinespace[1em]
                    \multirow[c]{2}{*}{\textsc{syn1}} & swimming team & 33.3 & \textbf{58.0} & \textbf{54.5} & 20.0 & 11.5 & - & 59 \\
                    & swim team & \textbf{66.7} & 42.0 & 45.5 & \textbf{80.0} & \textbf{88.5} & - & 100 \\
                    \addlinespace[1em]
                    \multirow[c]{2}{*}{\textsc{syn2}} & has gone & \textbf{99.8} & \textbf{99.7} & \textbf{98.5} & \textbf{97.3} & \textbf{98.4} & \textbf{100.0} & 48,762 \\
                    & has went & 0.2 & 0.3 & 1.5 & 2.7 & 1.6 & - & 617 \\
                    \addlinespace[1em]
                    \multirow[c]{2}{*}{\textsc{syn3}} & by accident & \textbf{91.0} & \textbf{95.4} & \textbf{98.4} & \textbf{92.5} & \textbf{94.7} & \textbf{83.3} & 9,278 \\
                    & on accident & 9.0 & 4.6 & 1.6 & 7.5 & 5.3 & 16.7 & 533 \\
                    \addlinespace[1em]
                    \bottomrule
                \end{tabularx}

            \tablenoteparagraph{\textbf{Table Note}: This table summarises the percentage distribution and total token frequency ($n$) of user-informed lexical variables and their variants across six inner-circle country-level communities: \texttt{r/newzealand} (\textsc{nz}), \texttt{r/australia} (\textsc{au}), \texttt{r/unitedkingdom} (\textsc{uk}), \texttt{r/ireland} (\textsc{ie}), \texttt{r/canada} (\textsc{ca}), and \texttt{r/usa} (\textsc{us}). Percentages for each community are rounded to one decimal place and sum to 100\%, with conservative variants underlined and orthographic variants for compounds combined and marked with an asterisk (*). The most frequent variant within each category-community group is indicated, while a hyphen (-) signifies a variant with zero frequency. All data is sourced from Reddit/Pushshift \citep{baumgartner_pushshift_2020}.}
        
    \end{table}

\subsubsection{Case Study: Irregular Verbs}

    One morphosyntactic construction where variation across the country-level communities was expected is in the irregular verbs. Some verbs, such as \textit{burn}, \textit{dream}, and \textit{learn}, alternate between the regular \textit{-ed} form and the irregular \textit{-t} form in the past and past participle. Based on the corpus studies in \citet{hundt_new_1998}, the distribution of these alternating forms was systematic and regular across written American, British, and \acrshort{NZE}es. The distributional patterns of these three verbs are presented in Table \ref{tab:irregular}. Using the same procedure as for the user-informed lexical alternations, I tagged each of the three verbs (\textit{burn}, \textit{dream}, and \textit{learn}) with their parts-of-speech tags and extracted all instances of the past tense (\textsc{vbd}), past participle (\textsc{vbn}), and adjectival (\textsc{jj}) forms where the alternating \textit{-ed}/\textit{-t} variants were also observed. I present the distributional pattern for the country-level communities in Table \ref{tab:irregular_verbs}. In brief, all country-level communities preferred the regular \textit{-ed} form over the irregular \textit{-t} form for the verb \textit{burn}. I observed an inverse distribution for the adjectival form; here, the \textit{-t} variant was dominant for \textit{burnt} and \textit{dreamt} (with the exception of \texttt{r/unitedkingdom}). Where I observed systematic variation was in the adjectival form of \textit{learn}, where the \textit{-t} form was the dominant variant in \texttt{r/newzealand}, \texttt{r/australia}, and \texttt{r/unitedkingdom}, while the \textit{-ed} form was the dominant variant in the remaining country-level communities.

        \begin{table}
        \scriptsize
        \centering
            
            \caption{Distribution of Irregular Past-tense Forms in English}
            \label{tab:irregular_compare}
            
            \renewcommand{\arraystretch}{1.4}
             \begin{tabularx}{\linewidth}{l*{9}{>{\centering\arraybackslash}X}}
            \toprule
            \multirow{2}{*}{Verb} & \multicolumn{3}{c}{New Zealand} & \multicolumn{3}{c}{United Kingdom} & \multicolumn{3}{c}{United States} \\
            & Hundt & Reddit & $\%\Delta$ & Hundt & Reddit & $\%\Delta$ & Hundt & Reddit & $\%\Delta$ \\
            \midrule
                \addlinespace[1em]
                \textit{burned} 
                & 55.0 & 68.7 & \multirow{2}{*}{13.7} 
                & 56.0 & 76.4 & \multirow{2}{*}{20.4} 
                & 95.0 & 88.9 & \multirow{2}{*}{6.1} \\
                \textit{burnt} 
                & 45.0 & 31.3 &  
                & 44.0 & 23.6 &  
                & 5.0  & 11.1 & \\
                \addlinespace[1em]
                \textit{dreamed} 
                & 80.0 & 91.7 & \multirow{2}{*}{11.7} 
                & 69.0 & 81.5 & \multirow{2}{*}{12.5} 
                & 95.0 & 100.0 & \multirow{2}{*}{5.0} \\
                \textit{dreamt} 
                & 20.0 & 8.3  &  
                & 31.0 & 18.5 &  
                & 5.0  & -    &  \\
                \addlinespace[1em]
                \textit{learned} 
                & 75.0 & 96.1 & \multirow{2}{*}{21.1} 
                & 78.0 & 96.6 & \multirow{2}{*}{18.6} 
                & 100.0 & 100.0 & \multirow{2}{*}{-} \\
                \textit{learnt} 
                & 25.0 & 3.9 & 
                & 22.0 & 3.4 & 
                & -    & -   &  \\
                \addlinespace[1em]
            \bottomrule
            \end{tabularx}

            \tablenoteparagraph{\textbf{Table Note}: This table compares verb usage across three newspaper sources representing distinct varieties of English: the Dominion Post/Evening Post (\texttt{DOM/EVP}) for \acrshort{NZE}, the Miami Herald (\texttt{Miami Herald}) for American English, and the Guardian (\texttt{Guardian}) for British English. Columns detail the root verb (\texttt{Verb}) alongside the corresponding frequency or usage data for each publication; all information is sourced from Reddit/Pushshift \citep{baumgartner_pushshift_2020} and \citet{hundt_new_1998}.}
        
    \end{table}
    
\subsubsection{Case Study: Model-\textit{of}}

    One user-informed morphosyntactic feature was the emergence of modal-\textit{of}. Combined with the three modal verbs (\textit{should}, \textit{would}, and \textit{could}), this construction is also known as the modal of lost opportunities. In contrast to the irregular forms of \textit{burn}, \textit{dream}, and \textit{learn} discussed in the previous section, I have limited literature on the distribution of modal-\textit{of} in \acrshort{NZE} (\citealp{bauer_can_2002}; \citealp{hay_new_2008}). I presented the distribution of modal-\textit{of} and the variant full (\textit{have}) and contracted (\textit{'ve}) forms in Table \ref{tab:modal_plus}. Of the three variants, the full form (\textit{have}) was the dominant variant across the six country-level communities. If I focused on the modal-\textit{of} variants in isolation, modal-\textit{of} did not appear to be a variant exclusive to \texttt{r/newzealand}.

    \begin{table}
        \scriptsize
        \centering
            
            \caption{Distribution of Irregular Verbs}
            \label{tab:irregular_verbs}
            
            \renewcommand{\arraystretch}{1.4}
                \begin{tabularx}{\textwidth}{*{8}{>{\centering\arraybackslash}X} |
                  >{\centering\arraybackslash}p{2cm} }
                    \toprule
                    POS & Form 
                    & \textsc{nz} & \textsc{au} & \textsc{uk} 
                    & \textsc{ie} & \textsc{ca} & \textsc{us}
                    & $n$ \\
                    \midrule
                    \addlinespace[1em]
                    
                    \multicolumn{9}{c}{\textit{burn}} \\
                    \addlinespace[1em]
                    
                    \textsc{vbd} & \textit{-ed} & \textbf{68.7} & \textbf{67.2} & \textbf{76.4} & \textbf{79.0} & \textbf{87.7} & \textbf{88.9} & 20,058 \\
                    & \textit{-t} & 31.3 & 32.8 & 23.6 & 21.0 & 12.3 & 11.1 & 5,670 \\
                    \addlinespace[1em]
                    
                    \textsc{vbn} & \textit{-ed} & \textbf{83.3} & \textbf{82.7} & \textbf{90.4} & \textbf{91.3} & \textbf{95.1} & \textbf{95.2} & 10,775 \\
                    & \textit{-t} & 16.7 & 17.3 & 9.6 & 8.7 & 4.9 & 4.8 & 1,163 \\
                    \addlinespace[1em]
                    
                    \textsc{jj} & \textit{-ed} & 3.5 & 4.4 & 4.2 & 5.0 & 8.5 & - & 384 \\
                    & \textit{-t} & \textbf{96.5} & \textbf{95.6} & \textbf{95.8} & \textbf{95.0} & \textbf{91.5} & \textbf{100.0} & 6,806 \\
                    \addlinespace[2em]
                    
                    \multicolumn{9}{c}{\textit{dream}} \\
                    \addlinespace[1em]
                    
                    \textsc{vbd} & \textit{-ed} & \textbf{91.7} & \textbf{86.3} & \textbf{81.5} & \textbf{83.3} & \textbf{92.3} & \textbf{100.0} & 1,942 \\
                    & \textit{-t} & 8.3 & 13.7 & 18.5 & 16.7 & 7.7 & - & 276 \\
                    \addlinespace[1em]
                    
                    \textsc{vbn} & \textit{-ed} & \textbf{98.0} & \textbf{95.5} & \textbf{94.5} & \textbf{95.3} & \textbf{97.3} & \textbf{100.0} & 1,592 \\
                    & \textit{-t} & 2.0 & 4.5 & 5.5 & 4.7 & 2.7 & - & 60 \\
                    \addlinespace[1em]
                    
                    \textsc{jj} & \textit{-ed} & 15.4 & - & \textbf{55.6} & 25.0 & 36.4 & - & 13 \\
                    & \textit{-t} & \textbf{84.6} & \textbf{100.0} & 44.4 & \textbf{75.0} & \textbf{63.6} & - & 39 \\
                    \addlinespace[2em]
                    
                    \multicolumn{9}{c}{\textit{learn}} \\
                    \addlinespace[1em]
                    
                    \textsc{vbd} & \textit{-ed} & \textbf{96.1} & \textbf{96.2} & \textbf{96.6} & \textbf{98.8} & \textbf{99.8} & \textbf{100.0} & 56,312 \\
                    & \textit{-t} & 3.9 & 3.8 & 3.4 & 1.2 & 0.2 & - & 998 \\
                    \addlinespace[1em]
                
                    \textsc{vbn} & \textit{-ed} & \textbf{74.8} & \textbf{73.2} & \textbf{75.4} & \textbf{92.1} & \textbf{97.9} & \textbf{97.0} & 44,985 \\
                    & \textit{-t} & 25.2 & 26.8 & 24.6 & 7.9 & 2.1 & 3.0 & 7457 \\
                    \addlinespace[1em]
                    
                    \textsc{jj} & \textit{-ed} & 32.9 & 39.6 & 40.1 & \textbf{57.7} & \textbf{82.6} & \textbf{100.0} & 23,39 \\
                    & \textit{-t} & \textbf{67.1} & \textbf{60.4} & \textbf{59.9} & 42.3 & 17.4 & - & 2,369 \\
                    \addlinespace[1em]
                    \bottomrule
                \end{tabularx}

            \tablenoteparagraph{\textbf{Table Note}: This table displays the percentage distribution and total token frequency ($n$) for specific user-informed lexical variables and their variants across six inner-circle country-level communities: \texttt{r/newzealand} (\textsc{nz}), \texttt{r/australia} (\textsc{au}), \texttt{r/unitedkingdom} (\textsc{uk}), \texttt{r/ireland} (\textsc{ie}), \texttt{r/canada} (\textsc{ca}), and \texttt{r/usa} (\textsc{us}). Percentages per community are rounded to one decimal place and sum to 100\%, with a hyphen (-) denoting zero instances of a variant. Conservative variants are underlined, and combined orthographic variants for compounds are marked with an asterisk (*). The most frequent variant within each category-community group is highlighted. All data is sourced from Reddit/Pushshift \citep{baumgartner_pushshift_2020}.}
        
    \end{table}

    \begin{table}
        \scriptsize
        \centering
            
            \caption{Distribution of Modal+ Constructions}
            \label{tab:modal_plus}
            
            \renewcommand{\arraystretch}{1.4}
                \begin{tabularx}{\textwidth}{*{8}{>{\centering\arraybackslash}X} |
                  >{\centering\arraybackslash}p{2cm} }
                    \toprule
                    POS & Form 
                    & \textsc{nz} & \textsc{au} & \textsc{uk} 
                    & \textsc{ie} & \textsc{ca} & \textsc{us}
                    & $n$ \\
                    \midrule
                    \addlinespace[1em]
                    \multicolumn{9}{c}{\textbf{should}} \\
                    \addlinespace[1em]
                    \textsc{vb} & have & 90.1 & 91.0 & 92.6 & 91.3 & 93.5 & 91.7 & 383,977 \\
                    \textsc{in} & of & 2.1 & 2.4 & 1.8 & 2.5 & 1.4 & 2.8 & 8,079 \\
                    \textsc{vbd} & ve & 7.8 & 6.7 & 5.6 & 6.2 & 5.1 & 5.6 & 25,140 \\
                    \addlinespace[2em]
                    \multicolumn{9}{c}{\textbf{would}} \\
                    \addlinespace[1em]
                    \textsc{vb} & have & 90.4 & 90.6 & 92.5 & 91.7 & 93.0 & 89.3 & 819,127 \\
                    \textsc{in} & of & 1.6 & 2.1 & 1.7 & 1.8 & 1.4 & 2.1 & 15,080 \\
                    \textsc{vbd} & ve & 8.0 & 7.3 & 5.8 & 6.5 & 5.7 & 8.7 & 57,916 \\
                    \addlinespace[2em]
                    \multicolumn{9}{c}{\textbf{could}} \\
                    \addlinespace[1em]
                    \textsc{vb} & have & 88.6 & 89.5 & 90.5 & 88.7 & 91.2 & 87.1 & 292,340 \\
                    \textsc{in} & of & 2.3 & 2.8 & 2.5 & 2.8 & 1.8 & 1.7 & 7,575 \\
                    \textsc{vbd} & ve & 9.1 & 7.7 & 7.0 & 8.4 & 7.1 & 11.2 & 24,905 \\
                    \addlinespace[1em]
                    \bottomrule
                \end{tabularx}

            \tablenoteparagraph{\textbf{Table Note}: This table presents the percentage distribution and total token frequency ($n$) for user-informed variables and their variants across six inner-circle country-level communities: \texttt{r/newzealand} (\textsc{nz}), \texttt{r/australia} (\textsc{au}), \texttt{r/unitedkingdom} (\textsc{uk}), \texttt{r/ireland} (\textsc{ie}), \texttt{r/canada} (\textsc{ca}), and \texttt{r/usa} (\textsc{us}). Columns show the variable, the specific variant, and the grouped percentage per community rounded to one decimal place, while rows group variants by variable; conservative forms are underlined, combined orthographic variants for compounds are marked with an asterisk (*), and the most frequent variant within each category-community group is highlighted. Percentages for each community sum to 100\%, with a hyphen (-) representing zero instances. All data is sourced from Reddit/Pushshift \citep{baumgartner_pushshift_2020}.}
        
    \end{table}

\section{Discussion}
\label{variable:discussion}

    The results from the user-informed lexical variables suggest that the intuitions of users from \texttt{r/newzealand} overestimated the number of lexical variables that have undergone complete change. Users have identified some lexical changes in progress in \texttt{r/newzealand} worth further investigation. For example, the inclusion of \textit{nappy} and \textit{tramp} in the innovative dominant variables (see Table \ref{tab:innovative_only}) was unexpected, since they were both associated with \acrshort{NZE} and are shared features with British and Australian Englishes \citep{orsman_new_1994}. It was unclear whether these distributional patterns were the result of a change in progress or if they resulted from some other phenomenon, such as non-local bias \citep{dunn_measuring_2020}.

    This variable approach did offer some new insights otherwise missed from the thematic analysis. Not included in my analysis was the marked second-person plural pronoun - \textit{yous} and its variant \textit{youse}. This was a user-informed variable that received significant attention because it was preferred to the alternative \textit{y'all}. I presented the distribution of the variants of the marked plural pronouns (\textit{y'all}, \textit{yinz}, \textit{you all}, \textit{you guys}, \textit{yous}, and \textit{youse}) in Table \ref{tab:pronouns}. The results suggest that the preferences of users in \texttt{r/newzealand} were not linked to production, even though \textit{yous} and \textit{youse} were in the minority compared to \textit{y'all}.
    
\begin{figure}
      \centering
      
      \begin{subfigure}[b]{0.8\textwidth}
          \centering
          \includegraphics[width=\textwidth]{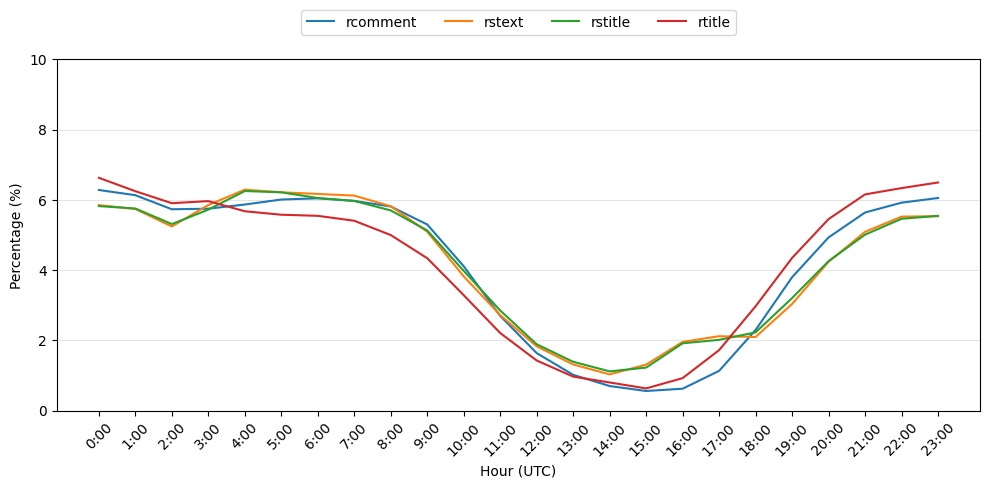}
          \subcaption{\texttt{r/newzealand}}
      \end{subfigure}
      \vspace{12pt}
      
      \begin{subfigure}[b]{0.8\textwidth}
          \centering
          \includegraphics[width=\textwidth]{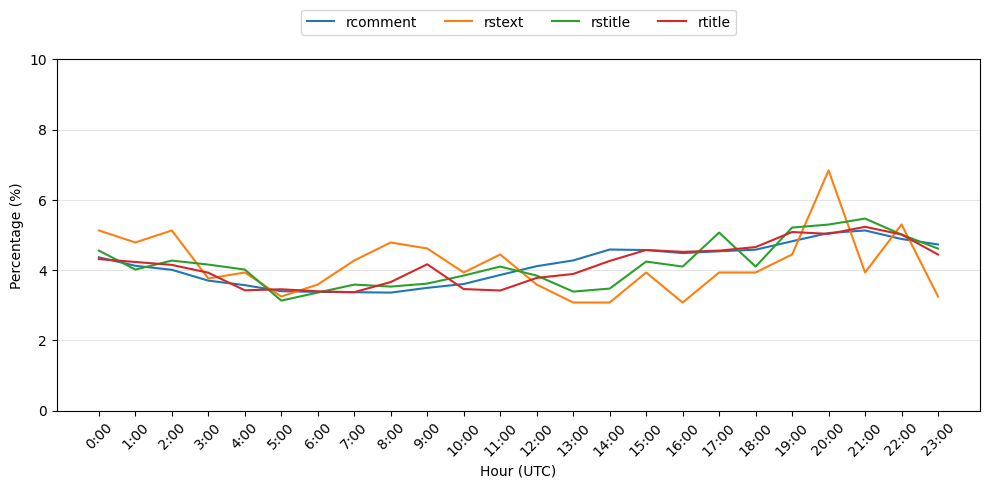}
          \subcaption{\texttt{r/MapsWithoutNZ}}
      \end{subfigure}

      \vspace{6pt}
      \caption{User Engagement by Hour between Communities}
      \label{fig:time_nz}
      
      \vspace{6pt}
      \captionsetup{font=footnotesize, labelformat=empty, justification=justified, singlelinecheck=false}
      \caption*{\setstretch{2}\textbf{Description}: This figure compares user activity patterns between \texttt{r/newzealand} (a) and \texttt{r/MapsWithoutNZ} (b) over a 24-hour cycle in \acrshort{UTC}, where 12:00 \acrshort{UTC} corresponds to midnight \acrfull{NZST} without daylight savings. In \texttt{r/newzealand}, activity declines from 09:00 \acrshort{UTC} (21:00 \acrshort{NZST}) to a minimum at approximately 15:00 \acrshort{UTC} (03:00 \acrshort{NZST}), whereas traffic in \texttt{r/MapsWithoutNZ} remains stable at roughly 4\% throughout the day with a slight increase around 20:00 \acrshort{UTC} (08:00 \acrshort{NZST}); notably, \acrshort{NZST} (\acrshort{UTC}+12) is a relatively isolated time zone shared only with seven Pacific nations and two Russian federal subjects. All data is sourced from Reddit/Pushshift \citep{baumgartner_pushshift_2020}.}
      
    \end{figure}

    While place-based communities, such as \texttt{r/newzealand}, are dedicated to a particular country, region, or city \citep{panek_understanding_2022}, the communities do not exclusively serve the interests of those living in those places. One aspect that I have yet to discuss is the role of time - or when a user engages with New Zealand-related communities. With reference to time, I visualised the level of engagement from users as a proportion of the day for \texttt{r/MapsWithoutNZ}, a meme-based community that is related to New Zealand, and the place-based community, \texttt{r/newzealand} (see Figure \ref{fig:time_nz}). Users in \texttt{r/newzealand} adhere to \acrshort{NZST}; meanwhile, users on \texttt{r/MapsWithoutNZ} do not.

\begin{figure}
      \centering
      
            \begin{subfigure}[b]{\textwidth}
                \centering
                \includegraphics[width=0.75\textwidth]{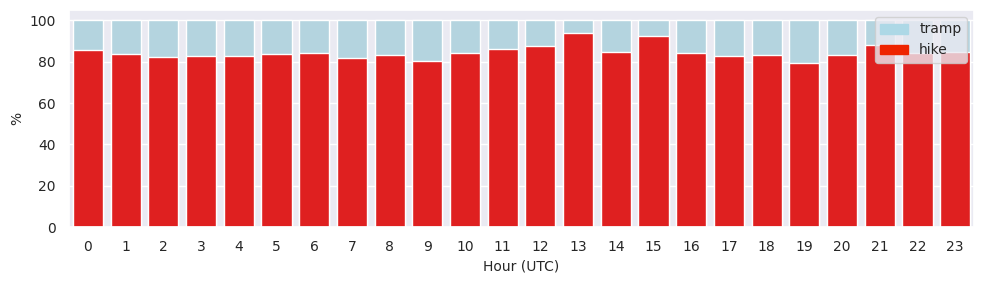}
                \subcaption{\textsc{tramp1} and \textsc{tramp2}}
            \end{subfigure}\vspace{12pt}
            \begin{subfigure}[b]{\textwidth}
                \centering
                \includegraphics[width=0.75\textwidth]{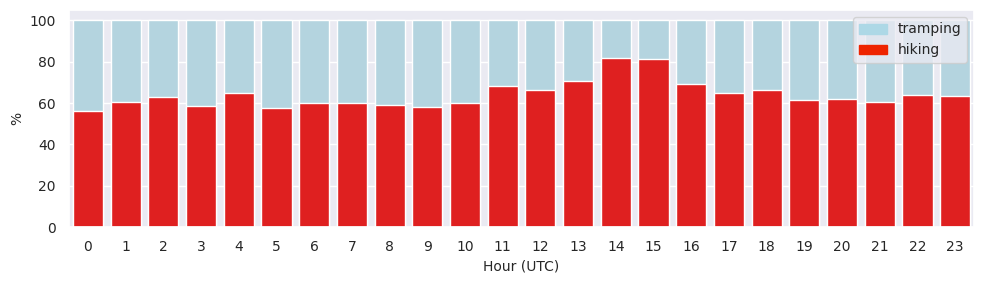}
                \subcaption{\textsc{tramp3}}
            \end{subfigure}\vspace{12pt}

      \vspace{6pt}
      \caption{Proportional Frequency by Hour (\textit{tramp} and \textit{tramping)}}
      \label{fig:time_lexical}
      
      \vspace{6pt}
      \captionsetup{font=footnotesize, labelformat=empty, justification=justified, singlelinecheck=false}
      \caption*{\setstretch{2}\textbf{Description}: This figure illustrates the proportion of user-informed variables \textit{tramp} (a) and \textit{tramping} (b) over a 24-hour period within \texttt{r/newzealand} comments (\gls{rcomm}). Observations indicate a minor increase in the proportional frequency of the innovative variants \textit{hike} and \textit{hiking} between 12:00 and 18:00 \acrshort{UTC}, a window corresponding approximately to midnight to 06:00 \acrshort{NZST} when not adjusted for daylight savings; all data is sourced from Reddit/Pushshift \citep{baumgartner_pushshift_2020}.}
      
    \end{figure}

\begin{figure}
      \centering
      
            \begin{subfigure}[b]{\textwidth}
                \centering
                \includegraphics[width=0.75\textwidth]{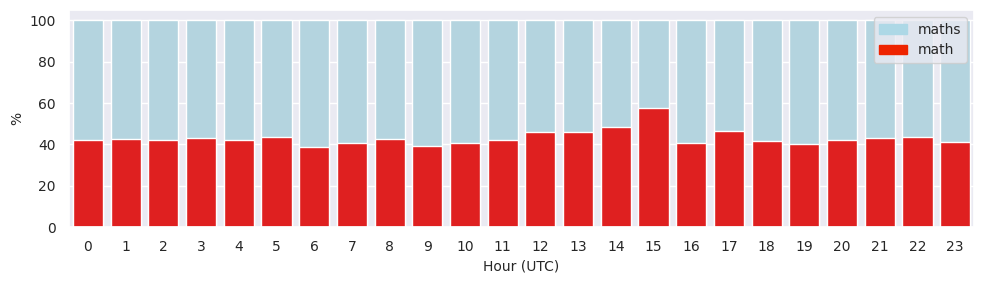}
                \subcaption{\textsc{maths}}
            \end{subfigure}\vspace{12pt}
            \begin{subfigure}[b]{\textwidth}
                \centering
                \includegraphics[width=0.75\textwidth]{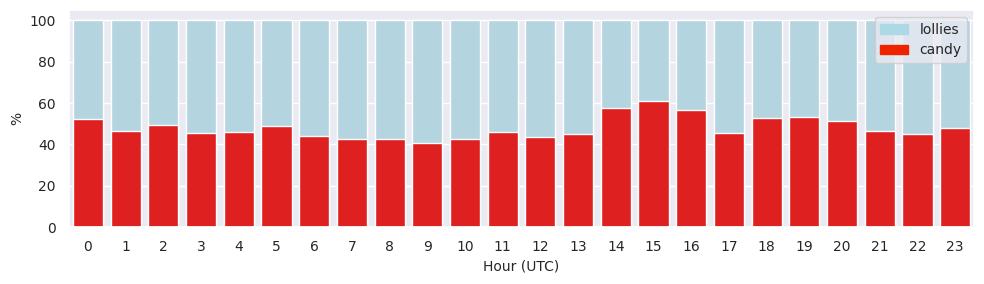}
                \subcaption{\textsc{lolly}}
            \end{subfigure}\vspace{12pt}
            \begin{subfigure}[b]{\textwidth}
                \centering
                \includegraphics[width=0.75\textwidth]{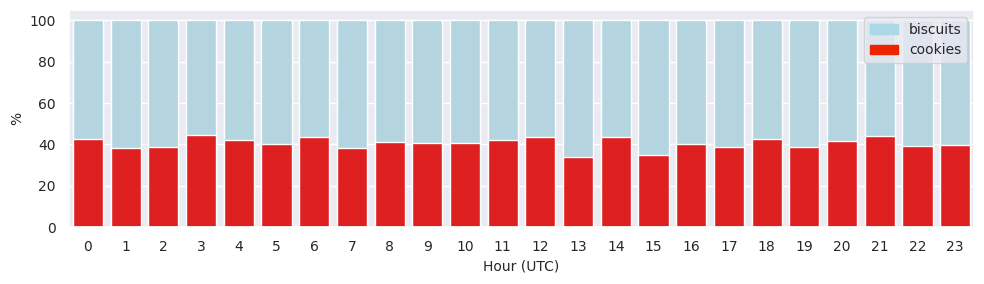}
                \subcaption{\textsc{biscuit}}
            \end{subfigure}

      \vspace{6pt}
      \caption{Proportional Frequency by Hour (Other Variables)}
      \label{fig:other_lexical}
      
      \vspace{6pt}
      \captionsetup{font=footnotesize, labelformat=empty, justification=justified, singlelinecheck=false}
      \caption*{\setstretch{2}\textbf{Description}: This figure displays the proportion of user-informed variables (a) \textsc{math}, (b) \textsc{lolly}, and (c) \textit{biscuit} over a 24-hour period in \texttt{r/newzealand} comments (\gls{rcomm}). Aside from the innovative variant \textit{cookies}, a minor increase in proportional frequency for the innovative variants \textit{math} and \textit{candy} was observed between 12:00 and 18:00 \acrshort{UTC}, which corresponds approximately to midnight through 06:00 \acrshort{NZST} without daylight savings adjustments; all data is sourced from Reddit/Pushshift \citep{baumgartner_pushshift_2020}.}
      
    \end{figure}

    In light of this new insight into the behaviour of users in \texttt{r/newzealand}, I considered the proportional distribution of the variable category \textsc{tramp} over time. Based on the distribution of this variable in \gls{rcomm} for the inner-circle country-level communities, \textit{tramping} was a word unique to \texttt{r/newzealand} (46.8\%), while the remaining five country-level communities had a proportional frequency of less than 4.5\%. I visualised the proportion of alternating lexical features over a 24-hour period in Figures \ref{fig:time_lexical} and \ref{fig:other_lexical}. In addition to (a) \textit{tramp} and (b) \textit{tramping} in Figure \ref{fig:time_lexical}, I included the variable categories of (a) \textit{maths}, (b) \textit{lollies}, and (c) \textit{biscuits} in Figure \ref{fig:other_lexical}. With the exception of \textit{biscuits}, the innovative variants all briefly increase in usage when users are the least active on \texttt{r/newzealand}. While this difference was marginal, it does suggest the presence of non-local users (or at least those who prefer the innovative variant) engaging in the place-based community at that time.

    \begin{table}
        \scriptsize
        \centering
            
            \caption{Marked Second Person Plural}
            \label{tab:pronouns}
            
            \renewcommand{\arraystretch}{1.4}
                \begin{tabularx}{\textwidth}{l*{6}{>{\centering\arraybackslash}X} |
                  >{\centering\arraybackslash}p{2cm}}
                    \toprule
                    Variant 
                    & \textsc{nz} & \textsc{au} & \textsc{uk} 
                    & \textsc{ie} & \textsc{ca} & \textsc{us}
                    & $n$ \\
                    \midrule
                    \addlinespace[1em]
                    ya'll & 14.0 & 10.4 & 5.8 & 10.4 & 13.9 & 22.6 & 21,033 \\
                    yinz & - & - & - & 0.1 & - & - & 37 \\
                    you all & 25.8 & 26.7 & 35.5 & 34.2 & 18.8 & 18.6 & 44,391 \\
                    you guys & 56.4 & 58.6 & 56.9 & 39.9 & 66.7 & 58.1 & 102,958 \\
                    yous & 2.4 & 1.3 & 1.4 & 13.1 & 0.4 & 0.8 & 4,917 \\
                    youse & 1.5 & 2.9 & 0.4 & 2.4 & 0.1 & - & 2,235 \\
                    \addlinespace[1em]
                    \bottomrule
                \end{tabularx}

            \tablenoteparagraph{\textbf{Table Note}: This table details the distributional patterns of user-informed lexical variables and their variants across six inner-circle country-level communities: \texttt{r/newzealand} (\textsc{nz}), \texttt{r/australia} (\textsc{au}), \texttt{r/unitedkingdom} (\textsc{uk}), \texttt{r/ireland} (\textsc{ie}), \texttt{r/canada} (\textsc{ca}), and \texttt{r/usa} (\textsc{us}). Columns provide the variable name, the specific variant, the percentage distribution within each community rounded to one decimal place, and the total token frequency ($n$) across all communities. Variants are grouped by variable, with conservative forms underlined and orthographic variants for compounds combined and marked by an asterisk (*); the most frequent variant within each category-community group is highlighted, and instances with zero frequency are indicated by a hyphen (-). All data is sourced from Reddit/Pushshift \citep{baumgartner_pushshift_2020}.}
        
    \end{table}

\section{Chapter Summary}
\label{variable:conclusion}

    The goals of the current chapter are: a) to evaluate the intuitions of the users; and b) to determine the distribution of these user-informed \acrshort{NZE} features across different place-based communities. Using the sociolinguistic variable approach, I observed that the distributional patterns of the user-informed lexical variables were systematic across the country-level communities. While user intuitions often overestimated the degree of lexical change, the extracted variables revealed distinct regional clusters and community-specific outliers that align with established dialect models. In the absence of a benchmark for \acrshort{NZE}, these user-informed variables provide a rudimentary measure of user expectations and linguistic production in place-based communities.

% -----------------------------
% Chapter 6: Dialect Modelling and Language Embeddings
% -----------------------------

\chapter{Dialect Modelling and Language Embeddings}
\markboth{Dialect Modelling and Language Embeddings}{}
\label{chap:dialect_classification}

\section{Chapter Outline}
\label{dialect:chapter_outline}

    This chapter examines place-based communities on Reddit by applying approaches from \acrshort{NLP}. I use text classification to understand corpus characteristics and language embeddings to determine linguistic variation in user-informed semantic variables between these communities. Section \ref{dialect:introduction} provides the background and motivation for this study. In Section \ref{dialect:data}, I describe the data for the outer-circle country-level communities and New Zealand city-level communities. The analytical pipeline and results for the text classification models are presented in Section \ref{dialect:classification_models}. Following a brief interim summary in Section \ref{dialect:interim_summary}, I introduce the analytical pipeline and results for the language embedding models in Section \ref{dialect:embedding_models}. I conclude the chapter in Section \ref{dialect:discussion} by synthesising the findings from both approaches and outlining the key results in Section \ref{dialect:conclusion}. The insights from this chapter contribute to my second secondary question: is there a relationship between geographic dialect communities and place-based communities?

\section{Background and Motivation }
\label{dialect:introduction}
    
    Based on the findings from Chapter \ref{chap:user_variables}: User-Informed Sociolinguistic Variables, the results suggest that the distribution pattern of the user-informed features identified in Chapter \ref{chap:user_intuitions}: User Intuitions and Place Identity was systematic. The results suggest that the intuitions of users on \texttt{r/newzealand} were correct to a degree. In some lexical alternations, I observed a higher proportion of innovative forms - such as \textit{soda}, \textit{diaper}, and \textit{hike}. As I later noted, the distribution of these lexical alternations was sensitive to non-linguistic factors, such as user activity across different time zones, which does not support a change in progress. However, this does offer evidence of non-local users in \texttt{r/newzealand} \citep{johnson_geography_2016}. In the case of the irregular verbs \textit{burnt} and \textit{learnt}, I also observed variable production consistent with the findings from \citep{hundt_new_1998}. The low token frequency in some country-level communities, such as \texttt{r/usa}, puts into question the significance of these results.
    
    After taking into account interactionist and variationist approaches to geographic dialect alignment in the previous two chapters, I now consider how I can scale my analysis by adopting approaches from \acrshort{NLP}. As observed in Chapter \ref{chap:user_variables}: User-Informed Sociolinguistic Variables, I was able to examine lexical or morphosyntactic variables where I could compare the distribution of variants based on their surface forms to a limited degree. I cannot solely rely on the surface forms of linguistic features to examine linguistic variation; I must also consider the relationship between words to determine their use within context. Semantic variation is a prime example where these relationships are necessary.
    
    The primary intersection between dialectology and \acrshort{NLP} is in the area of Natural Language Understanding, which involves tasks such as dialect identification and sentiment classification \citep{joshi_natural_2025}. The majority of work in this area involves some form of classification approach using linguistic features such as word or character $n$-grams \citep{taha_comprehensive_2024}. I found that text classification as a form of dialect modelling is an imperfect solution to exploring linguistic variation in a high-dimensional space \citep{dunn_stability_2022}. Georeferenced social media language data is typically characterised by a high volume of named entities, such as place-names and local abbreviations \citep{eisenstein_latent_2010}. In the case of text classification, reliance on these high-frequency features does not necessarily indicate latent linguistic variation. An alternative to word and character $n$-grams could include features such as function words or \acrshort{C2xG} \citep{dunn_global_2019}. Other confounds, such as sample size \citep{figueroa_text_2013}, have been shown to have an impact on classification performance.
    
    Recent advancements, such as the introduction of language embedding model architectures like Word2Vec \citep{mikolov_efficient_2013} and transformer-based language models like BERT \citep{devlin_bert_2019}, have revolutionised dialect modelling. One advantage of embedding models is that word vectors encode the relationships between words, which can be used to determine semantic variation and change. For example, vector representations trained on georeferenced Twitter\textsuperscript{X} data were shown to be stable across 84 language varieties \citep{dunn_variation_2023}. With past research limited to Twitter\textsuperscript{X}, Facebook, or YouTube \citep{nguyen_dialect_2021}, I extend this research to my place-based communities on Reddit. The goals of the current chapter are: a) to use text classification to distinguish the characteristics between place-based communities; and b) to evaluate the effectiveness of embedding models in determining semantic shifts in user-informed variables.
    
\section{Data}
\label{dialect:data}

    I continue to use the inner-circle country-level communities from Chapter \ref{chap:user_variables}: User-Informed Sociolinguistic Variables. The inner-circle classification is based on the Three Circles of English model \citep{kachru_standards_1985}. These communities are associated with the six inner-circle varieties of English (\texttt{r/canada}, \texttt{r/usa}, \texttt{r/ireland}, \texttt{r/unitedkingdom}, \texttt{r/australia}, and \texttt{r/newzealand}). My primary source of Reddit data comes from the Pushshift dumps, which are part of an ongoing data collection effort of the top 40,000 communities on Reddit via the Pushshift \acrshort{API} \citep{baumgartner_pushshift_2020}.
    
    I expand the selection of country-level communities to include place-based communities associated with outer-circle varieties of English. The outer circle refers to countries with a colonial history of English where the language continues to be an institutionalised official language. The six place-based communities are \texttt{r/Kenya}, \texttt{r/southafrica}, \texttt{r/india}, \texttt{r/pakistan}, \texttt{r/malaysia}, and \texttt{r/Philippines}. The largest place-based community is \texttt{r/Philippines}, with over 3.2 million subscribers, and the smallest is \texttt{r/Kenya}, with over 218,000 subscribers.

    In addition to the six inner-circle country-level communities, I also expand the analysis to New Zealand city-level communities. The six city-level communities are \texttt{r/auckland}, \texttt{r/Tauranga}, \texttt{r/thetron}, \texttt{r/Wellington}, \texttt{r/chch}, and \texttt{r/dunedin}. These six city-level communities correspond to the six largest urban areas in New Zealand \citep{stats_nz_place_2024}.

    \begin{table}
        \scriptsize
        \centering
            
            \caption{Summary of Outer-Circle Country-level Communities on Reddit}
            \label{tab:outer_circle}
            
            \renewcommand{\arraystretch}{1.4}
            \begin{tabularx}{\textwidth}{l*{5}{>{\centering\arraybackslash}X}}
            \toprule
            \textbf{Community} & \textbf{Created} & \textbf{Members} & \textbf{Population} & \textbf{Region} \\
            \midrule
                \addlinespace[1em]
                \texttt{r/Kenya} & Nov 14, 2009 & 218,884 & 56,433,000 & Eastern Africa \\
                \texttt{r/southafrica} & Oct 12, 2008 & 361,972 & 64,007,000 & Southern Africa \\
                \texttt{r/india} & Jan 25, 2008 & 3,216,182 & 1,450,936,000 & Southern Asia \\
                \texttt{r/pakistan} & Jan 26, 2008 & 575,437 & 251,269,000 & Southern Asia \\
                \texttt{r/malaysia} & Jan 26, 2008 & 1,263,289 & 35,558,000 & South-eastern Asia \\
                \texttt{r/Philippines} & Jul 4, 2008 & 3,270,391 & 115,844,000 & South-eastern Asia \\
                \addlinespace[1em]
            \bottomrule
            \end{tabularx}

            \tablenoteparagraph{\textbf{Table Note}: This table summarises the 16 outer-circle country-level communities identified for the study, including the community name (\texttt{Community}), its creation date (\textit{Created}), and subscriber count as of June 2025 (\textit{Members}). It also provides demographic and geographic context through estimated population figures recorded by the United Nations as of 2024 (\textit{Population}) and regional classifications based on United Nations geographic designations (\textit{Region}). Data is sourced from Reddit/Pushshift \citep{baumgartner_pushshift_2020} and the United Nations.}
        
    \end{table}

    \begin{table}
        \scriptsize
        \centering
            
            \caption{Summary of New Zealand City-level Communities on Reddit}
            \label{tab:city_level}
            
            \renewcommand{\arraystretch}{1.4}
            \begin{tabularx}{\textwidth}{l*{5}{>{\centering\arraybackslash}X}}
            \toprule
            \textbf{Community} & \textbf{Created} & \textbf{Members} & \textbf{Population} & \textbf{Region} \\
            \midrule
                \addlinespace[1em]
                \texttt{r/auckland} & Nov 22, 2009 & 230,631 & 1,530,500 & Auckland \\
                \texttt{r/Tauranga} & Dec 28, 2013 & 12,091 & 161,300 & Bay of Plenty \\
                \texttt{r/thetron} & Feb 17, 2011 & 43,310 & 189,700 & Waikato \\
                \texttt{r/Wellington} & Sep 2, 2010 & 125,402 & 208,800 & Wellington \\
                \texttt{r/chch} & Feb 14, 2011 & 66,506 & 400,600 & Canterbury \\
                \texttt{r/dunedin} & Sep 6, 2010 & 36,326 & 103,200 & Otago \\
                \addlinespace[1em]
            \bottomrule
            \end{tabularx}

            \tablenoteparagraph{\textbf{Table Note}: This table summarises the 16 New Zealand city-level communities identified for the study, providing details on the community name (\texttt{Community}), its creation date (\textit{Created}), and subscriber count as of October 2025 (\textit{Members}). To provide geographic and demographic context, the table also includes the urban population for each area according to the 2023 Census (\textit{Population}) and the corresponding regional council areas (\textit{Region}). Data is sourced from Reddit/Pushshift \citep{baumgartner_pushshift_2020} and Statistics New Zealand \citep{stats_nz_place_2024}.}
        
    \end{table}

\section{Classification Models}
\label{dialect:classification_models}

    Building on the established research regarding Twitter\textsuperscript{X}, I replicate the findings from \citet{dunn_stability_2022} on Reddit. More importantly, I aim to confirm that text classification as a form of dialect modelling is not fit-for-purpose for investigating variation and change. While I cannot rely on text classification to model dialects \citep{dunn_stability_2022}, I can use it as a diagnostic tool within my model development pipeline. I explore dialect modelling across two different conditions: country-level communities and city-level communities on Reddit. Of particular interest in the country-level communities is how I can distinguish \texttt{r/newzealand} from other communities associated with inner-circle varieties of English. Regarding the city-level communities, I want to determine the role of classifiers and internal variation within place-based communities associated with New Zealand. Beginning with shallow text classifiers, my aim is to apply the insights from these models to my language embedding models, where I can track the shift of features across a semantic space.

\subsection{Methodology}
\label{dialect:classifier_method}

    Text classification models follow a basic design pipeline \citep{kowsari_text_2019}: a) feature extraction; b) dimensionality reduction; c) model training; and d) evaluation. With a focus on country-level communities, I consider the role of geographic sampling bias, sample size, and the stability of my country-level classifiers over time.

\subsubsection{Feature Engineering}
\label{dialect:feature_engineering}

    One consideration is sample size per observation \citep{figueroa_text_2013}. I use the inner-circle country-level communities to illustrate the differences between text-types. The average length of \gls{rpost} was 72 words, while the average length of \gls{rstitle} was 60 words. Both text-types have a maximum length of 343 words and 325 words, respectively. The word length of these text-types will pose an issue for text classification. The average word length of each text-type in the city-level communities was considerably shorter than in the country-level communities.
    
    In order to account for named entities such as place names, I use the \texttt{spaCy} English pipeline optimised for central processing (\texttt{en\_core\_web\_lg}) to parse each sample using the named-entity recognition tagger to mask \acrfull{GPE} (\textit{Wellington}) and \acrfull{LOC} (\textit{park}). The pipeline was trained on the OntoNotes Release 5.0 \citep{weischedel_ralph_ontonotes_2013}. Using the same pipeline, I lemmatise and convert each token to lowercase while removing stopwords and punctuation. The final feature engineering step involves parsing multi-word entities by detecting collocations using the \texttt{Phraser} method from \texttt{Gensim}. I process the language data twice to produce concatenated bigrams (\textit{kia\_ora}) and trigrams (\textit{te\_reo\_māori}). I set the minimum count to one and the threshold to 50 across all conditions.

    \begin{table*}
            
            \caption{Summary of Communities}
            \label{tab:corpus_summary}

            \centering
            \scriptsize
            
            \renewcommand{\arraystretch}{1.4}
                \begin{tabularx}{\textwidth}{l
                *{3}{>{\centering\arraybackslash}X}c|
                *{4}{>{\centering\arraybackslash}X}|
                c}
                    \toprule
                    \multirow{2}{*}{Community} 
                    & \multicolumn{4}{c}{Words (Millions)} 
                    & \multicolumn{4}{c}{Mean} 
                    & \multirow{2}{*}{Max} \\
                    & \gls{rpost} & \gls{rstitle} & \gls{rstext}& \gls{rcomm} & \gls{rpost} & \gls{rstitle} & \gls{rstext}& \gls{rcomm} & \\
                    \midrule
                    \addlinespace[1em]
                    \multicolumn{10}{c}{Inner-Circle Country-level Communities} \\
                    \addlinespace[1em]
                    \texttt{r/canada} & 32.70 & 11.79 & 20.34 & 4,454.39 & 74.0 & 67.5 & 662.9 & 241.3 & 40,064 \\
                    \texttt{r/usa} & 10.00 & 1.25 & 1.84 & 30.12 & 72.9 & 54.8 & 587.0 & 293.0 & 36,112 \\
                    \texttt{r/ireland} & 16.79 & 12.52 & 35.56 & 1,881.12 & 67.8 & 53.3 & 458.9 & 178.3 & 23,271 \\
                    \texttt{r/unitedkingdom} & 26.41 & 9.84 & 16.65 & 1,746.86 & 72.7 & 67.3 & 758.6 & 236.7 & 39,598 \\
                    \texttt{r/australia} & 29.02 & 11.88 & 21.29 & 2,783.28 & 74.4 & 60.0 & 541.0 & 215.1 & 38,665 \\
                    \texttt{r/newzealand} & 10.84 & 7.66 & 30.72 & 1,744.57 & 67.2 & 56.2 & 608.6 & 198.7 & 33,629 \\
                    \addlinespace[2em]
                    \multicolumn{10}{c}{Outer-Circle Country-level Communities} \\
                    \addlinespace[1em]
                    \texttt{r/Kenya} & 1.07 & 1.03 & 5.07 & 71.62 & 63.0 & 42.3 & 417.0 & 150.6 & 31,527 \\
                    \texttt{r/southafrica} & 4.90 & 2.91 & 9.96 & 331.78 & 60.4 & 51.5 & 529.8 & 212.2 & 28,735 \\
                    \texttt{r/india} & 71.62 & 31.01 & 84.02 & 2,526.33 & 73.7 & 64.5 & 733.5 & 202.8 & 39,880 \\
                    \texttt{r/pakistan} & 9.51 & 4.93 & 13.03 & 507.43 & 66.3 & 56.9 & 533.9 & 212.0 & 36,387 \\
                    \texttt{r/malaysia} & 8.16 & 4.07 & 12.49 & 618.93 & 68.0 & 55.3 & 673.2 & 179.0 & 37,491 \\
                    \texttt{r/Philippines} & 21.83 & 17.22 & 46.30 & 1,897.41 & 65.2 & 52.0 & 514.9 & 138.7 & 39,636 \\
                    \addlinespace[2em]
                    \multicolumn{10}{c}{New Zealand City-level Communities} \\
                    \addlinespace[1em]
                    \texttt{r/auckland} & 1.69 & 2.94 & 17.78 & 350.77 & 68.5 & 44.4 & 406.4 & 164.9 & 39,004 \\
                    \texttt{r/Tauranga} & 0.04 & 0.07 & 0.46 & 5.67 & 61.9 & 34.8 & 302.4 & 159.5 & 9,886 \\
                    \texttt{r/thetron} & 0.11 & 0.18 & 1.23 & 13.27 & 62.7 & 37.4 & 334.2 & 154.6 & 14,363 \\
                    \texttt{r/Wellington} & 1.74 & 2.61 & 12.27 & 203.47 & 84.3 & 51.2 & 403.0 & 166.1 & 25,861 \\
                    \texttt{r/chch} & 0.44 & 0.77 & 4.92 & 69.54 & 66.4 & 39.2 & 340.4 & 164.3 & 9,956 \\
                    \texttt{r/dunedin} & 0.11 & 0.26 & 1.2 & 12.55 & 61.5 & 39.2 & 332.9 & 170.5 & 9,301 \\
                    \addlinespace[1em]
                    \bottomrule
                \end{tabularx}
    
            \tablenoteparagraph{\textbf{Table Note}: This table provides a descriptive summary of the datasets used, categorised by geographic level: inner-circle and outer-circle country-level communities, and New Zealand city-level communities. For each community, the table details the total word count in millions, mean word length, and the maximum observation length across four distinct text-types: submission titles (\gls{rpost}), selfpost titles (\gls{rstitle}), selfpost body text (\gls{rstext}), and comments (\gls{rcomm}). Word counts are rounded to two decimal places, while mean lengths are rounded to one decimal place; all data is sourced from Reddit/Pushshift \citep{baumgartner_pushshift_2020}.}
            
    \end{table*}
    
\subsubsection{Sampling Procedure}
\label{dialect:sampling_procedure}

    I adapt the population-based sampling technique in \citet{dunn_geographically-balanced_2020} to control for geographic sampling balance. Based on \gls{rcomm} alone, the largest country-level community (\texttt{r/canada}) is approximately 180 times the size of the smallest country-level community (\texttt{r/usa}). For this reason, I downscaled the class imbalance by using the minority class (in this case \texttt{r/usa}) as the baseline for my minimum group size. I calculate the minimum group size as a measure equivalent to a quarter of the total observations available in the minority class. With the minimum group size measure, I apply the following sampling procedures to simulate different forms of geographic sampling bias:
    
    \begin{itemize}
    
        \item \textbf{Balanced Sampling}: this procedure is to simulate class balance in the train and test data. The number of observations per class is based on the minimum group size figure which I calculate for each configuration.
        
        \item \textbf{Proportional Sampling}: this procedure is to simulate class imbalance. To simulate class imbalance, I adjust the fraction argument to account for corpus size. For the country-level communities, I set the percentage-based fraction to 0.001 for \gls{rcomm} and for the other text-types I set the fraction to 0.05. As for the city-level communities, I set the percentage-based fraction to 0.01 for \gls{rcomm} and similarly for the other text-types I set the fraction to 0.1. 
        
        \item \textbf{Random Sampling}: the number of items in the train and test data is based on the minimum group size multiplied by the number of groups. While I control for the total number of observations, the number of observations between classes is random.
        
    \end{itemize}

\subsubsection{Classifier Training}
\label{dialect:model_development}

    I use \acrshort{SVM} \citep{joachims_text_1998} as my shallow classifier. This is a supervised learning method that is effective in high-dimensional spaces (such as multi-class classification). I use the linear classifier with stochastic gradient descent from the \texttt{scikit-learn} \citep{pedregosa_scikit-learn_2011} Python library to train my shallow classifiers.

\subsubsection{Evaluation}
\label{dialect:evaluation}

    Some key metrics to evaluate the performance of a text classification model include accuracy, precision, and recall. Alternatively, macro average and weighted average $F_1$-scores are more suitable to assess the performance of a classification model while accounting for class imbalance \citep{dunn_natural_2022}. The $F_1$-score is calculated using the following formula:
    
    \vspace{24pt}

    $F_1$-score$=2{\times}\displaystyle\frac{Precision{\times}Recall}{Precision{+}Recall}$

    \vspace{24pt}

    In brief, the macro average $F_1$-score is the unweighted arithmetic mean of all classes, while the weighted average $F_1$-score takes into account the support, or number of observations, in the test data.

\subsubsection{Hypotheses and Predictions}

    In terms of the sampling procedure, I predict that the balanced sampling procedure will produce the best model performance for the country-level communities (both inner- and outer-circle) based on the weighted average $F_1$-score, followed by the proportional and, finally, random sampling strategies. I expect to see the greatest variance in model performance in the proportional sampling procedure, as a high weighted average $F_1$-score does not necessarily translate to high performance in multiclass classification. I can monitor this variance in reference to the macro average $F_1$-score, which I predict will be low in the proportional sampling procedure. As for the city-level communities, I predict low performance due to a high degree of similarity between speakers of \acrshort{NZE} \citep{bauer_can_2002}. I do not expect to observe differences between Reddit or Twitter\textsuperscript{X}.

\subsection{Country-level Communities}

    The results for the inner-circle country-level communities are presented in Table \ref{tab:f1_country_inner}. Based on the weighted average $F_1$-score, the best performing configuration for the selfpost body texts (\gls{rstext}) was the baseline model with no data cleaning procedures using the proportional sampling procedure ($F_1=0.81$). The macro average $F_1$-score was 0.81, suggesting good model performance. With reference to the suite of models where named entities were removed (\texttt{Processed}), I observed a decrease in weighted average and macro average $F_1$-scores across all three sampling conditions, suggesting that named entities were meaningful predictors.
    
    As for the comments (\gls{rcomm}), balanced sampling appeared to offer the best performance in the baseline condition based on the weighted average $F_1$-score (0.72). As the minority class, \texttt{r/usa} was the most impacted by class imbalance across both text-types. With reference to \gls{rstext}, \texttt{r/usa} saw the best performance in the balanced sampling condition ($F_1=0.79$). There was a drop in performance in the proportional sampling condition ($F_1=0.44$), before dropping to 0 in the random sampling condition. A similar result was observed for \gls{rcomm}. Even though \texttt{r/usa} served as the benchmark for the minimum group size, the differences in corpus size between \texttt{r/usa} and the other country-level communities were reflected in the $F_1$-scores.
    
    As a point of comparison, I now consider the results for the outer-circle country-level communities. The results are presented in Table \ref{tab:outer_circle}. Based on the weighted average $F_1$-score, the best performing configuration for both text-types was the baseline condition with no data cleaning using balanced sampling. The weighted average $F_1$-score for \gls{rstext} was 0.87 and for \gls{rcomm} was 0.83. Similar to \texttt{r/usa}, \texttt{r/Kenya} was the most impacted by class imbalance as the minority class. \texttt{r/Kenya} saw the best performance in the balanced sampling condition ($F_1=0.80$) but observed a drop in performance in the proportional ($F_1=0.06$) and random ($F_1=0.14$) sampling conditions. A similar result was observed for \gls{rcomm}. As with the inner-circle country-level communities, removing named entities resulted in a decrease in model performance.

    \begin{table*}
            
            \caption{Performance Metrics for Inner-Circle Communities}
            \label{tab:f1_country_inner}
            
            \begin{subtable}[t]{\textwidth}
            \centering
            \scriptsize
            \caption{\gls{rstext}}
            \renewcommand{\arraystretch}{1.4}
                \begin{tabularx}{\textwidth}{l
                *{3}{>{\centering\arraybackslash}X} |
                *{3}{>{\centering\arraybackslash}X}}
                    \toprule
                    \multirow{2}{*}{Community} & \multicolumn{3}{c}{Baseline} & \multicolumn{3}{c}{Processed}\\
                    & Balanced & Proportional & Random & Balanced & Proportional & Random \\
                    \midrule
                    \addlinespace[1em]
                    \texttt{r/canada} & 0.82 & \textbf{0.94} & 0.87 & 0.65 & 0.85 & 0.69 \\
                    \texttt{r/usa} & \textbf{0.79} & 0.44 & - & 0.68 & 0.53 & - \\
                    \texttt{r/ireland} & 0.70 & \textbf{0.88} & 0.77 & 0.53 & 0.80 & 0.68 \\
                    \texttt{r/unitedkingdom} & 0.71 & \textbf{0.85} & 0.71 & 0.61 & 0.76 & 0.55 \\
                    \texttt{r/australia} & 0.74 & \textbf{0.88} & 0.70 & 0.58 & 0.77 & 0.51 \\
                    \texttt{r/newzealand} & 0.73 & \textbf{0.89} & 0.76 & 0.56 & 0.79 & 0.61 \\ 
                    \addlinespace[1em]
                    Macro Average & 0.75 & \textbf{0.81} & 0.64 & 0.60 & 0.75 & 0.51 \\
                    Weighted Average & 0.75 & \textbf{0.88} & 0.76 & 0.60 & 0.79 & 0.61 \\
                    \addlinespace[1em]
                    \bottomrule
                \end{tabularx}
            \end{subtable}\vspace{12pt}
        
            \begin{subtable}[t]{\textwidth}
                \centering
                \scriptsize
                \caption{\gls{rcomm}}
                \renewcommand{\arraystretch}{1.4}
                \begin{tabularx}{\textwidth}{l
                *{3}{>{\centering\arraybackslash}X} |
                *{3}{>{\centering\arraybackslash}X}}
                    \toprule
                    \multirow{2}{*}{Community} & \multicolumn{3}{c}{Baseline} & \multicolumn{3}{c}{Processed}\\
                    & Balanced & Proportional & Random & Balanced & Proportional & Random \\
                    \midrule
                    \addlinespace[1em]
                    \texttt{r/canada} & 0.75 & \textbf{0.77} & \textbf{0.77} & 0.68 & 0.72 & 0.72 \\
                    \texttt{r/usa} & \textbf{0.79} & - & - & 0.74 & - & - \\
                    \texttt{r/ireland} & \textbf{0.71} & 0.69 & 0.63 & 0.60 & 0.58 & 0.54 \\
                    \texttt{r/unitedkingdom} & \textbf{0.68} & 0.66 & 0.63 & 0.62 & 0.58 & 0.56 \\
                    \texttt{r/australia} & \textbf{0.70} & \textbf{0.70} & 0.69 & 0.61 & 0.60 & 0.58 \\
                    \texttt{r/newzealand} & \textbf{0.68} & 0.62 & 0.64 & 0.56 & 0.43 & 0.45 \\ 
                    \addlinespace[1em]
                    Macro Average & \textbf{0.72} & 0.57 & 0.56 & 0.64 & 0.49 & 0.47 \\
                    Weighted Average & \textbf{0.72} & 0.71 & 0.70 & 0.64 & 0.62 & 0.61 \\
                    \addlinespace[1em]
                    \bottomrule
                \end{tabularx}
            \end{subtable}\vspace{6pt}
    
            \tablenoteparagraph{\textbf{Table Note}: This table presents model performance metrics for submission selftexts (\gls{rstext}) and comments (\gls{rcomm}) across inner-circle country-level communities, using shallow Linear \acrshort{SVM} classifiers. Evaluation covers balanced, proportional, and random sampling procedures cross-referenced with two data cleaning conditions: a baseline with no cleaning (\texttt{Baseline}) and standard processing (\texttt{Processed}). Results indicate that the best performing model configuration for \gls{rstext} utilized proportional sampling, while \gls{rcomm} performed best with balanced sampling; notably, both text-types achieved their highest weighted average $F_1$-scores in the \texttt{Baseline} condition. Observations with fewer than 500 words were excluded. Columns detail the community name and sampling procedures grouped by cleaning condition, while rows provide $F_1$-scores for each state. All data is sourced from Reddit/Pushshift \citep{baumgartner_pushshift_2020}.}
        
    \end{table*}

    \begin{table*}
        \scriptsize
        \centering
        \renewcommand{\arraystretch}{1.4}
            
            \caption{Performance Metrics for Outer-Circle Communities}
            \label{tab:f1_country_outer}

            \begin{subtable}[t]{\textwidth}
            \centering
            \caption{\gls{rstext}}
            \renewcommand{\arraystretch}{1.4}
                \begin{tabularx}{\textwidth}{l
                *{3}{>{\centering\arraybackslash}X} |
                *{3}{>{\centering\arraybackslash}X}}
                    \toprule
                    \multirow{2}{*}{Community} & \multicolumn{3}{c}{Baseline} & \multicolumn{3}{c}{Processed}\\
                    & Balanced & Proportional & Random & Balanced & Proportional & Random \\
                    \midrule
                    \addlinespace[1em]
                    \texttt{r/Kenya} & \textbf{0.85} & 0.06 & 0.14 & 0.71 & - & - \\
                    \texttt{r/southafrica} & \textbf{0.86} & 0.38 & 0.45 & 0.75 & 0.12 & 0.28 \\
                    \texttt{r/pakistan} & \textbf{0.90} & 0.37 & 0.34 & 0.77 & 0.06 & 0.05 \\
                    \texttt{r/india} & \textbf{0.84} & 0.80 & 0.82 & 0.69 & 0.76 & 0.77 \\
                    \texttt{r/philippines} & \textbf{0.91} & 0.78 & 0.80 & 0.82 & 0.69 & 0.70 \\
                    \texttt{r/malaysia} & \textbf{0.86} & 0.37 & 0.55 & 0.75 & 0.08 & 0.08 \\
                    \addlinespace[1em]
                    Macro Average & \textbf{0.87} & 0.46 & 0.52 & 0.75 & 0.28 & 0.31 \\
                    Weighted Average & \textbf{0.87} & 0.69 & 0.73 & 0.75 & 0.59 & 0.62 \\
                    \addlinespace[1em]
                    \bottomrule
                \end{tabularx}
            \end{subtable}\vspace{12pt}
        
            \begin{subtable}[t]{\textwidth}
                \centering
                \scriptsize
                \caption{\gls{rcomm}}
                \renewcommand{\arraystretch}{1.4}
                \begin{tabularx}{\textwidth}{l
                *{3}{>{\centering\arraybackslash}X} |
                *{3}{>{\centering\arraybackslash}X}}
                    \toprule
                    \multirow{2}{*}{Community} & \multicolumn{3}{c}{Baseline} & \multicolumn{3}{c}{Processed}\\
                    & Balanced & Proportional & Random & Balanced & Proportional & Random \\
                    \midrule
                    \addlinespace[1em]
                    \texttt{r/Kenya} & \textbf{0.80} & 0.03 & 0.03 & 0.72 & - & - \\
                    \texttt{r/southafrica} & \textbf{0.82} & 0.46 & 0.43 & 0.77 & 0.37 & 0.34 \\
                    \texttt{r/pakistan} & \textbf{0.87} & 0.50 & 0.49 & 0.78 & 0.31 & 0.27 \\
                    \texttt{r/india} & 0.79 & \textbf{0.80} & 0.79 & 0.72 & 0.77 & 0.77 \\
                    \texttt{r/philippines} & \textbf{0.88} & 0.72 & 0.69 & 0.83 & 0.64 & 0.61 \\
                    \texttt{r/malaysia} & \textbf{0.83} & 0.38 & 0.36 & 0.74 & 0.18 & 0.20 \\
                    \addlinespace[1em]
                    Macro Average & \textbf{0.83} & 0.48 & 0.47 & 0.76 & 0.38 & 0.36 \\
                    Weighted Average & \textbf{0.83} & 0.69 & 0.68 & 0.76 & 0.61 & 0.60 \\
                    \addlinespace[1em]
                    \bottomrule
                \end{tabularx}
            \end{subtable}\vspace{6pt}
    
            \tablenoteparagraph{\textbf{Table Note}: This table presents model performance metrics for submission selftexts (\gls{rstext}) and comments (\gls{rcomm}) across outer-circle country-level communities, using shallow Linear \acrshort{SVM} classifiers. Evaluation covers three sampling procedures—balanced, proportional, and random—cross-referenced with two data cleaning conditions: a baseline with no cleaning (\texttt{Baseline}) and standard processing (\texttt{Processed}). Results indicate that the best performing model configuration for both text-types, based on weighted average $F_1$-scores, utilized balanced sampling with no additional data cleaning; observations with fewer than 500 words were excluded. Columns detail the community name and the sampling procedures grouped by cleaning condition, while rows provide $F_1$-scores for each state. All data is sourced from Reddit/Pushshift \citep{baumgartner_pushshift_2020}.}
        
    \end{table*}

\subsubsection{Sample Size and Stability of Classifiers}

\begin{figure}
      \centering
      
        \begin{subfigure}[b]{0.75\textwidth}
            \centering
            \includegraphics[width=\textwidth]{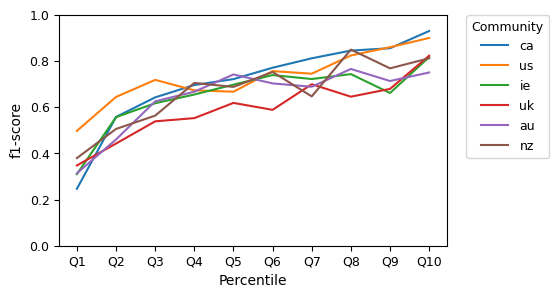}
            \subcaption{Baseline}
        \end{subfigure}
        
        \begin{subfigure}[b]{0.75\textwidth}
            \centering
            \includegraphics[width=\textwidth]{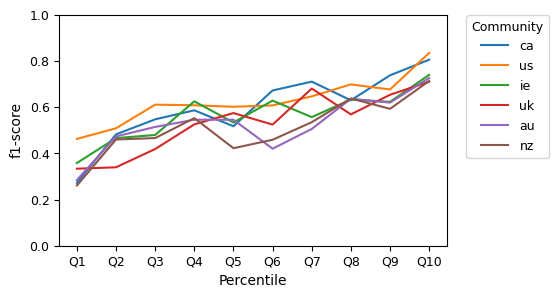}
            \subcaption{Processed}
        \end{subfigure}    
            
      \vspace{6pt}
      \caption{Lineplots of Classifier Performance by Sample Size}
      \label{fig:f1-sample_size}
      
        \vspace{6pt}
        \captionsetup{font=scriptsize, labelformat=empty, justification=justified, singlelinecheck=false}
        \caption*{\setstretch{2}\textbf{Description}: This figure presents subfigures comparing model performance metrics for inner-circle country-level communities based on comments (\gls{rcomm}) under two conditions: (a) a baseline with no data cleaning and (b) a standard data cleaning procedure. Shallow Linear \acrshort{SVM} classifiers were utilised, with the \textit{x}-axis representing $F_1$-scores and the \textit{y}-axis representing sample sizes grouped into percentiles; notably, $F_1$-scores increase as sample size increases, with the best performing configurations for both procedures occurring in the tenth percentile. The legend identifies the six communities as \texttt{r/canada} (\textsc{ca}), \texttt{r/usa} (\textsc{us}), \texttt{r/ireland} (\textsc{ie}), \texttt{r/unitedkingdom} (\textsc{uk}), \texttt{r/australia} (\textsc{au}), and \texttt{r/newzealand} (\textsc{nz}), with data sourced from Reddit/Pushshift \citep{baumgartner_pushshift_2020}.}
      
    \end{figure}

    Based on the results from the inner-circle and outer-circle country-level communities, sampling appeared to play a role in classification performance. To determine the role of sample size or the length of an observation, I split the comments (\gls{rcomm}) from the inner-circle country-level communities into ten quantiles. As a point of reference, the mean word length of the first quantile was 47.9 words, and the mean word length of the tenth quantile was 2,473.6 words. Regardless of the data cleaning procedure, sample size had an impact on model performance. Classifier performance improved from the second quantile, which had a mean word length of 111.9 words. I visualised the interaction between sample size and classifier performance in Figure \ref{fig:f1-sample_size}.
    
\begin{figure}
      \centering
      
            \includegraphics[width=\textwidth]{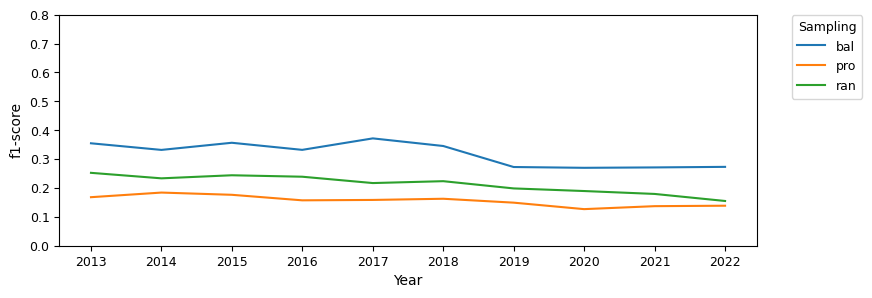} 

      \vspace{6pt}
      \caption{Stability of Dialect Classification Over Time}
      \label{fig:f1_time-comment}
      
        \vspace{6pt}
        \captionsetup{font=scriptsize, labelformat=empty, justification=justified, singlelinecheck=false}
        \caption*{\setstretch{2}\textbf{Description}: This figure displays model performance metrics for inner-circle country-level communities based on comments (\gls{rcomm}), comparing three data sampling procedures: balanced sampling (\textsc{bal}), proportional sampling (\textsc{pro}), and random sampling (\textsc{ran}). Shallow Linear \acrshort{SVM} classifiers were trained on all observations prior to January 2013, with a consistent held-out test set applied to each procedure and grouped by quarter. The visualisation reveals a decrease in $F_1$-scores over time, with the \textit{x}-axis representing the scores and the \textit{y}-axis representing years; all data is sourced from Reddit/Pushshift \citep{baumgartner_pushshift_2020}.}
      
    \end{figure}

    Lastly, I consider the stability of the classifiers over time. In order to determine the stability of inner-circle country-level communities, I trained a classifier using historic training data and iterated over a held-out test set for each period. As part of my analysis, I restricted my model to comments (\gls{rcomm}) generated before 1 January 2013 and aggregated the remaining observations into yearly quarters. I present the weighted average $F_1$-scores for inner-circle country-level community classification in \gls{rcomm} in Figure \ref{fig:f1_time-comment}. It was clear that dialect classification also remained stable over time on Reddit. I observed a gradual decrease in model performance over time for all country-level communities across the three sampling procedures.

\subsection{City-Level Communities}
\label{dialect:city_classification}

    I now shift my focus to the six New Zealand city-level communities: \texttt{r/auckland}, \texttt{r/Tauranga}, \texttt{r/thetron}, \texttt{r/Wellington}, \texttt{r/chch}, and \texttt{r/dunedin}. In contrast to the country-level communities, I should not observe significant differences between the New Zealand city-level communities. With reference to Table \ref{tab:f1_city-level}, I begin with the results of the baseline model with no data processing applied. The weighted average $F_1$-scores for \gls{rstext} and \gls{rcomm} were 0.53 in the baseline model. Considering the lack of regional variation in \acrshort{NZE} across the country, the classifier performance was still much higher than expected. After applying the same pipeline used for the country-level communities to remove named entities, the model performance fell to 0.31 for \gls{rstext} and 0.43 for \gls{rcomm}.

    To determine the impact of named entities on classification performance, I extracted the top 10 features based on classification weight for each city-level community. I present these features in Table \ref{tab:top_features-baseline}. Unsurprisingly, only seven of the sixty features across the city-level communities were not named entities. Even after applying the pipeline used to process the country-level communities, the number of non-named entities still only made up a minority of the top features in my city-level communities ($n=16$). Therefore, my next step was to refine the existing data cleaning pipeline.
    
    I updated the \acrshort{GPE} and \acrshort{LOC} entity patterns in \texttt{spaCy} with patterns specific to the context of New Zealand. This included an additional 111,606 street and road names from Land Information New Zealand, 46,308 place names from the New Zealand Gazetteer \citep{nga_pou_taunaha_o_aotearoa__new_zealand_geographic_board_new_2025}, and 3,882 geographic areas from Stats NZ. The Stats NZ geographic areas included Statistical Areas 1 and 2, urban-rural areas, wards and Māori wards, community board areas, territorial authorities, and regional council areas. As I want to account for orthographic variants with and without the \gls{tohuto} (such as \textit{Ōtautahi} and \textit{Otautahi}), I synthetically generated orthographic variants without the \gls{tohuto}. The updated pipeline included 61,796 New Zealand-specific place names.

    \begin{table*}
            
            \caption{Performance Metrics for City-level Communities}
            \label{tab:f1_city-level}
            
            \begin{subtable}[t]{\textwidth}
            \scriptsize
            \subcaption{\gls{rstext}}
            \renewcommand{\arraystretch}{1.4}
                  \begin{tabularx}{\textwidth}{l*{3}{>{\centering\arraybackslash}X}|
                  *{1}{>{\centering\arraybackslash}X}}
                    \toprule
                    Community & Baseline & Processed & Localised & Support \\
                    \midrule
                    \addlinespace[1em]
                    \texttt{r/auckland} & \textbf{0.44} & 0.22 & 0.19 & 44 \\
                    \texttt{r/Tauranga} & \textbf{0.57} & 0.40 & 0.38 & 45 \\
                    \texttt{r/thetron} & \textbf{0.48} & \textbf{0.48} & 0.26 & 45 \\
                    \texttt{r/Wellington} & \textbf{0.56} & 0.27 & 0.21 & 44 \\
                    \texttt{r/chch} & \textbf{0.45} & 0.19 & 0.21 & 44 \\
                    \texttt{r/dunedin} & \textbf{0.65} & 0.27 & 0.24 & 44 \\
                    \addlinespace[1em]
                    Macro Average & \textbf{0.53} &  0.31 & 0.25 & 266 \\
                    Weighted Average & \textbf{0.53} & 0.31 & 0.25 & 266 \\
                    \addlinespace[1em]
                    \bottomrule
                \end{tabularx}
            \end{subtable}\vspace{6pt}
            
            \begin{subtable}[t]{\textwidth}
            \scriptsize
            \subcaption{\gls{rcomm}}
            \renewcommand{\arraystretch}{1.4}
                  \begin{tabularx}{\textwidth}{l*{3}{>{\centering\arraybackslash}X}|
                  *{1}{>{\centering\arraybackslash}X}}
                    \toprule
                    Community & Baseline & Processed & Localised & Support \\
                    \midrule
                    \addlinespace[1em]
                    \texttt{r/auckland} & \textbf{0.42} & 0.33 & 0.30 & 382 \\
                    \texttt{r/Tauranga} & \textbf{0.65} & 0.53 & 0.51 & 382 \\
                    \texttt{r/thetron} & \textbf{0.52} & 0.45 & 0.34 & 383 \\
                    \texttt{r/Wellington} & \textbf{0.49} & 0.31 & 0.28 & 382 \\
                    \texttt{r/chch} & \textbf{0.48} & 0.38 & 0.33 & 382 \\
                    \texttt{r/dunedin} & \textbf{0.65} & 0.55 & 0.51 & 383 \\
                    \addlinespace[1em]
                    Macro Average & \textbf{0.53} & 0.43 & 0.38 & 2,294 \\
                    Weighted Average & \textbf{0.53} & 0.43 & 0.38 & 2,294 \\
                    \addlinespace[1em]
                    \bottomrule
                \end{tabularx}
            \end{subtable}\vspace{6pt}
    
            \tablenoteparagraph{\textbf{Table Note}: This table presents model performance metrics for submission selftexts (\gls{rstext}) and comments (\gls{rcomm}) across New Zealand city-level communities, utilizing shallow Linear \acrshort{SVM} classifiers with proportional sampling. Evaluation covers three data cleaning conditions: a baseline with no cleaning (\texttt{Baseline}), standard processing (\texttt{Processed}), and a localised condition with New Zealand-specific named entities removed (\texttt{Localised}). Results indicate that the best performing models for both text-types, based on weighted average $F_1$-scores, were achieved in the \texttt{Baseline} condition; observations with fewer than 500 words were excluded. Columns detail the community name, the cleaning procedure, and the number of observations in the held-out test set (\texttt{Support}), while rows provide $F_1$-scores with the highest values highlighted. All data is sourced from Reddit/Pushshift \citep{baumgartner_pushshift_2020}.}

    \end{table*}

    As a preliminary check, I applied the updated \texttt{EntityRecognizer} on the city-level communities, where it identified 6,349 \acrshort{GPE} and 2,290 \acrshort{LOC} patterns across the six communities. I only observed a minor increase from the default pipeline (\acrshort{GPE} $=6,335$; \acrshort{GPE} $=2,288$). Other changes to the data cleaning pipeline for the New Zealand city-level communities included removing strings that started with \texttt{u/} (user name) and \texttt{r/} (community name) and non-alphanumeric characters. Once I applied the \texttt{Localised} data cleaning procedures, I saw a drop in the weighted average $F_1$-score for \gls{rstext} to 0.25 and \gls{rcomm} to 0.38 (as seen in Table \ref{tab:f1_city-level}). This suggests the extra data cleaning steps achieved the intended task of removing New Zealand-specific place names. I present the top 10 features for the six city-level communities in Table \ref{tab:top_features-stacked}.

    \begin{table*}
        \scriptsize
        \centering
        \renewcommand{\arraystretch}{1.4}
            
            \caption{Predictive Features for City-level Communities in Baseline Models}
            \label{tab:top_features-baseline}

            \begin{subtable}[t]{\textwidth}
            \centering
            \caption{Baseline}
            \renewcommand{\arraystretch}{1.4}
                \begin{tabularx}{\textwidth}{*{6}{>{\centering\arraybackslash}X}}
                \toprule
                \texttt{r/auckland} & \texttt{r/Tauranga} & \texttt{r/thetron} & \texttt{r/Wellington} & \texttt{r/chch} & \texttt{r/dunedin} \\
                \midrule
                \addlinespace[1em]
                auckland* & tauranga* & hamilton* & wellington* & christchurch* & dunedin* \\
                lockdown & mount* & waikato* & hutt* & chch* & otago* \\
                shore* & papamoa* & wintec* & wcc* & riccarton* & halls \\
                eden* & pa* & raglan* & newtown* & ccc* & hall \\
                newmarket* & greerton* & cambridge* & petone* & rolleston* & carrington* \\
                passport & cameron* & frankton* & welly* & canterbury* & arana* \\
                queen* & tauriko* & rototuna* & aro* & uc* & hsfy* \\
                vaccination & bayfair* & river* & vic* & lincoln* & hayward* \\
                lockdowns & candidates & rapa* & porirua* & ecan* & knox* \\
                takapuna* & tcc* & hillcrest* & cuba* & hagley* & flats \\
                \addlinespace[1em]
                \bottomrule
            \end{tabularx}
            \end{subtable}\vspace{12pt}

            \begin{subtable}[t]{\textwidth}
            \centering
            \caption{Processed}
            \renewcommand{\arraystretch}{1.4}
                \begin{tabularx}{\textwidth}{*{6}{>{\centering\arraybackslash}X}}
                \toprule
                \texttt{r/auckland} & \texttt{r/Tauranga} & \texttt{r/thetron} & \texttt{r/Wellington} & \texttt{r/chch} & \texttt{r/dunedin} \\
                \midrule
                \addlinespace[1em]
                lockdown & mount* & hamilton* & wcc* & rolleston* & hall \\
                shore* & papamoa* & raglan* & wellington* & riccarton* & flat \\
                hamas* & tga* & wintec* & welly* & ccc* & hsfy* \\
                miq* & cameron\_road* & waikato* & hutt* & uc* & student \\
                britomart* & tcc* & river & wairarapa* & quake & arana* \\
                apartment & bayfair* & rototuna* & seat & ecan* & carrington* \\
                situation & tauranga* & frankton* & petone* & earthquake & dunedin* \\
                woman & welcome\_bay* & te\_rapa* & courtney\_place* & chch* & paper \\
                vaccine\_passport & vote & tron* & vic* & lincoln* & dcc* \\
                anger & omokoroa* & dinsdale* & kelburn* & wigram* & college \\
                \addlinespace[1em]
                \bottomrule
            \end{tabularx}
            \end{subtable}\vspace{12pt}
    
            \tablenoteparagraph{\textbf{Table Note}: This table lists the top 10 predictive features per class for New Zealand city-level communities, ranked by classifier weights using a Linear \acrshort{SVM} with proportional sampling and no initial data cleaning. Observations with fewer than 500 words were excluded; notably, standard data cleaning procedures (\texttt{Processed}) yielded little improvement in removing named entities, as only seven of the sixty identified features - including \textit{passport}, \textit{vaccination}, \textit{lockdowns}, \textit{halls}, \textit{hall}, and \textit{flats} - were not named entities. Columns represent the city-level communities, while rows show the predictive features with named entities marked by an asterisk (*); all data is sourced from Reddit/Pushshift \citep{baumgartner_pushshift_2020}.}
        
    \end{table*}

    \begin{table*}
        \scriptsize
        \centering
        \renewcommand{\arraystretch}{1.4}
            
            \caption{Top 10 Predictors for City-level Communities}
            \label{tab:top_features-stacked}
        
                \renewcommand{\arraystretch}{1.4}
                \begin{tabularx}{\textwidth}{*{6}{>{\centering\arraybackslash}X}}
                \toprule
                \texttt{r/auckland} & \texttt{r/Tauranga} & \texttt{r/thetron} & \texttt{r/Wellington} & \texttt{r/chch} & \texttt{r/dunedin} \\
                \midrule
                \addlinespace[1em]
                lockdown & mount* & wintec* & wcc* & ccc* & hall \\
                shore & tga* & river & welly* & uc* & flat \\
                hamas* & bayfair* & hamilton* & hutt* & quake & arana* \\
                miq* & tcc* & tron* & seat & ecan* & hsfy* \\
                britomart* & ward & ham* & wairarapa* & chch* & student \\
                apartment & vote & brain\_injury & vic* & earthquake & paper \\
                situation & commissioner & hcc* & courtney\_place* & cathedral & dcc* \\
                woman & cameron\_road* & lake & aro* & rebuild & tutorial \\
                vaccine\_passport & skate\_park & expressway & cable\_car & stadium & health\_sci \\
                anger & - & motel & tory* & rabbit & choice\_hall \\
                \addlinespace[1em]
                \bottomrule
            \end{tabularx}
    
            \tablenoteparagraph{\textbf{Table Note}: This table presents the top 10 predictive features for each city-level community class, ranked by classifier weights after the removal of New Zealand-specific named entities. The shallow classifiers were trained using a Linear \acrshort{SVM} with proportional sampling; the addition of 161,796 New Zealand-specific placenames to the named entity recognition pipeline successfully reduced the presence of named entities in the localised condition. Columns list the New Zealand city-level communities, while rows contain the predictive features, with remaining named entities indicated by an asterisk (*). All data is sourced from Reddit/Pushshift \citep{baumgartner_pushshift_2020}.}
        
    \end{table*}

\section{Interim Summary}
\label{dialect:interim_summary}

    On the whole, the text classifiers by themselves were not meaningful as dialect models. This because the differences observed between the New Zealand city-level communities was not the result of latent linguistic variation, but the presence of named entities. This is consistent with recent findings were named entities such as place-names, place-specific named entities, or people associated with these countries were the main predictors \citep{dunn_stability_2022}. In using text classification as a diagnostic tool, I was able to identify the role of sample size and named entities in text classification models. I determined the role of named entities in distinguishing these city-level communities. There was direct link between sample size and classifier performance \citep{figueroa_text_2013}. This means of the four text-types, only the selfpost body texts (\gls{rstext}) and comments (\gls{rcomm}) were suitable for language modelling.

\section{Embedding Models}
\label{dialect:embedding_models}

    Embedding models have been shown to be a promising alternative \citep{dunn_variation_2023}. One of the first Word2Vec embedding models made widely available was the Google News dataset, trained on 100 billion words with a vocabulary of 3 million words and phrases \citep{mikolov_efficient_2013}. This is not a feasible benchmark for most language varieties; however, even with limited data, Word2Vec models can be effectively trained on small, domain-specific corpora. As a point of comparison, Word2Vec embedding models have been trained on low-resourced languages using datasets ranging from 1.8 thousand to 41 million words \citep{stringham_evaluating_2020}. Of relevance to New Zealand, Word2Vec models identified meaningful relationships between hashtags in the 21-million-word Māori Loanword Twitter Corpus (\citealp{trye_maori_2019}; \citealp{trye_hybrid_2020}).

\subsection{Methodology}
\label{dialect:embedding_method}

    My aim is to train three candidate models using the New Zealand city-level communities (\acrshort{NZM}), the inner-circle country-level communities (\acrshort{ICM}), and the outer-circle country-level communities (\acrshort{OCM}). Including the baseline \acrshort{GOO} model \citep{mikolov_efficient_2013}, I have four candidate models that represent four different admixtures of English varieties. Using these language embedding models, I can then examine the user-informed semantic phenomena from Chapter \ref{chap:user_intuitions}: User Intuitions and Place Identity to determine semantic variation across my four candidate embedding models: \acrshort{NZM}, \acrshort{ICM}, \acrshort{OCM}, and \acrshort{GOO}.

\subsubsection{User-informed Semantic Variables}

    The thirteen user-informed semantic variables are: \textsc{bum}, \textsc{dick}, \textsc{fag}, \textsc{flannel}, \textsc{football}, \textsc{gay}, \textsc{kiwi}, \textsc{pudding}, \textsc{rubber}, \textsc{tea}, \textsc{tramp}, \textsc{tuna}, and \textsc{twink}. Due to the different semantic processes involved (broadening, metonymy, narrowing, pejoration, metaphor, and polysemy), I have adopted various strategies to gauge semantic shift within these variables. These strategies include concept-related targets (such as \textit{smoke} and \textit{sex} to prompt \textsc{bum} for \textit{to beg} and \textit{to have anal sex}), word synonym targets (such as \textit{richard} and \textit{jerk} to prompt \textsc{dick}), and a combination of the two, such as concept-related word synonym targets (\textit{towel} and \textit{shirt} to prompt \textsc{flannel}). I present the source and target prompts for the user-informed semantic variables in Table \ref{tab:source_target}.

    \begin{table}
        \scriptsize
        \centering
            
            \caption{Source and Targets for User-informed Semantic Variables}
            \label{tab:source_target}
            
            \renewcommand{\arraystretch}{1.4}
                \begin{tabularx}{\textwidth}{XXlll}
                    \toprule
                    Variable & Source & Conservative Target & Innovative Target & Selection Criteria \\
                    \midrule
                    \addlinespace[1em]
                    \textsc{bum} & bum & smoke & sex & concept-related \\
                    \textsc{dick} & dick & richard & jerk & word synonyms \\
                    \textsc{fag} & fag & cigarette & person & concept-related \\
                    \textsc{football} & football & soccer, rugby & soccer, rugby & word synonyms \\
                    \textsc{flannel} & flannel & towel & shirt & concept-related word synonyms \\
                    \textsc{gay} & gay & unhappy & lesbian & word synonym antonyms \\
                    \textsc{kiwi} & kiwi & eagle & apple & concept-related \\
                    \textsc{pudding} & pudding & dinner & custard & word synonyms \\
                    \textsc{rubber} & rubber & pencil & sex & concept-related \\
                    \textsc{tea} & tea & dinner & coffee & word synonyms \\
                    \textsc{tramp} & tramp & mountain & street & concept-related \\
                    \textsc{tuna} & tuna & salmon & eel & concept-related \\
                    \textsc{twink} & twink & pen & bear & concept-related \\
                    \addlinespace[1em]
                    \bottomrule
                \end{tabularx}

            \tablenoteparagraph{\textbf{Table Note}: This table outlines the 65 user-informed lexical variable word pairs used for model evaluation, categorised by the specific variable, the source prompt, and the associated conservative and innovative target prompts. The final column details the selection criteria applied to each target prompt; all relevant social media data is sourced from Reddit/Pushshift \citep{baumgartner_pushshift_2020}.}
        
    \end{table}

\subsubsection{Sampling Procedure}

    As I observed in Table \ref{tab:corpus_summary}, each community cohort differs in size. Similar to the classification models, the number of words per sample has an impact on the performance of the embedding models. Because I want the embedding models to capture more contextual information from the training data, I adjusted the training sets for the city-level and country-level communities. Specifically, I incorporated the observations from the minority classes (\texttt{r/usa} and \texttt{r/Kenya}) into their regional counterparts (\texttt{r/canada} and \texttt{r/southafrica}, respectively).

\subsubsection{Feature Engineering}

    In addition to the data cleaning procedures I adopted to prepare the Reddit data for the text classification models, I applied further cleaning steps in the development of my embedding models. Firstly, I removed all observations associated with a moderator and any author with the strings spam' or bot' to minimise the presence of machine-generated text. Secondly, I removed all non-Latin script characters (excluding diacritics), \acrshortpl{URL}, and strings beginning with `u/' or `r/'. Lastly, I added the 161,796 New Zealand-specific place names to the named entity recogniser pipeline in \texttt{spaCy} for the New Zealand city-level communities only. I also split observations longer than 500 words into 500-word chunks. After applying these procedures, the final training dataset for the New Zealand city-level communities included 201 million words. For the country-level communities, I followed the same cleaning procedures but used only the default entities from \texttt{spaCy} in the entity recognition pipeline. The final training datasets for the inner-circle and outer-circle country-level communities included 500 million words each.

\subsubsection{Model Development}

    For the New Zealand city-level communities, I train two embedding models using the two prevailing architectures: continuous bag-of-words (\acrshort{CBOW}) and skip-gram with negative sampling (\acrshort{SGNS}). For the country-level communities, I use only the \acrshort{CBOW} model. In terms of hyperparameters, I set the number of dimensions to 300 and the search window to 5. Additionally, I set the minimum word frequency to 5 to exclude low-frequency words.
    
\subsubsection{Evaluation}

    Unlike the text classification models, there are no clear metrics (such as the $F_1$-score) to determine the performance of embedding models. One approach to evaluating embedding models is to test human-perceived similarity \citep{wang_evaluating_2019}. This can be achieved by producing a list of similar and related words.Due to the widespread usage of hypocoristics in \acrshort{NZE} \citep{bardsley_hypocoristics_2009}, I evaluate the performance of my embedding models using this class of lexical features. Other features, such as borrowings from te reo Māori, were also considered but were not included due to the high likelihood that these words would be out of vocabulary for other place-based communities, as well as for the \acrshort{GOO} model \citep{mikolov_efficient_2013}.
    
    I created a list of word pairs consisting of 59 hypocoristics associated with \acrshort{NZE}, derived from 148 open-source Wiktionary entries and their semantic equivalents (such as \textit{tradie} and \textit{tradesman}). As these word pairs are conceptually related in \acrshort{NZE}, the distance between them should be smaller in a \acrshort{NZE}-specific embedding model. Inversely, if the relationship was not present in the training data, the distance between two conceptually distant words would be greater. Additionally, I include a list of homonymous words that are conceptually related and distinct to \acrshort{NZE}. These include: \textit{bach}, \textit{banger}, \textit{beehive}, \textit{bog}, \textit{dairy}, \textit{fizzy}, \textit{raro}, \textit{snag}, \textit{tramp}, and \textit{yarn}.

    From the 51 user-informed lexical variables, I derived a set of 65 word pairs as an additional evaluation set. If the user intuitions were accurate, I would expect the cosine similarity between the conservative and innovative variants to increase if there were indeed a change in progress. With reference to my findings from Chapter \ref{chap:user_variables}: Sociolinguistic Variables and Lexical Variants, I do not expect this list of 65 word pairs to be a useful indicator of model performance. However, this list may shed further light on the effectiveness of perceptual data for evaluative purposes. Using the \texttt{similarity()} method in \texttt{gensim}, I compute the cosine similarity between the word pairs and take the mean as a measure of model performance.

    \begin{table}
        \scriptsize
        \centering
            
            \caption{Hypocoristic Word-Pairs for Evaluation}
            \label{tab:hypo_wordpairs}
            
            \renewcommand{\arraystretch}{1.4}
            \begin{tabularx}{\textwidth}{*{2}{>{\centering\arraybackslash}X}|*{2}{>{\centering\arraybackslash}X}}
            \toprule
            Hypocoristic & Full-form Equivalent & Hypocoristic & Full-form Equivalent \\
            \midrule
                \addlinespace[1em]
                
                    aggro & angry & rager & party \\
                    agro & angry & rego & registration \\
                    arvo & afternoon & rigger & beer \\
                    banger & sausage & righto & okay \\
                    bizzo & business & scarfie & university\_otago \\
                    bludger & unemployed & scratchie & lotto \\
                    bottler & good & servo & petrol\_station \\
                    brownie & slice & sharpie & bird \\
                    chippy & potato\_chip & slipper & spank \\
                    cobbler & sheep & smoko & break \\
                    
                    cocky & cockatoo & sneaker & shoes \\
                    cuzzy & cousin & snapper & fish \\
                    damper & bread & sparky & electrician \\
                    dunger & car & spinner & coin \\
                    durrie & cigarette & stubbie & beer \\
                    fizzy & soda & swagger & traveller \\
                    flatter & roommate & tinnie & marijuana \\
                    greenie & environmentalist & togs & swimming \\
                    hospo & hospitality & townie & city \\
                    joker & man & trackie & pants \\

                    kiddo & child & tradie & tradesman \\
                    kindy & kindergarten & tramper & hiker \\
                    lippy & lipstick & trundler & trolley \\
                    maccas & mcdonalds & truckie & driver \\
                    mozzie & mosquito & turps & paint \\
                    nappy & diaper & vollie & volunteer \\
                    offsider & rugby & wellies & gumboot \\
                    packer & insulation & welly & wellington \\
                    pokie & poker & westie & auckland \\
                    pommy & brit & - & - \\
                    
                \addlinespace[1em]
            \bottomrule
            \end{tabularx}

            \tablenoteparagraph{\textbf{Table Note}: This table lists the 59 hypocoristic word pairs used to evaluate the Word2Vec embedding models, comprising the hypocoristic form from Wiktionary and its full-form equivalent. Out-of-vocabulary hypocoristics were excluded; in cases where the full-form equivalent was out of vocabulary, either concept-related or word synonym strategies were applied to ensure a suitable comparison. Data is sourced from Wiktionary.com.}
        
    \end{table}

\subsubsection{Hypotheses and Predictions}

    If there was a relationship between \acrshort{NZE}, as a geographic dialect community, and the New Zealand city-level communities, then I would expect to observe high mean cosine similarity for the 59 hypocoristic word pairs and the ten conceptually related homonymous word pairs. This would likely be followed by the Word2Vec embedding models trained on the inner-circle country-level communities, due to the presence of \texttt{r/newzealand} in the training data. The next in rank would be the Word2Vec embedding models trained on the outer-circle country-level communities due to register similarities, and finally, \acrshort{GOO}. In terms of model architectures, I predict that the \acrshort{SGNS} model will show greater cosine similarity, as \acrshort{SGNS} typically performs better on smaller datasets. After evaluating the performance of the Word2Vec embedding models, I will assess the user-informed semantic variables to test the intuitions of the users in \texttt{r/newzealand}.

\subsection{Results}

    I present the performance metrics for the candidate Word2Vec embedding models in Table \ref{tab:metrics_Word2Vec}. Beginning with the 59 hypocoristic word pairs from the Wiktionary entries, the mean cosine similarity scores for the skip-gram with negative sampling (\acrshort{SGNS}) architecture were higher across the board than those for the continuous bag-of-words (\acrshort{CBOW}) architecture. As predicted, the model trained on the New Zealand city-level communities (\acrshort{NZM}) using \acrshort{SGNS} had the highest mean cosine similarity at 0.512. This was followed by the embedding models trained on the inner-circle (\acrshort{ICM}) and outer-circle (\acrshort{OCM}) country-level communities, and lastly, the baseline \acrshort{GOO}. Excluding \acrshort{GOO}, the \acrshort{OCM} embedding model with the \acrshort{SGNS} architecture achieved a mean cosine similarity higher than the \acrshort{NZM} embedding model using the \acrshort{CBOW} architecture.

    \begin{table}
        \scriptsize
        \centering
            
            \caption{Performance Metrics for Word2Vec Embedding Models}
            \label{tab:metrics_Word2Vec}
            
            \renewcommand{\arraystretch}{1.4}
            \begin{tabularx}{\textwidth}{l*{6}{>{\centering\arraybackslash}X}c}
                \toprule
                \multirow{2}{*}{Model} & \multicolumn{3}{c}{Wiktionary Entries} & \multicolumn{3}{c}{User-Informed Variables} & \multirow{2}{*}{Vocabulary} \\
                & \acrshort{CBOW} & \acrshort{SGNS} & \texttt{OOV} & \acrshort{CBOW} & \acrshort{SGNS} & \texttt{OOV} & \\
                \midrule
                \addlinespace[1em]
                \acrshort{NZM} & 0.406 & 0.512 & - & 0.620 & 0.595 & - & 60,814 \\
                \acrshort{ICM} & 0.391 & 0.471 & 8 & 0.626 & 0.586 & - & 102,561\\
                \acrshort{OCM} & 0.294 & 0.445 & 36 & 0.584 & 0.578 & 1 & 154,708 \\
                \acrshort{GOO} & - & 0.325 & 10 & - & 0.504 & 3 & 3,000,000 \\
                \addlinespace[1em]
                \bottomrule
            \end{tabularx}

            \tablenoteparagraph{\textbf{Table Note}: This table evaluates Word2Vec embedding model performance based on 59 hypocoristic word pairs from Wiktionary entries and 65 user-informed lexical variable word pairs. Results indicate that the best performing model was trained on the New Zealand city-level communities dataset using the \acrshort{SGNS} architecture. Columns detail the training data source (\texttt{Model}), mean cosine similarity scores for both continuous bag-of-words (\acrshort{CBOW}) and skip-gram with negative sampling (\acrshort{SGNS}) architectures, and total vocabulary size; all social media data is sourced from Reddit/Pushshift \citep{baumgartner_pushshift_2020}.}
        
    \end{table}

    As for the 65 user-informed variable word pairs, the \acrshort{ICM} embedding model trained on the \acrshort{CBOW} architecture had the highest mean cosine similarity at 0.626. For comparison, the \acrshort{NZM} \acrshort{CBOW} model had a mean cosine similarity of 0.620. This does not necessarily indicate that the \acrshort{CBOW} architecture offered better performance; instead, the likely explanation for this unexpected result was that the user-informed variable word pairs were not fit-for-purpose for evaluating the Word2Vec embedding models, as these lexical variables are not specific to \acrshort{NZE}. Meanwhile, the hypocoristic word pairs are more meaningful, as they are a feature distinct to Australian and \acrshort{NZE}es.
    
    Of the five embedding models (including the \acrshort{CBOW} and \acrshort{SGNS} variants), the New Zealand city-level communities model was the smallest, with a vocabulary of 60,814 words and phrases - approximately 2\% of the \acrshort{GOO} embedding model. This was expected, as the embedding models for the New Zealand city-level communities were trained on 201 million words. By way of contrast, the \acrshort{ICM} embedding model had a vocabulary of 102,561 words and phrases, and the \acrshort{OCM} embedding model had a vocabulary of 154,708 words and phrases, both having been trained on corpora of 500 million words. Surprisingly, both \acrshort{OCM} and \acrshort{GOO} had the most instances of out-of-vocabulary (\texttt{OOV}) tokens for the 59 hypocoristic word pairs. This offered further evidence that \acrshort{NZM} better reflected the linguistic context of New Zealand.

    As an additional test, I compared the cosine similarity measures across the candidate embedding models on ten \acrshort{NZE} words that were semantically ambiguous in other varieties of English. I present the results of these semantically ambiguous words and their synonyms in Table \ref{tab:ambiguous_words}. The results from this small set of word pairs were consistent with my earlier findings. Using cosine similarity as a measure of semantic similarity within the word vector space, the \acrshort{NZM} model trained using the \acrshort{SGNS} architecture was able to determine the relationship between the ambiguous words and their synonyms. As a final check, I trained a text classification model with the \acrshort{NZM} embeddings, and the classification model performed poorly across all configurations. The results are presented in Table \ref{tab:f1_city-embedding}.

    \begin{table}
        \scriptsize
        \centering
            
            \caption{Performance Metrics for Word2Vec Embedding Models}
            \label{tab:ambiguous_words}
            
            \renewcommand{\arraystretch}{1.4}
            \begin{tabularx}{\textwidth}{ll*{6}{>{\centering\arraybackslash}X}c}
                \toprule
                \multirow{2}{*}{Source} & \multirow{2}{*}{Target} & \multicolumn{2}{c}{\acrshort{NZM}} & \multicolumn{2}{c}{\acrshort{ICM}} & \multicolumn{2}{c}{\acrshort{OCM}} & \multirow{2}{*}{\acrshort{GOO}} \\
                & & \acrshort{CBOW} & \acrshort{SGNS} & \acrshort{CBOW} & \acrshort{SGNS} & \acrshort{CBOW} & \acrshort{SGNS} & \\
                \midrule
                \addlinespace[1em]
                bach & crib & 0.359 & \textbf{0.444} & 0.232 & 0.306 & 0.213 & 0.327 & 0.107 \\
                banger & sausage & 0.480 & 0.401 & \textbf{0.497} & 0.425 & 0.384 & 0.473 & 0.211 \\
                beehive & government & 0.171 & 0.369 & 0.012 & 0.362 & -0.095 & \textbf{0.397} & 0.056 \\
                bog & toilet & 0.246 & 0.278 & \textbf{0.312} & 0.226 & 0.061 & 0.255 & 0.179 \\
                dairy & superette & 0.515 & \textbf{0.516} & - & - & - & - & 0.256 \\
                fizzy & soda & \textbf{0.834} & 0.821 & 0.688 & 0.648 & 0.474 & 0.650 & 0.432 \\
                raro & cordial & 0.628 & \textbf{0.638} & 0.413 & 0.496 & - & - & - \\
                snag & sausage & 0.450 & 0.332 & \textbf{0.598} & 0.463 & 0.194 & 0.193 & 0.063 \\
                tramp & hike & \textbf{0.647} & 0.556 & 0.475 & 0.453 & 0.158 & 0.360 & 0.150 \\
                yarn & conversation & 0.334 & \textbf{0.387} & 0.117 & 0.358 & -0.099 & 0.133 & 0.086 \\
                \addlinespace[1em]
                \bottomrule
            \end{tabularx}

            \tablenoteparagraph{\textbf{Table Note}: This table compares cosine similarity measures for ambiguous words in \acrshort{NZE} and their respective synonyms across four datasets and two model architectures. The training data includes New Zealand city-level communities (\acrshort{NZM}), inner-circle (\acrshort{ICM}) and outer-circle (\acrshort{OCM}) country-level communities, and the Google News dataset (\acrshort{GOO}). Similarity measures are provided for both continuous bag-of-words (\acrshort{CBOW}) and skip-gram with negative sampling (\acrshort{SGNS}) architectures; columns detail the ambiguous \texttt{Source} word and its \texttt{Target} synonym, with all data sourced from Reddit/Pushshift \citep{baumgartner_pushshift_2020} and the \acrshort{GOO} corpus \citep{mikolov_efficient_2013}.}
        
    \end{table}

    \begin{table*}
            
            \caption{Performance Metrics for City-level Embedding Classification Models}
            \label{tab:f1_city-embedding}
            
            \scriptsize
            \centering
            \renewcommand{\arraystretch}{1.4}
                \begin{tabularx}{\textwidth}{l
                *{3}{>{\centering\arraybackslash}X} |
                *{3}{>{\centering\arraybackslash}X}}
                    \toprule
                    \multirow{2}{*}{Community} & \multicolumn{3}{c}{\gls{rstext}} & \multicolumn{3}{c}{\gls{rcomm}}\\
                    & \acrshort{CBOW} & \acrshort{SGNS} & Support & \acrshort{CBOW} & \acrshort{SGNS} & Support \\
                    \midrule
                    \addlinespace[1em]
                    \texttt{r/auckland} & - & - & 44 & 0.25 & 0.26 & 382 \\
                    \texttt{r/Tauranga} & 0.29 & 0.26 & 45 & - & 0.12 & 382 \\
                    \texttt{r/thetron} & - & 0.04 & 44 & 0.02 & 0.31 & 382 \\
                    \texttt{r/Wellington} & - & 0.04 & 45 & 0.19 & - & 382 \\
                    \texttt{r/chch} & - & 0.21 & 44 & 0.24 & 0.25 & 382 \\
                    \texttt{r/dunedin} & - & 0.04 & 44 & - & 0.08 & 382 \\
                    \addlinespace[1em]
                    Macro Average & 0.05 & 0.10 & 266 & 0.12 & 0.17 & 2,294 \\
                    Weighted Average & 0.05 & 0.10 & 266 & 0.12 & 0.17 & 2,294 \\
                    \addlinespace[1em]
                    \bottomrule
                \end{tabularx}
    
            \tablenoteparagraph{\textbf{Table Note}: This table presents model performance metrics for submission selftexts (\gls{rstext}) and comments (\gls{rcomm}) across New Zealand city-level communities, using a shallow Linear \acrshort{SVM} classifier with balanced sampling. Input features were derived by averaging the \acrshort{NZM} word vector representations, with observations containing fewer than 500 words excluded. Columns detail the community name, $F_1$-scores for models trained using continuous bag-of-words (\acrshort{CBOW}) and skip-gram with negative sampling (\acrshort{SGNS}) architectures, and the number of observations in the held-out test set (\texttt{Support}); the low model performance observed across both text-types was anticipated. All data is sourced from Reddit/Pushshift \citep{baumgartner_pushshift_2020}.}
        
    \end{table*}

\subsubsection{Cosine Similarity}
\label{dialect:cosine_similarity}

    With the Word2Vec models, I can now evaluate the user-informed semantic variables identified in Chapter \ref{chap:user_intuitions}: User Intuitions and Place Identity. I restrict my analysis to the Word2Vec embedding models trained with the \acrshort{SGNS} architecture. I present the results from my semantic distance analysis in Tables \ref{tab:semantic_embeddings1} and \ref{tab:semantic_embeddings2}. I have grouped the results based on the greatest distance between two Word2Vec embedding models. Specific to the \acrshort{NZE} context, Group 1 includes target variants where the greatest distance was between \acrshort{NZM} and \acrshort{OCM}. I observed this pattern in three user-informed semantic variables: \textsc{pudding}, \textsc{rubber}, and \textsc{twink}. In all three instances, cosine similarity was greatest between the conservative variant in \acrshort{NZM} and the innovative variant in \acrshort{OCM}.

    Beyond the \acrshort{NZE} context, I observed significant contrast across the remaining models, which I have categorised into Group 2, Group 3, and Group 4. Group 2 included target variants where the greatest distance was between \acrshort{ICM} and \acrshort{OCM}. I observed this pattern in four user-informed semantic variables: \textsc{fag}, \textsc{tea}, \textsc{tramp}, and \textsc{tuna}. With the exception of \textsc{tuna}, cosine similarity was greatest between the conservative variant in \acrshort{ICM} and the innovative variant in \acrshort{OCM}. Group 3 included target variants where the greatest distance was between \acrshort{ICM} and \acrshort{GOO}. I observed this pattern in two user-informed variables: \textsc{football} and \textsc{gay}. For \textsc{football}, cosine similarity was greatest between the innovative variant in \acrshort{ICM} and the conservative variant in \acrshort{GOO}. I observed the inverse for \textsc{gay}.
    
    My final distributional pattern, Group 5, included target variants where the greatest distance was within a single model. I observed this pattern in the user-informed semantic variables \textsc{flannel}, \textsc{bum}, and \textsc{dick}. This suggests that these user-informed semantic variables were more or less polysemous, with both the conservative and innovative variants co-existing within these word vector representations.

    \begin{table}
        \scriptsize
        \centering
            
            \caption{Semantic Distance in Word Embedding Models (Groupings)}
            \label{tab:semantic_embeddings1}
            
            \renewcommand{\arraystretch}{1.4}
            \begin{tabularx}{\textwidth}{*{3}{>{\arraybackslash}X}*{4}{>{\centering\arraybackslash}X}}
            \toprule
            Variable & Source & Target & \acrshort{NZM} & \acrshort{ICM} & \acrshort{OCM} & \acrshort{GOO} \\
            \midrule
                \addlinespace[1em]
                \multicolumn{7}{c}{Group 1} \\
                \addlinespace[1em]
                \multirow[c]{2}{*}{\textsc{pudding}} & \multirow[c]{2}{*}{pudding}
                & \underline{dinner} & \textbf{0.543} & 0.460 & 0.493 & 0.353 \\
                & & custard & 0.837 & 0.786 & \textbf{0.854} & 0.581 \\
                \addlinespace[1em]
                \multirow[c]{2}{*}{\textsc{rubber}} & \multirow[c]{2}{*}{rubber}
                & \underline{pencil} & \textbf{0.513} & 0.398 & 0.417 & 0.224 \\
                & & sex & 0.087 & 0.042 & \textbf{0.190} & 0.087 \\
                \addlinespace[1em]
                \multirow[c]{2}{*}{\textsc{twink}} & \multirow[c]{2}{*}{twink}
                & \underline{pen} & \textbf{0.496} & 0.279 & 0.304 & 0.104 \\
                & & bear & 0.320 & 0.229 & \textbf{0.438} & 0.161 \\
                \addlinespace[2em]
                \multicolumn{7}{c}{Group 2} \\
                \addlinespace[1em]
                \multirow[c]{2}{*}{\textsc{fag}} & \multirow[c]{2}{*}{fag}
                & \underline{cigarette} & 0.391 & \textbf{0.534} & 0.294 & 0.335 \\
                & & person & 0.408 & 0.263 & \textbf{0.434} & 0.081 \\
                \addlinespace[1em]
                \multirow[c]{2}{*}{\textsc{tea}} & \multirow[c]{2}{*}{tea}
                & \underline{dinner} & 0.346 & \textbf{0.469} & 0.432 & 0.279 \\
                & & coffee & 0.501 & 0.587 & \textbf{0.604} & 0.564 \\
                \addlinespace[1em]
                \multirow[c]{2}{*}{\textsc{tramp}} & \multirow[c]{2}{*}{tramp}
                & \underline{mountain} & 0.509 & \textbf{0.516} & 0.382 & 0.256 \\
                & & street & 0.200 & 0.273 & \textbf{0.437} & 0.212 \\
                \addlinespace[1em]
                \multirow[c]{2}{*}{\textsc{tuna}} & \multirow[c]{2}{*}{tuna}
                & \underline{salmon} & 0.685 & 0.652 & \textbf{0.705} & 0.611 \\
                & & eel & 0.434 & \textbf{0.556} & 0.454 & 0.523 \\
                \addlinespace[2em]
                \bottomrule
            \end{tabularx}

            \tablenoteparagraph{\textbf{Table Note}: This table compares similarity measures across four Word2Vec models - New Zealand city-level communities (\acrshort{NZM}), inner-circle country-level communities (\acrshort{ICM}), outer-circle country-level communities (\acrshort{OCM}), and the Google News Dataset (\acrshort{GOO}) - for specific user-informed variables. The results are organised into two distributional patterns: Group 1, containing target variants where the greatest distance was observed between \acrshort{NZM} and \acrshort{OCM}, and Group 2, where the greatest distance occurred between \acrshort{ICM} and \acrshort{OCM}. Columns identify the variable category, source term, and target variants, while rows highlight user-informed conservative synonyms through underlining; all data is sourced from Reddit/Pushshift \citep{baumgartner_pushshift_2020} and the Google News Dataset \citep{mikolov_efficient_2013}.}
        
    \end{table}

    \begin{table}
        \scriptsize
        \centering
            
            \caption{Semantic Distance in Word Embedding Models (Groupings Continued)}
            \label{tab:semantic_embeddings2}
            
            \renewcommand{\arraystretch}{1.4}
            \begin{tabularx}{\textwidth}{*{3}{>{\arraybackslash}X}*{4}{>{\centering\arraybackslash}X}}
            \toprule
            Variable & Source & Target & \acrshort{NZM} & \acrshort{ICM} & \acrshort{OCM} & \acrshort{GOO} \\
            \midrule
                \addlinespace[1em]
                \multicolumn{7}{c}{Group 3} \\
                \addlinespace[1em]
                \multirow[c]{2}{*}{\textsc{football}} & \multirow[c]{2}{*}{football}
                & \underline{soccer} & 0.723 & 0.716 & 0.631 & \textbf{0.731} \\
                & & \underline{rugby} & 0.698 & \textbf{0.708} & 0.701 & 0.562 \\
                \addlinespace[1em]
                \multirow[c]{2}{*}{\textsc{gay}} & \multirow[c]{2}{*}{gay}
                & \underline{unhappy} & 0.231 & \textbf{0.260} & 0.241 & 0.195 \\
                & & lesbian & 0.617 & 0.676 & 0.648 & \textbf{0.812} \\
                \addlinespace[2em]
                \multicolumn{7}{c}{Group 4} \\
                \addlinespace[1em]
                \multirow[c]{2}{*}{\textsc{kiwi}} & \multirow[c]{2}{*}{kiwi}
                & \underline{eagle} & 0.167 & 0.243 & \textbf{0.299} & 0.196 \\
                & & apple & 0.132 & 0.181 & 0.276 & \textbf{0.306} \\
                \addlinespace[2em]
                \multicolumn{7}{c}{Group 5} \\
                \addlinespace[1em]
                \multirow[c]{2}{*}{\textsc{flannel}} & \multirow[c]{2}{*}{flannel}
                & \underline{towel} & 0.658 & \textbf{0.668} & 0.473 & 0.328 \\
                & & shirt & 0.479 & \textbf{0.515} & 0.493 & 0.418 \\
                \addlinespace[1em]
                \multirow[c]{2}{*}{\textsc{bum}} & \multirow[c]{2}{*}{bum}
                & \underline{smoke} & 0.312 & 0.292 & \textbf{0.323} & 0.133 \\
                & & sex & 0.165 & 0.161 & \textbf{0.195} & 0.151 \\
                \addlinespace[1em]
                \multirow[c]{2}{*}{\textsc{dick}} & \multirow[c]{2}{*}{dick}
                & \underline{richard} & 0.153 & 0.103 & 0.219 & \textbf{0.331} \\
                & & jerk & 0.513 & 0.492 & 0.474 & \textbf{0.550} \\
                \addlinespace[1em]
                \bottomrule
            \end{tabularx}

            \tablenoteparagraph{\textbf{Table Note}: This table compares similarity measures for specific user-informed variables across four Word2Vec representation models: New Zealand city-level communities (\acrshort{NZM}), inner-circle country-level communities (\acrshort{ICM}), outer-circle country-level communities (\acrshort{OCM}), and the Google News Dataset (\acrshort{GOO}). The results are categorised into three distributional patterns: Group 3, where the greatest distance exists between \acrshort{ICM} and \acrshort{GOO}; Group 4, where the greatest distance is between \acrshort{OCM} and \acrshort{GOO}; and Group 5, where the greatest distance occurs within a single model. Columns detail the variable category, the source term, and target variants, with conservative synonyms underlined; all data is sourced from Reddit/Pushshift \citep{baumgartner_pushshift_2020} and the Google News Dataset \citep{mikolov_efficient_2013}.}
        
    \end{table}

\subsubsection{Case Study: Tramp}
\label{dialect:tramp}

    So far, I have inferred semantic distance between Word2Vec embedding models as a measure of cosine similarity. As an interpretable measure of model performance, cosine measures alone offer little meaning in terms of geographic dialect alignment. I now focus on one user-informed semantic variable: \textsc{tramp}. In \acrshort{NZE}, \textit{tramp} either means a walking expedition (noun) or to walk long distances (verb) in back country or wilderness areas (see \textit{bush} and \textit{wop-wops}) \citep[p.296]{orsman_new_1994}. Simply put, \textit{tramp} is analogous to \textit{hike}. Outside the \acrshort{NZE} context, \textit{tramp} (noun) holds a pejorative meaning either as a vagrant or, more specifically in American English, a sexually promiscuous woman \citep{oxford_university_press_tramp_2025}. Therefore, this user-informed semantic variable offers a unique case study for both lexical and semantic variation across the country-level communities.
    
    With reference to the findings from Chapter \ref{chap:user_variables}: User-Informed Sociolinguistic Variables, I observed a degree of non-local bias in the distribution of the conservative variant (\textit{tramp}) and its innovative variant (\textit{hike}). When I measured the cosine similarity between \textit{tramp} and \textit{mountain} across the four Word2Vec embedding models, I observed a high cosine similarity between the source and conservative target in \acrshort{ICM}, \acrshort{NZM}, and \acrshort{OCM}, followed by the baseline \acrshort{GOO} embedding model. In order to confirm that \textit{tramp} in \acrshort{NZM} was in fact related to the conservative variant, I analysed the top 30 most similar words to \textit{tramp} across the Word2Vec embedding models. The \texttt{most\_similar()} method in Word2Vec provides a list of the nearest neighbours of a given input within the embedding space.

\begin{figure}
      \centering
      
            \includegraphics[height=0.6\textwidth]{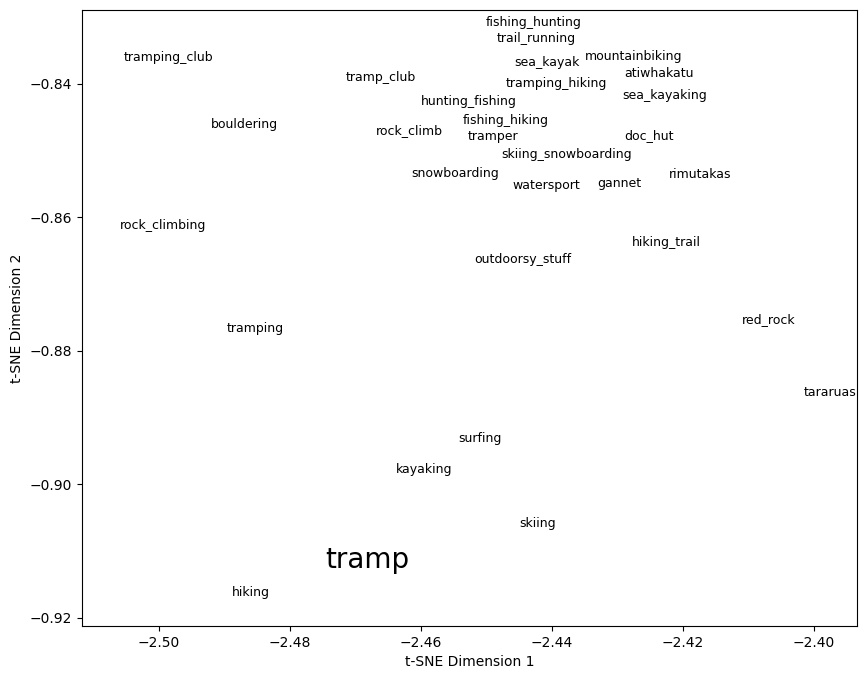}
            
      \vspace{6pt}
      \caption{Most Similar to \textit{tramp} in \acrshort{NZM}}
      \label{fig:tramp_nz-tsne}
      
      \vspace{6pt}
      \captionsetup{font=footnotesize, labelformat=empty, justification=justified, singlelinecheck=false}
      \caption*{\setstretch{2}\textbf{Description}: This visualisation presents the top 30 words most similar to \textit{tramp} by cosine similarity within the Word2Vec embedding model trained on New Zealand city-level communities (\acrshort{NZM}) using the skip-gram with negative sampling architecture. The terms most similar to \textit{tramp} are \textit{hiking}, \textit{kayaking}, \textit{skiing}, and \textit{surfing}; all data is sourced from Reddit/Pushshift \citep{baumgartner_pushshift_2020}.}
      
    \end{figure}

\begin{figure}
      \centering
      
            \includegraphics[height=0.6\textwidth]{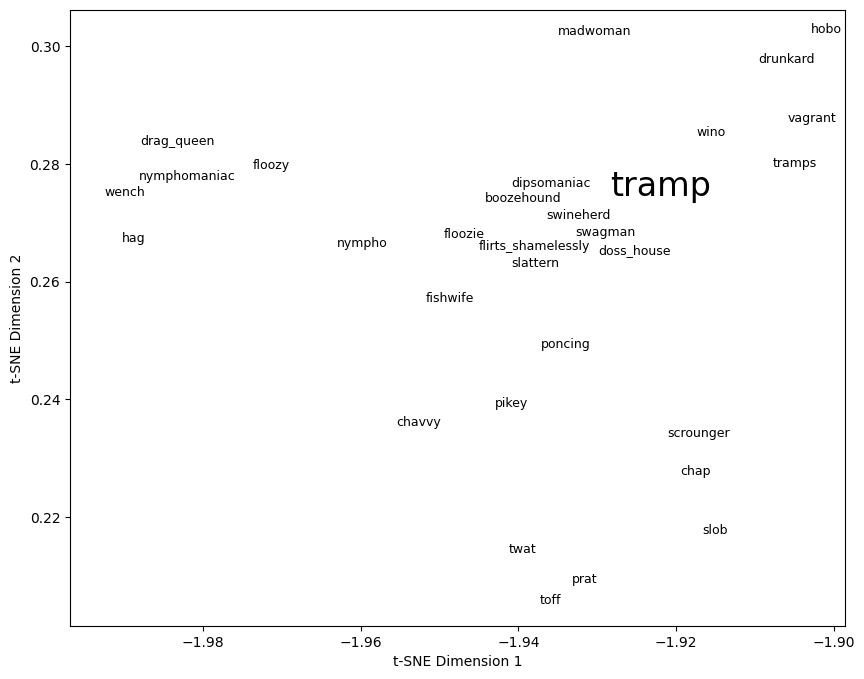}
            
      \vspace{6pt}
      \caption{Most Similar to \textit{tramp} in \acrshort{GOO}}
      \label{fig:tramp_g3-tsne}
      
      \vspace{6pt}
      \captionsetup{font=footnotesize, labelformat=empty, justification=justified, singlelinecheck=false}
      \caption*{\setstretch{2}\textbf{Description}: This visualisation presents the top 30 words most similar to \textit{tramp} by cosine similarity within the Word2Vec embedding model trained on the Google News dataset (\acrshort{GOO}; \citealp{mikolov_efficient_2013}). The highest-ranking terms include \textit{dipsomaniac}, \textit{boozehound}, \textit{swineherd}, \textit{swagman}, and \textit{doss\_house}; data is sourced from the \acrshort{GOO} corpus \citep{mikolov_efficient_2013}.}
      
    \end{figure}

\begin{figure}
      \centering
      
        \begin{subfigure}[t]{0.5\textwidth}
            \centering
            \includegraphics[width=\textwidth]{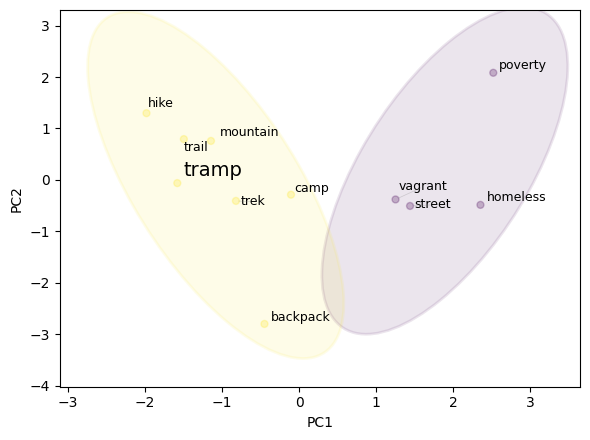}
            \subcaption{\acrshort{NZM}}
        \end{subfigure}%
        ~
        \begin{subfigure}[t]{0.5\textwidth}
            \centering
            \includegraphics[width=\textwidth]{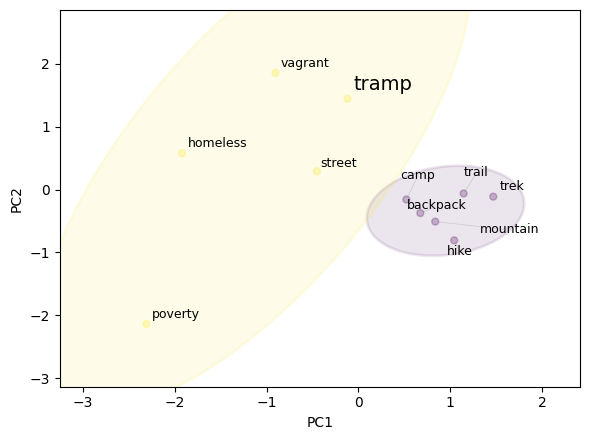}
            \subcaption{\acrshort{GOO}}
        \end{subfigure}
        
      \vspace{6pt}
      \caption{Clustering Analysis of \textit{tramp} across Embedding Models}
      \label{fig:tramp_pc}
      
      \vspace{6pt}
      \captionsetup{font=footnotesize, labelformat=empty, justification=justified, singlelinecheck=false}
      \caption*{\setstretch{2}\textbf{Description}: This figure displays subfigures comparing the \acrshort{NZM} and \acrshort{GOO} models for the term \textit{tramp}, utilising dimensionality reduction via \acrshort{PCA} and $k$-means clustering of semantically similar keywords. The analysis includes a conservative variant (\textit{mountain}, \textit{trek}, \textit{hike}, \textit{trail}), an innovative variant (\textit{poverty}, \textit{homeless}, \textit{street}), and neutral terms (\textit{camp}, \textit{backpack}, \textit{vagrant}). Results indicate that clusters in the \acrshort{NZM} model associate \textit{tramp} more closely with the conservative definition, whereas the \acrshort{GOO} model aligns the term more with the innovative definition; data is sourced from Reddit/Pushshift \citep{baumgartner_pushshift_2020} and the Google News Dataset \citep{mikolov_efficient_2013}.}
      
    \end{figure}

    I visualised the top 30 most similar words by cosine similarity for \textit{tramp} in \acrshort{NZM} and \acrshort{GOO} in Figures \ref{fig:tramp_nz-tsne} and \ref{fig:tramp_g3-tsne}. The top 30 nearest neighbours suggest that \textit{tramp} differs in meaning between \acrshort{NZM} and \acrshort{GOO}. To ensure that this variance was not only semantically meaningful but also statistically significant, I conducted $k$-means clustering on a selection of keywords.These keywords were related to the conservative variant (\textit{mountain}, \textit{trek}, \textit{hike}, \textit{trail}), the innovative variant (\textit{poverty}, \textit{homeless}, \textit{street}), and neutral words (\textit{camp}, \textit{backpack}, \textit{vagrant}). I used \acrshort{PCA} for dimensionality reduction. I found that \textit{tramp} in \acrshort{NZM} clustered with the keywords related to the conservative variant, while the opposite was true for \acrshort{GOO}. I visualised this cluster analysis in Figure \ref{fig:tramp_pc}.

    Finally, I examined the relationships of \textit{tramp} across the three candidate Word2Vec embedding models trained on the place-based communities and \acrshort{GOO} using the in-built analogical reasoning function. This approach is based on vector arithmetic, which I can use to interrogate the semantic relationships between words \citep{rogers_too_2017}. I present the top five features for each Word2Vec model in Table \ref{tab:tramp_analogy}. In brief, the results for both prompts found that the conservative variant of \textit{tramp} was restricted to \acrshort{NZM} and, to a lesser degree, \acrshort{ICM}. As for \acrshort{OCM} and \acrshort{GOO}, the relationships were more similar to the innovative variant of \textit{tramp}.

    \begin{table*}
        \scriptsize
        \centering
        \renewcommand{\arraystretch}{1.4}
            
            \caption{Analogical Reasoning Task}
            \label{tab:tramp_analogy}
        
                \begin{tabularx}{\textwidth}{*{4}{>{\centering\arraybackslash}X}}
                    \toprule
                    \acrshort{NZM} & \acrshort{ICM} & \acrshort{OCM} & \acrshort{GOO} \\
                    \midrule
                    \addlinespace[1em]
                    \multicolumn{4}{c}{\textbf{tramp} + \textbf{walk} - \textbf{hike}} \\
                    \addlinespace[1em]
                    jog & pour\_rain & doggy\_style & walking \\
                    polhill & unkempt & scream\_loudly & wander \\
                    crocs & walkin & semi\_naked & zimmer\_frame \\
                    mtb\_trail & toilet\_cubicle & scowl & walks \\
                    brisk\_walk & jogging & spit\_paan & poncing \\
                    \addlinespace[2em]
                    \multicolumn{4}{c}{\textbf{tramp} + \textbf{poverty} - \textbf{homeless}} \\
                    \addlinespace[1em]
                    skiing & skiing & pointi & dunghill \\
                    mountain\_biking & mountaineering & theire & abject\_poverty \\
                    hiking & snowboarding & stagnate\_decline & dung\_heaps \\
                    tramping & mountain\_biking & blatant\_sexism & prole \\
                    climbing & trekking & rasputin & serfdom \\
                    \addlinespace[1em]
                    \bottomrule
                \end{tabularx}
    
            \tablenoteparagraph{\textbf{Table Note}: This table presents the top features grouped by analogical reasoning prompts for four candidate models trained on varying datasets: New Zealand city-level communities (\acrshort{NZM}; 200 million words), inner-circle country-level communities (\acrshort{ICM}; 500 million words), outer-circle country-level communities (\acrshort{OCM}; 500 million words), and the Google News Dataset (\acrshort{GOO}; 100 billion words) \citep{mikolov_efficient_2013}. All social media data is sourced from Reddit/Pushshift \citep{baumgartner_pushshift_2020}.}
            
    \end{table*}

\section{Discussion}
\label{dialect:discussion}

    While I was able to distinguish country-level and city-level communities on Reddit using the text classification approach, the text classification models by themselves were not overly useful as dialect models. Instead, other confounds such as sampling bias and sample size had a greater impact on model performance \citep{figueroa_text_2013}. As predicted, balanced sampling improved model performance in the minority classes. This was particularly evident in the minority classes of \texttt{r/usa} and \texttt{r/Kenya}, where proportional and random sampling significantly impacted model performance. Removing named entities such as place names (\acrshort{GPE} and \acrshort{GPE}) using the \texttt{spaCy} pipeline resulted in a drop in performance across both inner-circle and outer-circle communities.

    As for the New Zealand city-level communities, I was able to distinguish the communities using the text classifiers. This result was unexpected, as there is limited linguistic variance within \acrshort{NZE} \citep{bauer_can_2002}. Similar to the country-level communities, I found that named entities were once again a significant predictor. These findings were consistent with existing research where named entities such as place-names, people, and organisations are over-represented on Twitter\textsuperscript{X} \citep{dunn_stability_2022}. I improved the interpretative performance of my classification models by injecting over 150,000 New Zealand-specific place-names and their orthographic variants into the entity recogniser during data cleaning.
    
    Encouragingly, these findings suggest that Reddit is not ``exceptional'' \citep[p.10]{panek_understanding_2022} in terms of its language use. This means I can transfer existing research from other social media platforms, such as Twitter\textsuperscript{X}, to the context of Reddit. Even though text classification does not offer meaningful insights into latent linguistic variation, it remains a low-cost yet effective diagnostic tool to examine corpus characteristics. I adapted the feature engineering and data cleaning procedures during the development of my Word2Vec embedding models, which in turn influenced my sampling and data cleaning protocols.
    
    Because the Word2Vec embedding models were trained on language data from my place-based communities, I was able to compare the vector representations between the New Zealand city-level communities (\acrshort{NZM}), the inner-circle (\acrshort{ICM}) and outer-circle (\acrshort{OCM}) country-level communities, and the baseline model trained on the Google News dataset (\acrshort{GOO}) \citep{mikolov_efficient_2013}. As a vector representation of words, cosine similarity does not necessarily imply linguistic variation and change. However, the findings from my case study on the user-informed semantic variable \textsc{tramp} suggest that the vector relationships indicate meaningful semantic variation across the place-based communities on Reddit.

\section{Chapter Summary}
\label{dialect:conclusion}

    While text classification was not meaningful as a dialect model for determining latent linguistic variation, it proved to be a useful diagnostic tool for examining the corpus characteristics of these communities. Specifically, it highlighted the significant role of named entities and sampling methods in distinguishing place-based datasets. On the other hand, the Word2Vec embedding models proved to be more fit-for-purpose for dialect modelling. By capturing the contextual relationships between words, these models provided a more nuanced view of lexical and semantic variance. In comparison to my earlier findings from Chapter \ref{chap:user_variables}: Sociolinguistic Variables and Lexical Variants, which took a traditional sociolinguistic variable approach, I now have evidence to suggest geographic dialect alignment between geographic dialect communities and place-based communities on Reddit. The embedding models successfully identified semantic nuances - such as the specific New Zealand usage of \textit{tramp} - that align with established geographic dialects. However, I am yet to fully address my goal of evaluating the user-informed semantic variables in their entirety. To understand whether these lexical features are truly undergoing the shifts suggested by user intuitions, I must examine these semantic relationships diachronically. This temporal analysis will be the focus of the following chapter.
    
% -----------------------------
% Chapter 7: Social Networks and Diachronic Embeddings
% -----------------------------

\chapter{Social Networks and Diachronic Embeddings}
\markboth{Social Networks and Diachronic Embeddings}{}
\label{chap:construction_grammar}

\section{Chapter Outline}
\label{c2xg:chapter_outline}

    This final results chapter builds upon the insights established in the preceding three chapters. I first address the critical issues of data availability and validity, before expanding the dialect models into diachronic embedding models. Following a brief background and motivation in Section \ref{c2xg:introduction}, Section \ref{c2xg:data} details the corpus dimensions for the New Zealand-related communities. I then utilise \acrshort{C2xG} to examine the grammatical and behavioural characteristics of these communities within a social network analysis (Section \ref{c2xg:social_network}). After an interim summary in Section \ref{c2xg:interim_summary}, I introduce the analytical pipeline and results for the diachronic embedding models, which are used to analyse change in the user-informed semantic variables (Section \ref{c2xg:diachronic_embeddings}). The chapter concludes with a synthesis of these findings in Section \ref{c2xg:discussion} and an outline of key results in Section \ref{c2xg:conclusion}. The insights generated here contribute directly to the resolution of \acrshort{SQ2} and \acrshort{SQ3}, specifically interrogating the relationship between geographic dialect communities and place-based digital networks.

\section{Background and Motivation }
\label{c2xg:introduction}

    \begin{figure}
      \centering
      
            \includegraphics[height=0.6\textwidth]{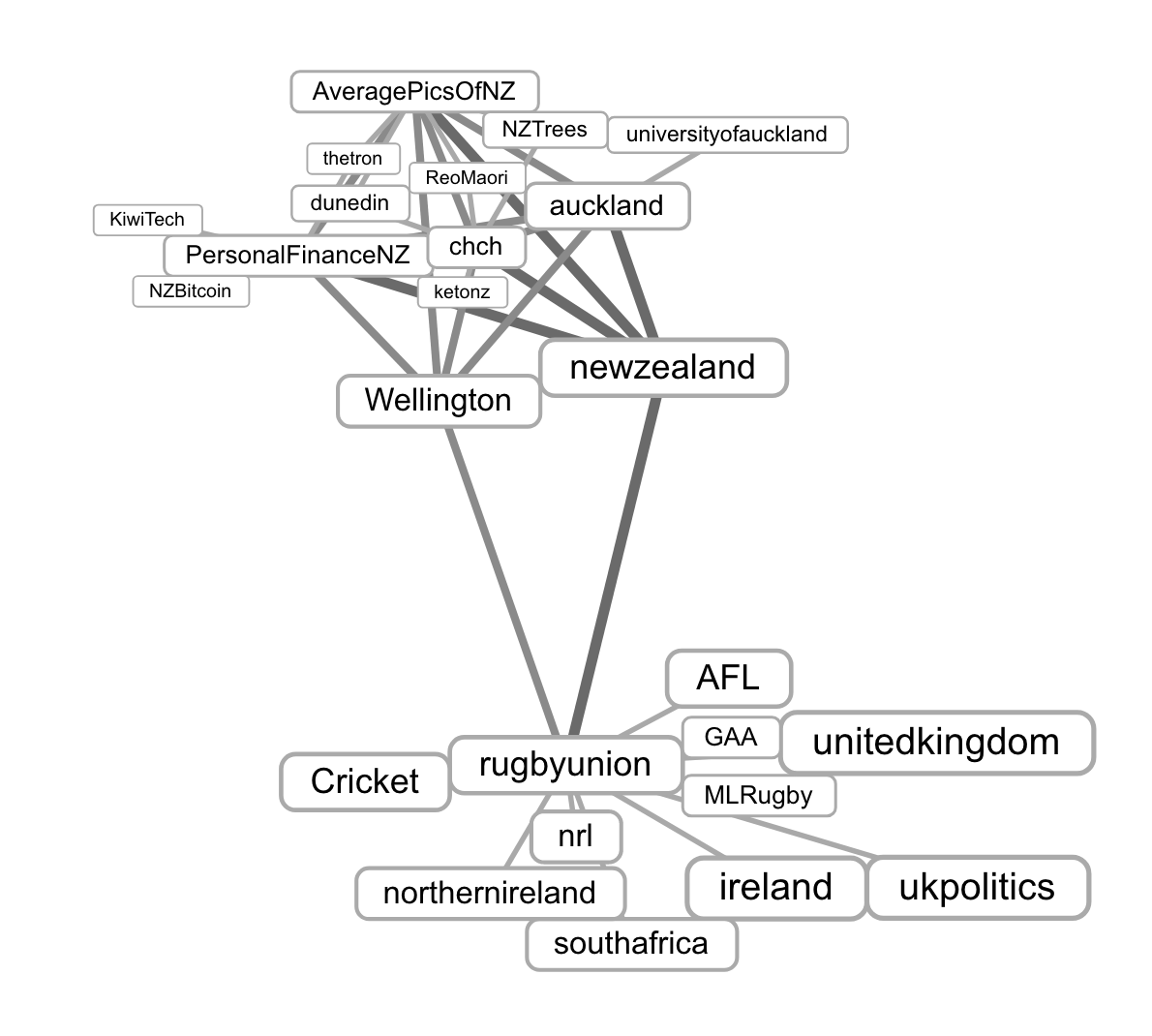}
            
      \vspace{6pt}
      \caption{Network Graph of Related Communities}
      \label{fig:nzreddit_network}
      
      \vspace{6pt}
      \captionsetup{font=footnotesize, labelformat=empty, justification=justified, singlelinecheck=false}
      \caption*{\setstretch{2}\textbf{Description}: This network graph visualises communities related to \texttt{r/newzealand} based on user engagement, originally developed by \texttt{u/avanka} to measure Jaccard similarity among users posting across multiple communities between August and September 2018. The visualisation demonstrates that users active on \texttt{r/newzealand} also engage with other place-based and special interest communities; data is sourced from \texttt{u/theworldisnoorange} via the SayIt platform (\href{https://anvaka.github.io/sayit/}{https://anvaka.github.io/sayit/}).}
      
    \end{figure}

    The results from Chapter \ref{chap:dialect_classification}: Dialect Modelling and Language Embeddings suggest that language embeddings provide an effective method for determining variation across place-based communities on Reddit. However, embedding models such as Word2Vec offer only a static representation of word vectors. To explore diachronic change, it is necessary to train diachronic embedding models that can capture semantic shifts over time \citep{kutuzov_diachronic_2018}.

    Before implementing these models, the issue of data availability must be addressed. Combining eligible observations from \texttt{r/newzealand} (530 million words) and the city-level communities (201 million words) provides a total corpus of 731 million words prior to data cleaning. To ensure these embeddings are truly representative, a model must incorporate both domain-specific and generalised language data \citep{roberts_assessing_2016}. Consequently, my objective is to expand the existing corpus beyond place-based communities by incorporating other relevant subreddits, while ensuring the language data remains distinctly characteristic of New Zealand.
    
    As illustrated in Figure \ref{fig:nzreddit_network}, the New Zealand Reddit ecosystem is not restricted to primary place-based communities like \texttt{r/newzealand} or city-level subreddits such as \texttt{r/auckland}, \texttt{r/Tauranga}, \texttt{r/thetron}, \texttt{r/Wellington}, \texttt{r/chch}, and \texttt{r/dunedin}. Rather, it exists as a complex network of communities. The goal of this chapter is therefore twofold: first, to determine the extent of this ecosystem and the role of place-related communities in developing a well-balanced language sample of New Zealand; and second, to utilise this expanded corpus to train a suite of diachronic embedding models to evaluate user-informed semantic variables.

\section{Data}
\label{c2xg:data}   

    I adopted a passive data collection approach by manually coding the selfposts and comments from \texttt{r/NZMetaHub} to establish a network of New Zealand-related communities. A significant advantage of utilising \texttt{r/NZMetaHub} as a starting point is that these communities are promoted by users with an active interest in New Zealand that extends beyond primary place-based affiliations. During the data collection process, I identified over 261 communities promoted as relevant to New Zealand. Of these, 69 ($n=23.8\%$) were either inactive or had been deleted. Ultimately, I located 32 of these communities within the Pushshift repository \citep{baumgartner_pushshift_2020}.
    
    Surprisingly, \texttt{r/Tauranga} - which was included in my city-level analyses - did not appear in \texttt{r/NZMetaHub}, nor was it featured in the list of promoted communities in \texttt{r/newzealand}. The final dataset comprised 33 New Zealand-related communities. Within this group, \texttt{r/newzealand} is the oldest community (created 23 March 2008), while \texttt{r/aucklandeats} is the most recent addition (created 3 August 2022). A summary of these communities is presented in Table \ref{tab:interestcommunity}, and the corpus dimensions are detailed in Table \ref{tab:reddit_ttr}.
    
    In addition to these 33 communities, I identified a further 14 subreddits associated with New Zealand in the Pushshift repository that were not promoted by users in \texttt{r/NZMetaHub}. One category of communities not featured in the typology provided by \citet{panek_understanding_2022} is the furtive or \acrfull{NSFW} category. This group features prominently among the top 40,000 Reddit communities, with 35.1\% ($n=13$) of the peripheral New Zealand-related communities falling into this category (see Table \ref{tab:interestcommunity}). While the primary purpose of these communities is to share content deemed not safe for work \citep{lagorio-chafkin_we_2018}, they frequently function as proxies for mobile dating applications, facilitating connections that transition from online interactions to offline encounters. Alongside these adult-content subreddits, \texttt{r/MedicalCannabisNZ} and \texttt{r/NZHauto} were also included as peripheral New Zealand-related communities.
    
    \begin{table}
        \scriptsize
        \centering
            
            \caption{Summary of New Zealand-related Communities}
            \label{tab:interestcommunity}
            
            \renewcommand{\arraystretch}{1.4}
                \begin{tabularx}{\textwidth}{l*{4}{>{\centering\arraybackslash}X}}
                \toprule
                \textbf{Community} & \textbf{Created} & \textbf{Members} & \textbf{Nickname} & \textbf{Rank} \\
                \midrule
                \addlinespace[1em]
                \texttt{r/Aleague} & Nov 7, 2010 & 185K &  & 1 \\
                \texttt{r/allblacks} & Jul 25, 2012 & 9.6K &  & 8 \\
                \texttt{r/antarctica} & Jul 15, 2010 & 37K &  & 4 \\
                \texttt{r/auckland} & Nov 22, 2009 & 220K &  & 1 \\
                \texttt{r/aucklandeats} & Aug 3, 2022 & 24K &  & 5 \\
                \texttt{r/AveragePicsOfNZ} & Jul 8, 2018 & 81K &  & 2 \\
                \texttt{r/CasualNZ} & Jan 5, 2018 & 11K &  & 7 \\
                \texttt{r/chch} & Feb 14, 2011 & 65K & Guests & 3 \\
                \texttt{r/ConservativeKiwi} & Apr 17, 2019 & 11K & Curious Chaps & 7 \\
                \texttt{r/Coronavirus\_NZ} & Mar 1, 2020 & 38K &  & 4 \\
                \texttt{r/diynz} & Feb 6, 2019 & 48K &  & 3 \\
                \texttt{r/DownUnderTV} & Oct 30, 2019 & 20K &  &  \\
                \texttt{r/dunedin} & Sep 6, 2010 & 35K &  & 4 \\
                \texttt{r/ImagesOfNewZealand} & Dec 6, 2015 & 712 &  & 24 \\
                \texttt{r/LegalAdviceNZ} & Dec 2, 2018 & 30K &  & 4 \\
                \texttt{r/MapsWithoutNZ} & Sep 16, 2015 & 121K &  & 2 \\
                \texttt{r/newzealand} & Mar 23, 2008 & 766K & kiwis & 1 \\
                \texttt{r/NewZealandWildlife} & Mar 24, 2019 & 56K &  & 3 \\
                \texttt{r/NZBitcoin} & Apr 6, 2013 & 23K &  & 5 \\
                \texttt{r/Nzcarfix} & Jul 2, 2024 & 7.1K &  & 9 \\
                \texttt{r/nzev} & Feb 5, 2020 & 8.6K &  & 8 \\
                \texttt{r/nzgardening} & Nov 5, 2015 & 17K &  & 6 \\
                \texttt{r/NZGoneWild} & Jan 27, 2012 & 160K &  &  \\
                \texttt{r/Nzhookups} &  &  &  & \\
                \texttt{r/nzpolitics} & Dec 14, 2010 & 5.7K &  & 10 \\
                \texttt{r/NZTrees} & Oct 2, 2011 & 26K &  &  \\
                \texttt{r/PartyParrot} & Jun 9, 2016 & 359K &  & 1 \\
                \texttt{r/PersonalFinanceNZ} & Jul 6, 2015 & 116K &  & 2 \\
                \texttt{r/Sheep} & Jun 22, 2010 & 27K &  & 4 \\
                \texttt{r/thetron} & Feb 17, 2011 & 42K &  & 3 \\
                \texttt{r/universityofauckland} & Aug 4, 2011 & 31K &  & 4 \\
                \texttt{r/Wellington} & Sep 2, 2010 & 119K &  & 2 \\
                \addlinespace[1em]
            \bottomrule
            \end{tabularx}

            \tablenoteparagraph{\textbf{Table Note}: This table summarises the New Zealand-related peripheral communities identified within the top 40,000 Reddit communities of the Pushshift Dump that were promoted in \texttt{r/NZMetaHub}, excluding \texttt{r/Tauranga}. The columns provide the community name (\texttt{Community}), its creation date (\texttt{Created}), subscriber count as of October 2025 (\texttt{Members}), community nickname (\texttt{Nickname}), and its respective rank on Reddit (\texttt{Rank}); all data is sourced from Reddit/Pushshift \citep{baumgartner_pushshift_2020}.}
        
    \end{table}

    \begin{table}
        \scriptsize
        \centering
            
            \caption{Summary of New Zealand Peripheral Communities}
            \label{tab:subversive}
            
            \renewcommand{\arraystretch}{1.4}
            \begin{tabularx}{\textwidth}{l*{4}{>{\centering\arraybackslash}X}}
                \toprule
                \textbf{Community} & \textbf{Created} & \textbf{Members} & \textbf{Nickname} & \textbf{Rank} \\
                \midrule
                \addlinespace[1em]
                \texttt{r/Aucklandgonewild} & Sep 11, 2021 & 31K & - & 18,531 \\
                \texttt{r/aucklandhookups1} & Feb 27, 2022 & - & - & 25,429 \\
                \texttt{r/chchNSFW} & Oct 15, 2021 & 9.5K & - & 23,803 \\
                \texttt{r/kiwifreaks} & Dec 15, 2022 & - & - & 11,978 \\
                \texttt{r/MedicalCannabisNZ} &  &  &  &  \\
                \texttt{r/NaughtyNewZealand} & Mar 22, 2020 & 32K & - & 10,054 \\
                \texttt{r/NewZealandGirls} & Jun 25, 2021 & 101K & - & 12,333 \\
                \texttt{r/newzealandonlyfans} & May 16, 2022 & 22K & - & 27,975 \\
                \texttt{r/NZgays} & Jan 27, 2023 & 20K & NZgays Members & 10,413 \\
                \texttt{r/NZGirlsGW} & Apr 21, 2020 & 225K & New Zealanders & 4,412 \\
                \texttt{r/NZHauto} &  &  &  &  \\
                \texttt{r/NZMeetUps} & Feb 26, 2023 & 17K & - & 11,282 \\
                \texttt{r/nzmilfs} & Feb 17, 2022 & 17K & - & 28,817 \\
                \texttt{r/NZSwingers} & May 26, 2021 & 27K & Swingers & 11,160 \\
                \addlinespace[1em]
                \bottomrule
            \end{tabularx}

            \tablenoteparagraph{\textbf{Table Note}: This table provides a summary of the 14 New Zealand-related peripheral communities identified within the top 40,000 Reddit communities of the Pushshift Dump that were not promoted in \texttt{r/NZMetaHub}. The columns detail the community name (\texttt{Community}), creation date (\texttt{Created}), subscriber count as of October 2025 (\texttt{Members}), community nickname (\texttt{Nickname}), and Reddit rank (\texttt{Rank}); notably, 12 of these communities were classified as Adult Only, with the exceptions being \texttt{r/MedicalCannabisNZ} and \texttt{r/NZHauto}. All data is sourced from Reddit/Pushshift \citep{baumgartner_pushshift_2020}.}
        
    \end{table}

\section{Social Network Analysis}
\label{c2xg:social_network}

    The primary benefit of including observations from New Zealand-related communities is that it increases the training corpus by 1.01 billion words, supplementing the existing data from \texttt{r/newzealand} and the six city-level communities. Unlike the place-based communities, however, there is limited a priori evidence to confirm that these New Zealand-related communities are composed of users physically located in New Zealand.

    For example, \texttt{r/PartyParrot} was established as an \acrshort{AAF} community to promote Sirocco - the \gls{kakapo_mao} featured in the BBC programme \textit{Last Chance to See} (2009) - but has since become the global community for a popular Slack emoji. Similarly, \texttt{r/MapsWithoutNZ} is a meme-based \acrshort{AAF} community. As observed in Chapter \ref{chap:user_variables} (Figure \ref{fig:time_nz}), the activity profiles of these users differ significantly, and neither community offers an obvious linguistic connection to New Zealand.

\begin{figure}
      \centering
      
        \begin{subfigure}[b]{\textwidth}
            \centering
            \includegraphics[width=0.8\textwidth]{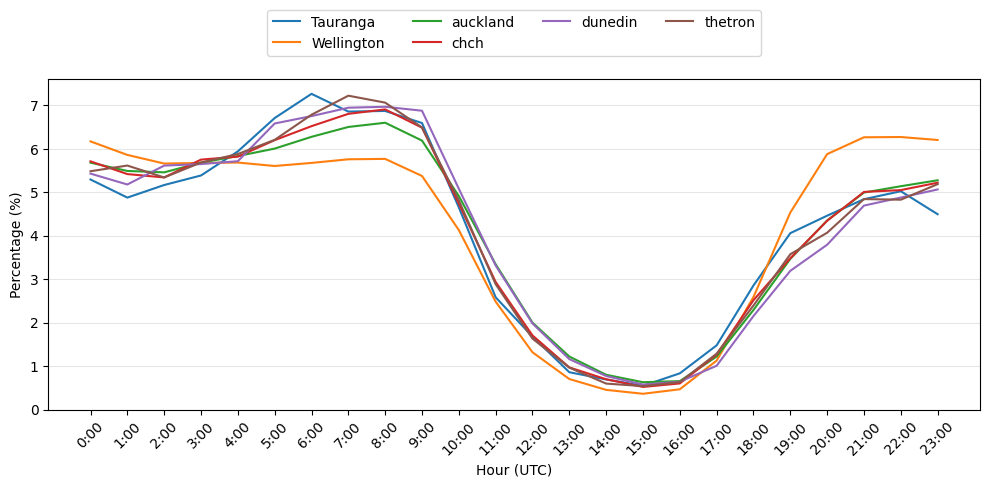}
            \vspace{6pt}
            \subcaption{Place-Based Communities}
        \end{subfigure}\vspace{12pt}
        \begin{subfigure}[b]{\textwidth}
            \centering
            \includegraphics[width=0.8\textwidth]{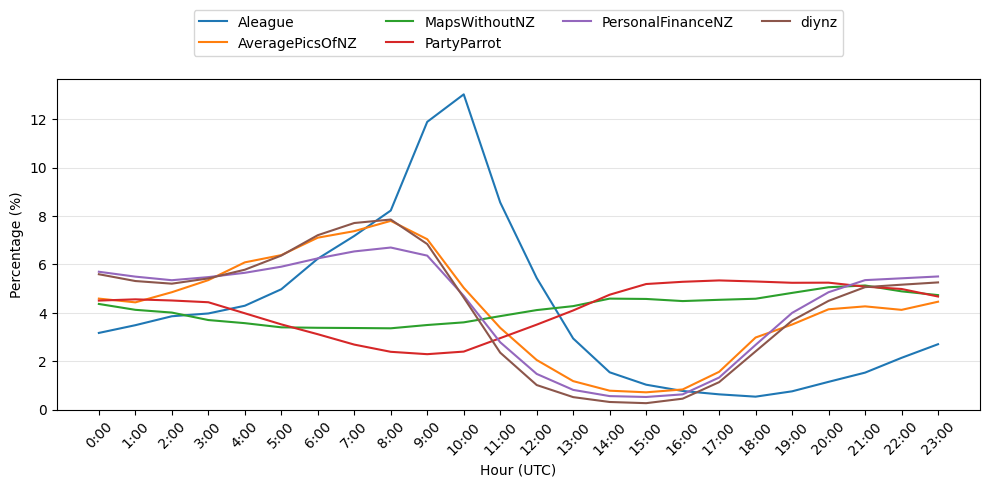}
            \vspace{6pt}
            \subcaption{Other Communities}
        \end{subfigure}

      \vspace{6pt}
      \caption{User Engagement by Hour in New Zealand-related Communities}
      \label{fig:time_city}
      
      \vspace{6pt}
      \captionsetup{font=footnotesize, labelformat=empty, justification=justified, singlelinecheck=false}
      \caption*{\setstretch{2}\textbf{Description}: This line plot displays user engagement based on comment timestamps in \acrshort{UTC} for (a) place-based communities and (b) other New Zealand-related communities. Consistent with \texttt{r/newzealand}, users in city-level communities adhere to \acrshort{NZST}, whereas users in other communities do not, suggesting a level of non-local activity; furthermore, the surge in comments within \texttt{r/PersonalFinanceNZ} at 10:00 \acrshort{UTC} indicates possible machine-generated activity. Data is sourced from Reddit/Pushshift \citep{baumgartner_pushshift_2020}.}
      
    \end{figure}

    To justify investigating the linguistic relationship between these communities, I visualised user engagement according to the time comments were posted (see Figure \ref{fig:time_city}). I included the six city-level communities alongside the most active non-place-based communities: \texttt{r/Aleague}, \texttt{r/MapsWithoutNZ}, \texttt{r/PersonalFinanceNZ}, \texttt{r/diynz}, \texttt{r/AveragePicsOfNZ}, and \texttt{r/PartyParrot}. As expected, the city-level communities adhered closely to \acrshort{NZST}, with a sharp drop in activity between midnight (12:00 \acrshort{UTC}) and early morning (18:00 \acrshort{UTC}). A similar adherence to \acrshort{NZST} was observable in \texttt{r/PersonalFinanceNZ}, \texttt{r/diynz}, and \texttt{r/AveragePicsOfNZ}, and to a lesser extent in \texttt{r/Aleague}. Conversely, \texttt{r/PartyParrot} and \texttt{r/MapsWithoutNZ} exhibited no such adherence to \acrshort{NZST}.

    Based on user activity alone, there is sufficient justification to interrogate the linguistic relationship of these communities. Finally, I replicated the analysis conducted by \texttt{u/avanka} (Figure \ref{fig:nzreddit_network}) to calculate the Jaccard distance between users in core and peripheral communities. I derived this distance based on the number of users who contributed either a selfpost (\gls{rstext}) or a comment (\gls{rcomm}) to two or more of the 47 identified communities.
    
\begin{figure}
      \centering
      
                \includegraphics[width=\textwidth]{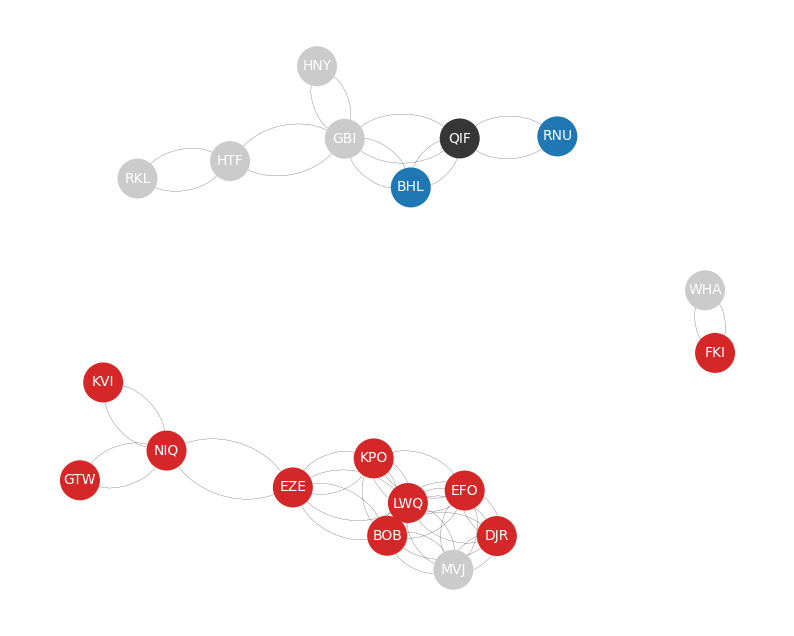}
                
      \caption{User Engagement in New Zealand-related Communities (\gls{rstext}}
      \label{fig:jaccard_rstext}
      
      \vspace{6pt}
      \captionsetup{font=footnotesize, labelformat=empty, justification=justified, singlelinecheck=false}
      \caption*{\setstretch{2}\textbf{Description}: This line plot displays user engagement based on comment timestamps in \acrshort{UTC} for (a) place-based communities and (b) other New Zealand-related communities. While users in city-level communities follow \acrshort{NZST} - consistent with \texttt{r/newzealand} - those in other communities do not, suggesting a level of non-local activity; notably, a surge in comments within \texttt{r/PersonalFinanceNZ} at 10:00 \acrshort{UTC} indicates potential machine-generated activity. Data is sourced from Reddit/Pushshift \citep{baumgartner_pushshift_2020}.}
      
    \end{figure}

\begin{figure}
      \centering
      
            \includegraphics[width=\textwidth]{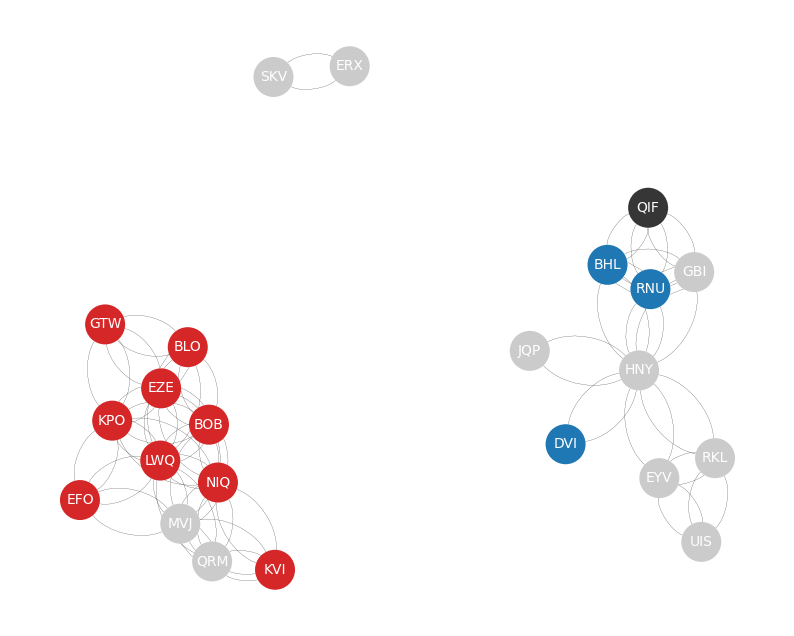}

      \vspace{6pt}
      \caption{User Engagement in New Zealand-related Communities (\texttt{rcomm)}}
      \label{fig:jaccard_rcomm}
      
      \vspace{6pt}
      \captionsetup{font=footnotesize, labelformat=empty, justification=justified, singlelinecheck=false}
      \caption*{\setstretch{2}\textbf{Description}: This visualisation illustrates the Jaccard distance (at 5\%) of users between communities within selfpost texts (\gls{rstext}), with nodes representing \texttt{r/newzealand} in black, city-level communities in blue, peripheral communities in red, and remaining communities in grey. The emergence of two primary clusters - one associated with place-based communities and the other with peripheral communities - suggests a limited degree of mutual engagement between users across the broader New Zealand-related network; all data is sourced from Reddit/Pushshift \citep{baumgartner_pushshift_2020}.}
      
    \end{figure}

    As shown in Figures \ref{fig:jaccard_rstext} and \ref{fig:jaccard_rcomm}, the clusters remain constant, with the emergence of two clusters connecting peripheral (largely Adult Only) communities and one major place-based cluster. It is evident that the peripheral communities form a user network largely distinct from the place-based ones, suggesting a limited degree of mutual engagement. Nevertheless, the linguistic role of New Zealand-related communities like \texttt{r/Aleague} remains unclear. These preliminary results provide a robust mandate to further interrogate the linguistic alignment of these New Zealand-related networks.

\subsection{Methodology}
\label{c2xg:methodology}

    As an intermediate stage prior to incorporating the New Zealand-related communities into the training of my diachronic Word2Vec embedding models, I investigate grammatical similarity at two distinct levels: user-level and community-level. At the user-level, I include all participants from the six city-level communities - \texttt{r/auckland}, \texttt{r/Tauranga}, \texttt{r/thetron}, \texttt{r/Wellington}, \texttt{r/chch}, and \texttt{r/dunedin}. This approach allows me to control for topic, ensuring that the observed linguistic features are not merely an artifact of specific subject matter. At the community-level, I consider all 33 New Zealand-related communities alongside the 14 peripheral communities that were excluded from \texttt{r/NZMetaHub}. The primary objective of this two-tiered investigation is to assess the suitability of integrating New Zealand-related communities into the development of my diachronic Word2Vec embedding models. By identifying shared grammatical and behavioural characteristics across these groups, I can justify their inclusion as part of a contiguous speech community, thereby providing a more robust and representative language sample of the New Zealand digital landscape.

\subsubsection{Computational Construction Grammar}

    \begin{table}
        \scriptsize
        \centering
        \renewcommand{\arraystretch}{1.4}
            
            \caption{Top Computational Construction Grammar Features}
            \label{tab:c2xg_constructions}
        
            \centering
            \renewcommand{\arraystretch}{1.4}

                \begin{tabularx}{\textwidth}{*{3}{>{\centering\arraybackslash}X}}
                    \toprule 
                    \texttt{LEX-Only} &
                    \texttt{SYN-Only} &
                    \texttt{SEM+} \\
                    \midrule
                    
                    \addlinespace[1em]
                    
                    {[ lex:\textit{of} -- lex:\textit{the} ]} & 
                    {[ syn:\texttt{PRP} -- syn:\texttt{DET} ]} & 
                    {[ syn:\texttt{PRP} -- sem:\textit{the} ]} \\
                    
                    {[ lex:\textit{in} < lex:\textit{the} ]} &
                    {[ syn:\texttt{MOD} > syn:\texttt{ADV} ]} &
                    {[ syn:\texttt{PRP} -- sem:\textit{the} ]} \\
                    
                    {[ syn:\texttt{MOD} > syn:\texttt{ADV} ]} &
                    {[ syn:\texttt{MOD} > syn:\texttt{ADV} ]} &
                    {[ syn:\texttt{PRP} -- lex:\textit{the} ]} \\
                    
                    {[ lex:\textit{if} < lex:\textit{you} ]} &
                    {[ syn:\textit{need} < syn:\textit{want} ]} &
                    {[ sem:\texttt{ADP} -- sem:\textit{the} ]} \\

                    {[ lex:\textit{to} > lex:\textit{be} ]} &
                    {[ syn:\textit{I} -- syn:\textit{think} ]} &
                    {[ sem:\texttt{ADP} -- lex:\textit{the} ]} \\
                    
                    {[ lex:\textit{on} < lex:\textit{the} ]} &
                    {[ syn:\textit{I} -- syn:\texttt{MOD} ]} &
                    {[ syn:\textit{need} < sem:\textit{want to} ]} \\
                    
                    \addlinespace[1em]
                    
                    \bottomrule
                \end{tabularx}
    
            \tablenoteparagraph{\textbf{Table Note}: This table details the top five features learned and parsed from New Zealand-related communities, categorised by lexical (\texttt{Lex-Only}), syntactic (\texttt{Syn-Only}), and combined (\texttt{SEM+}) feature sets; all data is sourced from Reddit/Pushshift \citep{baumgartner_pushshift_2020}.}
        
    \end{table}

    One limitation of the `bag-of-words' approach, as observed in previous text classification tasks, is that the relationships between words are severed. This effectively eliminates the latent linguistic variation that would otherwise be observable within the grammar. To evaluate the linguistic relationship between New Zealand-related communities more effectively, I adopt \acrshort{CxG} \citep{dunn_computational_2017} as my primary analytical tool. This allows me to account for morphosyntax, or the lexico-grammar, by treating lexicon and syntax as inseparable and interdependent components.
    
    \acrshort{CxG} is a usage-based approach to syntax \citep{goldberg_constructions_2006} that posits that a language's grammar is composed of a set of constructions (or \textit{constructicons}) - constraint-based representations that range in levels of abstractness \citep{dunn_computational_2017}. For example, the preference in \acrshort{NZE} for the construction \textit{in the weekend}, as opposed to the American \textit{on the weekend} or the traditional British \textit{at the weekend} \citep{trudgill_international_2017}, illustrates how specific lexical choices are bound to syntactic slots. In this framework, a grammar is not a list of rules but a network of slot-constraints. These constraints include lexical (item-specific), syntactic (form-based), and semantic (meaning-based) fillers. By analysing these multi-dimensional patterns, I can capture the nuanced alignment of digital communities that simpler models would overlook.
    
    I approximate grammatical distance by training a \acrshort{C2xG} model, where constructions serve as features and frequency of usage provides a metric for linguistic proximity \citep{dunn_computational_2017}. \acrshort{C2xG} is an approach to \acrshort{CxG} that integrates \acrshort{NLP} and unsupervised learning. I pooled the text observations from the 33 New Zealand-related communities to learn the specific constructional profile of New Zealand Reddit. While existing work in this field has largely focussed on Twitter/X and country-level varieties (\citealp{dunn_modeling_2019}; \citealp{dunn_stability_2022}; \citealp{dunn_language_2025}), my research shifts the focus to grammatical similarity, as I do not anticipate significant internal variance across these closely linked New Zealand-related communities.

    Regarding data cleaning, I removed observations that were deleted by the user or by \glspl{mod}. I then utilised the \acrshort{C2xG} Python package to learn a grammar based on the 33 New Zealand-related communities. Although the benchmark corpus size for training a \acrshort{C2xG} grammar is approximately 1 billion words, my training set of 744 million words proved sufficient for a robust model. After five rounds of training, the resulting grammar consisted of 1,582 distinct lexical constructions (\gls{lex}), 1,026 distinct syntactic constructions (\gls{syn}), and 8,284 combined semantic and structural constructions (\gls{sem}).
    
    Using these representations, I parsed the New Zealand-related communities at both the community and user levels. At the community-level, I parsed the features of each subreddit by text-type and month. At the user-level, I parsed features for individual users within the six city-level communities. I then employed cosine similarity as a measure of grammatical alignment between the New Zealand-related communities and the city-level behavioural groups.

    \begin{figure}
      \centering
      
        \begin{subfigure}[t]{0.5\textwidth}
            \centering
            \subcaption{}
            \includegraphics[width=\textwidth]{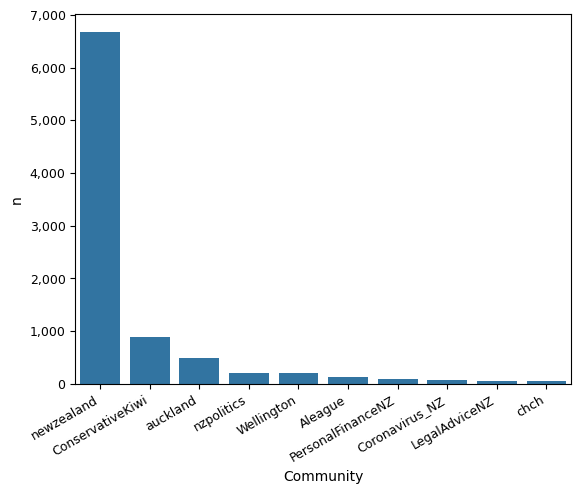}
        \end{subfigure}%
        ~
        \begin{subfigure}[t]{0.5\textwidth}
            \centering
            \subcaption{}
            \includegraphics[width=\textwidth]{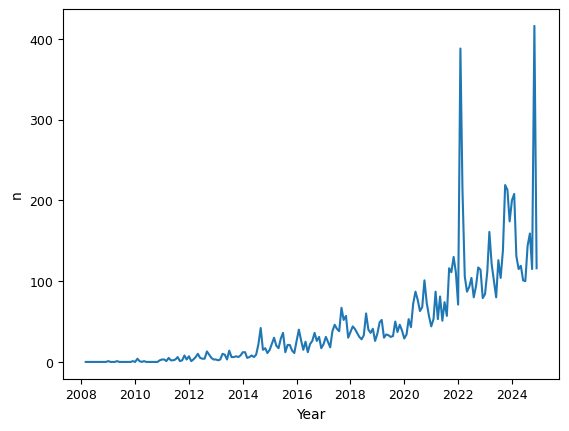}
        \end{subfigure}
        
      \vspace{6pt}
      \caption{Example of a Syntactic Construction}
      \label{fig:syn_feature}
      
      \vspace{6pt}
      \captionsetup{font=footnotesize, labelformat=empty, justification=justified, singlelinecheck=false}
      \caption*{\setstretch{2}\textbf{Description}: This barplot (a) and lineplot (b) illustrate the frequency and temporal distribution of the syntactic construction syn:\texttt{ADP} -- syn:\textit{the} -- syn:<200>, which captures New Zealand-related political discourse with a specific focus on Te Tiriti o Waitangi (the Treaty of Waitangi). Primarily sourced from \texttt{r/newzealand} and \texttt{r/ConservativeKiwi}, the construction exhibits a gradual increase alongside the growth of New Zealand-related Reddit communities, with significant surges observed during the Covid-19 pandemic in 2022, the 2023 government policy protests organised by Te Pāti Māori, and the late-2024 Hīkoi mō te Tiriti; associated tokens include \textit{over the treaty}, \textit{with the protest}, \textit{on the immigration}, and \textit{under the treaty}. Data is sourced from Reddit/Pushshift \citep{baumgartner_pushshift_2020}.}
      
    \end{figure}

    \begin{figure}
      \centering
      
        \begin{subfigure}[t]{0.5\textwidth}
            \centering
            \subcaption{}
            \includegraphics[width=\textwidth]{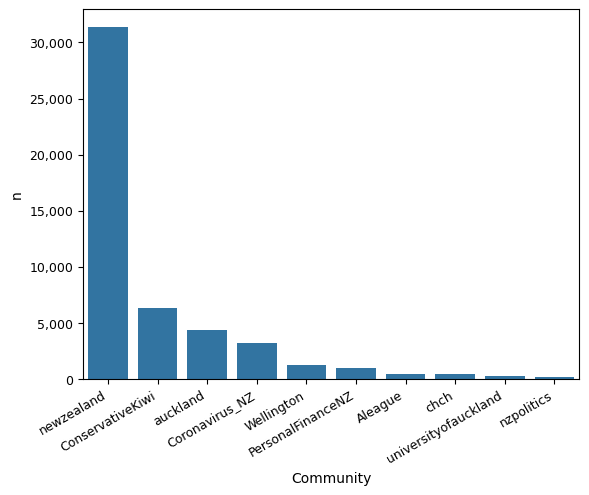}
        \end{subfigure}%
        ~
        \begin{subfigure}[t]{0.5\textwidth}
            \centering
            \subcaption{}
            \includegraphics[width=\textwidth]{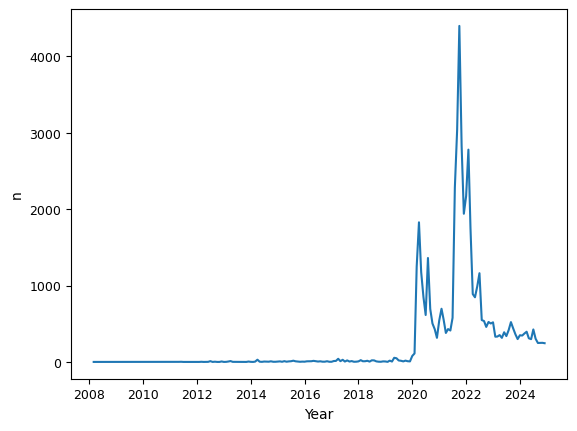}
        \end{subfigure}

      \vspace{6pt}
      \caption{Example of All (Combined) Construction}
      \label{fig:full_feature}
      
      \vspace{6pt}
      \captionsetup{font=footnotesize, labelformat=empty, justification=justified, singlelinecheck=false}
      \caption*{\setstretch{2}\textbf{Description}: This barplot (a) and lineplot (b) illustrate the combined frequency and temporal trajectory of the construction sem:\textit{the} > sem:<1089> related to the Covid-19 pandemic, where \acrshort{C2xG} captures both syntactic and semantic constraints. Primarily sourced from \texttt{r/newzealand} and \texttt{r/ConservativeKiwi}, the construction emerged in 2019 and reached peak usage in 2022; the associated semantic cluster includes tokens such as \textit{the lockdown}, \textit{the flu}, \textit{the miq}, \textit{the covid}, \textit{the pandemic}, \textit{the rona}, and various related forms. Data is sourced from Reddit/Pushshift \citep{baumgartner_pushshift_2020}.}
      
    \end{figure}

\subsubsection{User Behavioural Measures}

\begin{figure}
      \centering
      
            \includegraphics[width=0.6\textwidth]{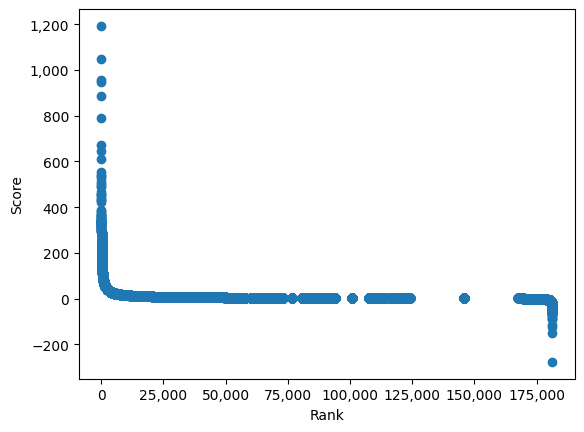}
            
      \vspace{6pt}
      \caption{Scatterplot of Users Ranked by Score}
      \label{fig:scatter_rank}
      
      \vspace{6pt}
      \captionsetup{font=footnotesize, labelformat=empty, justification=justified, singlelinecheck=false}
      \caption*{\setstretch{2}\textbf{Description}: This scatterplot displays the score and rank of users within city-level communities, sorted in descending order, revealing a noticeable Zipfian power law distribution among users with the highest scores; all data is sourced from Reddit/Pushshift \citep{baumgartner_pushshift_2020}.}
      
    \end{figure}

    In addition to grammatical similarity based on \acrshort{C2xG}, I examined user behaviour across the six city-level communities. This analysis included the duration since a user's first interaction, the total upvotes received (score), and the aggregate number of selfposts and comments. I identified 181,321 unique users who contributed to one or more of these communities. Initially, I visualised the total scores in descending order by rank (see Figure \ref{fig:scatter_rank}). The distribution exhibited a clear polarity: a small cohort of users possessed exceptionally high scores, while a few others sat at the lowest extreme (due to downvotes). The vast majority of users occupied the plateau between these two poles. To account for this non-normal distribution, I stratified users into deciles based on two primary behavioural measures: lifespan and engagement ratio.
    
\begin{figure}
      \centering
      
            \begin{subfigure}[b]{0.5\textwidth}
            \centering
            \includegraphics[width=\textwidth]{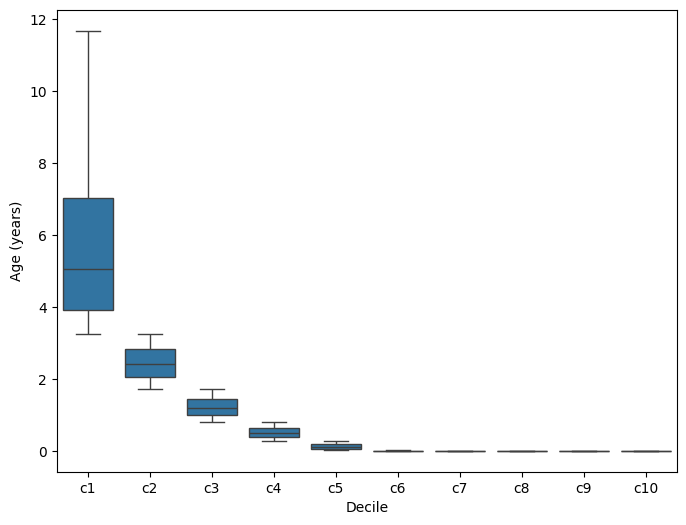}
            \subcaption{lifespan cohorts}
        \end{subfigure}%
        ~
        \begin{subfigure}[b]{0.5\textwidth}
            \centering
            \includegraphics[width=\textwidth]{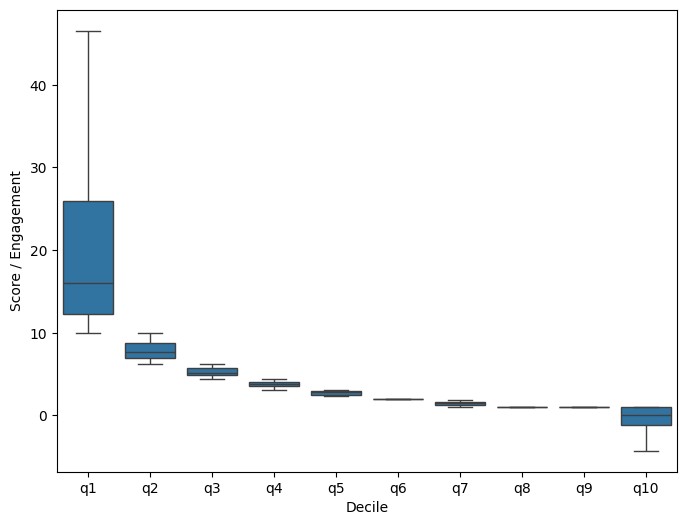}
            \subcaption{Engagement Ratio}
            \end{subfigure}
            
      \vspace{6pt}
      \caption{Boxplot of Behavioural Measures}
      \label{fig:boxplot_behaviour}
      
      \vspace{6pt}
      \captionsetup{font=footnotesize, labelformat=empty, justification=justified, singlelinecheck=false}
      \caption*{\setstretch{2}\textbf{Description}: This boxplot visualises the distribution of users' lifespan in years and engagement ratios by decile within city-level communities, illustrating that both behavioural measures follow a long right-tail distribution. The most long-lived users in the c1 decile exhibit a mean lifespan between four and six years, whereas younger cohorts from c6 onwards have a lifespan of no more than zero; additionally, fewer than half of the users in these city-level communities have a lifespan of one year or more. Data is sourced from Reddit/Pushshift \citep{baumgartner_pushshift_2020}.}
      
    \end{figure}

    I defined lifespan as the number of days elapsed between a user's first and most recent interaction within any of the city-level communities. Users were ranked by lifespan and grouped into deciles. Figure \ref{fig:boxplot_behaviour} provides a box plot representing the distribution of users across these deciles. Notably, the data for community moderators (\glspl{mod}) were excluded from this analysis to prevent distortion of the averages.
    
    The second measure, engagement ratio, was calculated as the ratio of a user's total score to their total number of interactions (the sum of selfposts and comments). A high engagement ratio signifies that a user is not only highly active but also consistently receives positive feedback (upvotes) from the community. Conversely, a low ratio indicates minimal interaction or a negative community reception (downvotes). As with lifespan, users were ranked and grouped into engagement ratio deciles. These are also represented in Figure \ref{fig:boxplot_behaviour}, with moderator data again removed to ensure the representativeness of the general user base.

    As observed in Figure \ref{fig:boxplot_behaviour}, the engagement ratio deciles also exhibit a distribution with a long right tail. Consistent with Figure \ref{fig:scatter_rank}, there is a polar distribution at each extremity. The decision to combine a user's total score and their total engagement into a single ratio was informed by a strong positive correlation between these two variables (Pearson's $R=0.91$). In contrast, there appears to be little to no relationship between a user's lifespan and their total score (Pearson's $R=0.26$) or their engagement (Pearson's $R=0.25$). This independence is even more pronounced when comparing lifespan to the engagement ratio (Pearson's $R=0.04$). Consequently, I maintain these behavioural measures as distinct dimensions of user activity. This separation allows for a more granular analysis of how temporal persistence (lifespan) and social resonance (engagement) independently influence grammatical alignment within the network.

\subsubsection{Evaluation}
\label{c2xg:c2xg_evaluation}

    Given that the typology of Reddit communities - specifically the distinction of place-based subreddits - plays a significant role in community stratification \citep{panek_understanding_2022}, I predict that topic and communicative purpose will be the primary determinants of network distance. Specifically, I anticipate that place-based communities, such as \texttt{r/auckland} and \texttt{r/Wellington}, will exhibit high linguistic similarity and thus form a tightly integrated cluster. Conversely, I predict that the peripheral communities (the \acrshort{NSFW} group) will form a distinct network, isolated from the core New Zealand-related communities. This expected divergence is based not only on disparate social functions but also on the fact that these peripheral groups were excluded from the initial training of the New Zealand Reddit \acrshort{C2xG} features. If this prediction holds, it would provide empirical evidence that geographic dialect alignment is most robust within those digital spaces explicitly dedicated to the social and cultural performance of \textit{place}.

\paragraph{OLS Regression}

    I utilised the \texttt{regression.linear\_model.OLS} method from the \texttt{statsmodels} Python library \citep{seabold_statsmodels_2010} to train a series of \acrfull{OLS} regression models. The comprehensive performance metrics for these models are presented in Appendix \ref{app:models}. The dependent variables consist of the mean cosine similarity measures for each behavioural grouping, categorised into four distinct models:

    \begin{itemize}[nolistsep]
        \item \gls{lcm}: Lexical (\gls{lex}) cosine similarity by lifespan group.
        \item \gls{ldm}: Lexical (\gls{lex}) cosine similarity by engagement ratio.
        \item \gls{scm}: Syntactic (\gls{syn}) cosine similarity by lifespan group.
        \item \gls{sdm}: Syntactic (\gls{syn}) cosine similarity by engagement ratio.
    \end{itemize}
    
    To determine the drivers of this similarity, I employed three predictor variables: mean lifespan (\textsc{age}), total upvote score (\textsc{score}), and total engagement (\textsc{count}). This multi-model approach enables an interrogation of whether linguistic alignment is primarily a function of a user's longevity within the community or their level of social resonance and activity.

\paragraph{Community Detection}

    The primary objective of the network analysis is to identify and characterise the relationships between distinct digital entities. In this study, the nodes represent individual New Zealand-related communities, while the edges are weighted based on the cosine similarity derived from \acrshort{C2xG} features. To visualise and interrogate these structures, I utilised the \texttt{igraph} Python package \citep{csardi_igraph_2006}. As a core component of the analytical pipeline, I employed the in-built Louvain method \citep{blondel_fast_2008} for community detection. The Louvain method is a greedy optimisation algorithm designed to extract hidden modular structures - or communities - from large-scale networks by maximising modularity. This approach allows for the empirical identification of clusters within the New Zealand Reddit ecosystem, revealing how disparate subreddits align linguistically to form a contiguous speech community.

\subsubsection{Prediction and Hypotheses}
\label{c2xg:prediction}

    As the focus of this phase is on the individual user within place-based communities, I predict that longevity and engagement are significant predictors of \acrshort{C2xG} linguistic similarity. Specifically, I hypothesise that as a user’s tenure within the community increases and their level of engagement deepens, their \acrshort{C2xG} linguistic similarity to the group norm will also increase. This prediction rests on the assumption that linguistic alignment is a function of sustained social interaction; users who have `vetted' the community over time and are heavily invested in its social capital are more likely to exhibit the latent grammatical markers that define the contiguous speech community.
    
\subsection{User-Level Behaviour}
\label{c2xg:behavioural_features}

\begin{figure}
      \centering
      
            \includegraphics[width=\textwidth]{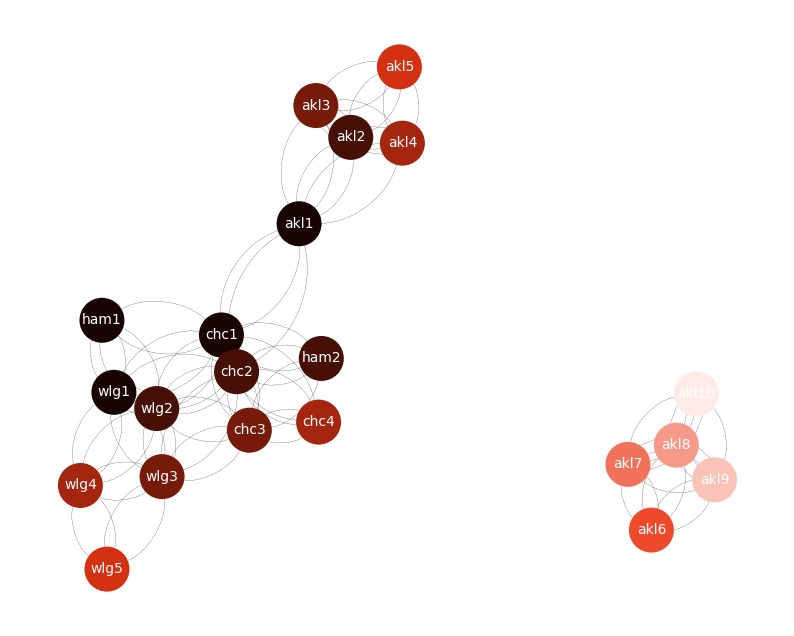}

      \vspace{6pt}
      \caption{Cosine Similarity of Lexical Features by Lifespan Cohort}
      \label{fig:igraph_cakelex}
      
      \vspace{6pt}
      \captionsetup{font=footnotesize, labelformat=empty, justification=justified, singlelinecheck=false}
      \caption*{\setstretch{2}\textbf{Description}: This network graph visualises the cosine similarity of \acrshort{C2xG} lexical features (\gls{lex}) between city-level communities, segmented by lifespan cohorts, with a threshold greater than 99\%. Nodes represent users grouped by lifespan decile, ranging from the most senior in the darkest shade of red (\texttt{1}) to the most junior in the lightest shade (\texttt{10}). The visualisation reveals two distinct clusters: a minor cluster comprising junior users (deciles 6–10) from \texttt{r/auckland} (akl), and a primary cluster consisting of senior users (deciles 1–5) from four city-level communities, excluding \texttt{r/Tauranga} (tau) and \texttt{r/dunedin} (dun); notably, the four most senior groups (decile 1) exhibit strong connectivity with users in the subsequent lifespan cohort (decile 2). All data is sourced from Reddit/Pushshift \citep{baumgartner_pushshift_2020}.}
      
    \end{figure}

\begin{figure}
      \centering
      
            \includegraphics[width=\textwidth]{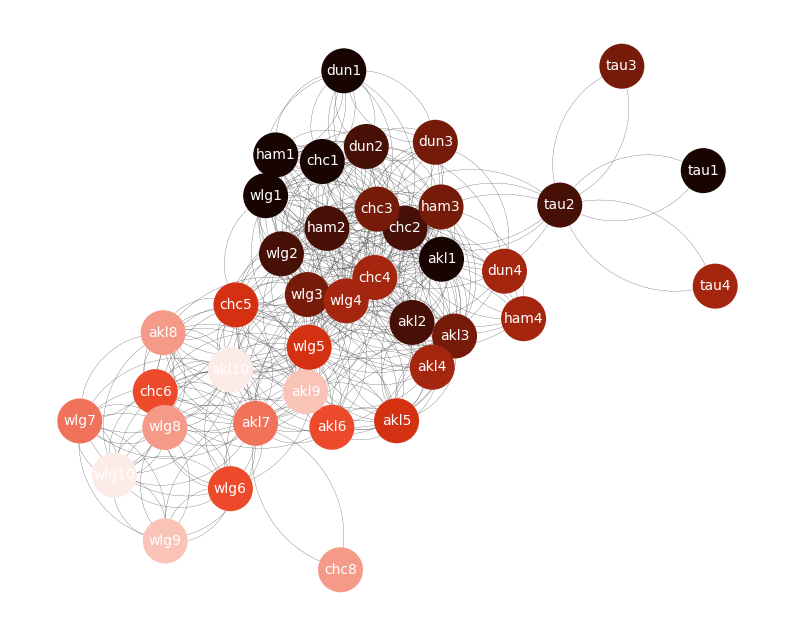}

      \vspace{6pt}
      \caption{Cosine Similarity of Syntactic Features by Lifespan Cohort}
      \label{fig:igraph_cakesyn}
      
      \vspace{6pt}
      \captionsetup{font=footnotesize, labelformat=empty, justification=justified, singlelinecheck=false}
      \caption*{\setstretch{2}\textbf{Description}: This network graph visualises the cosine similarity of \acrshort{C2xG} syntactic features (\gls{syn}) between city-level communities, segmented by lifespan cohorts, with a threshold of 99\%. Each node represents users grouped by lifespan decile, ranging from the most senior in the darkest shade of red (\texttt{1}) to the most junior in the lightest shade (\texttt{10}). The visualisation identifies one primary cluster with a core of most senior users from five city-level communities, which connects to a branching cluster of most senior users from \texttt{r/Tauranga} (tau); additionally, a secondary core within the primary cluster is formed by middle-aged users. Data is sourced from Reddit/Pushshift \citep{baumgartner_pushshift_2020}.}
      
    \end{figure}

    The purpose of this network analysis is to determine the grammatical relationship between New Zealand city-level communities based on two behavioural measures: lifespan cohort and engagement ratio. I began by analysing the network relationships between lifespan cohorts, utilising a threshold of cosine similarity greater than 99\% (see Figures \ref{fig:igraph_cakelex} and \ref{fig:igraph_cakesyn}). As this phase of the analysis relies heavily on the network graphs as visual evidence, detailed commentary is provided in the captions accompanying each figure.

    I observed that similar lifespan cohorts formed tightly integrated networks across the city-level communities. This suggests that users with comparable tenure in a digital space tend to employ language in a similar manner, according to the alignment of their \acrshort{C2xG} grammatical features. Furthermore, geographic origin emerged as a significant determinant of network ties, suggesting that even within digital environments, the physical \textit{place} remains a primary axis of linguistic organisation.

\begin{figure}
      \centering
      
            \includegraphics[width=\textwidth]{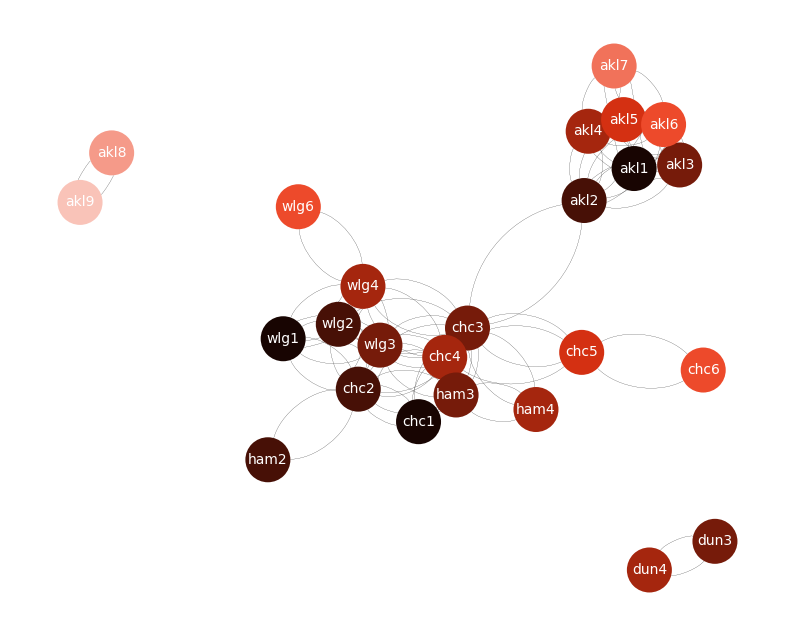}

      \vspace{6pt}
      \caption{Cosine Similarity of Lexical Features by Engagement Ratio}
      \label{fig:igraph_decilelex}
      
      \vspace{6pt}
      \captionsetup{font=footnotesize, labelformat=empty, justification=justified, singlelinecheck=false}
      \caption*{\setstretch{2}\textbf{Description}: This network graph visualises the cosine similarity of \acrshort{C2xG} lexical features (\gls{lex}) between city-level communities, segmented by engagement ratio deciles, with a threshold greater than 99\%. Nodes represent users ranging from the most engaged in the darkest shade of red (\texttt{1}) to the least engaged in the lightest shade (\texttt{10}). The visualisation identifies one primary cluster and two minor clusters; the minor clusters consist of a network of least engaged users (deciles 8–9) from \texttt{r/auckland} (akl) and moderately engaged users (deciles 3–4) from \texttt{r/dunedin} (dun). The primary cluster features a core of most to moderately engaged users (deciles 1–5) from three city-level communities - excluding \texttt{r/Tauranga}, \texttt{r/dunedin}, and \texttt{r/auckland} - with branches extending to include specific engagement deciles from \texttt{r/auckland}, \texttt{r/Wellington} (wlg), and \texttt{r/chch} (chc). Data is sourced from Reddit/Pushshift \citep{baumgartner_pushshift_2020}.}
      
    \end{figure}

\begin{figure}
      \centering
      
            \includegraphics[width=\textwidth]{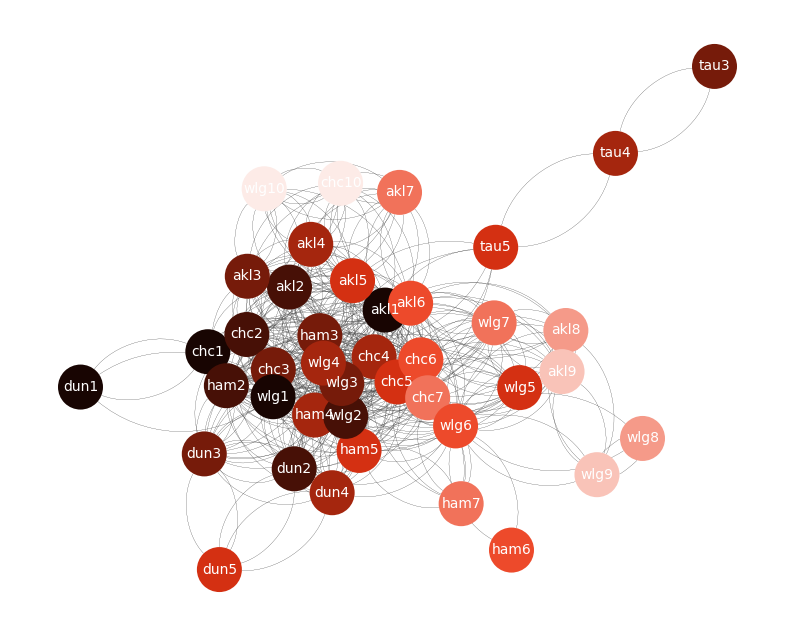}

      \vspace{6pt}
      \caption{Cosine Similarity of Syntactic Features by Engagement Ratio}
      \label{fig:igraph_decilesyn}
      
      \vspace{6pt}
      \captionsetup{font=footnotesize, labelformat=empty, justification=justified, singlelinecheck=false}
      \caption*{\setstretch{2}\textbf{Description}: This network graph visualises the cosine similarity of \acrshort{C2xG} syntactic features (\gls{syn}) between city-level communities, segmented by engagement ratio, with a threshold greater than 99\%. Each node represents a decile of users ranging from the most engaged in the darkest shade of red (\texttt{1}) to the least engaged in the lightest shade (\texttt{10}). The visualisation reveals a primary cluster and a branching cluster; the primary cluster consists of two internal groupings - one containing the most engaged users (deciles 1–5) and the other comprising moderate to least engaged users from five city-level communities - while \texttt{r/Tauranga} remains distinct as its own branch. Data is sourced from Reddit/Pushshift \citep{baumgartner_pushshift_2020}.}
      
    \end{figure}

    Regarding the engagement ratio cohorts, I visualised the network relationships - again using a 99\% cosine similarity threshold - in Figures \ref{fig:igraph_decilelex} and \ref{fig:igraph_decilesyn}. Similar to the lifespan results, users within the same engagement cohort formed close networks across the various city-level subreddits. In this instance, the role of geographic origin as a driver of network ties was even more pronounced. While these graphs offer significant insights into the internal characteristics of city-level communities, I am careful not to overstate these relationships, given the stringency of the 99\% similarity threshold required for inclusion in the network.

\subsubsection{Ordinary Least Squares Regression}

    The purpose of the \acrshort{OLS} regression models was to determine the impact of stratifying the city-level communities based on two behavioural measures: lifespan cohort and engagement ratio. The four dependent variables were defined as follows:
    
    \begin{itemize}[nolistsep]
        \item The mean cosine similarity of lexical features by lifespan cohort (\gls{lcm}).
        \item The mean cosine similarity of lexical features by engagement ratio cohort (\gls{ldm}).
        \item The mean cosine similarity of syntactic features by lifespan cohort (\gls{scm}).
        \item The mean cosine similarity of syntactic features by engagement ratio cohort (\gls{sdm}).
    \end{itemize}
    
    I fitted a suite of candidate models using three predictor variables: mean lifespan (\textsc{age}), total upvote score (\textsc{score}), and total engagement (\textsc{count}). Following a maximalist approach, I subsequently pruned the models with reference to the $F$-statistic, $R^2$, and adjusted $R^2$ values. Detailed performance metrics for these models are presented in Appendix \ref{app:models}. The results from the baseline model (a) ($\text{\textsc{age}} + \text{\textsc{score}} + \text{\textsc{count}}$) for \gls{lcm} ($R^2 = 0.756$; Adjusted $R^2 = 0.744$; $F = 33.111$) indicated that \textsc{age} and \textsc{count} both exerted a statistically significant ($p < 0.05$) positive effect on lexical alignment. Conversely, \textsc{score} exhibited a small but statistically significant ($p < 0.05$) negative effect. When \textsc{age} was removed in candidate model (b) ($\text{\textsc{score}} + \text{\textsc{count}}$), the $F$-statistic increased to $89.256$, though this resulted in a slight decrease in $R^2$ ($0.742$) and adjusted $R^2$ ($0.734$). The introduction of an interaction term in model (c) further reduced performance across all three measures ($R^2 = 0.742$; Adjusted $R^2 = 0.730$; $F = 58.569$), suggesting that \textsc{age} provides only marginal explanatory value in predicting \gls{lcm}.
    
    A consistent pattern emerged across the remaining three dependent variables, where the baseline model provided the most robust performance. The metrics for \gls{ldm} ($R^2 = 0.778$; Adjusted $R^2 = 0.767$; $F = 71.310$) are presented in Table \ref{tab:ols_ldm}, \gls{scm} ($R^2 = 0.662$; Adjusted $R^2 = 0.645$; $F = 39.829$) in Table \ref{tab:ols_scm}, and \gls{sdm} ($R^2 = 0.668$; Adjusted $R^2 = 0.651$; $F = 40.827$) in Table \ref{tab:ols_ldm}. Despite the cohorts being derived from these predictors, multicollinearity remained minimal. In summary, the \acrshort{OLS} results suggest that as temporal persistence (\textsc{age}) and volume of interaction (\textsc{count}) within a city-level community increase, grammatical similarity - measured by lexical and syntactic \acrshort{C2xG} cosine similarity - likewise increases. However, the inverse relationship observed with \textsc{score} suggests that higher community validation may correlate with a decrease in strict grammatical alignment.

\subsubsection{Classification Models}

    Following the \acrshort{OLS} regression analysis, I consider the impact of lifespan and engagement ratio groupings on text classification performance within the city-level communities. The primary objective of these classification models is to determine the degree of linguistic variance within each behavioural cohort. Utilising the proportional sampling procedure established in Chapter \ref{chap:dialect_classification}, I conducted city-level text classification for each lifespan and engagement ratio grouping, adjusting the fraction argument to account for varying corpus sizes. Given the reduced sample size for individual cohorts, I set the percentage-based fraction to $0.75$ across all text types to ensure sufficient training data.
    
    The model performance metrics for the lifespan cohorts are summarised in Table \ref{tab:cake_metrics}. Baseline metrics (without data cleaning) and the held-out test set support measures are provided in Table \ref{tab:cake_classification} of the Appendix. Based on these metrics, lifespan cohort \texttt{C8} exhibited the highest Weighted Average $F_1$-score ($0.70$), indicating high within-class variance - or greater distinctiveness - among city-level identities for this group. Conversely, cohort \texttt{C5} yielded the lowest Weighted Average $F_1$-score, suggesting lower within-group variance. Notably, I observed a quadratic distribution of the Weighted Average $F_1$-score across the lifespan cohorts, which suggests a normal distribution of within-group variance relative to user tenure.
    
    Regarding the engagement ratio cohorts, the performance metrics are detailed in Table \ref{tab:decile_metrics}, with the corresponding baseline models and support measures provided in Table \ref{tab:decile_classification}. Engagement ratio cohort \texttt{Q8} achieved the highest Weighted Average $F_1$-score ($0.72$), signifying high within-class variance. In contrast, cohort \texttt{Q7} produced the lowest score ($0.48$). With the exception of cohorts \texttt{Q8} and \texttt{Q9}, I observed a monotonic decrease in the Weighted Average $F_1$-score as the engagement ratio decreased. This indicates that as users become less engaged with the community, their city-specific linguistic markers become less distinct, leading to a higher degree of grammatical uniformity.

    \begin{table*}
        \scriptsize
        \centering
        \renewcommand{\arraystretch}{1.4}
            
            \caption{Performance Metrics by Lifespan Cohorts}
            \label{tab:cake_metrics}
        
            \centering
            \renewcommand{\arraystretch}{1.4}

                    \begin{tabularx}{\textwidth}{l*{10}{>{\centering\arraybackslash}X}}
                    \toprule 
                    Community & C1 & C2 & C3 & C4 & C5 & C6 & C7 & C8 & C9 & C10 \\ 
                    \midrule
                    \addlinespace[1em]
                    \texttt{r/auckland} & 0.77 & 0.72 & 0.72 & 0.73 & 0.74 & 0.76 & 0.81 & 0.79 & 0.81 & 0.79 \\ 
                    \texttt{r/Tauranga} & 0.24 & 0.30 & 0.13 & - & - & - & 0.89 & 0.67 & 0.57 & 0.33 \\ 
                    \texttt{r/thetron} & 0.67 & 0.50 & 0.36 & 0.20 & 0.16 & 0.17 & 0.62 & 0.57 & 0.71 & 0.40 \\ 
                    \texttt{r/Wellington} & 0.74 & 0.42 & 0.32 & 0.33 & 0.17 & 0.31 & 0.56 & 0.53 & 0.45 & 0.49 \\ 
                    \texttt{r/chch} & 0.59 & 0.22 & 0.14 & 0.12 & 0.06 & 0.05 & 0.52 & 0.33 & 0.42 & 0.49 \\ 
                    \texttt{r/dunedin} & 0.58 & 0.43 & 0.30 & 0.12 & 0.08 & 0.28 & 0.46 & 0.33 & 0.20 & 0.38 \\ 
                    \addlinespace[1em]
                    Macro Avg. & 0.60 & 0.43 & 0.33 & 0.25 & 0.20 & 0.26 & 0.64 & 0.54 & 0.53 & 0.48 \\
                    Weighted Avg. & 0.71 & 0.55 & 0.50 & 0.50 & 0.47 & 0.53 & 0.70 & 0.65 & 0.66 & 0.65 \\
                    \addlinespace[1em]
                    \bottomrule
                \end{tabularx}
    
            \tablenoteparagraph{\textbf{Table Note}: This table presents model performance metrics for New Zealand city-level communities across ten lifespan cohorts (\texttt{C1} to \texttt{C10}), where \texttt{C} denotes \textit{Cake} (lifespan). Utilising shallow linear \acrshort{SVM} classifiers with proportional sampling and localised data cleaning, the columns detail $F_1$-scores by community while rows provide macro and weighted average $F_1$-scores; observations with fewer than 500 words were excluded. The highest weighted average $F_1$-score occurred in \texttt{C1}, suggesting high within-class variance, while the lowest was in \texttt{C5}, indicating low within-class variance, with all data sourced from Reddit/Pushshift \citep{baumgartner_pushshift_2020}.}
        
    \end{table*}
    
    \begin{table*}
        \scriptsize
        \centering
        \renewcommand{\arraystretch}{1.4}
            
            \caption{Performance Metrics by Engagement Ratio}
            \label{tab:decile_metrics}
        
            \centering
            \renewcommand{\arraystretch}{1.4}
                \begin{tabularx}{\textwidth}{l*{10}{>{\centering\arraybackslash}X}}
                \toprule 
                    $F_1$-score & Q1 & Q2 & Q3 & Q4 & Q5 & Q6 & Q7 & Q8 & Q9 & Q10 \\ 
                \midrule
                \addlinespace[1em]
                    \texttt{r/auckland} & 0.74 & 0.74 & 0.75 & 0.74 & 0.73 & 0.74 & 0.76 & 0.85 & 0.82 & 0.77 \\ 
                    \texttt{r/Tauranga} & - & 0.11 & 0.24 & 0.35 & 0.28 & 0.07 & - & - & 0.60 & 0.12 \\ 
                    \texttt{r/thetron} & 0.10 & 0.59 & 0.69 & 0.64 & 0.49 & 0.31 & 0.11 & 0.62 & - & 0.07 \\ 
                    \texttt{r/Wellington} & 0.51 & 0.63 & 0.65 & 0.63 & 0.50 & 0.26 & 0.02 & 0.63 & 0.46 & 0.09 \\ 
                    \texttt{r/chch} & 0.15 & 0.31 & 0.44 & 0.47 & 0.36 & 0.18 & - & 0.19 & 0.17 & - \\ 
                    \texttt{r/dunedin} & - & 0.37 & 0.44 & 0.55 & 0.36 & 0.37 & 0.17 & 0.46 & 0.18 & - \\ 
                \addlinespace[1em]
                    Macro Avg. & 0.25 & 0.46 & 0.53 & 0.56 & 0.45 & 0.32 & 0.18 & 0.46 & 0.37 & 0.17 \\
                    Weighted Avg. & 0.56 & 0.62 & 0.65 & 0.65 & 0.58 & 0.52 & 0.48 & 0.72 & 0.65 & 0.49 \\
                \addlinespace[1em]
                \bottomrule
                \end{tabularx}
    
            \tablenoteparagraph{\textbf{Table Note}: This table presents model performance metrics for New Zealand city-level communities grouped by engagement ratio quantiles (\texttt{Q1} to \texttt{Q10}), using shallow linear \acrshort{SVM} classifiers with proportional sampling and localised data cleaning procedures. The columns detail the city-level communities (\texttt{Community}) and corresponding $F_1$-scores across the cohorts, while rows provide macro and weighted average $F_1$-scores; observations with fewer than 500 words were excluded. The highest weighted average $F_1$-score was observed in \texttt{Q8}, suggesting high within-class variance, whereas the lowest was found in \texttt{Q10}, indicating low within-class variance, with all data sourced from Reddit/Pushshift \citep{baumgartner_pushshift_2020}.}
        
    \end{table*}

\subsubsection{Summary}

    The behavioural measures of the city-level communities provide critical insight into their internal social characteristics. The \acrshort{OLS} regression models validated the relationships observed in the network graphs, confirming a positive correlation between lifespan, engagement volume, and grammatical similarity. Such findings are consistent with the expectations of a cohesive speech community. However, it is essential to consider that these results may partially represent an artifact of the training data used to learn the \acrshort{C2xG} features. 

\begin{figure}
      \centering
      
            \begin{subfigure}[t]{0.5\textwidth}
                \centering
                \includegraphics[width=\textwidth]{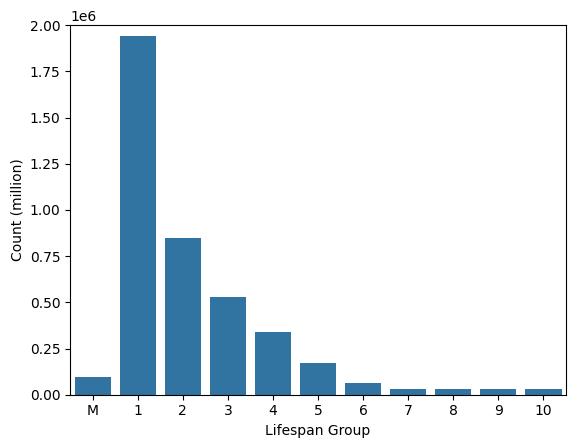}
                \caption{Lifespan Group}
            \end{subfigure}%
            ~ 
            \begin{subfigure}[t]{0.5\textwidth}
                \centering
                \includegraphics[width=\textwidth]{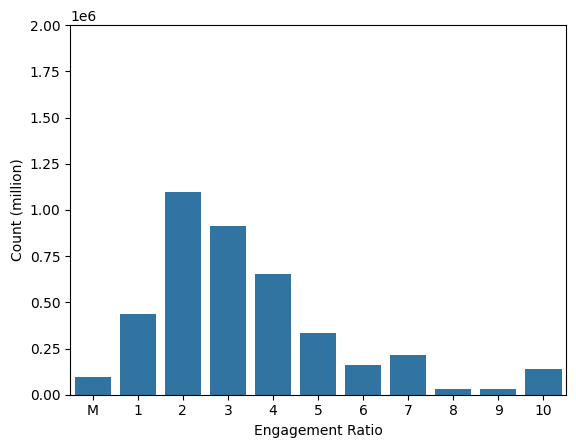}
                \caption{Engagement Ratio}
            \end{subfigure}\vspace{0.5cm}

      \caption{Observations by Lifespan and Engagement Ratio Cohorts}
      \label{fig:bias_c2xg}
      
      \vspace{6pt}
      \captionsetup{font=footnotesize, labelformat=empty, justification=justified, singlelinecheck=false}
      \caption*{\setstretch{2}\textbf{Description}: This barplot visualises the distribution of observations across lifespan and engagement ratio cohorts for the New Zealand-related communities utilised to train the \acrshort{C2xG} feature, indicating that more senior and highly engaged users are more significantly represented within the dataset; all data is sourced from Reddit/Pushshift \citep{baumgartner_pushshift_2020}.}
      
    \end{figure}

    As visualised in Figure \ref{fig:bias_c2xg}, which illustrates the number of observations (in millions) across lifespan and engagement ratio cohorts, there is a clear imbalance. Older users - those with the longest tenure - were significantly over-represented in the training data due to their inherent production advantage over newer users. In the context of \acrshort{C2xG} constructions, more senior users (\texttt{Decile 1}) were disproportionately represented compared to more junior users (\texttt{Deciles 6} to \texttt{10}) and the moderators (\glspl{mod}). 

    In contrast, the contributions of users stratified by engagement ratio followed a more recognisably normal distribution. While the restricted corpus size makes it impractical to account for these specific behavioural measures in the subsequent development of the Word2Vec embedding models, the results from this section nonetheless confirm that \acrshort{C2xG} serves as a highly effective measure of grammatical similarity across New Zealand-related communities. This reinforces the validity of the network of networks as a contiguous linguistic entity, even while acknowledging the disproportionate influence of long-term residents on the shared digital grammar.
    
\subsection{Community-level Analysis}
\label{c2xg:community_level}

    I now present the results for the \acrshort{C2xG} features trained across the 33 New Zealand-related communities. This community-level analysis incorporates the 14 peripheral subreddits, primarily associated with Adult Only content, to test the boundaries of the digital speech community. To determine the degree of alignment, I calculated the cosine similarity between each New Zealand-related community and \texttt{r/newzealand}, serving as the anchor for the national variety.

\begin{figure}
      \centering
      
            \includegraphics[width=\textwidth]{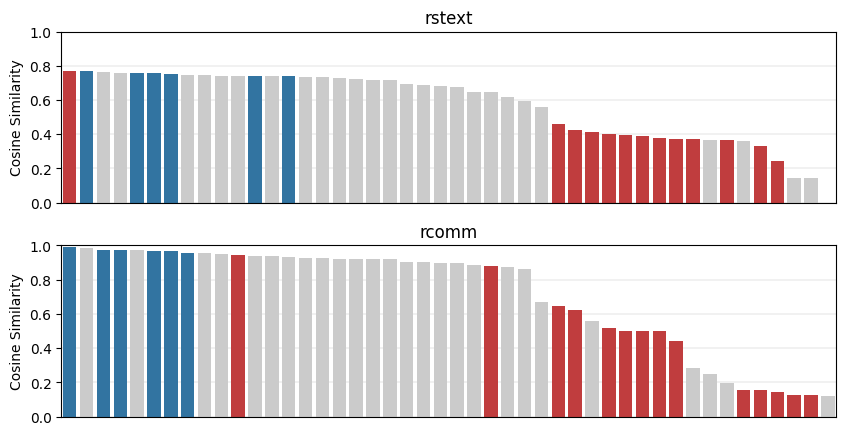}
            
      \vspace{6pt}
      \caption{Cosine Similarity of Lexical Features from \texttt{r/newzealand}}
      \label{fig:lex_nz}
      
      \vspace{6pt}
      \captionsetup{font=footnotesize, labelformat=empty, justification=justified, singlelinecheck=false}
      \caption*{\setstretch{2}\textbf{Description}: This barplot visualises the cosine distance of \acrshort{C2xG} lexical features from \texttt{r/newzealand} for selfpost body texts (\gls{rstext}) and comments (\gls{rcomm}), based on cosine similarity calculations between each community and \texttt{r/newzealand}. Bars are colour-coded to represent the six city-level communities in blue, 14 peripheral communities in red, and remaining communities in grey, with community names removed to improve interpretability; city-level communities exhibit the highest similarity to \texttt{r/newzealand} across both text-types, while peripheral communities - with the notable exception of the outlier \texttt{r/MedicalCannabisNZ} - show the least similarity, which was anticipated as they were not included in the training of \acrshort{C2xG} constructions. Data is sourced from Reddit/Pushshift \citep{baumgartner_pushshift_2020}.}
      
    \end{figure}

\begin{figure}
      \centering
      
            \includegraphics[width=\textwidth]{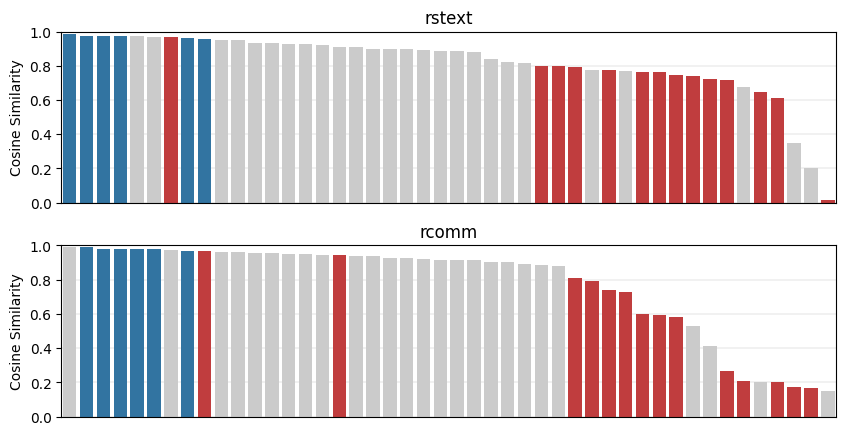}
            
      \vspace{6pt}
      \caption{Cosine Similarity of Syntactic Features from \texttt{r/newzealand}}
      \label{fig:syn_nz}
      
      \vspace{6pt}
      \captionsetup{font=footnotesize, labelformat=empty, justification=justified, singlelinecheck=false}
      \caption*{\setstretch{2}\textbf{Description}: This barplot visualises the cosine distance of \acrshort{C2xG} syntactic features from \texttt{r/newzealand} for selfpost body texts (\gls{rstext}) and comments (\gls{rcomm}), based on cosine similarity calculations between each community and \texttt{r/newzealand}. Bars are colour-coded to represent the six city-level communities in blue, 14 peripheral communities in red, and remaining communities in grey, with community names removed to improve interpretability; consistent with lexical features, city-level communities exhibit the highest similarity to \texttt{r/newzealand} across both text-types, while peripheral communities - with the notable exception of \texttt{r/MedicalCannabisNZ} - show the least similarity. Data is sourced from Reddit/Pushshift \citep{baumgartner_pushshift_2020}.}
      
    \end{figure}

    The results for the \acrshort{C2xG} lexical features are visualised in Figure \ref{fig:lex_nz}, and the syntactic features in Figure \ref{fig:syn_nz}, covering both selfpost body texts (\gls{rstext}) and comments (\gls{rcomm}). I found that the cosine similarity between the New Zealand-related communities and \texttt{r/newzealand} remained remarkably consistent across both feature sets and text types. Notably, the New Zealand city-level communities exhibited the highest degree of similarity to \texttt{r/newzealand}. In contrast, the 14 peripheral communities were consistently identified as the least similar.
    
    These preliminary results suggest that place-based communities are more grammatically aligned than other community types within the New Zealand Reddit ecosystem. This suggests that the performance of \textit{place} through language is a cohesive force that binds geographically oriented subreddits into a primary network, while interest-based or `furtive' communities operate on a different linguistic plane.

\paragraph{Community Detection}

\begin{figure}
      \centering
      
        \begin{subfigure}[b]{0.6\textwidth}
            \centering
            \includegraphics[width=\linewidth]{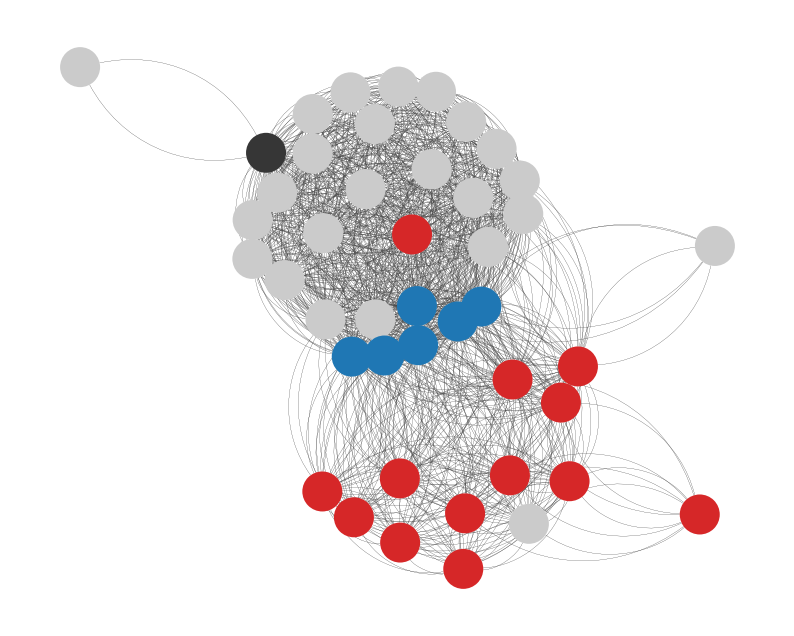}
            \subcaption{Cosine Similarity}
        \end{subfigure}
        \begin{subfigure}[b]{0.6\textwidth}
            \centering
            \includegraphics[width=\linewidth]{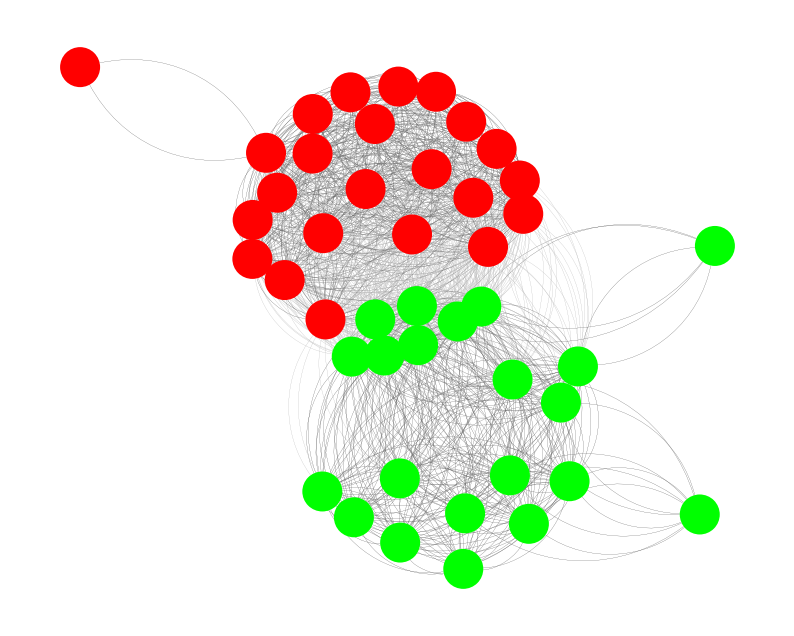}
            \subcaption{Latent Communities}
            \end{subfigure}
            
      \vspace{6pt}
      \caption{Network Graphs for \gls{rstext} by Lexical Features}
      \label{fig:lex3_nz}
      
      \vspace{6pt}
      \captionsetup{font=footnotesize, labelformat=empty, justification=justified, singlelinecheck=false}
      \caption*{\setstretch{2}\textbf{Description}: This figure displays subplots visualising network graphs for selfpost body texts (\gls{rstext}) based on cosine similarity (a) and latent communities (b) using \acrshort{C2xG} lexical features (\gls{lex}), with cosine similarity below 50\%. Nodes are colour-coded to represent \texttt{r/newzealand} in black, six city-level communities in blue, 14 peripheral communities in red, and remaining communities in grey; names have been removed to improve interpretability. Peripheral communities form a distinct second cluster (a), which is also identified by the Louvain method for community detection (b). Data is sourced from Reddit/Pushshift \citep{baumgartner_pushshift_2020}.}
      
    \end{figure}

\begin{figure}
      \centering
      
        \begin{subfigure}[b]{0.6\textwidth}
            \centering
            \includegraphics[width=\linewidth]{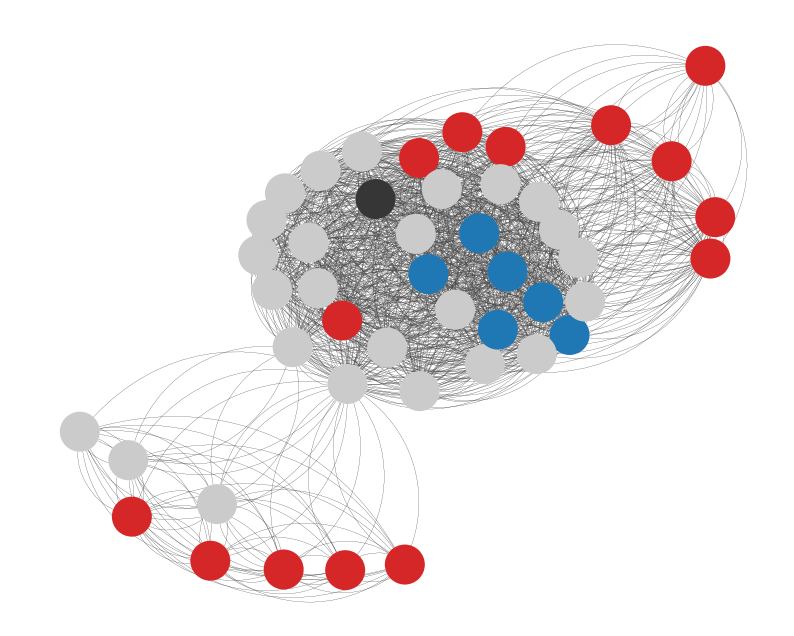}
            \subcaption{Cosine Similarity}
        \end{subfigure}
        \begin{subfigure}[b]{0.6\textwidth}
            \centering
            \includegraphics[width=\linewidth]{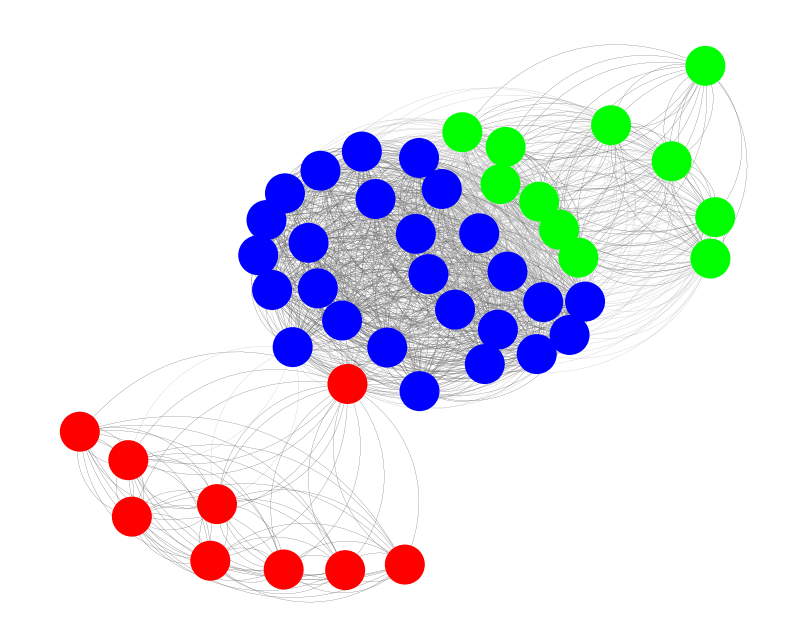}
            \subcaption{Latent Communities}
        \end{subfigure}
            
      \vspace{6pt}
      \caption{Network Graphs for \gls{rcomm} by Lexical Features}
      \label{fig:lex4_nz}
      
      \vspace{6pt}
      \captionsetup{font=footnotesize, labelformat=empty, justification=justified, singlelinecheck=false}
      \caption*{\setstretch{2}\textbf{Description}: This figure displays subplots visualising network graphs for comments (\gls{rcomm}) based on cosine similarity (a) and latent communities (b) using \acrshort{C2xG} lexical features (\gls{lex}), with cosine similarity below 50\%. Nodes are colour-coded to represent \texttt{r/newzealand} in black, six city-level communities in blue, 14 peripheral communities in red, and remaining communities in grey; community names have been removed to improve interpretability. Place-based communities form a primary cluster, while peripheral communities form secondary and tertiary clusters (a), which are also identified by the Louvain method for community detection (b). Data is sourced from Reddit/Pushshift \citep{baumgartner_pushshift_2020}.}
      
    \end{figure}

    Consistent with the user-level analysis for the city-level subreddits, I visualised the grammatical distance between the New Zealand-related communities using network graphs. In addition to mapping these relationships via cosine similarity, I employed the Louvain method to identify latent modular structures within the network. I provided detailed visualisations and commentary for the \acrshort{C2xG} lexical features for selfpost body texts (\gls{rstext}) in Figure \ref{fig:lex3_nz} and for comments (\gls{rcomm}) in Figure \ref{fig:lex4_nz}.
    
    Across both text types, the network relationships for the lexical features remained consistent: the 33 core New Zealand-related communities (including all place-based subreddits) formed a primary cluster, while the 14 peripheral communities were relegated to the edges of this main group. Encouragingly, those Adult Only subreddits already present within the core 33 communities shared strong network ties with the 14 peripheral communities. These results confirm that \acrshort{C2xG} lexical features provide a meaningful measure of the social and thematic boundaries within the New Zealand Reddit ecosystem.
    
\begin{figure}
      \centering
      
        \begin{subfigure}[b]{0.6\textwidth}
            \centering
            \includegraphics[width=\textwidth]
            {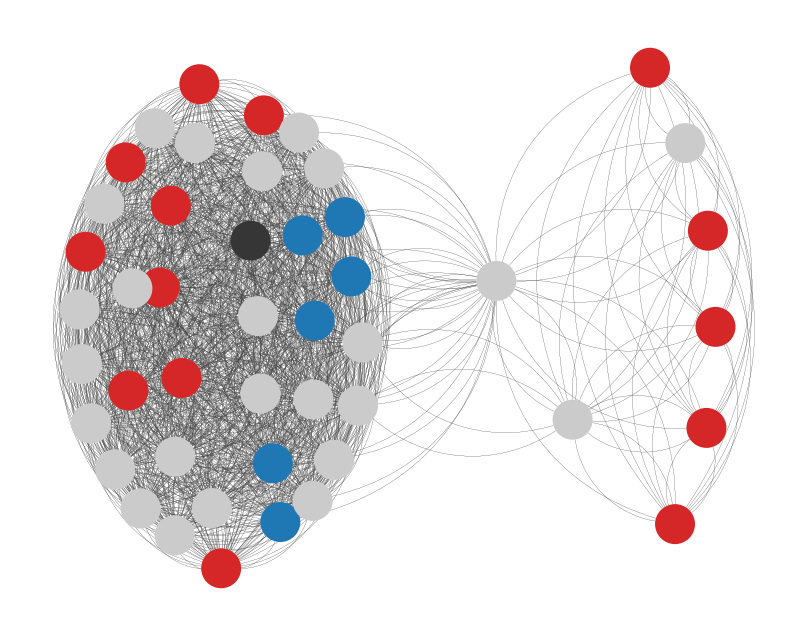}
            \subcaption{Cosine Similarity}
        \end{subfigure}
        \begin{subfigure}[b]{0.6\textwidth}
            \centering
            \includegraphics[width=\textwidth]{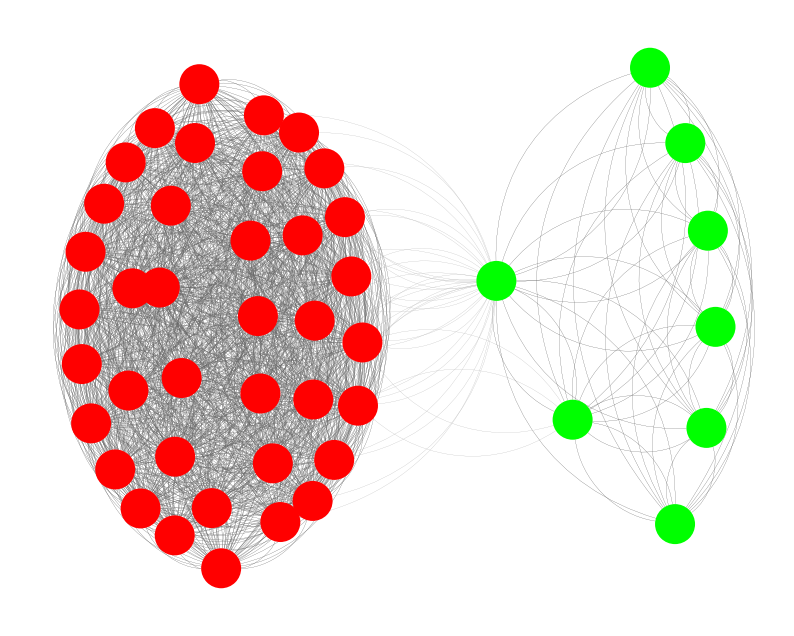}
            \subcaption{Latent Communities}
        \end{subfigure} 

      \vspace{6pt}
      \caption{Network Graphs for \gls{rcomm} by Syntactic Features}
      \label{fig:igraph_syn-cosine} 
      
      \vspace{6pt}
      \captionsetup{font=footnotesize, labelformat=empty, justification=justified, singlelinecheck=false}
      \caption*{\setstretch{2}\textbf{Description}: This figure displays subplots visualising network graphs for comments (\gls{rcomm}) based on cosine similarity (a) and latent communities (b) using \acrshort{C2xG} syntactic features (\gls{syn}), with cosine similarity below 50\%. Nodes are colour-coded to represent \texttt{r/newzealand} in black, six city-level communities in blue, 14 peripheral communities in red, and remaining communities in grey; names have been removed to improve interpretability. While the primary cluster appears to be an admixture of all community types, peripheral communities form a distinct second cluster (a), which is also identified by the Louvain method for community detection (b). Data is sourced from Reddit/Pushshift \citep{baumgartner_pushshift_2020}.}
      
    \end{figure}

    Next, I visualised the \acrshort{C2xG} syntactic features for comments (\gls{rcomm}) in Figure \ref{fig:igraph_syn-cosine}. While these results mirrored the lexical network in general terms, the 33 New Zealand-related communities - along with a small number of peripheral subreddits - formed a much tighter, more integrated cluster. The latent communities identified by the Louvain method confirmed these observations. Interestingly, when the combined (\gls{sem}) features were visualised, the 33 core communities and the 14 peripheral subreddits merged into a single, dense network with no obvious clustering.
    
    This phenomenon suggests that as grammatical complexity increases, network connectivity also increases across New Zealand Reddit. Such a result was anticipated, given the relative lack of morphosyntactic variation across \acrshort{NZE} varieties. Ultimately, the analysis confirms that while lexical features distinguish the core from the periphery, the shared syntactic and semantic architecture of these communities reinforces their status as a contiguous linguistic entity.

\paragraph{Classification Models}

\begin{figure}
      \centering
      
            \includegraphics[height=0.6\textwidth]{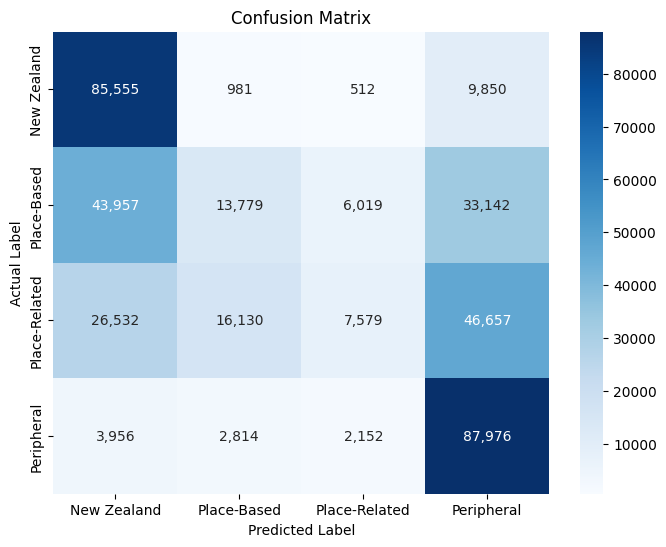}

      \vspace{6pt}
      \caption{Confusion Matrix from \acrshort{C2xG} Classification Model}
      \label{fig:confusion_matrix_sem}
      
      \vspace{6pt}
      \captionsetup{font=footnotesize, labelformat=empty, justification=justified, singlelinecheck=false}
      \caption*{\setstretch{2}\textbf{Description}: This figure displays a confusion matrix for the New Zealand-related groupings classification model, utilising SEM+ features parsed via \acrshort{C2xG} and trained on New Zealand-related communities; all data is sourced from Reddit/Pushshift \citep{baumgartner_pushshift_2020}.}
      
    \end{figure}

    Lastly, I utilised the \acrshort{C2xG} features to train a classification model designed to predict four custom, community-specific groupings: \texttt{r/newzealand} (National), the six city-level communities (Place-based), the remaining New Zealand-related communities (Place-related), and the 14 peripheral subreddits (Peripheral). The model performance metrics, presented in Table \ref{tab:f1_c2xg}, indicate that the \acrshort{C2xG} features achieved the highest class performance for \texttt{r/newzealand} across both selfpost body texts and comments. Unexpectedly, the model also demonstrated high precision in classifying the peripheral communities, suggesting they possess a distinct, albeit divergent, linguistic profile. The results indicate that these custom groupings maintain a continuous relationship governed by their underlying grammatical features. By inspecting the confusion matrix (see Figure \ref{fig:confusion_matrix_sem}), I confirmed that these communities are organised along a linguistic continuum. At one polar extreme lies \texttt{r/newzealand}, while the peripheral communities occupy the opposite extreme.
    
    The error distribution within the confusion matrix further clarifies this hierarchy: the place-based (city-level) communities showed a higher frequency of incorrect predictions with \texttt{r/newzealand}, suggesting a strong linguistic affiliation. In contrast, the place-related (interest-based) communities were more frequently misclassified as peripheral, indicating they occupy a space further removed from the national linguistic anchor. This alignment provides a clear empirical justification for the inclusion of city-level data in the diachronic models, as they represent the closest approximation to the national variety.

    \begin{table*}
            
            \caption{Performance Metrics for City-level Communities}
            \label{tab:f1_c2xg}
            
            \begin{subtable}[t]{\textwidth}
            \scriptsize
            \subcaption{\gls{rstext}}
            \renewcommand{\arraystretch}{1.4}
                  \begin{tabularx}{\textwidth}{l*{3}{>{\centering\arraybackslash}X}|
                  *{3}{>{\centering\arraybackslash}X}}
                    \toprule
                    \multirow{2}{*}{Community} & \multicolumn{3}{c}{$F_1$-score} & \multicolumn{3}{c}{Support}\\
                    & \gls{lex} & \gls{syn} & \gls{sem} & \gls{lex}& \gls{syn} & \gls{sem} \\
                    \midrule
                    \addlinespace[1em]
                    New Zealand & 0.59 & 0.60 & 0.57 & 9,294 & 9,772 & 43,707 \\
                    Place-Based & 0.13 & 0.22 & 0.00 & 9,294 & 9,772 & 43,707 \\
                    Place-Related & 0.10 & 0.00 & 0.22 & 9,294 & 9,772 & 43,707 \\
                    Peripheral & 0.63 & 0.64 & 0.62 & 9,294 & 9,772 & 43,707 \\
                    \addlinespace[1em]
                    Macro Average & 0.36 & 0.36 & 0.35 & 37,176 & 39,088 & 174,826 \\
                    Weighted Average & 0.36 & 0.36 & 0.35 & 37,176 & 39,088 & 174,826 \\
                    \addlinespace[1em]
                    \bottomrule
                \end{tabularx}
            \end{subtable}\vspace{6pt}
            
            \begin{subtable}[t]{\textwidth}
            \scriptsize
            \subcaption{\gls{rcomm}}
            \renewcommand{\arraystretch}{1.4}
                  \begin{tabularx}{\textwidth}{l*{3}{>{\centering\arraybackslash}X}|
                  *{3}{>{\centering\arraybackslash}X}}
                    \toprule
                    \multirow{2}{*}{Community} & \multicolumn{3}{c}{$F_1$-score} & \multicolumn{3}{c}{Support}\\
                    & \gls{lex}& \gls{syn} & \gls{sem} & \gls{lex}& \gls{syn} & \gls{sem} \\
                    \midrule
                    \addlinespace[1em]
                    New Zealand & 0.72 & 0.72 & 0.67 & 21,052 & 19,049 & 96,898 \\
                    Place-Based & 0.16 & 0.00 & 0.19 & 21,052 & 19,048 & 96,897 \\
                    Place-Related & 0.26 & 0.34 & 0.17 & 21,052 & 19,049 & 96,898 \\
                    Peripheral & 0.65 & 0.66 & 0.64 & 21,052 & 19,048 & 96,898 \\
                    \addlinespace[1em]
                    Macro Average & 0.45 & 0.43 & 0.42 & 84,208 & 76,194 & 387,591 \\
                    Weighted Average & 0.45 & 0.43 & 0.42 & 84,208 & 76,194 & 387,591 \\
                    \addlinespace[1em]
                    \bottomrule
                \end{tabularx}
            \end{subtable}\vspace{6pt}
    
            \tablenoteparagraph{\textbf{Table Note}: This table presents model performance metrics for submission selftext (\gls{rstext}) and comment (\gls{rcomm}) text-types, using shallow linear \acrshort{SVM} classifiers with proportional sampling and encoded \acrshort{C2xG} and month features. The columns detail city-level communities (\texttt{Community}), data cleaning procedures - including baseline (\texttt{Baseline}), standard processing (\texttt{Processed}), and New Zealand-specific entity removal (\texttt{Localised}) - and the test set size (\texttt{Support}). Rows provide $F_1$-scores for each condition, where the best-performing \gls{rstext} models for New Zealand city-level communities utilised lexical (\texttt{LEX}) and syntactic (\texttt{SYN}) features; all data is sourced from Reddit/Pushshift \citep{baumgartner_pushshift_2020}.}

    \end{table*}

\section{Interim Summary}
\label{c2xg:interim_summary}

    Utilising the \acrshort{C2xG} features derived from the collective linguistic data of New Zealand-related communities, I have successfully mapped the grammatical landscape of this digital ecosystem. Across lexical, syntactic, and combined feature sets, the analysis consistently identified \texttt{r/newzealand} as the national anchor, forming a core cluster with the six city-level communities. This central core is enveloped by the broader network of New Zealand-related subreddits, collectively constituting a contiguous speech community. The multi-methodological approach - incorporating cosine similarity, Louvain community detection, and text classification - suggests that while a continuous relationship exists between the core and the periphery, a clear boundary persists. As anticipated, the 14 peripheral communities remained distinct from the primary network cluster, although a secondary association with other adult-oriented subreddits within the core 33 was observable.
    
    Furthermore, the investigation into user behaviour revealed that while engagement metrics provide nuanced insights into internal community dynamics, temporal persistence (\textsc{age}) remains the most significant predictor of grammatical alignment. Consequently, I have secured sufficient empirical evidence to justify the incorporation of the 33 New Zealand-related communities into the development of my diachronic Word2Vec embedding models. The 14 peripheral communities will remain excluded to preserve the semantic and grammatical integrity of the New Zealand variety being modelled. This established dataset provides the robust foundation required to analyse semantic shifts in the subsequent phase of this research. 

\section{Diachronic Embedding Models}
\label{c2xg:diachronic_embeddings}

    Following the validation of the New Zealand-related communities as a contiguous speech community, I utilise Word2Vec embedding models to investigate diatopic and diachronic variation. The most direct approach to diachronic modelling involves training independent embeddings for discrete temporal windows. However, a significant methodological challenge arises: word embedding models are inherently stochastic \citep{kutuzov_diachronic_2018}. This stochasticity implies that any observed variation between word vectors across time slices may simply be an artifact of random initialisation during the learning and training phases, rather than a reflection of genuine diachronic semantic change. To mitigate this, two primary strategies exist: vector space alignment and incremental training. In this study, I evaluate these approaches to ensure that the semantic trajectories of the user-informed variables are representative of actual linguistic shifts within the New Zealand digital landscape. By controlling for the noise inherent in the training process, I can move from identifying where a word is used to understanding how its meaning evolves across the temporal span of the dataset.

\subsection{Methodology}

    I now synthesise the findings from the preceding three phases to develop the diachronic embedding models utilizing the New Zealand-related communities on Reddit. The primary objective of this phase is to quantify diachronic variation - or semantic change - within the 13 user-informed variables previously identified in Chapter \ref{chap:user_intuitions}. These variables serve as the linguistic focal points for examining how place identity and cultural concepts are negotiated over time within the New Zealand digital landscape. The development of these models relies on the validation provided by the \acrshort{C2xG} analysis, which confirmed that the 33 core communities constitute a contiguous speech community suitable for aggregate diachronic study. By applying the incremental training approach to this 1.74 billion-word corpus, I can observe the semantic trajectories of the user-informed semantic variables with a high degree of confidence that the shifts represent genuine linguistic evolution rather than stochastic noise. This phase moves the analysis from the \textit{who} and the \textit{where} to the \textit{when}, providing a longitudinal perspective on how the New Zealand Reddit ecosystem reflects broader social and cultural shifts in the physical world.

\subsubsection{Feature Engineering and Sampling}

    In combining the 33 New Zealand-related communities, the initial corpus comprised 4.26 billion words, including 2.51 billion from \texttt{r/newzealand}, 706 million from the city-level communities, and 1.01 billion from the remaining subreddits. To ensure the integrity of the diachronic models, I retained the feature engineering procedures developed in Chapter \ref{chap:user_variables}. This involved the removal of observations associated with \glspl{mod} within the place-based communities, as well as those suspected of containing machine-generated text.

    A significant refinement in this phase is the restriction of observations to between 06:00 and 24:00 \acrshort{NZST}. This serves as a rudimentary yet effective measure to control for the non-local production of language. Furthermore, I filtered out all observations of 500 words or longer to maintain a focus on organic social media discourse. Following these procedures, the refined dataset consists of approximately 1.35 billion words, distributed as follows: 832 million from \texttt{r/newzealand}, 182 million from city-level communities, and 339 million from other communities. To facilitate time-series training with an equal number of samples, I utilised a proportional sampling strategy to divide the corpus into four quartiles:

    \begin{enumerate}[nolistsep]
        \item Pre-October 2019
        \item November 2019 to January 2022
        \item February 2022 to August 2023
        \item Post-September 2023
    \end{enumerate}

    Each period contains 187 million words, with a composition of 69.7\% from \texttt{r/newzealand}, 20.8\% from other communities, and 9.5\% from the city-level communities. Consistent with previous models, the predominant text types remained selfpost body texts (\gls{rstext}) and comments (\gls{rcomm}). During the \texttt{spaCy} pre-processing pipeline, I removed non-Latin script characters, \acrshortpl{URL}, and strings associated with usernames (\texttt{u/}) and subreddits (\texttt{r/}). Given the geographic focus of the study, I employed a custom named entity recogniser within \texttt{spaCy}, incorporating an additional 161,796 New Zealand-specific place names to mask geographic identifiers. This ensures that the resulting embeddings capture latent semantic shifts rather than mere changes in the frequency of specific geographic mentions.

\subsubsection{Model Development}

    Based on the findings from Chapter \ref{chap:dialect_classification}, the Skip-gram with Negative Sampling (\acrshort{SGNS}) model architecture demonstrated superior performance in capturing the linguistic nuances of New Zealand city-level communities. Consequently, I employed the \acrshort{SGNS} architecture for the development of the diachronic embedding models. To ensure temporal alignment across the corpus quartiles, I adopted a combined strategy of sequential and incremental training. This approach initialises the weights of each successive model with those of the preceding period, thereby situated the vectors within a consistent coordinate system.

    Regarding hyperparameter configuration, I set the vector dimensionality to 300, providing sufficient capacity to capture complex semantic relationships without over-fitting the data. The context window was set to 5, a standard setting for capturing both local syntactic and broader semantic associations. To ensure the reliability of the embeddings and to eliminate noise from rare tokens, I applied a minimum word frequency threshold of 5. These parameters ensure that the resulting models are both robust and computationally efficient, providing a reliable basis for the analysis of the 13 user-informed semantic variables.
    
\subsubsection{Evaluation}

    To assess the robustness of the diachronic embedding models, I evaluated their performance using the same gold-standard test set established in Chapter \ref{chap:dialect_classification}. By utilising the results from the New Zealand city-level Word2Vec models as a benchmark, I established clear expectations for semantic stability across the temporal quartiles. If the word vectors remain stable and the incremental training successfully preserves the coordinate space, I anticipate a mean cosine similarity of approximately $0.512$ for the hypocoristics and $0.595$ for the user-informed sociolinguistic categories. Achieving these benchmarks would confirm that the refined 1.35 billion-word dataset - and the subsequent pre-processing and temporal filtering - has not compromised the model's ability to capture the specific lexical and cultural nuances of the New Zealand variety. Furthermore, consistency with these benchmarks provides the necessary empirical confirmation to interpret any observed vector movement as genuine diachronic drift rather than a failure of model convergence or data quality.

\subsection{Preliminary Results}

    I present the model performance metrics in Table \ref{tab:cosine-embedding}, focusing on two evaluation sets: the 59 hypocoristic word pairs and the 65 user-informed word pairs. A primary objective was to compare the performance of independently trained sequential Word2Vec models against models trained incrementally across the four time periods. In the sequential models, the mean cosine similarity for both evaluation sets remained remarkably stable. I observed a variance of only $0.018$ for the Wiktionary-derived hypocoristics and $0.005$ for the user-informed variables. This stability was mirrored in the vocabulary size, which remained constant across the periods, ranging between 57,956 and 61,924 tokens.
    
    In contrast, the incremental Word2Vec models provided a much more dynamic view of semantic drift. Across the four periods, the mean cosine similarity for the 59 hypocoristic word pairs decreased from $0.515$ to $0.461$. This downward trend suggests that the semantic distance between hypocoristic and non-hypocoristic forms is increasing over time. I observed a parallel decrease in the 65 user-informed word pairs, where mean cosine similarity dropped from $0.600$ in the first period to $0.550$ in the fourth. This indicates an increasing semantic distance between conservative and innovative forms within the New Zealand variety.
    
    Specifically, I identified a monotonic decrease in cosine similarity for five hypocoristic word pairs - \textit{agro}, \textit{banger}, \textit{rigger}, \textit{smoko}, and \textit{tradie} - and a monotonic increase in one: \textit{durrie}. When comparing only the initial and final periods, cosine similarity decreased for 37 pairs and increased for 12. Among the user-informed pairs, 12 exhibited a monotonic decrease, while a broader comparison of the start and end points showed a decrease in 40 pairs and an increase in 14. These results suggest that while the four time periods offer a condensed view, they capture a clear trajectory of semantic divergence within the New Zealand Reddit ecosystem.

    \begin{table*}
            
            \caption{Comparison of Sequential and Incremental Word2Vec Embedding Models}
            \label{tab:cosine-embedding}
            
            \scriptsize
            \centering
            \renewcommand{\arraystretch}{1.4}
                \begin{tabularx}{\textwidth}{*{4}{>{\centering\arraybackslash}X} |
                *{3}{>{\centering\arraybackslash}X}}
                    \toprule
                    \multirow{2}{*}{Period} & \multicolumn{3}{c}{Sequential} & \multicolumn{3}{c}{Incremental}\\
                    & Hypocoristics & User Var. & Vocabulary & Hypocoristics & User Var. & Vocabulary \\
                    \midrule
                    \addlinespace[1em]
                    P1 & 0.515 & 0.600 & 61,924 & 0.515 & 0.600 & 61,924 \\
                    P2 & 0.529 & 0.601 & 58,187 & 0.495 & 0.560 & 79,327 \\
                    P3 & 0.533 & 0.605 & 58,389 & 0.480 & 0.557 & 91,498 \\
                    P4 & 0.527 & 0.601 & 57,956 & 0.461 & 0.550 & 101,458 \\
                    \addlinespace[1em]
                    \bottomrule
                \end{tabularx}
    
            \tablenoteparagraph{\textbf{Table Note}: This table presents model performance metrics for submission selftext (\gls{rstext}) and comment (\gls{rcomm}) text-types, utilising a linear \acrshort{SVM} classifier trained with balanced sampling and input features derived from averaged \acrshort{NZM} word vector representations. The columns detail the period label (\texttt{Period}), mean cosine similarity measures for 59 hypocoristic word pairs from Wiktionary (\texttt{Hypocoristics}) and 65 user-informed pairs (\texttt{User Var.}), and total vocabulary size (\texttt{Vocabulary}), grouped by sequential and incremental model training procedures; observations with fewer than 500 words were excluded, and all data is sourced from Reddit/Pushshift \citep{baumgartner_pushshift_2020}.}
        
    \end{table*}

    \begin{figure}
      \centering
      
            \includegraphics[height=0.6\textwidth]{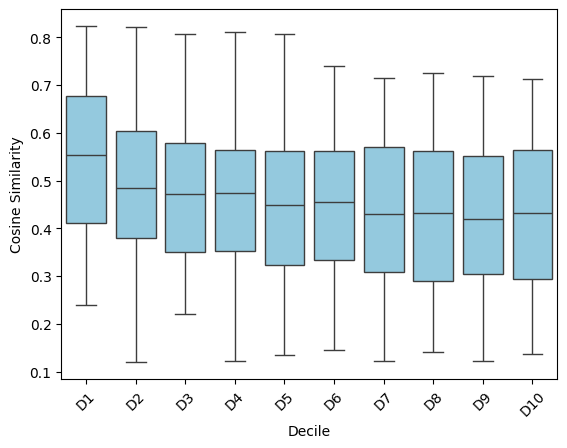}

      \vspace{6pt}
      \caption{Cosine Similarity of the Hypocoristic Word Pairs}
      \label{fig:diachronic_evaluation}
      
      \vspace{6pt}
      \captionsetup{font=footnotesize, labelformat=empty, justification=justified, singlelinecheck=false}
      \caption*{\setstretch{2}\textbf{Description}: Cosine similarity measures from the incrementally updated diachronic Word2Vec embedding models trained on the New Zealand-related communities presented as boxplots. The evaluation set was made up of the 59 hypocoristic word pairs from the Wiktionary Entries (\textit{tradie} and \textit{tradesman}). With reference to the median and interquartile ranges for each period, I observed a gradual decrease of cosine similarity over time which suggest the semantic distance of hypocoristic word pairs as a group has increased. \textbf{Data Source}: Reddit/Pushshift \citep{baumgartner_pushshift_2020}}
      
    \end{figure}

    To increase the number of training periods, I lowered the observation length threshold from 500 words to 150 words. This threshold was derived from the mean observation lengths across the sub-corpora: 198 words for \texttt{r/newzealand}, 162 words for the city-level communities, and 162 words for the other communities. This adjustment increased the total training data by 86.6\% to 2.52 billion words, comprising 1.5 billion words from \texttt{r/newzealand}, 394 million words from city-level communities, and 603 million words from other communities.
    
    Utilising the same proportional sampling strategy, I split the corpus into ten distinct periods with 161 million words per period. Although this resulted in a 13.7\% decrease in word count per period compared to the quartiles, the increase from four to ten periods provides a more granular resolution for identifying diachronic change. The composition of each 161-million-word period is as follows: 59.8\% from \texttt{r/newzealand}, 15.6\% from city-level communities, and 23.9\% from other communities.
    
    This sampling frame ensures that each temporal slice remains statistically comparable while allowing for a more detailed interrogation of the semantic trajectories of the 13 user-informed variables. By increasing the temporal frequency, I can better distinguish between short-term fluctuations and sustained linguistic shifts.

    \begin{table*}
        \scriptsize
        \centering
        \renewcommand{\arraystretch}{1.4}
            
            \caption{Performance Metrics for Diachronic Word2Vec Models}
            \label{tab:diachronic_metrics}

            \centering
            \renewcommand{\arraystretch}{1.4}
                \begin{tabularx}{\textwidth}{*{5}{>{\centering\arraybackslash}X}}
                \toprule 
                Period & Date Range & Hypocoristics & User-Informed & Vocabulary \\
                \midrule
                \addlinespace[1em]
                    P1 & Before 2016-11 & 0.541 & 0.617 & 63,538 \\
                    P2 & 2016-12---2018-12 & 0.477 & 0.562 & 78,576 \\
                    P3 & 2019-01---2020-06 & 0.468 & 0.546 & 90,532 \\
                    P4 & 2020-07---2021-06 & 0.453 & 0.536 & 99,295 \\
                    P5 & 2021-07---2022-01 & 0.443 & 0.527 & 106,874 \\
                    P6 & 2022-02---2022-08 & 0.440 & 0.534 & 113,620 \\
                    P7 & 2022-09---2023-04 & 0.433 & 0.533 & 119,932 \\
                    P8 & 2023-05---2023-11 & 0.428 & 0.534 & 125,631 \\
                    P9 & 2023-12---2024-05 & 0.428 & 0.520 & 130,911 \\
                    P10 & After 2024-06 & 0.424 & 0.525 & 136,010 \\
                \addlinespace[1em]
                \bottomrule
                \end{tabularx}
    
            \tablenoteparagraph{\textbf{Table Note}: This table outlines the period label (\texttt{Period}) and corresponding date range (\texttt{Date Range}), alongside the mean cosine similarity measures for 59 hypocoristic word pairs from Wiktionary (\texttt{Hypocoristics}) and 52 user-informed word pairs (\texttt{User Var.}). It also includes the total count of words and phrases in the vocabulary (\texttt{Vocabulary}), with all data sourced from Reddit/Pushshift \citep{baumgartner_pushshift_2020}.}
        
    \end{table*}

\subsection{User-informed Semantic Variables}

    I evaluate the degree of semantic shift for the 13 user-informed semantic variables using the diachronic embedding models trained via the incremental method. For each variable, I calculated the cosine similarity between the source word and its corresponding conservative and innovative variants across the ten temporal periods (P1 to P10). The target words remain consistent with those established in Chapter \ref{chap:user_variables}. A monotonic increase in cosine similarity over time indicates a decrease in semantic distance, while a monotonic decrease suggests an increase in semantic distance. If the users' intuitions regarding linguistic change are accurate, the models should demonstrate a monotonic decrease for the conservative variant and a monotonic increase for the innovative variant.
    
    The results are grouped by the nature of the observed diachronic change and are presented in Tables \ref{tab:semantic_embeddings1} and \ref{tab:semantic_embeddings2}. A trend consistent with user intuitions was observed for \textsc{bum}, \textsc{rubber}, and \textsc{tea}, each showing a monotonic decrease in the conservative variant and a monotonic increase in the innovative variant. Conversely, \textsc{fag} exhibited an inverse trend, with a monotonic increase in the conservative variant and a monotonic decrease in the innovative variant. This shift was likely influenced by platform-wide content moderation and community guidelines \citep{he_platform_2025}. For \textsc{football}, a monotonic increase was observed for both target variants, \textit{soccer} and \textit{rugby}, despite users identifying them as conservative and innovative respectively.
    
    These results suggest that while some variables align closely with community intuition, others are subject to external pressures or polysemous competition that complicates the binary model of conservative versus innovative forms.

    \begin{table*}
        \scriptsize
        \centering
        \renewcommand{\arraystretch}{1.4}
            
            \caption{Semantic Shift in User-Informed Semantic Variables}
            \label{tab:semantic_shift1}
        
            \centering
            \renewcommand{\arraystretch}{1.4}

            \begin{tabularx}{\textwidth}{*{3}{>{\arraybackslash}X}*{5}{>{\centering\arraybackslash}X}}
                    \toprule 
                    Variable & Source & Target & Start & End & Increase & Decrease & Difference \\ 
                    \midrule
                    \addlinespace[2em]
                    \multicolumn{8}{c}{Conservative Decreasing, Innovative Increasing} \\
                    \addlinespace[1em]
                    \multirow[c]{2}{*}{\textsc{bum}} & \multirow[c]{2}{*}{bum}
                        & \underline{smoke} & 0.379 & 0.205 & False & True & -0.174 \\
                        & & sex & 0.205 & 0.214 & True & False & 0.009 \\
                    \addlinespace[1em]
                    \multirow[c]{2}{*}{\textsc{rubber}} & \multirow[c]{2}{*}{rubber}
                        & \underline{pencil} & 0.592 & 0.230 & False & True & -0.362 \\
                        & & sex & 0.087 & 0.104 & True & False & 0.017 \\
                    \addlinespace[1em]
                    \multirow[c]{2}{*}{\textsc{tea}} & \multirow[c]{2}{*}{tea}
                        & \underline{dinner} & 0.486 & 0.350 & False & True & -0.136 \\
                        & & coffee & 0.484 & 0.495 & True & False & 0.011 \\
                    \addlinespace[2em]
                    \multicolumn{8}{c}{Conservative Increasing, Innovative Decreasing} \\
                    \addlinespace[1em]
                    \multirow[c]{2}{*}{\textsc{fag}} & \multirow[c]{2}{*}{fag}
                        & \underline{cigarette} & 0.308 & 0.427 & True & False & 0.119 \\
                        & & person & 0.315 & 0.164 & False & True & -0.151 \\
                    \addlinespace[2em]
                    \multicolumn{8}{c}{Conservative Increasing} \\
                    \addlinespace[1em]
                    \multirow[c]{2}{*}{\textsc{football}} & \multirow[c]{2}{*}{football}
                        & \underline{soccer} & 0.586 & 0.656 & True & False & 0.070 \\
                        & & \underline{rugby} & 0.520 & 0.577 & True & False & 0.057 \\
                    \addlinespace[1em]
                    \bottomrule
                \end{tabularx}
    
            \tablenoteparagraph{\textbf{Table Note}: This table details the user-informed variable (\texttt{Variable}), the source word (\texttt{Source}), and the conservative and innovative target words (\texttt{Target}), alongside cosine similarity measures from Period 1 (\texttt{Start}) to Period 10 (\texttt{End}). Additional columns indicate monotonic increase (\texttt{Increase}) or decrease (\texttt{Decrease}) over time and the difference between target words (\texttt{Difference}); user-informed conservative variants are underlined in the \texttt{Source} rows, with all data sourced from Reddit/Pushshift \citep{baumgartner_pushshift_2020}.}
        
    \end{table*}

    \begin{table*}[p]
        \scriptsize
        \centering
        \renewcommand{\arraystretch}{1.4}
            
            \caption{Semantic Shift in User-Informed Semantic Variables (Continued)}
            \label{tab:semantic_shift2}
        
            \centering
            \renewcommand{\arraystretch}{1.4}

            \begin{tabularx}{\textwidth}{*{3}{>{\arraybackslash}X}*{5}{>{\centering\arraybackslash}X}}
                    \toprule 
                    Variable & Source & Target & Start & End & Increase & Decrease & Difference \\ 
                    \midrule
                    \addlinespace[2em]
                    \multicolumn{8}{c}{Conservative and Innovative Decreasing} \\
                    \addlinespace[1em]
                        \multirow[c]{2}{*}{\textsc{dick}} & \multirow[c]{2}{*}{dick}
                        & \underline{richard} & 0.307 & 0.127 & False & True & -0.180 \\
                        & & jerk & 0.493 & 0.436 & False & True & -0.057 \\
                    \addlinespace[1em]
                        \multirow[c]{2}{*}{\textsc{flannel}} & \multirow[c]{2}{*}{flannel}
                        & \underline{towel} & 0.760 & 0.637 & False & True & -0.123 \\
                        & & shirt & 0.495 & 0.453 & False & True & -0.042 \\
                    \addlinespace[1em]
                        \multirow[c]{2}{*}{\textsc{gay}} & \multirow[c]{2}{*}{gay}
                        & \underline{unhappy} & 0.216 & 0.192 & False & True & -0.024 \\
                        & & lesbian & 0.608 & 0.558 & False & True & -0.050 \\
                    \addlinespace[1em]
                        \multirow[c]{2}{*}{\textsc{kiwi}} & \multirow[c]{2}{*}{kiwi}
                        & \underline{eagle} & 0.375 & 0.181 & False & True & -0.194 \\
                        & & apple & 0.207 & 0.098 & False & True & -0.109 \\
                    \addlinespace[1em]
                        \multirow[c]{2}{*}{\textsc{pudding}} & \multirow[c]{2}{*}{pudding}
                        & \underline{dinner} & 0.519 & 0.454 & False & True & -0.065 \\
                        & & custard & 0.817 & 0.524 & False & True & -0.293 \\
                    \addlinespace[1em]
                        \multirow[c]{2}{*}{\textsc{tramp}} & \multirow[c]{2}{*}{tramp}
                        & \underline{mountain} & 0.549 & 0.489 & False & True & -0.060 \\
                        & & street & 0.223 & 0.198 & False & True & -0.025 \\
                    \addlinespace[1em]
                        \multirow[c]{2}{*}{\textsc{tuna}} & \multirow[c]{2}{*}{tuna}
                        & \underline{salmon} & 0.743 & 0.549 & False & True & -0.194 \\
                        & & eel & 0.476 & 0.269 & False & True & -0.207 \\
                    \addlinespace[1em]
                        \multirow[c]{2}{*}{\textsc{twink}} & \multirow[c]{2}{*}{twink}
                        & \underline{pen} & 0.449 & 0.278 & False & True & -0.171 \\
                        & & bear & 0.237 & 0.232 & False & True & -0.005 \\
                    \addlinespace[1em]
                    
                    \bottomrule
                \end{tabularx}
    
            \tablenoteparagraph{\textbf{Table Note}: This table presents data for the variables \textit{snapper} and \textit{chippy} (\texttt{Source}), comparing conservative and innovative target words (\texttt{Target}) using cosine similarity measures from Period 1 (\texttt{Start}) to Period 10 (\texttt{End}). The columns also indicate monotonic increase (\texttt{Increase}) or decrease (\texttt{Decrease}) over time, as well as the difference between conservative and innovative target words (\texttt{Difference}), with user-informed conservative variants underlined in the \texttt{Source} rows; all data is sourced from Reddit/Pushshift \citep{baumgartner_pushshift_2020}.}
        
    \end{table*}

    Of the remaining eight user-informed semantic variables, I observed a monotonic decrease in both the conservative and innovative target variants. I present the results for these user-informed semantic variables in Table \ref{tab:semantic_shift2}. A decrease in cosine similarity between the source and target words suggest that semantic distance has increased. The results from these eight user-informed semantic variables suggest that the changes in cosine similarity were more global (across the entire embedding) and not just concentrated on the variables of interest.

\section{Discussion}
\label{c2xg:discussion}

    The expansion of the corpus to 2.52 billion words facilitated a more granular interrogation of diachronic change across ten temporal periods. Reducing the observation threshold to 150 words did not significantly degrade model performance, as verified by the hypocoristic and user-informed evaluation sets. However, the results for the 13 user-informed semantic variables largely failed to support the intuitions of the \texttt{r/newzealand} community. With only three variables - \textsc{bum}, \textsc{rubber}, and \textsc{tea} - exhibiting the anticipated monotonic shift, there is insufficient evidence to conclude that community perception of conservative versus innovative forms consistently matches corpus-driven semantic trajectories.

    These negative results do not invalidate the diachronic embedding method; rather, they highlight the complexities of using perceptual features as benchmarks for distributional models. The discrepancy suggests that what users perceive as a dying or emerging word may be influenced by social salience or frequency of encounter rather than a fundamental shift in the word's vector space.
    
    To further illustrate the effectiveness of the models, I examine two specific hypocoristics with unexpectedly high token frequencies: \textit{\gls{snapper}} and \textit{\gls{chippy}}. In both instances, the high frequency is attributed to semantic extension, resulting in significant polysemy within the New Zealand-related communities. For example, \textit{\gls{snapper}} functions as an animal (the fish), a transport identifier (the Wellington-based transit card), and a general hypocoristic. These case studies demonstrate that while user-informed categories may be socially significant, the diachronic models are most effective when capturing the competition between polysemous meanings within the New Zealand-specific context.
    
    \begin{table*}[p]
        \scriptsize
        \centering
        \renewcommand{\arraystretch}{1.4}
            
            \caption{Diachronic Cosine Similarity of \textit{snapper} and \textit{chippy}}
            \label{tab:snapper_chippy}
        
            \centering
            \renewcommand{\arraystretch}{1.4}

                    \begin{tabularx}{\textwidth}{ll*{4}{>{\centering\arraybackslash}X}c}
                    \toprule 
                    Source & Target & Start & End & Increase & Decrease & Difference \\ 
                    \midrule
                    \addlinespace[1em]
                        \multirow[c]{2}{*}{snapper}
                        & \underline{fish} & 0.388 & 0.429 & True & False & 0.041 \\
                        & card & 0.418 & 0.341 & False & True & -0.077 \\
                    \addlinespace[1em]
                    \multirow[c]{2}{*}{chippy}
                        & \underline{potato\_chip} & 0.821 & 0.24 & False & True & -0.581 \\
                        & prime\_minister & 0.196 & 0.367 & True & False & 0.171 \\
                    \addlinespace[1em]   
                    \bottomrule
                \end{tabularx}
    
            \tablenoteparagraph{\textbf{Table Note}: This table presents data for the variables \textit{snapper} and \textit{chippy} (\texttt{Source}), comparing conservative and innovative target words (\texttt{Target}) using cosine similarity measures from Period 1 (\texttt{Start}) to Period 10 (\texttt{End}). The columns also indicate monotonic increase (\texttt{Increase}) or decrease (\texttt{Decrease}) over time, as well as the difference between conservative and innovative target words (\texttt{Difference}), with user-informed conservative variants underlined in the \texttt{Source} rows; all data is sourced from Reddit/Pushshift \citep{baumgartner_pushshift_2020}.}
        
    \end{table*}

    The first example, \textit{\gls{snapper}} ($n=6,980$), represents a conflict between a stable biological referent (the fish) and a regional technological referent (the Wellington-based transit card). With reference to Table \ref{tab:snapper_chippy}, I observed a monotonic decrease in the similarity to the innovative variant (\textit{card}) over time. Conversely, there was a monotonic increase in similarity to the conservative variant (\textit{fish}).This semantic drift aligns with the historical trajectory of the Snapper ticketing system. While the card was launched in Wellington in 2008, its expansion into Auckland (2011) and Whangārei (2014) was short-lived, with withdrawals occurring between 2013 and 2018. Although Snapper remains the primary payment method for Wellington buses as of June 2025, the recent rollout of the national \texttt{Motu Move} system in other regions likely limited the Snapper vector's growth, allowing the more geographically universal meaning of the fish species to regain semantic dominance in the national corpus.
    
    The second example, \textit{\gls{chippy}} ($n=3,496$), serves as a primary hypocoristic in \acrshort{NZE}, significantly outperforming the variant \textit{chippie} and the generic \textit{potato chip}. The high token frequency of \textit{\gls{chippy}} can be attributed to its dual role as a snack food and as the widely utilised nickname for Chris Hipkins, the 41\textsuperscript{st} Prime Minister of New Zealand (January–November 2023). In Table \ref{tab:snapper_chippy}, the models show a monotonic increase in the conservative variant (\textit{potato\_chip}), while the innovative variant (\textit{prime\_minister}) showed an inverse trend. Despite the intense burst of media and social media attention during Hipkins' term in 2023, the semantic \textit{pull} of the prime ministerial meaning appears to have peaked and then receded relative to the enduring use of the snack-related term. This suggests that while political events create significant temporary `noise' in the vector space, they may not necessarily displace the entrenched, traditional meanings of hypocoristics in the long term.

    \begin{figure}
      \centering
      
            \includegraphics[height=0.6\textwidth]{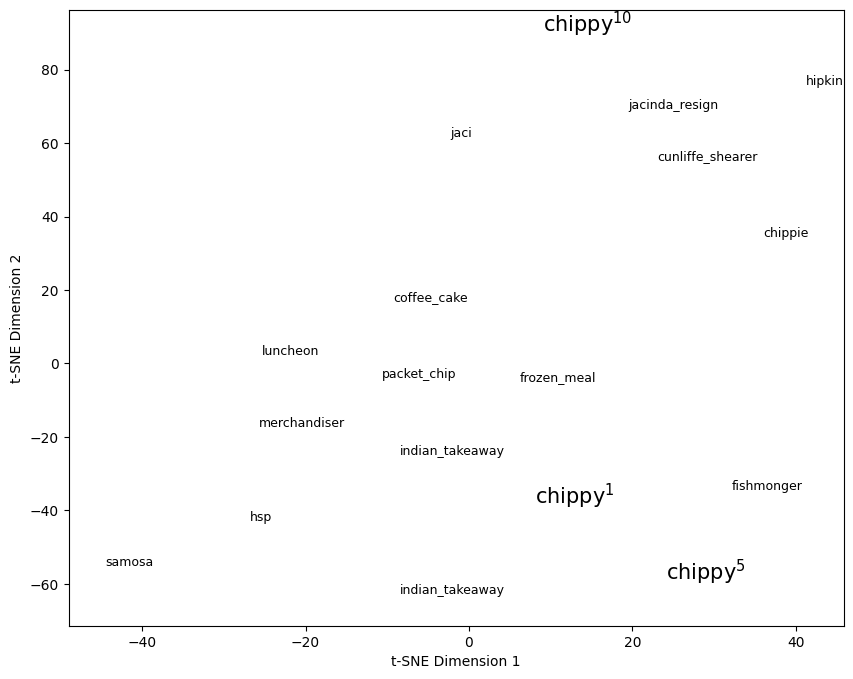}
            
      \vspace{6pt}
      \caption{Semantic Shift in \textit{chippy}}
      \label{fig:chippy_g3-tsne}
      
      \vspace{6pt}
      \captionsetup{font=footnotesize, labelformat=empty, justification=justified, singlelinecheck=false}
      \caption*{\setstretch{2}\textbf{Description}: This figure depicts the semantic shift of \textit{chippy} across three time periods, including chippy\textsuperscript{1} (before 2016-11), chippy\textsuperscript{5} (2021-07 to 2022-01), and chippy\textsuperscript{10} (after 2024-06), based on data sourced from Reddit/Pushshift \citep{baumgartner_pushshift_2020}.}
      
    \end{figure}

    \begin{figure}
      \centering
      
            \includegraphics[height=0.6\textwidth]{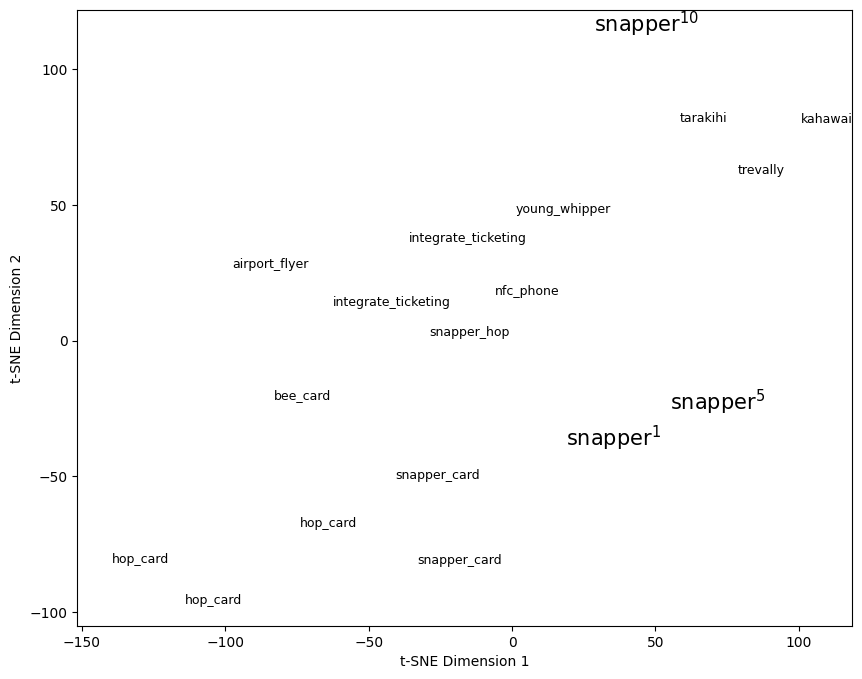}
            
      \vspace{6pt}
      \caption{Semantic Shift in \textit{snapper}}
      \label{fig:snapper_g3-tsne}
      
      \vspace{6pt}
      \captionsetup{font=footnotesize, labelformat=empty, justification=justified, singlelinecheck=false}
      \caption*{\setstretch{2}\textbf{Description}: This figure illustrates the semantic shift of \textit{snapper} over three distinct periods, specifically snapper\textsuperscript{1} (before 2016-11), snapper\textsuperscript{5} (2021-07 to 2022-01), and snapper\textsuperscript{10} (after 2024-06), with data sourced from Reddit/Pushshift \citep{baumgartner_pushshift_2020}.}
      
    \end{figure}

\section{Chapter Summary}
\label{c2xg:conclusion}

    Through the application of \acrshort{C2xG} \citep{dunn_computational_2017}, I have demonstrated that New Zealand-related communities on Reddit constitute a contiguous speech community. This linguistic unity suggests that the New Zealand variety is not confined to specifically place-based subreddits but is a pervasive characteristic of the wider national digital ecosystem. This foundational finding permitted the development of diachronic Word2Vec embedding models across an expanded corpus, allowing for a longitudinal interrogation of the 13 user-informed semantic variables.
    
    The results indicated that user intuitions regarding semantic change were largely inconsistent with the computational evidence. Only three of the thirteen variables exhibited the predicted monotonic shift away from conservative variants toward innovative ones. While this suggests a disconnect between community perception and actual usage patterns, it is important to note the impact of the expanding digital lexicon. The total vocabulary size effectively doubled between the initial period ($n=63,538$) and the final period ($n=136,010$), a factor that inevitably influences the density and movement of the vector space.
    
    The case studies of \textit{snapper} and \textit{chippy} provided a more nuanced validation of the models. These examples confirmed that semantic shifts are often driven by external socio-political events and infrastructural changes, even when they do not follow a simple linear path of innovation. Ultimately, while the user-informed variables proved to be complex benchmarks, the diachronic models successfully captured the dynamic nature of \acrshort{NZE} as it adapts to the unique cultural and technological pressures of the Reddit environment

% -----------------------------
% Chapter 8: Conclusion
% -----------------------------

\chapter{Conclusion}
\markboth{Conclusion}{}
\label{chap:conclusion}

\section{Chapter Outline}
\label{disc:outline}

    I now synthesise the findings from across the four phases of my research to address the primary research question (Section \ref{disc:primary_rq}) and the secondary research questions (Section \ref{disc:secondary_rq}). This is followed by a discussion of the limitations inherent to these studies (Section \ref{disc:limitations}). Finally, I propose avenues for future work in Section \ref{disc:future_work} before concluding with my final remarks in Section \ref{disc:final_remarks}.

\section{Introduction}

    Over the course of the last four chapters, I have explored geographic dialect alignment in place-based communities across Reddit. In Chapter \ref{chap:user_intuitions}: User Intuitions and Place Identity, I first considered the language ideologies and attitudes of users in \texttt{r/newzealand}. From this, I derived a set of user-informed sociolinguistic variables to examine geographic dialect alignment based on user perception. In Chapter \ref{chap:user_variables}: User-Informed Sociolinguistic Variables, I evaluated these intuitions by investigating the distributional patterns of these variables across six inner-circle, country-level communities on Reddit.

    Once I had established the distributional patterns of these user-informed variables, I utilised \acrshort{NLP} methods to examine the corpus characteristics of place-based communities. These insights informed the training of language embedding models to analyse the user-informed semantic variables in Chapter \ref{chap:dialect_classification}: Dialect Modelling and Embedding Models, which allowed me to define a network of place-based communities associated with New Zealand. Finally, in Chapter \ref{chap:construction_grammar}: Social Networks and Diachronic Embeddings, I expanded my analysis beyond place-based communities to establish a broader corpus of New Zealand-related communities.

\section{Primary Research Question}
\label{disc:primary_rq}

    I now address the primary research question: \textit{to what extent can we observe geographic dialect alignment in place-informed social media communities?} The underlying hypothesis is that a significant relationship exists between language use and place identity within these communities, and that this relationship is fundamentally connected to the sociolinguistic context of New Zealand. I evaluate geographic dialect alignment through two contrasting yet complementary approaches: the analysis of user-informed sociolinguistic variables and the implementation of language embedding models.

\subsection{Sociolinguistic Variables}

    I began my investigation by developing a set of user-informed lexical features associated with \acrshort{NZE} in Chapter \ref{chap:user_intuitions}: User Intuitions and Place Identity. I subsequently transformed these features into a set of 51 lexical variables and three morphosyntactic variables in Chapter \ref{chap:user_variables}: User-Informed Sociolinguistic Variables. I then determined the distribution of these variables across six inner-circle, country-level communities. A primary benefit of this approach was the ability to examine these variables within their original production context. Even though user intuitions were largely incorrect, the distributional patterns of these variables were systematic and meaningful, providing insights into potential changes in progress within \texttt{r/newzealand} as they relate to \acrshort{NZE} (specifically regarding terms such as \textit{nappy} and \textit{tramp}).

\subsection{Embedding Models}

    In contrast to the text classification models, the Word2Vec embedding models offered significantly greater explanatory power regarding geographic dialect alignment. Utilising the user-informed variables alongside open-source Wiktionary entries, I was able to assess linguistic variation between models trained on New Zealand city-level communities (\acrshort{NZM}), inner-circle country-level communities (\acrshort{ICM}), outer-circle country-level communities (\acrshort{OCM}), and a baseline model trained on the Google News dataset (\acrshort{GOO}; \citealp{mikolov_efficient_2013}). Based on my evaluation data, I successfully distinguished the four candidate models, finding that the \acrshort{NZM} model was the most sensitive to the semantic distance between 59 hypocoristic word pairs (e.g., \textit{sparky} and \textit{electrician}) based on mean cosine similarity. Conversely, the \acrshort{ICM}, \acrshort{OCM}, and \acrshort{GOO} models were markedly less sensitive to these specific semantic relationships.
    
    The embedding models further enabled an examination of the differences between word vector representations of the user-informed semantic variables. For example, \textit{rubber} and \textit{twink} exhibited higher cosine similarity to their conservative variants (target words: \textit{pencil} and \textit{pen}) in the \acrshort{NZM} model than in the other candidate models. In those broader models, both terms aligned more closely with innovative variants (target words: \textit{sex} and \textit{bear}). Using Word2Vec, I was able to observe distinct semantic shifts in these user-informed variables. I then extended this analysis to capture diachronic semantic shift by incorporating data from New Zealand-related communities, observing fluctuations in cosine similarity between the semantic variables and their conservative variants through incremental training over nine distinct periods.

\subsection{Other Features}

    While not necessarily reflecting specific linguistic features, text classification models did initially appear to distinguish country-level communities relative to the baseline models in Chapter \ref{chap:dialect_classification}: Dialect Modelling. Consistent with the findings of \citet{dunn_stability_2022}, these classifiers relied heavily on named entities - such as place names - which offer little explanatory power regarding latent linguistic variation and change. Nevertheless, text classification served as a useful diagnostic tool for the Reddit data, allowing me to determine the impact of sampling procedures and sample size for both the country-level and New Zealand city-level communities.
    
    One feature that demands further consideration is \acrshort{C2xG}, which I examined in Chapter \ref{chap:construction_grammar}: Construction Grammar and User Networks. Studies have shown \acrshort{C2xG} to be a meaningful predictor for distinguishing geographic dialects \citep{dunn_global_2019}. Understandably, the \acrshort{C2xG} features themselves were poor classifiers for the New Zealand city-level communities due to the low morphosyntactic variance between New Zealand-related communities (with the exception of peripheral communities). However, \acrshort{C2xG} features appeared to be a useful measure for grouping users based on behavioural metrics, such as lifespan cohorts and engagement ratios. This is consistent with recent findings where \acrshort{C2xG} features were combined with non-linguistic variables such as language contact and travel distance \citep{dunn_language_2025}.

\section{Secondary Research Questions}
\label{disc:secondary_rq}

    Now that I have established that distinct patterns of language use exist across country-level communities, I must determine the degree to which these differences are specific to New Zealand Reddit as a register, or if they are authentically associated with \acrshort{NZE}. The presence of dialect features within a specific community does not automatically imply that users are employing \acrshort{NZE} in a holistic sense. Therefore, the purpose of the secondary research questions was to explore the motivations for geographic dialect alignment in place-informed social media communities - specifically, to ask what motivates users in New Zealand-related communities to use language in a way that aligns with their place identity.

\subsection{SQ 1: Place Identity}
\label{disc:secrq1}

    \acrshort{SQ1} asks: Do users in place-based communities associate language use with a place identity? By establishing language use explicitly associated with New Zealand, the \acrshortpl{OP} and users of \texttt{r/newzealand} co-construct and reproduce a collective New Zealand identity within the community. Users associate \acrshort{NZE} with their mass identity of place, and the Discourse Models of New Zealand (alongside the endonym `kiwi') integrate these situated meanings. This alignment occurs because specific language use affords behavioural and existential insideness to some users, while the absence of these features is linked to behavioural and existential outsideness. Consequently, language use associated with \acrshort{NZE} contributes to place-making in \texttt{r/newzealand}, as defined by \citet{johnstone_place_2004}. Comments from self-identified outsiders further re-affirm this behavioural outsideness. These New Zealand-related communities effectively re-settle users on Reddit; moving the interaction from a mere coordinate to a lived experience, affirming that ``places are located, but are not locations'' \citep[pp.1-2]{scheider_places_2010}.

\subsection{SQ 2: Dialect Alignment}
\label{disc:secrq2}

    \acrshort{SQ2} asks: Is there a relationship between geographic dialect communities and place-based communities? The distribution of user-informed variables and the results of the embedding models confirm that features associated with \acrshort{NZE} are distinctly observable within \texttt{r/newzealand} and the New Zealand city-level subreddits. However, the absence of specific sociodemographic metadata for Reddit users means I cannot definitively rule out the influence of non-local users on the observed variation. Consequently, I cannot conclude a direct, one-to-one relationship between these digital communities and the physical variety of \acrshort{NZE}. Instead, these place-based communities should be conceptualised along a continuum of alignment, where users exhibit varying degrees of likelihood in utilizing features associated with \acrshort{NZE} based on their level of community engagement and local affiliation.

\subsection{SQ 3: Speech Community}
\label{disc:secrq3}

    \acrshort{SQ3} asks: Do place-related communities form a contiguous speech community? This question addresses the challenge of data availability inherent to non-dominant varieties of English by determining the actual extent of New Zealand Reddit. Place-based communities - such as \texttt{r/newzealand} and the six city-level subreddits - comprise only a small fraction of the broader ecosystem. Through a passive data collection approach and an analysis of \texttt{r/NZMetaHub}, I identified over 261 New Zealand-related communities, 32 of which were ranked within the top 40,000 communities in the Pushshift repository \citep{baumgartner_pushshift_2020}. To assess whether these communities function as a contiguous speech community, I utilised \acrshort{C2xG} as a measure of grammatical similarity. The results indicate that New Zealand-related communities exist on a continuum of alignment; many are closely aligned with the linguistic norms of \texttt{r/newzealand}, while others - primarily Adult Only communities - exist on the periphery. Ultimately, these communities function as a network of networks, mirroring the structure of a traditional speech community as defined by \citet{hymes_scope_2020}.

\subsection{Section Summary}

    After addressing the secondary research questions, I now return to the primary research question: To what extent can we observe geographic dialect alignment in place-informed social media communities? Based on the cumulative evidence provided by the secondary questions, the answer is a resounding yes. There is strong empirical evidence to suggest that geographic dialect alignment is observable within place-informed social communities, and that this relationship is fundamentally anchored to the sociolinguistic context of New Zealand. In the case of New Zealand-related communities on Reddit, a clear and systematic relationship exists between dialectal performance and place-related group affiliation. This suggests that digital spaces do not merely `flatten' linguistic variation, but rather provide new territories for the expression and maintenance of local identity.

\section{Limitations}
\label{disc:limitations}

    The primary criticism of perceptual dialectology is that language perception rarely reflects language production \citep{preston_language_2010}. As the objective of this thesis was to address the gap in the literature between user perceptions and \acrshort{NLP} applications on social media \citep{nguyen_dialect_2021}, I acknowledge the inherent limitations of user intuitions and the fact that they may not mirror actual linguistic production - an inconsistency explored in detail in Chapter \ref{chap:user_variables}: User-Informed Sociolinguistic Variables. Beyond the challenges of perception, I identify four specific limitations of this research: a reliance on user-informed variables; constraints in data availability; potential model bias; and limited engagement with foundational sociolinguistic and variationist theory.

\subsection{User-Informed Variables}

    The primary limitation of this research was the reliance on user-informed variables to guide the analysis, as the content of user comments ultimately dictated the granularity of the findings. From the two primary selfposts, I identified only 51 lexical and 13 semantic features suitable for analysis. In the case of morphosyntax, the selection was limited to a paltry three features; this sample was by no means representative of the complex morphosyntactic constructions associated with \acrshort{NZE} described in Chapter \ref{chap:literature}: Literature Review. Arguably, the lexical and semantic variables also constituted an under-representative sample of the \acrshort{NZE} lexicon. Furthermore, the priming effect from the \acrshortpl{OP} meant that the majority of identified variables highlighted contrasts between New Zealand and American English, potentially obscuring other regional nuances. However, a significant benefit of this user-informed approach was the mitigation of confirmation bias; by allowing users to define the variables, I avoided `cherry-picking' features to support my hypothesis - a rigour exemplified by the unexpected results for variables such as \textit{chippy} and \textit{snapper}.

\subsection{Data Availability}

    Regarding data availability, a significant limitation was that the corpus was restricted by the parameters of the Pushshift repository \citep{baumgartner_pushshift_2020}. Of the 261 New Zealand-related communities identified by users in \texttt{r/NZMetaHub}, only 32 were available within the Pushshift dumps. While these 33 subreddits (including \texttt{r/Tauranga}) were ranked within the top 40,000 communities globally, they represent only a limited subset of the broader New Zealand-related ecosystem. Furthermore, the identification of 14 peripheral New Zealand-related communities suggests a need for more refined data collection strategies to fully map the extent of New Zealand’s Reddit ecosystem as a cohesive network. Relatedly, the modest size of the New Zealand-related corpus impacted the temporal granularity of the diachronic embedding models. Although I successfully identified diachronic semantic shifts in two instances - \textit{chippy} and \textit{snapper} (see Section \ref{c2xg:discussion}) - it remains unclear whether the fluctuations in cosine similarity for other user-informed variables reflect genuine language change-in-progress or are simply artifacts of the vector representations in smaller datasets. Ultimately, a larger data volume would be required to analyse diachronic shifts with greater statistical resolution and granularity.

\subsection{Model Bias}
    
    A primary objective of this research was to investigate the role of geographic dialects within Reddit communities; as such, a degree of geographic focus was intentional. However, unexpected model bias was introduced throughout the data processing pipeline. For instance, the \texttt{spaCy} \texttt{EntityRecognizer} pipeline misclassified numerous non-English terms - particularly those from te reo Māori - as companies, agencies, or institutions. The \texttt{spaCy} English pipeline was trained on the OntoNotes 5.0 corpus, which is comprised of annotated American English sources such as newswire and telephone conversations \citep{weischedel_ralph_ontonotes_2013}. As a pre-trained model, it likely contained limited representations of \acrshort{NZE} and te reo Māori. While the \texttt{EntityRecognizer} did not play a central role in the final analysis, its failure highlights how model bias can be incrementally introduced at any stage of an \acrshort{NLP} pipeline when applied to non-dominant or indigenous linguistic contexts.

\subsection{Variation Theory}

    The final limitation of this thesis concerns the integration of sociolinguistic and variationist theory within the context of geographic dialect alignment. While the discourse analysis suggests that users index their identity through language use, metalinguistic commentary alone does not necessarily provide sufficient evidence of enregisterment \citep{johnstone_place_2004}. Substantive engagement between sociolinguistic theory and \acrshort{NLP} remains limited by the field's foundational preference for spoken language - the `vernacular' \citep{labov_principles_1972}. This preference is particularly evident in \acrshort{NZE} literature, which has historically focused on sound change and phonological variation \citep{gordon_new_2004}. Admittedly, researchers have successfully applied spoken-language concepts - such as linguistic diffusion \citep{eisenstein_diffusion_2014} and Communication Accommodation Theory \citep{danescu-niculescu-mizil_mark_2011} - to social media data. However, the findings in Chapter \ref{chap:user_variables}: User-Informed Sociolinguistic Variables demonstrate that the distribution of lexical and morphosyntactic variables cannot be analysed in isolation from the specific affordances of Reddit engagement. As this thesis represents an early integration of sociolinguistics and \acrshort{NLP} within the New Zealand context, I have prioritised methodological validity to provide a robust foundation for bridging this theoretical gap in future research.

\section{Implications and Contributions}
\label{disc:implications}

    In light of these limitations, the findings of this thesis significantly contribute to our understanding of language variation and change within place-informed Reddit communities. This research represents the first study to integrate \acrshort{NLP} methodologies with perceptual dialectology to investigate the sociolinguistics of \acrshort{NZE}. Specifically, the case studies presented showcase distinct yet complementary methodological approaches to analysing geographic dialect alignment on social media, bridging the existing research gap regarding the role of user perceptions in digital linguistic change. To summarise, this research makes three primary contributions to computational sociolinguistics:

    \begin{enumerate}[nolistsep]
        \item The development of place-informed social media dialectology through advanced language modelling.
        \item The integration of user perceptions as a valid metric for evaluating language variation and change in digital spaces.
        \item The production of a robust corpus of New Zealand-related Reddit communities, providing a significant resource for future linguistic inquiry.
    \end{enumerate}

\subsection{Place-Informed Dialectology}

    In contrast to prevailing spatial approaches to social media dialectology, I treated place as a form of implicit geographic information to determine geographic dialect alignment on Reddit. While spatial data - in the form of the absolute physical locations of users - is often the preferred type of geographic information in social media language research \citep{nguyen_computational_2016}, the treatment of space as a social construction (place) offers greater alignment with existing sociolinguistic theories on language and identity. Counterintuitively, spatial information (absolute physical location) often fails to offer the granularity required to determine nuanced language variation. In the case of New Zealand Reddit as a place, I identified an extensive network of communities connected to \texttt{r/newzealand}. The addition of 32 New Zealand-related communities increased my training data by 20.8\%, which in turn enhanced the temporal granularity of my diachronic embedding models. By utilising \acrshort{C2xG} features, I demonstrated that these New Zealand-related communities share a level of grammatical similarity suggesting that the networked communities form a contiguous speech community. In summary, users and their networks play a crucial role in providing implicit geographic data that transcends mere physical location.

\subsection{Perception-Production Pipeline}

    As an intra-disciplinary study of geographic dialect alignment, I developed an integrated analytical pipeline that incorporates approaches from both variationist and interactionist sociolinguistics, as well as computational linguistics. The key point of difference in this integrated pipeline is that it accounts for the interactional and representational contexts of language data within a corpus. Unsurprisingly, the role of user perception remains an area that is under-explored in current literature (\citealp{rymes_youtube-based_2018}; \citealp{cutler_metapragmatic_2020}; \citealp{kim_i_2025}). With this research gap in mind, the integrated pipeline attempts to reconnect the producers of language with their discourses \citep{matheson_discourse_2023}. I utilised this framework to explore and evaluate language variation and change within a social media platform. In the case of geographic dialect alignment, this perception-production pipeline serves as a vital tool for determining data quality and availability in non-dominant language varieties, such as \acrshort{NZE}.

\subsection{Modelling New Zealand English}
    
    My third and final contribution was the development of a New Zealand Reddit corpus derived from the Pushshift repository \citep{baumgartner_pushshift_2020}. The unprocessed corpus contains 4.26 billion words, spanning a temporal window of 15 years between 2008 and 2023. To contextualise the scale of this resource:

    \begin{itemize}[nolistsep]
        \item The Wellington Corpus of Written New Zealand English contains one million words \citep{bauer_introducing_1994};
        \item The Twitter\textsuperscript{X} subset of the \acrshort{CGLU} \citep{dunn_mapping_2020} contains 7.98 million words as of 2025 \citep{dunn_language_2025};
        \item The Hansard corpus of New Zealand parliamentary debates contains 57 million words \citep{ford_rethinking_2018}.
    \end{itemize}

    This New Zealand Reddit corpus represents a massive expansion in available data and possesses immense potential as a resource for researchers seeking to investigate the complexities of language variation and change within the Aotearoa New Zealand context.

\section{Future Work}
\label{disc:future_work}

    My original motivation for exploring geographic dialect alignment was to address specific research gaps within dialect variation and social media research \citep{nguyen_dialect_2021}. While I have successfully addressed several of these gaps - namely place-informed dialectology, variation and change, and the role of user perceptions - other areas remain ripe for further exploration. In this penultimate section, I discuss future research directions, including both short-term and long-term projects designed to build upon the findings of this thesis. Having already addressed the limitations of my case studies, I will now focus on the promising avenues of inquiry that have emerged throughout the research process.

\subsection{Large Language Models}

   Encoder- and transformer-based large language models were noticeably absent from this research, representing a significant area for future inquiry. A primary limitation of Word2Vec embedding models is that they provide only a static representation of word vectors, meaning the analysis remains limited to the surface forms of words. In contrast, contextualised word embeddings - such as those produced by transformer-based models - allow for the examination of word representations across diverse contexts \citep{vaswani_attention_2017}. This would facilitate a more nuanced investigation of variation and change in polysemous words (for instance, the varying senses of \textit{tramp}). However, the analytical pipeline utilised for static Word2Vec models cannot be simply transposed to pre-trained large language models like \texttt{BERT} \citep{devlin_bert_2019} (see Table \ref{tab:ambiguous_llm}). The computational resources required to fine-tune existing models - let alone train an embedding model specific to \acrshort{NZE} from scratch - pose a substantial barrier. Consequently, further research must involve interrogating the suitability and methodology of using transformer-based models to evaluate linguistic variation and change in non-dominant varieties.

    \begin{table}
        \scriptsize
        \centering
            
            \caption{Performance Metrics for Transformer-based Embedding Models}
            \label{tab:ambiguous_llm}
            
            \renewcommand{\arraystretch}{1.4}
            \begin{tabularx}{\textwidth}{ll*{8}{>{\centering\arraybackslash}X}}
                \toprule
                \multirow{2}{*}{Source} & \multirow{2}{*}{Target} & \multicolumn{2}{c}{\texttt{BERT}} & \multicolumn{2}{c}{\texttt{RoBERTa}} & \multicolumn{2}{c}{\texttt{XLM-RoBERTa}} & \multicolumn{2}{c}{\texttt{GPT-2}} \\
                & & Base & Large & Base & Large & Base & Large & Base & Large \\
                \midrule
                \addlinespace[1em]
                bach & crib & 0.375 & 0.149 & 0.838 & 0.971 & 0.979 & 0.984 & - & - \\
                banger & sausage & 0.334 & 0.413 & 0.887 & 0.977 & 0.982 & 0.950 & 0.965 & 0.317 \\
                beehive & government & 0.212 & 0.385 & 0.829 & 0.960 & 0.962 & 0.956 & - & - \\
                bog & toilet & 0.733 & 0.804 & 0.904 & 0.975 & 0.987 & 0.987 & 0.993 & 0.306 \\
                dairy & superette & 0.395 & 0.413 & 0.851 & 0.966 & 0.974 & 0.983 & 0.981 & 0.195 \\
                fizzy & soda & 0.369 & 0.499 & 0.936 & 0.981 & 0.990 & 0.976 & 0.981 & 0.534 \\
                raro & cordial & 0.369 & 0.129 & 0.875 & 0.978 & 0.992 & 0.989 & 0.984 & 0.313 \\
                snag & sausage & 0.291 & 0.503 & 0.890 & 0.973 & 0.992 & 0.947 & 0.971 & 0.299 \\
                tramp & hike & 0.244 & 0.172 & 0.891 & 0.976 & 0.980 & 0.971 & 0.987 & 0.375 \\
                yarn & conversation & 0.730 & 0.892 & 0.832 & 0.936 & 0.987 & 0.995 & 0.995 & 0.249 \\
                \addlinespace[1em]
                \bottomrule
            \end{tabularx}

        \tablenoteparagraph{\textbf{Table Note}: This table displays the ambiguous word (Source) and its related synonym (Target) alongside cosine similarity measures in \acrshort{NZE}, grouped by model architecture and training data from Reddit/Pushshift \citep{baumgartner_pushshift_2020} and Google300 \citep{mikolov_efficient_2013}. The pre-trained language models utilised include BERT \citep{devlin_bert_2019}, RoBERTa \citep{zhuang_robustly_2021}, XLM-RoBERTa \citep{conneau_unsupervised_2020}, and GPT-2 \citep{radford_language_2019}. High cosine similarity between the source and target prompts does not necessarily indicate semantic similarity, as it may be an artefact of high dimensionality \citep{kgosietsile_cosine_2025}.}
        
    \end{table}

\subsection{New Zealand English Benchmark}

    There is an increasing awareness of the need for rigour in evaluating how \acrshort{NLP} systems perform across different dialects, varieties, and closely-related languages \citep{faisal_dialectbench_2024}. The goal of evaluative frameworks, such as \texttt{dialectbench} \citep{faisal_dialectbench_2024}, is to determine the performance of various language models in language understanding and generation \citep{joshi_natural_2025}. However, existing benchmarks rarely account for the intersection of language variation and change, which remains the primary focus of sociolinguistic inquiry \citep{nguyen_computational_2016}. In the absence of a benchmark designed specifically for \acrshort{NZE}, I utilised user-informed lexical variables and a list of hypocoristic word pairs from Wiktionary to evaluate the Word2Vec embedding models. While these two word lists served their purpose in evaluating the models within this specific context, more intentional development is required to create an \acrshort{NLP} benchmark that is truly fit-for-purpose for the \acrshort{NZE} context. Such a benchmark would need to account for both stable dialectal features and the diachronic semantic shifts identified in this research.

\subsection{Other Platforms and Modalities}

    While I restricted this analysis to Reddit to control for register, it is likely that the findings are generalisable across other social media platforms. For instance, text classification as a form of dialect modelling has performed poorly across both Reddit and Twitter\textsuperscript{X} due to a shared reliance on named entities \citep{dunn_stability_2022}. Crucially, the objective of this research was not to reinforce the concept of ``platform exceptionalism'' \citep[p.10]{panek_understanding_2022}. On the contrary, I aimed to demonstrate that meaningful patterns of language variation and change are observable within their social contexts, even under specific machine constraints. Current scholarship calls for a shift beyond single-source data to consider how linguistic patterns manifest across multiple platforms, thereby addressing questions of generalisability \citep{nguyen_dialect_2021}. My initial attempts to compare and contrast production phenomena between Twitter\textsuperscript{X} and Reddit using the \acrshort{CGLU} \citep{dunn_mapping_2020} proved inconclusive, primarily due to the stark register differences between the two platforms. Consequently, future research should explore methodologies to compare and contrast disparate sources of social media language data while rigorously controlling for corpus characteristics such as register \citep{biber_register_2009}.

\subsection{Origins of Written New Zealand English}

    The findings from my research phases demonstrate that language embedding models, such as Word2Vec, provide an effective approach for analysing language variation and change in large corpora. Just as linguists have traced the phonological development of \acrshort{NZE} through historical recordings \citep{gordon_new_2004}, there is immense potential to utilise language embeddings to expand upon existing research into the development of written \acrshort{NZE} \citep{bauer_introducing_1994}. A particularly promising avenue lies in applying these models to historical repositories. For instance, Papers Past\footnote{\href{https://paperspast.natlib.govt.nz/}{https://paperspast.natlib.govt.nz/}} contains over 105 million articles published in New Zealand dating back to 1861, including over a million documents from books, letters, diaries, magazines, and newspapers. Furthermore, the Parliamentary Debates (Hansard) repository, dating back to 1867, comprises over 500 million words. Mapping these datasets through the perception-production pipeline would allow for a comprehensive diachronic analysis of \acrshort{NZE}, tracing its evolution from colonial origins to its contemporary digital forms.

\section{Final Remarks}
\label{disc:final_remarks}

    In Chapter \ref{chap:introduction}: Introduction, I conceptualised geographic dialect alignment as the components required to produce an ideal language sample representing an underlying population. However, the findings of this research suggest that the `underlying population' is not a monolith. When we ask what features speakers of \acrshort{NZE} expect to see in language models, we must look beyond lexical and grammatical markers. As Ranginui Walker (Whakatōhea) suggested, while the term New Zealand serves as a useful shorthand, it does not simply equate to the complex tapestry of both the land and the people associated with it \citep{ballantyne_place_2011}. \acrshort{NZE} is more than a regional variety; it is an expression of the social, cultural, and political contexts it encompasses. As with any form of inquiry, answering one question inevitably leads to another. I conclude this thesis with two guiding questions that must inform the future of our discipline: What is \acrshort{NZE}, and who is New Zealand?

% -----------------------------
% References
% -----------------------------

\fancyhf{}
\fancyfoot[C]{\thepage}

\addcontentsline{toc}{chapter}{\protect\numberline{References}}
\renewcommand{\refname}{References}
\bibliographystyle{apalike}
\bibliography{references}

% -----------------------------
% References
% -----------------------------

\appendix

\chapter{New Zealand English Features}
\label{app:nze}

    The purpose of this section is to provide a comprehensive overview for readers unfamiliar with \acrshort{NZE}. In the absence of phonological and prosodic features in written language, I focus on the lexical and grammatical features that distinguish \acrshort{NZE} from other varieties.

\section{Lexis}

    To provide an overview of the various lexical features in \acrshort{NZE}, the following sections describe the key word-formation processes identified in the data.

\subsection{Abbreviations and Acronyms}

    An abbreviation is the reduced form of a word, while an acronym, also known as an initialism, consists of the first letter of one or more words \citep{cannon_abbreviations_1989}. An example of an abbreviation is \textit{\gls{chch}} \citep{deverson_chch_2005}. Some abbreviations and acronyms may develop into common nouns and become grammatically productive \citep{cannon_abbreviations_1989}. As an example, \textit{\gls{jafa}} \citep{deverson_jafa_2005}, a pejorative term for residents of Auckland, can take on the regular plural (\textit{\glspl{jafa}}).

\subsection{Analogy}

    Analogy refers to the pattern of linguistic regularisation whereby a new word, or neologism, is coined by modelling on an existing word or linguistic structure \citep{mattiello_analogical_2016}. One example is the second-person plural \textit{\gls{youse}} \citep{orsman_new_1994}, which differs in usage from other varieties of English and is often associated with Māori English \citep{quinn_variation_1995}. Other examples include \textit{\gls{blockholder}}, which originated from the term \textit{freeholder} through semantic analogy \citep{oxford_university_press_blockholder_2023}. Neologisms formed through phonological analogy are more frequently observed, as seen in \textit{\gls{chur}} \citep{oxford_university_press_chur_2023} and \textit{\gls{new_zild}} \citep{orsman_new_1994}; these are eye-dialect forms used to reflect contemporary \acrshort{NZE} phonology.

\subsection{Back-formations, Blending, and Clipping}

    Back-formations are a type of clipping (see below), though it is a relatively infrequent word-formation process \citep{cannon_back-formations_1986}. More specifically, a back-formation may change word classes, such as from noun to verb as in \textit{\gls{bludger}} \citep{orsman_new_1994} to \textit{\gls{bludge}} \citep{orsman_new_1994}. In some cases, back-formations may also transform words within classes, such as from proper noun to common noun as in \textit{\gls{grundies}} \citep{deverson_grundies_2005}.

    Blending involves the telescoping of two or more independent lexical forms \citep{cannon_blends_1986}. One example is \textit{\gls{jandal}} \citep{orsman_new_1994}, which is a blend of \textit{Japanese} and \textit{sandal}. Originally a trademarked name, \textit{jandal} is now the common noun for that form of footwear. Of interest to New Zealand social media is \textit{\gls{marmageddon}} \citep{cuming_marmageddon_2013}, which was coined during the national shortage of the savoury yeast spread from 2012 to 2013. The neologism is a blend of \textit{Marmite} and \textit{Armageddon}.
    
    Clipping, also known as shortening or truncation, is a word-formation process that removes components of a word to produce a shorter word \citep{jamet_morphophonological_2009}. Clippings can be formed by simply removing syllables or speech sounds from a word, such as from utility vehicle to \textit{\gls{ute}} \citep{deverson_ute_2005}. A common greeting associated with Australian and \acrshort{NZE}es is \textit{\gls{gday}} \citep{orsman_new_1994}, variably rendered as \textit{Gidday}, which is a clipping of \textit{good day}.

\subsection{Borrowing and Calques}
\label{lexis:borrowings}

    \cite{haugen_analysis_1950} defines borrowing as ``the attempted reproduction in one language of patterns previously found in another''. The development of borrowings is a form of contact-induced change \citep{winford_contact_2010}. Some words borrowed from other languages into \acrshort{NZE} include \textit{\gls{berm}} \citep{orsman_new_1994} and the fruit \textit{feijoa} \citep{deverson_feijoa_2005} from Brazilian Portuguese. Another form of borrowing is the calque \citep{haugen_analysis_1950}. Examples include terms with origins in the British Empire, such as \textit{\gls{mufti}} \citep{orsman_new_1994} from Arabic. Some food-related borrowings and calques include \textit{\gls{kransky}} \citep{stevenson_kransky_2010} from Slovene (\textit{Kranjska klobasa}) and \textit{\gls{saveloy}} \citep{stevenson_saveloy_2010} from French (\textit{cervelas}).

\subsection{Compounding, Phrasal Verbs, and Listemes}
\label{lexis:compounding}

    Compounding occurs when two or more words combine to form a new word. \cite{lieber_english_2005} distinguishes two types of compounding: synthetic and root compounds. Synthetic compounds (also known as verbal, deverbal, or verbal nexus compounds) are compounds where the second stem is derived from a verb, whereas root compounds do not have this parameter. An example of this is \textit{\gls{bushwalk}} \citep{deverson_bush_2005}, where the noun modifies a verb. In contrast, a root compound consists of a noun, or nouns, modified by an adjective, as in \textit{\gls{longdrop}} \citep{orsman_new_1994}, or by another noun, as in \textit{\gls{gumboot}} \citep{pollock_tea_2013}, among other combinations. Orthographically, compounds are variably represented with a space, a hyphen, or no space, as observed with \textit{\gls{greenstone}} \citep{orsman_new_1994} in the forms \textit{green stone}, \textit{green-stone}, and \textit{greenstone}.
    
    \citet{di_sciullo_definition_1987} described these list-like syntactic units (such as phrasal verbs, idiomatic phrases, proverbs, quotations, and other syntactic objects such as interjections) as \textit{listemes}. I illustrate this point with the phrasal verb \textit{\gls{pack_a_sad}} \citep{orsman_new_1994}, which can be inflected for number and tense, as in \textit{packs a sad} (third-person singular), \textit{packed a sad} (past tense), and \textit{packing a sad} (present participle). Although idiomatic phrases are not themselves part of the lexis in the sense that they do not function as a single word-level unit, they contribute to the lexicon, such as \textit{\gls{box_of_birds}} \citep[p.31]{orsman_new_1994}.

\subsection{Conversion and Extension}

    Conversion occurs when a word shifts to a different semantic meaning without derivational affixes, also known as zero-derivation \citep{cannon_functional_1985}. I can observe this process in words like \textit{\gls{rarked_up}} \citep{stevenson_rark_2010}, which shifts from noun to adjective, or from adjective to noun. Another common shift is from a phrasal verb to a noun, as seen in \textit{\gls{goashore}} \citep[p.108]{orsman_new_1994}. A more specific form of conversion is functional shift, which occurs when a word shifts syntactic function without changing its internal structure. A related process is extension, which occurs when the semantic meaning of a word broadens or expands beyond its original definition \citep{yan_lexical_2019}. This is particularly prevalent in words related to agriculture and the natural environment.

\subsection{Hypocoristics}
\label{lexis:hypocoristics}

    One process which deserves special attention is the use of hypocoristics in \acrshort{NZE}. Hypocorism, a form of affixation, is a highly productive word-formation strategy in the varieties of English in Australia \citep{simpson_hypocoristics_2008} and New Zealand \citep{bardsley_hypocoristics_2009}. Common nouns, personal names, and place names are often clipped to a one-syllable base form followed by a vowel. \cite{kiesling_english_2020} claimed that the presence of hypocoristics is the distinguishing feature between Australian English and \acrshort{NZE}; however, \cite{bardsley_hypocoristics_2009} noted that this is indeed a shared feature between the two varieties, with over 1,150 instances of hypocoristics documented by the New Zealand Dictionary Centre in \acrshort{NZE} alone.

\subsection{Other Processes}

    I conclude this section with two more word-formation processes that have contributed to the development of \acrshort{NZE} lexis. These are:

\subsubsection{Onomatopoeia}

    Onomatopoeia refers to the process where a word resembles the sound of its source. \cite{fandrych_non-morphematic_2004} described onomatopoeia as a non-morphematic word-formation process. While there are few examples of this process unique to \acrshort{NZE}, notable instances include \textit{\gls{morepork}} \citep[p.172]{orsman_new_1994}, which originates from the bird’s distinctive call, and \textit{\gls{wop_wops}}, which is said to mimic the sound of a helicopter heard in remote areas of New Zealand \citep{orsman_new_1994}.

\subsubsection{Proper Nouns}

    Proper nouns may refer to personal names and place names. As of March 2025, there were 54,576 place names associated with 121 different geographic feature types \citep{nga_pou_taunaha_o_aotearoa__new_zealand_geographic_board_new_2025}. Of those geographic feature types, only a few names are associated with cities ($n=19$) and towns ($n=163$). Proper nouns may also include organisations and brand names. These proper nouns may, in turn, be adopted as common nouns through a combination of the processes discussed in the sections above.

\subsubsection{Eponyms}

    An eponym refers to neologisms derived from a personal name, as well as any proper noun that has become a common noun \citep{lalic_eponyms_2004}.

\section{Morphosyntactic Features}

    With reference to this literature, I now describe some of the morphological, syntactic, and miscellaneous grammatical features associated with \acrshort{NZE}.

\subsection{Morphological Features}

    I begin with morphological features. \citet{hundt_new_1998} selected three features to illustrate morphological variability in \acrshort{NZE}: adjectival and adverbial morphemes, irregular verbs, and the \textit{s}-genitive. While the formal variants of these morphological variables are clearly defined, it is also necessary to account for their grammatical function. For example, the irregular \textit{t}-forms of \textit{burn}, \textit{learn}, and \textit{dream} may appear as the past tense, the past participle, or the adjectival form \citep{quirk_comprehensive_1985}. Where appropriate, I have included observations made by Bauer (\citeyear{bauer_new_1987}; \citeyear{bauer_grammatical_2007}), Quinn (\citeyear{quinn_variation_1995}; \citeyear{quinn_variation_2000}), and \citet{hay_new_2008}.
    
\subsubsection{Adjectival and Adverbial Morphemes}

   My first morphological feature of interest is adjectival and adverbial morphemes. \cite{hundt_new_1998} observed diachronic and stylistic variation, but not regional variation, in adjectives in \acrshort{NZE}. In standardised varieties of English, adjectives made up of two or more syllables (e.g. \textit{unhappy}) generally use a periphrastic construction to form the comparative (\textit{more unhappy}) or the superlative (\textit{most unhappy}). In contrast, the comparative and superlative forms of monosyllabic adjectives are typically inflected with the suffixes \textit{-er} and \textit{-est} to form the respective constructions (e.g. \textit{cold}, \textit{colder}, \textit{coldest}) \citep{quirk_comprehensive_1985}. In the case of disyllabic adjectives such as \textit{easy}, both periphrastic and inflectional constructions are grammatical (\textit{more easy}, \textit{easier}; \textit{most easy}, \textit{easiest}), though there is often a preference for the inflectional form in certain contexts \citep{barber_english_1993}. Based on corpus data from the \textit{New York Times} and \textit{The Times}, \cite{bauer_watching_1994} observed the following rule:
    
    \begin{quote}
        \footnotesize
        Disyllabic adjectives which end in the suffix \textit{-ly} take periphrastic comparison, other adjectives ending in \textit{-y} and also those ending in syllabic \textit{-le} take suffixed comparison, all others take periphrastic comparison except for a few remnants.
    \end{quote}
    
    When \cite{bauer_watching_1994} extended his analysis to the Wellington Corpus of Written \acrshort{NZE}, he found that 92\% of 564 tokens of comparative disyllabic adjective constructions (with the exclusion of adjectives ending in \textit{-ed} or \textit{-ing}) fit the rule prescribed above. Of the exceptions, the disyllabic adjectives \textit{narrow}, \textit{quiet}, and \textit{shallow} took only the inflectional suffix. \cite{quinn_variation_1995} noted that a third variant—a combined periphrastic and inflectional construction, also referred to as the double comparative—is a viable variant. While considered ungrammatical in standardised varieties of English, the double comparative construction is primarily observed in the most colloquial registers of spoken English \citep{smitterberg_adjective_2006}. I illustrate these variants with the disyllabic adjective \textit{gentle} taken from \cite{quinn_variation_2000}, where the inflectional (\ref{syn:gentle}a), periphrastic (\ref{syn:gentle}b), and double comparative (\ref{syn:gentle}c) variants can be observed.

    \vspace{-12pt}
    \begin{center}
    \singlespacing
    \begin{align}
        \textnormal{Mono-, Disyllabic, and Double Comparatives} \label{syn:gentle} \\
        \notag \\
        \textnormal{\texttt{gentler}} \tag{a} \\
        \textnormal{\texttt{more gentle}} \tag{b} \\
        \textnormal{\texttt{more gentler}} \tag{c} \\
        \notag \\
        \eqname{\citep{quinn_variation_2000}}
    \end{align}
    \end{center}

    \cite{quinn_variation_1995} found high acceptability ratings for the double comparative construction amongst her informants. The double comparative construction can also be observed in some monosyllabic adjectives \citep{hundt_new_1998}. Curiously, this three-way contrast is not observed in monosyllabic adjectives such as \textit{nice}, where the periphrastic construction, *\textit{more nice}, would be ungrammatical in \acrshort{NZE}. \cite{bauer_grammatical_2007} noted that in restricted environments, unmarked adverbs preceding an adjective are grammatical in \acrshort{NZE}. This means both examples (\ref{syn:adverb}a) and (\ref{syn:adverb}b) are grammatical, whereas only (\ref{syn:adverb}a) would be grammatical in standardised varieties of English \citep{quirk_comprehensive_1985}. While \cite{hickey_legacies_2005} identified unmarked adverbs, such as \textit{awful} in \textit{He's awful busy these days}, as a `non-standard feature' in various Englishes, this specific construction is seldom observed in \acrshort{NZE}, suggesting that these are distinct processes.

    \vspace{-12pt}
    \begin{center}
    \singlespacing
    \begin{align}
        \textnormal{Marked and Unmarked Adverbs} \label{syn:adverb} \\
        \notag \\
        \textnormal{\texttt{It was \textit{really} funny}} \tag{a} \\
        \textnormal{\texttt{It was \textit{real} funny}} \tag{b} \\
        \notag \\
        \eqname{\citep{bauer_grammatical_2007}}
    \end{align}
    \end{center}

\subsubsection{Irregular Verbs} 

    \begin{table}
        \scriptsize
        \centering
            
            \caption{Distribution of Irregular Past-tense Forms in English}
            \label{tab:irregular}
            
            \renewcommand{\arraystretch}{1.4}
             \begin{tabularx}{\linewidth}{l*{3}{>{\centering\arraybackslash}X}}
            \toprule
            \textbf{Verb} & \textbf{DOM/EVP} & \textbf{Miami Herald} & \textbf{Guardian}\\
            \midrule
                \addlinespace[1em]
                \textit{burnt} & 45.0 & 5.0 & 44.0 \\ 
                \textit{burned} & 55.0 & 95.0 & 56.0 \\ 
                \addlinespace[1em]
                \textit{learnt} & 25.0 & - & 22.0 \\ 
                \textit{learned} & 75.0 & 100.0 & 78.0 \\ 
                \addlinespace[1em]
                \textit{dreamt} & 20.0 & 5.0 & 31.0 \\ 
                \textit{dreamed} & 80.0 & 95.0 & 69.0 \\ 
                \addlinespace[1em]
            \bottomrule
            \end{tabularx}

            \tablenoteparagraph{\textbf{Table Notes}: This table compares the distribution of regular and irregular past tense forms for the root verbs \textit{burn}, \textit{learn}, and \textit{dream} across three newspaper corpora representing distinct varieties of written English: the Dominion Post/Evening Post (\texttt{DOM/EVP}) for \acrshort{NZE}, the Miami Herald (\texttt{Miami Herald}) for American English, and the Guardian (\texttt{Guardian}) for British English. Rows group variants by root verb, with percentages for each corpus-verb grouping summing to 100.0\%. The data, sourced from \citet{hundt_new_1998}, reveals that the distributional patterns for \acrshort{NZE} align more closely with British English than with American English.}
        
    \end{table}

    The past tense and past participle of regular verbs in most varieties of English are formed by the suffix \textit{-ed}. However, there remain approximately 250 irregular verbs that do not follow this generalised pattern \citep{quirk_comprehensive_1985}. These irregular verbs have been grouped into seven classes based on specific characteristics, including the type of suffixation, the form of the past tense ($V$-$ed_1$) and past participle ($V$-$ed_2$), and vowel identity. For the purpose of my analysis, I focus on Class 1 to Class 3, where the greatest variability in past participle forms is observed.
    
    \cite{quirk_comprehensive_1985} noted that the \textit{t}-form is rarely observed in American English, while it is often preferred in British English. \cite{bauer_new_1987} examined this in the irregular past tense forms of \textit{burn}, \textit{learn}, \textit{smell}, \textit{spell}, \textit{spill}, and \textit{spoil}. The distribution of the three most frequent irregular past tense verbs (\textit{burn}, \textit{learn}, and \textit{dream}) from three newspapers representative of three varieties of English (New Zealand, General American, and British) is presented in Table \ref{tab:irregular}. The data for the \textit{Miami Herald} and the \textit{Guardian} were sourced from \citet{gloderer_morphological_1993}, as cited by \citet{hundt_new_1998}.
    
    Of special note is the Class 2 irregular verb \textit{prove}, where the \textit{n}-form of the past participle is the preferred form in American English usage, in contrast to the \textit{ed}-form, which is more common in British English. \citet{gordon_word_1980} noted that ``the past participle of `prove' is `proved', as is its past tense. Do NOT use the form `proven', which can be used only in the specialised phrase `not proven'.'' While this view does not reflect actual usage in \acrshort{NZE} \citep{bauer_new_1987}, it offers insight into early attempts to ‘standardise’ the grammar of \acrshort{NZE}. 

\subsubsection{\textit{s}-Genitive}

    \begin{table}
        \scriptsize
        \centering
            
            \caption{Distribution of \textit{s}-Genitive in English}
            \label{tab:s_genitive}
            
            \renewcommand{\arraystretch}{1.4}
            \begin{tabularx}{\textwidth}{l*{6}{>{\centering\arraybackslash}X}}
            \toprule
            \textbf{Noun Class} & \textbf{WCNZE} & \textbf{LOB} & \textbf{FLOB} & \textbf{Brown} & \textbf{Frown} & \textbf{ACE}\\
            \midrule
            \addlinespace[1em]
            Personal names & 27.0 & 38.0 & 40.0 & 37.0 & 38.0 & 31.0 \\ 
            Personal nouns & 19.0 & 22.0 & 14.0 & 19.0 & 16.0 & 18.0 \\ 
            Collective nouns & 21.0 & 15.0 & 18.0 & 15.0 & 16.0 & 17.0 \\ 
            (Higher) animals & 1.0 & 0.4 & 0.5 & 0.5 & 0.05 & 0.4 \\ 
            Geographic nouns & 18.0 & 14.0 & 16.0 & 16.0 & 17.0 & 21.0 \\ 
            Temporal nouns & 8.0 & 7.0 & 7.0 & 6.5 & 5.0 & 6.0 \\ 
            Other nouns & 6.0 & 3.0 & 4.5 & 6.0 & 8.0 & 6.0 \\ 
            \addlinespace[1em]
            \bottomrule
            \end{tabularx}

            \tablenoteparagraph{\textbf{Table Notes}: This table contrasts the distribution of the $s$-genitive across different noun classes (\texttt{Noun Class}) using data from six written corpora representing various English varieties: the Wellington Corpus of \acrshort{NZE} (\texttt{WCNZE}), the Lancaster-Oslo-Bergen Corpus (\texttt{LOB}), the Freiburg-LOB Corpus of British English (\texttt{FLB}), the Brown Corpus (\texttt{Brown}), the Freiburg-Brown Corpus of American English (\texttt{Frown}), and the Australian Corpus of English (\texttt{ACE}). Rows provide the percentage distribution within each corpus grouping, summing to 100.0\%. Sourced from \citet{hundt_new_1998}, the data indicates that \acrshort{NZE} (\texttt{WCNZE}) is characterised by a higher proportion of $s$-genitive forms in collective nouns, displaying a distributional pattern that aligns more closely with Australian English (\texttt{ACE}) than with other varieties.}
        
    \end{table}

    The \textit{s}-genitive construction in English, also known as the Saxon genitive, stems from the Germanic prototypical inflectional genitive morpheme. \cite{hundt_new_1998} noted that there is regional and stylistic variation in the use of the synthetic \textit{s}-genitive compared with the analytic \textit{of}-phrase construction. The greatest predictor for the choice between these two variants is the semantic category of the head noun phrase.

\subsubsection{Other Inflectional Morphemes}

    In addition to the three morphological variables discussed above, \cite{bauer_new_1987} noted that the plural forms of \textit{roof} and \textit{wharf} in \acrshort{NZE} are typically \textit{rooves} and \textit{wharves}, in contrast to the \textit{roofs} and \textit{wharfs} observed in standard forms in England. The plural form of \textit{hoof} is variably \textit{hoofs} or \textit{hooves}. Another notable aspect of \acrshort{NZE} morphology is that the plural forms of words borrowed from te reo Māori often remain unmarked \citep{bauer_grammatical_2007}.

\subsection{Syntactic Features}

    Of the grammatical phenomena discussed in \citet{hundt_new_1998}, syntactic variants were the least straightforward to define from a variationist sociolinguistic perspective. Unlike morphological variables where variation is clearly defined (such as irregular verbs), syntactic analysis must instead rely on functional comparability. An example of this is the English dative construction, as described in \ref{syn:dative}. \citet{hundt_new_1998} identified eight syntactic phenomena of special interest to \acrshort{NZE}, but I concentrate my description on three features: aspect, agreement, and the mandative subjunctive. Once again, I have included observations made by Bauer (\citeyear{bauer_new_1987}; \citeyear{bauer_grammatical_2007}), Quinn (\citeyear{quinn_variation_1995}; \citeyear{quinn_variation_2000}), and \citet{hay_new_2008} where appropriate.

    \vspace{-12pt}
    \begin{center}
    \singlespacing
    \begin{align}
        \textnormal{Functionally Comparable Dative Constructions} \label{syn:dative} \\
        \notag \\
        \textnormal{\texttt{Thelma gave Louise money}} \tag{a} \\
        \textnormal{\texttt{Thelma gave money to Louise}} \tag{b}
    \end{align}
    \end{center}
    
\subsubsection{Agreement of Collective Nouns}

    Collective nouns (\textit{committee}, \textit{team}, \textit{government}, \textit{army}, \textit{family}), also known as corporate nouns, are grammatically singular. In standardised registers of English, singular agreement is preferred when the collective noun is viewed as a unit and plural agreement when the collective noun is viewed as a number of individuals \citep{bauer_grammatical_2007}. Neither \cite{hundt_new_1998} nor \cite{bauer_grammatical_2007} discussed the patterns of agreement in collective nouns borrowed from te reo Māori, such as \textit{\gls{whanau_mao}} or \textit{\gls{iwi}}. Variable agreement can also be observed in the copula following \textit{there} in \acrshort{NZE} \citep{bauer_grammatical_2007}. In standardised registers of English, \textit{there is} precedes a singular noun phrase and \textit{there are} precedes a plural noun phrase. Both \textit{there is}, and the contracted form \textit{there's}, are invariably observed before singular and plural noun phrases with a preference for the contracted form in spoken language \citep{bauer_grammatical_2007}.

\subsubsection{Aspect}

    With reference to the examples in (\ref{aspect:1}), \citet{bauer_grammatical_2007} found that innovative aspect-markings found in other varieties of English are notably absent from \acrshort{NZE}. \citet{hundt_new_1998} observed an increased usage of the progressive aspect as a feature of colonial varieties of English except for American English. In addition to these observations, there is an increasing frequency of \textit{be going to} in place of will \citep{hay_new_2008}, \textit{going to} is contracted to \textit{gonna}.

\subsubsection{Mandative Subjunctive}

    \begin{table}
        \scriptsize
        \centering
            
            \caption{Distribution of the Mandative Subjunctive in English}
            \label{tab:mandative_subjunctive}
            
            \renewcommand{\arraystretch}{1.4}
             \begin{tabularx}{\linewidth}{l*{3}{>{\centering\arraybackslash}X}}
            \toprule
            \textbf{Variants} & \textbf{DOM/EVP} & \textbf{Guardian} & \textbf{Miami Herald}\\
            \midrule
            \addlinespace[1em]
            Mandative Subjunctive & 70.0 & 88.0 & 35.0 \\ 
            \textit{should} & 21.0 & 8.0 & 55.0 \\ 
            Indicative & 9.0 & 4.0 & 10.0 \\ 
            \addlinespace[1em]
            \bottomrule
            \end{tabularx}

            \tablenoteparagraph{\textbf{Table Notes}: This table compares the distribution of mandative subjunctive variants (\texttt{Variants}) across three newspaper sources representing distinct varieties of written English: the Dominion Post/Evening Post (\texttt{DOM/EVP}) for \acrshort{NZE}, the Miami Herald (\texttt{Miami Herald}) for American English, and the Guardian (\texttt{Guardian}) for British English. Rows provide the percentage distribution for each corpus grouping, with each summing to 100.0\%. Sourced from \citet{hundt_new_1998}, the data indicates that the distributional patterns for \acrshort{NZE} align more closely with American English than with British English.}
        
    \end{table}

    One example is where in subordinate clauses after expressions of demand, recommendation, or intention, that there is a preference in \acrshort{NZE} for the mandative subjunctive (\ref{subj}a), while the periphrastic \textit{should} followed by an infinitive (\ref{subj}b) (the \textit{should} variant) or the indicative (\ref{subj}c) are preferred in British English where the mandative subjunctive is the formal variant \citet{hundt_new_1998}. This follows a similar pattern to American English as shown in Table \ref{tab:mandative_subjunctive} where the mandative subjunctive is preferred to the other variants. When \citet{hundt_new_1998} compared the distribution of the mandative subjunctive and the \textit{should} construction between \acrshort{NZE}, Australian English, and American English, they found a higher occurrence of the mandative subjunctive in  Australian English (77.7\%) than \acrshort{NZE} (66.7\%). The distribution of the mandative subjunctive suggest that both Australian and \acrshort{NZE}es pattern more closely with American English than British English.

    \vspace{-12pt}
    \begin{center}
    \singlespacing
    \begin{align}
        \textnormal{Mandative Subjunctive Constructions} \label{subj} \\
        \notag \\
        \textnormal{\texttt{I propose that he \textit{talk} to a specialist}} \tag{a} \\ 
        \textnormal{\texttt{I propose that he \textit{should talk} to a specialist}} \tag{b} \\
        \textnormal{\texttt{I propose that he \textit{talks} to a specialist}} \tag{c} \\
        \notag \\
        \eqname{\citep{hundt_new_1998}}
    \end{align}
    \end{center}

\subsubsection{Modals and Marginal Modals}

    \citet{trudgill_international_2017} noted that there appears to be no clear distinction between \textit{shall} and \textit{will} in \acrshort{NZE}. The use of \textit{of} as a marginal modal, henceforth modal-\textit{of}, is a noticeable development in \acrshort{NZE}. This stems from the reanalysis of the contracted form of \textit{have} to \textit{of} as in \textit{should've} to \textit{should of} \citep{hay_new_2008}. Further evidence of this reanalysis is presented in \ref{syn:negation}a and \ref{syn:negation}b where I observe the modal-\textit{of} \citep{hay_new_2008}.

    \vspace{-12pt}
    \begin{center}
    \singlespacing
    \begin{align}
        \textnormal{`\textit{of}' Reanalysis} \label{syn:negation} \\
        \notag \\
        \textnormal{\texttt{Nigel should of not done that}} \tag{a} \\ 
        \textnormal{\texttt{What would of she done}} \tag{b} \\
        \notag \\
        \eqname{\citep{hay_new_2008}}
    \end{align}
    \end{center}

\subsubsection{\textit{for-to}-Infinitive Construction}

    \citet{hundt_new_1998} noted one syntactic construction in \acrshort{NZE} which patterns more closely with the American English variant is the \textit{for-to}-infinitive construction. As I illustrate in \ref{syn:for_to}, prepositional verbs with \textit{for} (such as \textit{arrange for}, \textit{call for}, and \textit{wait for}) may take an infinitive clause \ref{syn:for_to}a in American English, whereas as British English would use a fused infinitive (also known as raising construction in \ref{syn:for_to}b) instead. Examples of phrasal verbs following the \textit{for-to}-infinitive construction in \acrshort{NZE} include \textit{appeal for}, \textit{argue for}, and \textit{vote for}. 

    \vspace{-12pt}
    \begin{center}
    \singlespacing
    \begin{align}
        \textnormal{\textit{for-to}-Infinitive Construction} \label{syn:for_to} \\
        \notag \\
        \textnormal{\texttt{I would like for you to go on holiday}} \tag{a} \\ 
        \textnormal{\texttt{I would like you to go on holiday}} \tag{b} \\
        \notag \\
        \eqname{\citep{hundt_new_1998}}
    \end{align}
    \end{center}

\subsection{Other Grammatical Features}

    In this section, I describe other grammatical features not discussed in \citet{hundt_new_1998}. It is important to note that these features are not exclusive to \acrshort{NZE}. Instead, these grammatical features offer me a view of \acrshort{NZE} as a composite of these disparate constructions. I now focus my description on the following features of interest to \acrshort{NZE}. These include discourse particle \textit{eh}, intensifier \textit{as}, negative \textit{ought}, personal pronouns, preposition adverbials, and relative clauses.

\subsubsection{Intensifier `as'} 
\label{review:intensifier_as}

    The use of \textit{as} as a post-modifying construction of an adjective (\textsc{adj}+\textit{as}) is a feature that is distinct to Australian English and \acrshort{NZE} \citep{sowa_sweet_2009}.

\subsubsection{Negative `\textit{ought}'}
    
    \acrshort{NZE} patterns closely to other varieties of terms of negation and negative contraction with one notable exception in the modal verb \textit{ought} where the negative and negative contracted form are relatively uncommon \citep{bauer_grammatical_2007}.

\subsubsection{Personal Pronouns} 

    While there is a preference for the second-person plural to remain unmarked in \acrshort{NZE} \citep{bauer_new_1987}, the marked variant (\textit{youse} or \textit{yous}) can be observed and most widespread in Māori English \citep{stubbe_talking_2000}.

\subsubsection{Prepositional Adverbials} 

     The co-occurrence of prepositional adverbials co-occur with the verb \textit{to be} is a distinct feature of \acrshort{NZE} (and Scots) as illustrated in (\ref{phrasal_verb}a), (\ref{phrasal_verb}b), and (\ref{phrasal_verb}c) \citep{hickey_legacies_2005}. Despite being described as phrasal verbs (as in \textit{bugger up} and \textit{deal to}) by \citet{hickey_legacies_2005}, these prepositional adverbials are not true phrasal verbs.

    \vspace{-12pt}
    \begin{center}
    \singlespacing
    \begin{align}
        \textnormal{Propositional Verbs} \label{phrasal_verb} \\
        \notag \\
        \textnormal{\texttt{I'm \textit{off} to town}} \tag{a} \\ 
        \textnormal{\texttt{He's \textit{down} the pub}} \tag{b} \\
        \textnormal{\texttt{He's \textit{out} the door}} \tag{c} \\
        \notag \\
        \eqname{\citep{hickey_legacies_2005}}
    \end{align}
    \end{center}
    
\subsubsection{Relative Clauses}

    \citet{sigley_choosing_1997} noted that relative pronoun choice in \acrshort{NZE} was not vastly different from American or British English. The general exception to this rule is that \textit{which} is dispreferred in restrictive clauses. There is a preference for \textit{which} when the relative clause is referring to a human.

\chapter{Miscellaneous Figures and Tables}
\label{app:misc}

    \begin{table}
        \scriptsize
        \renewcommand{\arraystretch}{1.4}
        \centering
        \caption{\label{tab:country_ttr} Corpus Dimensions of Country-Level Communities}
        \begin{tabular}{lcccccccc}
        \hline
        \multicolumn{1}{c}{} & \multicolumn{2}{c}{rpost} & \multicolumn{2}{c}{rstitle} & \multicolumn{2}{c}{rstext} & \multicolumn{2}{c}{rcomm} \\
        \multicolumn{1}{c}{} & \textit{n} & \textsc{ttr}  & \textit{n} & \textsc{ttr} & \textit{n} & \textsc{ttr} & \textit{n} & \textsc{ttr} \\
        \hline
        \\
        r/canada & 441,965 & 0.097 & 174,600 & 0.109 & 30,678 & 0.029 & 18,461,782 & 0.010 \\
        r/usa & 137,208 & 0.143 & 22,872 & 0.207 & 3,131 & 0.085 & 102,781 & 0.031 \\
        r/ireland & 247,498 & 0.110 & 234,968 & 0.120 & 77,478 & 0.024 & 10,550,629 & 0.015 \\
        r/unitedkingdom & 363,259 & 0.101 & 146,188 & 0.114 & 21,951 & 0.033 & 7,381,335 & 0.011 \\
        r/australia & 390,341 & 0.102 & 197,969 & 0.121 & 39,351 & 0.029 & 12,937,852 & 0.013 \\
        r/newzealand & 161,387 & 0.124 & 136,330 & 0.130 & 50,483 & 0.027 & 8,781,356 & 0.015 \\
        r/Kenya & 16,911 & 0.214 & 24,352 & 0.235 & 12,164 & 0.057 & 475,592 & 0.042 \\
        r/southafrica & 81,221 & 0.156 & 56,548 & 0.169 & 18,796 & 0.041 & 1,563,293 & 0.022 \\
        r/india & 971,301 & 0.092 & 480,930 & 0.097 & 114,554 & 0.023 & 12,454,839 & 0.017 \\
        r/pakistan & 143,366 & 0.141 & 86,625 & 0.148 & 24,409 & 0.041 & 2,393,973 & 0.022 \\
        r/malaysia & 119,919 & 0.141 & 73,520 & 0.156 & 18,546 & 0.040 & 3,457,047 & 0.025 \\
        r/Philippines & 335,025 & 0.126 & 331,219 & 0.134 & 89,933 & 0.031 & 13,678,914 & 0.026 \\
        \\
        \hline
        \end{tabular}
    \end{table} 

    \begin{table}
        \scriptsize
        \renewcommand{\arraystretch}{1.4}
        \centering
        \caption{\label{tab:reddit_ttr} Corpus Dimensions of New Zealand-related Communities}
        \begin{tabular}{lcccccccc}
        \hline
        \multicolumn{1}{c}{} & \multicolumn{2}{c}{rpost} & \multicolumn{2}{c}{rstitle} & \multicolumn{2}{c}{rstext} & \multicolumn{2}{c}{rcomm} \\
        \multicolumn{1}{c}{} & \textit{n} & \textsc{ttr}  & \textit{n} & \textsc{ttr} & \textit{n} & \textsc{ttr} & \textit{n} & \textsc{ttr} \\
        \hline
        \\
        r/Aleague & 40,540 & 0.157 & 23,078 & 0.169 & 13,950 & 0.046 & 1,544,712 & 0.030 \\
        r/allblacks & 1,258 & 0.329 & 808 & 0.404 & 663 & 0.138 & 28,140 & 0.057 \\
        r/antarctica & 3,111 & 0.263 & 3,775 & 0.298 & 2,487 & 0.078 & 46,819 & 0.044 \\
        r/auckland & 24,673 & 0.166 & 66,252 & 0.178 & 43,759 & 0.031 & 2,127,594 & 0.024 \\
        r/aucklandeats & 464 & 0.445 & 3,216 & 0.360 & 3,113 & 0.074 & 66,001 & 0.065 \\
        r/AveragePicsOfNZ & 5,780 & 0.298 & 959 & 0.483 & 519 & 0.399 & 37,467 & 0.117 \\
        r/CasualNZ & 1,069 & 0.469 & 2,735 & 0.103 & 2,559 & 0.079 & 129,274 & 0.044 \\
        r/chch & 6,672 & 0.234 & 19,629 & 0.258 & 14,469 & 0.051 & 423,332 & 0.035 \\
        r/ConservativeKiwi & 19,788 & 0.194 & 11,555 & 0.233 & 8,309 & 0.053 & 723,197 & 0.026 \\
        r/Coronavirus\_NZ & 6,738 & 0.201 & 2,321 & 0.310 & 889 & 0.108 & 118,740 & 0.033 \\
        r/diynz & 3,851 & 0.182 & 13,177 & 0.255 & 10,970 & 0.038 & 204,011 & 0.031 \\
        r/DownUnderTV & 569 & 0.272 & 8,452 & 0.311 & 6,061 & 0.153 & 43,043 & 0.085 \\
        r/dunedin & 1,827 & 0.342 & 6,626 & 0.310 & 3,606 & 0.083 & 73,625 & 0.049 \\
        r/ImagesOfNewZealand & 21,392 & 0.185 & 9 & 0.780 & 3 & 0.596 & 23,439 & 0.036 \\
        r/LegalAdviceNZ & 266 & 0.360 & 10,472 & 0.213 & 10,038 & 0.022 & 167,432 & 0.019 \\
        r/MapsWithoutNZ & 16,994 & 0.202 & 3,512 & 0.298 & 585 & 0.311 & 109,616 & 0.094 \\
        r/newzealand & 188,212 & 0.121 & 199,674 & 0.127 & 98,576 & 0.022 & 12,065,075 & 0.014 \\
        r/NewZealandWildlife & 3,799 & 0.261 & 2,279 & 0.359 & 1,837 & 0.131 & 59,456 & 0.074 \\
        r/NZBitcoin & 1,183 & 0.270 & 4,037 & 0.202 & 1,273 & 0.096 & 33,084 & 0.045 \\
        r/NZcarfix & 107 & 0.523 & 1,584 & 0.374 & 1,558 & 0.082 & 33,239 & 0.052 \\
        r/nzev & 1,323 & 0.307 & 2,027 & 0.353 & 1,855 & 0.078 & 62,340 & 0.041 \\
        r/nzgardening & 1,074 & 0.275 & 3,518 & 0.332 & 3,369 & 0.071 & 48,537 & 0.054 \\
        r/nzgonewild & 79,683 & 0.140 & 27,404 & 0.170 & 166 & 0.380 & 471,200 & 0.084 \\
        r/NZhookups & 14,510 & 0.156 & 44,854 & 0.131 & 15,610 & 0.069 & 92,228 & 0.056 \\
        r/nzpolitics & 2,571 & 0.259 & 1,511 & 0.352 & 1,312 & 0.082 & 67,320 & 0.035 \\
        r/NZTrees & 8,949 & 0.181 & 12,326 & 0.236 & 6,288 & 0.068 & 173,552 & 0.043 \\
        r/PartyParrot & 30,653 & 0.200 & 7,519 & 0.272 & 940 & 0.158 & 292,062 & 0.054 \\
        r/PersonalFinanceNZ & 3,759 & 0.201 & 35,185 & 0.170 & 25,208 & 0.022 & 811,741 & 0.019 \\
        r/sheep & 3,898 & 0.269 & 2,879 & 0.322 & 2,250 & 0.059 & 31,533 & 0.055 \\
        r/Tauranga & 571 & 0.418 & 1,880 & 0.427 & 1,517 & 0.120 & 35,581 & 0.065 \\
        r/thetron & 1,715 & 0.337 & 4,748 & 0.351 & 3,685 & 0.088 & 85,831 & 0.053 \\
        r/universityofauckland & 1,819 & 0.244 & 25,582 & 0.235 & 18,612 & 0.029 & 183,589 & 0.026 \\
        r/Wellington & 20,681 & 0.121 & 50,925 & 0.155 & 30,443 & 0.035 & 1,224,800 & 0.027 \\
        \\
        \hline
        \end{tabular}
    \end{table}

\chapter{Statistical Models}
\label{app:models}

        \begin{table}
        \centering
        \begin{minipage}[b]{\textwidth}
            \begin{subtable}{\textwidth}
                \scriptsize
                \centering
                \subcaption{${lcm}\sim{age}+{score}+{count}$}
                \renewcommand{\arraystretch}{1.4}
                        \begin{tabular}{p{0.4\linewidth}p{0.3\linewidth}}
                        \\
                        \hline
                        \textbf{Dependent variable: lcm} & \\
                        \hline
                        \\
                        Intercept & 0.916$^{***}$ (0.013) \\
                    age & 0.000$^{*}$ (0.000) \\
                    count & 0.000$^{*}$ (0.000) \\
                    score & -0.000$^{***}$ (0.000) \\
                        \\
                    Observations & 65 \\
                    $R^2$ & 0.756 \\
                    Adjusted $R^2$ & 0.744 \\
                    Residual Std. Error & 0.058 (df=61) \\
                    F Statistic & 63.111$^{***}$ (df=3; 61) \\
                        \\
                        \hline 
                        \\
                    \end{tabular}
                \end{subtable}\vspace{6pt}
                \begin{subtable}{\textwidth}
                \scriptsize
                \centering
                \subcaption{\label{tab:ols_lcm2} ${lcm}\sim{score}+{count}$}\vspace{6pt}
                \renewcommand{\arraystretch}{1.4}
                    \begin{tabular}{p{0.4\linewidth}p{0.3\linewidth}}
                        \hline
                        \\
                    Intercept & 0.935$^{***}$ (0.008) \\
                    count & 0.000$^{**}$ (0.000) \\
                    score & -0.000$^{***}$ (0.000) \\
                    \\
                    Observations & 65 \\
                    $R^2$ & 0.742 \\
                    Adjusted $R^2$ & 0.734 \\
                    Residual Std. Error & 0.059 (df=62) \\
                    F Statistic & 89.256$^{***}$ (df=2; 62) \\
                     \\
                    \hline \\
                \end{tabular}
                \end{subtable}\vspace{6pt}
                \begin{subtable}{\textwidth}
                \scriptsize
                \centering
                \subcaption{${lcm}\sim{score}*{count}$}\vspace{6pt}
                \renewcommand{\arraystretch}{1.4}
                    \begin{tabular}{p{0.4\linewidth}p{0.3\linewidth}}
                        \hline
                        \\
                        Intercept & 0.935$^{***}$ (0.008) \\
                    count & 0.000$^{**}$ (0.000) \\
                    score & -0.000$^{***}$  (0.000) \\
                    score:count & -0.000$^{}$ (0.000) \\
                    \\
                    Observations & 65 \\
                    $R^2$ & 0.742 \\
                    Adjusted $R^2$ & 0.730 \\
                    Residual Std. Error & 0.059 (df=61) \\
                    F Statistic & 58.569$^{***}$ (df=3; 61) \\
                     \\
                    \hline \\
                    \textit{Note:} & \multicolumn{1}{r}{$^{*}$p$<$0.1; $^{**}$p$<$0.05; $^{***}$p$<$0.01} \\
                \end{tabular}
                \end{subtable}\vspace{24pt}
                \end{minipage}
        \caption{\label{tab:ols_lcm} \acrshort{OLS} Regression Model for \gls{lex} lifespan cohorts.}
    \end{table}

            \begin{table}
        \centering
        \begin{minipage}[b]{\textwidth}
            \begin{subtable}{\textwidth}
                \scriptsize
                \centering
                \subcaption{${ldm}\sim{age}+{score}+{count}$}
                \renewcommand{\arraystretch}{1.4}
                        \begin{tabular}{p{0.4\linewidth}p{0.3\linewidth}}
                        \\
                        \hline
                        \textbf{Dependent variable: ldm} & \\
                        \hline
                        \\
                        Intercept & 0.924$^{***}$ (0.012) \\
                    age & 0.000$^{*}$ (0.000) \\
                    count & 0.000$^{*}$ (0.000) \\
                    score & -0.000$^{***}$ (0.000) \\
                        \\
                    Observations & 65 \\
                    $R^2$ & 0.778 \\
                    Adjusted $R^2$ & 0.767 \\
                    Residual Std. Error & 0.054 (df=61) \\
                    F Statistic & 71.310$^{***}$ (df=3; 61) \\
                        \\
                        \hline 
                        \\
                    \end{tabular}
                \end{subtable}\vspace{6pt}
                \begin{subtable}{\textwidth}
                \scriptsize
                \centering
                \subcaption{${ldm}\sim{score}+{count}$}\vspace{6pt}
                \renewcommand{\arraystretch}{1.4}
                    \begin{tabular}{p{0.4\linewidth}p{0.3\linewidth}}
                        \hline
                        \\
                    Intercept & 0.943$^{***}$ (0.007) \\
                    count & 0.000$^{**}$ (0.000) \\
                    score & -0.000$^{***}$ (0.000) \\
                        \\
                    Observations & 65 \\
                    $R^2$ & 0.764 \\
                    Adjusted $R^2$ & 0.757 \\
                    Residual Std. Error & 0.055 (df=62) \\
                    F Statistic & 100.612$^{***}$ (df=2; 62) \\
                     \\
                    \hline \\
                \end{tabular}
                \end{subtable}\vspace{6pt}
                \begin{subtable}{\textwidth}
                \scriptsize
                \centering
                \subcaption{${ldm}\sim{score}*{count}$}\vspace{6pt}
                \renewcommand{\arraystretch}{1.4}
                    \begin{tabular}{p{0.4\linewidth}p{0.3\linewidth}}
                        \hline
                        \\
                        Intercept & 0.942$^{***}$ (0.007) \\
                    count & 0.000$^{**}$ (0.000) \\
                    score & -0.000$^{***}$ (0.000) \\
                    score:count & -0.000$^{}$ (0.000) \\
                    \\
                    Observations & 65 \\
                    $R^2$ & 0.765 \\
                    Adjusted $R^2$ & 0.753 \\
                    Residual Std. Error & 0.056 (df=61) \\
                    F Statistic & 66.041$^{***}$ (df=3; 61) \\
                     \\
                    \hline \\
                    \textit{Note:} & \multicolumn{1}{r}{$^{*}$p$<$0.1; $^{**}$p$<$0.05; $^{***}$p$<$0.01} \\
                \end{tabular}
                \end{subtable}\vspace{24pt}
                \end{minipage}
        \caption{\label{tab:ols_ldm} \acrshort{OLS} Regression Model for \gls{lex} Engagement Ratio.}
    \end{table}

            \begin{table}
        \centering
        \begin{minipage}[b]{\textwidth}
            \begin{subtable}{\textwidth}
                \scriptsize
                \centering
                \subcaption{${scm}\sim{age}+{score}+{count}$}
                \renewcommand{\arraystretch}{1.4}
                        \begin{tabular}{p{0.4\linewidth}p{0.3\linewidth}}
                        \\
                        \hline
                        \textbf{Dependent variable: scm} & \\
                        \hline
                        \\
                        Intercept & 0.977$^{***}$ (0.014) \\
                    age & -0.000$^{}$ (0.000) \\
                    count & 0.000$^{***}$ (0.000) \\
                    score & -0.000$^{***}$ (0.000) \\
                        \\
                    Observations & 65 \\
                    $R^2$ & 0.662 \\
                    Adjusted $R^2$ & 0.645 \\
                    Residual Std. Error & 0.063 (df=61) \\
                    F Statistic & 39.829$^{***}$ (df=3; 61) \\
                        \\
                        \hline 
                        \\
                    \end{tabular}
                \end{subtable}\vspace{6pt}
                \begin{subtable}{\textwidth}
                \scriptsize
                \centering
                \subcaption{${scm}\sim{score}+{count}$}\vspace{6pt}
                \renewcommand{\arraystretch}{1.4}
                    \begin{tabular}{p{0.4\linewidth}p{0.3\linewidth}}
                        \hline
                        \\
                        Intercept & 0.961$^{***}$ (0.008) \\
                    count & 0.000$^{***}$ (0.000) \\
                    score & -0.000$^{***}$ (0.000) \\
                        \\
                    Observations & 65 \\
                    $R^2$ & 0.651 \\
                    Adjusted $R^2$ & 0.640 \\
                    Residual Std. Error & 0.064 (df=62) \\
                    F Statistic & 57.859$^{***}$ (df=2; 62) \\
                     \\
                    \hline \\
                \end{tabular}
                \end{subtable}\vspace{6pt}
                \begin{subtable}{\textwidth}
                \scriptsize
                \centering
                \subcaption{${scm}\sim{score}*{count}$}\vspace{6pt}
                \renewcommand{\arraystretch}{1.4}
                    \begin{tabular}{p{0.4\linewidth}p{0.3\linewidth}}
                        \hline
                        \\
                        Intercept & 0.968$^{***}$ (0.007) \\
                    count & 0.000$^{***}$ (0.000) \\
                    score & -0.000$^{***}$ (0.000) \\
                    score:count & 0.000$^{***}$ (0.000) \\
                        \\
                    Observations & 65 \\
                    $R^2$ & 0.738 \\
                    Adjusted $R^2$ & 0.725 \\
                    Residual Std. Error & 0.056 (df=61) \\
                    F Statistic & 57.372$^{***}$ (df=3; 61) \\
                     \\
                    \hline \\
                    \textit{Note:} & \multicolumn{1}{r}{$^{*}$p$<$0.1; $^{**}$p$<$0.05; $^{***}$p$<$0.01} \\
                \end{tabular}
                \end{subtable}\vspace{24pt}
                \end{minipage}
        \caption{\label{tab:ols_scm} \acrshort{OLS} Regression Model for \gls{syn} lifespan cohorts.}
    \end{table}

            \begin{table}
        \centering
        \begin{minipage}[b]{\textwidth}
            \begin{subtable}{\textwidth}
                \scriptsize
                \centering
                \subcaption{${sdm}\sim{age}+{score}+{count}$}
                \renewcommand{\arraystretch}{1.4}
                        \begin{tabular}{p{0.4\linewidth}p{0.3\linewidth}}
                        \\
                        \hline
                        \textbf{Dependent variable: sdm} & \\
                        \hline
                        \\
                    Intercept & 0.979$^{***}$ (0.014) \\
                    age & -0.000$^{}$ (0.000) \\
                    count & 0.000$^{***}$ (0.000) \\
                    score & -0.000$^{***}$ (0.000) \\
                        \\
                    Observations & 65 \\
                    $R^2$ & 0.668 \\
                    Adjusted $R^2$ & 0.651 \\
                    Residual Std. Error & 0.062 (df=61) \\
                    F Statistic & 40.827$^{***}$ (df=3; 61) \\
                        \\
                        \hline 
                        \\
                    \end{tabular}
                \end{subtable}\vspace{6pt}
                \begin{subtable}{\textwidth}
                \scriptsize
                \centering
                \subcaption{${sdm}\sim{score}+{count}$}\vspace{6pt}
                \renewcommand{\arraystretch}{1.4}
                    \begin{tabular}{p{0.4\linewidth}p{0.3\linewidth}}
                        \hline
                        \\
                    Intercept & 0.963$^{***}$ (0.008) \\
                    count & 0.000$^{***}$ (0.000) \\
                    score & -0.000$^{***}$  (0.000) \\
                        \\
                    Observations & 65 \\
                    $R^2$ & 0.656 \\
                    Adjusted $R^2$ & 0.645 \\
                    Residual Std. Error & 0.063 (df=62) \\
                    F Statistic & 59.042$^{***}$ (df=2; 62) \\
                     \\
                    \hline \\
                \end{tabular}
                \end{subtable}\vspace{6pt}
                \begin{subtable}{\textwidth}
                \scriptsize
                \centering
                \subcaption{${sdm}\sim{score}*{count}$}\vspace{6pt}
                \renewcommand{\arraystretch}{1.4}
                    \begin{tabular}{p{0.4\linewidth}p{0.3\linewidth}}
                        \hline
                        \\
                        Intercept & 0.969$^{***}$ (0.007) \\
                    count & 0.000$^{***}$ (0.000) \\
                    score & -0.000$^{***}$ (0.000) \\
                    score:count & 0.000$^{***}$ (0.000) \\
                        \\
                    Observations & 65 \\
                    $R^2$ & 0.746 \\
                    Adjusted $R^2$ & 0.734 \\
                    Residual Std. Error & 0.054 (df=61) \\
                    F Statistic & 59.764$^{***}$ (df=3; 61) \\
                     \\
                    \hline \\
                    \textit{Note:} & \multicolumn{1}{r}{$^{*}$p$<$0.1; $^{**}$p$<$0.05; $^{***}$p$<$0.01} \\
                \end{tabular}
                \end{subtable}\vspace{24pt}
                \end{minipage}
        \caption{\label{tab:ols_sdm} \acrshort{OLS} Regression Model for \gls{syn} Engagement Ratio.}
    \end{table}

    \begin{table*}
        \scriptsize
        \centering
        \renewcommand{\arraystretch}{1.4}
            
            \caption{Performance Metrics for Engagement Cohorts}
            \label{tab:decile_classification}
            
            \begin{subtable}[t]{\textwidth}
            \centering
            \caption{Baseline}
            \renewcommand{\arraystretch}{1.4}
                \begin{tabularx}{\textwidth}{l*{10}{>{\centering\arraybackslash}X}}
                    \toprule
                    $F_1$-score & Q1 & Q2 & Q3 & Q4 & Q5 & Q6 & Q7 & Q8 & Q9 & Q10 \\ 
                    \midrule
                    \addlinespace[1em]
                    \texttt{r/auckland} & 0.79 & 0.77 & 0.78 & 0.79 & 0.79 & 0.77 & 0.78 & 0.86 & 0.84 & 0.79 \\
                    \texttt{r/Tauranga} & - & 0.15 & 0.31 & 0.45 & 0.49 & 0.13 & 0.09 & 0.29 & 0.73 & 0.12 \\
                    \texttt{r/thetron} & 0.15 & 0.44 & 0.57 & 0.62 & 0.49 & 0.23 & 0.07 & 0.62 & - & - \\
                    \texttt{r/Wellington} & 0.70 & 0.72 & 0.75 & 0.74 & 0.71 & 0.56 & 0.32 & 0.67 & 0.55 & 0.37 \\
                    \texttt{r/chch} & 0.43 & 0.57 & 0.62 & 0.63 & 0.55 & 0.30 & 0.11 & 0.29 & 0.37 & 0.10 \\
                    \texttt{r/dunedin} & 0.35 & 0.48 & 0.71 & 0.76 & 0.79 & 0.52 & 0.31 & 0.62 & 0.57 & 0.11 \\
                    \addlinespace[1em]
                    Macro Avg. & 0.40 & 0.52 & 0.62 & 0.67 & 0.64 & 0.42 & 0.28 & 0.56 & 0.51 & 0.25 \\
                    Weighted Avg. & 0.68 & 0.69 & 0.73 & 0.74 & 0.72 & 0.62 & 0.56 & 0.75 & 0.71 & 0.58 \\
                    \addlinespace[1em]
                    \bottomrule
                \end{tabularx}
            \end{subtable}\vspace{12pt}
        
            \begin{subtable}[t]{\textwidth}
                \centering
                \scriptsize
                \caption{Localised}
                \renewcommand{\arraystretch}{1.4}
                \begin{tabularx}{\textwidth}{l*{7}{>{\centering\arraybackslash}X}}
                    \toprule
                    Cohort & \texttt{r/auckland} & \texttt{r/Tauranga} & \texttt{r/thetron} & \texttt{r/Wellington} & \texttt{r/chch} & \texttt{r/dunedin} & Total \\ 
                    \midrule
                    \addlinespace[1em]
                    Decile 1 & 766 & 16 & 37 & 455 & 172 & 33 & 1,479 \\
                    Decile 2 & 820 & 36 & 78 & 556 & 238 & 70 & 1,798 \\
                    Decile 3 & 785 & 43 & 92 & 510 & 236 & 83 & 1,749 \\
                    Decile 4 & 871 & 51 & 107 & 523 & 250 & 105 & 1,907 \\
                    Decile 5 & 682 & 43 & 68 & 348 & 185 & 73 & 1,399 \\
                    Decile 6 & 508 & 28 & 38 & 197 & 104 & 48 & 923 \\
                    Decile 7 & 838 & 42 & 54 & 238 & 143 & 60 & 1,375 \\
                    Decile 8 & 207 & 6 & 9 & 60 & 29 & 9 & 320 \\
                    Decile 9 & 210 & 7 & 8 & 60 & 31 & 10 & 326 \\
                    \addlinespace[1em]
                    \bottomrule
                \end{tabularx}
            \end{subtable}\vspace{6pt}
    
            \tablenoteparagraph{\textbf{Table Notes}: This table presents the model performance metrics, specifically $F_1$-scores, for text classification tasks utilising proportional sampling based on engagement ratios. It evaluates the effectiveness of the classifier across different engagement-based strata to determine how representative sampling influences predictive accuracy; all data is sourced from Reddit.}
        
    \end{table*}

    \begin{table*}
        \scriptsize
        \centering
        \renewcommand{\arraystretch}{1.4}
            
            \caption{Performance Metrics for Lifespan Cohorts}
            \label{tab:cake_classification}
            
            \begin{subtable}[t]{\textwidth}
            \centering
            \caption{Baseline}
            \renewcommand{\arraystretch}{1.4}
                \begin{tabularx}{\textwidth}{l*{10}{>{\centering\arraybackslash}X}}
                    \toprule 
                    Community & C1 & C2 & C3 & C4 & C5 & C6 & C7 & C8 & C9 & C10 \\ 
                    \midrule
                    \addlinespace[1em]
                    \texttt{r/auckland} & 0.81 & 0.77 & 0.76 & 0.77 & 0.77 & 0.80 & 0.84 & 0.80 & 0.82 & 0.81 \\ 
                    \texttt{r/Tauranga} & 0.45 & 0.37 & 0.27 & 0.19 & 0.14 & - & 0.75 & 0.50 & 0.57 & 0.33 \\ 
                    \texttt{r/thetron} & 0.60 & 0.45 & 0.31 & 0.19 & 0.08 & 0.17 & 0.62 & 0.33 & 0.50 & 0.40 \\ 
                    \texttt{r/Wellington} & 0.83 & 0.63 & 0.56 & 0.59 & 0.48 & 0.58 & 0.73 & 0.59 & 0.55 & 0.63 \\ 
                    \texttt{r/chch} & 0.70 & 0.43 & 0.36 & 0.38 & 0.14 & 0.29 & 0.60 & 0.44 & 0.45 & 0.44 \\ 
                    \texttt{r/dunedin} & 0.82 & 0.64 & 0.52 & 0.51 & 0.39 & 0.53 & 0.46 & 0.18 & 0.36 & 0.27 \\ 
                    \addlinespace[1em]
                    Macro Avg. & 0.70 & 0.55 & 0.46 & 0.44 & 0.33 & 0.40 & 0.67 & 0.47 & 0.54 & 0.48 \\
                    Weighted Avg. & 0.78 & 0.66 & 0.62 & 0.63 & 0.58 & 0.65 & 0.77 & 0.67 & 0.70 & 0.69 \\
                    \addlinespace[1em]
                    \bottomrule
                \end{tabularx}
            \end{subtable}\vspace{12pt}
        
            \begin{subtable}[t]{\textwidth}
                \centering
                \scriptsize
                \caption{Localised}
                \renewcommand{\arraystretch}{1.4}
                \begin{tabularx}{\textwidth}{l*{7}{>{\centering\arraybackslash}X}}
                    \toprule
                    Cohort & \texttt{r/auckland} & \texttt{r/Tauranga} & \texttt{r/thetron} & \texttt{r/Wellington} & \texttt{r/chch} & \texttt{r/dunedin} & Total \\ 
                    \midrule
                    \addlinespace[1em]
                    Decile 1 & 987 & 51 & 113 & 747 & 345 & 144 & 2,387 \\
                    Decile 2 & 1,165 & 67 & 125 & 583 & 290 & 104 & 2,334 \\
                    Decile 3 & 1,064 & 58 & 99 & 514 & 239 & 90 & 2,064 \\
                    Decile 4 & 963 & 48 & 77 & 439 & 213 & 62 & 1,802 \\
                    Decile 5 & 816 & 27 & 46 & 338 & 165 & 46 & 1,438 \\
                    Decile 6 & 438 & 14 & 21 & 180 & 77 & 25 & 755 \\
                    Decile 7 & 187 & 5 & 10 & 89 & 34 & 10 & 335 \\
                    Decile 8 & 187 & 6 & 10 & 82 & 35 & 10 & 330 \\
                    Decile 9 & 199 & 5 & 9 & 71 & 38 & 9 & 331 \\
                    \addlinespace[1em]
                    \bottomrule
                \end{tabularx}
            \end{subtable}\vspace{6pt}
    
            \tablenoteparagraph{\textbf{Table Notes}: This table details the model performance metrics, reported as $F_1$-scores, for text classification across various lifespan cohorts using proportional sampling. It further provides the support values for the held-out test set used in the engagement cohort models, noting that significant class imbalance exists across both city-level communities and specific engagement cohorts. All data is sourced from Reddit.}
        
    \end{table*}
    
\printnoidxglossary[title={Place Names and Other Proper Nouns},type=name]
\printnoidxglossary[title={New Zealand English Glossary},type=dialect]
\printnoidxglossary[title={te reo Māori Glossary},type=maori]
\glsnogroupskiptrue\printnoidxglossary[title=Miscellaneous Glossary,sort=standard]
\glsnogroupskiptrue\printnoidxglossary[title={List of General Abbreviations},type=\acronymtype,sort=standard]
\glsnogroupskiptrue\printnoidxglossary[title={List of Statistical and Model Abbreviations},type=tech,sort=standard]
\glsnogroupskiptrue\printnoidxglossary[title={List of Miscellaneous Abbreviations},type=misc,sort=def]

\thispagestyle{plain}

\end{CJK}
\end{document}

%% file: foo.tex
% -----------------------------
% Abbreviations and Acronyms
% -----------------------------

% -----------------------------
% General Abbreviations
% -----------------------------

\newacronym[plural=APIs,firstplural=Application Programing Interfaces (APIs)]{API}{API}{Application Programming Interface}
\newacronym{AAF}{AAF}{Appreciation, Affinity, Fandom}
\newacronym{AAVE}{AAVE}{African American Vernacular English}
\newacronym{AI}{AI}{artificial intelligence}
\newacronym{CGLU}{CGLU}{Corpus of Global Language-Use}
\newacronym{CMC}{CMC}{Computer Mediated Communication}
\newacronym{IRC}{IRC}{Internet Relay Chat}
\newacronym{LBSN}{LBSN}{Locative-based Social Network}
\newacronym[plural=LLMs,firstplural=large language models (LLMs)]{LLM}{LLM}{large language model}
\newacronym{ML}{ML}{machine learning}
\newacronym[plural=MSAs,firstplural=Metropolitan Statistical Areas (MSAs)]{MSA}{MSA}{Metropolitan Statistical Area}
\newacronym{NLG}{NLG}{Natural Language Generation}
\newacronym{NLP}{NLP}{Natural Language Processing}
\newacronym{NLU}{NLU}{Natural Language Understanding}
\newacronym[plural=NORMs,firstplural=Non-Mobile Older Rural Males (NORMs)]{NORM}{NORM}{Non-Mobile Older Rural Male}
\newacronym{NSFW}{NSFW}{Not Safe for Work}
\newacronym{NZE}{NZE}{New Zealand English}
\newacronym{NZSL}{NZSL}{New Zealand Sign Language}
\newacronym{GIS}{GIS}{Geographic Information Systems}
\newacronym{GISc}{GIScience}{Geographic Information Science}
\newacronym[plural=OPs,firstplural=Original Posters (OPs)]{OP}{OP}{\gls{op}}
\newacronym{AMA}{AMA}{\gls{ama}}
\newacronym{UTC}{UTC}{Coordinated Universal Time}
\newacronym{NZST}{NZST}{New Zealand Standard Time}
\newacronym[plural=URLs,firstplural=Uniform Resource Locators (URLs)]{URL}{URL}{Uniform Resource Locator}
\newacronym[plural=GIFs,firstplural=Graphics Interchange Formats (GIFs)]{GIF}{GIF}{Graphics Interchange Format}

% -----------------------------
% Statistical & Model Abbreviations
% -----------------------------

\newacronym[type=tech]{OLS}{\textsc{OLS}}{Ordinary Least Squares}
\newacronym[type=tech]{PCA}{\textsc{PCA}}{Principal Component Analysis}
\newacronym[type=tech]{RNN}{\textsc{RNN}}{Recurrent Neural Network}
\newacronym[type=tech]{BERT}{\textsc{BERT}}{Bidirectional Encoder Representations from Transformers}
\newacronym[type=tech]{CBOW}{\textsc{CBOW}}{continuous-bag-of-words}
\newacronym[type=tech]{CxG}{\textsc{CxG}}{Construction Grammar}
\newacronym[type=tech]{C2xG}{\textsc{C2xG}}{Computational Construction Grammar}
\newacronym[type=tech]{GIW}{\textsc{GIW}}{Gewichteter Identitätswert}
\newacronym[type=tech]{GloVe}{\textsc{GloVe}}{Global Vectors for Word Representation}
\newacronym[type=tech]{SGNS}{\textsc{SGNS}}{skip-gram with negative sampling}
\newacronym[type=tech]{SVM}{\textsc{SVM}}{Support Vector Machine}
\newacronym[plural=GPEs,firstplural=Geopolitical Entities (GPEs)]{GPE}{GPE}{Geopolitical Entity}
\newacronym[plural=LOCs,firstplural=Physical Locations (LOCs)]{LOC}{LOC}{Physical Location}

% -----------------------------
% Miscellaneous Abbreviations
% -----------------------------

\newacronym[type=misc]{SQ1}{SQ1}{Secondary Research Question 1}
\newacronym[type=misc]{SQ2}{SQ2}{Secondary Research Question 2}
\newacronym[type=misc]{SQ3}{SQ3}{Secondary Research Question 3}

\newacronym[type=misc]{rpost}{\texttt{rpost}}{Submission post titles}
\newacronym[type=misc]{rstitle}{\texttt{rstitle}}{Selfpost titles}
\newacronym[type=misc]{rstext}{\texttt{rstext}}{Selfpost body texts}
\newacronym[type=misc]{rcomm}{\texttt{rcomm}}{Comments}

\newacronym[type=misc]{NZM}{\texttt{NewZealand}}{New Zealand city-only Word2Vec model}
\newacronym[type=misc]{ICM}{\texttt{InnerCircle}}{Inner-circle countries Word2Vec model}
\newacronym[type=misc]{OCM}{\texttt{OuterCircle}}{Outer-circle countries Word2Vec model}
\newacronym[type=misc]{GOO}{\texttt{Google300}}{Google News Corpus Word2Vec model}

\newacronym[type=misc]{lex}{\texttt{lex}}{Lexical features}
\newacronym[type=misc]{syn}{\texttt{syn}}{Syntactic features}
\newacronym[type=misc]{sem}{\texttt{sem}}{Semantic features}

\newacronym[type=misc]{lcm}{\texttt{lcm}}{Lifespan groupings by lexical features}
\newacronym[type=misc]{ldm}{\texttt{ldm}}{Engagement ratio by lexical features}
\newacronym[type=misc]{scm}{\texttt{scm}}{Lifespan groupings by syntactic features}
\newacronym[type=misc]{sdm}{\texttt{sdm}}{Engagement ratio by syntactic features}

% -----------------------------
% General Glossary
% -----------------------------

\newglossaryentry{mod}
{
    name=mod,
    description={Community Moderator},
    plural={mods},
}

\newglossaryentry{op}
{
    name=original poster,
    description={Original poster of a submission post},
    plural={OPs},
}

\newglossaryentry{ama}
{
    name=Ask Me Anything,
    description={a live, text-based, question and answer session},
    plural={OPs},
}

\newglossaryentry{flair}
{
    name=flair,
    description={A customisable, community-specific tag or badge},
    plural={flairs},
}

% -----------------------------
% Placenames and Proper Nouns
% -----------------------------

\newglossaryentry{jafa}
{
    name=JAFA,
    description={(\textit{colloquial, derogatory}) \textit{noun} an Aucklander \citep{deverson_jafa_2005}},
    plural={JAFAs},
    type=dialect
}

\newglossaryentry{alps}
{
    name=alps,
    description={\textit{location} In New Zealand, the \gls{southern_alps}, a South Island mountain chain (pp.4)},
    type=name
}

\newglossaryentry{chch}
{
    name=chch,
    description={\textit{abbr.} Christchurch \citep{deverson_chch_2005}},
    type=name
}

% -----------------------------
% Analogy
% -----------------------------

% -----------------------------
% A
% -----------------------------

\newglossaryentry{arvo}
{
    name=arvo,
    description={\textit{noun} (this) afternoon \citep[p.6]{orsman_new_1994}. p},
    type=dialect
}

% -----------------------------
% B
% -----------------------------

\newglossaryentry{banger}
{
    name=banger,
    description={\textit{noun} a (large) firecracker \citep[p.11]{orsman_new_1994}. p},
    type=dialect
}

\newglossaryentry{berm}
{
    name=berm,
    description={\textit{noun} a grass strip often mown, between the hard surface of a (usually suburban) road and the footpath \citep[p.18]{orsman_new_1994}. p},
    type=dialect
}

\newglossaryentry{belgium}
{
    name=belgium,
    description={\textit{noun} },
    type=dialect
}

\newglossaryentry{bellbird}
{
    name=bellbird,
    description={\textit{noun} },
    plural={bellbirds},
    type=dialect
}

\newglossaryentry{beetroot}
{
    name=beetroot,
    description={\textit{noun} },
    type=dialect
}

\newglossaryentry{beech_tree}
{
    name=beech tree,
    description={\textit{noun} },
    plural={beech trees},
    type=dialect
}

\newglossaryentry{beehive}
{
    name=beehive,
    description={\textit{noun} as \textbf{the Beehive}, a nickname for the dome-roofed building in Wellington \citep[p.16]{orsman_new_1994}. p},
    type=dialect
}

\newglossaryentry{blockholder}
{
    name=blockholder,
    description={\textit{noun} the owner or tenant of a building section or lot \citep{oxford_university_press_blockholder_2023}},
    type=dialect
}

\newglossaryentry{bludger}
{
    name=bludger,
    description={\textit{noun} a persistent scrounger \citep[p.24]{orsman_new_1994}. p},
    type=dialect
}

\newglossaryentry{bludge}
{
    name=bludge,
    description={\textit{verb} often with on, to scrounge; sponge (on) \citep[p.24]{orsman_new_1994}. p},
    type=dialect
}

\newglossaryentry{bush}
{
    name=bush,
    description={\textit{noun} Usually as \textbf{the bush} (land covered with) native or indigenous rain forest \citep[p.40]{orsman_new_1994}. p},
    type=dialect
}

\newglossaryentry{bushwalk}
{
    name=bushwalk,
    description={\textit{noun} a walk in the bush \citep{deverson_bush_2005}. p},
    type=dialect
}

\newglossaryentry{brownie}
{
    name=brownie,
    description={\textit{noun} (also \textit{browny}; mainly South Island rural) a currant or raisin loaf sweetened with (brown) sugar \citep[p.34]{orsman_new_1994}. p},
    type=dialect
}

\newglossaryentry{back_blocks}
{
    name=back blocks,
    description={\textit{noun} A (surveyed) block of grazing land remote from a main access route, or homestead \citep[p.9]{orsman_new_1994}. p},
    type=dialect
}

\newglossaryentry{box_of_birds}
{
    name=box of birds,
    description={\textit{idiomatic phrase} to be or feel fine, fit, in health \citep[p.31]{orsman_new_1994}. p},
    type=dialect
}

\newglossaryentry{bolt}
{
    name=bolt,
    description={\textit{noun} },
    type=dialect
}

\newglossaryentry{byre}
{
    name=byre,
    description={\textit{noun} (also \textbf{cow-byre}; Otago-Southland Scottish) cowshed \citep[p.45]{orsman_new_1994}. p},
    type=dialect
}

\newglossaryentry{byelection}
{
    name=byelection,
    description={\textit{noun} },
    type=dialect
}

\newglossaryentry{bog}
{
    name=bog,
    description={\textit{noun} },
    type=dialect
}

\newglossaryentry{bottlo}
{
    name=bottlo,
    description={\textit{noun} (also \textbf{bottle-oh}) bottle store \citep{oxford_university_press_bottle-oh_2023}. p},
    type=dialect
}

% -----------------------------
% C
% -----------------------------

\newglossaryentry{campervan}
{
    name=campervan,
    description={\textit{noun} },
    type=dialect
}

\newglossaryentry{compo}
{
    name=compo,
    description={\textit{noun} compensation \citep{oxford_university_press_compo_2024}. p},
    type=dialect
}

\newglossaryentry{car_park}
{
    name=car park,
    description={\textit{noun} },
    type=dialect
}

\newglossaryentry{chippy}
{
    name=chippy,
    description={\textit{noun} (also \textbf{chippie}) a potato crisp \citep[p.53]{orsman_new_1994}. pp},
    type=dialect
}

\newglossaryentry{chocka}
{
    name=chocka,
    description={\textit{adjective} (also \textbf{chocker}) tightly packed, from \textbf{chock-a-bloc} \citep{oxford_university_press_chocker_2023}. p},
    type=dialect
}

\newglossaryentry{chur}
{
    name=chur,
    description={\textit{interjection} Used to express good wishes on meeting or departing, or to express thanks, approval (`cheers!') \citep{oxford_university_press_chur_2023}. p},
    type=dialect
}

\newglossaryentry{crossing}
{
    name=crossing,
    description={\textit{noun} },
    type=dialect
}

\newglossaryentry{crib}
{
    name=crib,
    description={\textit{noun} (originally and mainly southern South Island) a weekend or beach cottage \citep[p.65]{orsman_new_1994}. p},
    type=dialect
}

\newglossaryentry{ctv}
{
    name=ctv,
    description={\textit{abbreviation} },
    type=name
}

% -----------------------------
% D
% -----------------------------

\newglossaryentry{de_facto}
{
    name=de facto,
    description={\textit{adjective} a  common-law spouse \citep[p.73]{orsman_new_1994}. p},
    type=dialect
}

\newglossaryentry{DOC}
{
    name=DOC,
    description={\textit{abbreviation} Acronym of the Department of Conservation \citep[p.77]{orsman_new_1994}. p},
    type=name
}

\newglossaryentry{duvet}
{
    name=duvet,
    description={\textit{noun} },
    type=dialect
}

\newglossaryentry{dux}
{
    name=dux,
    description={\textit{noun} },
    type=dialect
}

% -----------------------------
% E
% -----------------------------

\newglossaryentry{eh}
{
    name=eh,
    description={\textit{particle} a frequent speech particle reinforcing or emphasising a question or request \citep[p.87]{orsman_new_1994}. p},
    type=dialect
}

\newglossaryentry{egg}
{
    name=egg,
    description={\textit{noun} an obnoxious person \citep{group_think_dont_2024}. p},
    type=dialect
}

\newglossaryentry{electorate}
{
    name=electorate,
    description={\textit{noun} the area represented by a Member of Parliament; a constituency \citep[p.87]{orsman_new_1994}. p},
    type=dialect
}

\newglossaryentry{expressway}
{
    name=expressway,
    description={\textit{noun} },
    type=dialect
}

% -----------------------------
% F
% -----------------------------

\newglossaryentry{feijoa}
{
    name=feijoa,
    description={\textit{noun} any evergreen shrub or tree of the genus Feijoa, bearing edible guava-like fruit \citep{deverson_feijoa_2005}. p},
    type=dialect
}

\newglossaryentry{fern}
{
    name=fern,
    description={\textit{noun} },
    plural={ferns},
    type=dialect
}

\newglossaryentry{fiordland}
{
    name=Fiordland,
    description={\textit{location} The area of fiord-like sounds on the south west South Island coast; also its hinterland \citep[p.95]{orsman_new_1994}. p},
    type=name
}

\newglossaryentry{footpath}
{
    name=footpath,
    description={\textit{noun} },
    type=dialect
}

\newglossaryentry{footy}
{
    name=footy,
    description={\textit{noun} rugby \citep{oxford_university_press_footy_2024}. p},
    type=dialect
}

\newglossaryentry{flatting}
{
    name=flatting,
    description={\textit{noun} },
    type=dialect
}

\newglossaryentry{flatmate}
{
    name=flatmate,
    description={\textit{noun} },
    type=dialect
}

\newglossaryentry{fucken}
{
    name=fucken,
    description={\textit{adjective} },
    type=dialect
}

% -----------------------------
% G
% -----------------------------

\newglossaryentry{girdle}
{
    name=girdle,
    description={\textit{noun} },
    type=dialect
}

\newglossaryentry{goashore}
{
    name=goashore,
    description={\textit{noun} a three-legged pot \citep[p.108]{orsman_new_1994}. p},
    type=dialect
}

\newglossaryentry{glowworm}
{
    name=glowworm,
    description={\textit{noun} (also \textbf{New Zealand glow-worm} the larva of a fungus gnat having luminous internal tubules attached to the gut \citep[p.107]{orsman_new_1994}. p},
    plural={glowworms},
    type=dialect
}

\newglossaryentry{glut}
{
    name=glut,
    description={\textit{noun} },
    type=dialect
}

\newglossaryentry{greenstone}
{
    name=greenstone,
    description={\textit{noun} a nephrite rock of the south-west South Island only \citep[p.112]{orsman_new_1994}. p},
    type=dialect
}

\newglossaryentry{gumboot}
{
    name=gumboot tea,
    description={\textit{noun} an ordinary black tea \citep{pollock_tea_2013}. p},
    type=dialect
}

\newglossaryentry{gday}
{
    name=gday,
    description={\textit{interjection} (also \textbf{gidday}) a familiar (usually male) greeting \citep[p.105]{orsman_new_1994}. p},
    type=dialect
}

\newglossaryentry{grundies}
{
    name=Grundies,
    description={\textit{noun} underpants \citep{deverson_grundies_2005}. p},
    type=dialect
}

\newglossaryentry{gruts}
{
    name=gruts,
    description={\textit{noun} },
    type=dialect
}

% -----------------------------
% H
% -----------------------------

\newglossaryentry{hospo}
{
    name=hospo,
    description={\textit{noun} },
    type=dialect
}

\newglossaryentry{hoon}
{
    name=hoon,
    description={\textit{noun} a fool; a stupid person; \textit{verb} to do things in an irresponsible way \citep[p.128]{orsman_new_1994}. p},
    type=dialect
}

% -----------------------------
% I
% -----------------------------

% -----------------------------
% J
% -----------------------------

% -----------------------------
% K
% -----------------------------

\newglossaryentry{kindy}
{
    name=kindy,
    description={\textit{noun} a familiar name for kindergarten \citep[p.144]{orsman_new_1994}. p},
    type=dialect
}

\newglossaryentry{kiwiana}
{
    name=kiwiana,
    description={\textit{adjective} },
    type=dialect
}

\newglossaryentry{kransky}
{
    name=kransky,
    description={\textit{noun} a sausage made from veal and smoked pork, usually eaten fried or grilled \citep{stevenson_kransky_2010}. p},
    type=dialect
}

% -----------------------------
% L
% -----------------------------

\newglossaryentry{lash}
{
    name=lash,
    description={\textit{verb} },
    type=dialect
}

\newglossaryentry{longdrop}
{
    name=long drop,
    description={\textit{noun} an outdoor privy built over a hole in the ground \citep[p.157]{orsman_new_1994}},
    type=dialect
}

\newglossaryentry{lemonade}
{
    name=lemonade,
    description={\textit{noun} },
    type=dialect
}

\newglossaryentry{like}
{
    name=like,
    description={\textit{noun} },
    type=dialect
}

\newglossaryentry{little_boy}
{
    name=little boy,
    description={\textit{noun} },
    type=dialect
}

\newglossaryentry{lux}
{
    name=lux,
    description={\textit{noun} },
    type=dialect
}

% -----------------------------
% M
% -----------------------------

\newglossaryentry{maccas}
{
    name=maccas,
    description={\textit{name} },
    type=name
}

\newglossaryentry{marmageddon}
{
    name=Marmageddon,
    description={\textit{noun} national shortage of Marmite between 2012-2013 \citep{cuming_marmageddon_2013}. p},
    type=dialect
}

\newglossaryentry{marquee}
{
    name=marquee,
    description={\textit{noun} },
    type=dialect
}

\newglossaryentry{messages}
{
    name=messages,
    description={\textit{noun} },
    type=dialect
}

\newglossaryentry{mo_eng}
{
    name=mo,
    description={\textit{abbreviation} abbreviation of moustache \citep[p.169]{orsman_new_1994}. p},
    type=dialect
}

\newglossaryentry{mind}
{
    name=mind,
    description={\textit{noun} },
    type=dialect
}

\newglossaryentry{morepork}
{
    name=morepork,
    description={\textit{noun} a small brown owl \citep[p.172]{orsman_new_1994}. p},
    type=dialect
}

\newglossaryentry{morning_tea}
{
    name=morning tea,
    description={\textit{noun} },
    type=dialect
}

\newglossaryentry{mufti}
{
    name=mufti,
    description={\textit{noun} `civilian' clothes worn by school students on a days when they would normally be required to wear a school uniform \citep[p.173]{orsman_new_1994}. p},
    type=dialect
}

\newglossaryentry{munted}
{
    name=munted,
    description={\textit{adjective} },
    type=dialect
}

\newglossaryentry{miq}
{
    name=miq,
    description={\textit{abbreviation} },
    type=name
}

\newglossaryentry{motorway}
{
    name=motorway,
    description={\textit{noun} },
    type=dialect
}

% -----------------------------
% N
% -----------------------------

\newglossaryentry{new_zild}
{
    name=New Zild,
    description={(\textit{jocular}) \textit{noun} New Zealand English; \textit{adjective} pertaining to or characteristic of stereotyped New Zealandness, or New Zealand speech \citep[p.181]{orsman_new_1994}. p},
    type=name
}

\newglossaryentry{nogging}
{
    name=nogging,
    description={\textit{noun} },
    type=dialect
}

\newglossaryentry{nor_wester}
{
    name=nor'wester,
    description={},
    type=dialect
}

% -----------------------------
% O
% -----------------------------

\newglossaryentry{old_mate}
{
    name=old mate,
    description={\textit{noun} },
    type=dialect
}

% -----------------------------
% P
% -----------------------------

\newglossaryentry{pack_a_sad}
{
    name=pack a sad,
    description={\textit{idiomatic phrase} to become severely depressed to reveal a dismal or depressed state of mind \citep[p.191]{orsman_new_1994}. p},
    type=dialect
}

\newglossaryentry{pikelet}
{
    name=pikelet,
    description={\textit{noun} },
    type=dialect
}

\newglossaryentry{pasifika}
{
    name=pasifika,
    description={\textit{noun} },
    type=dialect
}

\newglossaryentry{public_transport}
{
    name=public transport,
    description={\textit{noun} },
    type=dialect
}

\newglossaryentry{polytech}
{
    name=polytech,
    description={\textit{noun} },
    type=dialect
}

% -----------------------------
% Q
% -----------------------------

% -----------------------------
% R
% -----------------------------

\newglossaryentry{rarked_up}
{
    name=rarked up,
    description={\textit{verb} annoy or irritate (someone) \citep{stevenson_rark_2010}. p},
    type=dialect
}

\newglossaryentry{rego}
{
    name=rego,
    description={\textit{noun} },
    type=dialect
}

\newglossaryentry{reserve_bank}
{
    name=reserve bank,
    description={\textit{noun} },
    type=dialect
}

\newglossaryentry{righto}
{
    name=righto,
    description={\textit{interjection} },
    type=dialect
}

\newglossaryentry{roundabout}
{
    name=roundabout,
    description={\textit{noun} },
    type=dialect
}

% -----------------------------
% S
% -----------------------------

\newglossaryentry{sandfly}
{
    name=sandfly,
    description={\textit{noun} },
    type=dialect
}

\newglossaryentry{savvyb}
{
    name=savvyb,
    description={\textit{noun} sauvignon blanc \citep{british_oracle_savvy_2019}. p},
    type=dialect
}

\newglossaryentry{saveloy}
{
    name=saveloy,
    description={\textit{noun} a seasoned red pork sausage, dried and smoked, and sold ready to eat \citep{stevenson_saveloy_2010}. p},
    type=dialect
}

\newglossaryentry{scarfie}
{
    name=scarfie,
    description={\textit{noun} },
    plural={scarfies},
    type=dialect
}

\newglossaryentry{servo}
{
    name=servo,
    description={\textit{noun} a service station \citep{oxford_university_press_servo_2023}. p},
    type=dialect
}

\newglossaryentry{shaws}
{
    name=shaws,
    description={\textit{noun} },
    type=dialect
}

\newglossaryentry{shitty}
{
    name=shitty,
    description={\textit{adjective} },
    type=dialect
}

\newglossaryentry{silver_fern}
{
    name=silver fern,
    description={\textit{noun} },
    plural={silver ferns},
    type=dialect
}

\newglossaryentry{snapper}
{
    name=snapper,
    description={\textit{noun} a fish. p},
    type=dialect
}

\newglossaryentry{slaters}
{
    name=slaters,
    description={\textit{noun} },
    type=dialect
}

\newglossaryentry{skux}
{
    name=skux,
    description={\textit{noun} a womaniser \citep{momoisea_lani_2015}. p},
    type=dialect
}

\newglossaryentry{southern_alps}
{
    name=southern alps,
    description={\textit{location} Southern Alps, South Island. p},
    type=name
}

\newglossaryentry{southland}
{
    name=Southland,
    description={\textit{location} Southland Region, South Island. p},
    type=name
}

\newglossaryentry{sooky}
{
    name=sooky,
    description={\textit{adjective} },
    type=dialect
}

\newglossaryentry{superheater}
{
    name=superheater,
    description={\textit{noun} },
    type=dialect
}

\newglossaryentry{sulky}
{
    name=sulky,
    description={\textit{noun} },
    type=dialect
}

\newglossaryentry{suss}
{
    name=suss,
    description={\textit{noun} },
    type=dialect
}

\newglossaryentry{sparky}
{
    name=sparky,
    description={\textit{noun} },
    type=dialect
}

\newglossaryentry{spagbol}
{
    name=spagbol,
    description={\textit{noun} spaghetti bolognese \citep{oxford_university_press_spag_2025}. p},
    type=dialect
}

\newglossaryentry{sweetbix}
{
    name=sweetbix,
    description={\textit{interjection} another way to say something is `sweet' \citep{liz_quilty_sweetbix_2007}. p},
    type=dialect
}

% -----------------------------
% T
% -----------------------------

\newglossaryentry{takeaway}
{
    name=takeaway,
    description={\textit{noun} },
    type=dialect
}

\newglossaryentry{tall_poppy}
{
    name=tall poppy,
    description={\textit{noun} see \gls{tall_poppy_syndrome}. p},
    type=dialect
}

\newglossaryentry{tall_poppy_syndrome}
{
    name=tall poppy syndrome,
    description={\textit{noun} a perceived tendency to disparage prominent or successful people \citep{oxford_university_press_tall_2025}. p},
    type=dialect
}

\newglossaryentry{tailing}
{
    name=tailing,
    description={\textit{noun} the docking or removing of a lamb's tail \citep[p.279]{orsman_new_1994}. p},
    type=dialect
}

\newglossaryentry{torch}
{
    name=torch,
    description={\textit{noun} a portable battery-powered electric lamp \citep{deverson_torch_2005}. p},
    type=dialect
}

\newglossaryentry{tomato_sauce}
{
    name=tomato sauce,
    description={\textit{noun} },
    type=dialect
}

\newglossaryentry{tradie}
{
    name=tradie,
    description={\textit{noun} },
    type=dialect
}

\newglossaryentry{tron}
{
    name=tron,
    description={\textit{location} },
    type=name
}

\newglossaryentry{tussock}
{
    name=tussock,
    description={\textit{noun} },
    type=dialect
}

\newglossaryentry{tvnz}
{
    name=tvnz,
    description={\textit{abbreviation} },
    type= name
}

\newglossaryentry{tramp}
{
    name=tramp,
    description={\textit{noun} extensive recreational walking in rough country \citep[296]{orsman_new_1994}. p},
    type=dialect
}

\newglossaryentry{tramping}
{
    name=tramping,
    description={\textit{noun} extensive recreational walking in rough country \citep[296]{orsman_new_1994}. p},
    type=dialect
}

\newglossaryentry{tucker_fucker}
{
    name=tucker fucker,
    description={\textit{noun} a microwave oven; a bad cook \citep{miller_lingo_2009}. p},
    type=dialect
}

% -----------------------------
% U
% -----------------------------

\newglossaryentry{ute}
{
    name=ute,
    description={\textit{noun} a utility or small truck \citep{deverson_ute_2005}. p},
    type=dialect
}

% -----------------------------
% V
% -----------------------------

\newglossaryentry{jab}
{
    name=jab,
    description={\textit{noun} },
    type=dialect
}

\newglossaryentry{jandal}
{
    name=jandal,
    description={\textit{noun} a simple sandal (often rubber of plastic) held to the foot by two straps joined across the arch to the sides and to a single strap fitted between two toes \citep[p.135]{orsman_new_1994}. p},
    plural={jandals},
    type=dialect
}

\newglossaryentry{joker}
{
    name=joker,
    description={\textit{noun} },
    type=dialect
}

% -----------------------------
% W
% -----------------------------

\newglossaryentry{waitakere_ranges}
{
    name=waitākere ranges,
    description={\textit{location} Waitākere Ranges, North Island. p},
    type=name
}

\newglossaryentry{lake_wanaka}
{
    name=lake wānaka,
    description={\textit{location} Lake Wānaka, South Island. p},
    type=name
}

\newglossaryentry{welly}
{
    name=welly,
    description={\textit{location}},
    type=name
}

\newglossaryentry{wee}
{
    name=wee,
    description={\textit{adjective} in frequent general use in New Zealand replacing little \citep[p.309]{orsman_new_1994}. p},
    type=dialect
}

\newglossaryentry{wop_wops}
{
    name=wop wops,
    description={\textit{noun} the remote or primitive rural districts; the \gls{back_blocks} \citep[p.317]{orsman_new_1994}. p},
    type=dialect
}

\newglossaryentry{wof}
{
    name=wof,
    description={\textit{abbreviation} },
    type=dialect
}

\newglossaryentry{whinge}
{
    name=whinge,
    description={\textit{verb} },
    type=dialect
}

% -----------------------------
% X
% -----------------------------

% -----------------------------
% Y
% -----------------------------

\newglossaryentry{youse}
{
    name=youse,
    description={(also \textbf{yous}) used mainly with a plural noun as a term of address to more than one person \citep[p.320]{orsman_new_1994}. p},
    type=dialect
}

% -----------------------------
% Z
% -----------------------------

% -----------------------------
% te reo Māori
% -----------------------------

% -----------------------------
% A
% -----------------------------

\newglossaryentry{aotea}
{
    name=aotea,
    description={\textit{noun} },
    type=name
}

\newglossaryentry{aotearoa}
{
    name=aotearoa,
    description={\textit{location} North Island or all of New Zealand. \citep[p.6]{ryan_revised_1983}. p},
    type=name
}

\newglossaryentry{aoraki}
{
    name=aoraki,
    description={\textit{location} (also \textbf{Aorangi}) Mount Cook \citep[p.6]{ryan_revised_1983}. p},
    type=name
}

\newglossaryentry{aroha}
{
    name=aroha,
    description={\textit{noun} },
    type=maori
}

% -----------------------------
% H
% -----------------------------

\newglossaryentry{haka}
{
    name=haka,
    description={\textit{noun} fierce dance with chant \citep[p.9]{ryan_revised_1983}. p},
    type=maori
}

\newglossaryentry{harakeke}
{
    name=harakeke,
    description={\textit{noun} flax \citep[p.9]{ryan_revised_1983}. p},
    type=maori
}

\newglossaryentry{hau}
{
    name=hau,
    description={\textit{noun} wind; famous; essence \citep[p.10]{ryan_revised_1983}. p},
    type=maori
}

\newglossaryentry{he_aha}
{
    name={he aha te mea nui te ao? he tangata he tangata, he tangata!},
    description={\textit{idiomatic phrase} What is the most important thing in the world? It is people, it is people, it is people!},
    type=maori
}

\newglossaryentry{he_waka}
{
    name=he waka eke noa,
    description={\textit{idiomatic phrase} a \gls{waka} was used by the community so this \gls{whakatauki} can mean `what's yours is mine, and what's mine is yours' \citep{moorfield_he_2025}. p},
    type=maori
}

\newglossaryentry{huia_eng}
{
    name=huia,
    description={\textit{noun} see \gls{huia_mao}},
    type=dialect
}

\newglossaryentry{huia_mao}
{
    name=hūia,
    description={\textit{noun} bird of that name \citep[p.13]{ryan_revised_1983}. p},
    type=maori
}

\newglossaryentry{hori}
{
    name=hori,
    description={\textit{noun} },
    type=maori
}

\newglossaryentry{hoiho}
{
    name=hoiho,
    description={\textit{noun} },
    type=maori
}

% -----------------------------
% I
% -----------------------------

\newglossaryentry{iwi}
{
    name=iwi,
    description={\textit{noun} tribe; bone; people; strength \citep[p.14]{ryan_revised_1983}. p},
    type=maori
}

% -----------------------------
% K
% -----------------------------

\newglossaryentry{kaika}
{
    name=kāika,
    description={\textit{noun} variation of \gls{kainga} \citep{deverson_kaika_2005}. p},
    type=maori
}

\newglossaryentry{kainga}
{
    name=kāinga,
    description={\textit{noun} home; village \citep[p.15]{ryan_revised_1983}. p},
    type=maori
}

\newglossaryentry{kaikoura}
{
    name=kaikoura,
    description={\textit{location} Kaikōura, Canterbury Region. p},
    type=name
}

\newglossaryentry{kauri}
{
    name=kauri,
    description={\textit{noun} kauri tree; resin \citep[p.18]{ryan_revised_1983}. p},
    type=maori
}

\newglossaryentry{kakapo_eng}
{
    name=kakapo,
    description={\textit{noun} see \gls{kakapo_mao}. p},
    type=dialect
}

\newglossaryentry{kakapo_mao}
{
    name=kākāpō,
    description={\textit{noun} ground parrot \citep[p.16]{ryan_revised_1983}. p},
    type=maori
}

\newglossaryentry{kanuka}
{
    name=kānuka,
    description={\textit{noun} white tea-tree \citep{moorfield_kanuka_2025}. p},
    type=maori
}

\newglossaryentry{karanga}
{
    name=karanga,
    description={\textit{verb} (-tia) call; \textit{noun} relative \citep[p.17]{ryan_revised_1983}. p},
    type=maori
}

\newglossaryentry{kia_ora}
{
    name=kia ora,
    description={\textit{interjection} hello! cheers! good luck! best wishes! \citep{moorfield_kia_2025}. p},
    type=maori
}

\newglossaryentry{kea}
{
    name=kea,
    description={\textit{noun} mountain parrot \citep[p.18]{ryan_revised_1983}. p},
    type=maori
}

\newglossaryentry{kotare}
{
    name=kōtare,
    description={\textit{noun} kingfisher \citep[p.21]{ryan_revised_1983}. p},
    type=maori
}

\newglossaryentry{koauau}
{
    name=kōauau,
    description={\textit{noun} flute \citep[p.19]{ryan_revised_1983}. p},
    type=maori
}

\newglossaryentry{kowhai}
{
    name=kōwhai,
    description={\textit{adjective} yellow; \textit{noun} shrub \citep[p.22]{ryan_revised_1983}. p},
    type=maori
}

\newglossaryentry{kiwi}
{
    name=kiwi,
    description={\textit{noun} flightless bird \citep[p.18]{ryan_revised_1983}. p},
    type=maori
}

% -----------------------------
% M
% -----------------------------

\newglossaryentry{manuka}
{
    name=mānuka,
    description={\textit{noun} tea tree \citep[p.24]{ryan_revised_1983}. p},
    type=maori
}

\newglossaryentry{mahi}
{
    name=mahi,
    description={\textit{verb} (-a) do; work; to make; deed; undertaking \citep[p.23]{ryan_revised_1983}. pp},
    type=maori
}

\newglossaryentry{mihi}
{
    name=mihi,
    description={\textit{noun} greet; admire \citep[p.26]{ryan_revised_1983}. p},
    type=maori
}

\newglossaryentry{mita}
{
    name=mita,
    description={\textit{noun} rhythm, intonation, pronunciation and sound of a language, accent, diction, elocution, dialect, register \citep{moorfield_mita_2025}. p},
    type=maori
}

\newglossaryentry{moho}
{
    name=mōho,
    description={\textit{noun}},
    type=maori
}

\newglossaryentry{mo_mao}
{
    name=mō,
    description={\textit{particle} for, about \citep[p.26]{ryan_revised_1983}. p},
    type=maori
}

\newglossaryentry{marae}
{
    name=marae,
    description={\textit{noun} meeting-ground \citep[p.24]{ryan_revised_1983}. p},
    type=maori
}

\newglossaryentry{maori_mao}
{
    name=māori,
    description={\textit{noun} ordinary; fresh; native people \citep[p.24]{ryan_revised_1983}. p},
    type=maori
}

\newglossaryentry{maori_eng}
{
    name=maori,
    description={\textit{noun} see \gls{maori_mao}. p},
    type=dialect
}

\newglossaryentry{matariki}
{
    name=matariki,
    description={\textit{name} (also \textbf{mātāriki}) Pleiades stars \citep[p.25]{ryan_revised_1983}. p},
    type=name
}

\newglossaryentry{manuhiri}
{
    name=manuhiri,
    description={\textit{noun} guest; visitor \citep[p.24]{ryan_revised_1983}. p},
    type=maori
}

\newglossaryentry{mairehau}
{
    name=mairehau,
    description={\textit{noun} },
    type=maori
}

\newglossaryentry{maoritanga}
{
    name=Māoritanga,
    description={\textit{noun} },
    type=maori
}

% -----------------------------
% N
% -----------------------------

\newglossaryentry{nga_mihi}
{
    name=ngā mihi,
    description={\textit{idiomatic phrase} acknowledgements \citep{moorfield_nga_2025}. p},
    type=maori
}

\newglossaryentry{ngahere}
{
    name=ngahere,
    description={\textit{noun} forest \citep[p.29]{ryan_revised_1983}. p},
    type=maori
}

\newglossaryentry{ngati_mao}
{
    name=ngāti,
    description={\textit{noun} used with tribal names \citep[p.29]{ryan_revised_1983}. p},
    type=maori
}

\newglossaryentry{ngaio}
{
    name=ngaio,
    description={\textit{noun} ngaio tree \citep[p.29]{ryan_revised_1983}. p},
    type=maori
}

% -----------------------------
% O
% -----------------------------

\newglossaryentry{otago}
{
    name=otago,
    description={\textit{location} Otago Region, South Island. p},
    type=name
}

% -----------------------------
% P
% -----------------------------

\newglossaryentry{pa_eng}
{
    name=pa,
    description={\textit{noun} see \gls{pa_mao}. p},
    type=dialect
}

\newglossaryentry{pa_mao}
{
    name=pā,
    description={\textit{noun} fortified village \citep[p.31]{ryan_revised_1983}. p},
    type=maori
}

\newglossaryentry{whare_karakia}
{
    name=whare karakia,
    description={\textit{noun} church \citep[p.61]{ryan_revised_1983}. p},
    type=maori
}

\newglossaryentry{pakeha_eng}
{
    name=pakeha,
    description={\textit{noun} see \gls{pakeha_mao}. p},
    type=dialect
}

\newglossaryentry{pakeha_mao}
{
    name=pākehā,
    description={\textit{noun} not Māori; European \citep[p.32]{ryan_revised_1983}. p},
    type=maori
}

\newglossaryentry{papatuanuku}
{
    name=Papatūānuku,
    description={\textit{name} (also \textbf{Papatuanuku}) mother earth \citep[p.33]{ryan_revised_1983}. p},
    type=name
}

\newglossaryentry{para}
{
    name=para,
    description={\textit{noun} garbage; pulp; scraps; edible fern \citep[p.33]{ryan_revised_1983}. p},
    type=maori
}

\newglossaryentry{patu}
{
    name=patu,
    description={\textit{verb} (-a) to beat; \textit{noun} weapon; beater \citep[p.34]{ryan_revised_1983}. p},
    type=maori
}

\newglossaryentry{pounamu}
{
    name=pounamu,
    description={\textit{noun} greenstone; bottle; \textit{verb} to preserve 37},
    type=maori
}

\newglossaryentry{po_eng}
{
    name=po,
    description={\textit{noun} see \gls{po_mao}. p},
    type=dialect
}

\newglossaryentry{po_mao}
{
    name=pō,
    description={\textit{noun} night; death; after life; realm of death \citep[p.35]{ryan_revised_1983}. p},
    type=maori
}

\newglossaryentry{poi}
{
    name=poi,
    description={\textit{noun} ball; \textit{verb} (-a) swing about \citep[p.36]{ryan_revised_1983}. p},
    type=maori
}

\newglossaryentry{purerehua}
{
    name=pūrerehua,
    description={\textit{noun} butterfly \citep[p.38]{ryan_revised_1983}. p},
    type=maori
}

% -----------------------------
% T
% -----------------------------

\newglossaryentry{taranaki}
{
    name=taranaki,
    description={\textit{location} Taranaki Region, North Island (also \textbf{Mount Taranaki}). p},
    type=name
}

\newglossaryentry{tauranga}
{
    name=tauranga,
    description={\textit{location} Tauranga, Bay of Plenty Region. p},
    type=name
}

\newglossaryentry{taupo_eng}
{
    name=taupo,
    description={\textit{location} see \gls{taupo_mao}. p},
    type=name
}

\newglossaryentry{taupo_mao}
{
    name=taupō,
    description={\textit{location} Taupō, Waikato Region (also \textbf{Lake Taupō}). p},
    type=name
}

\newglossaryentry{tahi}
{
    name=tahi,
    description={\textit{numeric} one \citep[p.43]{ryan_revised_1983}. p},
    type=maori
}

\newglossaryentry{takiwa}
{
    name=takiwā,
    description={\textit{noun} area; space; time; place \citep[p.44]{ryan_revised_1983}. p},
    type=maori
}

\newglossaryentry{tapu}
{
    name=tapu,
    description={\textit{noun} sacred; forbidden \citep[p.45]{ryan_revised_1983}. p},
    type=maori
}

\newglossaryentry{tohuto}
{
    name=tohutō,
    description={\textit{noun} macron - a symbol to mark long vowels \citep{moorfield_tohuto_2025}. p},
    type=maori
}

\newglossaryentry{totara_eng}
{
    name=totara,
    description={\textit{noun} see \gls{totara_mao}. p},
    type=dialect
}

\newglossaryentry{totara_mao}
{
    name=tōtara,
    description={\textit{noun} tree; canoe made of it \citep[p.50]{ryan_revised_1983}. p},
    type=maori
}

\newglossaryentry{toi}
{
    name=toi,
    description={\textit{noun} point; central part of a citadel \citep[p.49]{ryan_revised_1983}. p},
    type=maori
}

\newglossaryentry{timaru}
{
    name=timaru,
    description={\textit{location} Timaru (also \textbf{Te Tihi o Maru}), Canterbury Region. p},
    type=name
}

\newglossaryentry{tangi}
{
    name=tangi,
    description={\textit{verb} (-hia) to cry; weep; mourn \citep[p.45]{ryan_revised_1983}. p},
    type=maori
}

\newglossaryentry{tui_mao}
{
    name=tūī,
    description={\textit{noun} bird of that name \citep[p.51]{ryan_revised_1983}. p},
    type=maori
}

\newglossaryentry{tui_eng}
{
    name=tui,
    description={\textit{noun} see \gls{tui_mao}. p},
    type=dialect
}

\newglossaryentry{tuna}
{
    name=tuna,
    description={\textit{noun} long-finned eel},
    type=maori
}

% -----------------------------
% R
% -----------------------------

\newglossaryentry{ra}
{
    name=rā,
    description={\textit{noun} day; date; sun; a sail \citep[p.39]{ryan_revised_1983}. p},
    type=maori
}

\newglossaryentry{rakau}
{
    name=rākau,
    description={\textit{noun} tree; stick; wooden; bat; weapon \citep[p.39]{ryan_revised_1983}. p},
    type=maori
}

\newglossaryentry{ranginui}
{
    name=Ranginui,
    description={\textit{name} },
    type=maori
}

\newglossaryentry{rimu}
{
    name=rimu,
    description={\textit{noun} red pine; seaweed \citep[p.41]{ryan_revised_1983}. p},
    type=maori
}

\newglossaryentry{rotorua}
{
    name=rotorua,
    description={\textit{location} Rotorua, Bay of Plenty Region (also \textbf{Lake Rotorua}). p},
    type=name
}

\newglossaryentry{ruapehu}
{
    name=ruapehu,
    description={\textit{location} Mount Ruapehu, North Island. p},
    type=name
}

% -----------------------------
% W
% -----------------------------

\newglossaryentry{waikato}
{
    name=waikato,
    description={\textit{location} Waikato Region, North Island (also \textbf{Waikato River}). p},
    type=name
}

\newglossaryentry{wanaka}
{
    name=wanaka,
    description={\textit{location} Wānaka, Otago Region. p},
    type=name
}

\newglossaryentry{waikanae}
{
    name=waikanae,
    description={\textit{location} Waikanae, Wellington Region. p},
    type=name
}

\newglossaryentry{waitangi}
{
    name=waitangi,
    description={\textit{location} Waitangi, Northland Region. p},
    type=name
}

\newglossaryentry{waka}
{
    name=waka,
    description={\textit{noun} vehicle; canoe; box for feathers \citep[p.55]{ryan_revised_1983}. p},
    type=maori
}

\newglossaryentry{weta_eng}
{
    name=weta,
    description={\textit{noun} see \gls{weta_mao}. p},
    type=dialect
}

\newglossaryentry{weta_mao}
{
    name=wētā,
    description={\textit{noun} insect \citep[p.55]{ryan_revised_1983}. p},
    type=maori
}

\newglossaryentry{whakatauki}
{
    name=whakataukī,
    description={\textit{noun} proverb \citep[p.59]{ryan_revised_1983}. pp},
    type=maori
}

\newglossaryentry{whanau_eng}
{
    name=whanau,
    description={\textit{noun} see \gls{whanau_mao}. p},
    type=dialect
}

\newglossaryentry{whanau_mao}
{
    name=whānau,
    description={\textit{verb} (-a) to be born, give birth; \textit{noun} family \citep[p.60]{ryan_revised_1983}. p},
    type=maori
}

\newglossaryentry{whangarei}
{
    name=whangarei,
    description={\textit{location} Whangārei, Northland Region. p},
    type=name
}

\newglossaryentry{whenua}
{
    name=whenua,
    description={\textit{noun} land; country \citep[p.62]{ryan_revised_1983}. p},
    type=maori
}

\newglossaryentry{whetu}
{
    name=whetū,
    description={\textit{noun} star \citep[p.62]{ray_language_1999}. p},
    type=maori
}

%% file: thesis.bbl
\begin{thebibliography}{}

\bibitem[Abell and Gordon, 1990]{abell_this_1990}
Abell, M. and Gordon, E. (1990).
\newblock This objectionable colonial dialect': historical and contemporary attitudes to {New} {Zealand} speech.
\newblock In Bell, A. and Holmes, J., editors, {\em New {Zealand} {Ways} of {Speaking} {English}}. Multilingual Matters, Clevedon, England; Bristol, PA.

\bibitem[{Adam}, 2019]{adam_they_2019}
{Adam} (2019).
\newblock They {Had} {Us} {In} the {First} {Half}.

\bibitem[Adams, 2022]{adams_scraping_2022}
Adams, N.~N. (2022).
\newblock '{Scraping}' {Reddit} posts for academic research? {Addressing} some blurred lines of consent in growing internet-based research trend during the time of {COVID}-19.
\newblock {\em International journal of social research methodology}, 27(1).
\newblock https://doi.org/10.1080/13645579.2022.2111816.

\bibitem[Agha, 2003]{agha_social_2003}
Agha, A. (2003).
\newblock The social life of cultural value.
\newblock {\em Language \& Communication}, 23(3):231--273.
\newblock https://doi.org/10.1016/S0271-5309(03)00012-0.

\bibitem[Agnew, 1987]{agnew_place_1987}
Agnew, J.~A. (1987).
\newblock {\em Place and {Politics}: {The} {Geographical} {Mediation} of {State} and {Society}}, volume~1 of {\em Routeledge {Library} {Editions}: {Political} {Geography}}.
\newblock Routledge, Abingdon, England; New York, NY, 3 edition.

\bibitem[Ainsworth, 2004]{ainsworth_regional_2004}
Ainsworth, H. (2004).
\newblock Regional {Variation} in {New} {Zealand} {English}: the {Taranaki} {Sing}-{Song} {Accent}.
\newblock Retrieved from https://doi.org/10.26686/wgtn.16945720.v1.

\bibitem[Amaya et~al., 2021]{amaya_new_2021}
Amaya, A., Bach, R., Keusch, F., and Kreuter, F. (2021).
\newblock New {Data} {Sources} in {Social} {Science} {Research}: {Things} to {Know} {Before} {Working} {With} {Reddit} {Data}.
\newblock {\em Social Science Computer Review}, 39(5):943--960.
\newblock https://doi.org/10.1177/0894439319893305.

\bibitem[Androutsopoulos and Ziegler, 2004]{androutsopoulos_exploring_2004}
Androutsopoulos, J. and Ziegler, E. (2004).
\newblock Exploring language variation on the {Internet}: {Regional} speech in a chat community.
\newblock In {\em Language variation in {Europe}: papers from the second international conference on language variation in {Europe}, {ICLaVE}}, volume~2, pages 99--111.

\bibitem[Antonakaki et~al., 2021]{antonakaki_survey_2021}
Antonakaki, D., Fragopoulou, P., and Ioannidis, S. (2021).
\newblock A survey of {Twitter} research: {Data} model, graph structure, sentiment analysis and attacks.
\newblock {\em Expert Systems with Applications}, 164:114006.
\newblock https://doi.org/10.1016/j.eswa.2020.114006.

\bibitem[{AxstaBludsta}, 2011]{axstabludsta_nek_2011}
{AxstaBludsta} (2011).
\newblock Nek {Minute}.
\newblock Retrieved from https://youtu.be/CTZyorJVeqI.

\bibitem[Ballantyne, 2011]{ballantyne_place_2011}
Ballantyne, T. (2011).
\newblock On {Place}, {Space} and {Mobility} in {Nineteenth}-{Century} {New} {Zealand}.
\newblock {\em New Zealand Journal of History}, 45(1):50--70.
\newblock Retrieved from https://muse.jhu.edu/article/879358/.

\bibitem[Ballard et~al., 2025]{ballard_new_2025}
Ballard, E., Charters, H., Meyerhoff, M., and Watson, C. (2025).
\newblock New {Zealand}, {Multicultural} {Auckland} {English}.
\newblock In {\em The {Wiley} {Blackwell} {Encyclopedia} of {World} {Englishes}}, pages 1--10. John Wiley \& Sons, Hoboken, NJ.
\newblock https://doi.org/10.1002/9781119518297.eowe00118.

\bibitem[Bamman et~al., 2014]{bamman_distributed_2014}
Bamman, D., Dyer, C., and Smith, N.~A. (2014).
\newblock Distributed {Representations} of {Geographically} {Situated} {Language}.
\newblock In Toutanova, K. and Wu, H., editors, {\em Proceedings of the 52nd {Annual} {Meeting} of the {Association} for {Computational} {Linguistics} ({Volume} 2: {Short} {Papers})}, pages 828--834, Baltimore, Maryland. Association for Computational Linguistics.

\bibitem[Barber et~al., 1993]{barber_english_1993}
Barber, C.~L., Beal, J.~C., and Shaw, P.~A. (1993).
\newblock {\em The {English} {Language}: {A} {Historical} {Introduction}}.
\newblock Cambridge University Press, New York, NY, 2 edition.

\bibitem[Bardsley, 2006]{bardsley_specialist_2006}
Bardsley, D. (2006).
\newblock A {Specialist} {Study} in {New} {Zealand} {English} {Lexis}: {The} {Rural} {Sector}.
\newblock {\em International Journal of Lexicography}, 19(1):41--72.
\newblock https://doi.org/10.1093/ijl/eci052.

\bibitem[Bardsley, 2009]{bardsley_lexicography_2009}
Bardsley, D. (2009).
\newblock Lexicography in {New} {Zealand}.
\newblock Technical report, New Zealand Dictionary Centre, Wellington, New Zealand.

\bibitem[Bardsley and Simpson, 2009]{bardsley_hypocoristics_2009}
Bardsley, D. and Simpson, J. (2009).
\newblock Hypocoristics in {New} {Zealand} and {Australian} {English}.
\newblock In Peters, P., Collins, P., and Smith, A., editors, {\em Comparative {Studies} in {Australian} and {New} {Zealand} {English}: {Grammar} and beyond}, pages 49--70. John Benjamins Publishing Company, Amsterdam, The Netherlands; Philadelphia, PA.
\newblock https://doi.org/10.1075/veaw.g39.04bar.

\bibitem[Bartlett, 1992]{bartlett_regional_1992}
Bartlett, C.~M. (1992).
\newblock Regional {Variation} in {New} {Zealand} {English}: {The} {Case} of {Southland}.
\newblock {\em New Zealand English Newsletter}, 6:5--15.

\bibitem[Bauer, 1987]{bauer_new_1987}
Bauer, L. (1987).
\newblock New {Zealand} {English} morphology: {Some} experimental evidence.
\newblock {\em Te Reo – The Journal of the Linguistic Society of New Zealand}, 30(1):37--53.
\newblock Retrieved from https://nzlingsoc.org/journal\_article/new-zealand-english-morphology-some-experimental-evidence/.

\bibitem[Bauer, 1994a]{bauer_introducing_1994}
Bauer, L. (1994a).
\newblock Introducing the {Wellington} {Corpus} of {Written} {New} {Zealand} {English}.
\newblock {\em Te Reo – The Journal of the Linguistic Society of New Zealand}, 37:21--28.
\newblock Retrieved from https://nzlingsoc.org/journal\_article/introducing-the-wellington-corpus-of-written-new-zealand-english/.

\bibitem[Bauer, 1994b]{bauer_watching_1994}
Bauer, L. (1994b).
\newblock {\em Watching {English} {Change}: {An} {Introduction} to the {Study} of {Linguistic} {Change} in {Standard} {Englishes} in the 20th {Century}}.
\newblock Longman, Harlow, England.
\newblock https://doi.org/10.4324 /9781315844169.

\bibitem[Bauer, 2007]{bauer_grammatical_2007}
Bauer, L. (2007).
\newblock Some {Grammatical} {Features} of {New} {Zealand} {English}.
\newblock {\em New Zealand English Journal}, 21:1--25.
\newblock Retrieved from https://search.informit.org/doi/10.3316/informit.555852519597653.

\bibitem[Bauer and Bauer, 2002]{bauer_can_2002}
Bauer, L. and Bauer, W. (2002).
\newblock Can we watch regional dialects developing in colonial {English}?: {The} case of {New} {Zealand}.
\newblock {\em English World-Wide}, 23(2):169--193.
\newblock https://doi.org/10.1075/eww.23.2.02bau.

\bibitem[Bauer and Bauer, 2003]{bauer_playground_2003}
Bauer, L. and Bauer, W. (2003).
\newblock {\em Playground {Talk}: {Dialects} and {Change} in {New} {Zealand} {English}}.
\newblock School of Linguistics and Applied Language Studies, Victoria University of Wellington, Wellington, New Zealand.

\bibitem[Baum and Petrie, 1966]{baum_statistical_1966}
Baum, L.~E. and Petrie, T. (1966).
\newblock Statistical {Inference} for {Probabilistic} {Functions} of {Finite} {State} {Markov} {Chains}.
\newblock {\em The Annals of Mathematical Statistics}, 37(6):1554--1563.
\newblock Retrieved from https://www.jstor.org/stable/2238772.

\bibitem[Baumgartner et~al., 2020]{baumgartner_pushshift_2020}
Baumgartner, J., Zannettou, S., Keegan, B., Squire, M., and Blackburn, J. (2020).
\newblock The {Pushshift} {Reddit} {Dataset}.
\newblock In {\em Proceedings of the {International} {AAAI} {Conference} on {Web} and {Social} {Media}}, volume~14, pages 830--839, Atlanta, GA. PKP Publishing Services Network.
\newblock https://doi.org/10.1609/icwsm.v14i1.7347.

\bibitem[Bayard, 1989]{bayard_me_1989}
Bayard, D. (1989).
\newblock ‘{Me} {Say} {That}? {No} {Way}!': {The} social correlates of {American} lexical diffusion in {New} {Zealand} {English}.
\newblock {\em Te Reo}, 32(1):17--60.
\newblock Retrieved from https://nzlingsoc.org/journal\_article/me-say-that-no-way-the-social-correlates-of-american-lexical-diffusion-in-new-zealand-english/.

\bibitem[Bayard, 1991]{bayard_antipodean_1991}
Bayard, D. (1991).
\newblock Antipodean {Accents} and the ``{Cultural} {Cringe}'': {New} {Zealand} and {American} {Attitudes} {Toward} {NZE} and {Other} {English} {Accents}.
\newblock {\em Te Reo}, 34(1):15--52.
\newblock Retrieved from https://nzlingsoc.org/journal\_article/antipodean-accents-and-the-cultural-cringe-new-zealand-and-american-attitudes-toward-nze-and-other-english-accents/.

\bibitem[Beaman, 2021]{beaman_identity_2021}
Beaman, K.~V. (2021).
\newblock Identity and mobility in linguistic change across the lifespan: {The} case of {Swabian} {German}.
\newblock In Ziegler, A., Edler, S., and Oberdorfer, G., editors, {\em Urban {Matters}: {Current} approaches in variationist sociolinguistics}, pages 27--60. John Benjamins Publishing Company, Amsterdam, The Netherlands; Philadelphia, PA.
\newblock https://doi.org/10.1075/silv.27.02bea.

\bibitem[Bell, 1996]{bell_inventing_1996}
Bell, C. (1996).
\newblock {\em Inventing {New} {Zealand}: {Everyday} {Myths} of {Pakeha} {Identity}}.
\newblock Penguin Books, Harmondsworth, England.

\bibitem[Berger and Luckmann, 1966]{berger_social_1966}
Berger, P.~L. and Luckmann, T. (1966).
\newblock {\em The {Social} {Construction} of {Reality}: {A} {Treatise} in the {Sociology} of {Knowledge}}.
\newblock Anchor, London, England; New York, NY; Camberwell, Victoria, Australia; Toronto, Ontario, Canada; New Delhi, India; Auckland, New Zealand; Rosebank, South Africa.

\bibitem[Biber and Conrad, 2009]{biber_register_2009}
Biber, D. and Conrad, S. (2009).
\newblock {\em Register, {Genre}, and {Style}}.
\newblock Cambridge {Textbooks} in {Linguistics}. Cambridge University Press, Cambridge, England.
\newblock https://doi.org/10.1017/CBO9780511814358.

\bibitem[Birznieks, 2020]{birznieks_perpetuation_2020}
Birznieks, L. (2020).
\newblock The {Perpetuation} of {Western} {Dominance} through {Online} {Discourse}: {A} {Critical} {Discourse} {Analysis} of {Reddit} {Comment} {Threads}.
\newblock Retrieved from https://studenttheses.uu.nl/handle/20.500.12932/36478.

\bibitem[Blaschke et~al., 2024]{blaschke_what_2024}
Blaschke, V., Purschke, C., Schuetze, H., and Plank, B. (2024).
\newblock What {Do} {Dialect} {Speakers} {Want}? {A} {Survey} of {Attitudes} {Towards} {Language} {Technology} for {German} {Dialects}.
\newblock In Ku, L.-W., Martins, A., and Srikumar, V., editors, {\em Proceedings of the 62nd {Annual} {Meeting} of the {Association} for {Computational} {Linguistics} ({Volume} 2: {Short} {Papers})}, pages 823--841, Bangkok, Thailand. Association for Computational Linguistics.
\newblock https://doi.org/10.18653/v1/2024.acl-short.74.

\bibitem[Blommaert, 2013]{blommaert_writing_2013}
Blommaert, J. (2013).
\newblock Writing as a sociolinguistic object.
\newblock {\em Journal of Sociolinguistics}, 17(4):440--459.
\newblock https://doi.org/10.1111/josl.12042.

\bibitem[Blondel et~al., 2008]{blondel_fast_2008}
Blondel, V.~D., Guillaume, J.-L., Lambiotte, R., and Lefebvre, E. (2008).
\newblock Fast unfolding of communities in large networks.
\newblock {\em Journal of Statistical Mechanics: Theory and Experiment}, 2008(10):P10008.
\newblock https://dx.doi.org/10.1088/1742-5468/2008/10/P10008.

\bibitem[Bloomfield, 1933]{bloomfield_language_1933}
Bloomfield, L. (1933).
\newblock {\em Language}.
\newblock George Allen \& Unwin, London, England.

\bibitem[Braun and Clarke, 2006]{braun_using_2006}
Braun, V. and Clarke, V. (2006).
\newblock Using thematic analysis in psychology.
\newblock {\em Qualitative Research in Psychology}, 3(2):77--101.
\newblock https://doi.org/10.1191/1478088706qp063oa.

\bibitem[{British Oracle}, 2019]{british_oracle_savvy_2019}
{British Oracle} (2019).
\newblock Savvy {B}.
\newblock Retrieved from http://savvy-b.urbanup.com/14111231.

\bibitem[Bucher, 2013]{bucher_objects_2013}
Bucher, T. (2013).
\newblock Objects of {Intense} {Feeling}: {The} {Case} of the {Twitter} {API}.
\newblock {\em Computational Culture}, 3.
\newblock Retrieved from http://computationalculture.net/objects-of-intense-feeling-the-case-of-the-twitter-api/.

\bibitem[Burrows, 2002]{burrows_delta_2002}
Burrows, J. (2002).
\newblock ‘{Delta}’: a {Measure} of {Stylistic} {Difference} and a {Guide} to {Likely} {Authorship}.
\newblock {\em Literary \& Linguistic Computing}, 17(3):267--287.
\newblock https://doi.org/10.1093/llc/17.3.267.

\bibitem[Cabitza et~al., 2023]{cabitza_toward_2023}
Cabitza, F., Campagner, A., and Basile, V. (2023).
\newblock Toward a {Perspectivist} {Turn} in {Ground} {Truthing} for {Predictive} {Computing}.
\newblock {\em Proceedings of the AAAI Conference on Artificial Intelligence}, 37(6):6860--6868.

\bibitem[Calude, 2023]{calude_linguistics_2023}
Calude, A.~S. (2023).
\newblock {\em The {Linguistics} of {Social} {Media}: {An} {Introduction}}.
\newblock Taylor \& Francis, Abingdon, England; New York, NY.
\newblock https://doi.org/10.4324/9781003321873.

\bibitem[Calude et~al., 2024]{calude_arehashtagswords_2024}
Calude, A.~S., Long, M., and Burnette, J. (2024).
\newblock \#{AreHashtagsWords}? {Structure}, position, and syntactic integration of hashtags in ({English}) tweets.
\newblock {\em Linguistics Vanguard}, 10(1):105--114.
\newblock https://doi.org/10.1515/lingvan-2023-0044.

\bibitem[Cannon, 1985]{cannon_functional_1985}
Cannon, G. (1985).
\newblock Functional shift in {English}.
\newblock {\em Linguistics}, 23(3):411--432.
\newblock https://doi.org/10.1515/ling.1985.23.3.411.

\bibitem[Cannon, 1986]{cannon_blends_1986}
Cannon, G. (1986).
\newblock Blends in {English} word formation.
\newblock {\em Linguistics}, 24(4):725--754.
\newblock https://doi.org/10.1515/ling.1986.24.4.725.

\bibitem[Cannon, 1989]{cannon_abbreviations_1989}
Cannon, G. (1989).
\newblock Abbreviations and {Acronyms} in {English} {Word}-{Formation}.
\newblock {\em American Speech}, 64(2):99--127.
\newblock https://doi.org/10.2307/455038.

\bibitem[Cannon and Bailey, 1986]{cannon_back-formations_1986}
Cannon, G. and Bailey, G. (1986).
\newblock Back-{Formations} in {English} {Word}-{Formation}.
\newblock {\em Meta}, 31(4):427--438.

\bibitem[Carmichael, 2023]{carmichael_locating_2023}
Carmichael, K. (2023).
\newblock Locating place in variationist sociolinguistics: {Making} the case for ethnographically informed multidimensional place orientation metrics.
\newblock {\em Journal of Linguistic Geography}, 11(2):65--77.
\newblock https://doi.org/10.1017/jlg.2023.2.

\bibitem[Carmichael and Reed, 2025]{carmichael_language_2025}
Carmichael, K. and Reed, P.~E. (2025).
\newblock {\em Language and {Place}}.
\newblock Cambridge University Press, Cambridge, England; New York, NY; Port Melbourne, Victoria, Australia; New Delhi, India; Singapore, Singapore.
\newblock https://doi.org/10.1017/9781009380874.

\bibitem[Chambers, 2000]{chambers_region_2000}
Chambers, J.~K. (2000).
\newblock Region and language variation.
\newblock {\em English World-Wide}, 21(2):169--199.
\newblock https://doi.org/10.1075/eww.21.2.02cha.

\bibitem[Chambers and Trudgill, 1998]{chambers_dialectology_1998}
Chambers, J.~K. and Trudgill, P. (1998).
\newblock {\em Dialectology}.
\newblock Cambridge {Textbooks} in {Linguistics}. Cambridge University Press, Cambridge, England, 2 edition.
\newblock https://doi.org/10.1017/CBO9780511805103.

\bibitem[Charmaz, 2006]{charmaz_constructing_2006}
Charmaz, K. (2006).
\newblock {\em Constructing {Grounded} {Theory}: {A} {Practical} {Guide} through {Qualitative} {Analysis}}.
\newblock Introducing {Qualitative} {Methods}. SAGE Publications, London, England; Thousand Oaks, CA; New Delhi, India.

\bibitem[Chomsky, 1965]{chomsky_aspects_1965}
Chomsky, N. (1965).
\newblock {\em Aspects of the {Theory} of {Syntax}}.
\newblock The MIT Press, Cambridge, MA.

\bibitem[Church and Hanks, 1990]{church_word_1990}
Church, K.~W. and Hanks, P. (1990).
\newblock Word {Association} {Norms}, {Mutual} {Information}, and {Lexicography}.
\newblock {\em Computational Linguistics}, 16(1):22--29.
\newblock Retrieved from https://aclanthology.org/J90-1003/.

\bibitem[Conneau et~al., 2020]{conneau_unsupervised_2020}
Conneau, A., Khandelwal, K., Goyal, N., Chaudhary, V., Wenzek, G., Guzmán, F., Grave, E., Ott, M., Zettlemoyer, L., and Stoyanov, V. (2020).
\newblock Unsupervised {Cross}-lingual {Representation} {Learning} at {Scale}.
\newblock In Jurafsky, D., Chai, J., Schluter, N., and Tetreault, J., editors, {\em Proceedings of the 58th {Annual} {Meeting} of the {Association} for {Computational} {Linguistics}}, pages 8440--8451, Online. Association for Computational Linguistics.
\newblock https://doi.org/10.18653/v1/2020.acl-main.747.

\bibitem[Coupland, 2001]{coupland_introduction_2001}
Coupland, N. (2001).
\newblock Introduction: {Sociolinguistic} {Theory} and {Social} {Theory}.
\newblock In Coupland, N., Sarangi, S., and Candlin, C.~N., editors, {\em Sociolinguistics and {Social} {Theory}}, Language in {Social} {Life} {Series}, pages 1--26. Longman, Harlow, England.

\bibitem[Csárdi and Nepusz, 2006]{csardi_igraph_2006}
Csárdi, G. and Nepusz, T. (2006).
\newblock The igraph software package for complex network research.
\newblock {\em InterJournal, Complex Systems}, 1695:1--9.
\newblock Retrieved from https://igraph.org/.

\bibitem[Cuming, 2013]{cuming_marmageddon_2013}
Cuming, A. (2013).
\newblock Marmageddon no more.
\newblock {\em Waikato Times}.
\newblock Retrieved from https://www.stuff.co.nz/waikato-times/news/8450129/Marmageddon-no-more.

\bibitem[Cutler, 2020]{cutler_metapragmatic_2020}
Cutler, C. (2020).
\newblock Metapragmatic comments and orthographic performances of a {New} {York} accent on {YouTube}.
\newblock {\em World Englishes}, 39(1):36--53.
\newblock https://doi.org/10.1111/weng.12444.

\bibitem[Cutler et~al., 2022a]{cutler_digital_2022}
Cutler, C., Ahmar, M., and Bahri, S., editors (2022a).
\newblock {\em Digital {Orality}: {Vernacular} {Writing} in {Online} {Spaces}}.
\newblock Springer International Publishing, Cham, Switzerland.
\newblock https://doi.org/10.1007/978-3-031-10433-6.

\bibitem[Cutler et~al., 2022b]{cutler_introduction_2022}
Cutler, C., Ahmar, M., and Bahri, S. (2022b).
\newblock Introduction: {The} {Oralization} of {Digital} {Written} {Communication}.
\newblock In Cutler, C., Ahmar, M., and Bahri, S., editors, {\em Digital {Orality}: {Vernacular} {Writing} in {Online} {Spaces}}, pages 3--31. Springer International Publishing, Cham, Switzerland.
\newblock https://doi.org/10.1007/978-3-031-10433-6\_1.

\bibitem[Danescu-Niculescu-Mizil et~al., 2011]{danescu-niculescu-mizil_mark_2011}
Danescu-Niculescu-Mizil, C., Gamon, M., and Dumais, S. (2011).
\newblock Mark my words! linguistic style accommodation in social media.
\newblock In {\em Proceedings of the 20th international conference on {World} wide web}, {WWW} '11, pages 745--754, New York, NY, USA. Association for Computing Machinery.
\newblock https://doi.org/10.1145/1963405.1963509.

\bibitem[De~Bres, 2010]{de_bres_attitudes_2010}
De~Bres, J. (2010).
\newblock Attitudes of non-{Maori} {New} {Zealanders} towards the use of {Maori} in {New} {Zealand} {English}.
\newblock {\em New Zealand English Journal}, 24:2--14.
\newblock Retrieved from https://search.informit.org/doi/10.3316/informit.206778324257255.

\bibitem[de~Bres and Nicholas, 2021]{de_bres_sexiest_2021}
de~Bres, J. and Nicholas, S.~A. (2021).
\newblock The sexiest accent in the world: {Linguistic} insecurity and prejudice in media coverage of the {New} {Zealand} accent.
\newblock {\em Te Reo – The Journal of the Linguistic Society of New Zealand}, 64(1):15--32.
\newblock Retrieved from https://nzlingsoc.org/journal\_article/the-sexiest-accent-in-the-world-linguistic-insecurity-and-prejudice-in-media-coverage-of-the-new-zealand-accent/.

\bibitem[Degani, 2012]{degani_language_2012}
Degani, M. (2012).
\newblock Language contact in {New} {Zealand}: {A} focus on {English} lexical borrowings in {Māori}.
\newblock {\em Academic Journal of Modern Philology}, 1:13--24.

\bibitem[Desmarais, 2020]{desmarais_men_2020}
Desmarais, A.-M. (2020).
\newblock Men who knit: {A} social media critical discourse study ({SM}-{CDS}) on the legitimisation of men within {Reddit}’sr/knitting community.
\newblock Retrieved from https://hdl.handle.net/10292/13594.

\bibitem[Deverson, 2000]{deverson_handling_2000}
Deverson, T. (2000).
\newblock Handling {New} {Zealand} {English} lexis.
\newblock In Bell, A. and Kuiper, K., editors, {\em New {Zealand} {English}}, pages 23--39. Victoria University Press, Amsterdam, The Netherlands; Philadelphia, PA.

\bibitem[Deverson and Kennedy, 2005a]{deverson_bush_2005}
Deverson, T. and Kennedy, G. (2005a).
\newblock bush.
\newblock In {\em The {New} {Zealand} {Oxford} {Dictionary}}. Oxford University Press.
\newblock https://doi.org/10.1093/acref/9780195584516.001.0001.

\bibitem[Deverson and Kennedy, 2005b]{deverson_chch_2005}
Deverson, T. and Kennedy, G. (2005b).
\newblock {ChCh}.
\newblock In {\em The {New} {Zealand} {Oxford} {Dictionary}}. Oxford University Press.
\newblock https://doi.org/10.1093/acref/9780195584516.001.0001.

\bibitem[Deverson and Kennedy, 2005c]{deverson_feijoa_2005}
Deverson, T. and Kennedy, G. (2005c).
\newblock feijoa.
\newblock In {\em The {New} {Zealand} {Oxford} {Dictionary}}. Oxford University Press.
\newblock https://doi.org/10.1093/acref/9780195584516.001.0001.

\bibitem[Deverson and Kennedy, 2005d]{deverson_grundies_2005}
Deverson, T. and Kennedy, G. (2005d).
\newblock grundies.
\newblock In {\em The {New} {Zealand} {Oxford} {Dictionary}}. Oxford University Press.
\newblock https://doi.org/10.1093/acref/9780195584516.001.0001.

\bibitem[Deverson and Kennedy, 2005e]{deverson_jafa_2005}
Deverson, T. and Kennedy, G. (2005e).
\newblock {JAFA}.
\newblock In {\em The {New} {Zealand} {Oxford} {Dictionary}}. Oxford University Press.
\newblock https://doi.org/10.1093/acref/9780195584516.001.0001.

\bibitem[Deverson and Kennedy, 2005f]{deverson_ute_2005}
Deverson, T. and Kennedy, G. (2005f).
\newblock ute.
\newblock In {\em The {New} {Zealand} {Oxford} {Dictionary}}. Oxford University Press.
\newblock https://doi.org/10.1093/acref/9780195584516.001.0001.

\bibitem[Devlin et~al., 2019]{devlin_bert_2019}
Devlin, J., Chang, M.-W., Lee, K., and Toutanova, K. (2019).
\newblock {BERT}: {Pre}-training of {Deep} {Bidirectional} {Transformers} for {Language} {Understanding}.
\newblock In Burstein, J., Doran, C., and Solorio, T., editors, {\em Proceedings of the 2019 {Conference} of the {North} {American} {Chapter} of the {Association} for {Computational} {Linguistics}: {Human} {Language} {Technologies}}, volume~1 of {\em Long and {Short} {Papers}}, pages 4171--4186, Minneapolis, MN. Association for Computational Linguistics.
\newblock https://doi.org/10.18653/v1/N19-1423.

\bibitem[Di~Sciullo and Williams, 1987]{di_sciullo_definition_1987}
Di~Sciullo, A.-M. and Williams, E. (1987).
\newblock {\em On the definition of word}.
\newblock Number~14 in Linguistic inquiry monographs. The MIT Press, Cambridge, MA.

\bibitem[Dijkstra et~al., 2021]{dijkstra_using_2021}
Dijkstra, J., Heeringa, W., Jongbloed-Faber, L., and Van~de Velde, H. (2021).
\newblock Using {Twitter} {Data} for the {Study} of {Language} {Change} in {Low}-{Resource} {Languages}. {A} {Panel} {Study} of {Relative} {Pronouns} in {Frisian}.
\newblock {\em Frontiers in Artificial Intelligence}, 4.
\newblock https://doi.org/10.3389/frai.2021.644554.

\bibitem[Donald, 2018]{donald_its_2018}
Donald, S. (2018).
\newblock It's a colonial thing: {New} {Zealand} cultural identity and the use of 'colony' as a social category in intercultural communication.
\newblock {\em New Zealand Studies in Applied Linguistics}, 24(1):5--17.
\newblock Retrieved from https://search.informit.org/doi/abs/10.3316/INFORMIT.740485668341927.

\bibitem[Dourish, 2004]{dourish_what_2004}
Dourish, P. (2004).
\newblock What we talk about when we talk about context.
\newblock {\em Personal and Ubiquitous Computing}, 8(1):19--30.
\newblock https://doi.org/10.1007/s00779-003-0253-8.

\bibitem[Dourish, 2006]{dourish_re-space-ing_2006}
Dourish, P. (2006).
\newblock Re-space-ing place: "place" and "space" ten years on.
\newblock In {\em Proceedings of the 2006 20th anniversary conference on {Computer} supported cooperative work}, pages 299--308, New York, NY. Association for Computing Machinery.
\newblock https://doi.org/10.1145/1180875.1180921.

\bibitem[Duhamel and Meyerhoff, 2015]{duhamel_end_2015}
Duhamel, M.-F. and Meyerhoff, M. (2015).
\newblock An end of egalitarianism? {Social} evaluations of language difference in {New} {Zealand}.
\newblock {\em Linguistics Vanguard}, 1(1):235--248.
\newblock https://doi.org/10.1515/lingvan-2014-1005.

\bibitem[Dunn, 2017]{dunn_computational_2017}
Dunn, J. (2017).
\newblock Computational learning of construction grammars.
\newblock {\em Language and Cognition}, 9(2):254--292.
\newblock https://doi.org/10.1017/langcog.2016.7.

\bibitem[Dunn, 2019a]{dunn_global_2019}
Dunn, J. (2019a).
\newblock Global {Syntactic} {Variation} in {Seven} {Languages}: {Toward} a {Computational} {Dialectology}.
\newblock {\em Frontiers in Artificial Intelligence}, 2:15.
\newblock https://doi.org/10.3389/frai.2019.00015.

\bibitem[Dunn, 2019b]{dunn_modeling_2019}
Dunn, J. (2019b).
\newblock Modeling {Global} {Syntactic} {Variation} in {English} {Using} {Dialect} {Classification}.
\newblock In Zampieri, M., Nakov, P., Malmasi, S., Ljubešić, N., Tiedemann, J., and Ali, A., editors, {\em Proceedings of the {Sixth} {Workshop} on {NLP} for {Similar} {Languages}, {Varieties} and {Dialects}}, pages 42--53, Ann Arbor, Michigan. Association for Computational Linguistics.
\newblock https://doi.org/10.18653/v1/W19-1405.

\bibitem[Dunn, 2020]{dunn_mapping_2020}
Dunn, J. (2020).
\newblock Mapping languages: the {Corpus} of {Global} {Language} {Use}.
\newblock {\em Language Resources and Evaluation}, 54(4):999--1018.
\newblock https://doi.org/10.1007/s10579-020-09489-2.

\bibitem[Dunn, 2022]{dunn_natural_2022}
Dunn, J. (2022).
\newblock {\em Natural {Language} {Processing} for {Corpus} {Linguistics}}.
\newblock Cambridge University Press, Cambridge, England; New York, NY; Port Melbourne, Victoria, Australia; New Delhi, India; Singapore, Singapore.
\newblock https://doi.org/10.1017/9781009070447.

\bibitem[Dunn, 2023]{dunn_variation_2023}
Dunn, J. (2023).
\newblock Variation and {Instability} in {Dialect}-{Based} {Embedding} {Spaces}.
\newblock In Scherrer, Y., Jauhiainen, T., Ljubešić, N., Nakov, P., Tiedemann, J., and Zampieri, M., editors, {\em Tenth {Workshop} on {NLP} for {Similar} {Languages}, {Varieties} and {Dialects} ({VarDial} 2023)}, pages 67--77, Dubrovnik, Croatia. Association for Computational Linguistics.

\bibitem[Dunn, 2024]{dunn_computational_2024}
Dunn, J. (2024).
\newblock {\em Computational {Construction} {Grammar}: {A} {Usage}-{Based} {Approach}}.
\newblock Cambridge University Press, Cambridge, England; New York, NY; Port Melbourne, Victoria, Australia; New Delhi, India; Singapore, Singapore.
\newblock https://doi.org/10.1017/9781009233743.

\bibitem[Dunn and Adams, 2019]{dunn_mapping_2019}
Dunn, J. and Adams, B. (2019).
\newblock Mapping {Languages} and {Demographics} with {Georeferenced} {Corpora}.
\newblock In {\em Proceedings of {GeoComputation} 2019}, pages 1--16, Queenstown, New Zealand. Centre for Computational Geography.
\newblock https://doi.org/10.17608/k6.auckland.9869252.v2.

\bibitem[Dunn and Adams, 2020]{dunn_geographically-balanced_2020}
Dunn, J. and Adams, B. (2020).
\newblock Geographically-{Balanced} {Gigaword} {Corpora} for 50 {Language} {Varieties}.
\newblock In Calzolari, N., Béchet, F., Blache, P., Choukri, K., Cieri, C., Declerck, T., Goggi, S., Isahara, H., Maegaard, B., Mariani, J., Mazo, H., Moreno, A., Odijk, J., and Piperidis, S., editors, {\em Proceedings of the {Twelfth} {Language} {Resources} and {Evaluation} {Conference}}, pages 2528--2536, Marseille, France. European Language Resources Association.
\newblock Retrieved from https://aclanthology.org/2020.lrec-1.308/.

\bibitem[Dunn et~al., 2020]{dunn_measuring_2020}
Dunn, J., Coupe, T., and Adams, B. (2020).
\newblock Measuring {Linguistic} {Diversity} {During} {COVID}-19.
\newblock In {\em Proceedings of {The} {Fourth} {Workshop} on the {Fourth} {Workshop} on {Natural} {Language} {Processing} and {Computational} {Social} {Science}}, pages 1--10. Association for Computational Linguistics.
\newblock https://doi.org/10.18653/v1/2020.nlpcss-1.1.

\bibitem[Dunn and Wong, 2022]{dunn_stability_2022}
Dunn, J. and Wong, S. (2022).
\newblock Stability of {Syntactic} {Dialect} {Classification} over {Space} and {Time}.
\newblock In {\em Proceedings of the 29th {International} {Conference} on {Computational} {Linguistics}}, pages 26--36. International Committee on Computational Linguistics.
\newblock https://aclanthology.org/2022.coling-1.3/.

\bibitem[Dunn and Wong, 2025]{dunn_language_2025}
Dunn, J. and Wong, S. (2025).
\newblock Language {Contact} and {Population} {Contact} as {Sources} of {Dialect} {Similarity}.
\newblock {\em Languages}, 10(8):188.
\newblock https://doi.org/10.3390/languages10080188.

\bibitem[Duranti, 2003]{duranti_language_2003}
Duranti, A. (2003).
\newblock Language as {Culture} in {U}.{S}. {Anthropology}: {Three} {Paradigms}.
\newblock {\em Current Anthropology}, 44(3):323--347.
\newblock https://doi.org/10.1086/368118.

\bibitem[Díaz-Campos and Balasch, 2023]{diaz-campos_handbook_2023}
Díaz-Campos, M. and Balasch, S. (2023).
\newblock {\em The {Handbook} of {Usage}‐{Based} {Linguistics}}.
\newblock John Wiley \& Sons, Hoboken, NJ.
\newblock https://doi.org/10.1002/9781119839859.

\bibitem[Eckert, 2012]{eckert_three_2012}
Eckert, P. (2012).
\newblock Three {Waves} of {Variation} {Study}: {The} {Emergence} of {Meaning} in the {Study} of {Sociolinguistic} {Variation}.
\newblock {\em Annual Review of Anthropology}, 41:87--100.
\newblock https://doi.org/10.1146/annurev-anthro-092611-145828.

\bibitem[Eckert and Wenger, 2005]{eckert_communities_2005}
Eckert, P. and Wenger, E. (2005).
\newblock Communities of practice in sociolinguistics.
\newblock {\em Journal of Sociolinguistics}, 9(4):582--589.
\newblock https://doi.org/10.1111/j.1360-6441.2005.00307.x.

\bibitem[Edosomwan et~al., 2011]{edosomwan_history_2011}
Edosomwan, S., Prakasan, S.~K., Kouame, D., Watson, J., and Seymour, T. (2011).
\newblock The history of social media and its impact on business.
\newblock {\em Journal of Applied Management and entrepreneurship}, 16(3):79--91.

\bibitem[Eisenstein et~al., 2010]{eisenstein_latent_2010}
Eisenstein, J., O'Connor, B., Smith, N.~A., and Xing, E.~P. (2010).
\newblock A {Latent} {Variable} {Model} for {Geographic} {Lexical} {Variation}.
\newblock In Li, H. and Màrquez, L., editors, {\em Proceedings of the 2010 {Conference} on {Empirical} {Methods} in {Natural} {Language} {Processing}}, pages 1277--1287. Association for Computational Linguistics.
\newblock https://aclanthology.org/D10-1124/.

\bibitem[Eisenstein et~al., 2014]{eisenstein_diffusion_2014}
Eisenstein, J., O'Connor, B., Smith, N.~A., and Xing, E.~P. (2014).
\newblock Diffusion of {Lexical} {Change} in {Social} {Media}.
\newblock {\em PLOS ONE}, 9(11):e113114.
\newblock https://doi.org/10.1371/journal.pone.0113114.

\bibitem[Ellis and Larsen-Freeman, 2009]{ellis_language_2009}
Ellis, N.~C. and Larsen-Freeman, D. (2009).
\newblock {\em Language as a {Complex} {Adaptive} {System}}.
\newblock John Wiley \& Sons.

\bibitem[Evans, 2015]{evans_locative_2015}
Evans, L. (2015).
\newblock {\em Locative {Social} {Media}}.
\newblock Palgrave Macmillan, Basingstoke, England; New York, NY.
\newblock https://doi.org/10.10 7/9781137456113.

\bibitem[Faisal et~al., 2024]{faisal_dialectbench_2024}
Faisal, F., Ahia, O., Srivastava, A., Ahuja, K., Chiang, D., Tsvetkov, Y., and Anastasopoulos, A. (2024).
\newblock {DIALECTBENCH}: {An} {NLP} {Benchmark} for {Dialects}, {Varieties}, and {Closely}-{Related} {Languages}.
\newblock In Ku, L.-W., Martins, A., and Srikumar, V., editors, {\em Proceedings of the 62nd {Annual} {Meeting} of the {Association} for {Computational} {Linguistics} ({Volume} 1: {Long} {Papers})}, pages 14412--14454, Bangkok, Thailand. Association for Computational Linguistics.
\newblock https://doi.org/10.18653/v1/2024.acl-long.777.

\bibitem[Fandrych, 2004]{fandrych_non-morphematic_2004}
Fandrych, I.~M. (2004).
\newblock Non-morphematic word-formation processes: a multi-level approach to acronyms, blends, clippings and onomatopoeia.
\newblock Retrieved from http://hdl.handle.net/11660/2696.

\bibitem[Ferrer et~al., 2021]{ferrer_discovering_2021}
Ferrer, X., Nuenen, T.~v., Such, J.~M., and Criado, N. (2021).
\newblock Discovering and {Categorising} {Language} {Biases} in {Reddit}.
\newblock {\em Proceedings of the International AAAI Conference on Web and Social Media}, 15(1):140--151.
\newblock https://doi.org/10.1609/icwsm.v15i1.18048.

\bibitem[Figueroa and Zeng-Treitler, 2013]{figueroa_text_2013}
Figueroa, R.~L. and Zeng-Treitler, Q. (2013).
\newblock Text {Classification} {Performance}: {Is} the {Sample} {Size} the {Only} {Factor} to be {Considered}?
\newblock In {\em {MEDINFO} 2013}, volume 192 of {\em Studies in {Health} {Technology} and {Informatics}}, pages 1193--1193. IOS Press.
\newblock https://doi.org/10.3233/978-1-61499-289-9-1193.

\bibitem[Fishman, 1971]{fishman_sociology_1971}
Fishman, J.~A. (1971).
\newblock The {Sociology} {Of} {Language}: {An} {Interdisciplinary} {Social} {Science} {Approach} to {Language} in {Society}.
\newblock In Fishman, J.~A., editor, {\em Advances in the {Sociology} of {Language}}, volume Volume 1 Basic concepts, theories and problems: alternative approaches, pages 217--404. Mouton, The Hague, The Netherlands; Paris, France.

\bibitem[Ford, 2018]{ford_rethinking_2018}
Ford, G. (2018).
\newblock Rethinking lay people’s theories of the economy.
\newblock http://dx.doi.org/10.26021/5163.

\bibitem[Francis and Kučera, 1979]{francis_brown_1979}
Francis, W.~N. and Kučera, H. (1979).
\newblock Brown corpus manual.
\newblock {\em Letters to the Editor}, 5(2):7.
\newblock Retrieved from https://listings.lib.msu.edu/public-corpora/cd421/manuals/brown/INDEX.HTM.

\bibitem[Gallie, 1956]{gallie_essentially_1956}
Gallie, W.~B. (1956).
\newblock Essentially {Contested} {Concepts}.
\newblock In {\em Proceedings of the {Aristotelian} {Society}}, volume~56, pages 167--198, London, England. The Aristotelian Society.
\newblock Retrieved from https://www.jstor.org/stable/4544562.

\bibitem[Ganin and Lempitsky, 2015]{ganin_unsupervised_2015}
Ganin, Y. and Lempitsky, V. (2015).
\newblock Unsupervised {Domain} {Adaptation} by {Backpropagation}.
\newblock In {\em Proceedings of the 32nd {International} {Conference} on {Machine} {Learning}}, volume~37, pages 1180--1189, Lille, France. PMLR.
\newblock Retrieved from https://proceedings.mlr.press/v37/ganin15.

\bibitem[Gee, 1991]{gee_linguistic_1991}
Gee, J.~P. (1991).
\newblock A {Linguistic} {Approach} to {Narrative}.
\newblock {\em Journal of Narrative and Life History}, 1(1):15--39.
\newblock https://doi.org/10.1075/jnlh.1.1.03ali.

\bibitem[Gee, 2005]{gee_introduction_2005}
Gee, J.~P. (2005).
\newblock {\em Introduction to {Discourse} {Analysis}: {Theory} and {Method}}.
\newblock Taylor \& Francis, London, England; New York, NY, 2 edition.

\bibitem[Gee, 2014]{gee_situated_2014}
Gee, J.~P. (2014).
\newblock Situated {Meaning}.
\newblock In {\em Unified {Discourse} {Analysis}}, pages 48--56. Routledge, London, England.

\bibitem[Ghaseminejad~Raeini, 2025]{ghaseminejad_raeini_evolution_2025}
Ghaseminejad~Raeini, M. (2025).
\newblock The evolution of language models: {From} {N}-{Grams} to {LLMs}, and beyond.
\newblock {\em Natural Language Processing Journal}, 12:100168.
\newblock https://doi.org/10.1016/j.nlp.2025.100168.

\bibitem[Gieryn, 2000]{gieryn_space_2000}
Gieryn, T.~F. (2000).
\newblock A {Space} for {Place} in {Sociology}.
\newblock {\em Annual Review of Sociology}, 26(Volume 26, 2000):463--496.
\newblock https://doi.org/10.1146/annurev.soc.26.1.463.

\bibitem[Gliniecka, 2023]{gliniecka_ethics_2023}
Gliniecka, M. (2023).
\newblock The {Ethics} of {Publicly} {Available} {Data} {Research}: {A} {Situated} {Ethics} {Framework} for {Reddit}.
\newblock {\em Social Media + Society}, 9(3):20563051231192021.
\newblock https://doi.org/10.1177/20563051231192021.

\bibitem[Gloderer, 1993]{gloderer_morphological_1993}
Gloderer, G. (1993).
\newblock {\em Morphological {Regularization} of {Irregular} {Verbs}: {A} {Comparison} of {British} and {American} {English}}.
\newblock University of Freiburg, Freiburg, Germany.

\bibitem[Goebl, 1993]{goebl_dialectometry_1993}
Goebl, H. (1993).
\newblock Dialectometry: {A} {Short} {Overview} of the {Principles} and {Practice} of {Quantitative} {Classification} of {Linguistic} {Atlas} {Data}.
\newblock In Köhler, R. and Riegar, B.~B., editors, {\em Contributions to {Quantitative} {Linguistics}}, pages 277--315. Springer Science+Business Media, Dordrecht, The Netherlands.
\newblock https://doi.org/10.1007/978-94-011-1769-2\_20.

\bibitem[Goebl, 2006]{goebl_recent_2006}
Goebl, H. (2006).
\newblock Recent {Advances} in {Salzburg} {Dialectometry}.
\newblock {\em Literary and Linguistic Computing}, 21(4):411--435.
\newblock https://doi.org/10.1093/llc/fql042.

\bibitem[Goldberg, 2006]{goldberg_constructions_2006}
Goldberg, A.~E. (2006).
\newblock {\em Constructions at {Work}: {The} {Nature} of {Generalization} in {Language}}.
\newblock Oxford University Press, Oxford, England.

\bibitem[Gonçalves et~al., 2018]{goncalves_mapping_2018}
Gonçalves, B., Loureiro-Porto, L., Ramasco, J.~J., and Sánchez, D. (2018).
\newblock Mapping the {Americanization} of {English} in space and time.
\newblock {\em PLOS ONE}, 13(5):e0197741.
\newblock https://doi.org/10.1371/journal.pone.0197741.

\bibitem[{Google}, 2025]{google_google_2025}
{Google} (2025).
\newblock Google {Laboratory}.
\newblock Retrieved from https://colab.research.google.com/.

\bibitem[Gordan, 1957]{gordan_hunting_1957}
Gordan, I.~A. (1957).
\newblock Hunting {New} {Zealandisms}.
\newblock {\em New Zealand Listener}, page~4.

\bibitem[Gordon et~al., 2004]{gordon_new_2004}
Gordon, E., Hay, J., Campbell, L., Maclagan, M., Sudbury, A., and Trudgill, P. (2004).
\newblock {\em New {Zealand} {English}: its origins and evolution}.
\newblock Cambridge University Press, Cambridge, England; New York, NY.
\newblock https://doi.org/10.1017/CBO9780511486678.007.

\bibitem[Gordon, 1980]{gordon_word_1980}
Gordon, I. (1980).
\newblock {\em A {Word} in {Your} {Ear}}.
\newblock Heinemann, Auckland, New Zealand.

\bibitem[Graham et~al., 2014]{graham_where_2014}
Graham, M., Hale, S.~A., and Gaffney, D. (2014).
\newblock Where in the {World} {Are} {You}? {Geolocation} and {Language} {Identification} in {Twitter}.
\newblock {\em The Professional Geographer}, 66(4):568--578.
\newblock https://doi.org/10.1080/00330124.2014.907699.

\bibitem[Graham, 1998]{graham_end_1998}
Graham, S. (1998).
\newblock The end of geography or the explosion of place? {Conceptualizing} space, place and information technology.
\newblock {\em Progress in Human Geography}, 22(2):165--185.
\newblock http://dx.doi.org/10.1191/030913298671334137.

\bibitem[Grieve et~al., 2025]{grieve_sociolinguistic_2025}
Grieve, J., Bartl, S., Fuoli, M., Grafmiller, J., Huang, W., Jawerbaum, A., Murakami, A., Perlman, M., Roemling, D., and Winter, B. (2025).
\newblock The sociolinguistic foundations of language modeling.
\newblock {\em Frontiers in Artificial Intelligence}, 7:1472411.
\newblock https://doi.org/10.3389/frai.2024.1472411.

\bibitem[Grieve et~al., 2019]{grieve_mapping_2019}
Grieve, J., Montgomery, C., Nini, A., Murakami, A., and Guo, D. (2019).
\newblock Mapping {Lexical} {Dialect} {Variation} in {British} {English} {Using} {Twitter}.
\newblock {\em Frontiers in Artificial Intelligence}, 2:11.
\newblock https://doi.org/10.3389/frai.2019.00011.

\bibitem[Grieve et~al., 2018]{grieve_mapping_2018}
Grieve, J., Nini, A., and Guo, D. (2018).
\newblock Mapping {Lexical} {Innovation} on {American} {Social} {Media}.
\newblock {\em Journal of English Linguistics}, 46(4):293--319.
\newblock https://doi.org/10.1177/0075424218793191.

\bibitem[Grootaers, 1962]{grootaers_new_1962}
Grootaers, W.~A. (1962).
\newblock New {Methods} to {Interpret} {Linguistic} {Maps}.
\newblock In {\em Proceedings of the {Ninth} {International} {Congress} of {Linguists}}, pages 259--259. De Gruyter Mouton, Cambridge, MA.
\newblock https://doi.org/10.1515/9783112317419-056.

\bibitem[{Group Think}, 2024]{group_think_dont_2024}
{Group Think} (2024).
\newblock Don’t be a dingus: {The} best {New} {Zealand} schoolyard insults that time forgot.
\newblock {\em The Spinoff}.
\newblock Retrieved from https://thespinoff.co.nz/society/26-01-2024/dont-be-a-dingus-new-zealands-best-schoolyard-insults-that-time-forgot.

\bibitem[Gumperz, 1962]{gumperz_types_1962}
Gumperz, J.~J. (1962).
\newblock Types of {Linguistic} {Communities}.
\newblock {\em Anthropological Linguistics}, 4(1):28--40.

\bibitem[Gumperz, 1968]{gumperz_speech_1968}
Gumperz, J.~J. (1968).
\newblock The {Speech} {Community}.
\newblock In {\em International {Encyclopedia} of the {Social} {Sciences}}, pages 381--386. Macmillan, New York, NY.

\bibitem[Gumperz, 1982]{gumperz_discourse_1982}
Gumperz, J.~J. (1982).
\newblock {\em Discourse {Strategies}}.
\newblock Number~1 in Studies in {Interactional} {Sociolinguistics}. Cambridge University Press, Cambridge, England.
\newblock https://doi.org/10.1017/CBO9780511611834.

\bibitem[Gumperz, 1996]{gumperz_introduction_1996}
Gumperz, J.~J. (1996).
\newblock Introduction to part {IV}.
\newblock In Gumperz, J.~J. and Levinson, S.~C., editors, {\em Rethinking {Linguistic} {Relativity}}. Cambridge University Press, Cambridge, England.

\bibitem[Hagen, 2023]{hagen_no_2023}
Hagen, S. (2023).
\newblock No {Space} for {Reddit} {Spacing}: {Mapping} the {Reflexive} {Relationship} {Between} {Groups} on 4chan and {Reddit}.
\newblock {\em Social Media + Society}, 9(4):20563051231216960.
\newblock https://doi.org/10.1177/20563051231216960.

\bibitem[Halliday and Matthiessen, 2004]{halliday_introduction_2004}
Halliday, M. and Matthiessen, C. (2004).
\newblock {\em An {Introduction} to {Functional} {Grammar}}.
\newblock Routledge, London, England; New York, NY, 3 edition.
\newblock https://doi.org/10.4324/9780203783771.

\bibitem[Hamilton, 2023]{hamilton_blue_2023}
Hamilton, P. (2023).
\newblock Blue {Grinch} / {That} {Feeling} {When} {Knee} {Surgery} {Is} {Tomorrow}.
\newblock Retrieved May 23, 2025, from https://knowyourmeme.com/memes/blue-grinch-that-feeling-when-knee-surgery-is-tomorrow/.

\bibitem[Hamilton et~al., 2016]{hamilton_diachronic_2016}
Hamilton, W.~L., Leskovec, J., and Jurafsky, D. (2016).
\newblock Diachronic {Word} {Embeddings} {Reveal} {Statistical} {Laws} of {Semantic} {Change}.
\newblock In Erk, K. and Smith, N.~A., editors, {\em Proceedings of the 54th {Annual} {Meeting} of the {Association} for {Computational} {Linguistics}}, volume~1 of {\em Long {Papers}}, pages 1489--1501, Berlin, Germany. Association for Computational Linguistics.
\newblock https://doi.org/10.18653/v1/P16-1141.

\bibitem[Hamre, 2024]{hamre_geographic_2024}
Hamre, C. (2024).
\newblock {\em A {Geographic} {Analysis} of {Lexical} {Variation} in {North} {American} {English} {Using} {Reddit} {Corpora}}.
\newblock Doctoral dissertation, Available from ProQuest Dissertations \& Theses Global database.
\newblock (UMI No. 30820387).

\bibitem[Harris et~al., 2020]{harris_array_2020}
Harris, C.~R., Millman, K.~J., van~der Walt, S.~J., Gommers, R., Virtanen, P., Cournapeau, D., Wieser, E., Taylor, J., Berg, S., Smith, N.~J., Kern, R., Picus, M., Hoyer, S., van Kerkwijk, M.~H., Brett, M., Haldane, A., del Río, J.~F., Wiebe, M., Peterson, P., Gérard-Marchant, P., Sheppard, K., Reddy, T., Weckesser, W., Abbasi, H., Gohlke, C., and Oliphant, T.~E. (2020).
\newblock Array programming with {NumPy}.
\newblock {\em Nature}, 585(7825):357--362.
\newblock https://doi.org/10.1038/s41586-020-2649-2.

\bibitem[Harris, 1954]{harris_distributional_1954}
Harris, Z.~S. (1954).
\newblock Distributional {Structure}.
\newblock {\em WORD}, 10(2-3):146--162.
\newblock https://doi.org/10.1080/00437956.1954.11659520.

\bibitem[Haugen, 1950]{haugen_analysis_1950}
Haugen, E. (1950).
\newblock The {Analysis} of {Linguistic} {Borrowing}.
\newblock {\em Language}, 26(2):210--231.
\newblock https://doi.org/10.2307/410058.

\bibitem[Hay et~al., 2008]{hay_new_2008}
Hay, J., Maclagan, M.~A., and Gordon, E. (2008).
\newblock {\em New {Zealand} {English}}.
\newblock Dialects of {English}. Edinburgh University Press, Edinburgh, Scotland.
\newblock https://doi.org/10.1515/9780748630882.

\bibitem[He et~al., 2025]{he_platform_2025}
He, Q., Hong, Y., and Raghu, T.~S. (2025).
\newblock Platform {Governance} with {Algorithm}-{Based} {Content} {Moderation}: {An} {Empirical} {Study} on {Reddit}.
\newblock {\em Information Systems Research}, 36(2):1078--1095.
\newblock https://doi.org/10.1287/isre.2021.0036.

\bibitem[Hearst, 1994]{hearst_review_1994}
Hearst, M.~A. (1994).
\newblock A {Review} of {Statistical} {Language} {Learning}.
\newblock {\em AI Magazine}, 15(3):88--88.
\newblock https://doi.org/10.1609/aimag.v15i3.1106.

\bibitem[Hedderich et~al., 2019]{hedderich_using_2019}
Hedderich, M.~A., Yates, A., Klakow, D., and de~Melo, G. (2019).
\newblock Using {Multi}-{Sense} {Vector} {Embeddings} for {Reverse} {Dictionaries}.
\newblock In Dobnik, S., Chatzikyriakidis, S., and Demberg, V., editors, {\em Proceedings of the 13th {International} {Conference} on {Computational} {Semantics} - {Long} {Papers}}, pages 247--258, Gothenburg, Sweden. Association for Computational Linguistics.
\newblock https://doi.org/10.18653/v1/W19-0421.

\bibitem[Hickey, 2005]{hickey_legacies_2005}
Hickey, R., editor (2005).
\newblock {\em Legacies of {Colonial} {English}: {Studies} in {Transported} {Dialects}}.
\newblock Studies in {English} {Language}. Cambridge University Press, Cambridge, England.
\newblock https://doi.org/10.1017/CBO9780511486920.

\bibitem[Hogg, 2016]{hogg_social_2016}
Hogg, M.~A. (2016).
\newblock Social {Identity} {Theory}.
\newblock In McKeown, S., Haji, R., and Ferguson, N., editors, {\em Understanding {Peace} and {Conflict} {Through} {Social} {Identity} {Theory}: {Contemporary} {Global} {Perspectives}}, Peace {Psychology} {Book} {Series}, pages 3--17. Springer International Publishing, Cham, Switzerland.
\newblock https://doi.org/10.1007/978-3-319-29869-6\_1.

\bibitem[Holmes and Wilson, 2013]{holmes_introduction_2013}
Holmes, J. and Wilson, N. (2013).
\newblock {\em An {Introduction} to {Sociolinguistics}}.
\newblock Taylor \& Francis, Abingdon, England; New York, NY, 4 edition.

\bibitem[Hovy and Spruit, 2016]{hovy_social_2016}
Hovy, D. and Spruit, S.~L. (2016).
\newblock The {Social} {Impact} of {Natural} {Language} {Processing}.
\newblock In Erk, K. and Smith, N.~A., editors, {\em Proceedings of the 54th {Annual} {Meeting} of the {Association} for {Computational} {Linguistics} ({Volume} 2: {Short} {Papers})}, pages 591--598, Berlin, Germany. Association for Computational Linguistics.
\newblock https://doi.org/10.18653/v1/P16-2096.

\bibitem[Hu, 2018]{hu_geo-text_2018}
Hu, Y. (2018).
\newblock Geo-text data and data-driven geospatial semantics.
\newblock {\em Geography Compass}, 12(11):e12404.
\newblock https://doi.org/10.1111/gec3.12404.

\bibitem[Huang et~al., 2016]{huang_understanding_2016}
Huang, Y., Guo, D., Kasakoff, A., and Grieve, J. (2016).
\newblock Understanding {U}.{S}. regional linguistic variation with {Twitter} data analysis.
\newblock {\em Computers, Environment and Urban Systems}, 59:244--255.
\newblock https://doi.org/10.1016/j.compenvurbsys.2015.12.003.

\bibitem[Hundt, 1998]{hundt_new_1998}
Hundt, M. (1998).
\newblock {\em New {Zealand} {English} {Grammar} - {Fact} or {Fiction}?: {A} corpus-based study in morphosyntactic variation}, volume G23 of {\em Varieties of {English} {Around} the {World}}.
\newblock John Benjamins Publishing Company, Amsterdam, The Netherlands; Philadelphia, PA, 1 edition.

\bibitem[Hunter, 2007]{hunter_matplotlib_2007}
Hunter, J.~D. (2007).
\newblock Matplotlib: {A} {2D} {Graphics} {Environment}.
\newblock {\em Computing in Science \& Engineering}, 9(3):90--95.
\newblock https://doi.org/10.1109/MCSE.2007.55.

\bibitem[Hymes, 1962]{hymes_ethnography_1962}
Hymes, D. (1962).
\newblock The ethnography of speaking.
\newblock In Gladwin, T. and Sturtevant, W.~C., editors, {\em Anthropology and human behavior}, pages 13--53. Anthropological Society of Washington, Washington, D.C.

\bibitem[Hymes, 1967]{hymes_models_1967}
Hymes, D. (1967).
\newblock Models of the {Interaction} of {Language} and {Social} {Setting}.
\newblock {\em Journal of Social Issues}, 23(2):8--28.
\newblock https://doi.org/10.1111/j.1540-4560.1967.tb00572.x.

\bibitem[Hymes, 1968]{hymes_linguistic_1968}
Hymes, D. (1968).
\newblock Linguistic problems in defining the concept of ‘tribe’.
\newblock In Helm, J., editor, {\em Essays on the problem of tribe}, pages 23--48. University of Washington Press for the American Ethnological Society.

\bibitem[Hymes, 1972]{hymes_scope_1972}
Hymes, D. (1972).
\newblock The {Scope} of {Sociolinguistics}.
\newblock In Shuy, R.~W., editor, {\em 23rd {Annual} {Round} {Table} {Sociolinguistics}: {Current} {Trends} and {Prospects}}, number~25 in Monograph {Series} on {Language} and {Linguistics}, pages 313--333. Georgetown University Press, Washington, D.C.

\bibitem[Hymes, 1997]{hymes_scope_1997}
Hymes, D. (1997).
\newblock The {Scope} of {Sociolinguistics}.
\newblock In Coupland, N. and Jaworski, A., editors, {\em Sociolinguistics: {A} {Reader}}, Modern {Linguistics} {Series}, pages 12--22. Macmillan Education UK, New York, NY, 1 edition.
\newblock https://doi.org/10.1007/978-1-349-25582-5\_16.

\bibitem[Hymes, 2020]{hymes_scope_2020}
Hymes, D. (2020).
\newblock The {Scope} of {Sociolinguistics}.
\newblock {\em International Journal of the Sociology of Language}, 2020(263):67--76.
\newblock https://doi.org/10.1515/ijsl-2020-2084.

\bibitem[Ilbury, 2020]{ilbury_sassy_2020}
Ilbury, C. (2020).
\newblock “{Sassy} {Queens}”: {Stylistic} orthographic variation in {Twitter} and the enregisterment of {AAVE}.
\newblock {\em Journal of Sociolinguistics}, 24(2):245--264.
\newblock https://doi.org/10.1111/josl.12366.

\bibitem[Ilbury, 2022]{ilbury_tale_2022}
Ilbury, C. (2022).
\newblock A tale of two cities: {The} discursive construction of ‘place’ in gentrifying {East} {London}.
\newblock {\em Language in Society}, 51(3):511--534.
\newblock https://doi.org/10.1017/S0047404521000130.

\bibitem[Ilbury et~al., 2024]{ilbury_using_2024}
Ilbury, C., Grieve, J., and Hall, D. (2024).
\newblock Using social media to infer the diffusion of an urban contact dialect: {A} case study of {Multicultural} {London} {English}.
\newblock {\em Journal of Sociolinguistics}.
\newblock https://doi.org/10.1111/josl.12653.

\bibitem[{InternetNZ}, 2025]{internetnz_internetnz_2025}
{InternetNZ} (2025).
\newblock {InternetNZ} {\textbar} {Ipurangi} {Aotearoa} {New} {Zealand}'s {Internet} {Insights}: 2024 survey findings.
\newblock Technical report, Internet New Zealand, Auckland, New Zealand.
\newblock Retrieved from https://internetnz.nz/new-zealands-internet-insights/new-zealands-internet-insights-2024/.

\bibitem[Ison et~al., 2025]{ison_i_2025}
Ison, J., Santamaria, E., Caluzzi, G., McAllister, C., Hooker, L., Wilson, I., Theobald, J., Laslett, A.-M., and Riordan, B. (2025).
\newblock “{I} {Don}’t {Know} for {Certain}”: {A} {Content} {Analysis} of {Reddit} {Posters}’ {Accounts} of {Drink} {Spiking}.
\newblock {\em Violence Against Women}, page 10778012251347585.
\newblock https://doi.org/10.1177/10778012251347585.

\bibitem[Jamet, 2009]{jamet_morphophonological_2009}
Jamet, D. (2009).
\newblock A morphophonological approach to clipping in {English}.
\newblock {\em Lexis. Journal in English Lexicology}, 1.
\newblock https://doi.org/10.4000/lexis.884.

\bibitem[Jautz, 2008]{jautz_gratitude_2008}
Jautz, S. (2008).
\newblock Gratitude in {British} and {New} {Zealand} radio programmes: {Nothing} but gushing?
\newblock In Schneider, K.~P. and Barron, A., editors, {\em Variational {Pragmatics}: {A} focus on regional varieties in pluricentric languages}, pages 141--178. John Benjamins Publishing Company, Amsterdam, The Netherlands; Philadelphia, PA.
\newblock https://doi.org/10.1075/pbns.178.

\bibitem[Jeszenszky et~al., 2024]{jeszenszky_effects_2024}
Jeszenszky, P., Steiner, C., and Leemann, A. (2024).
\newblock Effects of mobility on dialect change: {Introducing} the linguistic mobility index.
\newblock {\em PLOS ONE}, 19(4):e0300735.
\newblock https://doi.org/10.1371/journal.pone.0300735.

\bibitem[Joachims, 1998]{joachims_text_1998}
Joachims, T. (1998).
\newblock Text categorization with {Support} {Vector} {Machines}: learning with many relevant features.
\newblock In {\em Proceedings of the 10th {European} {Conference} on {Machine} {Learning}}, {ECML}'98, pages 137--142, Chemnitz, Germany. Springer-Verlag, Berlin, Heidelberg.
\newblock https://doi.org/10.1007/BFb0026683.

\bibitem[Johnson et~al., 2016]{johnson_geography_2016}
Johnson, I.~L., Sengupta, S., Schöning, J., and Hecht, B. (2016).
\newblock The {Geography} and {Importance} of {Localness} in {Geotagged} {Social} {Media}.
\newblock In {\em Proceedings of the 2016 {CHI} {Conference} on {Human} {Factors} in {Computing} {Systems}}, {CHI} '16, pages 515--526. Association for Computing Machinery.
\newblock https://doi.org/10.1145/2858036.2858122.

\bibitem[Johnstone, 2004]{johnstone_place_2004}
Johnstone, B. (2004).
\newblock Place, {Globalization}, and {Linguistic} {Variation}, {Barbara} {Johnstone}.
\newblock In Fought, C., editor, {\em Sociolinguistic {Variation}: {Critical} {Reflections}}, pages 65--83. Oxford University Press, New York, NY.
\newblock https://doi.org/10.1093/oso/9780195170399.003.0005.

\bibitem[Johnstone, 2009]{johnstone_pittsburghese_2009}
Johnstone, B. (2009).
\newblock Pittsburghese shirts: {Commodification} and the enregisterment of an urban dialect.
\newblock {\em American Speech}, 84(2):157--175.
\newblock https://doi.org/10.1215/00031283-2009-013.

\bibitem[Jones, 1994]{jones_natural_1994}
Jones, K.~S. (1994).
\newblock Natural {Language} {Processing}: {A} {Historical} {Review}.
\newblock In Zampolli, A., Calzolari, N., and Palmer, M., editors, {\em Current {Issues} in {Computational} {Linguistics}: {In} {Honour} of {Don} {Walker}}, pages 3--16. Springer Netherlands, Dordrecht, The Netherlands.
\newblock https://doi.org/10.1007/978-0-585-35958-8\_1.

\bibitem[Jones, 2015]{jones_toward_2015}
Jones, T. (2015).
\newblock Toward a {Description} of {African} {American} {Vernacular} {English} {Dialect} {Regions} {Using} “{Black} {Twitter}”.
\newblock {\em American Speech}, 90(4):403--440.
\newblock https://doi.org/10.1215/00031283-3442117.

\bibitem[Joshi et~al., 2025]{joshi_natural_2025}
Joshi, A., Dabre, R., Kanojia, D., Li, Z., Zhan, H., Haffari, G., and Dippold, D. (2025).
\newblock Natural {Language} {Processing} for {Dialects} of a {Language}: {A} {Survey}.
\newblock {\em ACM Comput. Surv.}, 57(6):149:1--149:37.
\newblock https://dl.acm.org/doi/10.1145/3712060.

\bibitem[Jurafsky and Martin, 2026]{jurafsky_speech_2026}
Jurafsky, D. and Martin, J.~H. (2026).
\newblock Speech and {Language} {Processing}: {An} {Introduction} to {Natural} {Language} {Processing}, {Computational} {Linguistics}, and {Speech} {Recognition}, with {Language} {Models}.
\newblock Retrieved from https://web.stanford.edu/{\textasciitilde}jurafsky/slp3/.

\bibitem[Jurgens et~al., 2017]{jurgens_incorporating_2017}
Jurgens, D., Tsvetkov, Y., and Jurafsky, D. (2017).
\newblock Incorporating {Dialectal} {Variability} for {Socially} {Equitable} {Language} {Identification}.
\newblock In Barzilay, R. and Kan, M.-Y., editors, {\em Proceedings of the 55th {Annual} {Meeting} of the {Association} for {Computational} {Linguistics} ({Volume} 2: {Short} {Papers})}, pages 51--57, Vancouver, Canada. Association for Computational Linguistics.
\newblock https://doi.org/10.18653/v1/P17-2009.

\bibitem[Jünger, 2021]{junger_brief_2021}
Jünger, J. (2021).
\newblock A brief history of {APIs}: {Limitations} and opportunities for online research.
\newblock In Engel, U., Quan-Haase, A., Liu, S.~X., and Lyberg, L., editors, {\em Handbook of {Computational} {Social} {Science}: {Data} {Science}, {Statistical} {Modelling}, and {Machine} {Learning} {Methods}}, volume~2, pages 17--32. Taylor \& Francis, Abingdon, England; New York, NY.
\newblock https://doi.org/10.4324/9781003025245-3.

\bibitem[Kachru, 1985]{kachru_standards_1985}
Kachru, B.~B. (1985).
\newblock Standards, codification and sociolinguistic realism: {The} {English} language in the outer circle.
\newblock In Quirk, R. and Widdowson, H.~G., editors, {\em English in the world: {Teaching} and learning the language and literatures}, page 11‐30. Cambridge University Press, Cambridge, England.

\bibitem[Kgosietsile, 2025]{kgosietsile_cosine_2025}
Kgosietsile, T. (2025).
\newblock Cosine {Similarity} {Preserving} {Curse} of {Dimensionality} {Reduction} for {Managing} {Computational} {Complexity}.
\newblock In Noor, A., Saroha, K., Pricop, E., Sen, A., and Trivedi, G., editors, {\em Emerging {Trends} and {Technologies} on {Intelligent} {Systems}}, pages 253--265, Singapore. Springer Nature.
\newblock https://doi.org/10.1007/978-981-97-5703-9\_21.

\bibitem[Kiesling, 1998]{kiesling_mens_1998}
Kiesling, S.~F. (1998).
\newblock Men’s {Identities} and {Sociolinguistic} {Variation}: {The} {Case} of {Fraternity} {Men}.
\newblock {\em Journal of Sociolinguistics}, 2(1):69--99.
\newblock https://doi.org/10.1111/1467-9481.00031.

\bibitem[Kiesling, 2020]{kiesling_english_2020}
Kiesling, S.~F. (2020).
\newblock English in {Australia} and {New} {Zealand}.
\newblock In Nelson, C.~L., Proshina, Z.~G., and Davis, D.~R., editors, {\em The {Handbook} of {World} {Englishes}}, pages 70--86. Wiley-Blackwell, Hoboken, NJ, 2 edition.
\newblock https://doi.org/10.1002/9780470757598.ch5.

\bibitem[Kim and Zhang, 2025]{kim_i_2025}
Kim, J. and Zhang, L. (2025).
\newblock “{I} {Want} to {Be} {Born} with {That} {Pronunciation}”: {Metalinguistic} {Comments} {About} {K}-{Pop} {Idols}’ {Inner} {Circle} {Accents}.
\newblock {\em Languages}, 10(4):75.
\newblock https://doi.org/10.3390/languages10040075.

\bibitem[Kim et~al., 2014]{kim_temporal_2014}
Kim, Y., Chiu, Y.-I., Hanaki, K., Hegde, D., and Petrov, S. (2014).
\newblock Temporal {Analysis} of {Language} through {Neural} {Language} {Models}.
\newblock In Danescu-Niculescu-Mizil, C., Eisenstein, J., McKeown, K., and Smith, N.~A., editors, {\em Proceedings of the {ACL} 2014 {Workshop} on {Language} {Technologies} and {Computational} {Social} {Science}}, pages 61--65, Baltimore, MD, USA. Association for Computational Linguistics.

\bibitem[Klein, 2000]{klein_no_2000}
Klein, N. (2000).
\newblock {\em No {Logo}: {Taking} {Aim} at the {Brand} {Bullies}}.
\newblock Flamingo, London, England.

\bibitem[Koplenig, 2017]{koplenig_impact_2017}
Koplenig, A. (2017).
\newblock The impact of lacking metadata for the measurement of cultural and linguistic change using the {Google} {Ngram} data sets—{Reconstructing} the composition of the {German} corpus in times of {WWII}.
\newblock {\em Digital Scholarship in the Humanities}, 32(1):169--188.

\bibitem[Kowsari et~al., 2019]{kowsari_text_2019}
Kowsari, K., Jafari~Meimandi, K., Heidarysafa, M., Mendu, S., Barnes, L., and Brown, D. (2019).
\newblock Text {Classification} {Algorithms}: {A} {Survey}.
\newblock {\em Information}, 10(4):150.
\newblock https://doi.org/10.3390/info10040150.

\bibitem[Krovetz, 1997]{krovetz_homonymy_1997}
Krovetz, R. (1997).
\newblock Homonymy and {Polysemy} in {Information} {Retrieval}.
\newblock In {\em 35th {Annual} {Meeting} of the {Association} for {Computational} {Linguistics} and 8th {Conference} of the {European} {Chapter} of the {Association} for {Computational} {Linguistics}}, pages 72--79, Madrid, Spain. Association for Computational Linguistics.
\newblock https://doi.org/10.3115/976909.979627.

\bibitem[Kulshrestha et~al., 2012]{kulshrestha_geographic_2012}
Kulshrestha, J., Kooti, F., Nikravesh, A., and Gummadi, K. (2012).
\newblock Geographic {Dissection} of the {Twitter} {Network}.
\newblock In {\em Proceedings of the {International} {AAAI} {Conference} on {Web} and {Social} {Media}}, volume~6, pages 202--209. PKP Publishing Services Network, Dublin, Ireland.
\newblock https://doi.org/10.1609/icwsm.v6i1.14280.

\bibitem[Kutuzov et~al., 2018]{kutuzov_diachronic_2018}
Kutuzov, A., Øvrelid, L., Szymanski, T., and Velldal, E. (2018).
\newblock Diachronic {Word} {Embeddings} and {Semantic} {Shifts}: {A} {Survey}.
\newblock In Bender, E.~M., Derczynski, L., and Isabelle, P., editors, {\em Proceedings of the 27th {International} {Conference} on {Computational} {Linguistics}}, pages 1384--1397, Santa Fe, NM. Association for Computational Linguistics.
\newblock Retrieved from https://aclanthology.org/C18-1117/.

\bibitem[Kytö and Romaine, 2006]{smitterberg_adjective_2006}
Kytö, M. and Romaine, S. (2006).
\newblock Adjective comparison in nineteenth-century {English}.
\newblock In Smitterberg, E., Rydén, M., and Kytö, M., editors, {\em Nineteenth-{Century} {English}: {Stability} and {Change}}, Studies in {English} {Language}, pages 194--214. Cambridge University Press, Cambridge, England.
\newblock https://doi.org/10.1017/CBO9780511486944.008.

\bibitem[Labov, 1972a]{labov_sociolinguistic_1972}
Labov, W. (1972a).
\newblock {\em Sociolinguistic {Patterns}}.
\newblock University of Pennsylvania Press, Philadelphia, PA.

\bibitem[Labov, 1972b]{labov_principles_1972}
Labov, W. (1972b).
\newblock Some {Principles} of {Linguistic} {Methodology}.
\newblock {\em Language in Society}, 1(1):97--120.
\newblock Retrieved from https://www.jstor.org/stable/4166672.

\bibitem[Labov, 2006]{labov_social_2006}
Labov, W. (2006).
\newblock {\em The {Social} {Stratification} of {English} in {New} {York} {City}}.
\newblock Cambridge University Press, Cambridge, England; New York, NY, 2 edition.
\newblock https://doi.org/10.1017/CBO9780511618208 (Originally published 1966).

\bibitem[Labov et~al., 2008]{labov_dialects_2008}
Labov, W., Ash, S., and Boberg, C. (2008).
\newblock The dialects of {North} {American} {English}.
\newblock In {\em The {Atlas} of {North} {American} {English}}, pages 119--151. De Gruyter Mouton, Berlin, Germany; New York, NY.
\newblock https://doi.org/10.1515/9783110167467.d.119.

\bibitem[Labov et~al., 1968]{labov_study_1968}
Labov, W., Cohen, P., Robins, C., and Lewis, J. (1968).
\newblock A study of non-standard {English} of {Negro} and {Puerto} {Rican} speakers in {New} {York} {City}. {Volume} {II}: {The} {Use} of {Language} in the {Speech} {Community}.
\newblock Technical Report CRP-3288, Spons Agency - Office of Education (DHEW), Washington, D.C. Bureau of Research, Washington, D.C.
\newblock Retrieved from https://searchworks.stanford.edu/view/2344843.

\bibitem[Labov and Waletzky, 1967]{labov_narrative_1967}
Labov, W. and Waletzky, J. (1967).
\newblock Narrative analysis: oral versions of personal experience.
\newblock In Helm, J., editor, {\em Proceedings of the 1966 {Annual} {Spring} {Meeting} of the {American} {Ethnological} {Society}}.
\newblock https://psycnet.apa.org/doi/10.1075/jnlh.7.02nar.

\bibitem[Lackaff and Moner, 2016]{lackaff_local_2016}
Lackaff, D. and Moner, W.~J. (2016).
\newblock Local languages, global networks: {Mobile} design for minority language users.
\newblock In {\em Proceedings of the 34th {ACM} {International} {Conference} on the {Design} of {Communication}}, pages 1--9, New York, NY. Association for Computing Machinery.
\newblock https://doi.org/10.1145/2987592.2987612.

\bibitem[Lafkioui, 2008]{lafkioui_dialectometry_2008}
Lafkioui, M. (2008).
\newblock Dialectometry {Analyses} of {Berber} {Lexis}.
\newblock {\em Folia Orientalia}, 44:71--88.
\newblock Retrieved from https://journals.pan.pl/dlibra/show-content?id=125653.

\bibitem[Lagorio-Chafkin, 2018]{lagorio-chafkin_we_2018}
Lagorio-Chafkin, C. (2018).
\newblock {\em We {Are} the {Nerds}: {The} {Birth} and {Tumultuous} {Life} of {Reddit}, the {Internet}'s {Culture} {Laboratory}}.
\newblock Hachette Books, New York, NY, illustrated edition edition.

\bibitem[Lalić, 2004]{lalic_eponyms_2004}
Lalić, G. (2004).
\newblock Eponyms in english.
\newblock {\em Romanian Journal of English Studies}, 1:64--69.

\bibitem[{lavacano201014}, 2011]{lavacano201014_stay_2011}
{lavacano201014} (2011).
\newblock Stay {Classy}, {X}.
\newblock Retrieved May 23, 2025, from https://knowyourmeme.com/memes/stay-classy-x.

\bibitem[Lendvai, 2025]{lendvai_reddit_2025}
Lendvai, G.~F. (2025).
\newblock Reddit in scholarly reception: a bibliometric assessment of the front page of the internet.
\newblock {\em Quality \& Quantity}.
\newblock https://doi.org/10.1007/s11135-025-02416-z.

\bibitem[Leuckert and Leuckert, 2020]{leuckert_towards_2020}
Leuckert, S. and Leuckert, M. (2020).
\newblock Towards a digital sociolinguistics: {Communities} of {Practice} on {Reddit}.
\newblock In Rüdiger, S. and Dayter, D., editors, {\em Corpus {Approaches} to {Social} {Media}}, pages 15--40. John Benjamins Publishing Company, Amsterdam, The Netherlands; Philadelphia, PA.

\bibitem[Lieber, 2005]{lieber_english_2005}
Lieber, R. (2005).
\newblock English {Word}-{Formation} {Processes}.
\newblock In Štekauer, P. and Lieber, R., editors, {\em Handbook of {Word}-{Formation}}, volume~64 of {\em Studies in {Natural} {Language} and {Linguistic} {Theory}}, pages 375--427. Springer, Dordrecht, The Netherlands.
\newblock https://doi.org/10.1007/1-4020-3596-9\_16.

\bibitem[{Liz Quilty}, 2007]{liz_quilty_sweetbix_2007}
{Liz Quilty} (2007).
\newblock Sweetbix.
\newblock Retrieved from http://sweetbix.urbanup.com/2340215.

\bibitem[Lyons, 1970]{lyons_new_1970}
Lyons, J. (1970).
\newblock {\em New {Horizons} in {Linguistics}}.
\newblock Penguin Books, Harmondsworth, England; New York, NY; Ringwood, Victoria, Australia; Markham, Ontario, Canada; Auckland, New Zealand.

\bibitem[Macalister, 1999]{macalister_trends_1999}
Macalister, J. (1999).
\newblock Trends in {New} {Zealand} {English}: some observations on the presence of {Maori} words in the lexicon.
\newblock {\em New Zealand English Journal}, 13:38--49.
\newblock Retrieved from https://search.informit.org/doi/abs/10.3316/INFORMIT.778553791609998.

\bibitem[Macalister, 2006]{macalister_weka_2006}
Macalister, J. (2006).
\newblock Of weka and waiata: {Familiarity} with borrowings from te reo {Maori}.
\newblock {\em Te Reo}, 49:101--124.
\newblock Retrieved from https://nzlingsoc.org/journal\_article/of-weka-and-waiata-familiarity-with-borrowings-from-te-reo-maori/.

\bibitem[Maclagan, 1998]{maclagan_h-dropping_1998}
Maclagan, D. (1998).
\newblock /h/-dropping in {Early} {New} {Zealand} {English}.
\newblock {\em New Zealand English Journal}, 12:34--42.
\newblock Retrieved from https://search.informit.org/doi/10.3316/informit.778423360811191.

\bibitem[Maclagan et~al., 2008]{maclagan_maori_2008}
Maclagan, M., King, J., and Gillon, G. (2008).
\newblock Maori {English}.
\newblock {\em Clinical Linguistics \& Phonetics}, 22(8):658--670.
\newblock https://doi.org/10.1080/02699200802222271.

\bibitem[Maclagan and Warren, 2020]{maclagan_englishes_2020}
Maclagan, M. and Warren, P. (2020).
\newblock The {English}(es) of {New} {Zealand}.
\newblock In Kirkpatrick, A., editor, {\em The {Routledge} {Handbook} of {World} {Englishes}}. Routledge, Abingdon, England; New York, NY, 2 edition.
\newblock https://doi.org/10.4324/9781003128755.

\bibitem[Mahler, 2020]{mahler_lexical_2020}
Mahler, H. (2020).
\newblock Lexical {Emergence} on {Reddit}: {An} {Analysis} of {Lexical} {Change} on the “{Front} {Page} of the {Internet}”.
\newblock {\em Lexis. Journal in English Lexicology}, 16:1--21.
\newblock https://doi.org/10.4000/lexis.4917.

\bibitem[Manvi et~al., 2024]{manvi_large_2024}
Manvi, R., Khanna, S., Burke, M., Lobell, D., and Ermon, S. (2024).
\newblock Large language models are geographically biased.
\newblock In {\em Proceedings of the 41st {International} {Conference} on {Machine} {Learning}}, volume 235, pages 34654--34669, Vienna, Austria. Journal of Machine Learning Research.
\newblock https://doi.org/10.5555/3692070.3693479.

\bibitem[Markov, 2006]{markov_example_2006}
Markov, A.~A. (2006).
\newblock An {Example} of {Statistical} {Investigation} of the {Text} {Eugene} {Onegin} {Concerning} the {Connection} of {Samples} in {Chains}.
\newblock {\em Science in Context}, 19(4):591--600.
\newblock https://doi.org/10.1017/S0269889706001074.

\bibitem[Martí et~al., 2019]{marti_social_2019}
Martí, P., Serrano-Estrada, L., and Nolasco-Cirugeda, A. (2019).
\newblock Social {Media} data: {Challenges}, opportunities and limitations in urban studies.
\newblock {\em Computers, Environment and Urban Systems}, 74:161--174.
\newblock https://doi.org/10.1016/j.compenvurbsys.2018.11.001.

\bibitem[Matheson, 2023]{matheson_discourse_2023}
Matheson, D. (2023).
\newblock Discourse analysis after the computational turn: a mixed bag.
\newblock {\em Communication Research and Practice}, 9(1):3--15.
\newblock https://doi.org/10.1080/22041451.2023.2190531.

\bibitem[Mattiello, 2016]{mattiello_analogical_2016}
Mattiello, E. (2016).
\newblock Analogical neologisms in {English}.
\newblock {\em Italian Journal of Linguistics}, 28(2):103--142.
\newblock Retrieved from https://hdl.handle.net/11568/772593.

\bibitem[McKinney, 2010]{mckinney_data_2010}
McKinney, W. (2010).
\newblock Data {Structures} for {Statistical} {Computing} in {Python}.
\newblock In van~der Walt, S. and Millman, J., editors, {\em Proceedings of the 9th {Python} in {Science} {Conference}}, pages 56--61, Austin, TX. SciPy Proceedings.
\newblock https://doi.org/10.25080/Majora-92bf1922-00a.

\bibitem[Merrit, 2012]{merrit_analysis_2012}
Merrit, E. (2012).
\newblock An {Analysis} of the {Discourse} of {Internet} {Trolling}: {A} {Case} {Study} of {Reddit}.com.
\newblock Retrieved from http://hdl.handle.net/10166/1058.

\bibitem[Meyerhoff, 1993]{meyerhoff_lexical_1993}
Meyerhoff, M. (1993).
\newblock Lexical {Shift} in {Working} {Class} {New} {Zealnd} {English}: {Variation} in the {Use} of {Lexical} {Pairs}.
\newblock {\em English World-Wide}, 14(2):231--248.
\newblock https://doi.org/10.1075/eww.14.2.04mey.

\bibitem[Meyerhoff and Niedzielski, 2003]{meyerhoff_globalisation_2003}
Meyerhoff, M. and Niedzielski, N. (2003).
\newblock The globalisation of vernacular variation.
\newblock {\em Journal of Sociolinguistics}, 7(4):534--555.
\newblock https://doi.org/10.1111/j.1467-9841.2003.00241.x.

\bibitem[Mikolov et~al., 2013]{mikolov_efficient_2013}
Mikolov, T., Chen, K., Corrado, G., and Dean, J. (2013).
\newblock Efficient {Estimation} of {Word} {Representations} in {Vector} {Space}.
\newblock In Bengio, Y. and LeCun, Y., editors, {\em Proceedings of {Workshop} at {ICLR}}, Scottsdale, AZ.
\newblock https://doi.org/10.48550/arXiv.1301.3781.

\bibitem[Miller, 2009]{miller_lingo_2009}
Miller, J. (2009).
\newblock {\em The {Lingo} {Dictionary}: {Of} {Favourite} {Australian} {Words} and {Phrases}}.
\newblock Exisle Publishing, Wollombi, NSW, Australia.

\bibitem[Milroy, 1987]{milroy_language_1987}
Milroy, L. (1987).
\newblock {\em Language and {Social} {Networks}}.
\newblock Basil Blackwell, Oxford, England; New York, NY, 2 edition.

\bibitem[Mir and Rathinam, 2023]{mir_rise_2023}
Mir, A.~A. and Rathinam, S. (2023).
\newblock The rise of scientific literature on {Twitter} research: a bibliometric analysis and some insights.
\newblock {\em Global Knowledge, Memory and Communication}, 74(3-4):1045--1068.
\newblock https://doi.org/10.1108/GKMC-03-2023-0083.

\bibitem[Mislove et~al., 2011]{mislove_understanding_2011}
Mislove, A., Lehmann, S., Ahn, Y.-Y., Onnela, J.-P., and Rosenquist, J. (2011).
\newblock Understanding the {Demographics} of {Twitter} {Users}.
\newblock {\em Proceedings of the International AAAI Conference on Web and Social Media}, 5(1):554--557.
\newblock https://doi.org/10.1609/icwsm.v5i1.14168.

\bibitem[Momoisea, 2015]{momoisea_lani_2015}
Momoisea, L. (2015).
\newblock Lani {Writes}: {The} origins of {Poly}-slang.
\newblock {\em Metro}.
\newblock Retrieved from https://www.metromag.co.nz/society/society-etc/lani-writes-the-origins-of-poly-slang.

\bibitem[Monka et~al., 2020]{monka_place_2020}
Monka, M., Quist, P., and Skovse, A.~R. (2020).
\newblock Place attachment and linguistic variation: {A} quantitative analysis of language and local attachment in a rural village and an urban social housing area.
\newblock {\em Language in Society}, 49(2):173--205.
\newblock https://doi.org/10.1017/S0047404519000733.

\bibitem[Moorfield, 2025a]{moorfield_he_2025}
Moorfield, J.~C. (2025a).
\newblock he waka eke noa.
\newblock In {\em Te {Aka} {Māori} {Dictionary}}. Pearson Education New Zealand.
\newblock Retrieved from https://maoridictionary.co.nz/word/48434.

\bibitem[Moorfield, 2025b]{moorfield_kia_2025}
Moorfield, J.~C. (2025b).
\newblock Kia ora!
\newblock In {\em Te {Aka} {Māori} {Dictionary}}. Pearson Education New Zealand.
\newblock Retrieved from https://maoridictionary.co.nz/word/2608.

\bibitem[Moorfield, 2025c]{moorfield_nga_2025}
Moorfield, J.~C. (2025c).
\newblock ngā mihi.
\newblock In {\em Te {Aka} {Māori} {Dictionary}}. Pearson Education New Zealand.
\newblock Retrieved from https://maoridictionary.co.nz/word/10748.

\bibitem[Moorfield, 2025d]{moorfield_tohuto_2025}
Moorfield, J.~C. (2025d).
\newblock tohutō.
\newblock In {\em Te {Aka} {Māori} {Dictionary}}. Pearson Education New Zealand.
\newblock Retrieved from https://maoridictionary.co.nz/word/8328.

\bibitem[Morgan, 2014]{morgan_speech_2014}
Morgan, M.~H. (2014).
\newblock {\em Speech {Communities}}.
\newblock Key {Topics} in {Linguistic} {Anthropology}. Cambridge University Press, Cambridge, England; New York, NY.
\newblock https://doi.org/10.1017/CBO9781139151269.

\bibitem[Morin et~al., 2020]{morin_dialect_2020}
Morin, C., Desagulier, G., and Grieve, J. (2020).
\newblock Dialect syntax in {Construction} {Grammar}: {Theoretical} benefits of a constructionist approach to double modals in {English}.
\newblock {\em Belgian Journal of Linguistics}, 34(1):248--258.

\bibitem[Nerbonne and Heeringa, 2007]{nerbonne_geographic_2007}
Nerbonne, J. and Heeringa, W. (2007).
\newblock Geographic {Distributions} of {Linguistic} {Variation} {Reflect} {Dynamics} of {Differentiation}.
\newblock In Featherston, S. and Sternefeld, W., editors, {\em Roots: {Linguistics} in {Search} of its {Evidential} {Base}}, number~96 in Studies in {Generative} {Grammar}, pages 267--298. De Gruyter Mouton, Berlin, Germany.
\newblock https://doi.org/10.1515/9783110198621.267.

\bibitem[Nerbonne et~al., 1996]{nerbonne_phonetic_1996}
Nerbonne, J., Heeringa, W., Van~den Hout, E., Van~der Kooi, P., Otten, S., and Van~de Vis, W. (1996).
\newblock Phonetic distance between {Dutch} dialects.
\newblock In Durieux, G., Daelemans, W., and Gillis, S., editors, {\em Proceedings of the {Sixth} {Computational} {Linguistics} in the {Netherlands} {Meeting}}, pages 185--202, Antwerp, Belgium. University of Antwerp. Center for Dutch Language and Speech.
\newblock Retrieved from https://www.clinjournal.org/CLIN\_proceedings/VI/nerbonne.pdf.

\bibitem[Nerbonne and Kleiweg, 2007]{nerbonne_toward_2007}
Nerbonne, J. and Kleiweg, P. (2007).
\newblock Toward a dialectological yardstick*.
\newblock {\em Journal of Quantitative Linguistics}, 14(2-3):148--166.
\newblock https://doi.org/10.1080/09296170701379260.

\bibitem[Nguyen, 2021]{nguyen_dialect_2021}
Nguyen, D. (2021).
\newblock Dialect {Variation} on {Social} {Media}.
\newblock In Zampieri, M. and Nakov, P., editors, {\em Similar {Languages}, {Varieties}, and {Dialects}: {A} {Computational} {Perspective}}, Studies in {Natural} {Language} {Processing}, pages 204--218. Cambridge University Press, Cambridge, United Kingdom.
\newblock https://doi.org/10.1017/9781108565080.014.

\bibitem[Nguyen et~al., 2016]{nguyen_computational_2016}
Nguyen, D., Doğruöz, A.~S., Rosé, C.~P., and de~Jong, F. (2016).
\newblock Computational {Sociolinguistics}: {A} {Survey}.
\newblock {\em Computational Linguistics}, 42(3):537--593.
\newblock https://doi.org/10.1162/COLI\_a\_00258.

\bibitem[{Ngā Pou Taunaha o Aotearoa {\textbar} New Zealand Geographic Board}, 2025]{nga_pou_taunaha_o_aotearoa__new_zealand_geographic_board_new_2025}
{Ngā Pou Taunaha o Aotearoa {\textbar} New Zealand Geographic Board} (2025).
\newblock New {Zealand} {Gazetteer}.
\newblock Retrieved from https://www.linz.govt.nz/products-services/place-names/place-names-new-zealand.

\bibitem[Niedzielski and Preston, 2000]{niedzielski_folk_2000}
Niedzielski, N.~A. and Preston, D.~R. (2000).
\newblock {\em Folk {Linguistics}}.
\newblock Number 122 in Trends in {Linguistics}. {Studies} and {Monographs}. Walter de Gruyter, Berlin, Germany; New York, NY.

\bibitem[{NovaXP}, 2013]{novaxp_doge_2013}
{NovaXP} (2013).
\newblock Doge.
\newblock Retrieved May 23, 2025, from https://knowyourmeme.com/memes/doge.

\bibitem[Orsman and Orsman, 1994]{orsman_new_1994}
Orsman, E. and Orsman, H. (1994).
\newblock {\em The {New} {Zealand} {Dictionary}}.
\newblock New House, Auckland, New Zealand, standard edition.

\bibitem[{Otago Daily Times}, 2011]{otago_daily_times_nek_2011}
{Otago Daily Times} (2011).
\newblock Nek minute, {Dunedin} video in top 2011 {Google} searches.
\newblock {\em Otago Daily Times}.
\newblock Retrieved from https://www.odt.co.nz/news/dunedin/nek-minute-dunedin-video-top-2011-google-searches.

\bibitem[{Oxford University Press}, 2023a]{oxford_university_press_blockholder_2023}
{Oxford University Press} (2023a).
\newblock blockholder.
\newblock In {\em Oxford {English} {Dictionary}}. Oxford University Press.
\newblock https://doi.org/10.1093/OED/8346884735.

\bibitem[{Oxford University Press}, 2023b]{oxford_university_press_bottle-oh_2023}
{Oxford University Press} (2023b).
\newblock bottle-oh.
\newblock In {\em Oxford {English} {Dictionary}}. Oxford University Press.
\newblock https://doi.org/10.1093/OED/1172259375.

\bibitem[{Oxford University Press}, 2023c]{oxford_university_press_chocker_2023}
{Oxford University Press} (2023c).
\newblock chocker.
\newblock In {\em Oxford {English} {Dictionary}}. Oxford University Press.
\newblock https://doi.org/10.1093/OED/7811084612.

\bibitem[{Oxford University Press}, 2023d]{oxford_university_press_chur_2023}
{Oxford University Press} (2023d).
\newblock chur.
\newblock In {\em Oxford {English} {Dictionary}}. Oxford University Press.
\newblock https://doi.org/10.1093/OED/7527832248.

\bibitem[{Oxford University Press}, 2023e]{oxford_university_press_cringe_2023}
{Oxford University Press} (2023e).
\newblock cringe.
\newblock In {\em Oxford {English} {Dictionary}}. Oxford University Press.
\newblock https://doi.org/10.1093/OED/5403283771.

\bibitem[{Oxford University Press}, 2023f]{oxford_university_press_finna_2023}
{Oxford University Press} (2023f).
\newblock finna.
\newblock In {\em Oxford {English} {Dictionary}}. Oxford University Press.
\newblock https://doi.org/10.1093/OED/4275745882.

\bibitem[{Oxford University Press}, 2023g]{oxford_university_press_servo_2023}
{Oxford University Press} (2023g).
\newblock servo.
\newblock In {\em Oxford {English} {Dictionary}}. Oxford University Press.
\newblock https://doi.org/10.1093/OED/1763988289.

\bibitem[{Oxford University Press}, 2023h]{oxford_university_press_troll_2023}
{Oxford University Press} (2023h).
\newblock troll.
\newblock In {\em Oxford {English} {Dictionary}}. Oxford University Press.
\newblock https://doi.org/10.1093/OED/2603475121.

\bibitem[{Oxford University Press}, 2024a]{oxford_university_press_bruh_2024}
{Oxford University Press} (2024a).
\newblock bruh.
\newblock In {\em Oxford {English} {Dictionary}}. Oxford University Press.
\newblock https://doi.org/10.1093/OED/7213310981.

\bibitem[{Oxford University Press}, 2024b]{oxford_university_press_cheeky_2024}
{Oxford University Press} (2024b).
\newblock cheeky.
\newblock In {\em Oxford {English} {Dictionary}}. Oxford University Press.
\newblock https://doi.org/10.1093/OED/2161579907.

\bibitem[{Oxford University Press}, 2024c]{oxford_university_press_compo_2024}
{Oxford University Press} (2024c).
\newblock compo.
\newblock In {\em Oxford {English} {Dictionary}}. Oxford University Press.
\newblock https://doi.org/10.1093/OED/7811084612.

\bibitem[{Oxford University Press}, 2024d]{oxford_university_press_footy_2024}
{Oxford University Press} (2024d).
\newblock footy.
\newblock In {\em Oxford {English} {Dictionary}}. Oxford University Press.
\newblock https://doi.org/10.1093/OED/5966077543.

\bibitem[{Oxford University Press}, 2024e]{oxford_university_press_maths_2024}
{Oxford University Press} (2024e).
\newblock maths.
\newblock In {\em Oxford {English} {Dictionary}}. Oxford University Press.
\newblock https://doi.org/10.1093/OED/1848133693.

\bibitem[{Oxford University Press}, 2024f]{oxford_university_press_vibes_2024}
{Oxford University Press} (2024f).
\newblock vibes.
\newblock In {\em Oxford {English} {Dictionary}}. Oxford University Press.
\newblock https://doi.org/10.1093/OED/5038453448.

\bibitem[{Oxford University Press}, 2025a]{oxford_university_press_dishcloth_2025}
{Oxford University Press} (2025a).
\newblock dishcloth.
\newblock In {\em Oxford {English} {Dictionary}}. Oxford University Press.
\newblock https://doi.org/10.1093/OED/6305997783.

\bibitem[{Oxford University Press}, 2025b]{oxford_university_press_math_2025}
{Oxford University Press} (2025b).
\newblock math.
\newblock In {\em Oxford {English} {Dictionary}}. Oxford University Press.
\newblock https://doi.org/10.1093/OED/8481846890.

\bibitem[{Oxford University Press}, 2025c]{oxford_university_press_spag_2025}
{Oxford University Press} (2025c).
\newblock spag bol.
\newblock In {\em Oxford {English} {Dictionary}}. Oxford University Press.
\newblock https://doi.org/10.1093/OED/1533522246.

\bibitem[{Oxford University Press}, 2025d]{oxford_university_press_tramp_2025}
{Oxford University Press} (2025d).
\newblock tramp.
\newblock In {\em Oxford {English} {Dictionary}}. Oxford University Press.
\newblock https://doi.org/10.1093/OED/1778125019.

\bibitem[Pabst, 2022]{pabst_putting_2022}
Pabst, K. (2022).
\newblock Putting “the {Other} {Maine}” on the {Map}: {Language} {Variation}, {Local} {Affiliation}, and {Co}-occurrence in {Aroostook} {County} {English}.
\newblock http://hdl.handle.net/1807/125473.

\bibitem[Panek, 2022]{panek_understanding_2022}
Panek, E.~T. (2022).
\newblock {\em Understanding {Reddit}}.
\newblock Routledge, Abingdon, England; New York, NY.
\newblock https://doi.org/10.4324/9781003150800.

\bibitem[Parsons, 1949]{parsons_structure_1949}
Parsons, T. (1949).
\newblock {\em The {Structure} of {Social} {Action}: {A} {Study} in {Social} {Theory} with {Special} {Reference} to a {Group} of {Recent} {European} {Writers}}.
\newblock The Free Press, Glencoe, IL.

\bibitem[Patrick, 2004]{patrick_speech_2004}
Patrick, P.~L. (2004).
\newblock The {Speech} {Community}.
\newblock In Chambers, J.~K. and Schilling-Estes, N., editors, {\em The {Handbook} of {Language} {Variation} and {Change}}, pages 573--597. John Wiley \& Sons, Hoboken, NJ, 2 edition.
\newblock https://doi.org/10.1002/9780470756591.ch23.

\bibitem[Pechenick et~al., 2015]{pechenick_characterizing_2015}
Pechenick, E.~A., Danforth, C.~M., and Dodds, P.~S. (2015).
\newblock Characterizing the {Google} {Books} {Corpus}: {Strong} {Limits} to {Inferences} of {Socio}-{Cultural} and {Linguistic} {Evolution}.
\newblock {\em PLOS ONE}, 10(10):e0137041.

\bibitem[Pedregosa et~al., 2011]{pedregosa_scikit-learn_2011}
Pedregosa, F., Varoquaux, G., Gramfort, A., Michel, V., Thirion, B., Grisel, O., Blondel, M., Prettenhofer, P., Weiss, R., Dubourg, V., Vanderplas, J., Passos, A., Cournapeau, D., Brucher, M., Perrot, M., and Duchesnay, E. (2011).
\newblock Scikit-learn: {Machine} {Learning} in {Python}.
\newblock {\em The Journal of Machine Learning Research}, 12:2825--2830.
\newblock https://doi.org/10.5555/1953048.2078195.

\bibitem[Pennington et~al., 2014]{pennington_glove_2014}
Pennington, J., Socher, R., and Manning, C. (2014).
\newblock {GloVe}: {Global} {Vectors} for {Word} {Representation}.
\newblock In Moschitti, A., Pang, B., and Daelemans, W., editors, {\em Proceedings of the 2014 {Conference} on {Empirical} {Methods} in {Natural} {Language} {Processing} ({EMNLP})}, pages 1532--1543, Doha, Qatar. Association for Computational Linguistics.
\newblock https://doi.org/10.3115/v1/D14-1162.

\bibitem[Perez and Granger, 2007]{perez_ipython_2007}
Perez, F. and Granger, B.~E. (2007).
\newblock {IPython}: {A} {System} for {Interactive} {Scientific} {Computing}.
\newblock {\em Computing in Science \& Engineering}, 9(3):21--29.
\newblock https://doi.org/10.1109/MCSE.2007.53.

\bibitem[Phillips, 1950]{phillips_cultural_1950}
Phillips, A. (1950).
\newblock The {Cultural} {Cringe}.
\newblock {\em Meanjin}, 9(4):299--302.
\newblock Retrieved from https://search.informit.org/doi/abs/10.3316/INFORMIT.692761906212773.

\bibitem[Phillipson, 2014]{phillipson_americanization_2014}
Phillipson, R. (2014).
\newblock Americanization and {Englishization} {As} {Processes} of {Global} {Occupation}.
\newblock In Orelus, P.~W., editor, {\em Affirming {Language} {Diversity} in {Schools} and {Society}}, pages 188--214. Routledge, New York, NY.

\bibitem[Pickles, 2002]{pickles_kiwi_2002}
Pickles, K. (2002).
\newblock Kiwi {Icons} and the {Re}-{Settlement} of {New} {Zealand} as {Colonial} {Space}.
\newblock {\em New Zealand Geographer}, 58(2):5--16.
\newblock https://doi.org/10.1111/j.1745-7939.2002.tb01631.x.

\bibitem[Pickles, 2011]{pickles_transnational_2011}
Pickles, K. (2011).
\newblock Transnational {History} and {Cultural} {Cringe}: {Some} {Issues} for {Consideration} in {New} {Zealand}, {Australia} and {Canada}.
\newblock {\em History Compass}, 9(9):657--673.
\newblock https://doi.org/10.1111/j.1478-0542.2011.00794.x.

\bibitem[Pollock, 2013]{pollock_tea_2013}
Pollock, K. (2013).
\newblock Tea, coffee and soft drinks – {Tea} and coffee.
\newblock Retrieved from https://teara.govt.nz/en/tea-coffee-and-soft-drinks.

\bibitem[Preston, 1986]{preston_five_1986}
Preston, D.~R. (1986).
\newblock Five {Visions} of {America}.
\newblock {\em Language in Society}, 15(2):221--240.
\newblock Retrieved from https://www.jstor.org/stable/4167748.

\bibitem[Preston, 2010]{preston_language_2010}
Preston, D.~R. (2010).
\newblock Language, people, salience, space: perceptual dialectology and language regard.
\newblock {\em Dialectologia: revista electrònica}, 5:87--131.
\newblock Retrieved from https://raco.cat/index.php/Dialectologia/article/view/198838.

\bibitem[Price, 2016]{price_where_2016}
Price, R. (2016).
\newblock `{Where} are they now?': {Nek} {Minnit} guy.
\newblock {\em Stuff}.
\newblock Retrieved from https://www.stuff.co.nz/life-style/life/76472397/where-are-they-now-nek-minnit-guy.

\bibitem[Proferes et~al., 2021]{proferes_studying_2021}
Proferes, N., Jones, N., Gilbert, S., Fiesler, C., and Zimmer, M. (2021).
\newblock Studying reddit: {A} systematic overview of disciplines, approaches, methods, and ethics.
\newblock {\em Social Media+ Society}, 7(2):20563051211019004.
\newblock https://doi.org/10.1177/20563051211019004.

\bibitem[Punske and Butler, 2019]{punske_me_2019}
Punske, J. and Butler, E. (2019).
\newblock Do me a syntax: {Doggo} memes, language games and the internal structure of {English}.
\newblock {\em Ampersand}, 6:100052.
\newblock https://doi.org/10.1016/j.amper.2019.100052.

\bibitem[Purba and Nainggolan, 2025]{purba_rise_2025}
Purba, C.~N. and Nainggolan, A.~I. (2025).
\newblock Rise of '{Rizz}': {A} {Morphological} {Study} of {Its} {Formation} and {Usage}.
\newblock {\em Innovative: Journal Of Social Science Research}, 5(2):1598--1607.

\bibitem[{Python Software Foundation}, 2025]{python_software_foundation_python_2025}
{Python Software Foundation} (2025).
\newblock Python 3.
\newblock Retrieved from https://www.python.org/.

\bibitem[Quinn, 1995]{quinn_variation_1995}
Quinn, H. (1995).
\newblock Variation in {NZE} syntax and morphology: a study of the acceptance and use of grammatical variants among {Canterbury} and {West} {Coast} teenagers.
\newblock Retrieved from https://doi.org/10.26021/14978.

\bibitem[Quinn, 2000]{quinn_variation_2000}
Quinn, H. (2000).
\newblock Variation in {New} {Zealand} {English} syntax and morphology.
\newblock In Bell, A. and Kuiper, K., editors, {\em New {Zealand} {English}}, Varieties of {English} {Around} the {World}, pages 173--197. John Benjamins Publishing Company, Amsterdam, The Netherlands; Philadelphia, PA.
\newblock https://doi.org/10.1075/veaw.g25.11qui.

\bibitem[Quirk et~al., 1985]{quirk_comprehensive_1985}
Quirk, R., Greenbaum, S., Leech, G., and Svartvik, J. (1985).
\newblock {\em A {Comprehensive} {Grammar} of the {English} {Language}}.
\newblock Longman, London, England; New York, NY.

\bibitem[Radford et~al., 2019]{radford_language_2019}
Radford, A., Wu, J., Child, R., Luan, D., Amodei, D., Sutskever, I., Dean, J., and Ghemawat, S. (2019).
\newblock Language {Models} are {Unsupervised} {Multitask} {Learners}.
\newblock Technical report, OpenAI, San Francisco, CA.
\newblock Retrieved from https://cdn.openai.com/better-language-models/language\_models\_are\_unsupervised\_multitask\_learners.pdf.

\bibitem[Reed, 2016]{reed_sounding_2016}
Reed, P.~E. (2016).
\newblock {\em Sounding {Appalachian}: /{aI}/ {Monophthongization}, {Rising} {Pitch} {Accents}, and {Rootedness}}.
\newblock Doctoral dissertation, Available from ProQuest Dissertations \& Theses Global database.
\newblock (UMI No. 10127065).

\bibitem[Reed, 2020a]{reed_importance_2020}
Reed, P.~E. (2020a).
\newblock The {Importance} of {Rootedness} in the {Study} of {Appalachian} {English}: {Case} {Study} {Evidence} for a {Proposed} {Rootedness} {Metric}.
\newblock {\em American Speech}, 95(2):203--226.
\newblock https://doi.org/10.1215/00031283-7706532.

\bibitem[Reed, 2020b]{reed_place_2020}
Reed, P.~E. (2020b).
\newblock Place and language: {Links} between speech, region, and connection to place.
\newblock {\em WIREs Cognitive Science}, 11(3):e1524.
\newblock https://doi.org/10.1002/wcs.1524.

\bibitem[Relph, 1976]{relph_place_1976}
Relph, E.~C. (1976).
\newblock {\em Place and placelessness}.
\newblock Research in planning and design. 1. Pion, London, England.

\bibitem[Rensink, 2008]{rensink_informant_2008}
Rensink, W.~G. (2008).
\newblock Informant {Classification} of {Dialects}.
\newblock In Preston, D.~R., editor, {\em Handbook of {Perceptual} {Dialectology}}, volume~1, pages 3--7. John Benjamins Publishing Company, Amsterdam, The Netherlands; Philadelphia, PA.
\newblock https://doi.org/10.1075/z.hpd1.07ren.

\bibitem[Roberts, 2016]{roberts_assessing_2016}
Roberts, K. (2016).
\newblock Assessing the {Corpus} {Size} vs. {Similarity} {Trade}-off for {Word} {Embeddings} in {Clinical} {NLP}.
\newblock In Rumshisky, A., Roberts, K., Bethard, S., and Naumann, T., editors, {\em Proceedings of the {Clinical} {Natural} {Language} {Processing} {Workshop}}, pages 54--63, Osaka, Japan. The COLING 2016 Organizing Committee.
\newblock https://aclanthology.org/W16-4208/.

\bibitem[Rocha-Silva et~al., 2024]{rocha-silva_passive_2024}
Rocha-Silva, T., Nogueira, C., and Rodrigues, L. (2024).
\newblock Passive data collection on {Reddit}: a practical approach.
\newblock {\em Research Ethics}, 20(3):453--470.
\newblock https://doi.org/10.1177/17470161231210542.

\bibitem[Rogers et~al., 2017]{rogers_too_2017}
Rogers, A., Drozd, A., and Li, B. (2017).
\newblock The (too {Many}) {Problems} of {Analogical} {Reasoning} with {Word} {Vectors}.
\newblock In Ide, N., Herbelot, A., and Màrquez, L., editors, {\em Proceedings of the 6th {Joint} {Conference} on {Lexical} and {Computational} {Semantics}}, pages 135--148, Vancouver, Canada. Association for Computational Linguistics.
\newblock https://doi.org/10.18653/v1/S17-1017.

\bibitem[Ruette et~al., 2014]{ruette_semantic_2014}
Ruette, T., Geeraerts, D., Peirsman, Y., and Speelman, D. (2014).
\newblock Semantic weighting mechanisms in scalable lexical sociolectometry.
\newblock In Szmrecsanyi, B. and Wälchli, B., editors, {\em Aggregating {Dialectology}, {Typology}, and {Register} {Analysis}: {Linguistic} {Variation} in {Text} and {Speech}}, number~28 in linguae \& littera, pages 205--230. Walter de Gruyter, Berlin, Germany; Boston, MA.
\newblock https://doi.org/10.1515/9783110317558.205.

\bibitem[Ryan, 1983]{ryan_revised_1983}
Ryan, J.~C. (1983).
\newblock {\em The {Revised} {Dictionary} of {Māori}}.
\newblock Heinemann, Auckland, New Zealand; Tadworth, England; Portsmouth, NH.

\bibitem[Rymes and Leone-Pizzighella, 2018]{rymes_youtube-based_2018}
Rymes, B. and Leone-Pizzighella, A. (2018).
\newblock {YouTube}-based accent challenge narratives: {Web} 2.0 as a context for studying the social value of accent.
\newblock {\em International Journal of the Sociology of Language}, 2018(250):137--163.
\newblock https://doi.org/10.1515/ijsl-2017-0058.

\bibitem[Sadiq et~al., 2023]{sadiq_deepfake_2023}
Sadiq, S., Aljrees, T., and Ullah, S. (2023).
\newblock Deepfake {Detection} on {Social} {Media}: {Leveraging} {Deep} {Learning} and {FastText} {Embeddings} for {Identifying} {Machine}-{Generated} {Tweets}.
\newblock {\em IEEE Access}, 11:95008--95021.
\newblock https://doi.org/10.1109/ACCESS.2023.3308515.

\bibitem[Sawyer, 2005]{sawyer_social_2005}
Sawyer, R.~K. (2005).
\newblock {\em Social {Emergence}: {Societies} {As} {Complex} {Systems}}.
\newblock Cambridge University Press, Cambridge, England.
\newblock https://doi.org/10.1017/CBO9780511734892.

\bibitem[Scheider and Janowicz, 2010]{scheider_places_2010}
Scheider, S. and Janowicz, K. (2010).
\newblock Places as media of containment.
\newblock In {\em Proceedings of the 6th {International} {Conference} on {Geographic} {Information} {Science}}, pages 1--5, Zürich, Switzerland.
\newblock Retrieved from http://geographicknowledge.de/pdf/PlaceAsMedia.pdf.

\bibitem[Schleicher, 1853]{schleicher_ersten_1853}
Schleicher, A. (1853).
\newblock Die ersten {Spaltungen} des indogermanischen {Urvolkes} [{The} first splits of the {Indo}-{European} people].
\newblock {\em Allgemeine Monatsschrift für Wissenschaft und Literatur [General Monthly Journal for Science and Literature]}, 3:786--787.

\bibitem[Schmidt, 1872]{schmidt_verwandtschaftsverhaltnisse_1872}
Schmidt, J. (1872).
\newblock {\em Die {Verwandtschaftsverhältnisse} der indogermanischen {Sprachen} [{The} {Relationships} of the {Indo}-{Germanic} {Languages}]}.
\newblock Hermann Böhlau, Weimar, Germany.

\bibitem[Schneider, 2005]{schneider_no_2005}
Schneider, K.~P. (2005).
\newblock No problem, you're welcome, anytime: {Responding} to thanks in {Ireland}, {England}, and the {US}.
\newblock In Barron, A. and Schneider, K.~P., editors, {\em The {Pragmatics} of {Irish} {English}}, volume 164 of {\em Trends in {Linguistics}. {Studies} and {Monographs} [{TiLSM}] {Series}}, pages 101--139. De Gruyter Mouton, Berlin, Germany; New York, NY, 1 edition.
\newblock https://doi.org/10.1515/9783110898934.101.

\bibitem[Schoux~Casey, 2013]{schoux_casey_postvocalic_2013}
Schoux~Casey, C. (2013).
\newblock {\em Postvocalic /r/ in {New} {Orleans}: {Language}, place and commodification}.
\newblock Doctoral dissertation, Available from ProQuest Dissertations \& Theses Global database.
\newblock (UMI No. 3577179).

\bibitem[Schwartz and Halegoua, 2015]{schwartz_spatial_2015}
Schwartz, R. and Halegoua, G.~R. (2015).
\newblock The spatial self: {Location}-based identity performance on social media.
\newblock {\em New Media \& Society}, 17(10):1643--1660.
\newblock https://doi.org/10.1177/1461444814531364.

\bibitem[Seabold and Perktold, 2010]{seabold_statsmodels_2010}
Seabold, S. and Perktold, J. (2010).
\newblock Statsmodels: {Econometric} and {Statistical} {Modeling} with {Python}.
\newblock In {\em Proceedings of the 9th {Python} in {Science} {Conference}}, pages 92--96, Austin, TX. SciPy Proceedings.
\newblock https://doi.org/10.25080/Majora-92bf1922-011.

\bibitem[Seamon, 1979]{seamon_phenomenology_1979}
Seamon, D. (1979).
\newblock Phenomenology, geography and geographical education.
\newblock {\em Journal of Geography in Higher Education}, 3(2):40--50.
\newblock https://doi.org/10.1080/03098267908708726.

\bibitem[Seaver, 2015]{seaver_nice_2015}
Seaver, N. (2015).
\newblock The nice thing about context is that everyone has it.
\newblock {\em Media, Culture \& Society}, 37(7):1101--1109.
\newblock https://doi.org/10.1177/0163443715594102.

\bibitem[Sebba, 2008]{sebba_spelling_2008}
Sebba, M. (2008).
\newblock Spelling rebellion.
\newblock In Androutsopoulos, J. and Georgakopoulou, A., editors, {\em Discourse {Constructions} of {Youth} {Identities}}, pages 151--172. John Benjamins Publishing Company, Amsterdam, The Netherlands; Philadelphia, PA.
\newblock https://doi.org/10.1075/pbns.110.09seb.

\bibitem[Seidman, 2017]{seidman_contested_2017}
Seidman, S. (2017).
\newblock {\em Contested {Knowledge}: {Social} {Theory} {Today}}.
\newblock John Wiley \& Sons, Chichester, England.

\bibitem[Shannon, 1948]{shannon_mathematical_1948}
Shannon, C.~E. (1948).
\newblock A mathematical theory of communication.
\newblock {\em The Bell System Technical Journal}, 27(3):379--423.
\newblock https://doi.org/10.1002/j.1538-7305.1948.tb01338.x.

\bibitem[Sigley, 1997]{sigley_choosing_1997}
Sigley, R.~J. (1997).
\newblock Choosing {Your} {Relatives}: {Relative} {Clauses} in {New} {Zealand} {English}.
\newblock Retrieved from https://ir.wgtn.ac.nz/handle/123456789/23581.

\bibitem[Silverstein, 2003]{silverstein_indexical_2003}
Silverstein, M. (2003).
\newblock Indexical order and the dialectics of sociolinguistic life.
\newblock {\em Language \& Communication}, 23(3):193--229.
\newblock https://doi.org/10.1016/S0271-5309(03)00013-2.

\bibitem[Simpson, 2008]{simpson_hypocoristics_2008}
Simpson, J. (2008).
\newblock Hypocoristics in {Australian} {English}.
\newblock In Kortmann, B. and Burridge, K., editors, {\em The {Pacific} and {Australasia}}, number~3 in Varieties of {English}, pages 398--414. De Gruyter Mouton, Berlin, Germany; New York, NY.
\newblock https://doi.org/10.1515/9783110208412.2.398.

\bibitem[Singer et~al., 2014]{singer_evolution_2014}
Singer, P., Flöck, F., Meinhart, C., Zeitfogel, E., and Strohmaier, M. (2014).
\newblock Evolution of reddit: from the front page of the internet to a self-referential community?
\newblock In {\em Proceedings of the 23rd {International} {Conference} on {World} {Wide} {Web}}, pages 517--522. Association for Computing Machinery.
\newblock https://doi.org/10.1145/2567948.2576943.

\bibitem[Sloan and Morgan, 2015]{sloan_who_2015}
Sloan, L. and Morgan, J. (2015).
\newblock Who {Tweets} with {Their} {Location}? {Understanding} the {Relationship} between {Demographic} {Characteristics} and the {Use} of {Geoservices} and {Geotagging} on {Twitter}.
\newblock {\em PLoS ONE}, 10(11):e0142209.
\newblock https://doi.org/10.1371/journal.pone.0142209.

\bibitem[Sloan et~al., 2013]{sloan_knowing_2013}
Sloan, L., Morgan, J., Housley, W., Williams, M., Edwards, A., Burnap, P., and Rana, O. (2013).
\newblock Knowing the {Tweeters}: {Deriving} {Sociologically} {Relevant} {Demographics} from {Twitter}.
\newblock {\em Sociological Research Online}, 18(3):74--84.
\newblock https://doi.org/10.5153/sro.3001.

\bibitem[Solomon, 1999]{solomon_phonological_1999}
Solomon, J. (1999).
\newblock {\em Phonological and {Syntactic} {Variation} in the {Spanish} of {Valladolid}, {Yucatán}}.
\newblock Doctoral dissertation, Available from ProQuest Dissertations \& Theses Global database.
\newblock (UMI No. 9958201).

\bibitem[Sorokin, 1943]{sorokin_sociocultural_1943}
Sorokin, P.~A. (1943).
\newblock {\em Sociocultural {Causality}, {Space}, {Time}: {A} {Study} of {Referential} {Principles} of {Sociology} and {Social} {Science}}.
\newblock Sociological {Series}. Duke University Press, Durham, NC.

\bibitem[Sowa, 2009]{sowa_sweet_2009}
Sowa, J. (2009).
\newblock ‘{Sweet} as!’: the intensifier as in {New} {Zealand} and {Australian} {English}.
\newblock {\em English Today}, 25(2):58--61.
\newblock https://doi.org/10.1017/S0266078409000212.

\bibitem[Spärck~Jones, 1972]{sparck_jones_statistical_1972}
Spärck~Jones, K. (1972).
\newblock A statistical interpretation of term specificity and its application in retrieval.
\newblock {\em Journal of Documentation}, 28(1):11--21.
\newblock https://doi.org/10.1108/eb026526.

\bibitem[{Stats NZ}, 2024]{stats_nz_place_2024}
{Stats NZ} (2024).
\newblock Place and ethnic group summaries.
\newblock Retrieved from https://tools.summaries.stats.govt.nz/.

\bibitem[Steed et~al., 2022]{steed_upstream_2022}
Steed, R., Panda, S., Kobren, A., and Wick, M. (2022).
\newblock Upstream {Mitigation} {Is} \textit{ {N}}ot {All} {You} {Need}: {Testing} the {Bias} {Transfer} {Hypothesis} in {Pre}-{Trained} {Language} {Models}.
\newblock In Muresan, S., Nakov, P., and Villavicencio, A., editors, {\em Proceedings of the 60th {Annual} {Meeting} of the {Association} for {Computational} {Linguistics} ({Volume} 1: {Long} {Papers})}, pages 3524--3542, Dublin, Ireland. Association for Computational Linguistics.
\newblock https://doi.org/10.18653/v1/2022.acl-long.247.

\bibitem[Steels, 2000]{steels_language_2000}
Steels, L. (2000).
\newblock Language as a {Complex} {Adaptive} {System}.
\newblock In Schoenauer, M., Deb, K., Rudolph, G., Yao, X., Lutton, E., Merelo, J.~J., and Schwefel, H.-P., editors, {\em Parallel {Problem} {Solving} from {Nature} {PPSN} {VI}}, volume 1917 of {\em Lecture {Notes} in {Computer} {Science}}, pages 17--26, Berlin, Germany. Springer.
\newblock https://doi.org/10.1007/3-540-45356-3\_2.

\bibitem[Steen et~al., 2023]{steen_you_2023}
Steen, E., Yurechko, K., and Klug, D. (2023).
\newblock You {Can} ({Not}) {Say} {What} {You} {Want}: {Using} {Algospeak} to {Contest} and {Evade} {Algorithmic} {Content} {Moderation} on {TikTok}.
\newblock {\em Social Media + Society}, 9(3):1--17.
\newblock https://doi.org/10.1177/20563051231194586.

\bibitem[Stevenson, 2010a]{stevenson_kransky_2010}
Stevenson, A. (2010a).
\newblock kransky.
\newblock In {\em Oxford {Dictionary} of {English}}. Oxford University Press, 3 edition.
\newblock https://doi.org/10.1093/acref/9780199571123.001.0001.

\bibitem[Stevenson, 2010b]{stevenson_rark_2010}
Stevenson, A. (2010b).
\newblock rark.
\newblock In {\em Oxford {Dictionary} of {English}}. Oxford University Press, 3 edition.
\newblock https://doi.org/10.1093/acref/9780199571123.001.0001.

\bibitem[Stevenson, 2010c]{stevenson_saveloy_2010}
Stevenson, A. (2010c).
\newblock saveloy.
\newblock In {\em Oxford {Dictionary} of {English}}. Oxford University Press, 3 edition.
\newblock https://doi.org/10.1093/acref/9780199571123.001.0001.

\bibitem[Stigler, 1980]{stigler_stiglers_1980}
Stigler, S.~M. (1980).
\newblock Stigler's {Law} of {Eponymy}.
\newblock {\em Transactions of the New York Academy of Sciences}, 39(1 Series II):147--157.
\newblock https://doi.org/10.1111/j.2164-0947.1980.tb02775.x.

\bibitem[Stine and Agarwal, 2020]{stine_comparative_2020}
Stine, Z.~K. and Agarwal, N. (2020).
\newblock Comparative {Discourse} {Analysis} {Using} {Topic} {Models}: {Contrasting} {Perspectives} on {China} from {Reddit}.
\newblock In {\em International {Conference} on {Social} {Media} and {Society}}, pages 73--84, New York, NY, USA. Association for Computing Machinery.
\newblock https://doi.org/10.1145/3400806.3400816.

\bibitem[Stringham and Izbicki, 2020]{stringham_evaluating_2020}
Stringham, N. and Izbicki, M. (2020).
\newblock Evaluating {Word} {Embeddings} on {Low}-{Resource} {Languages}.
\newblock In Eger, S., Gao, Y., Peyrard, M., Zhao, W., and Hovy, E., editors, {\em Proceedings of the {First} {Workshop} on {Evaluation} and {Comparison} of {NLP} {Systems}}, pages 176--186, Online. Association for Computational Linguistics.
\newblock https://doi.org/10.18653/v1/2020.eval4nlp-1.17.

\bibitem[Stubbe and Holmes, 2000]{stubbe_talking_2000}
Stubbe, M. and Holmes, J. (2000).
\newblock Talking {Maori} or {Pakeha} in {English}: signalling identity in discourse.
\newblock In Bell, A. and Kuiper, K., editors, {\em New {Zealand} {English}}, pages 249--278. John Benjamins Publishing Company, Amsterdam, The Netherlands; Philadelphia, PA.
\newblock https://doi.org/10.1075/veaw.g25.14stu.

\bibitem[Stæhr et~al., 2019]{staehr_dialect_2019}
Stæhr, A.~C., Monka, M., Quist, P., and Larsen, A. (2019).
\newblock Dialect in the {Media}: {Mediatization} and {Processes} of {Standardization}.
\newblock In {\em Standardization as {Sociolinguistic} {Change}}, pages 169--189. Routledge.

\bibitem[Szmrecsanyi, 2011]{szmrecsanyi_corpus-based_2011}
Szmrecsanyi, B. (2011).
\newblock Corpus-based dialectometry: a methodological sketch.
\newblock {\em Corpora}, 6(1):45--76.
\newblock https://doi.org/10.3366/cor.2011.0004.

\bibitem[Séguy, 1971]{seguy_relation_1971}
Séguy, J. (1971).
\newblock La relation entre la distance spatiale et la distance lexicale [{The} relationship between spatial distance and lexical distance].
\newblock {\em Revue de linguistique romane}, 35(139-140):335--357.
\newblock https://doi.org/10.5169/seals-399508.

\bibitem[Søgaard et~al., 2014]{sogaard_selection_2014}
Søgaard, A., Plank, B., and Hovy, D. (2014).
\newblock Selection bias, label bias, and bias in ground truth.
\newblock In {\em Proceedings of {COLING} 2014, the 25th {International} {Conference} on {Computational} {Linguistics}: {Tutorial} {Abstracts}}, pages 11--13, Dublin, Ireland. Association for Computational Linguistics.

\bibitem[Taha et~al., 2024]{taha_comprehensive_2024}
Taha, K., Yoo, P.~D., Yeun, C., Homouz, D., and Taha, A. (2024).
\newblock A comprehensive survey of text classification techniques and their research applications: {Observational} and experimental insights.
\newblock {\em Computer Science Review}, 54:100664.
\newblock https://doi.org/10.1016/j.cosrev.2024.100664.

\bibitem[Tarnarutckaia and Ensslin, 2020]{tarnarutckaia_myth_2020}
Tarnarutckaia, E. and Ensslin, A. (2020).
\newblock The myth of the “clarté française”: {Language} ideologies and metalinguistic discourse of videogame speech accents on {Reddit}.
\newblock {\em Discourse, Context \& Media}, 33:100352.
\newblock https://doi.org/10.1215/00031283-2322637.

\bibitem[{The pandas development team}, 2025]{the_pandas_development_team_pandas-devpandas_2025}
{The pandas development team} (2025).
\newblock pandas-dev/pandas: {Pandas}.
\newblock https://doi.org/10.5281/zenodo.3509134.

\bibitem[{The “Five Graces Group”} et~al., 2009]{the_five_graces_group_language_2009}
{The “Five Graces Group”}, Beckner, C., Blythe, R., Bybee, J., Christiansen, M.~H., Croft, W., Ellis, N.~C., Holland, J., Ke, J., Larsen-Freeman, D., and Schoenemann, T. (2009).
\newblock Language {Is} a {Complex} {Adaptive} {System}: {Position} {Paper}.
\newblock {\em Language Learning}, 59(s1):1--26.
\newblock https://doi.org/10.1111/j.1467-9922.2009.00533.x.

\bibitem[Trudgill, 1971]{trudgill_social_1971}
Trudgill, P. (1971).
\newblock The {Social} {Differentiation} of {English} in {Norwich}.
\newblock Retrieved from http://hdl.handle.net/1842/16333.

\bibitem[Trudgill, 1983]{trudgill_dialect_1983}
Trudgill, P. (1983).
\newblock {\em On {Dialect}: {Social} and {Geographical} {Perspectives}}.
\newblock New York University Press, New York, NY; London, England.

\bibitem[Trudgill and Hannah, 2017]{trudgill_international_2017}
Trudgill, P. and Hannah, J. (2017).
\newblock {\em International {English}: {A} {Guide} to {Varieties} of {English} {Around} the {World}}.
\newblock Routledge, London, England, 6 edition.
\newblock https://doi.org/10.4324/9781315192932.

\bibitem[Trye et~al., 2019]{trye_maori_2019}
Trye, D., Calude, A., Bravo-Marquez, F., and Keegan, T.~T. (2019).
\newblock Māori {Loanwords}: {A} {Corpus} of {New} {Zealand} {English} {Tweets}.
\newblock In Alva-Manchego, F., Choi, E., and Khashabi, D., editors, {\em Proceedings of the 57th {Annual} {Meeting} of the {Association} for {Computational} {Linguistics}: {Student} {Research} {Workshop}}, pages 136--142. Association for Computational Linguistics.
\newblock https://doi.org/10.18653/v1/P19-2018.

\bibitem[Trye et~al., 2020]{trye_hybrid_2020}
Trye, D., Calude, A.~S., Bravo-Marquez, F., and Keegan, T.~T. (2020).
\newblock Hybrid {Hashtags}: \#{YouKnowYoureAKiwiWhen} {Your} {Tweet} {Contains} {Māori} and {English}.
\newblock {\em Frontiers in Artificial Intelligence}, 3.
\newblock https://doi.org/10.3389/frai.2020.00015.

\bibitem[Tuan, 1979]{tuan_space_1979}
Tuan, Y.-F. (1979).
\newblock Space and {Place}: {Humanistic} {Perspective}.
\newblock In Gale, S. and Olsson, G., editors, {\em Philosophy in {Geography}}, pages 387--427. Springer Netherlands, Dordrecht, The Netherlands.
\newblock https://doi.org/10.1007/978-94-009-9394-5\_19.

\bibitem[Turner, 1966]{turner_english_1966}
Turner, G.~W. (1966).
\newblock {\em The {English} {Language} in {Australia} and {New} {Zealand}}.
\newblock English {Language} {Series}. Longmans, Auckland, New Zealand.

\bibitem[van~der Nagel, 2018]{van_der_nagel_networks_2018}
van~der Nagel, E. (2018).
\newblock ‘{Networks} that work too well’: intervening in algorithmic connections.
\newblock {\em Media International Australia}, 168(1):81--92.
\newblock https://doi.org/10.1177/1329878X18783002.

\bibitem[van Rossum and de~Boer, 1991]{van_rossum_interactively_1991}
van Rossum, G. and de~Boer, J. (1991).
\newblock Interactively testing remote servers using the {Python} programming language.
\newblock {\em CWI Quarterly}, 4(4):283--304.
\newblock Retrieved from https://ir.cwi.nl/pub/18204.

\bibitem[Vaswani et~al., 2017]{vaswani_attention_2017}
Vaswani, A., Shazeer, N., Parmar, N., Uszkoreit, J., Jones, L., Gomez, A.~N., Kaiser, L., and Polosukhin, I. (2017).
\newblock Attention is all you need.
\newblock In {\em Proceedings of the 31st {International} {Conference} on {Neural} {Information} {Processing} {Systems}}, pages 6000--6010, Red Hook, NY. Curran Associates.
\newblock https://dl.acm.org/doi/10.5555/3295222.3295349.

\bibitem[Wang et~al., 2019]{wang_evaluating_2019}
Wang, B., Wang, A., Chen, F., Wang, Y., and Kuo, C.-C.~J. (2019).
\newblock Evaluating word embedding models: methods and experimental results.
\newblock {\em APSIPA Transactions on Signal and Information Processing}, 8:e19.

\bibitem[Waskom, 2021]{waskom_seaborn_2021}
Waskom, M.~L. (2021).
\newblock seaborn: statistical data visualization.
\newblock {\em Journal of Open Source Software}, 6(60):3021.
\newblock https://doi.org/10.21105/joss.03021.

\bibitem[Weinreich et~al., 1968]{weinreich_empirical_1968}
Weinreich, U., Labov, W., and Herzog, M. (1968).
\newblock Empirical {Foundations} for a {Theory} of {Language} {Change}.
\newblock In Lehmann, W.~P. and Malkiel, Y., editors, {\em Directions for {Historical} {Linguistics}}, pages 95--195. University of Texas Press, Austin, TX.

\bibitem[{Weischedel, Ralph} et~al., 2013]{weischedel_ralph_ontonotes_2013}
{Weischedel, Ralph}, {Palmer, Martha}, {Marcus, Mitchell}, {Hovy, Eduard}, {Pradhan, Sameer}, {Ramshaw, Lance}, {Xue, Nianwen}, {Taylor, Ann}, {Kaufman, Jeff}, {Franchini, Michelle}, {El-Bachouti, Mohammed}, {Belvin, Robert}, and {Houston, Ann} (2013).
\newblock {OntoNotes} {Release} 5.0.
\newblock https://doi.org/10.35111/xmhb-2b84.

\bibitem[Wick, 1997]{wick_speech_1997}
Wick, N.~B. (1997).
\newblock {\em Speech community in the virtual world: the case of one listserv}.
\newblock Doctoral dissertation, Available from ProQuest Dissertations \& Theses Global database.
\newblock (UMI No. 9807043).

\bibitem[Williams and Dagli, 2017]{williams_twitter_2017}
Williams, J. and Dagli, C. (2017).
\newblock Twitter {Language} {Identification} {Of} {Similar} {Languages} {And} {Dialects} {Without} {Ground} {Truth}.
\newblock In Nakov, P., Zampieri, M., Ljubešić, N., Tiedemann, J., Malmasi, S., and Ali, A., editors, {\em Proceedings of the {Fourth} {Workshop} on {NLP} for {Similar} {Languages}, {Varieties} and {Dialects} ({VarDial})}, pages 73--83, Valencia, Spain. Association for Computational Linguistics.

\bibitem[Williams et~al., 2017]{williams_towards_2017}
Williams, M.~L., Burnap, P., and Sloan, L. (2017).
\newblock Towards an {Ethical} {Framework} for {Publishing} {Twitter} {Data} in {Social} {Research}: {Taking} into {Account} {Users}’ {Views}, {Online} {Context} and {Algorithmic} {Estimation}.
\newblock {\em Sociology}, 51(6):1149--1168.
\newblock https://doi.org/10.1177/0038038517708140.

\bibitem[Winford, 2010]{winford_contact_2010}
Winford, D. (2010).
\newblock Contact and {Borrowing}.
\newblock In Hickey, R., editor, {\em The {Handbook} of {Language} {Contact}}, Blackwell {Handbooks} in {Linguistics}, pages 170--187. Wiley-Blackwell, Chichester, England.
\newblock https://doi.org/10.1002/9781444318159.ch8.

\bibitem[Withers, 2009]{withers_place_2009}
Withers, C. W.~J. (2009).
\newblock Place and the "{Spatial} {Turn}" in {Geography} and in {History}.
\newblock {\em Journal of the History of Ideas}, 70(4):637--658.
\newblock Retrieved from https://www.jstor.org/stable/20621915.

\bibitem[Wolfram and Schilling-Estes, 2016]{wolfram_american_2016}
Wolfram, W. and Schilling-Estes, N. (2016).
\newblock {\em American {English}: {Dialects} and {Variation}}.
\newblock Blackwell Publishing, Chichester, England, 3 edition.

\bibitem[Würschinger, 2021]{wurschinger_social_2021}
Würschinger, Q. (2021).
\newblock Social {Networks} of {Lexical} {Innovation}. {Investigating} the {Social} {Dynamics} of {Diffusion} of {Neologisms} on {Twitter}.
\newblock {\em Frontiers in Artificial Intelligence}, 4.
\newblock https://doi.org/10.3389/frai.2021.648583.

\bibitem[Yan and Song, 2019]{yan_lexical_2019}
Yan, H.-c. and Song, Y.-q. (2019).
\newblock On the {Lexical} {Features} of {New} {Zealand} {English}.
\newblock {\em Journal of Literature and Art Studies}, 9(4):392--397.
\newblock https://doi.org/10.17265/2159-5836/2019.04.005.

\bibitem[Zampieri et~al., 2020]{zampieri_natural_2020}
Zampieri, M., Nakov, P., and Scherrer, Y. (2020).
\newblock Natural language processing for similar languages, varieties, and dialects: {A} survey.
\newblock {\em Natural Language Engineering}, 26(6):595--612.

\bibitem[Zhuang et~al., 2021]{zhuang_robustly_2021}
Zhuang, L., Wayne, L., Ya, S., and Jun, Z. (2021).
\newblock A {Robustly} {Optimized} {BERT} {Pre}-training {Approach} with {Post}-training.
\newblock In Li, S., Sun, M., Liu, Y., Wu, H., Liu, K., Che, W., He, S., and Rao, G., editors, {\em Proceedings of the 20th {Chinese} {National} {Conference} on {Computational} {Linguistics}}, pages 1218--1227, Huhhot, China. Chinese Information Processing Society of China.
\newblock Retrieved from https://aclanthology.org/2021.ccl-1.108/.

\bibitem[Zimmerbauer, 2011]{zimmerbauer_image_2011}
Zimmerbauer, K. (2011).
\newblock From {Image} to {Identity}: {Building} {Regions} by {Place} {Promotion}.
\newblock {\em European Planning Studies}, 19(2):243--260.
\newblock https://doi.org/10.1080/09654313.2011.532667.

\bibitem[Zorić, 2024]{zoric_constructing_2024}
Zorić, L. (2024).
\newblock Constructing language identity on {Reddit}.
\newblock Retrieved from https://urn.nsk.hr/urn:nbn:hr:131:831276.

\end{thebibliography}
